\documentclass{article} \usepackage{iclr2022_conference,times}

\usepackage{xcolor}         \usepackage[utf8]{inputenc} \usepackage[T1]{fontenc}    \usepackage{hyperref}       \usepackage{url}            \usepackage{booktabs}       \usepackage{amsfonts}       \RequirePackage{amsmath}
\RequirePackage{amssymb}
\RequirePackage{mathtools}
\ifx\proof\undefined 
\RequirePackage{amsthm}
\fi
\RequirePackage{bm} 
\RequirePackage{url}

\newcommand{\Dcal}{\mathcal{D}}
\newcommand{\Ecal}{\mathcal{E}}

\newcommand{\Gcal}{\mathcal{G}}

\newcommand{\Ical}{\mathcal{I}}

\newcommand{\Pcal}{\mathcal{P}}

\newcommand{\Tcal}{{\mathcal{T}}}

        \newcommand{\NN}{\mathbb{N}}   \newcommand{\RR}{\mathbb{R}}

\newcommand*{\argmin}{\mathop{\mathrm{argmin}}}
\newcommand*{\argmax}{\mathop{\mathrm{argmax}}}

\ifx\BlackBox\undefined
\newcommand{\BlackBox}{\rule{1.5ex}{1.5ex}}  \fi

\ifx\QED\undefined
\def\QED{~\rule[-1pt]{5pt}{5pt}\par\medskip}
\fi

\ifx\proof\undefined
\newenvironment{proof}{\par\noindent{\bf Proof\ }}{\hfill\BlackBox\\[2mm]}
\fi
\ifx\proofsketch\undefined
\newenvironment{proofsketch}{\par\noindent{\bf Proof Sketch\ }}{\hfill\BlackBox\\[2mm]}
\fi

\ifx\theorem\undefined
\newtheorem{theorem}{Theorem}
\fi
\ifx\example\undefined
\newtheorem{example}{Example}
\fi
\ifx\property\undefined
\newtheorem{property}{Property}
\fi
\ifx\lemma\undefined
\newtheorem{lemma}[theorem]{Lemma}
\fi
\ifx\proposition\undefined
\newtheorem{proposition}[theorem]{Proposition}
\fi
\ifx\remark\undefined
\newtheorem{remark}[theorem]{Remark}
\fi
\ifx\corollary\undefined
\newtheorem{corollary}{Corollary}
\fi
\ifx\definition\undefined
\newtheorem{definition}{Definition}
\fi
\ifx\conjecture\undefined
\newtheorem{conjecture}[theorem]{Conjecture}
\fi
\ifx\fact\undefined
\newtheorem{fact}[theorem]{Fact}
\fi
\ifx\claim\undefined
\newtheorem{claim}[theorem]{Claim}
\fi
\ifx\assum\undefined

\fi
\ifx\assum\undefined

\fi

\def\-{\raisebox{.75pt}{-}}

\newcommand{\n}[1]{$n$\nobreakdash-\hspace{0pt}}

\usepackage{cleveref}
\usepackage{caption}
\usepackage{subcaption}
\usepackage{tikz}
\usetikzlibrary{matrix}
\usetikzlibrary{arrows,shadows}
\usepackage{pgf,tikz}
\usetikzlibrary{positioning}
\usepackage{numprint}
\usepackage{enumitem}
\usepackage{cleveref}
\usepackage{csquotes}

\newcommand{\isep}{\mathrel{{.}\,{.}}\nobreak}
\newcommand{\thebenchmark}{GeneDisco}
\title{\thebenchmark{}: A Benchmark for Experimental Design in Drug Discovery}

\author{Arash Mehrjou$^1$, Ashkan Soleymani$^2$, Andrew Jesson$^3$, Pascal Notin$^3$, \\
  \textbf{Yarin Gal}$^3$, \textbf{Stefan Bauer}$^1$$^,$$^4$$^,$$^5$, \textbf{Patrick Schwab}$^1$ \\
  $^1$GlaxoSmithKline, Artificial Intelligence \& Machine Learning \\
  $^2$ MIT, $^3$ Department of Computer Science, University of Oxford \\
  $^4$ CIFAR Azrieli Global Scholar, $^5$ KTH Stockholm \\
}

\iclrfinalcopy \begin{document}

\maketitle

\begin{abstract}
In vitro cellular experimentation with genetic interventions, using for example CRISPR technologies, is an essential step in early-stage drug discovery and target validation that serves to assess initial hypotheses about causal associations between biological mechanisms and disease pathologies. With billions of potential hypotheses to test, the experimental design space for in vitro genetic experiments is extremely vast, and the available experimental capacity - even at the largest research institutions in the world - pales in relation to the size of this biological hypothesis space. Machine learning methods, such as active and reinforcement learning, could aid in optimally exploring the vast biological space by integrating prior knowledge from various information sources as well as extrapolating to yet unexplored areas of the experimental design space based on available data. However, there exist no standardised benchmarks and data sets for this challenging task and little research has been conducted in this area to date. Here, we introduce \thebenchmark{}, a benchmark suite for evaluating active learning algorithms for experimental design in drug discovery. \thebenchmark{} contains a curated set of multiple publicly available experimental data sets as well as open-source implementations of state-of-the-art active learning policies for experimental design and exploration.
\end{abstract}
\vspace{-1.00em}

\section{Introduction}
The discovery and development of new therapeutics is one of the most challenging human endeavours with success rates of around 5\% \citep{hay2014clinical,wong2019estimation}, timelines that span on average over a decade \citep{dickson2009cost, dickson2004key}, and monetary costs exceeding two billion United States (US) dollars \citep{dimasi2016innovation, berdigaliyev2020overview}. The successful discovery of drugs at an accelerated pace is critical to satisfy current unmet medical needs \citep{rawlins2004cutting, ringel2020breaking}, and, with thousands of potential treatments currently in development \citep{PharmaReport2018}, increasing the probability of success of new medicines by establishing causal links between drug targets and diseases \citep{nelson2015support} could introduce an additional hundreds of new and potentially life-changing therapeutic options for patients every year.

\begin{figure}[t]
\vspace{-3.25em}
\centering
    \includegraphics[width=0.70\textwidth]{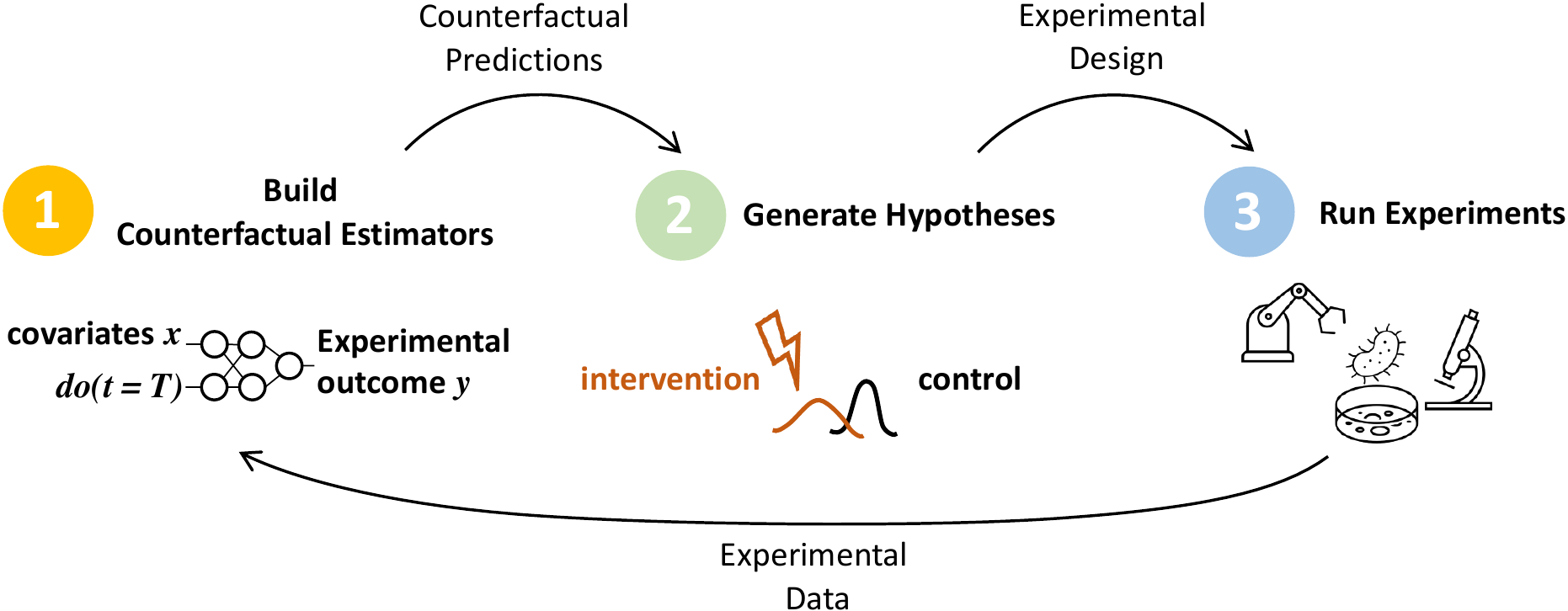}
    \caption{In the setting considered in this work, counterfactual estimators of experimental outcomes (step 1, left) are used to propose experimental hypotheses (step 2, center) for validation in in vitro experiments with genetic interventions (step 3, right), such as CRISPR knockouts, in order to discover potential causal associations between biological entities that could be relevant for the development of novel therapeutics. The trained counterfactual estimators can be used to direct the experimental search towards the space of biological interest, and thus more efficiently explore the vast space of genetic interventions. After every cycle, experimental data are generated that could lead to mechanistic scientific discoveries forming the basis for new therapeutics development, and guide subsequent experiment cycles with enhanced counterfactual estimators.}
    \label{fig:workflow}
\vspace{-1.5em}
\end{figure}

However, given the current estimate of around \numprint{20000} protein-coding genes \citep{pertea2018chess}, a continuum of potentially thousands of cell types and states under a multitude of environmental conditions \citep{trapnell2015defining,MACLEAN201832,worzfeld2017unique}, and tens of thousands of cellular measurements that could be taken \citep{hasin2017multi,chappell2018single}, the combinatorial space of biological exploration spans hundreds of billions of potential experimental configurations, and vastly exceeds the experimental capacity of even the world's largest research institutes. Machine learning methods, such as active and reinforcement learning, could potentially aid in optimally exploring the space of genetic interventions by prioritising experiments that are more likely to yield mechanistic insights of therapeutic relevance (\Cref{fig:workflow}), but, given the lack of openly accessible curated experimental benchmarks, there does not yet exist to date a concerted effort to leverage the machine learning community for advancing research in this important domain.

To bridge the gap between machine learning researchers versed in causal inference and the challenging task of biological exploration, we introduce \thebenchmark{}, an open benchmark suite for evaluating batch active learning algorithms for experimental design in drug discovery. \thebenchmark{} consists of several curated datasets, tasks and associated performance metrics, open implementations of state-of-the-art active learning algorithms for experimental design, and an accessible open-source code base for evaluating and comparing new batch active learning methods for biological discovery.

Concretely, the contributions presented in this work are as follows:

\begin{itemize}[noitemsep]
\item We introduce GeneDisco, an open benchmark suite for batch active learning for drug discovery that provides curated datasets, tasks, performance evaluation and open source implementations of state-of-the-art algorithms for experimental exploration.
\item We perform an extensive experimental baseline evaluation that establishes the relative performance of existing state-of-the-art methods on all the developed benchmark settings using a total of more than \numprint{20000} central processing unit (CPU) hours of compute time.
\item We survey and analyse the current state-of-the-art of active learning for biological exploration in the context of the generated experimental results, and present avenues of heightened potential for future research based on the developed benchmark.
\end{itemize}

\section{Related Work}

\textbf{Background.} Drug discovery is a challenging endeavour with (i) historically low probabilities of successful development into clinical-stage therapeutics \citep{hay2014clinical,wong2019estimation}, and, for many decades until recently \citep{ringel2020breaking}, (ii) declining industry productivity commonly referred to as \enquote{Eroom's law} \citep{scannell2012diagnosing}. Seminal studies by \citet{nelson2015support} and \citet{king2019drug} respectively first reported and later independently confirmed that the probability of clinical success of novel therapeutics increases up to three-fold if a medicine's molecular target is substantiated by high-confidence causal evidence from genome-wide association studies (GWAS) \citet{visscher201710}. With the advent of molecular technologies for genetic perturbation, such as Clustered Regularly Interspaced Short Palindromic Repeats (CRISPR) \citep{jehuda2018genome}, there now exist molecular tools for establishing causal evidence supporting the putative mechanism of a potential therapeutic hypothesis by means of in vitro experimentation beyond GWAS early on during the target identification and target validation stages of drug discovery \citep{rubin2019coupled,itokawa2016testing,harrer2019artificial,vamathevan2019applications}. Among other applications \citep{chen2018rise,ekins2019exploiting,mak2019artificial}, machine learning methods, such as active and reinforcement learning, could potentially aid in discovering the molecular targets with the highest therapeutic potential faster.

\textbf{Machine Learning for Drug Discovery.} There exist a number of studies that proposed, advanced or evaluated the use of machine learning algorithms for drug discovery: \citet{costa2010machine} used decision tree meta-classifier to identify genes associated with morbidity using protein-protein, metabolic, transcriptional, and sub-cellular interactions as input. \citet{jeon2014systematic} used a support vector machine (SVM) on protein expressions to predict target or non-target for breast pancreatic or ovarian cancers, and \citet{ament2018transcriptional} used least absolute shrinkage and selection operator (LASSO) regularised linear regression on transcription factor binding sites and transcriptome profiling to predict transcriptional changes associated with Huntington's disease. In the domain of human muscle ageing, \citet{mamoshina2018machine} used a SVM on deep features extracted from gene expression signatures in tissue samples from young and old subjects to discover molecular targets with putative involvement in muscle ageing. More recently, \citet{stokes2020deep} utilised deep learning to discover Halicin as a repurposed molecule with antibiotic activity in mice.

\textbf{Reinforcement and Active Learning.} 
The use of deep reinforcement learning for de novo molecular design has been extensively studied \citep{olivecrona2017molecular,popova2018deep,putin2018reinforced,lim2018molecular,blaschke2020memory,gottipati2020learning,horwood2020molecular}. 
Active learning for de novo molecular design has seen less attention \citep{dixit2016perturb, green2020bradshaw}, however, active learning for causal inference has seen increasing application toward causal-effect estimation \citep{deng2011active, schwab2018perfect, sundin2019active, schwab2020doseresponse, bhattacharyya2020learning, parbhoo2021ncore, qin2021budgeted, jesson2021causal, chau2021bayesimp}, and causal graph discovery \citep{ke2019learning, tong2001active, murphy2001active, hauser2014two, ghassami2018budgeted, ness2017bayesian, agrawal2019abcd, lorch2021dibs, annadani2021variational, scherrer2021learning}. 

\textbf{Benchmarks.}
Benchmark datasets play an important role in developing machine learning methodologies. 
Examples include ImageNet \citep{deng2009imagenet} or MSCOCO \citep{lin2014microsoft} for computer vision, as well as cart-pole \citep{barto1983neuronlike} or reinforcement learning \citep{ahmed2020causalworld}.
Validation of active learning for causal inference methods depends largely on synthetic data experiments due to the difficulty or impossibility of obtaining real world counterfactual outcomes. 
For causal-effect active learning, real world data with synthetic outcomes such as IHDP \citep{hill2011bayesian} or ACIC2016 \cite{dorie2019automated} are used. 
For active causal discovery, \emph{in silico} data such as DREAM4 \citep{prill2011dream} or the gene regulatory networks proposed by \citet{marbach2009generating} are used. 
Non synthetic data has been limited to protein signalling networks \citep{sachs2005causal} thus far.
Most similar to our work is the benchmark for de novo molecular design of \citet{brown2019guacamol}.

In contrast to existing works, we develop an open benchmark to evaluate the use of machine learning for efficient experimental exploration in an iterative batch active learning setting. To the best of our knowledge, this is the first study (i) to introduce a curated open benchmark for the challenging task of biological discovery, and (ii) to comprehensively survey and evaluate state-of-the-art active learning algorithms in this setting.

\section{Methodology}
\textbf{Problem Setting.} We consider the setting in which we are given a dataset consisting of covariates $X \in \RR^p$ with input feature dimensionality $p \in \NN$ and treatment descriptors $T \in \RR^q$ with treatment descriptor dimensionality $q \in \NN^+$ that indicate similarity between interventions. 
Our aim is to estimate the expectation of the conditional distribution of an unseen counterfactual outcome $Y_t \in \RR$ given observed covariates $X=x$ and intervention $do(T=t)$, $\hat{y}_t = \widehat{g}(X=x, do(T=t)) \approx \mathbb{E}[Y \mid X=x, do(T=t)]$.
This setting corresponds to the Rubin-Neyman potential outcomes framework \citep{rubin2005causal} adapted to the context of genetic interventions with a larger number of  parametric interventions. In the context of a biological experiment with genetic interventions, $y_t$ is the experimental outcome relative to a non-interventional control (e.g., change in pro-inflammatory effect) that is measured upon perturbation of the cellular system with intervention $t$, $x$ is a descriptor of the properties of the model system and/or cell donor (e.g., the immuno-phenotype of the cell donor), and $t$ is a descriptor of the genetic intervention (e.g., a CRISPR knockout on gene STAT1) that indicates similarity to other genetic interventions that could potentially be applied. In general, certain ranges of $y_t$ may be preferable for further pursuit of an intervention $T=t$ that inhibits a given molecular target - often, but not necessarily always, larger absolute values that move the experimental result more are of higher mechanistic interest. We note that the use of an empty covariate vector $X = x_0$ with $p=0$ is permissible if all experiments are performed in the same model system with the same donor properties. In in vitro experimentation, the set of all possible genetic interventions $\Dcal_\textrm{pool}=\{t_i\}_{i=1}^{n_{\textrm{pool}}}$ is typically known a-priori and of finite size (e.g., knockout interventions on all \numprint{20000} protein-coding genes).

\textbf{Batch Active Learning.} In the considered setting, reading out additional values for yet unexplored interventions $t$ requires a lab experiment and can be costly and time-consuming. Lab experiments are typically conducted in parallelised manner, for example by performing a batch of multiple interventions at once in an experimental plate. Our overall aim is to leverage the counterfactual estimator $\widehat{g}$ trained on the available dataset to simulate future experiments with the aim of maximising an objective function $\mathcal{L}$ in the next iteration with as few interventions as possible. For the purpose of this benchmark, we consider the counterfactual mean squared error (MSE) of $\widehat{g}$ in predicting experimentally observed true outcomes $y_t$ from predicted outcomes $\hat{y}_t$ as the optimisation objective $\mathcal{L}_\text{MSE} = \text{MSE}(y_t, \hat{y}_t)$. Depending on context, other objective functions, such as for example maximising the number of discovered molecular targets with certain properties of interest (e.g., druggability \citep{keller2006practical}) may also be sensible in the context of biological exploration. At every time point, a new counterfactual estimator $\widehat{g}$ is trained with the entire available experimental dataset, and used to propose the batch of $b$ interventions to be explored in the next iteration with the batch size $b \in \NN^+$. When using $\mathcal{L}_\text{MSE}$, this setting corresponds to batch active learning with the optimisation objective of maximally improving the counterfactual estimator $\widehat{g}$.

\textbf{Acquisition Function.} An acquisition function $\Dcal^k = \alpha(g(t), \Dcal^k_\textrm{avail})$ takes the model and the set of all available interventions $\Dcal^k_\textrm{avail}$ in cycle $k$ as input and outputs the set of interventions $\Dcal^k$ that are most informative after the $k$th experimental cycle with cycle index $k\in [K] = [0 \isep K]$ where $K\in\NN^+$ is the maximum number of cycles that can be run. Formally speaking, the acquisition function $\alpha:\Pcal({\Dcal_\textrm{avail}})\times \Gcal \to \Pcal({\Dcal_\textrm{avail}})$ takes a subset of all possible interventions that have not been tried so far ($\Dcal_\textrm{avail}$), together with the trained model ($\widehat{g}$) derived from the cumulative data collected over the previous cycles, and outputs a subset of the available interventions $\Dcal^k$ that are likely to be most useful under $\mathcal{L}$ to obtain $\widehat{g} \in \Gcal$ as a better estimate of $\mathbb{E}[Y \mid X=x, do(T=t)]$ with $\Gcal$ being the space of the models which can be, e.g. the space of models and (hyper-)parameters.

\section{Datasets, Metrics \& Baselines}
\label{sec:datasetandbaselines}

The GeneDisco benchmark curates and standardizes two types of datasets: three standardized feature sets describing interventions $t$ (inputs to counterfactual estimators; {\Cref{Section 4: Treatment Descriptors}}), and four different in vitro genome-wide CRISPR experimental assays (predicted counterfactual outcomes; {\Cref{Section 4: Assays}}), each measuring a specific outcome $y_t$  following possible interventions $T$.
We perform an extensive evaluation across these datasets, leveraging two different model types ({\Cref{Section 4: Models}}) and nine different acquisition functions ({\Cref{Section 4: Acquisition functions}}). Since all curated assay datasets contained only outcomes for only one model system, we used the empty covariate set $X = x_0$ for all evaluated benchmark configurations. The metrics used to evaluate the various experimental conditions (acquisition functions and model types) include model performance (\Cref{fig:bnn_feat_string_alldatasets_allbathcsizes_small}) and the ratio of discovered interesting hits (\Cref{fig:hitratio_bnn_feat_string_alldatasets_allbathcsizes_small})  as a function of number of samples queried.

\subsection{Treatment Descriptors} 
\label{Section 4: Treatment Descriptors}
The treatment descriptors $T$ characterize a genetic intervention and generally should correspond to data sources that are informative as to a genes' functional similarity - i.e. defining which genes if acted upon, would potentially respond similarly to perturbation. Any dataset considered for use as a treatment descriptor must be available for all potentially available interventions $\Dcal_\textrm{pool}$ in the considered experimental setting. In \thebenchmark{}, we provide three standardised gene descriptor sets for genetic interventions, and furthermore enable users to provide custom treatment descriptors via a standardised programming interface:

\textbf{Achilles.} The Achilles project generated dependency scores across cancer cell lines by assaying 808 cell lines covering a broad range of tissue types and cancer types \citep{Dempster720243}. The genetic intervention effects are based on interventional CRISPR screens performed across the included cell lines. When using the Achilles treatment descriptors, each genetic intervention is summarized using a gene representation $T$ with $q=808$ corresponding to the dependency scores measured in each cell line. In Achilles, after processing and normalisation (see  \cite{Dempster720243}), the final dependency scores provided are such that the median negative control (non-essential) gene effect for each cell line is 0, and the median positive control (essential) gene effect for each cell line is -1. The rationale for using treatment descriptors based on the Achilles dataset is that genetic effects measured across the various tissues and cancer types in the 808 cell line assays included in \citep{Dempster720243} could serve as a similarity vector in functional space that may extrapolate to other biological contexts due to its breadth.

\textbf{Search Tool for the Retrieval of Interacting Genes/Proteins (STRING) Network Embeddings.} The STRING \citep{szklarczyk2021string} database collates known and predicted protein-protein interactions (PPIs) for both physical as well as for functional associations. In order to derive a vector representation  suitable to serve as a genetic intervention descriptor $T$, we utilised the network embeddings of the PPIs contained in STRING as provided by \citet{cho2016compact,cho2015diffusion} with dimensionality $q = 799$. PPI network embeddings could be an informative descriptor of functional gene similarity since proteins that functionally interact with the same network partners may serve similar biological functions \citep{vazquez2003global,saha2014funpred}.

\textbf{Cancer Cell Line Encyclopedia (CCLE).} The CCLE \citep{nusinow2020quantitative} project collected quantitative proteomics data from thousands of proteins by mass spectrometry across 375 diverse cancer cell lines. The generated protein quantification profiles with dimensionality $q=420$ could indicate similarity of genetic interventions since similar expression profiles across a broad range of biological contexts may indicate functional similarity.

\textbf{Custom Treatment Descriptors.} Additional, user-defined treatment descriptors can be evaluated in \thebenchmark{} by implementing the standardised dataset interface provided within.

\subsection{Assays}
\label{Section 4: Assays}

As ground-truth interventional outcome datasets, we leverage various genome-wide CRISPR screens, primarily from the domain of immunology, that evaluated the causal effect of intervening on a large number of genes in cellular model systems in order to identify the genetic perturbations that induce a desired phenotype of interest.

\subsubsection{Regulation of Human T cells proliferation}

\paragraph{Experimental setting.} This assay is based on \citet{Shifrut2018Tcells}. After isolating CD8$^{+}$ T cells from human donors, \citet{Shifrut2018Tcells} performed a genome-wide loss-of-function screen to identify genes that impact the proliferation of T cells following stimulation with T cell receptors.
\vspace{-10pt}
\paragraph{Measurement.} The measured outcome is the proliferation of T cells in response to T cell receptor stimulation. Cells were labeled before stimulation with CFSE (a fluorescent cell staining dye). Proliferation of cells is measured 4 days following stimulation by FACS sorting (a flow cytometry technique to sort cells based on their fluorescence).
\vspace{-10pt}
\paragraph{Importance.} Human T cells play a central role in immunity and cancer immunotherapies. The identification of genes driving T cell proliferation could serve as the basis for new preclinical drug candidates or adoptive T cell therapies that help eliminate cancerous tumors.

\subsubsection{Interleukin-2 production in primary human T cells}

\paragraph{Experimental setting.} This dataset is based on a genome-wide CRISPR interference (CRISPRi) screen in primary human T cells to uncover the genes regulating the production of Interleukin-2 (IL-2). CRISPRi screens test for loss-of-function genes by reducing their expression levels. IL-2 is a cytokine produced by CD4$^{+}$ T cells and is a major driver of T cell expansion during adaptive immune responses. Assays were performed on primary T cells from 2 different donors. The detailed experimental protocol is described in \citet{Schmidt2021Cytokine}.
\vspace{-10pt}
\paragraph{Measurement.} Log fold change (high/low sorting bins) in IL-2 normalized read counts (averaged across the two donors). Sorting was done via flow cytometry after intracellular cytokine staining. 
\vspace{-10pt}
\paragraph{Importance.} IL-2 is central to several immunotherapies against cancer and autoimmunity.

\subsubsection{Interferon-$\gamma$ production in primary human T cells}

\paragraph{Experimental setting.} This data is also based on \citet{Schmidt2021Cytokine}, except that this experiment was performed to understand genes driving production of Interferon-$\gamma$ (IFN-$\gamma$). IFN-$\gamma$ is a cytokine produced by CD4$^{+}$ and CD8$^{+}$ T cells that induces additional T cells. 
\vspace{-10pt}
\paragraph{Measurement.} Log fold change (high/low sorting bins) in IFN-$\gamma$ normalized read counts (averaged across the two donors).
\vspace{-10pt}
\paragraph{Importance.} IFN-$\gamma$ is critical to cancerous tumor killing and resistance to IFN-$\gamma$ is one escape mechanism for malignant cells.

\subsubsection{Vulnerability of Leukemia cells to NK cells}

\paragraph{Experimental setting.} This genome-wide CRISPR screen was performed in the K562 cell line to identify genes regulating the sensitivity of leukemia cells to cytotoxic activity of primary human NK cells. Detailed protocol is described in \citet{zhuang2019LeukemiaNK}.
\vspace{-10pt}
\paragraph{Measurement.} Log fold counts of gRNAs in surviving K562 cells (after exposition to NK cells) compared to control (no exposition to NK cells). Gene scores are normalized fold changes for all gRNAs targeting this gene.
\vspace{-10pt}
\paragraph{Importance.} Better understanding and control over the genes that drive the vulnerability of leukemia cells to NK cells will help improve anti-cancer treatment efficacy for leukemia patients, for example by preventing relapse during hematopoeitic stem cell transplantation. 

\subsection{Models} 
\label{Section 4: Models}

Parametric or non-parametric models can be used to model the conditional expected outcomes, $\mathbb{E}[Y \mid X=x, do(T=t)]$. 
Parametric models assume that the outcome $Y$ has density $f(y \mid t, \omega)$ conditioned on the intervention $t$ and the parameters of the model $\omega$ (we drop $x_0$ for compactness). 
A common assumption for continuous outcomes is a Gaussian distribution with density $f(y \mid t, \omega) = \mathcal{N}(y \mid \widehat{g}(t; \omega), \sigma^2)$, which assumes that $y$ is a deterministic function of $\widehat{g}(t; \omega)$ with additive Gaussian noise scaled by $\sigma^2$. 
Bayesian methods treat the model parameters $\omega$ as instances of the random variable $\Omega \in \mathcal{W}$ and aim to model the posterior density of the parameters given the data, $f(\omega \mid \Dcal)$. 
For high-dimensional, large-sample data, such as we explore here, a variational approximation to the posterior is often used, $q(\omega \mid \Dcal)$ \citep{mackay1992practical, hinton1993keeping, barber1998ensemble, gal2016dropout}. 
In this work we use Bayesian Neural Networks (BNNs) to approximate the posterior over model parameters. 
A BNN gives $\widehat{g}^{k-1}(t) = \frac{1}{m} \sum_{j=1}^m \widehat{g}(t; \omega_j^{k-1})$, where $\widehat{g}(t; \omega_j^{k-1})$ is a unique functional estimator of $\mathbb{E}[Y \mid X=x, do(T=t)]$ induced by $\omega_j^{k-1} \sim q(\omega \mid \Dcal^{k-1}_\textrm{cum})$: a sample from the approximate posterior over parameters given the cumulative data at acquisition step $k-1$.
We also use non-parametric, non-Bayesian, Random Forest Regression \citep{breiman2001random}.
A Random Forest gives $\widehat{g}^{k-1}(t) = \frac{1}{m} \sum_{j=1}^m \widehat{g}_j^{k-1}(t)$, where $\widehat{g}_j^{k-1}(t)$ is a unique functional estimator of $\mathbb{E}[Y \mid X=x, do(T=t)]$ indexed by the $j$th sample in the ensemble of $m$ trees trained on $\Dcal^{k-1}_\textrm{cum})$.
In the following, we will define our acquisition functions in terms of parametric models, but the definitions are easily adapted for non-parametric models as well.

\subsection{Acquisition Functions}
\label{Section 4: Acquisition functions}

\textbf{Random.} As a baseline we look at random acquisition. 
Random acquisition at cycle $k$ can be seen as uniformly sampling data from $\Dcal^k_\textrm{avail}$:
\begin{equation}
\small
    \alpha_{\textrm{Random}}(\widehat{g}^{k-1}(t), \Dcal^k_\textrm{avail}) = \{t_1, \dots t_b\} \sim \left\{ t_i ; \frac{1}{n_{\textrm{avail}}} \right\}_{i = 1}^{n_{\textrm{avail}}}.
\end{equation}
Here, the acquisition function samples $b$ elements without replacement from the set of $n_{\textrm{avail}}$ elements. 
The set element ($t_i$) is on the left of the semicolon, and the probability of the element being acquired ($\frac{1}{n_{\textrm{avail}}}$) is on the right of the semicolon. 
This convention will be used again below.

\textbf{BADGE.} BADGE looks to maximize the diversity of acquired samples, but, in contrast to Coreset, it additionally takes the uncertainty of the prediction into account \citep{ash2019deep}. 
If the true label $y$ were observed, BADGE would proceed by maximizing the diversity of samples based on the gradient of the loss function $l$ with respect to the weights of the final layer of the most recently trained model $\tilde{w}^{k-1}$: $\frac{\partial}{\partial \tilde{\omega}^{k-1}} l(y, \widehat{g}(t; \omega))$. Intuitively, it asks how much would our parameters change if we observed the labeled outcome for this example?
However, the true label $y$ is not yet observed.
\citet{ash2019deep} explore BADGE in the classification setting. For a two class problem, where $f(y \mid t, \omega) = \textrm{Bernoulli}(y \mid \widehat{g}(t; \omega))$, they  propose using the class with the highest predicted probability, $\widehat{y} = \argmax_{y \in {0, 1}}{f(y \mid t, \omega)}$, to approximate the gradient as $\frac{\partial}{\partial \tilde{\omega}^{k-1}} l(\widehat{y}, \widehat{g}(t; \omega))$. 
This does not directly translate to the regression setting, as under our modelling assumptions the $y$ with the highest predicted likelihood corresponds exactly to $\widehat{g}(t; \omega)$, which would lead to a loss of zero, and gradients of zero. 
As a starting point, we instead take $\widehat{y}$ as a random sample from  $f(y \mid t, \omega) = \mathcal{N}(y \mid \widehat{g}(t; \omega), \sigma^2)$. 
We then use the same $k$-means++ algorithm as \citet{ash2019deep} to approximate:
\begin{equation}
    \begin{split}
        \small
        \alpha&_{\textrm{BADGE}}(\widehat{g}^{k-1}(t), \Dcal^k_\textrm{avail}) \\
        &= \argmin_{\{t_1, \dots t_b\} \in \Dcal^k_\textrm{avail}} \argmax_{t_i \in \Dcal^k_\textrm{avail}} \argmin_{t_j \in \Dcal^k_\textrm{avail} \cup \Dcal^{k-1}_\textrm{cum}} \Delta\left(
            \frac{\partial l(\widehat{y}, \widehat{g}(t_i; \omega^{k-1}))}{\partial \tilde{\omega}^{k-1}},
            \frac{\partial l(\widehat{y}, \widehat{g}(t_j; \omega^{k-1}))}{\partial \tilde{\omega}^{k-1}}
        \right)
    \end{split}
\end{equation}
where $\Delta$ is again the Euclidean distance. 

\textbf{Bayesian Active Learning by Disagreement (BALD).} Given an uncertainty aware model, such as a BNN or Random Forest
we can now take an information theoretic approach to selecting interventions from the pool data. \citet{houlsby2011bayesian} frame active learning as looking to maximize the information gain about the model parameters if we observe the outcome $Y = y$ given model inputs. Formally, the information gain is given by the mutual information between the random variables $Y$ and $\Omega$ given the intervention $t$ and acquired training data $\Dcal^{k-1}_\textrm{cum}$ up until acquisition step $k$:
\begin{equation}
\small
    \begin{split}
        \Ical(Y ; \Omega \mid t, \Dcal^{k-1}_\textrm{cum}) &= H(Y \mid t, \Dcal^{k-1}_\textrm{cum}) - H(Y \mid \Omega, t, \Dcal^{k-1}_\textrm{cum}) \\
        &= H(Y \mid t, \Dcal^{k-1}_\textrm{cum}) - \mathbb{E}_{f(\omega \mid \Dcal^{k-1}_\textrm{cum})} H(Y \mid \omega, t).
    \end{split}
\end{equation}
Under the assumed model we have
\begin{equation}
\small
    \begin{split}
        \Ical(Y ; \Omega \mid t, \Dcal^{k-1}_\textrm{cum}) 
        &= \frac{1}{2} \log{\left( \frac{\sigma^2 + \mathbb{E}_{f(\omega \mid \Dcal^{k-1}_\textrm{cum})} \left[ \widehat{g}(t; \omega)^2 \right] - \mathbb{E}_{f(\omega \mid \Dcal^{k-1}_\textrm{cum})} \left[ \widehat{g}(t; \omega) \right]^2 }{\sigma^2} \right)},
    \end{split}
\end{equation}
which leads to the following estimator setting $\sigma^2 = 1$
\begin{equation}
\small
    \widehat{\Ical}(Y ; \Omega \mid t, \Dcal^{k-1}_\textrm{cum}) = \frac{1}{2} \log{\left( 1 + \frac{1}{m} \sum_{j=1}^{m} \left( \widehat{g}(t; \omega_j^{k-1}) - \frac{1}{m} \sum_{j=1}^{m} \widehat{g}(t; \omega_j^{k-1}) \right)^2 \right)}.
\end{equation}
We look at two acquisition functions for BALD. First, we consider the naive batch acquisition $\alpha_{\textrm{BALD}}$ proposed by \citet{gal2017deep} which acquires the the top $b$ examples from $\Dcal^k_\textrm{avail}$:
\begin{equation}
\small
    \alpha_{\textrm{BALD}}(\widehat{g}^{k-1}(t), \Dcal^k_\textrm{avail}) = \argmax_{\{t_1, \dots t_b\} \in \Dcal^k_\textrm{avail}} \sum_{i=1}^b \widehat{\Ical}(Y ; \Omega \mid t_i, \Dcal^{k-1}_\textrm{cum}).
\end{equation}
This method will be referred to as \texttt{topuncertain} in the plots later. And second, we consider $\alpha_{\textrm{SoftBALD}}$ which randomly samples $b$ interventions from $\Dcal^k_\textrm{avail}$ weighted by a tempered softmax function \citep{kirsch2021simple}:
\begin{equation}
\small
    \alpha_{\textrm{SoftBALD}}(\widehat{g}^{k-1}(t), \Dcal^k_\textrm{avail}) 
    = \{t_1, \dots t_b\} \sim \left\{ t_i ; \frac{
            \exp{\left( \frac{1}{\textrm{Temp}}\widehat{\Ical}(Y ; \Omega \mid t_i, \Dcal^{k-1}_\textrm{cum}) \right)}
        }{
            \sum_{l=1}^{n_{\textrm{avail}}} \exp{\left( \frac{1}{\textrm{Temp}}\widehat{\Ical}(Y ; \Omega \mid t_l, \Dcal^{k-1}_\textrm{cum})\right)}} 
        \right\}_{i = 1}^{n_{\textrm{avail}}},
\end{equation}
where $\textrm{Temp} > 0$ is a user defined constant. This method will be referred to as \texttt{softuncertain} in the plots. As $\textrm{Temp} \to \infty$, $\alpha_{\textrm{SoftBALD}}$ will behave more like $\alpha_{\textrm{Random}}$. And as $\textrm{Temp} \to 0$, $\alpha_{\textrm{SoftBALD}}$ will behave more like $\alpha_{\textrm{BALD}}$. The remaining acquisition functions (Coreset, Margin Sample (\texttt{Margin}), Adversarial Basic Iteractive Method (\texttt{AdvBIM}), $k$-means Sampling (\texttt{kmeansdata and kmeansembed})) included in the benchmark are described in detail in \Cref{sec:acq_fun_continued}.

\section{Experimental Evaluation}
\label{Section 5 - Experiments}

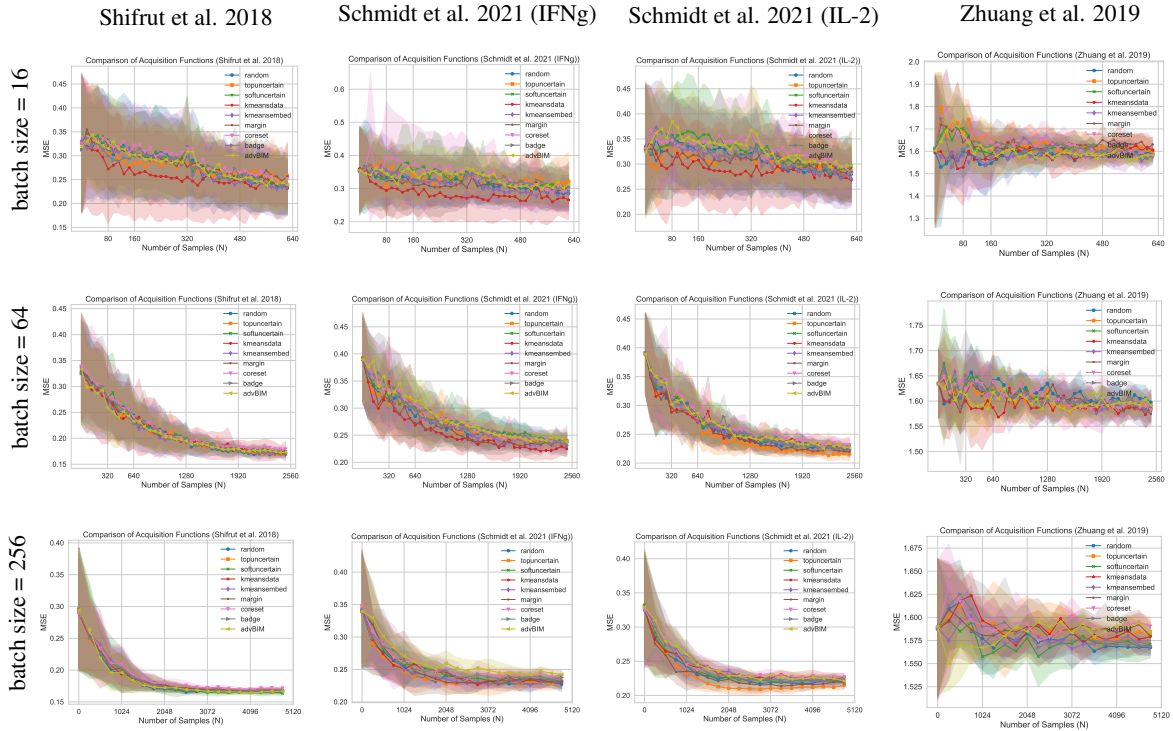
\begin{figure*}
    \vspace{-7em}
        \centering
        
        \makebox[0.72\paperwidth]{\begin{tikzpicture}[ampersand replacement=\&]
            \matrix (fig) [matrix of nodes]{ 
\begin{subfigure}{0.27\columnwidth}
                    \hspace{-17mm}
                    \centering
                    \resizebox{\linewidth}{!}{
                        \begin{tikzpicture}
                            \node (img)  {\includegraphics[width=\textwidth]{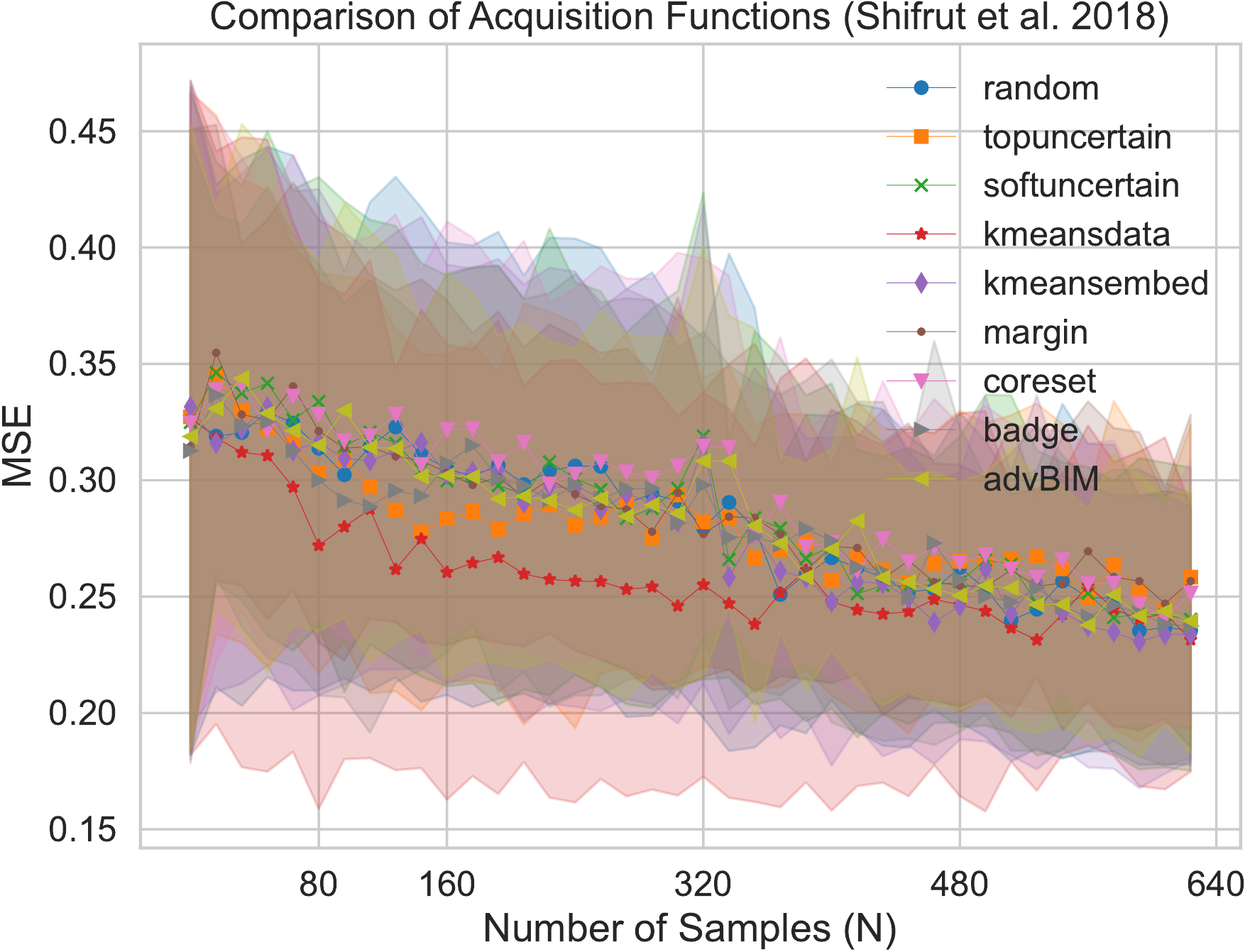}};
                        \end{tikzpicture}
                    }
                \end{subfigure}
                \&
                 \begin{subfigure}{0.27\columnwidth}
                    \hspace{-23mm}
                    \centering
                    \resizebox{\linewidth}{!}{
                        \begin{tikzpicture}
                            \node (img)  {\includegraphics[width=\textwidth]{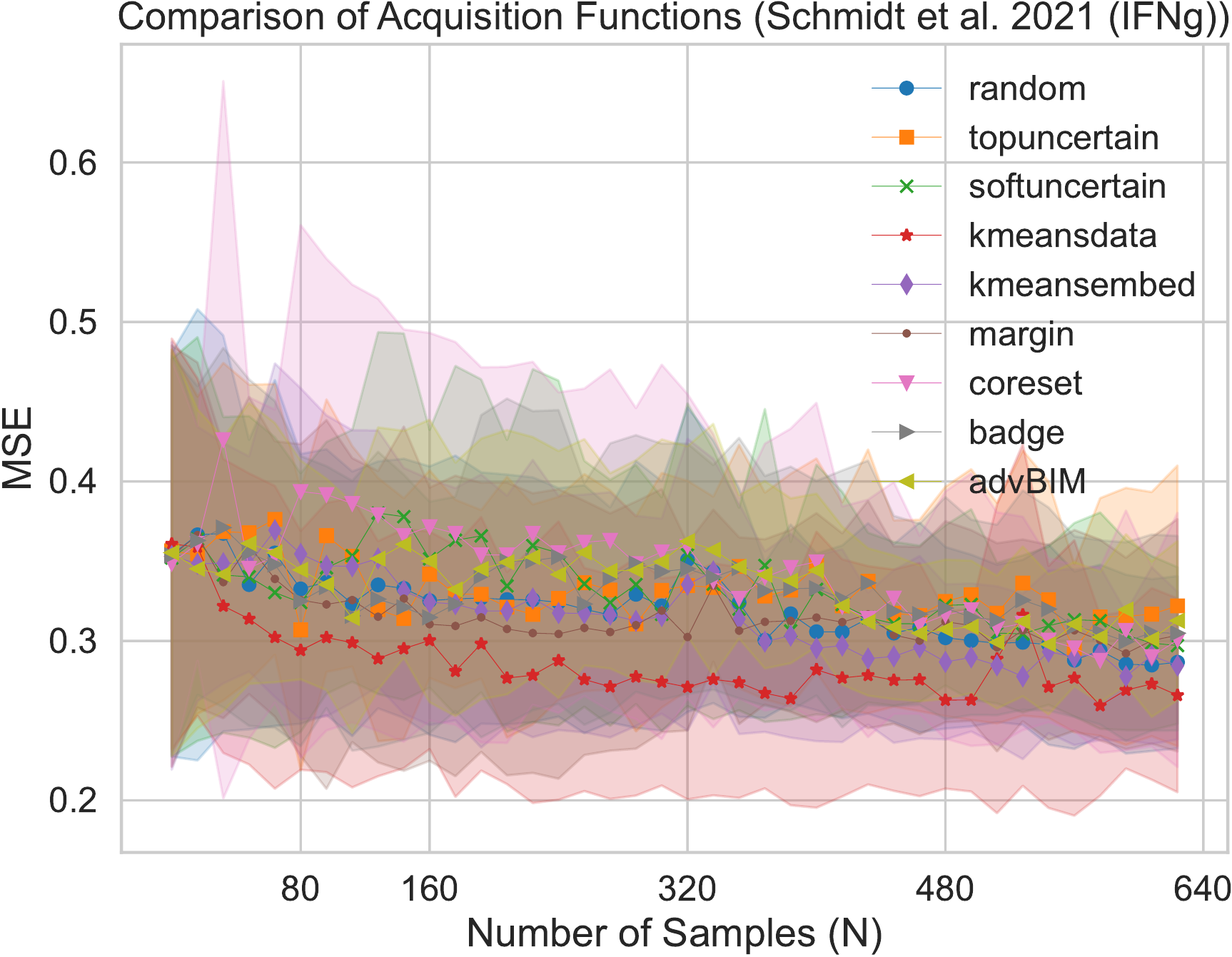}};
                        \end{tikzpicture}
                    }
                \end{subfigure}
                \&
                 \begin{subfigure}{0.27\columnwidth}
                    \hspace{-28mm}
                    \centering
                    \resizebox{\linewidth}{!}{
                        \begin{tikzpicture}
                            \node (img)  {\includegraphics[width=\textwidth]{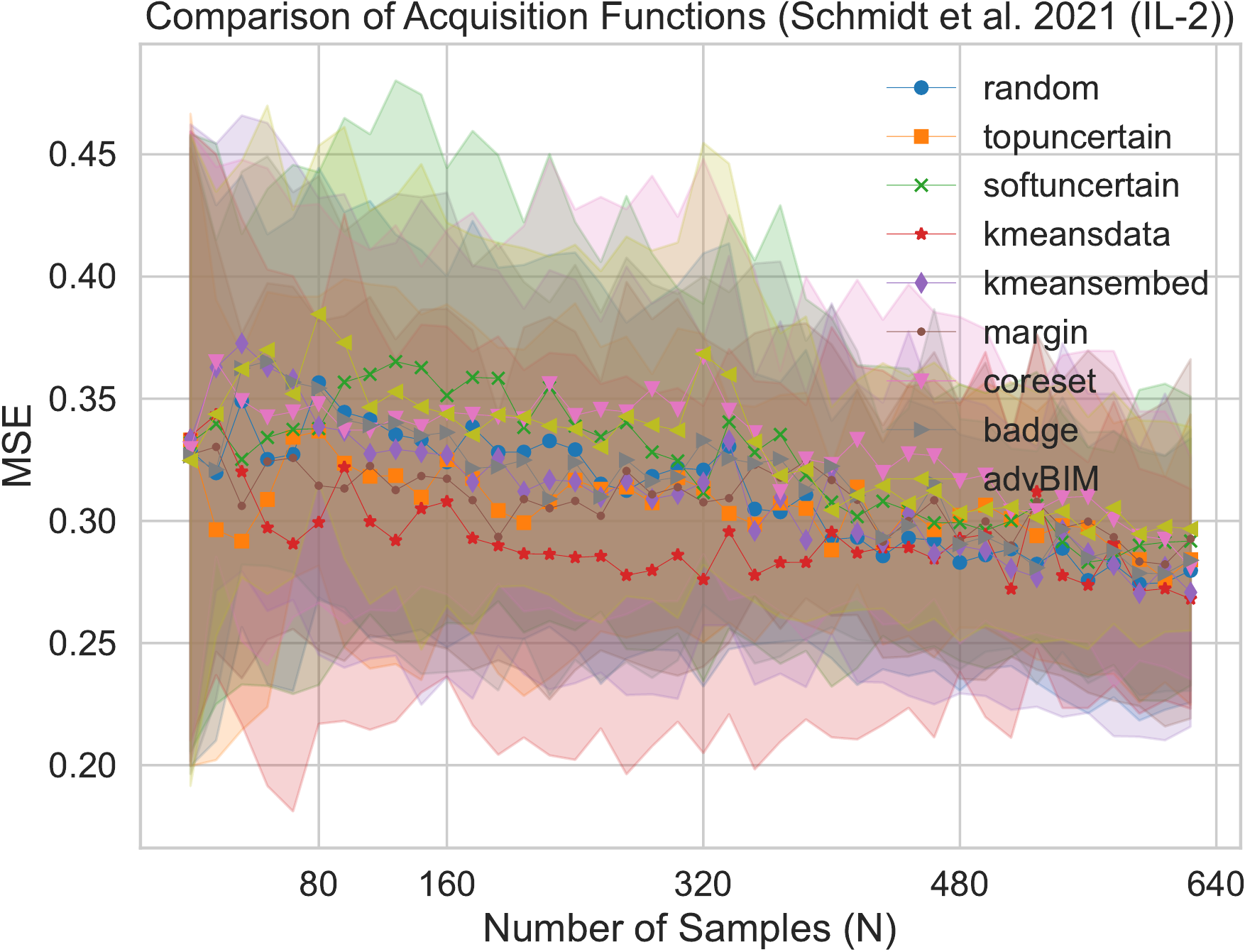}};
                        \end{tikzpicture}
                    }
                \end{subfigure}
                \&
                \begin{subfigure}{0.28\columnwidth}
                    \hspace{-32mm}
                    \centering
                    \resizebox{\linewidth}{!}{
                        \begin{tikzpicture}
                            \node (img)  {\includegraphics[width=\textwidth]{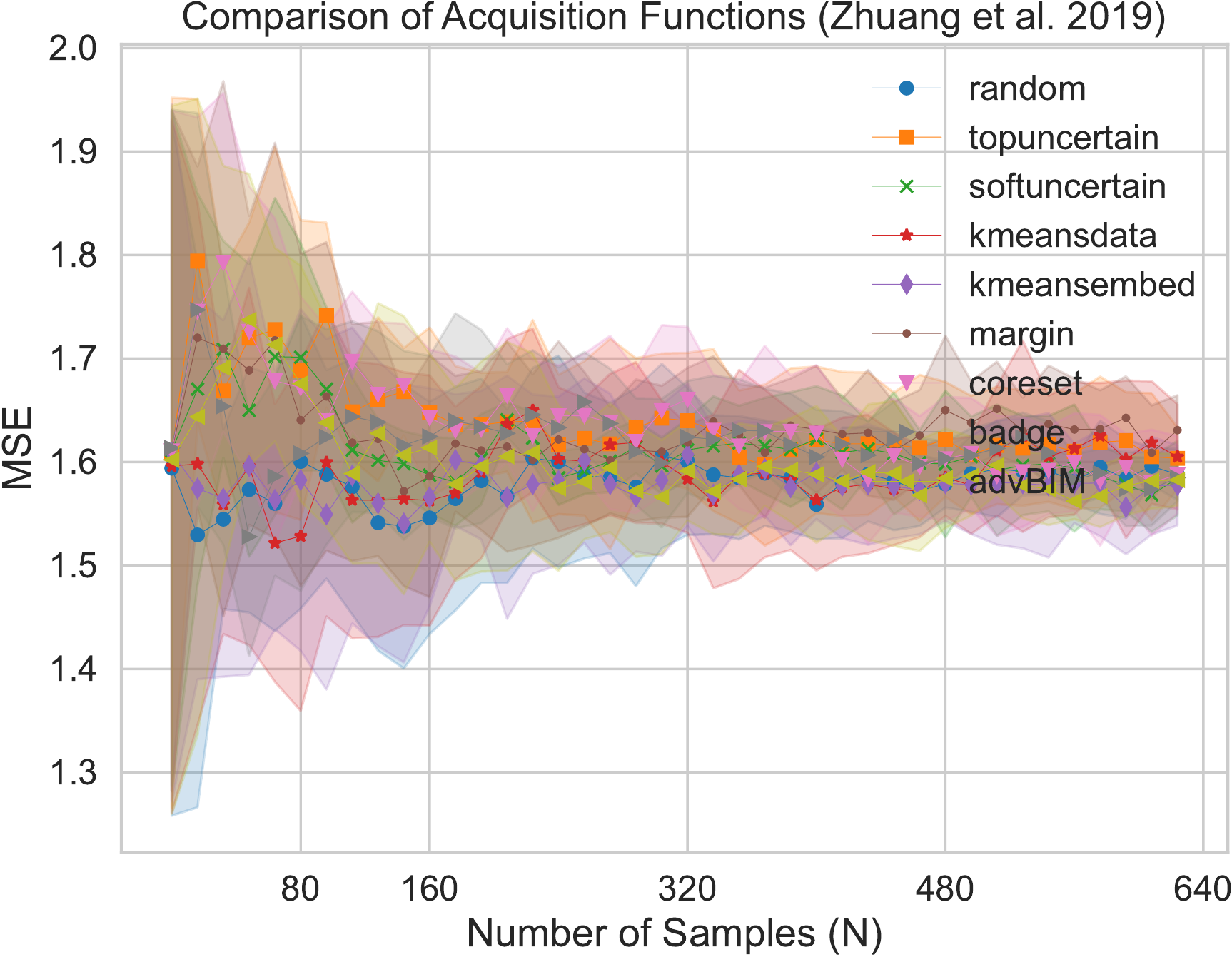}};
                        \end{tikzpicture}
                    }
                \end{subfigure}
                \&
            \\
\begin{subfigure}{0.27\columnwidth}
                    \hspace{-17mm}
                    \centering
                    \resizebox{\linewidth}{!}{
                        \begin{tikzpicture}
                            \node (img)  {\includegraphics[width=\textwidth]{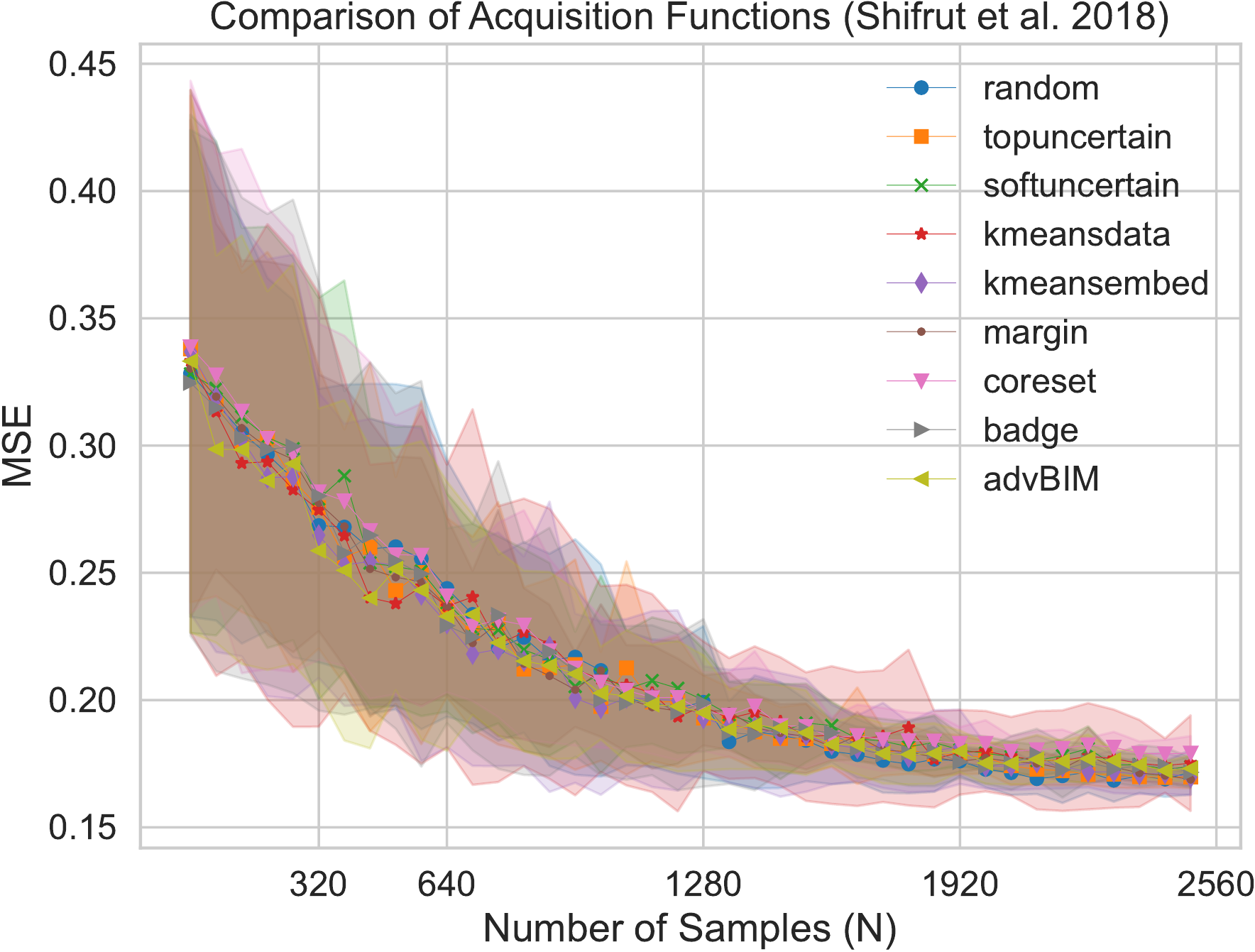}};
                        \end{tikzpicture}
                    }
                \end{subfigure}
                \&
                \begin{subfigure}{0.27\columnwidth}
                    \hspace{-23mm}
                    \centering
                    \resizebox{\linewidth}{!}{
                        \begin{tikzpicture}
                            \node (img)  {\includegraphics[width=\textwidth]{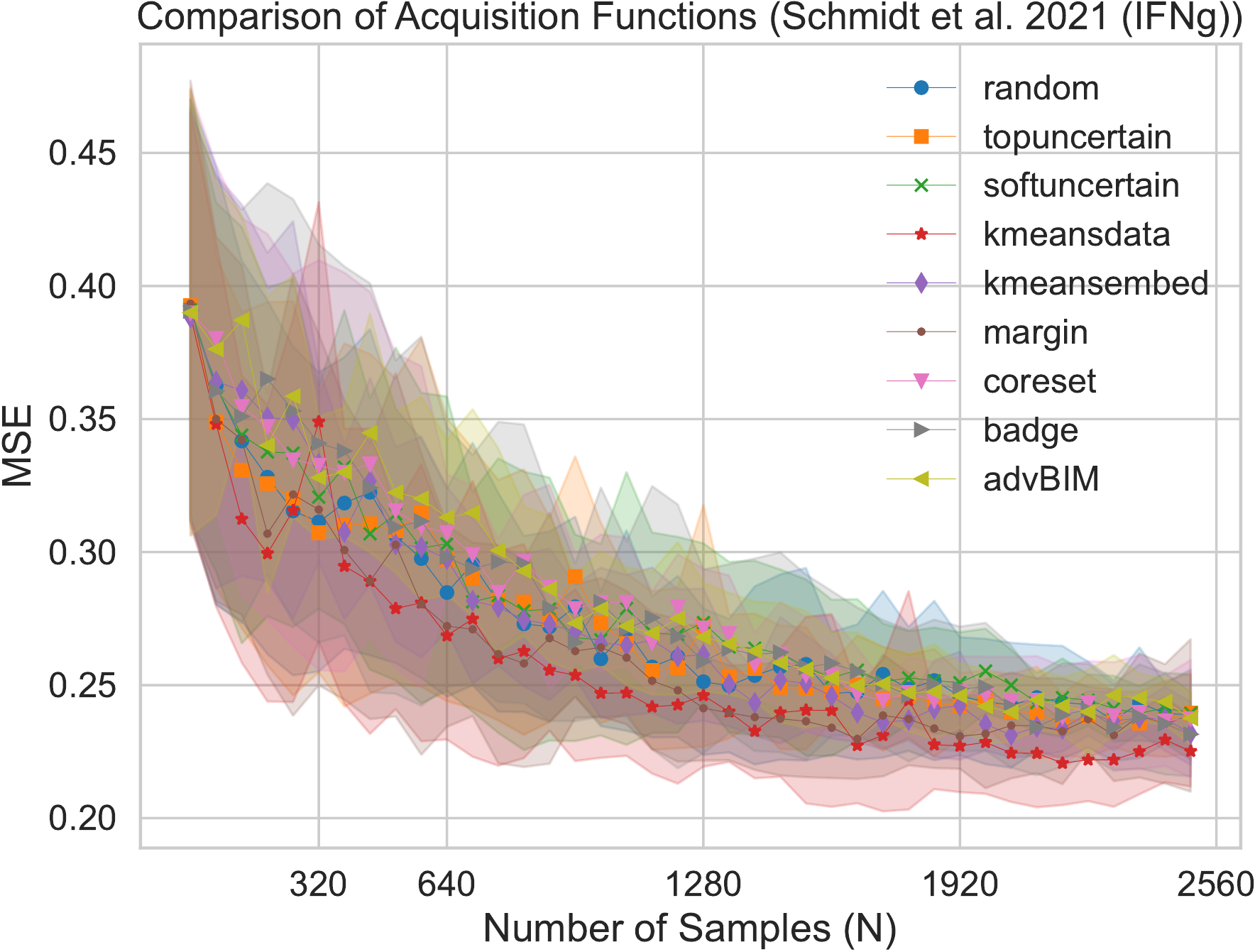}};
                        \end{tikzpicture}
                    }
                \end{subfigure}
                \&
                \begin{subfigure}{0.27\columnwidth}
                    \hspace{-28mm}
                    \centering
                    \resizebox{\linewidth}{!}{
                        \begin{tikzpicture}
                            \node (img)  {\includegraphics[width=\textwidth]{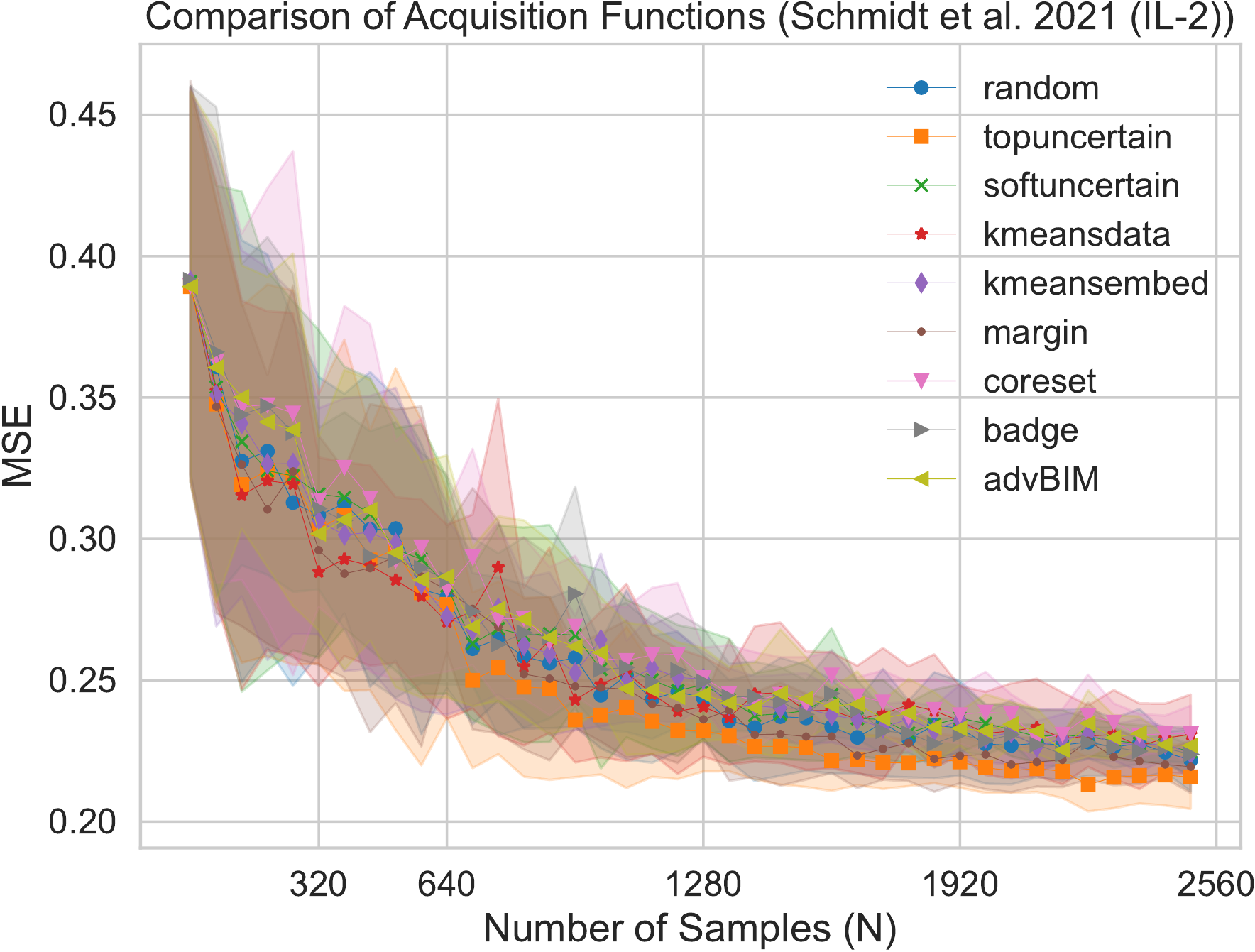}};
                        \end{tikzpicture}
                    }
                \end{subfigure}
                \&
                \begin{subfigure}{0.28\columnwidth}
                    \hspace{-32mm}
                    \centering
                    \resizebox{\linewidth}{!}{
                        \begin{tikzpicture}
                            \node (img)  {\includegraphics[width=\textwidth]{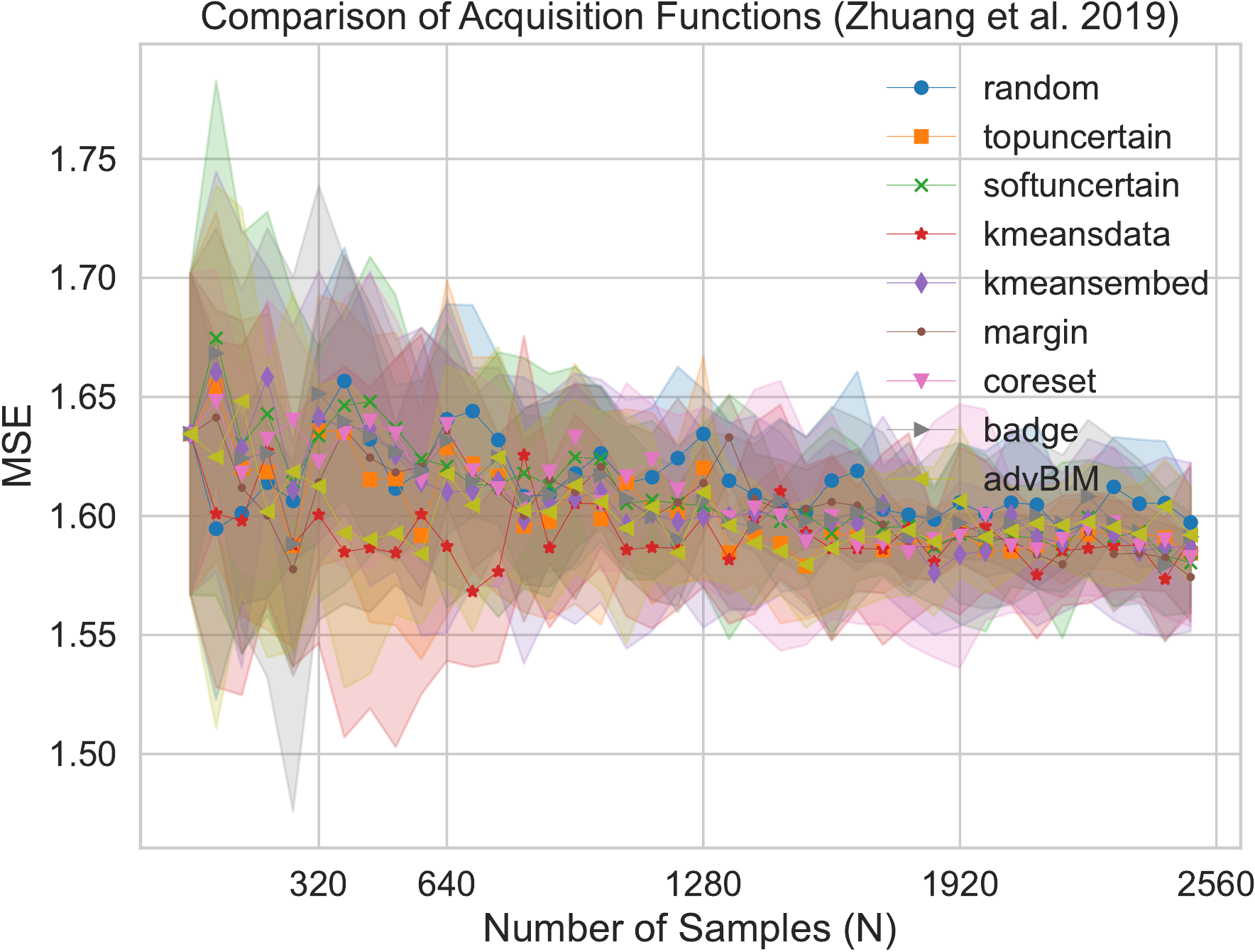}};
                        \end{tikzpicture}
                    }
                \end{subfigure}
                \&
                \\
\begin{subfigure}{0.275\columnwidth}
                    \hspace{-17mm}
                    \centering
                    \resizebox{\linewidth}{!}{
                        \begin{tikzpicture}
                            \node (img)  {\includegraphics[width=\textwidth]{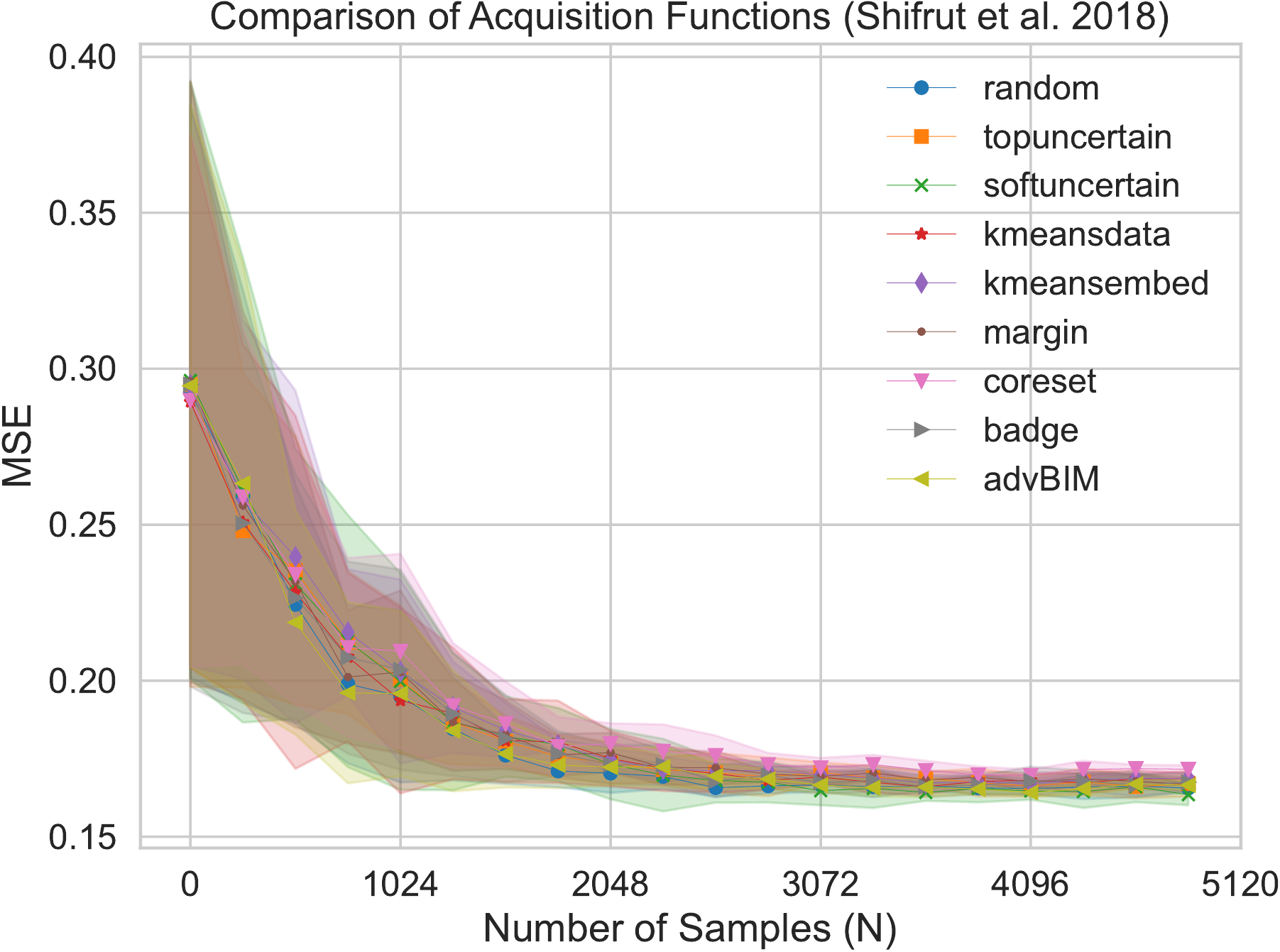}};
                        \end{tikzpicture}
                    }
                \end{subfigure}
                \&
                \begin{subfigure}{0.27\columnwidth}
                    \hspace{-23mm}
                    \centering
                    \resizebox{\linewidth}{!}{
                        \begin{tikzpicture}
                            \node (img)  {\includegraphics[width=\textwidth]{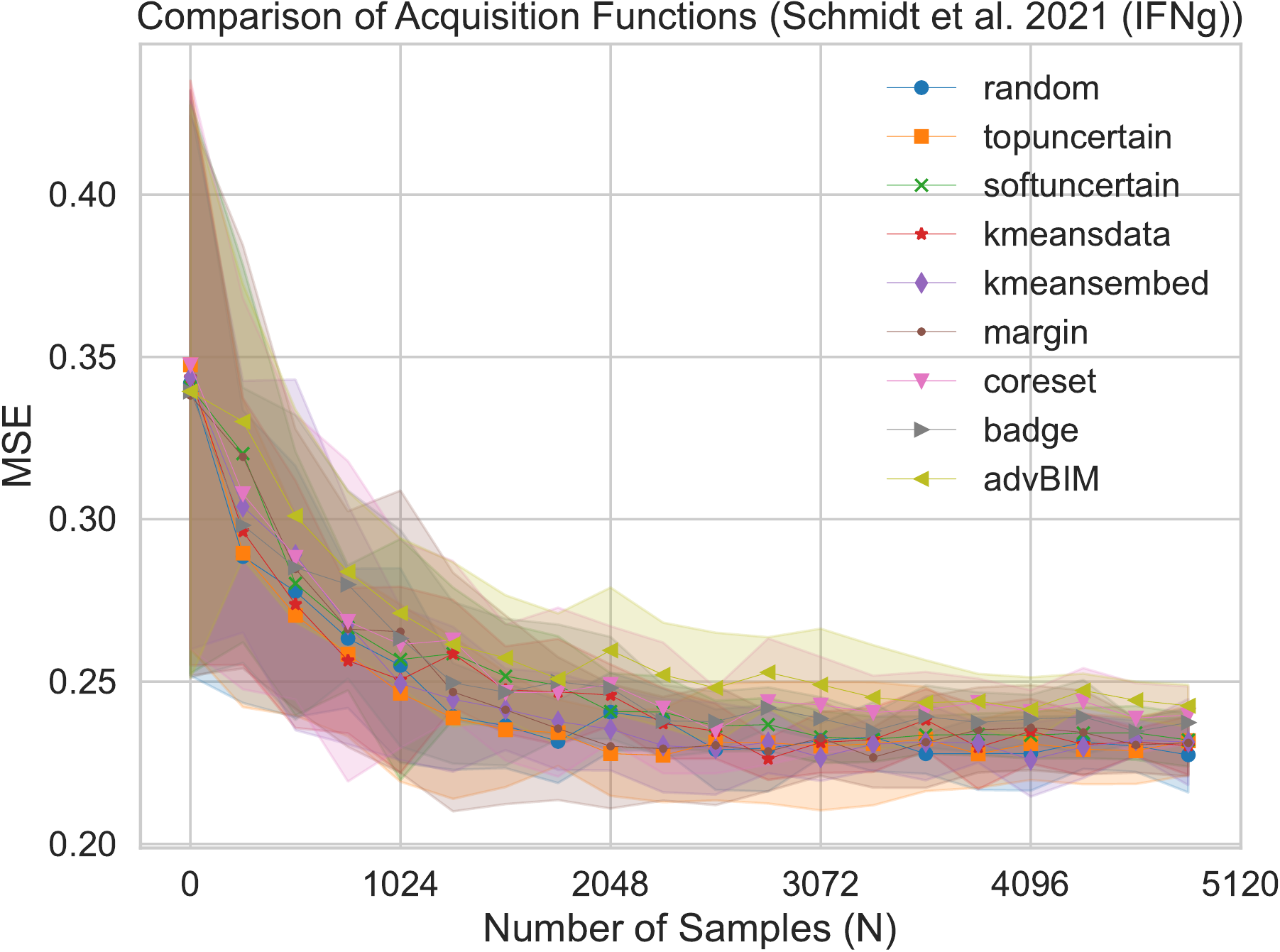}};
                        \end{tikzpicture}
                    }
                \end{subfigure}
                \&
                \begin{subfigure}{0.27\columnwidth}
                    \hspace{-28mm}
                    \centering
                    \resizebox{\linewidth}{!}{
                        \begin{tikzpicture}
                            \node (img)  {\includegraphics[width=\textwidth]{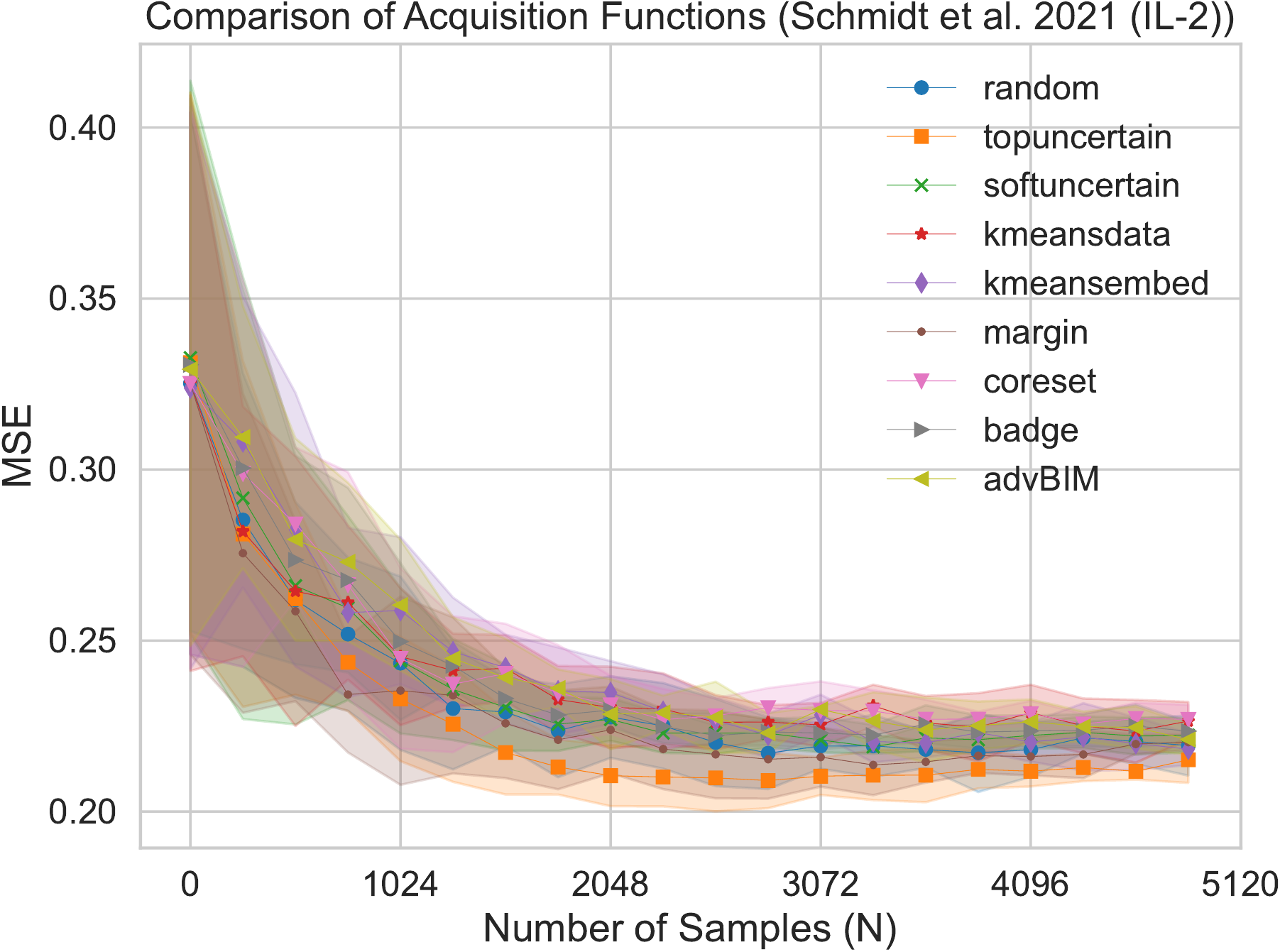}};
                        \end{tikzpicture}
                    }
                \end{subfigure}
                \&
                \begin{subfigure}{0.29\columnwidth}
                    \hspace{-32mm}
                    \centering
                    \resizebox{\linewidth}{!}{
                        \begin{tikzpicture}
                            \node (img)  {\includegraphics[width=\textwidth]{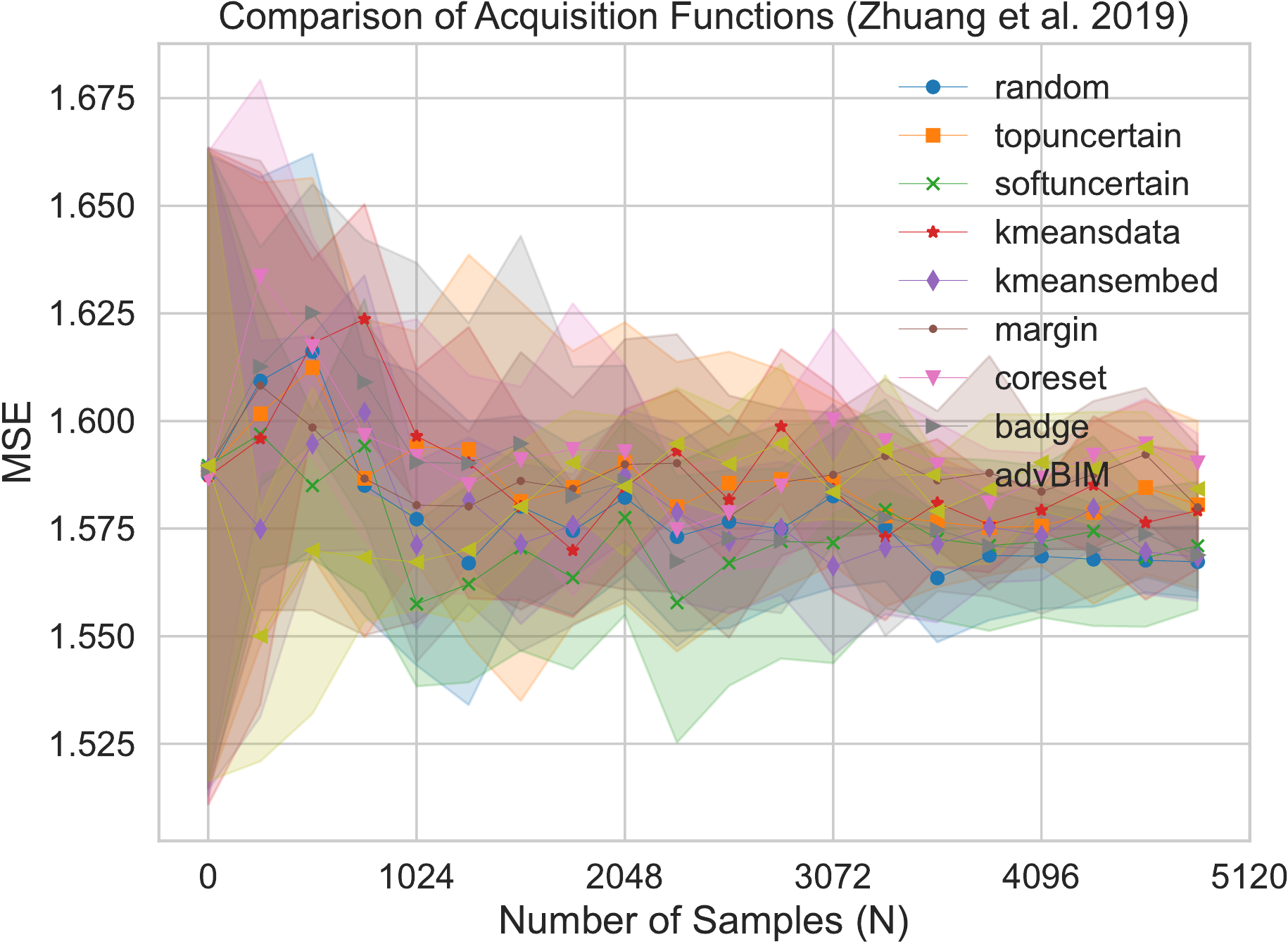}};
                        \end{tikzpicture}
                    }
                \end{subfigure}
                \&
                \\
            \\
           
            \\
            };
            \node [draw=none, rotate=90] at ([xshift=-8mm, yshift=2mm]fig-1-1.west) {\small batch size = 16};
            \node [draw=none, rotate=90] at ([xshift=-8mm, yshift=2mm]fig-2-1.west) {\small batch size = 64};
            \node [draw=none, rotate=90] at ([xshift=-8mm, yshift=2mm]fig-3-1.west) {\small batch size = 256};
            \node [draw=none] at ([xshift=-6mm, yshift=3mm]fig-1-1.north) {\small Shifrut et al. 2018};
            \node [draw=none] at ([xshift=-9mm, yshift=3mm]fig-1-2.north) {\small Schmidt et al. 2021 (IFNg)};
            \node [draw=none] at ([xshift=-11mm, yshift=3mm]fig-1-3.north) {\small Schmidt et al. 2021 (IL-2)};
            \node [draw=none] at ([xshift=-13mm, yshift=2.5mm]fig-1-4.north) {\small Zhuang et al. 2019};
\end{tikzpicture}}
        \vspace{-3em}
        \caption{The evaluation of the model trained with {STRING} treatment descriptors at each active learning cycle for 4 datasets and 3 acquisition batch sizes. In each plot, the x-axis is the active learning cycles multiplied by the acquisition bath size that gives the total number of data points collected so far. The y-axis is the test MSE error evaluated on the test data.}
        \vspace{-5mm}
        \label{fig:bnn_feat_string_alldatasets_allbathcsizes_small}
    \end{figure*} 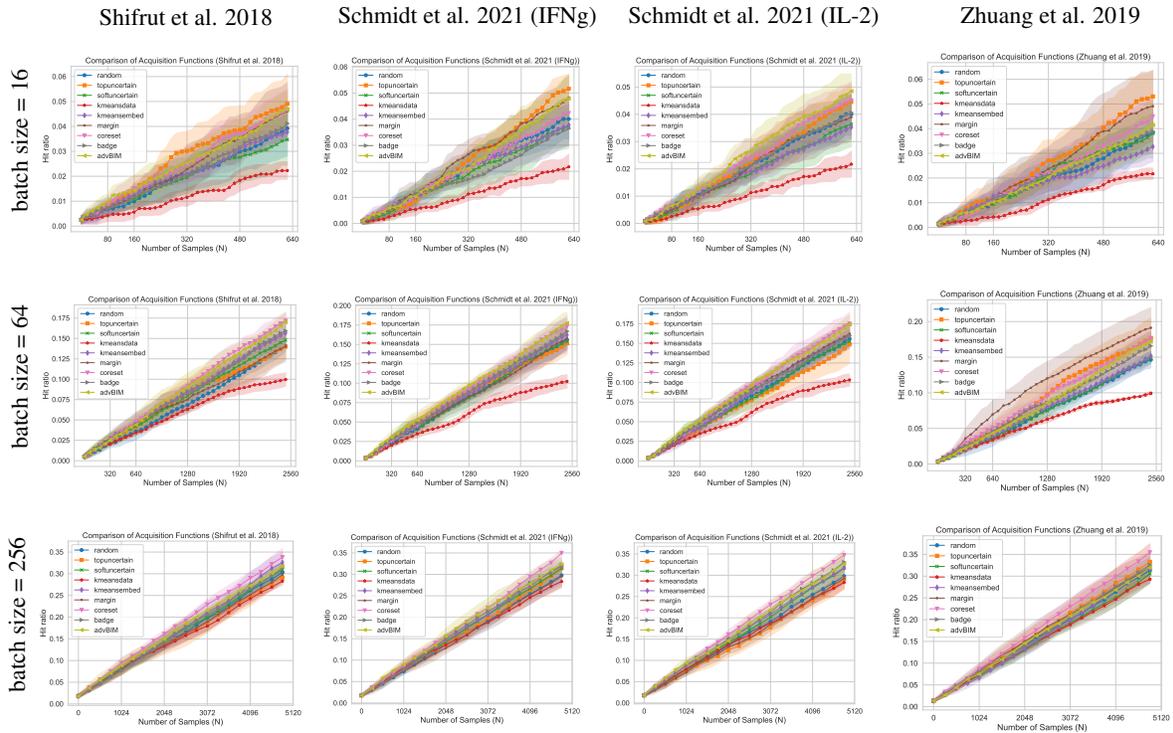
\begin{figure*}
    \vspace{-4em}
        \centering
        \makebox[0.72\paperwidth]{\begin{tikzpicture}[ampersand replacement=\&]
            \matrix (fig) [matrix of nodes]{ 
\begin{subfigure}{0.27\columnwidth}
                    \hspace{-17mm}
                    \centering
                    \resizebox{\linewidth}{!}{
                        \begin{tikzpicture}
                            \node (img)  {\includegraphics[width=\textwidth]{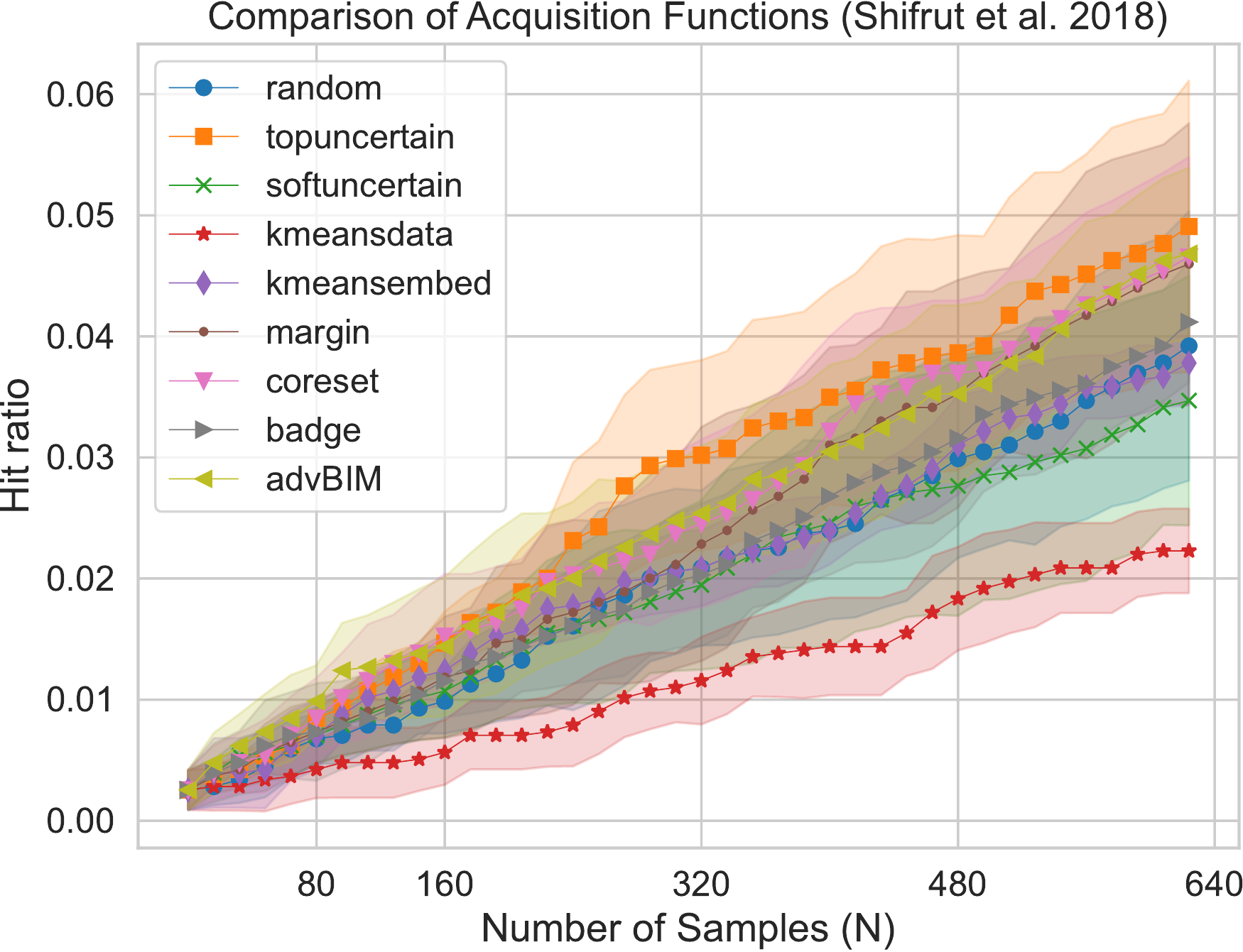}};
                        \end{tikzpicture}
                    }
                \end{subfigure}
                \&
                 \begin{subfigure}{0.27\columnwidth}
                    \hspace{-23mm}
                    \centering
                    \resizebox{\linewidth}{!}{
                        \begin{tikzpicture}
                            \node (img)  {\includegraphics[width=\textwidth]{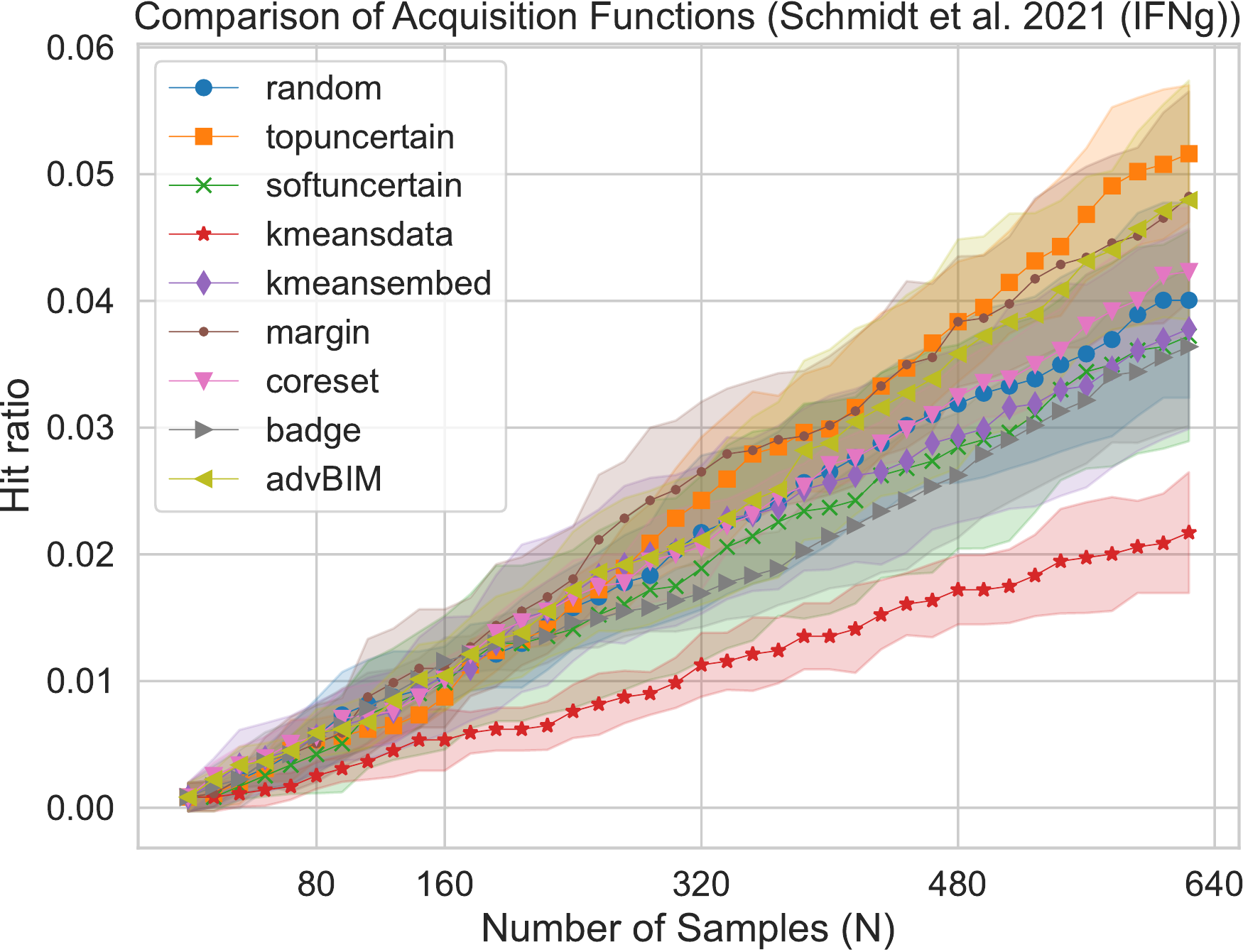}};
                        \end{tikzpicture}
                    }
                \end{subfigure}
                \&
                 \begin{subfigure}{0.27\columnwidth}
                    \hspace{-28mm}
                    \centering
                    \resizebox{\linewidth}{!}{
                        \begin{tikzpicture}
                            \node (img)  {\includegraphics[width=\textwidth]{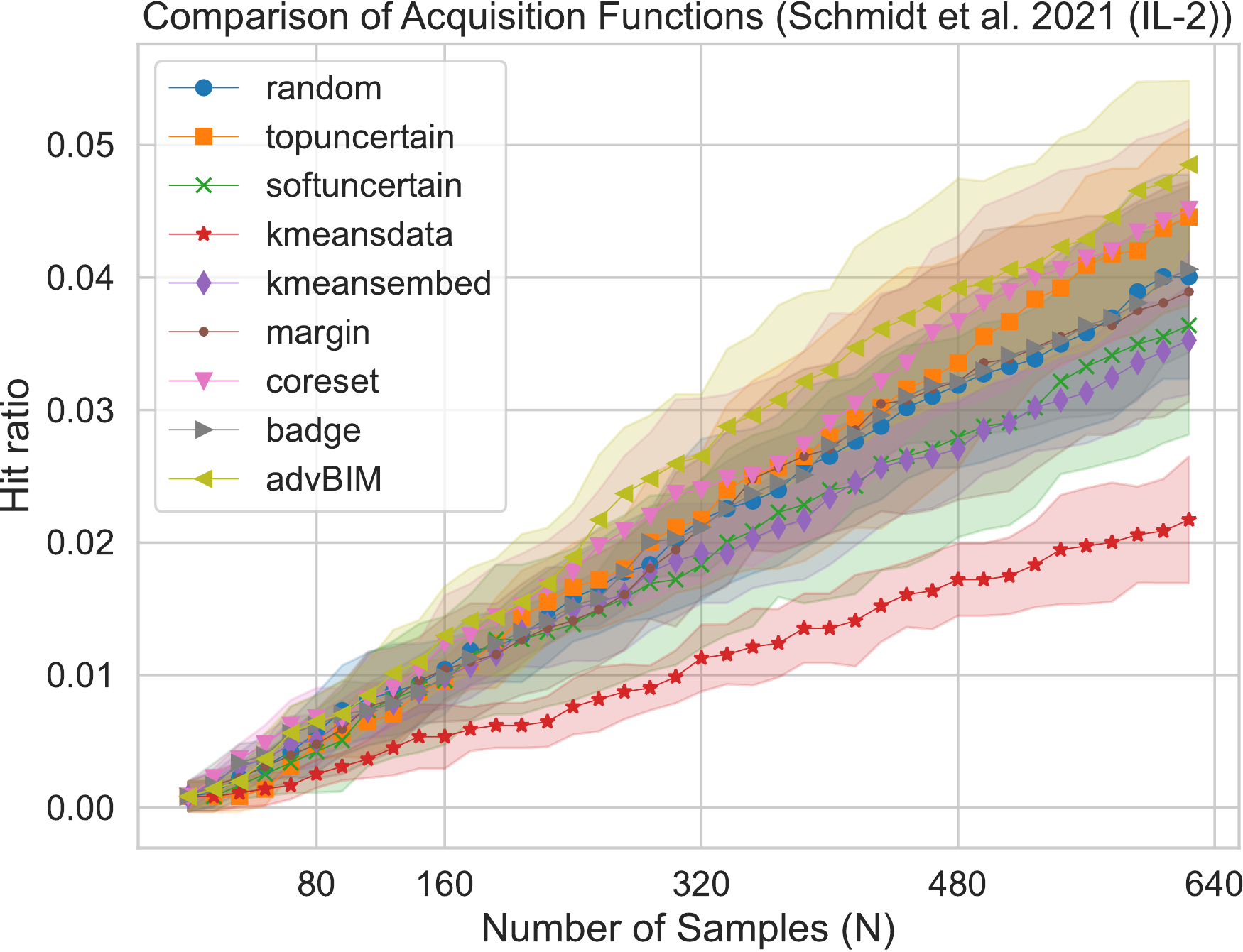}};
                        \end{tikzpicture}
                    }
                \end{subfigure}
                \&
                \begin{subfigure}{0.28\columnwidth}
                    \hspace{-32mm}
                    \centering
                    \resizebox{\linewidth}{!}{
                        \begin{tikzpicture}
                            \node (img)  {\includegraphics[width=\textwidth]{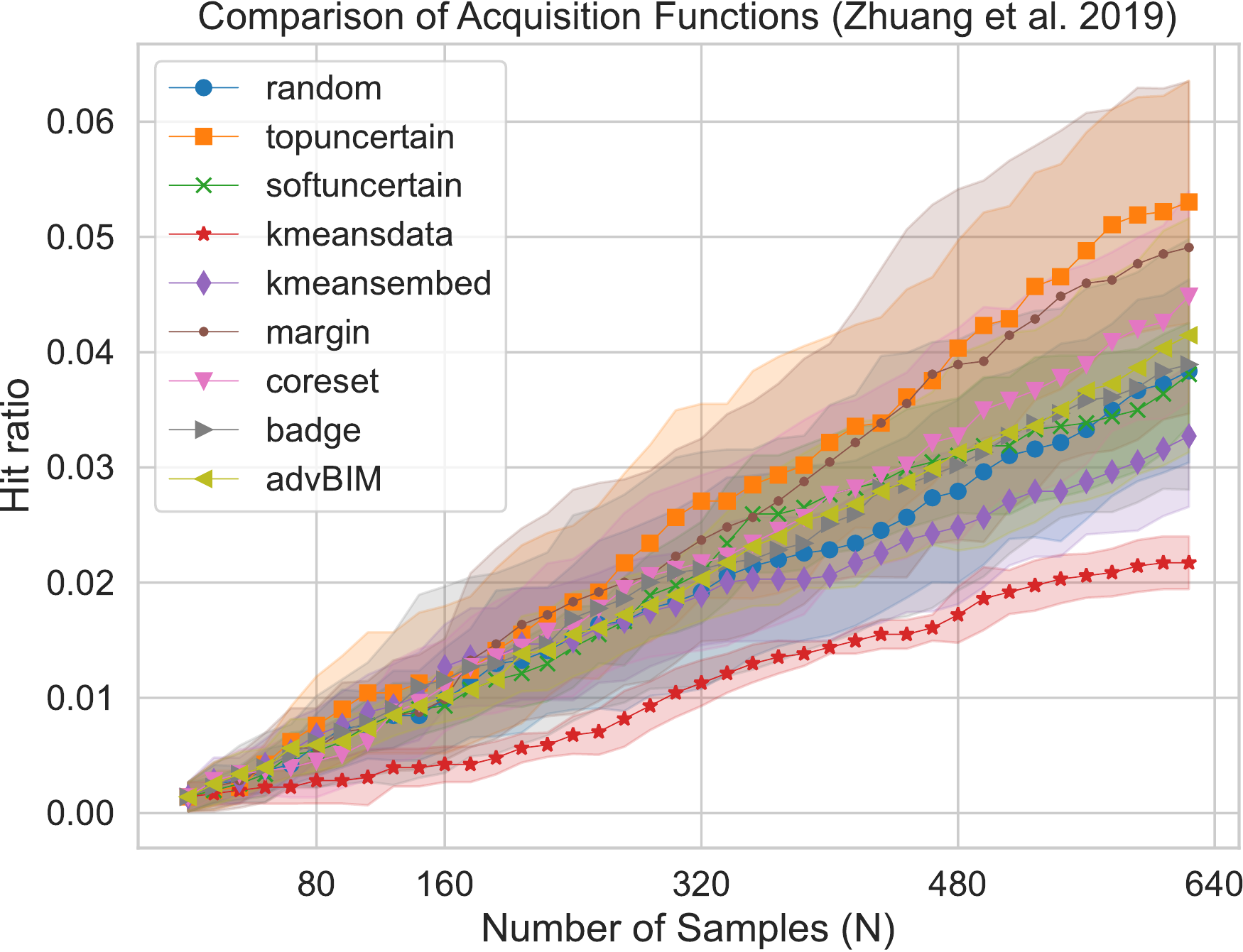}};
                        \end{tikzpicture}
                    }
                \end{subfigure}
                \&
            \\
\begin{subfigure}{0.27\columnwidth}
                    \hspace{-17mm}
                    \centering
                    \resizebox{\linewidth}{!}{
                        \begin{tikzpicture}
                            \node (img)  {\includegraphics[width=\textwidth]{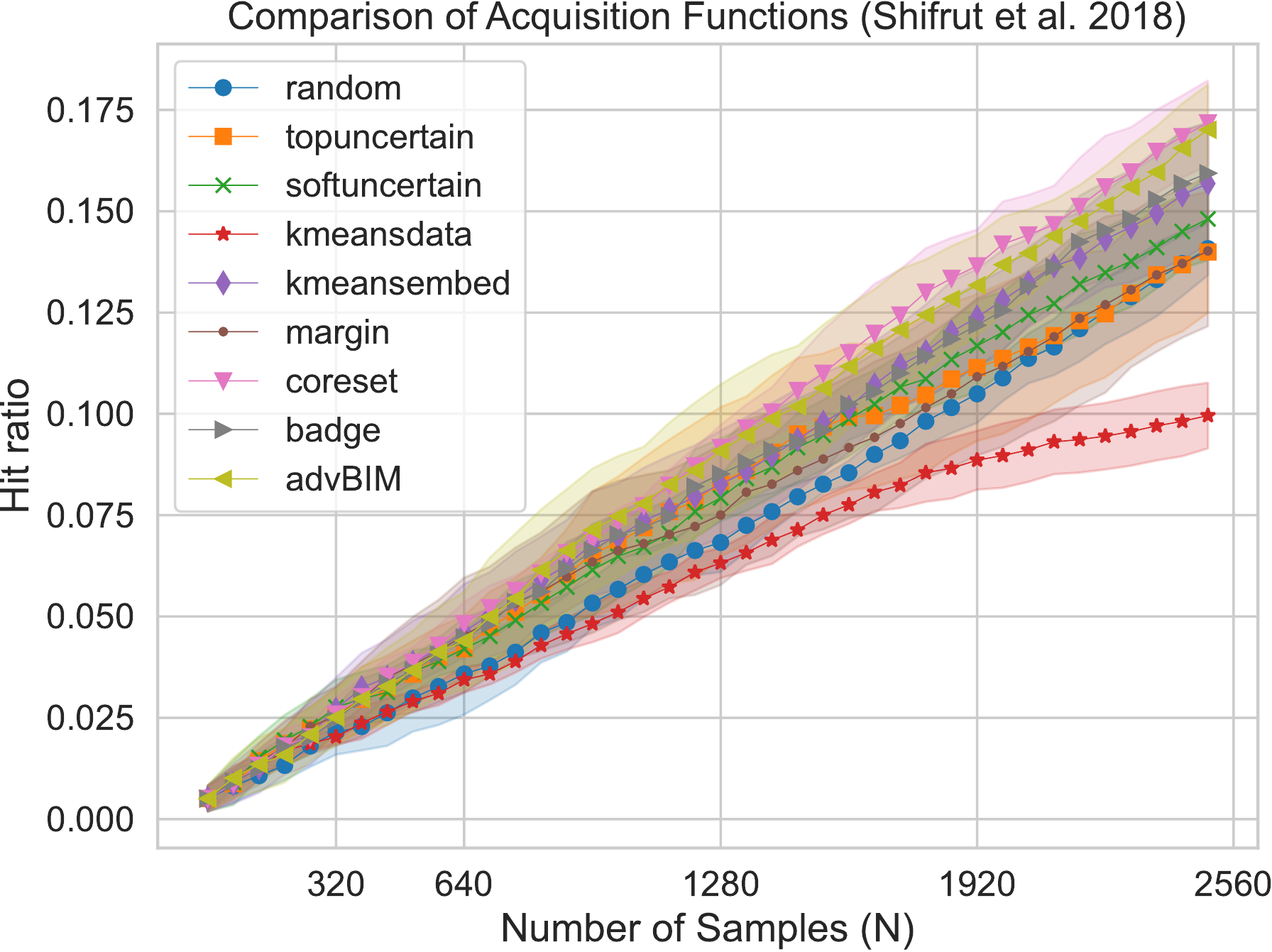}};
                        \end{tikzpicture}
                    }
                \end{subfigure}
                \&
                \begin{subfigure}{0.27\columnwidth}
                    \hspace{-23mm}
                    \centering
                    \resizebox{\linewidth}{!}{
                        \begin{tikzpicture}
                            \node (img)  {\includegraphics[width=\textwidth]{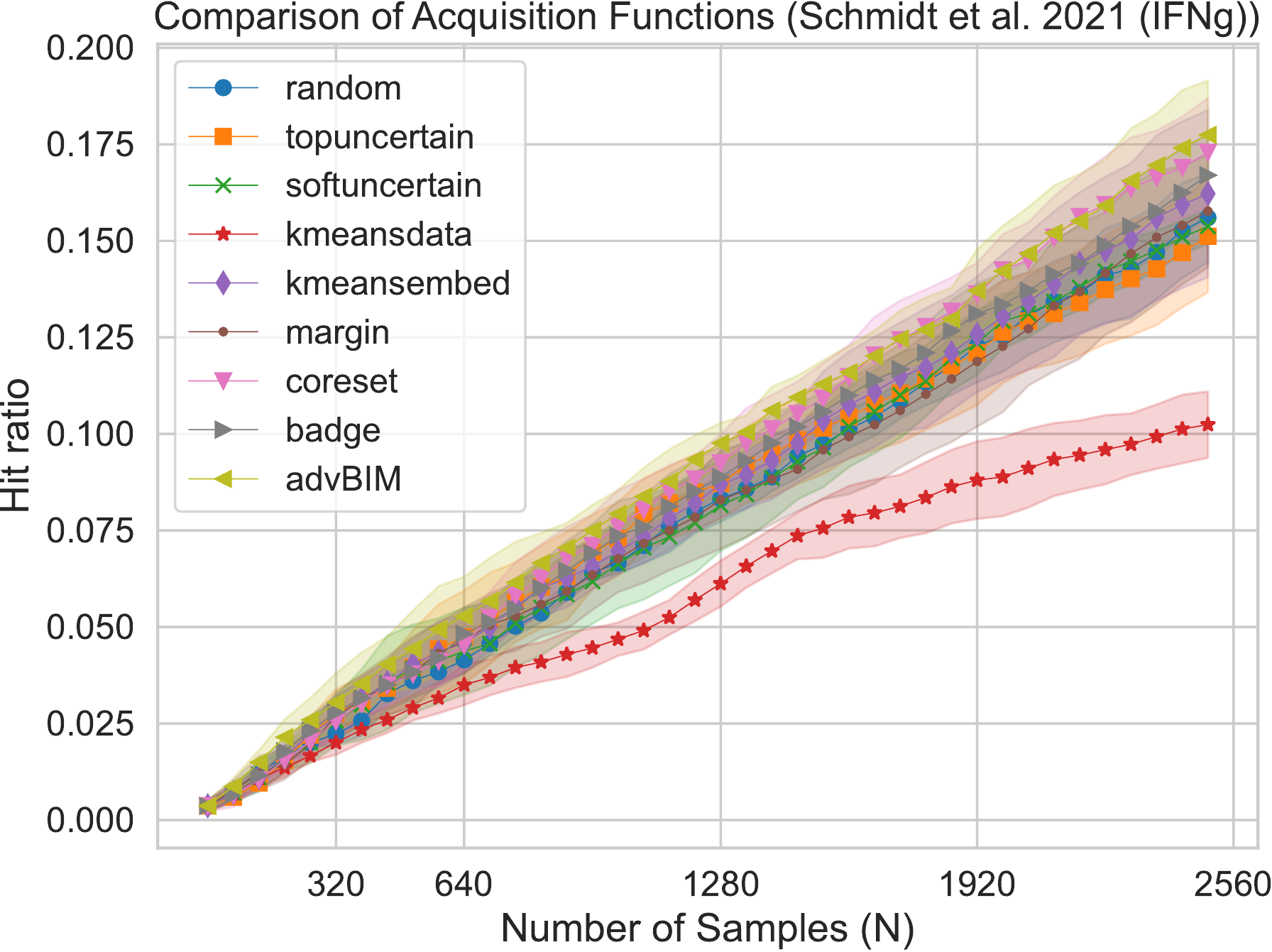}};
                        \end{tikzpicture}
                    }
                \end{subfigure}
                \&
                \begin{subfigure}{0.27\columnwidth}
                    \hspace{-28mm}
                    \centering
                    \resizebox{\linewidth}{!}{
                        \begin{tikzpicture}
                            \node (img)  {\includegraphics[width=\textwidth]{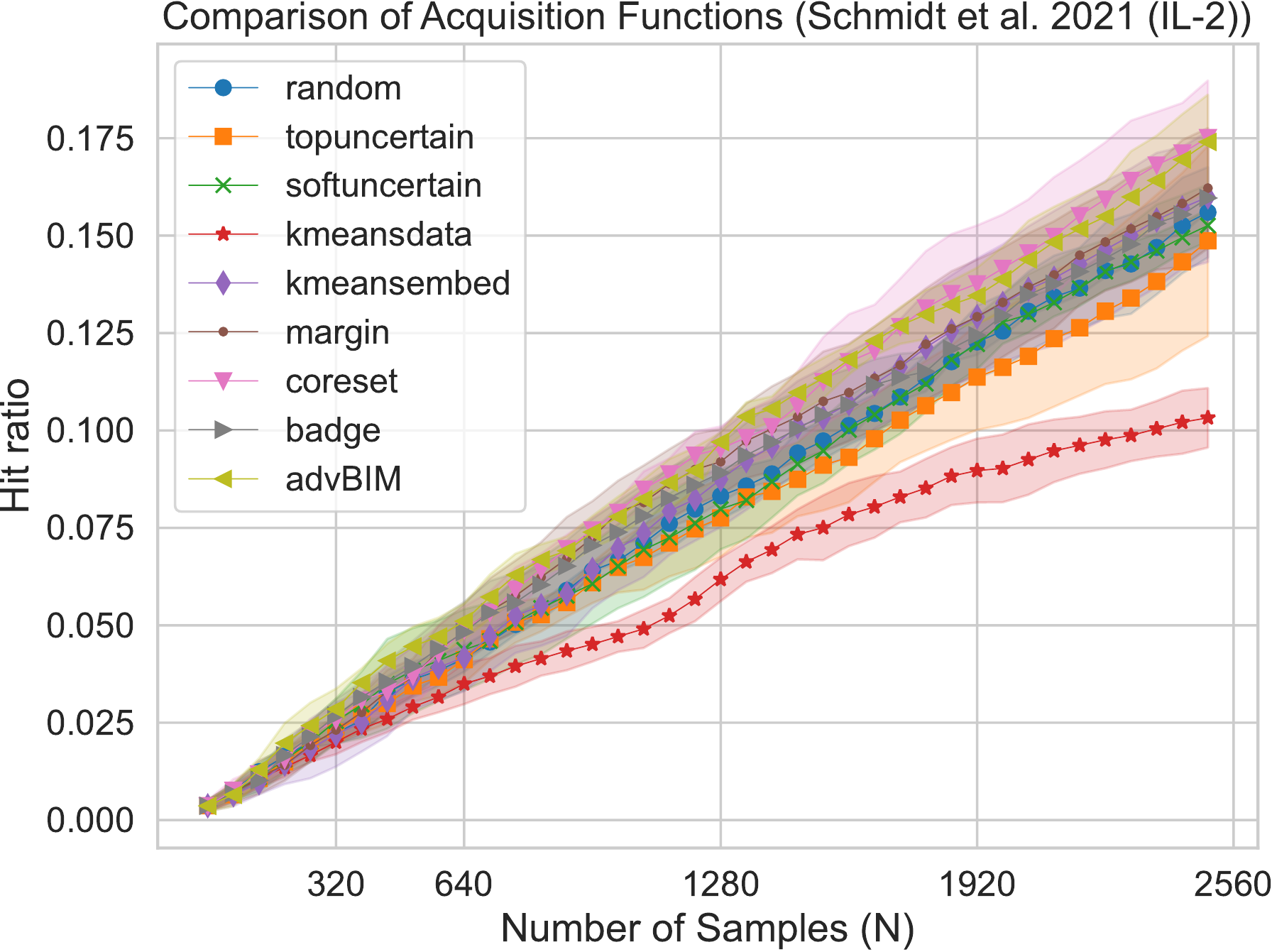}};
                        \end{tikzpicture}
                    }
                \end{subfigure}
                \&
                \begin{subfigure}{0.28\columnwidth}
                    \hspace{-32mm}
                    \centering
                    \resizebox{\linewidth}{!}{
                        \begin{tikzpicture}
                            \node (img)  {\includegraphics[width=\textwidth]{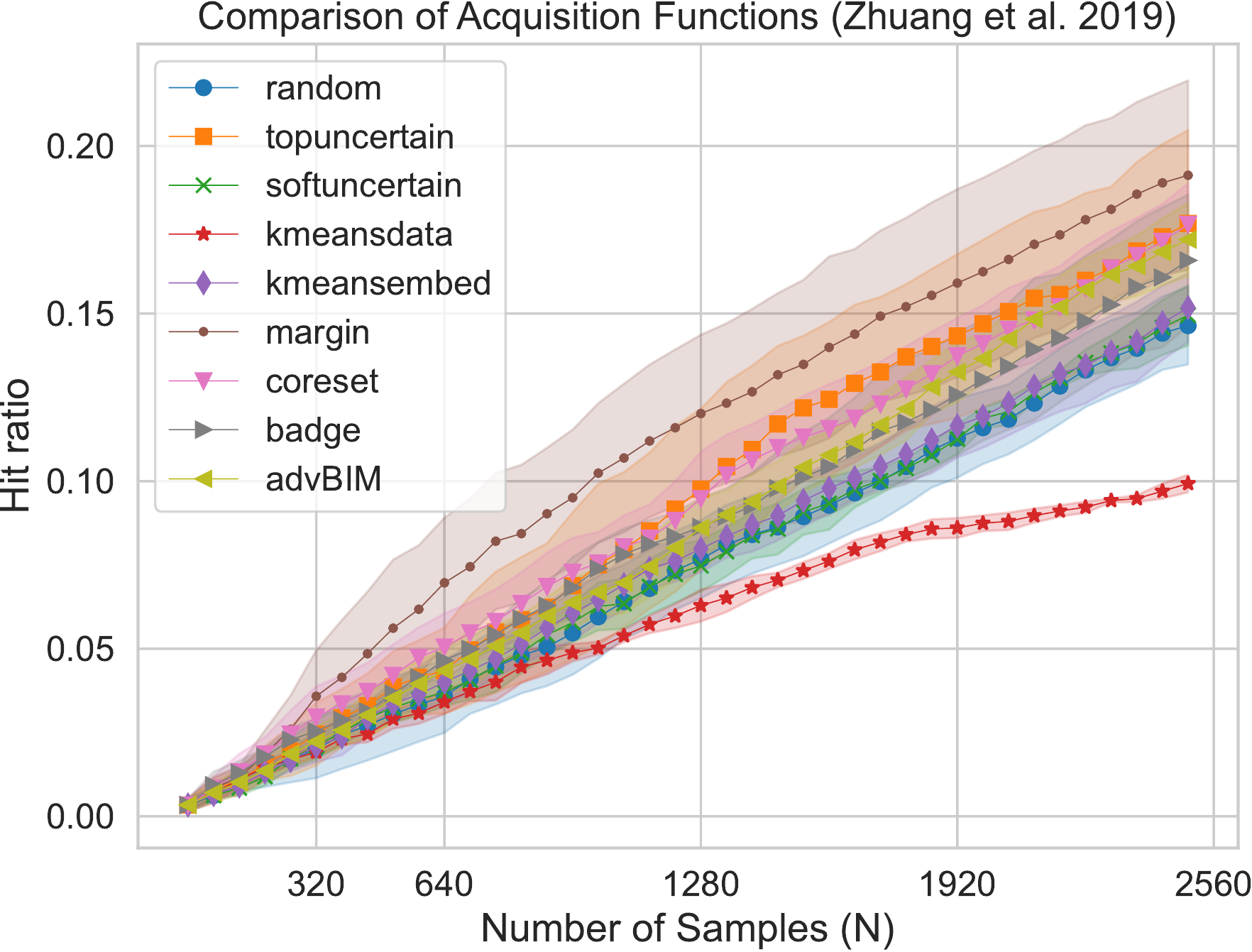}};
                        \end{tikzpicture}
                    }
                \end{subfigure}
                \&
                \\
\begin{subfigure}{0.275\columnwidth}
                    \hspace{-17mm}
                    \centering
                    \resizebox{\linewidth}{!}{
                        \begin{tikzpicture}
                            \node (img)  {\includegraphics[width=\textwidth]{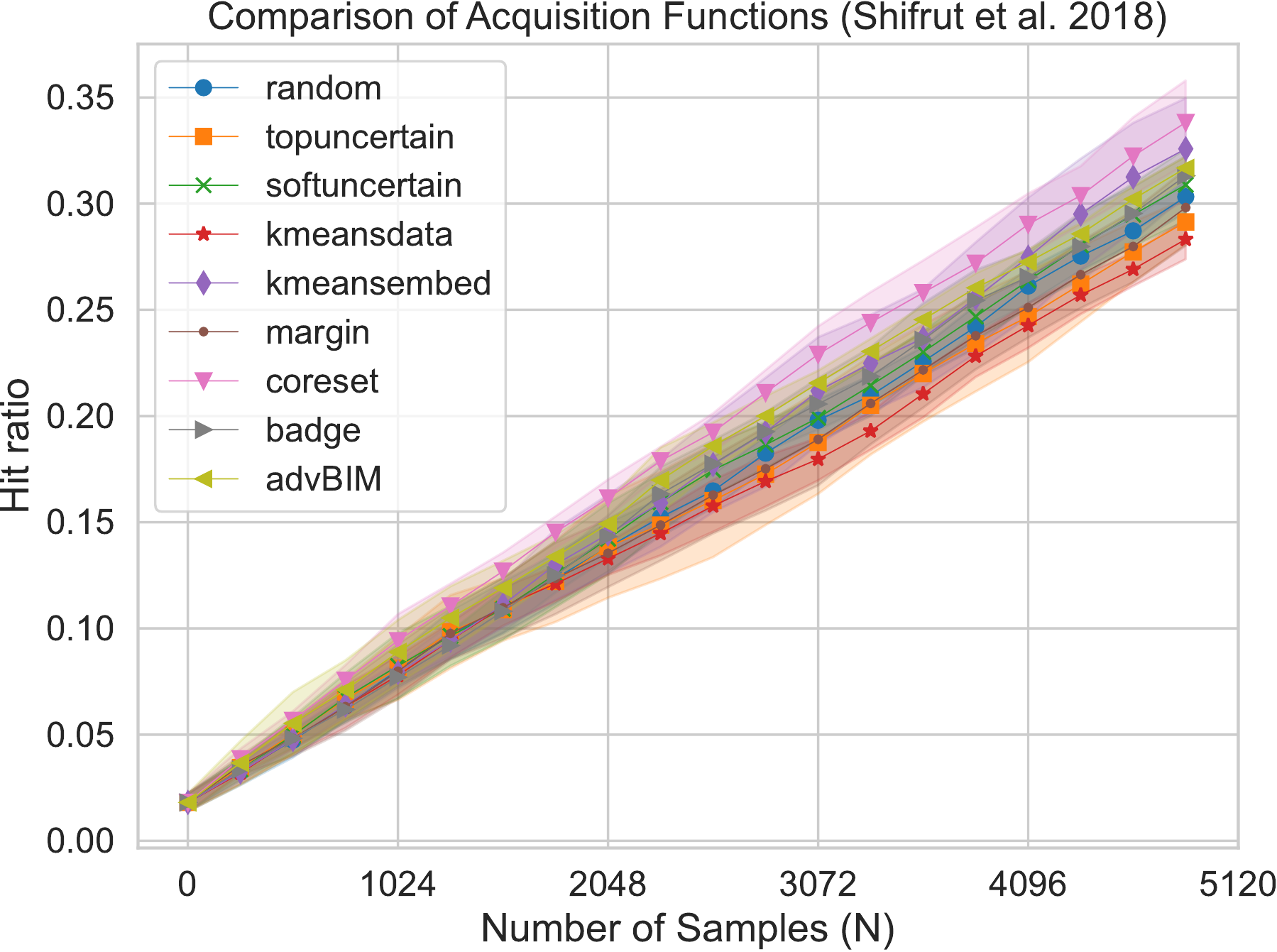}};
                        \end{tikzpicture}
                    }
                \end{subfigure}
                \&
                \begin{subfigure}{0.27\columnwidth}
                    \hspace{-23mm}
                    \centering
                    \resizebox{\linewidth}{!}{
                        \begin{tikzpicture}
                            \node (img)  {\includegraphics[width=\textwidth]{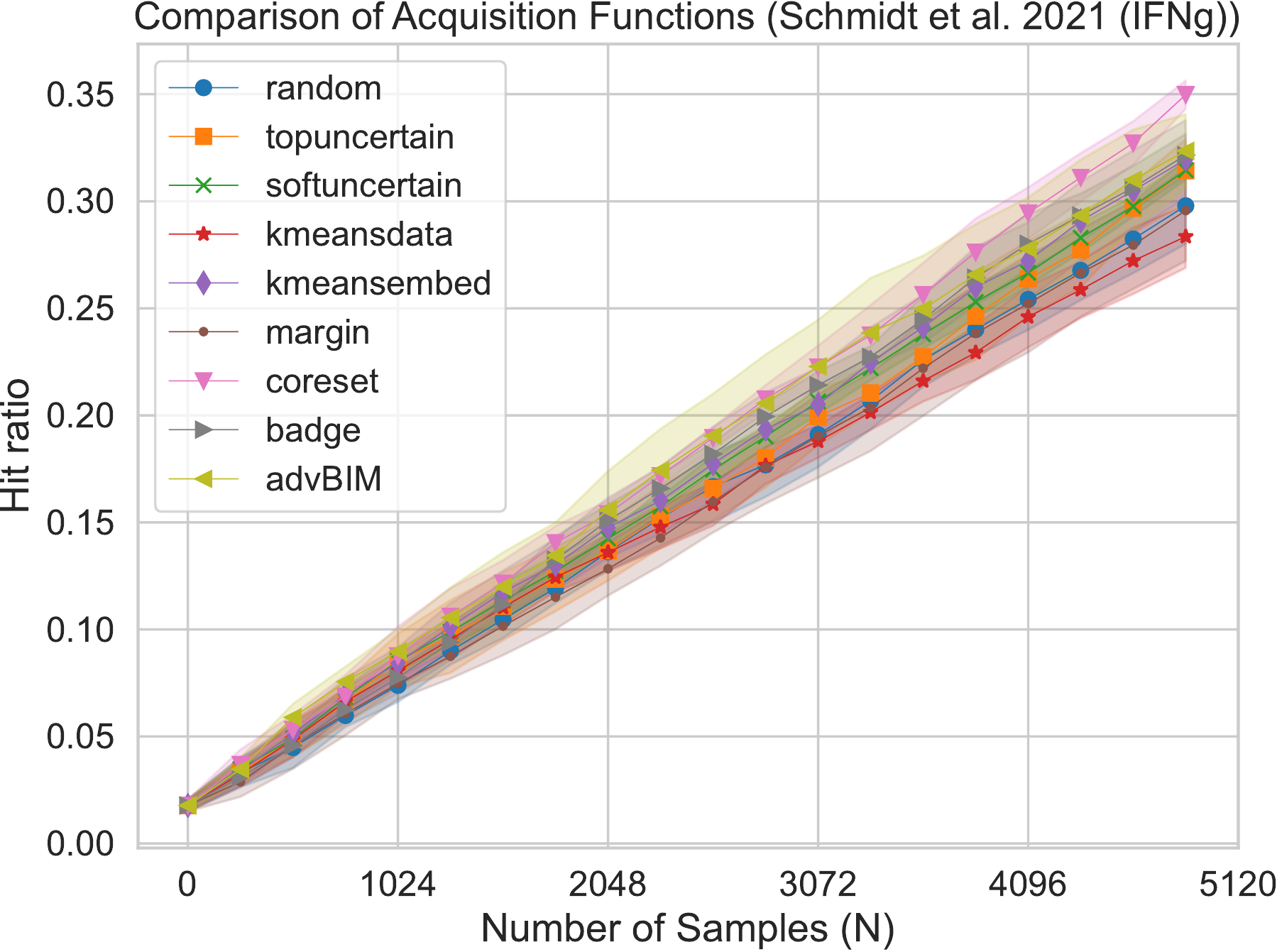}};
                        \end{tikzpicture}
                    }
                \end{subfigure}
                \&
                \begin{subfigure}{0.27\columnwidth}
                    \hspace{-28mm}
                    \centering
                    \resizebox{\linewidth}{!}{
                        \begin{tikzpicture}
                            \node (img)  {\includegraphics[width=\textwidth]{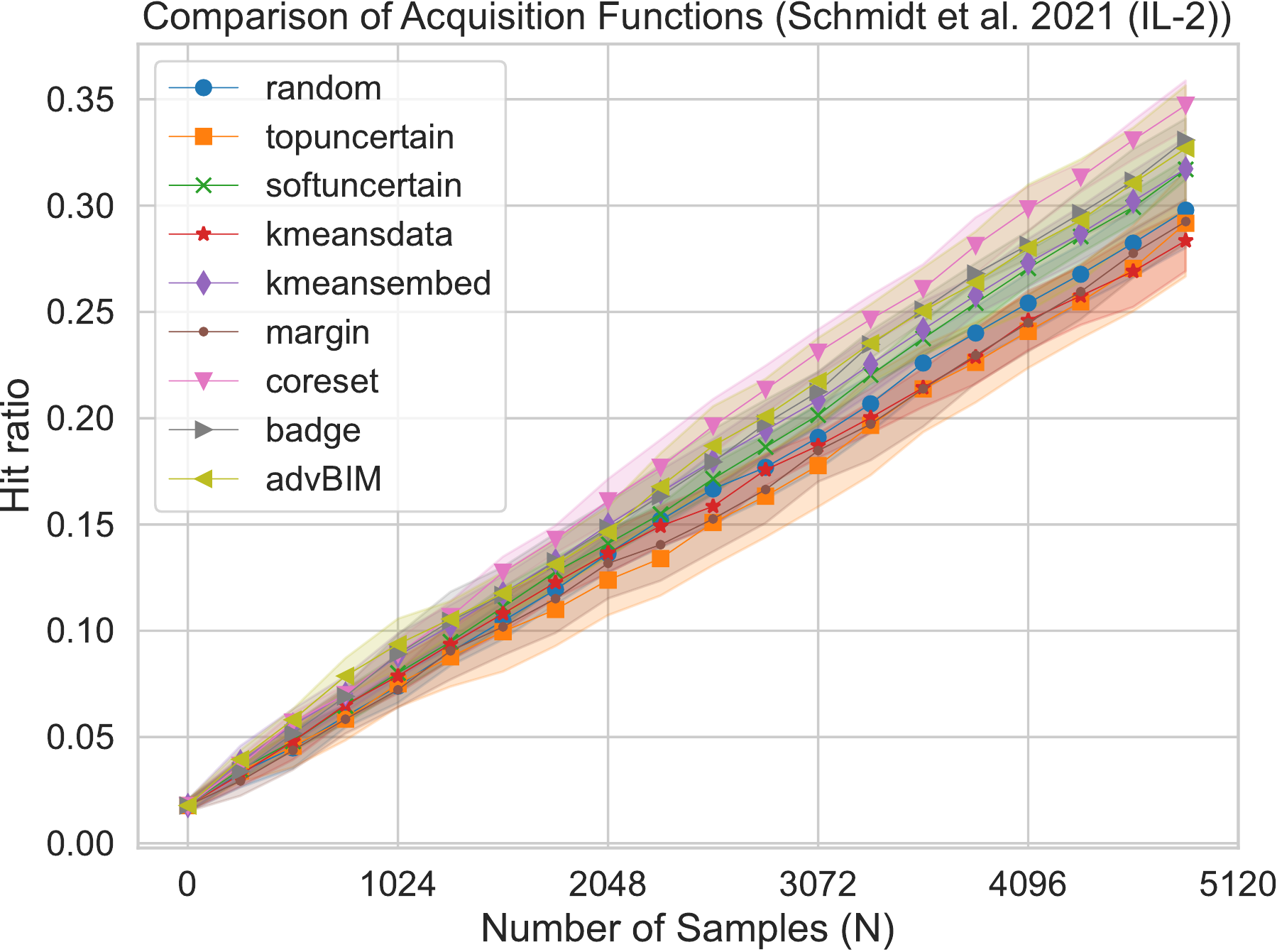}};
                        \end{tikzpicture}
                    }
                \end{subfigure}
                \&
                \begin{subfigure}{0.29\columnwidth}
                    \hspace{-32mm}
                    \centering
                    \resizebox{\linewidth}{!}{
                        \begin{tikzpicture}
                            \node (img)  {\includegraphics[width=\textwidth]{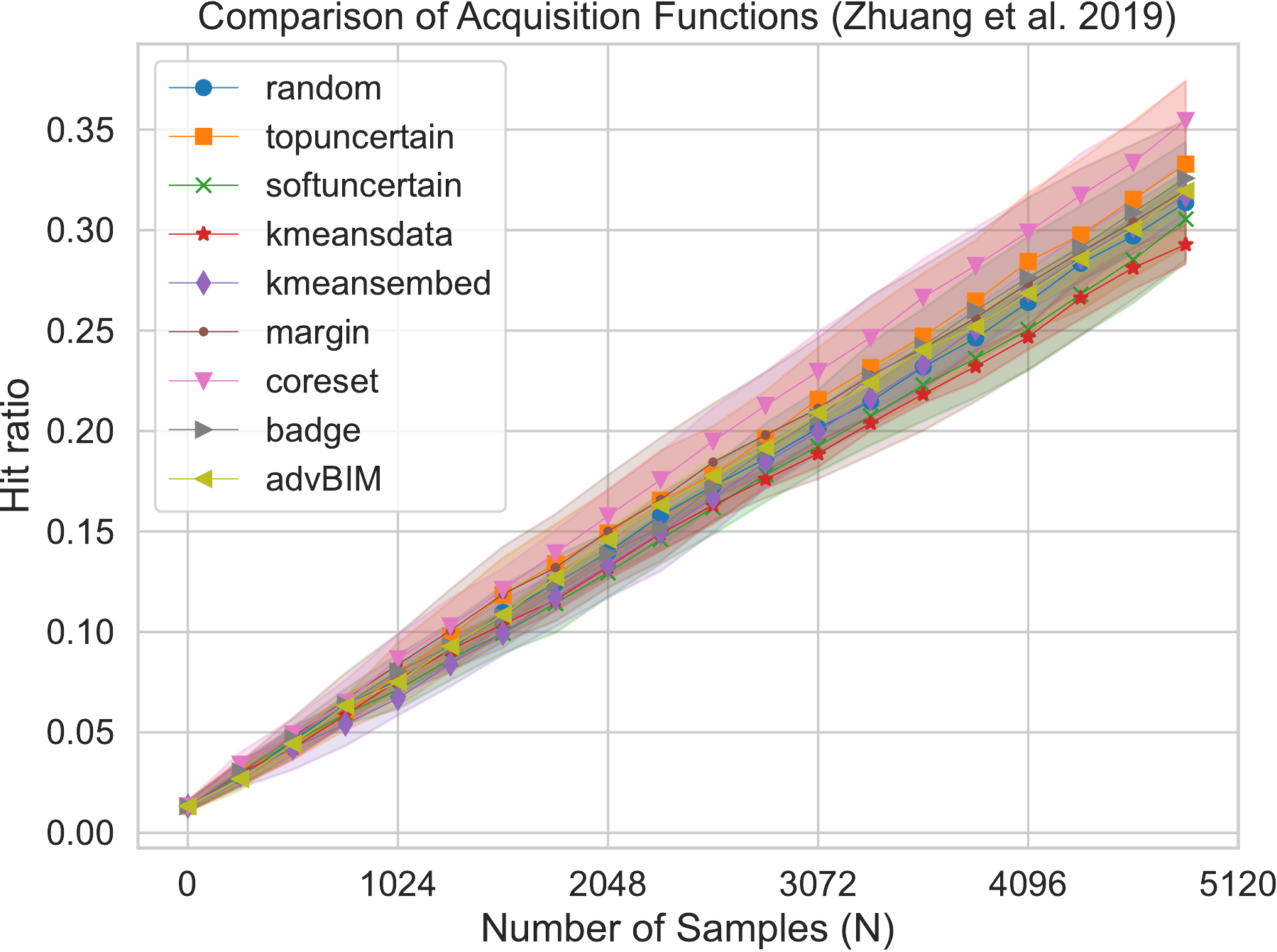}};
                        \end{tikzpicture}
                    }
                \end{subfigure}
                \&
                \\
            \\
           
            \\
            };
            \node [draw=none, rotate=90] at ([xshift=-8mm, yshift=2mm]fig-1-1.west) {\small batch size = 16};
            \node [draw=none, rotate=90] at ([xshift=-8mm, yshift=2mm]fig-2-1.west) {\small batch size = 64};
            \node [draw=none, rotate=90] at ([xshift=-8mm, yshift=2mm]fig-3-1.west) {\small batch size = 256};
            \node [draw=none] at ([xshift=-6mm, yshift=3mm]fig-1-1.north) {\small Shifrut et al. 2018};
            \node [draw=none] at ([xshift=-9mm, yshift=3mm]fig-1-2.north) {\small Schmidt et al. 2021 (IFNg)};
            \node [draw=none] at ([xshift=-11mm, yshift=3mm]fig-1-3.north) {\small Schmidt et al. 2021 (IL-2)};
            \node [draw=none] at ([xshift=-13mm, yshift=2.5mm]fig-1-4.north) {\small Zhuang et al. 2019};
\end{tikzpicture}}
        \vspace{-3em}
        \caption{The hit ratio of different acquisition for BNN model, different target datasets, and different acquisition batch sizes. We use {STRING} treatment descriptors here. The x-axis shows the number of data points collected so far during the active learning cycles. The y-axis shows the ratio of the set of interesting genes that have been found by the acquisition function up until each cycle.}
        \vspace{-5mm}
        \label{fig:hitratio_bnn_feat_string_alldatasets_allbathcsizes_small}
    \end{figure*} 
\textbf{Setup.} In order to assess current state-of-the-art methods on the \thebenchmark{} benchmark, we perform an extensive baseline evaluation of 9 acquisition functions, 6 acquisition batch sizes and 4 experimental assays using in excess of \numprint{20000} CPU hours of compute time. Due to the space limit, we include the results for 3 batch sizes in the main text and present the results for all batch sizes in the appendix.
The employed counterfactual estimator $\hat{g}$ is a multi-layer perceptron (MLP) that has one hidden layer with ReLU activation and a linear output layer. The size of the hidden layer is determined at each active learning cycle by k-fold cross validation against $20\%$ of the acquired batch. At each cycle, the model is trained for at most 100 epochs but early stopping may interrupt training earlier if the validation error does not decrease. Each experiment is repeated with 5 random seeds to assess experimental variance. To choose the number of active learning cycles, we use the following strategy: the number of cycles are bounded to 40 for the acquisition batches of sizes 16, 32 and 64 due to the computational limits. For larger batch sizes, the number of cycles are reduced proportionally so that the same total number of data points are acquired throughout the cycles. At each cycle, the model is trained from scratch using the data collected up to that cycle, i.e. a trained model is not transferred to the future cycles. 
The test data is a random $20\%$ subset of the whole data that is left aside before the active learning process initiates, and is kept fixed across all experimental settings (i.e., for different datasets and different batch sizes) to enable a consistent comparison of the various acquisition functions, counterfactual estimator and treatment descriptor configurations.

\textbf{Results.} The model performance based on the STRING treatment descriptors for different acquisition functions, acquisition batch sizes and datasets are presented in~\Cref{fig:bnn_feat_string_alldatasets_allbathcsizes_small}. The same metrics in the Achilles and CCLE treatment descriptors are provided in \Cref{sec:appendix:experiments}. To showcase the effect of the model class, we additionally repeat the experiments using a random forest model as an uncertainty aware ensemble model with a reduced set of acquisition functions that are compatible with non-differentiable models (\Cref{sec:appendix:experiments}). 
To investigate the types of genes chosen by different acquisition functions, we defined a subset of {potentially interesting} genes as the top $5\%$ with the largest absolute target value. These are the genes that could potentially be of therapeutic value due to their outsized causal influence on the phenotype of interest. The hit ratio out of the set of interesting genes chosen by different acquisition functions are presented in \Cref{fig:hitratio_bnn_feat_string_alldatasets_allbathcsizes_small} for the STRING treatment descriptors, and in \Cref{sec:appendix:experiments} for Achilles and CCLE. Benchmark results of interest include that model-independent acquisition methods using diversity heuristics (random, kmeansdata) perform relatively better in terms of model improvement than acquisition functions based on model uncertainty (e.g., topuncertain, softuncertain) when using lower batch acquisition sizes than in regimes with larger batch acquisition sizes potentially due to diversity being inherently higher in larger batch acquisition regimes due to the larger set of included interventions in an intervention space with a limited amount of similar interventions. Notably, while diversity-focused, model-independent acquisition functions, such as random and kmeansdata, perform well in terms of model performance, they underperform in terms of interesting hits discovered as a function of acquired interventional samples (\Cref{fig:hitratio_bnn_feat_string_alldatasets_allbathcsizes_small}). Based on these results, there appears to be a trade-off between model improvement and hit discovery in experimental exploration with counterfactual estimators that may warrant research into approaches to manage this trade-off to maximize long-term discovery rates.
\vspace{-1em}
\section{Discussion and Conclusion} 
\vspace{-0.75em}
The ranking of acquisition functions in \thebenchmark{} depends on several confounding factors, such as the choice of evaluation metric to compare different approaches, the characteristics of the dataset of interest, and the choice of the model class and its hyperparameters. An extrapolation of results obtained in \thebenchmark{} to new settings may not be possible under significantly different experimental conditions.
There is a subtle interplay between the predictive strength of the model and the acquisition function used to select the next set of interventions, as certain acquisition functions are more sensitive to the ability to the model to estimate its own epistemic uncertainty.
From a practical standpoint, \thebenchmark{} assumes the availability of a labeled set that is sufficiently representative to train and validate the different models required by the successive active learning cycles. However, model validation might be challenging when this set is small (e.g., during the early active learning cycles) or when the labelling process is noisy. Label noise is unfortunately common in interventional biological experiments, such as the ones considered in \thebenchmark{}. Experimental noise introduces additional trade-offs for consideration in experimental design not considered in \thebenchmark{}, such as choosing the optimal budget allocation between performing experiment replicates (technical and biological) to mitigate label noise or collecting more data points via additional active learning cycles.

\thebenchmark{} addresses the current lack of standardised benchmarks for developing batch active learning methods for experimental design in drug discovery. \thebenchmark{} consists of several curated datasets for experimental outcomes and genetic interventions, provides open source implementations of state-of-the-art acquisition functions for batch active learning, and includes a thorough assessment of these methods across a wide range of hyperparameter settings. We aim to attract the broader active learning community with an interest in causal inference by providing a robust and user-friendly benchmark that diversifies the benchmark repertoire over standard vision datasets. New models and acquisition functions for batch active learning in experimental design are of critical importance to realise the potential of machine learning for improving drug discovery. As future research, we aim to expand \thebenchmark{} to enable multi-modal learning and support simultaneous optimization across multiple output phenotypes of interest.

\newpage

\section*{Reproducibility Statement}

This work introduces a new curated and standardized benchmark, GeneDisco, for batch active learning for drug discovery. The benchmark includes four publicly available datasets, which have previously been published in a peer review process. 
Using a total of more than 20,000 central processing unit (CPU) hours of compute time, we perform an extensive evaluation of state-of-the-art acquisition functions for batch active learning on the GeneDisco benchmark, across a wide range of hyperparameters. 
To the best of our knowledge,  this  is  the  first  comprehensive survey and evaluation of active learning algorithms on real-world interventional genetic experiment data. Similar to developments in other fields e.g. for the learning of disentangled representations \citep{locatello2019challenging} or generative adversarial networks \citep{lucic2017gans}, we hope that our large scale experiments across a diverse set of real-world datasets provide an evidence basis to better understand the settings in which different active learning approaches work or do not work for drug discovery applications. 

All used models and acquisition functions are described in detail and referenced in \Cref{Section 4: Acquisition functions}
and \Cref{Section 4: Models}. For all introduced datasets, we include a detailed description and the details on the train, test and validation splits at the beginning of  \Cref{Section 5 - Experiments}. 

For all experimental results we report the range of hyper-parameters considered and the methods of selecting hyper-parameters as well as the exact number of training and evaluation runs (\Cref{sec:appendix:experiments}). We additionally provide error bars over multiple random seeds and the code was executed on a cloud cluster with Intel CPUs. We provide detailed results for all investigated settings in the appendix (\ref{sec:appendix:experiments}).

\bibliographystyle{iclr2022_conference}

\clearpage

\newpage
\appendix

\section{Notations}
\label{sec:notation}

Here is the list of notations used in this manuscript.

\begin{itemize}
    \item $\Dcal_\textrm{pool}$: The pool of unlabeled data.
    \item $\Dcal_\textrm{acq}^k$: Acquired data at $k$-th AL cycle.
    \item $\Dcal_\textrm{train}^k$: Cumulative training data after $k$-th cycle of AL.
    \item $\Dcal_\textrm{val}$: Validation data.
    \item $\Dcal_\textrm{test}$: Held-out test data.
    \item $b$: Acquisition batch size.
    \item $K$: Total number of AL cycles.
    \item $k=[1,2,\ldots, K]$: Index of the AL cycle.
    \item $[K]=[1,2,\ldots, K]$.
    \item $e\in \Ecal$: Inherent noise (aleatoric uncertainty).
    \item $T\in \Tcal$: Treatment variable.
    \item $\mathbb{E}[Y \mid X=x, do(T=t)]$: The conditional expected outcomes.
    \item $\hat{g}(t; \omega)$: The model parameterised by $\omega\in\Omega$ to estimate $\mathbb{E}[Y \mid X=x, do(T=t)]$.
    \item $X$: random variable $X$ with distribution $X\sim F(X)$ and density $f(x)$.
\end{itemize}

\section{Acquisition Functions cont.}
\label{sec:acq_fun_continued}

\paragraph{Coreset.} Coreset acquisition looks to maximize the diversity of acquired samples. 
This is done by finding the data points in $\Dcal^k_\textrm{avail}$ that are furthest from the labelled data points in $\Dcal^{k-1}_\textrm{cum}$. 
The robust K-centers algorithm of \citet{sener2017active} approximates a solution to: 
\begin{equation}
\small
    \alpha_{\textrm{CORESET}}(\widehat{g}^{k-1}(t), \Dcal^k_\textrm{avail}) = \argmin_{\{t_1, \dots t_b\} \in \Dcal^k_\textrm{avail}} \argmax_{t_i \in \Dcal^k_\textrm{avail}} \argmin_{t_j \in \Dcal^k_\textrm{avail} \cup \Dcal^{k-1}_\textrm{cum}} \Delta(t_i, t_j).
\end{equation}
Euclidean distances, $\Delta(t_i, t_j)$, are calculated between the output of the penultimate layer of $\widehat{g}(t; \omega)$.

\paragraph{Margin Sample.} Margin sampling is designed for classifiers where selection is based on the distance of a sample from the classifiers decision boundary \citep{roth2006margin}. 
As a proxy, the difference between the predicted probability of the most and second most probable classes is used. 
The distance between the most probable and the second most probable classes for a multi-class classification problem can be seen as how confident the model is about the label of that class. 
However, The concept of a decision boundary is ill-defined for regression tasks. 
One option to approximate margin sampling could be to model the aleatoric uncertainty of the model by predicting the conditional variance of the outcome $\sigma^2(t; \omega)$ and select data based on the magnitude of this value.
Here, we instead look at the difference in the maximum and minimum values of the predicted outcome as a measure of the model's confidence and select data based on the magnitude of this value. Formally, we have
\begin{equation}
    \widehat{\mathcal{M}}(Y ; \Omega \mid t_i, \Dcal^{k-1}_\textrm{cum}) = \max_{j \in \{1, \dots m\}}(\widehat{g}(t; \omega_j^{k-1})) - \min_{j \in \{1, \dots m\}}(\widehat{g}(t; \omega_j^{k-1})),
\end{equation}
and the acquisition function:
\begin{equation}
    \alpha_{\textrm{Margin}}(\widehat{g}^{k-1}(t), \Dcal^k_\textrm{avail}) = \argmax_{\{t_1, \dots t_b\} \in \Dcal^k_\textrm{avail}} \sum_{i=1}^b \widehat{\mathcal{M}}(Y ; \Omega \mid t_i, \Dcal^{k-1}_\textrm{cum}).
\end{equation}
Note that this approximation is similar to BALD under the assumption of a uniformly distributed outcome: $f(y \mid t, \omega) = \mathcal{U}(y \mid \widehat{g}(t; \omega))$.

\paragraph{Adversarial Basic Iteractive Method (AdvBIM).} Some of the adversarial algorithms can act as active learning acquisition functions by nominating the adversarial samples. Here, we extended the famous Adversarial BIM method for our regression task as an example.
BIM was introduced by \citep{kurakin2016adversarial} to iteratively perturb adversarial samples to maximize the cost function $J$ subject to an $l_p$ norm constraint as
\begin{eqnarray}
	\hat{t}^{(0)} = t,
	\hat{t}^{(i)} = \text{clip}_{t,e} (\hat{t}^{(i-1)} + \text{sign}(\nabla_{\hat{t}^{(i-1)}} J(\theta, \hat{t}^{(i-1)} , y)))
\end{eqnarray}
(intermediate results are clipped to stay in $e$-neighbourhood of the primary data point $t$). This technique bypasses the intractable problem of finding the distance from the decision boundary by iteratively perturbing the features until crossing the boundary \citep{tramer2017ensemble}.
In our regression task, we perturb the features in the gradients' direction to increase the conditional variance of the outcome, i.e., 
\begin{eqnarray}
\hat{t}^{(0)} = t,
\hat{t}^{(i)} = \text{clip}_{t,e} (\hat{t}^{(i-1)} + \text{sign}(\nabla_{\hat{t}^{(i-1)}} \textrm{Var}_{\omega}(\hat{g}(t, \omega)))) \text{ for $i = \{1, \dots, m\}$},
\end{eqnarray}
where $||\hat{t}_i - t||_2 < \gamma * ||t||_2$ with the hyperparameter $\gamma$. After creating adversarial samples for each data point in $D_\textrm{avail}^{k}$, 
$\alpha_{\textrm{AdversarialBIM}}$ acquires the samples by
\begin{equation}
\small
\alpha_{\textrm{AdversarialBIM}}(\widehat{g}^{k-1}(t), \Dcal^k_\textrm{avail}) = \bigcup_{t_i\in \Dcal^k_\textrm{avail}} \argmin_{t_j\in \Dcal^k_\textrm{avail}} \Delta(\hat{t}^{(m)}_i, t_j),
\end{equation} 
where $\Delta$ is the euclidean distance.

\paragraph{$k$-means Sampling.}
This method nominates samples by returning the closest sample to each center of the unlabeled data clusters. In order to do so, one may run Kmeans++ clustering algorithm with the number of clusters equal to $b$ over either the unlabeled data points $D_\textrm{avail}^{k}$ or the output of the penultimate layer of $\widehat{g}(t; \omega)$. We refer to the former as $\texttt{kmeansdata}$ and to the latter as $\texttt{kmeansembed}$ in the experiments. Assuming $\{\mu_1, \dots, \mu_b\}$ are the centers of the clustering, we have
\begin{equation}
\small
\alpha_{\textrm{Kmeans}}(\widehat{g}^{k-1}(t), \Dcal^k_\textrm{avail}) = \bigcup_{i = 1}^{b} \argmin_{t_j\in \Dcal^k_\textrm{avail}} \Delta(\mu_i, t_j),
\end{equation}
where $\Delta$ is euclidean distance over the data points or the penultimate layer of $\widehat{g}(t; \omega)$.

\section{Detailed experimental results}
\label{sec:appendix:experiments}

\subsection{Bayesian Neural Network (BNN) Model}
\label{sec:appendix_experiments_bnn}
\begin{figure*}
    \vspace{-2mm}
        \centering
        \makebox[0.72\paperwidth]{\begin{tikzpicture}[ampersand replacement=\&]
            \matrix (fig) [matrix of nodes]{ 
\begin{subfigure}{0.27\columnwidth}
                    \hspace{-17mm}
                    \centering
                    \resizebox{\linewidth}{!}{
                        \begin{tikzpicture}
                            \node (img)  {\includegraphics[width=\textwidth]{figs/bnnplots/data_shifrut_2018_feat_string_bs16.pdf}};
                        \end{tikzpicture}
                    }
                \end{subfigure}
                \&
                 \begin{subfigure}{0.27\columnwidth}
                    \hspace{-23mm}
                    \centering
                    \resizebox{\linewidth}{!}{
                        \begin{tikzpicture}
                            \node (img)  {\includegraphics[width=\textwidth]{figs/bnnplots/data_schmidt_2021_ifng_feat_string_bs16.pdf}};
                        \end{tikzpicture}
                    }
                \end{subfigure}
                \&
                 \begin{subfigure}{0.27\columnwidth}
                    \hspace{-28mm}
                    \centering
                    \resizebox{\linewidth}{!}{
                        \begin{tikzpicture}
                            \node (img)  {\includegraphics[width=\textwidth]{figs/bnnplots/data_schmidt_2021_il2_feat_string_bs16.pdf}};
                        \end{tikzpicture}
                    }
                \end{subfigure}
                \&
                \begin{subfigure}{0.28\columnwidth}
                    \hspace{-32mm}
                    \centering
                    \resizebox{\linewidth}{!}{
                        \begin{tikzpicture}
                            \node (img)  {\includegraphics[width=\textwidth]{figs/bnnplots/data_zhuang_2019_nk_feat_string_bs16.pdf}};
                        \end{tikzpicture}
                    }
                \end{subfigure}
                \&
            \\
\begin{subfigure}{0.27\columnwidth}
                    \hspace{-17mm}
                    \centering
                    \resizebox{\linewidth}{!}{
                        \begin{tikzpicture}
                            \node (img)  {\includegraphics[width=\textwidth]{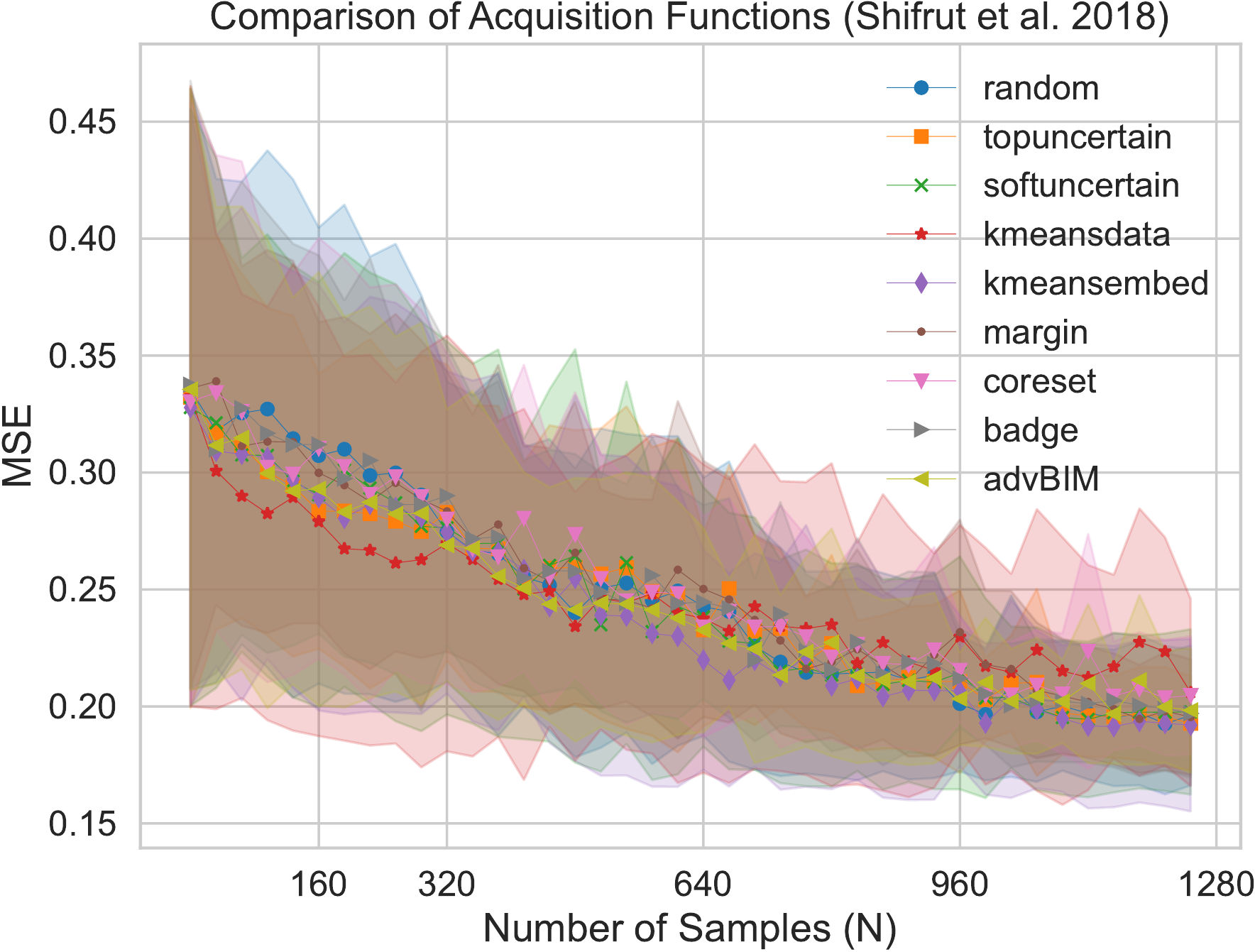}};
                        \end{tikzpicture}
                    }
                \end{subfigure}
                \&
                \begin{subfigure}{0.27\columnwidth}
                    \hspace{-23mm}
                    \centering
                    \resizebox{\linewidth}{!}{
                        \begin{tikzpicture}
                            \node (img)  {\includegraphics[width=\textwidth]{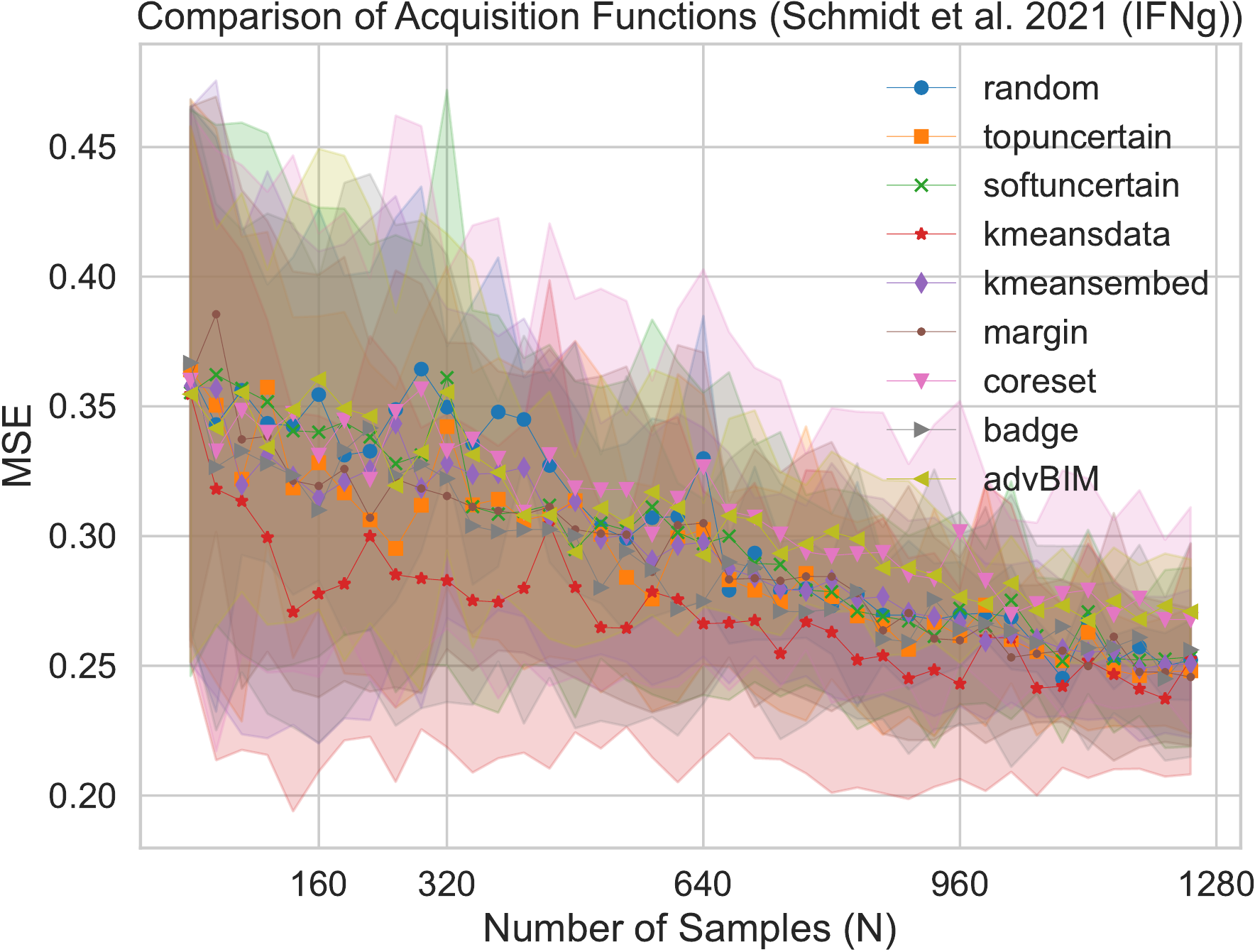}};
                        \end{tikzpicture}
                    }
                \end{subfigure}
                \&
                \begin{subfigure}{0.27\columnwidth}
                    \hspace{-28mm}
                    \centering
                    \resizebox{\linewidth}{!}{
                        \begin{tikzpicture}
                            \node (img)  {\includegraphics[width=\textwidth]{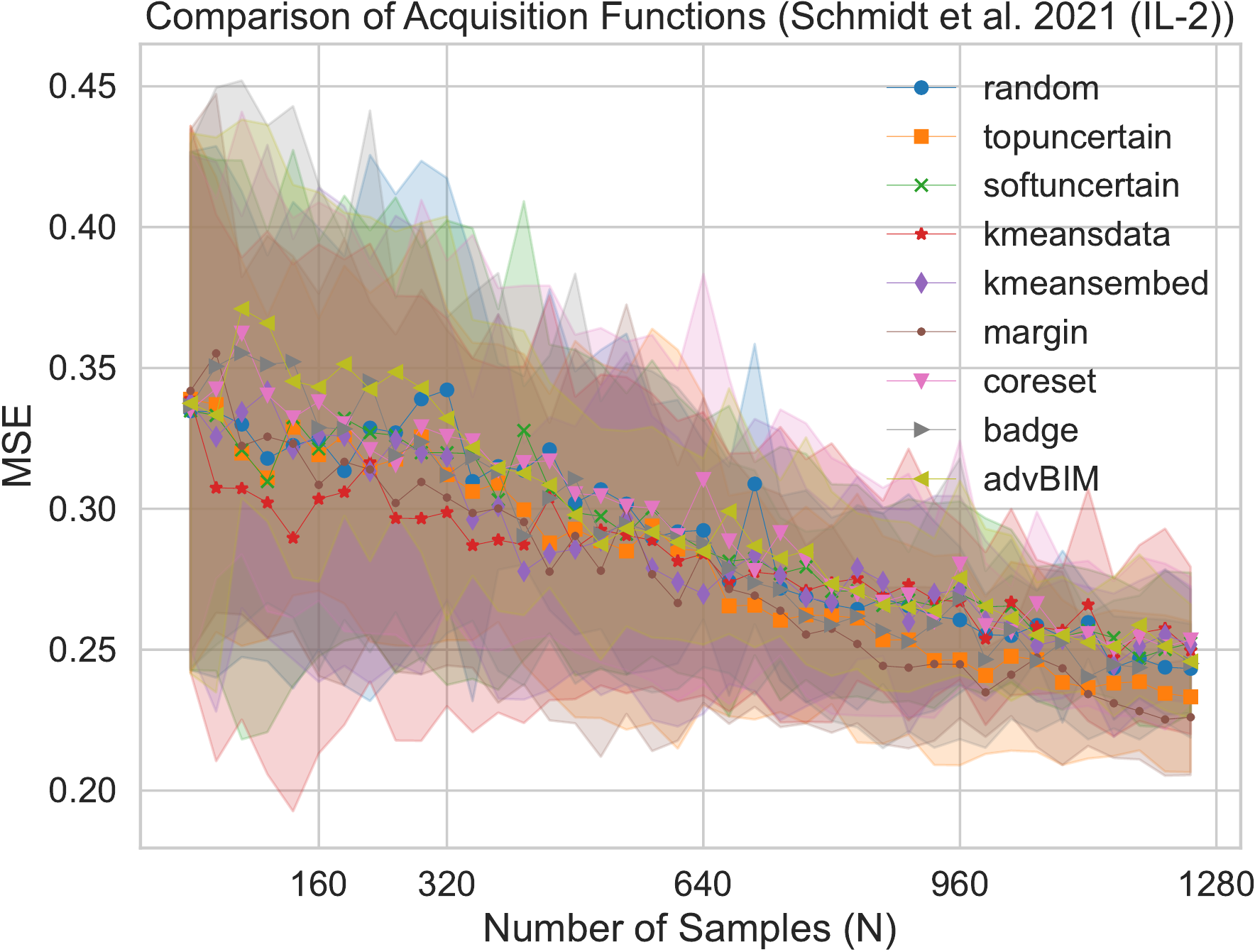}};
                        \end{tikzpicture}
                    }
                \end{subfigure}
                \&
                \begin{subfigure}{0.28\columnwidth}
                    \hspace{-32mm}
                    \centering
                    \resizebox{\linewidth}{!}{
                        \begin{tikzpicture}
                            \node (img)  {\includegraphics[width=\textwidth]{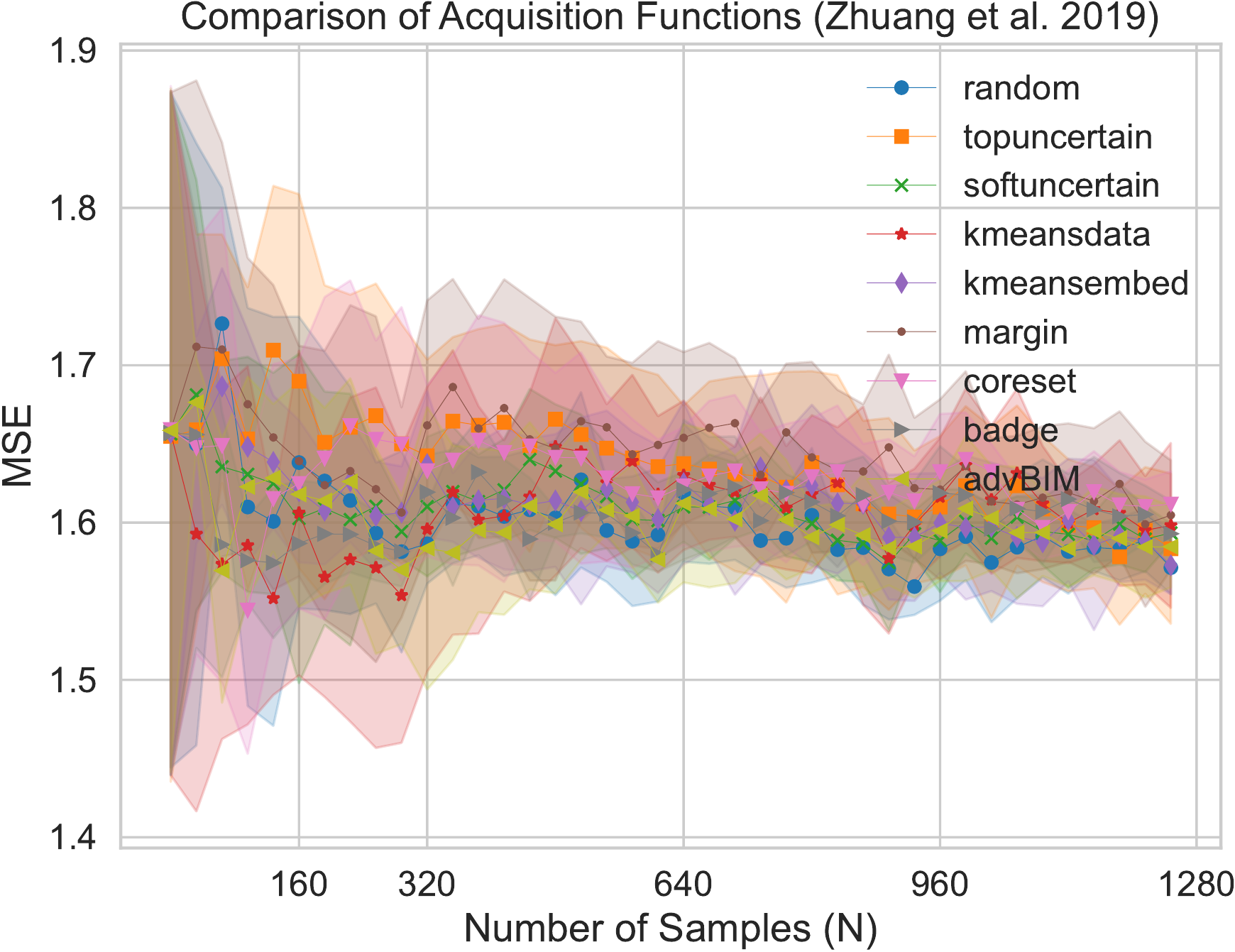}};
                        \end{tikzpicture}
                    }
                \end{subfigure}
                \&
                \\
\begin{subfigure}{0.27\columnwidth}
                    \hspace{-17mm}
                    \centering
                    \resizebox{\linewidth}{!}{
                        \begin{tikzpicture}
                            \node (img)  {\includegraphics[width=\textwidth]{figs/bnnplots/data_shifrut_2018_feat_string_bs64.pdf}};
                        \end{tikzpicture}
                    }
                \end{subfigure}
                \&
                \begin{subfigure}{0.27\columnwidth}
                    \hspace{-23mm}
                    \centering
                    \resizebox{\linewidth}{!}{
                        \begin{tikzpicture}
                            \node (img)  {\includegraphics[width=\textwidth]{figs/bnnplots/data_schmidt_2021_ifng_feat_string_bs64.pdf}};
                        \end{tikzpicture}
                    }
                \end{subfigure}
                \&
                \begin{subfigure}{0.27\columnwidth}
                    \hspace{-28mm}
                    \centering
                    \resizebox{\linewidth}{!}{
                        \begin{tikzpicture}
                            \node (img)  {\includegraphics[width=\textwidth]{figs/bnnplots/data_schmidt_2021_il2_feat_string_bs64.pdf}};
                        \end{tikzpicture}
                    }
                \end{subfigure}
                \&
                \begin{subfigure}{0.28\columnwidth}
                    \hspace{-32mm}
                    \centering
                    \resizebox{\linewidth}{!}{
                        \begin{tikzpicture}
                            \node (img)  {\includegraphics[width=\textwidth]{figs/bnnplots/data_zhuang_2019_nk_feat_string_bs64.pdf}};
                        \end{tikzpicture}
                    }
                \end{subfigure}
                \&
                \\
\begin{subfigure}{0.27\columnwidth}
                    \hspace{-17mm}
                    \centering
                    \resizebox{\linewidth}{!}{
                        \begin{tikzpicture}
                            \node (img)  {\includegraphics[width=\textwidth]{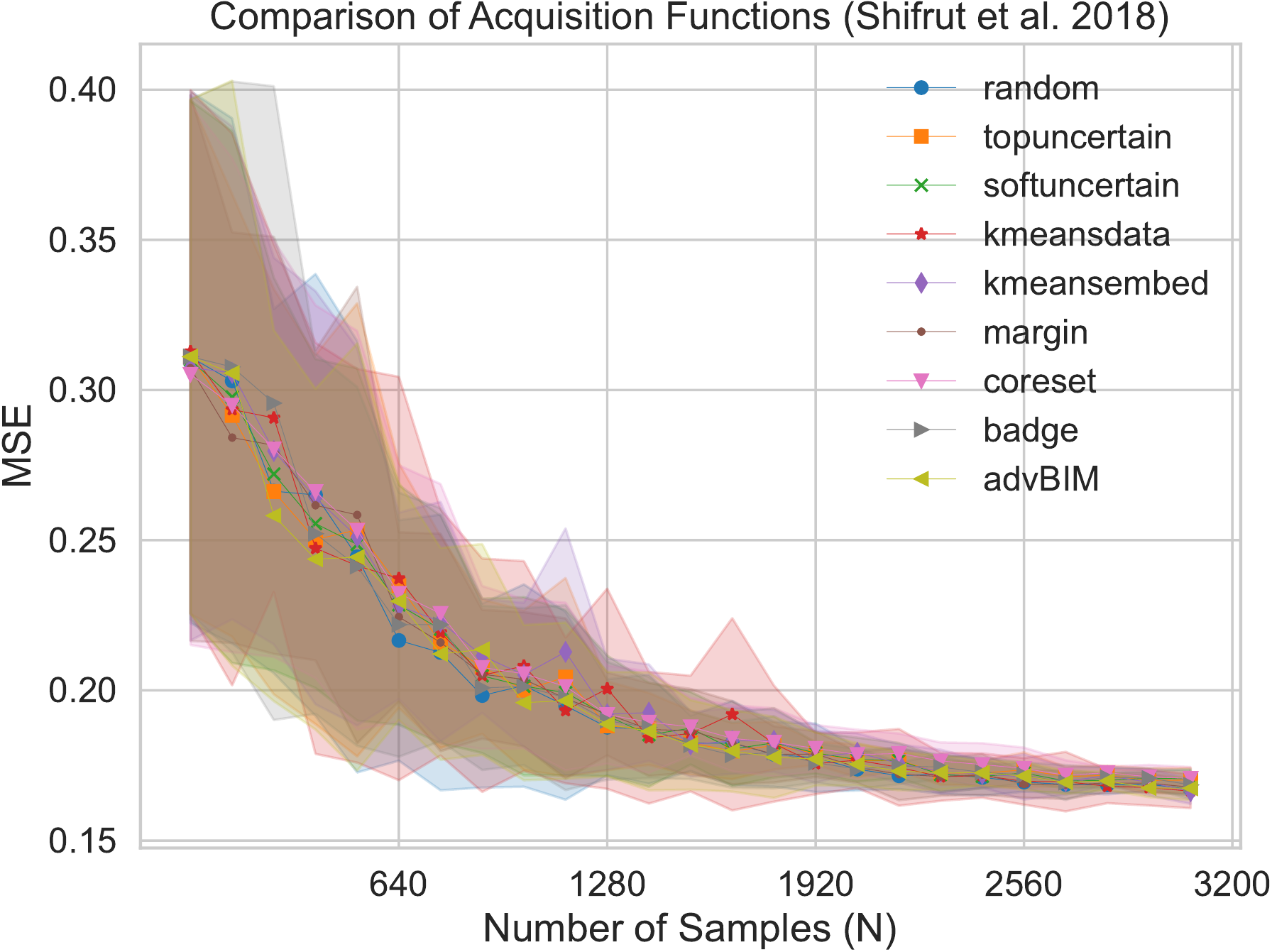}};
                        \end{tikzpicture}
                    }
                \end{subfigure}
                \&
                \begin{subfigure}{0.27\columnwidth}
                    \hspace{-23mm}
                    \centering
                    \resizebox{\linewidth}{!}{
                        \begin{tikzpicture}
                            \node (img)  {\includegraphics[width=\textwidth]{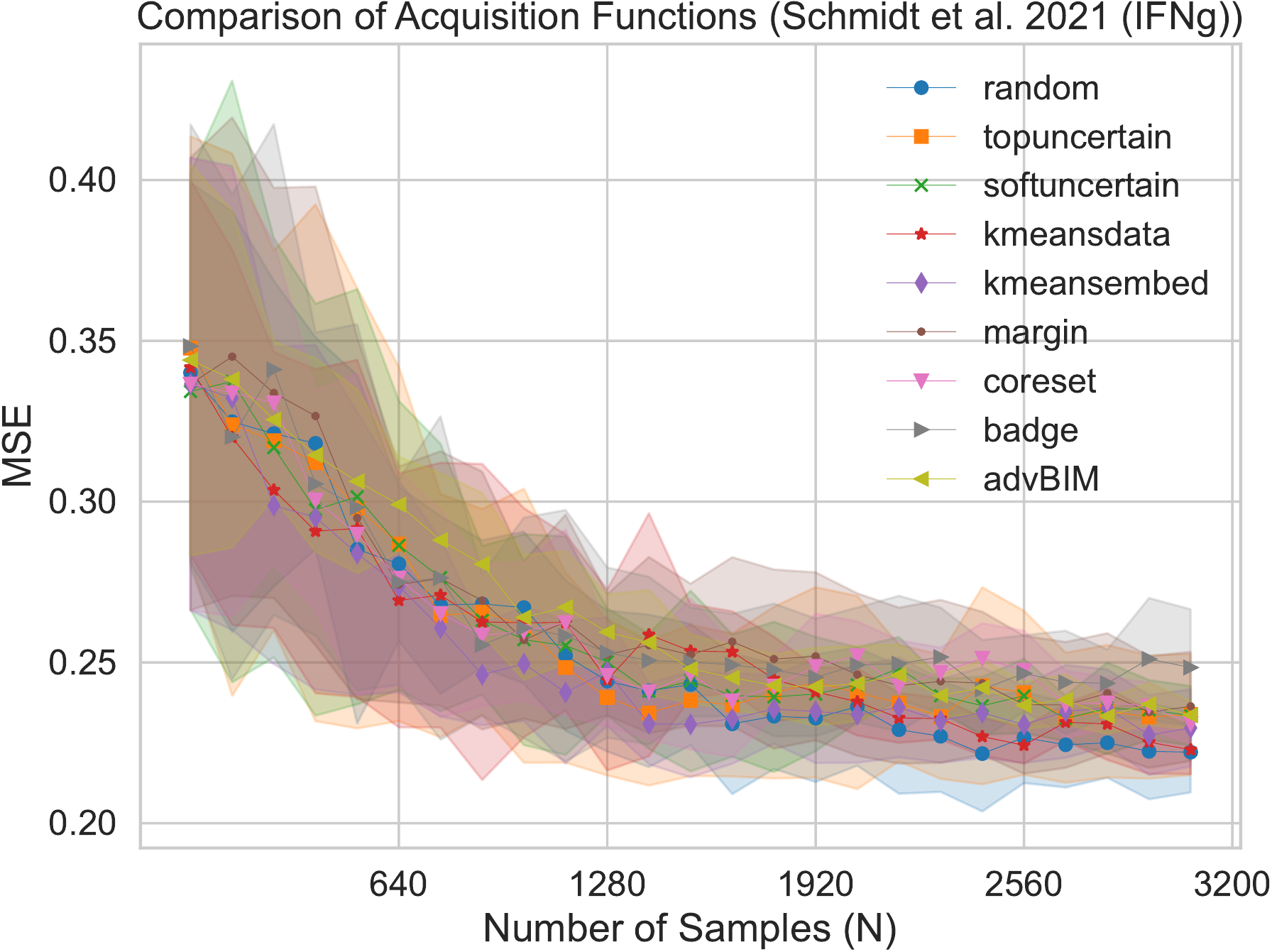}};
                        \end{tikzpicture}
                    }
                \end{subfigure}
                \&
                \begin{subfigure}{0.28\columnwidth}
                    \hspace{-28mm}
                    \centering
                    \resizebox{\linewidth}{!}{
                        \begin{tikzpicture}
                            \node (img)  {\includegraphics[width=\textwidth]{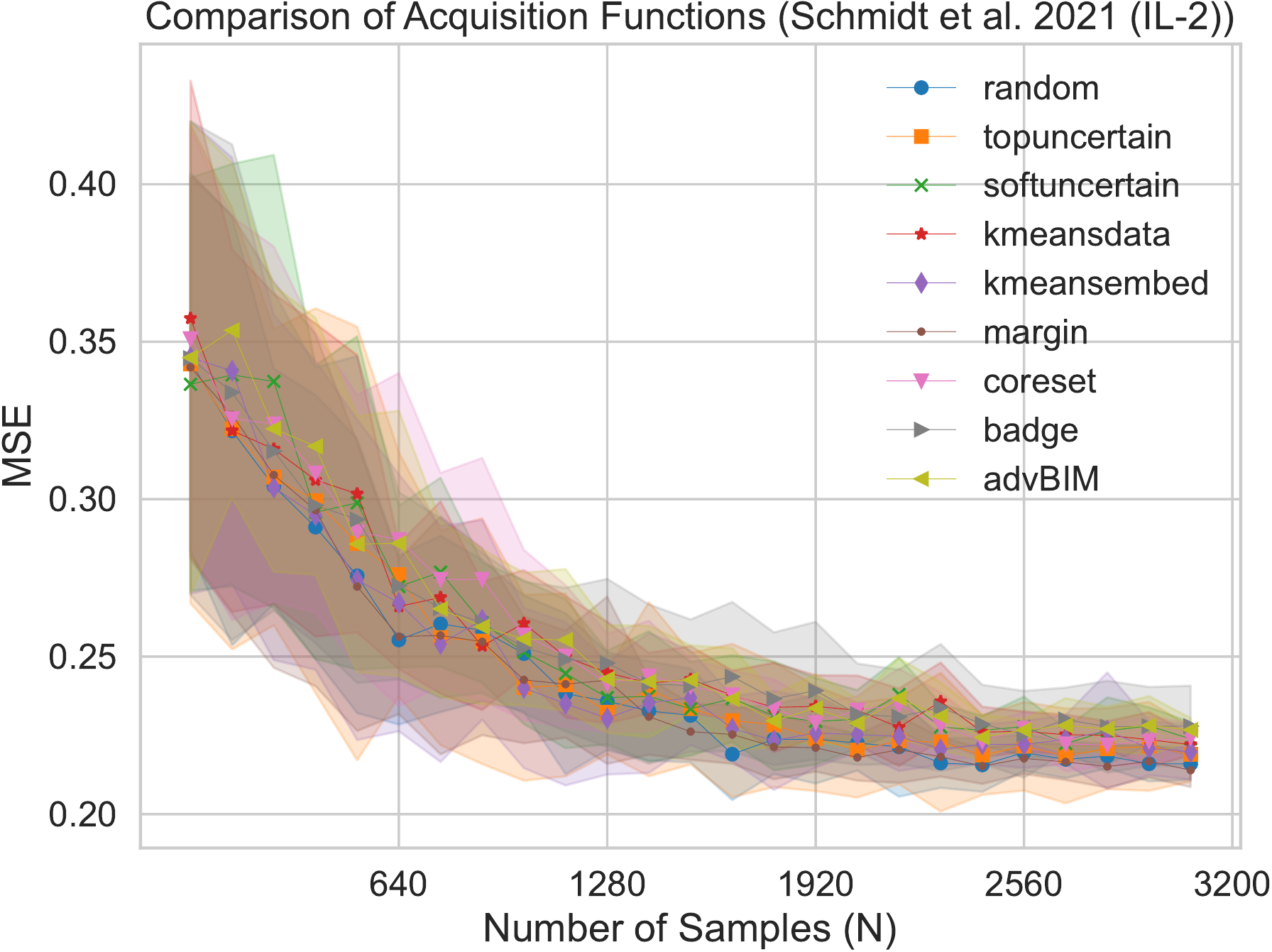}};
                        \end{tikzpicture}
                    }
                \end{subfigure}
                \&
                \begin{subfigure}{0.29\columnwidth}
                    \hspace{-32mm}
                    \centering
                    \resizebox{\linewidth}{!}{
                        \begin{tikzpicture}
                            \node (img)  {\includegraphics[width=\textwidth]{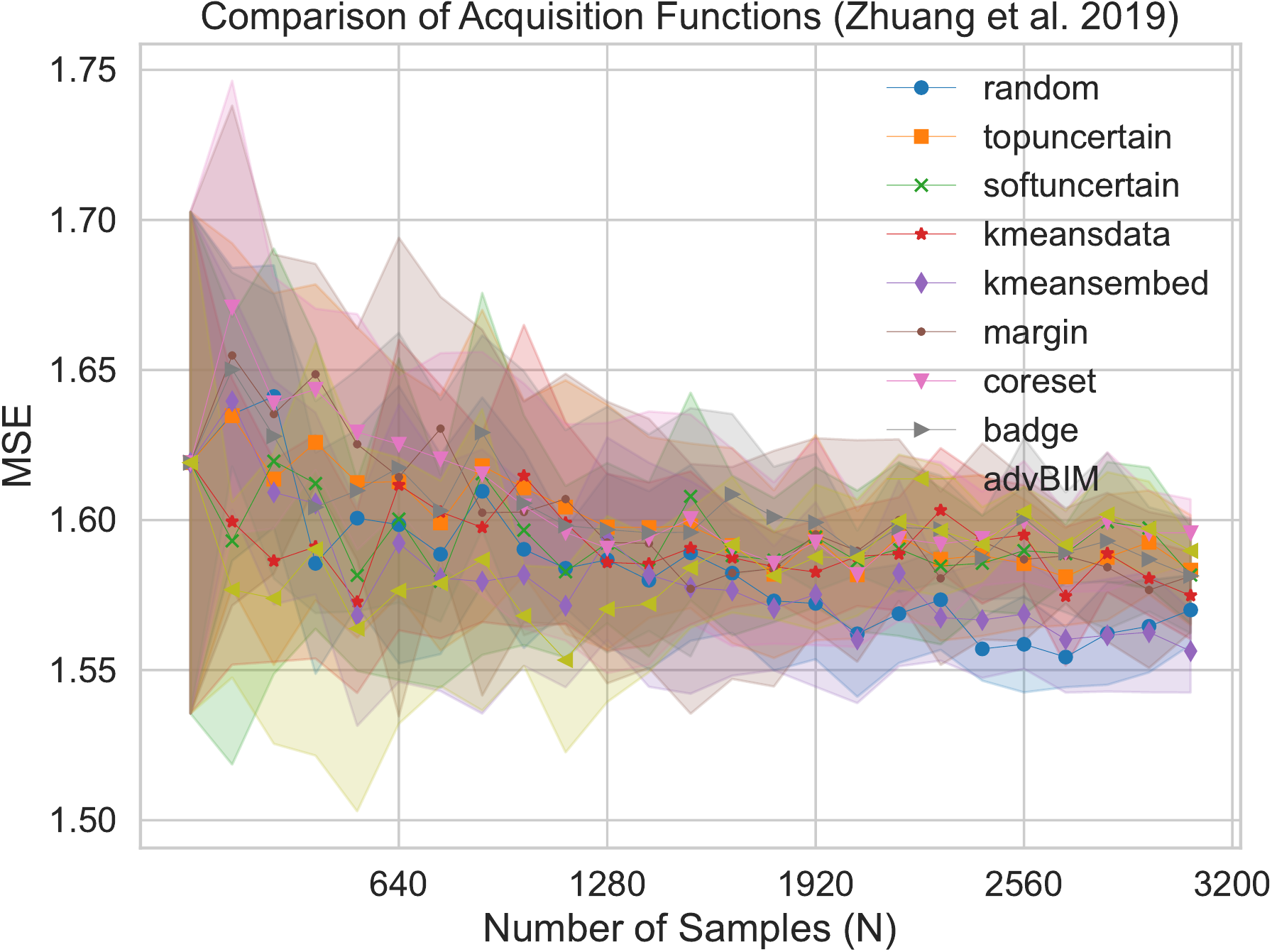}};
                        \end{tikzpicture}
                    }
                \end{subfigure}
                \&
                \\
\begin{subfigure}{0.275\columnwidth}
                    \hspace{-17mm}
                    \centering
                    \resizebox{\linewidth}{!}{
                        \begin{tikzpicture}
                            \node (img)  {\includegraphics[width=\textwidth]{figs/bnnplots/data_shifrut_2018_feat_string_bs256.pdf}};
                        \end{tikzpicture}
                    }
                \end{subfigure}
                \&
                \begin{subfigure}{0.27\columnwidth}
                    \hspace{-23mm}
                    \centering
                    \resizebox{\linewidth}{!}{
                        \begin{tikzpicture}
                            \node (img)  {\includegraphics[width=\textwidth]{figs/bnnplots/data_schmidt_2021_ifng_feat_string_bs256.pdf}};
                        \end{tikzpicture}
                    }
                \end{subfigure}
                \&
                \begin{subfigure}{0.27\columnwidth}
                    \hspace{-28mm}
                    \centering
                    \resizebox{\linewidth}{!}{
                        \begin{tikzpicture}
                            \node (img)  {\includegraphics[width=\textwidth]{figs/bnnplots/data_schmidt_2021_il2_feat_string_bs256.pdf}};
                        \end{tikzpicture}
                    }
                \end{subfigure}
                \&
                \begin{subfigure}{0.29\columnwidth}
                    \hspace{-32mm}
                    \centering
                    \resizebox{\linewidth}{!}{
                        \begin{tikzpicture}
                            \node (img)  {\includegraphics[width=\textwidth]{figs/bnnplots/data_zhuang_2019_nk_feat_string_bs256.pdf}};
                        \end{tikzpicture}
                    }
                \end{subfigure}
                \&
                \\
\begin{subfigure}{0.28\columnwidth}
                    \hspace{-17mm}
                    \centering
                    \resizebox{\linewidth}{!}{
                        \begin{tikzpicture}
                            \node (img)  {\includegraphics[width=\textwidth]{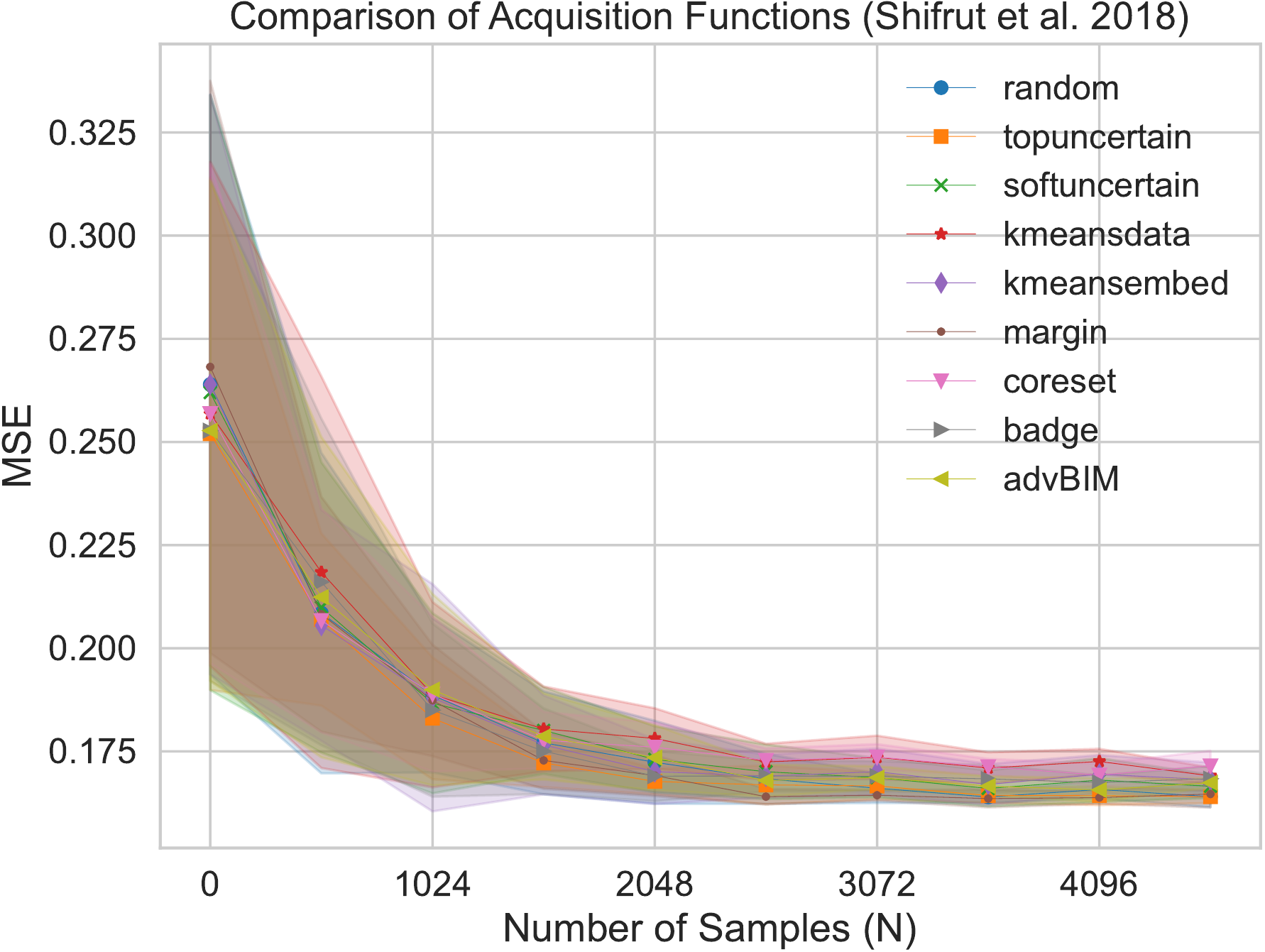}};
                        \end{tikzpicture}
                    }
                \end{subfigure}
                \&
                \begin{subfigure}{0.27\columnwidth}
                    \hspace{-23mm}
                    \centering
                    \resizebox{\linewidth}{!}{
                        \begin{tikzpicture}
                            \node (img)  {\includegraphics[width=\textwidth]{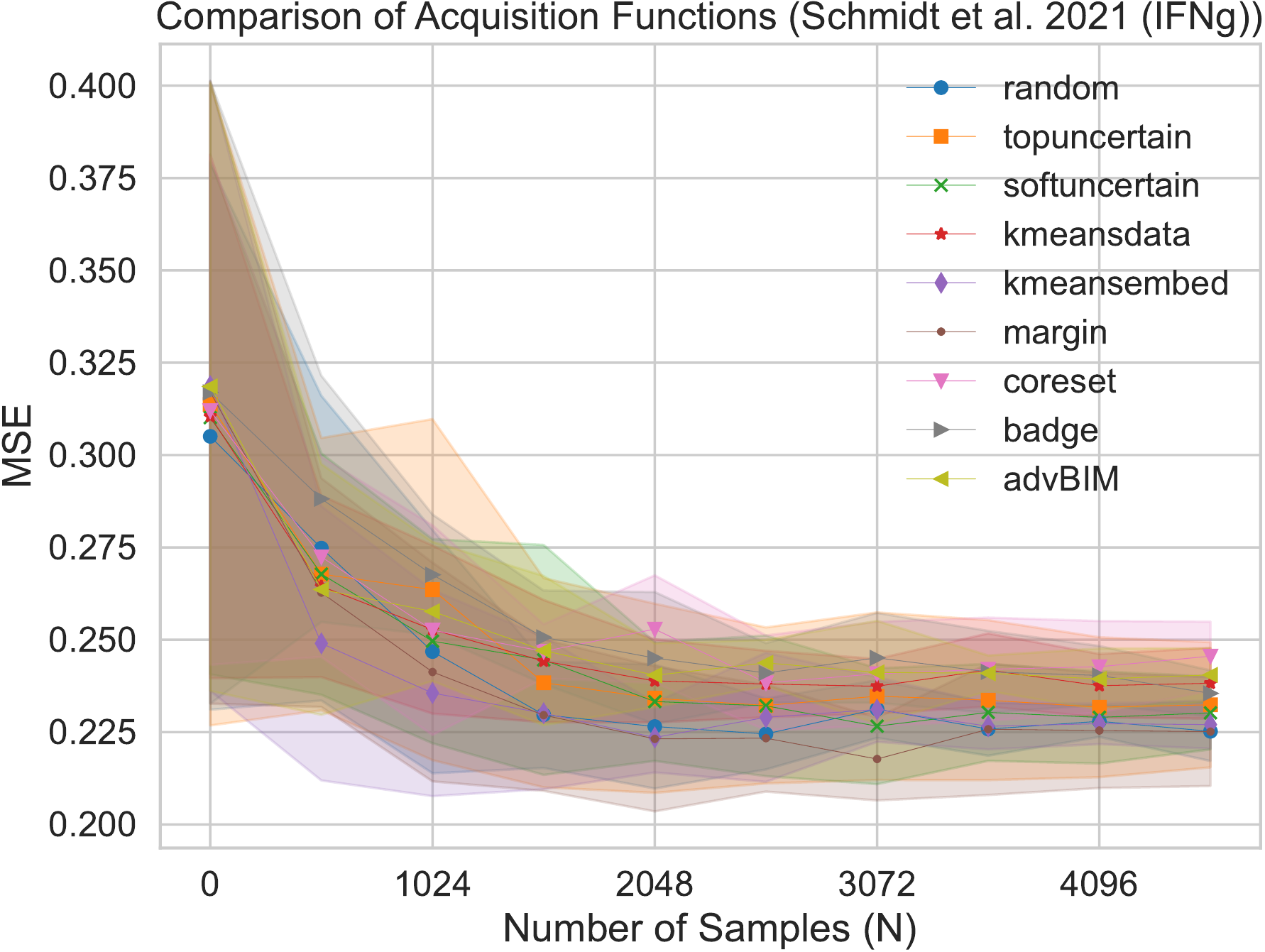}};
                        \end{tikzpicture}
                    }
                \end{subfigure}
                \&
                \begin{subfigure}{0.27\columnwidth}
                    \hspace{-28mm}
                    \centering
                    \resizebox{\linewidth}{!}{
                        \begin{tikzpicture}
                            \node (img)  {\includegraphics[width=\textwidth]{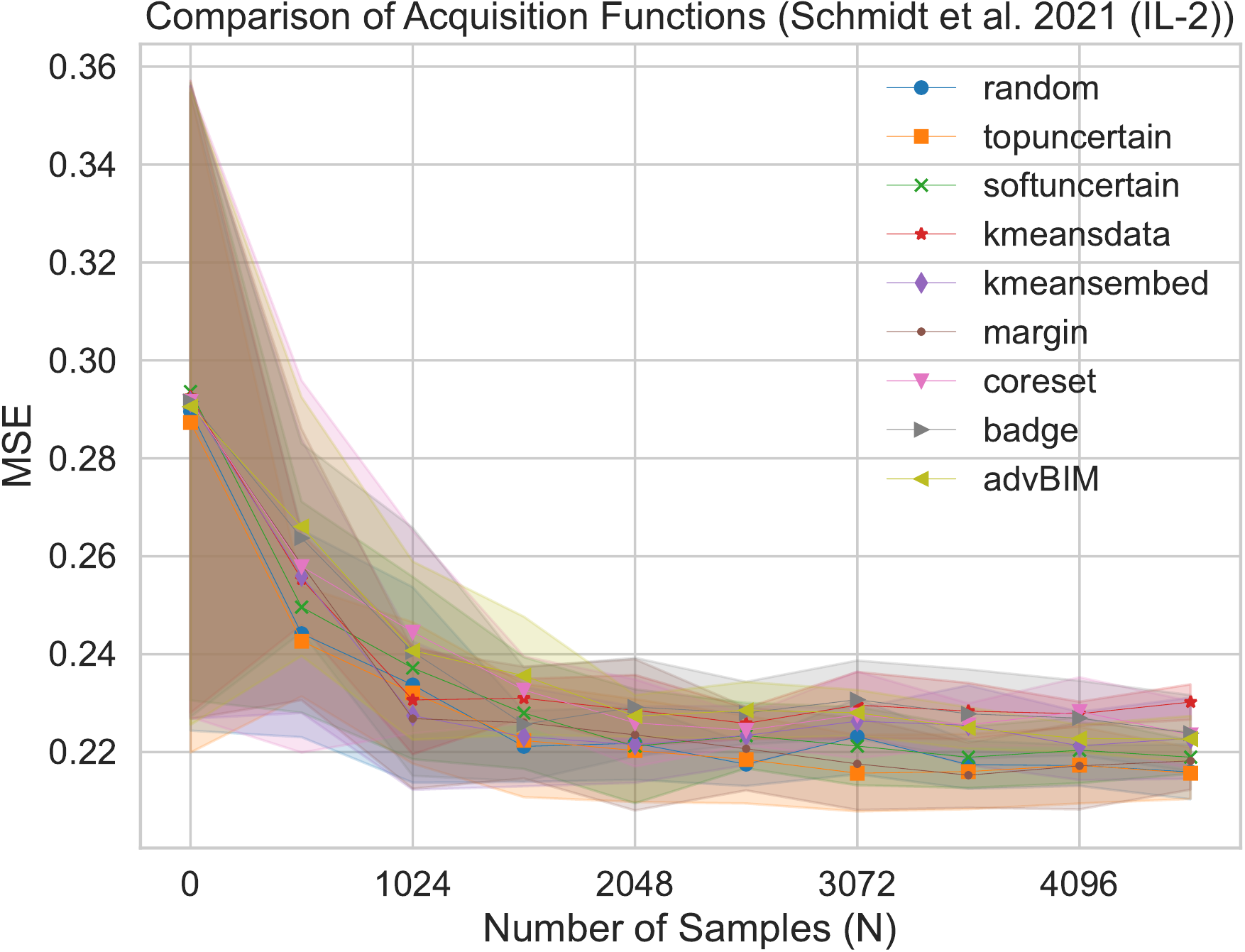}};
                        \end{tikzpicture}
                    }
                \end{subfigure}
                \&
                \begin{subfigure}{0.28\columnwidth}
                    \hspace{-32mm}
                    \centering
                    \resizebox{\linewidth}{!}{
                        \begin{tikzpicture}
                            \node (img)  {\includegraphics[width=\textwidth]{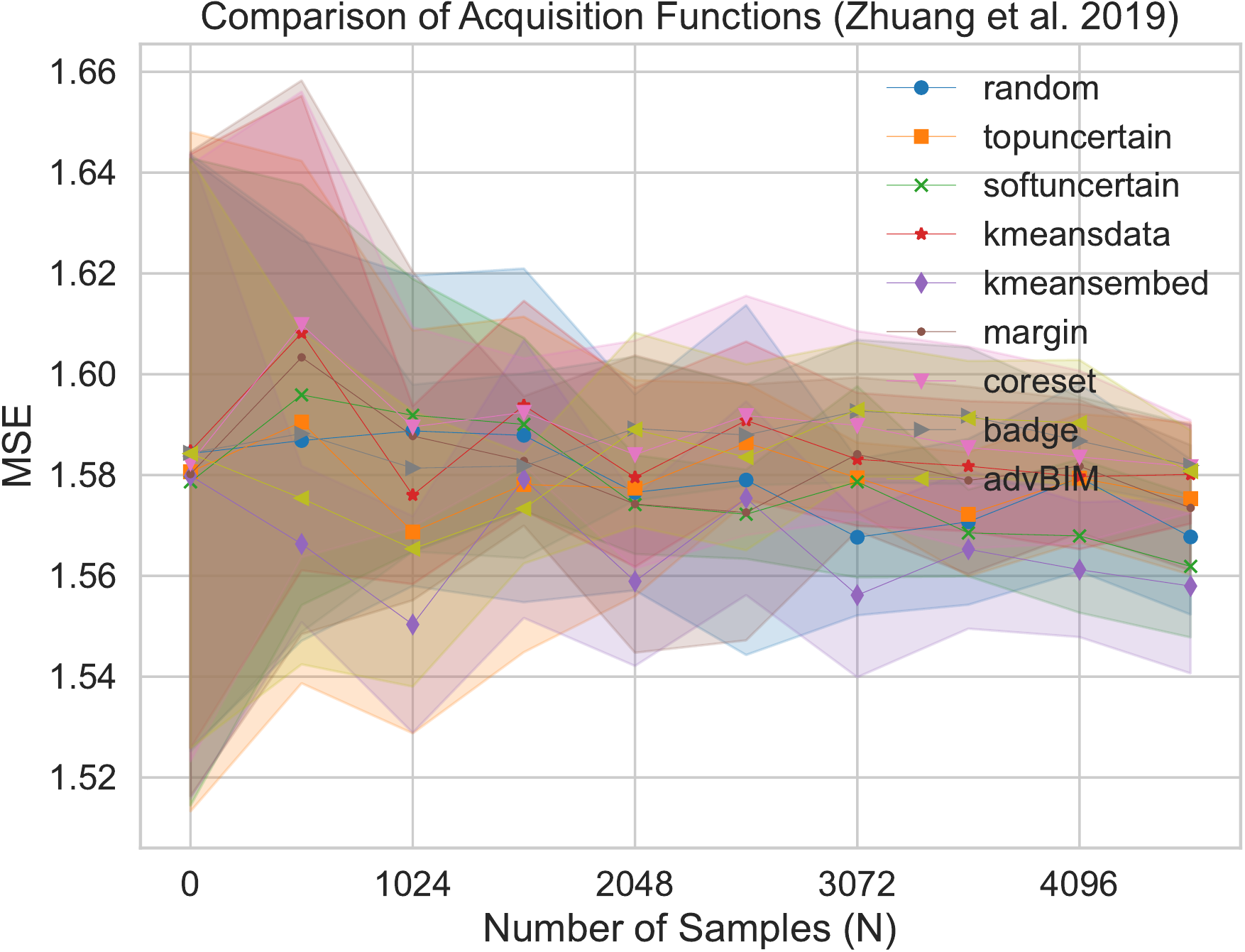}};
                        \end{tikzpicture}
                    }
                \end{subfigure}
                \&
                \\
            \\
           
            \\
            };
            \node [draw=none, rotate=90] at ([xshift=-8mm, yshift=2mm]fig-1-1.west) {\small batch size = 16};
            \node [draw=none, rotate=90] at ([xshift=-8mm, yshift=2mm]fig-2-1.west) {\small batch size = 32};
            \node [draw=none, rotate=90] at ([xshift=-8mm, yshift=2mm]fig-3-1.west) {\small batch size = 64};
            \node [draw=none, rotate=90] at ([xshift=-8mm, yshift=2mm]fig-4-1.west) {\small batch size = 128};
            \node [draw=none, rotate=90] at ([xshift=-8mm, yshift=2mm]fig-5-1.west) {\small batch size = 256};
            \node [draw=none, rotate=90] at ([xshift=-8mm, yshift=2mm]fig-6-1.west) {\small batch size = 512};
            \node [draw=none] at ([xshift=-6mm, yshift=3mm]fig-1-1.north) {\small Shifrut et al. 2018};
            \node [draw=none] at ([xshift=-9mm, yshift=3mm]fig-1-2.north) {\small Schmidt et al. 2021 (IFNg)};
            \node [draw=none] at ([xshift=-11mm, yshift=3mm]fig-1-3.north) {\small Schmidt et al. 2021 (IL-2)};
            \node [draw=none] at ([xshift=-13mm, yshift=2.5mm]fig-1-4.north) {\small Zhuang et al. 2019};
\end{tikzpicture}}
        \vspace{-7mm}
        \caption{The evaluation of the model trained with {STRING} treatment descriptors at each active learning cycle for 4 datasets and 6 acquisition batch sizes. In each plot, the x-axis is the active learning cycles multiplied by the acquisition bath size that gives the total number of data points collected so far. The y-axis is the test MSE error evaluated on the test data.}
        \vspace{-5mm}
        \label{fig:bnn_feat_string_alldatasets_allbathcsizes}
    \end{figure*}

\begin{figure*}
    \vspace{-2mm}
        \centering
        \makebox[0.72\paperwidth]{\begin{tikzpicture}[ampersand replacement=\&]
            \matrix (fig) [matrix of nodes]{ 
\begin{subfigure}{0.27\columnwidth}
                    \hspace{-17mm}
                    \centering
                    \resizebox{\linewidth}{!}{
                        \begin{tikzpicture}
                            \node (img)  {\includegraphics[width=\textwidth]{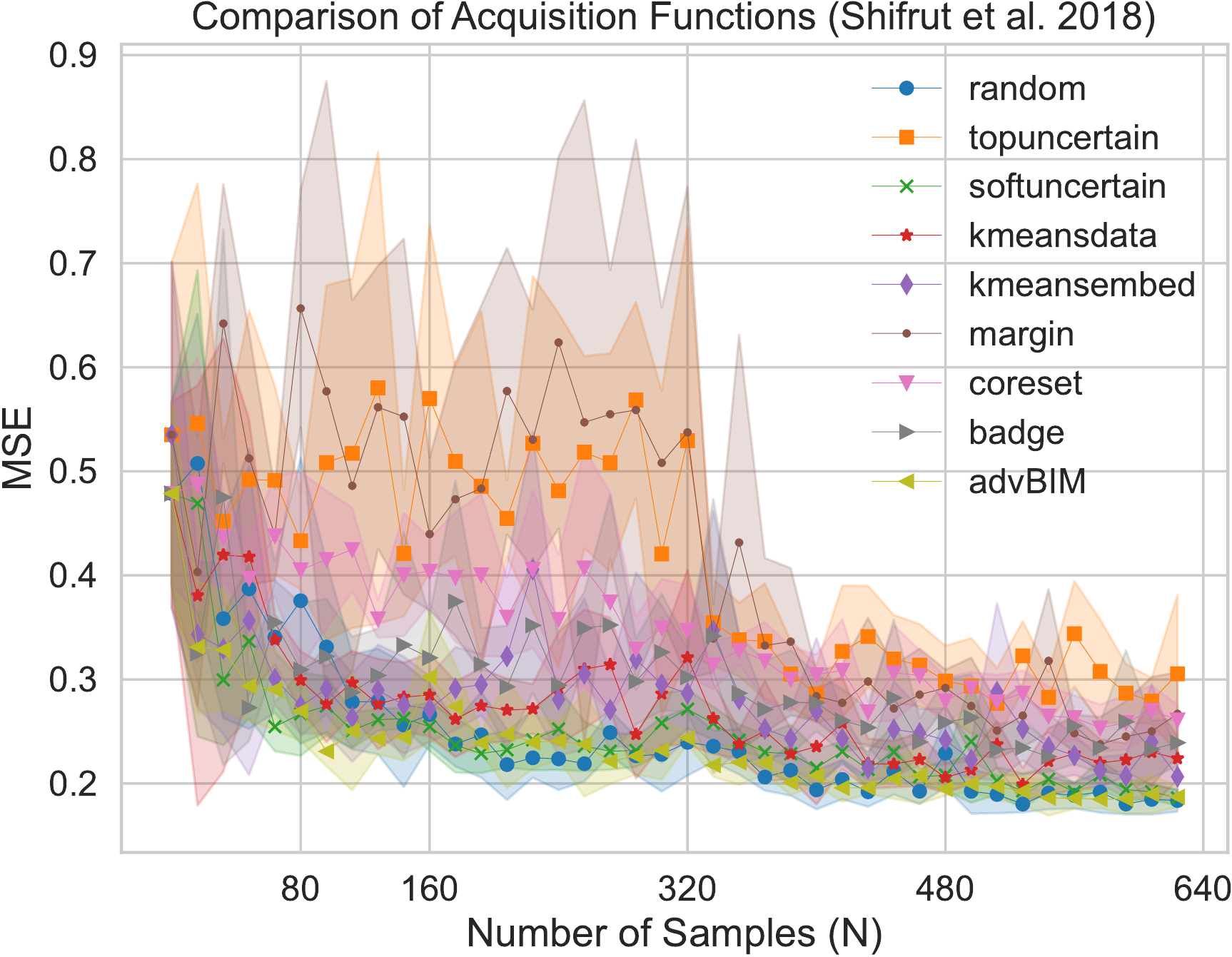}};
                        \end{tikzpicture}
                    }
                \end{subfigure}
                \&
                 \begin{subfigure}{0.27\columnwidth}
                    \hspace{-23mm}
                    \centering
                    \resizebox{\linewidth}{!}{
                        \begin{tikzpicture}
                            \node (img)  {\includegraphics[width=\textwidth]{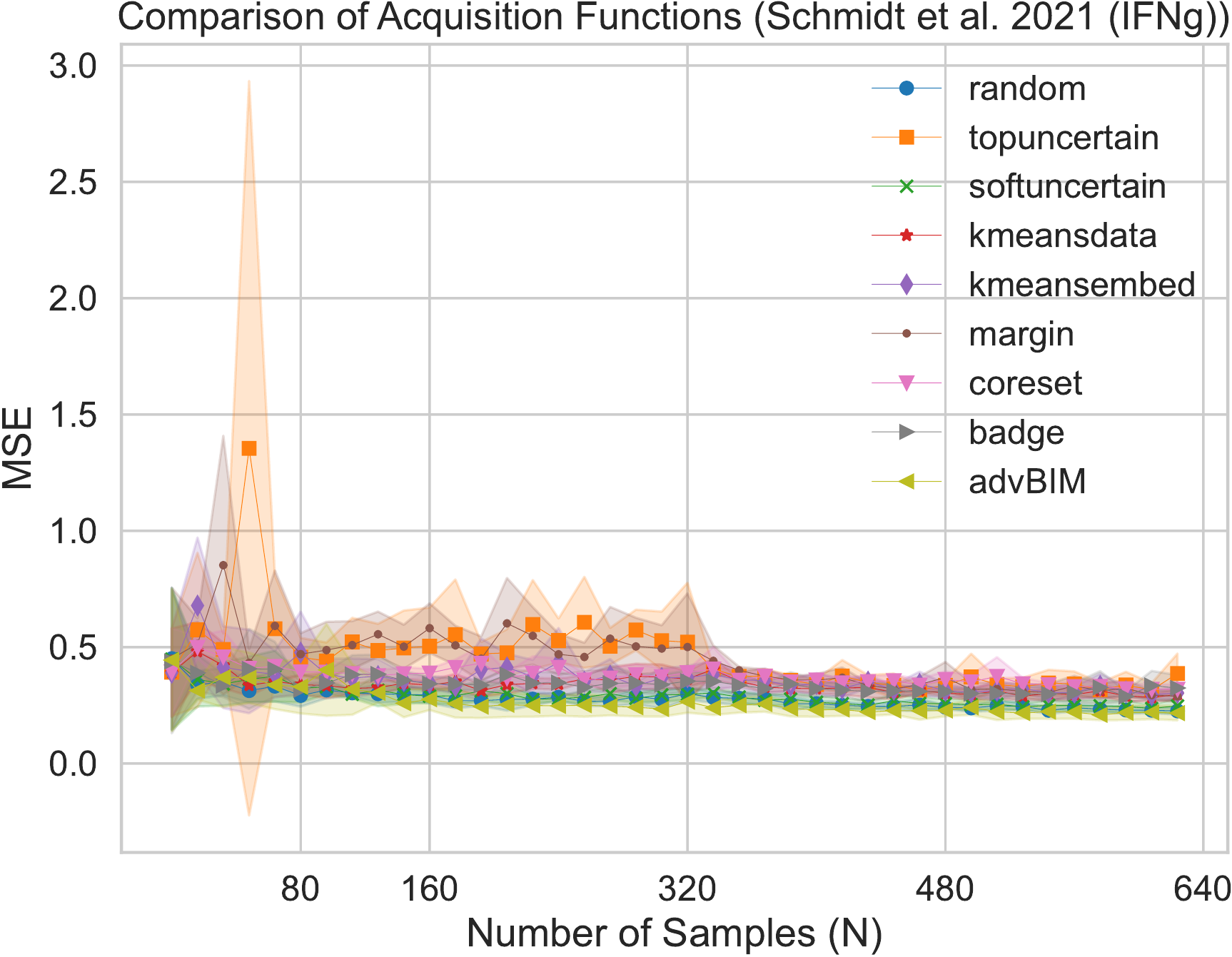}};
                        \end{tikzpicture}
                    }
                \end{subfigure}
                \&
                 \begin{subfigure}{0.27\columnwidth}
                    \hspace{-28mm}
                    \centering
                    \resizebox{\linewidth}{!}{
                        \begin{tikzpicture}
                            \node (img)  {\includegraphics[width=\textwidth]{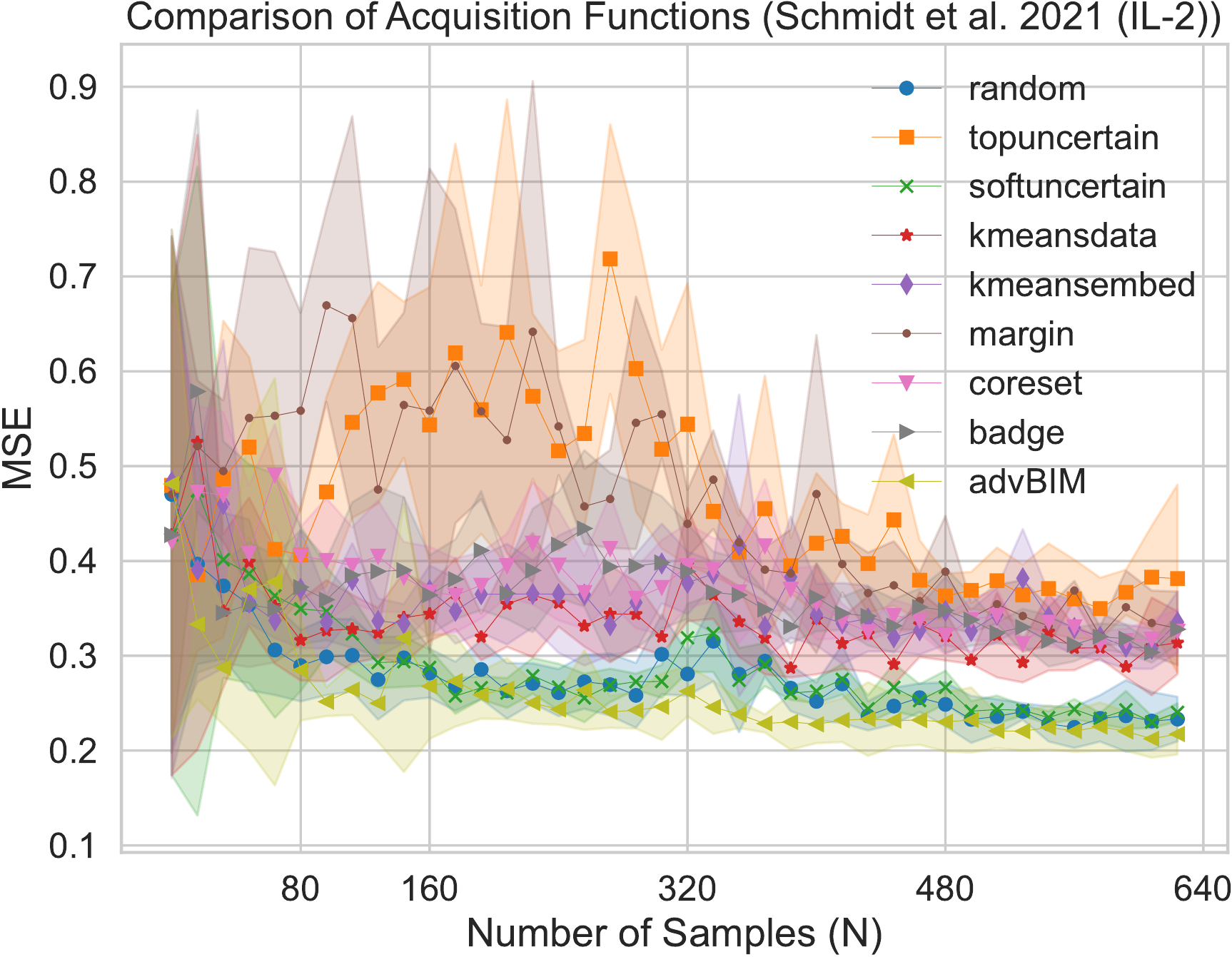}};
                        \end{tikzpicture}
                    }
                \end{subfigure}
                \&
                \begin{subfigure}{0.28\columnwidth}
                    \hspace{-32mm}
                    \centering
                    \resizebox{\linewidth}{!}{
                        \begin{tikzpicture}
                            \node (img)  {\includegraphics[width=\textwidth]{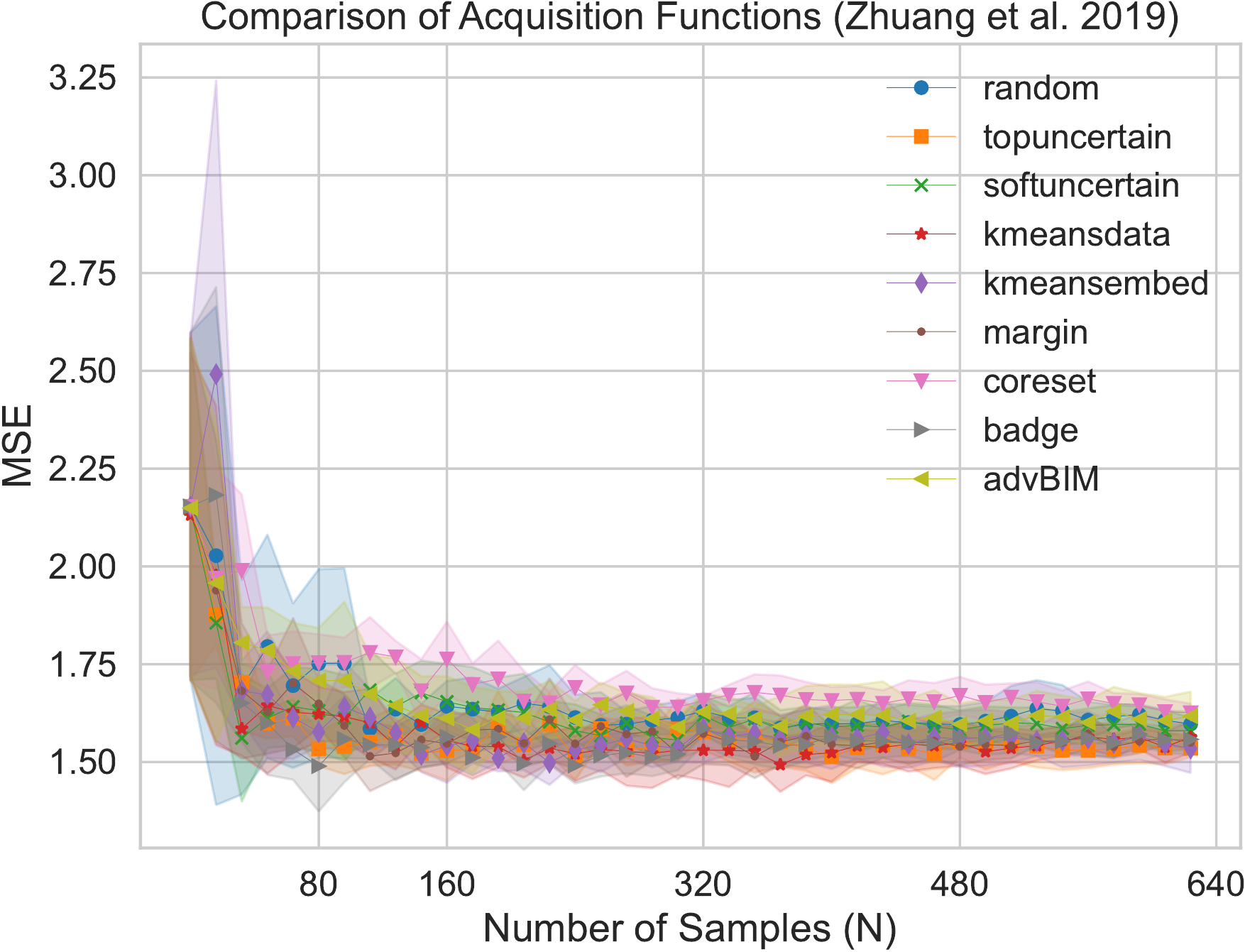}};
                        \end{tikzpicture}
                    }
                \end{subfigure}
                \&
            \\
\begin{subfigure}{0.27\columnwidth}
                    \hspace{-17mm}
                    \centering
                    \resizebox{\linewidth}{!}{
                        \begin{tikzpicture}
                            \node (img)  {\includegraphics[width=\textwidth]{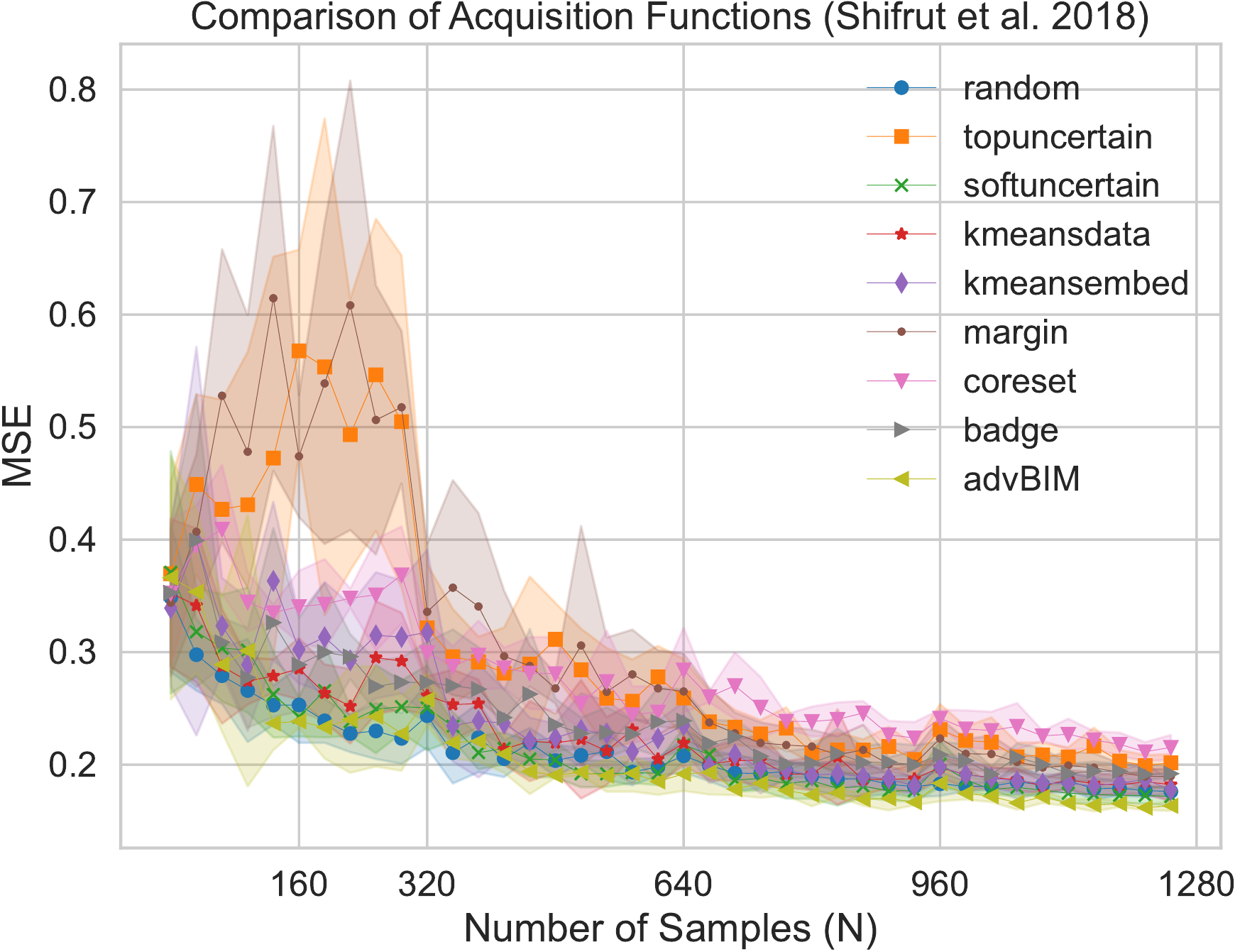}};
                        \end{tikzpicture}
                    }
                \end{subfigure}
                \&
                \begin{subfigure}{0.27\columnwidth}
                    \hspace{-23mm}
                    \centering
                    \resizebox{\linewidth}{!}{
                        \begin{tikzpicture}
                            \node (img)  {\includegraphics[width=\textwidth]{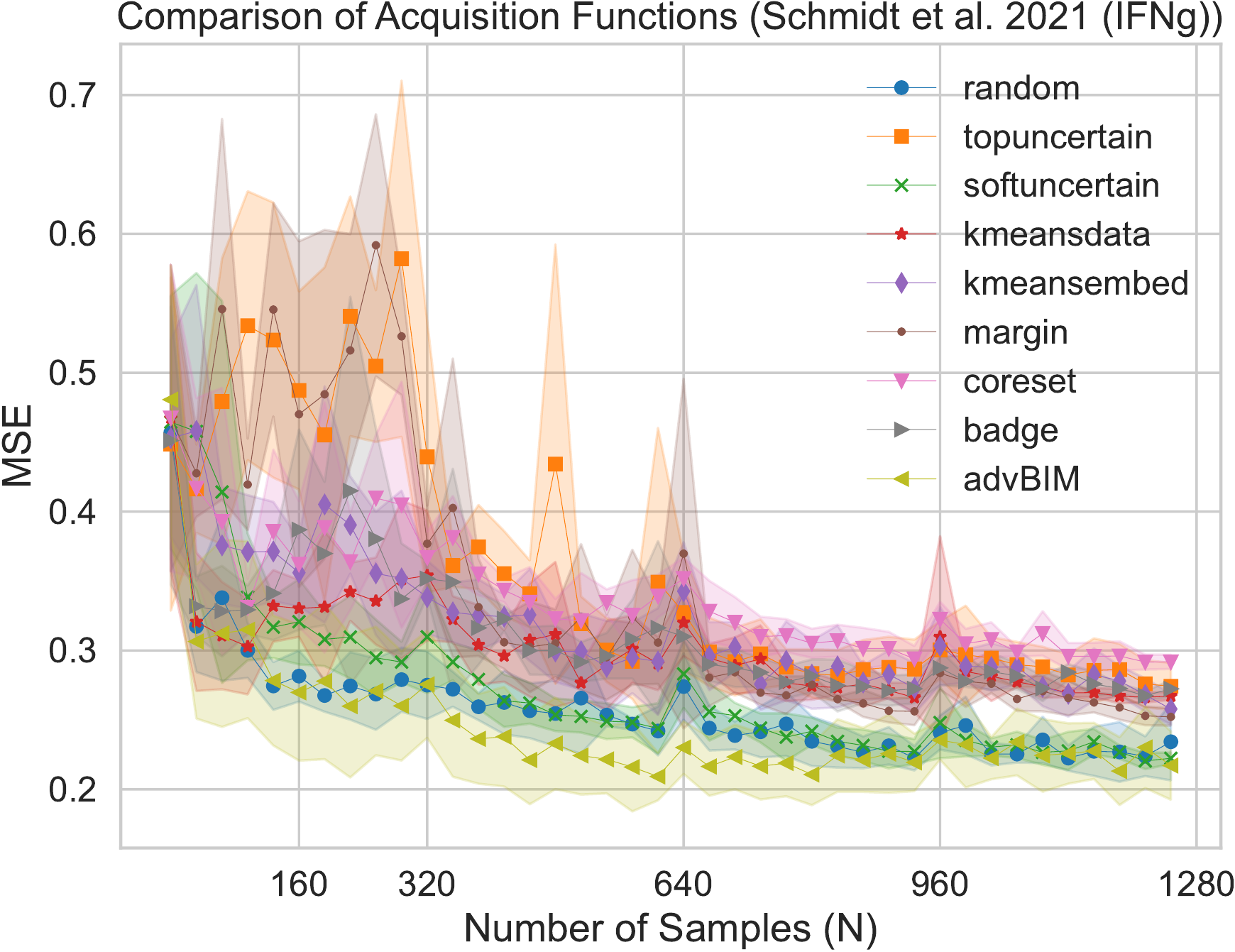}};
                        \end{tikzpicture}
                    }
                \end{subfigure}
                \&
                \begin{subfigure}{0.27\columnwidth}
                    \hspace{-28mm}
                    \centering
                    \resizebox{\linewidth}{!}{
                        \begin{tikzpicture}
                            \node (img)  {\includegraphics[width=\textwidth]{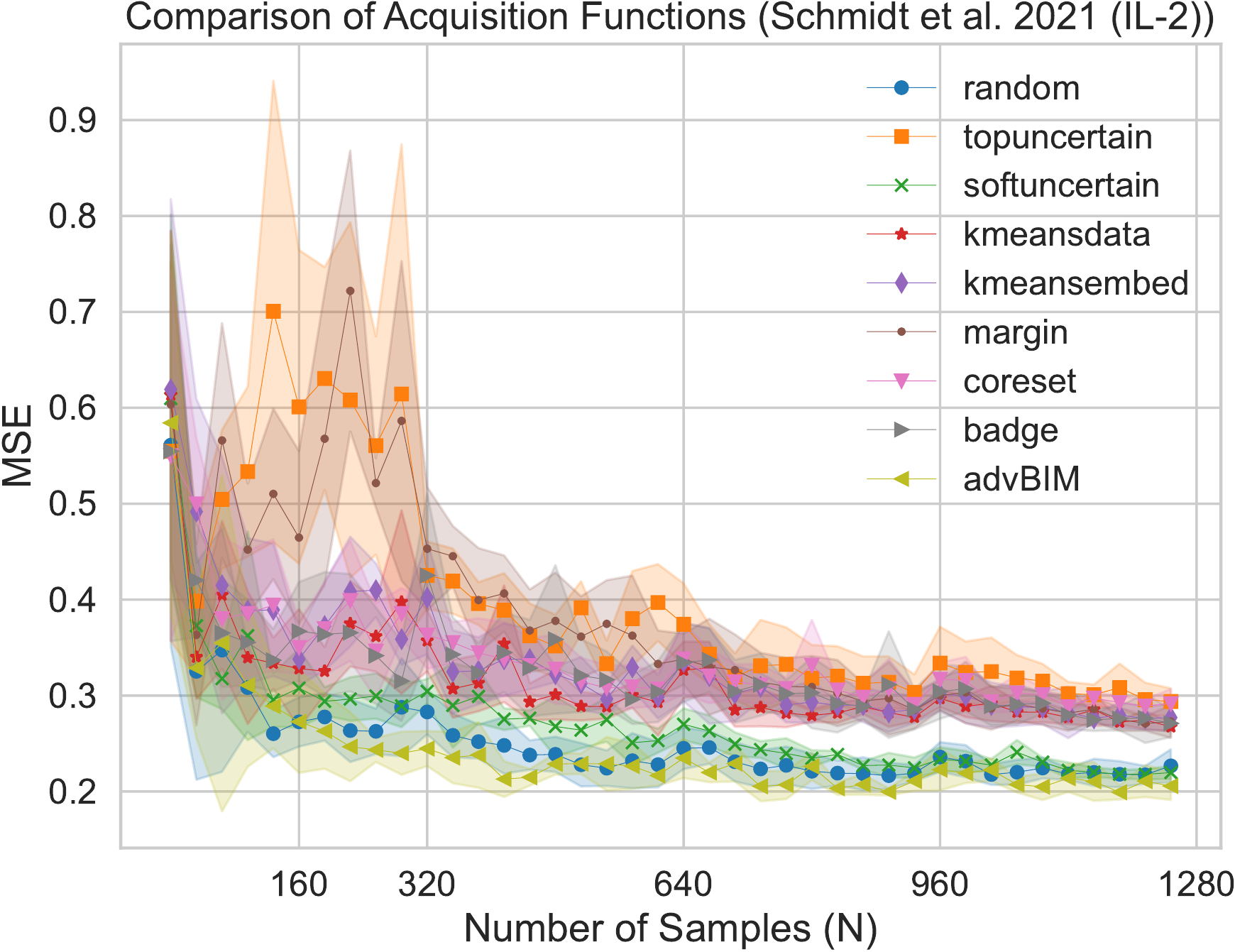}};
                        \end{tikzpicture}
                    }
                \end{subfigure}
                \&
                \begin{subfigure}{0.28\columnwidth}
                    \hspace{-32mm}
                    \centering
                    \resizebox{\linewidth}{!}{
                        \begin{tikzpicture}
                            \node (img)  {\includegraphics[width=\textwidth]{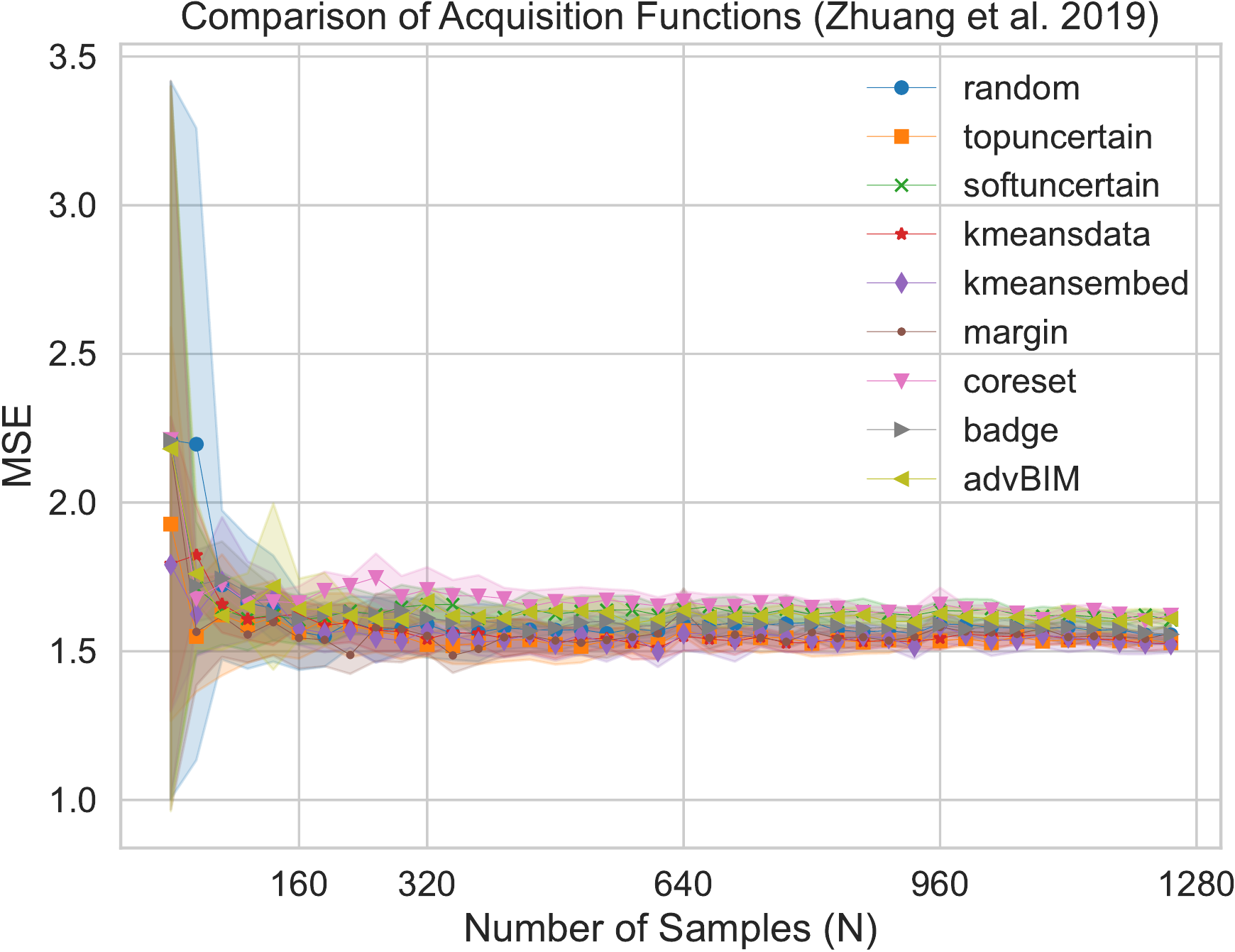}};
                        \end{tikzpicture}
                    }
                \end{subfigure}
                \&
                \\
\begin{subfigure}{0.27\columnwidth}
                    \hspace{-17mm}
                    \centering
                    \resizebox{\linewidth}{!}{
                        \begin{tikzpicture}
                            \node (img)  {\includegraphics[width=\textwidth]{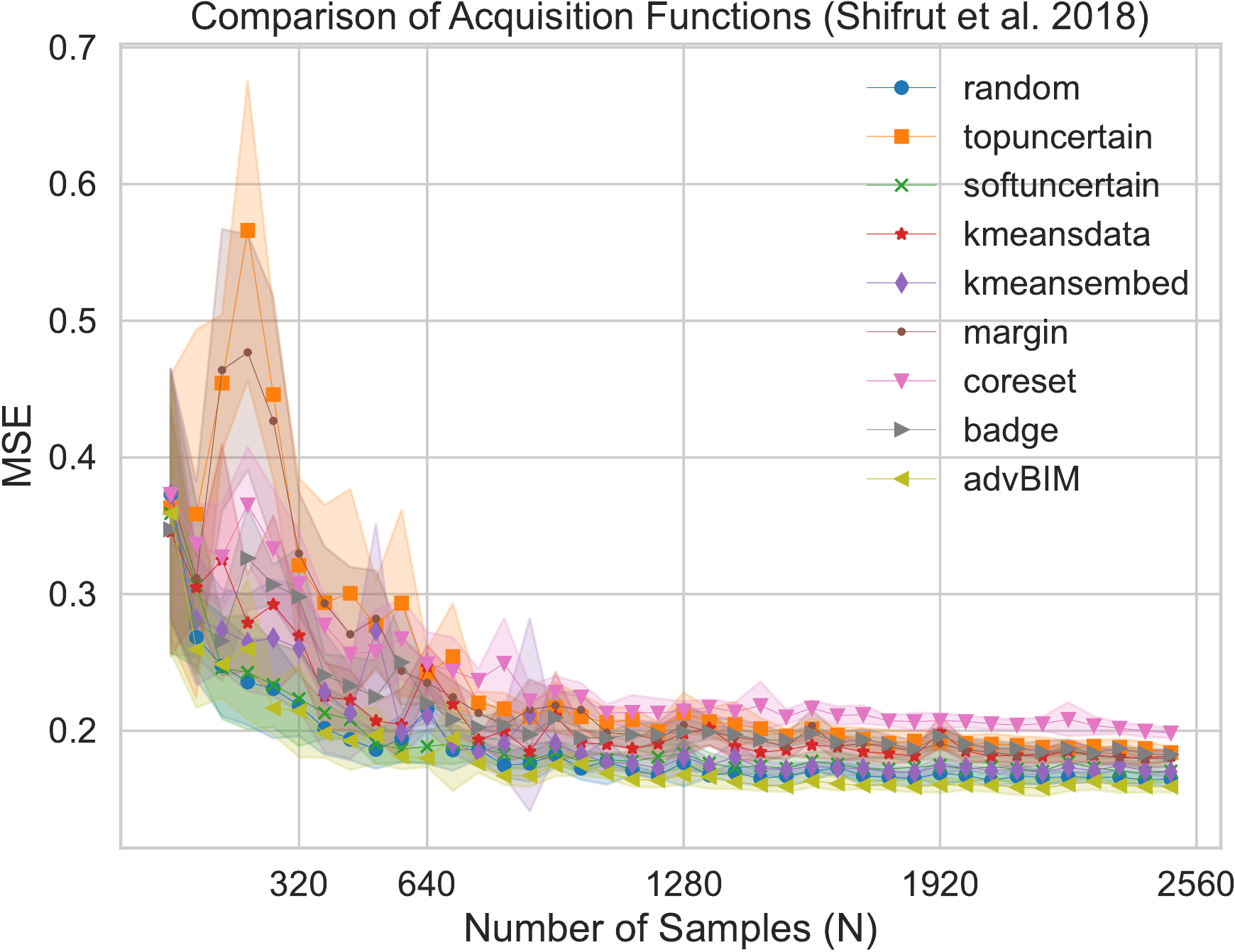}};
                        \end{tikzpicture}
                    }
                \end{subfigure}
                \&
                \begin{subfigure}{0.27\columnwidth}
                    \hspace{-23mm}
                    \centering
                    \resizebox{\linewidth}{!}{
                        \begin{tikzpicture}
                            \node (img)  {\includegraphics[width=\textwidth]{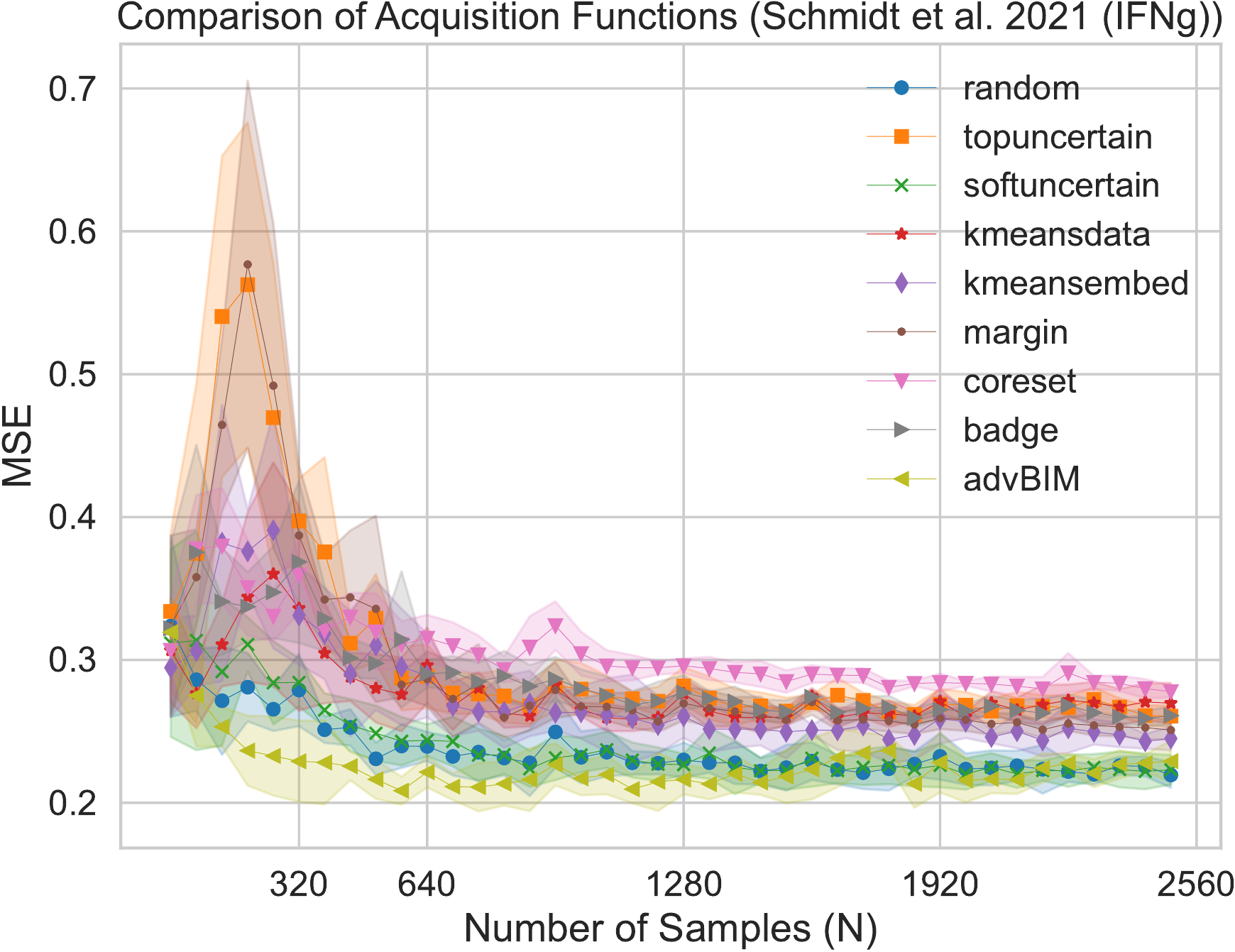}};
                        \end{tikzpicture}
                    }
                \end{subfigure}
                \&
                \begin{subfigure}{0.27\columnwidth}
                    \hspace{-28mm}
                    \centering
                    \resizebox{\linewidth}{!}{
                        \begin{tikzpicture}
                            \node (img)  {\includegraphics[width=\textwidth]{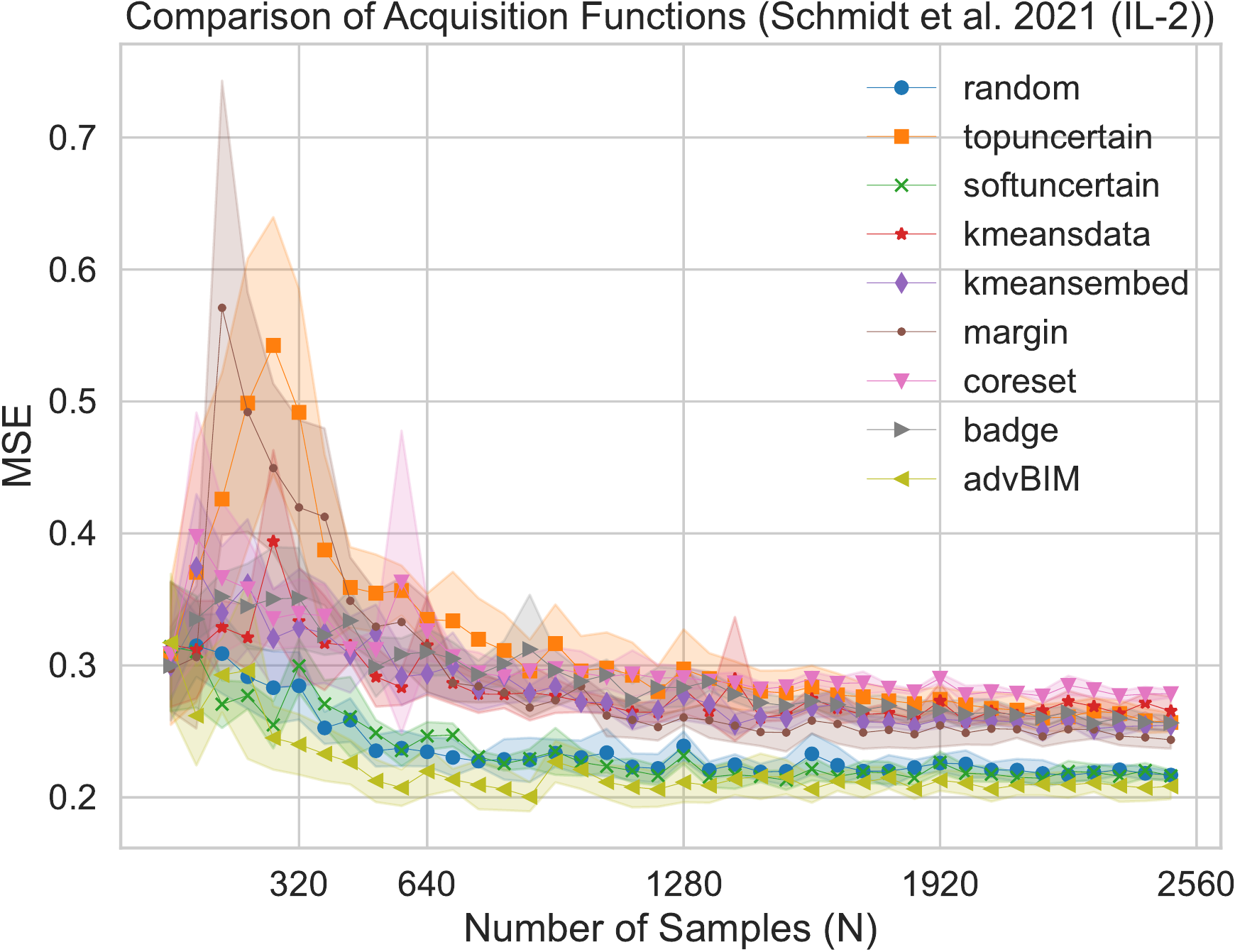}};
                        \end{tikzpicture}
                    }
                \end{subfigure}
                \&
                \begin{subfigure}{0.28\columnwidth}
                    \hspace{-32mm}
                    \centering
                    \resizebox{\linewidth}{!}{
                        \begin{tikzpicture}
                            \node (img)  {\includegraphics[width=\textwidth]{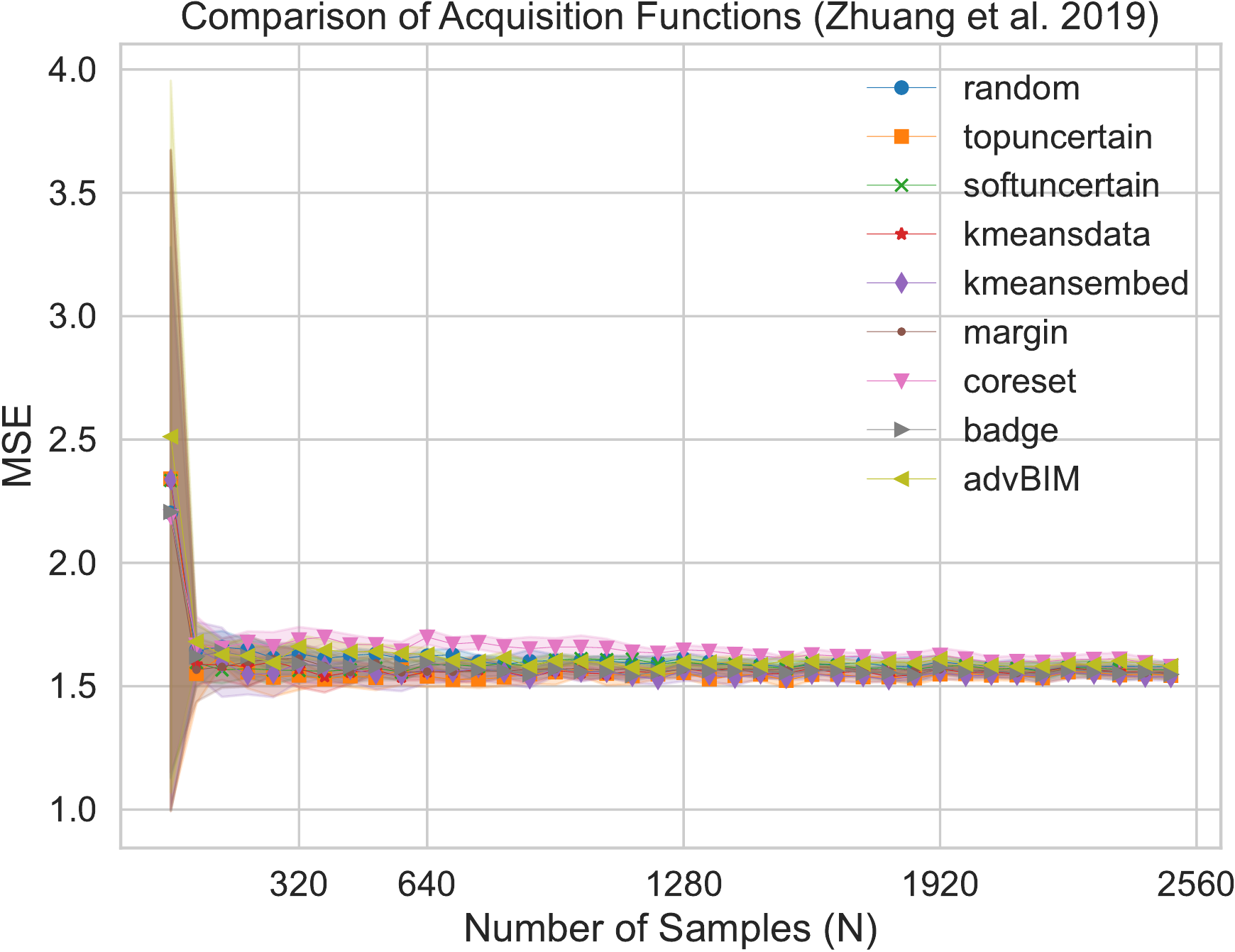}};
                        \end{tikzpicture}
                    }
                \end{subfigure}
                \&
                \\
\begin{subfigure}{0.27\columnwidth}
                    \hspace{-17mm}
                    \centering
                    \resizebox{\linewidth}{!}{
                        \begin{tikzpicture}
                            \node (img)  {\includegraphics[width=\textwidth]{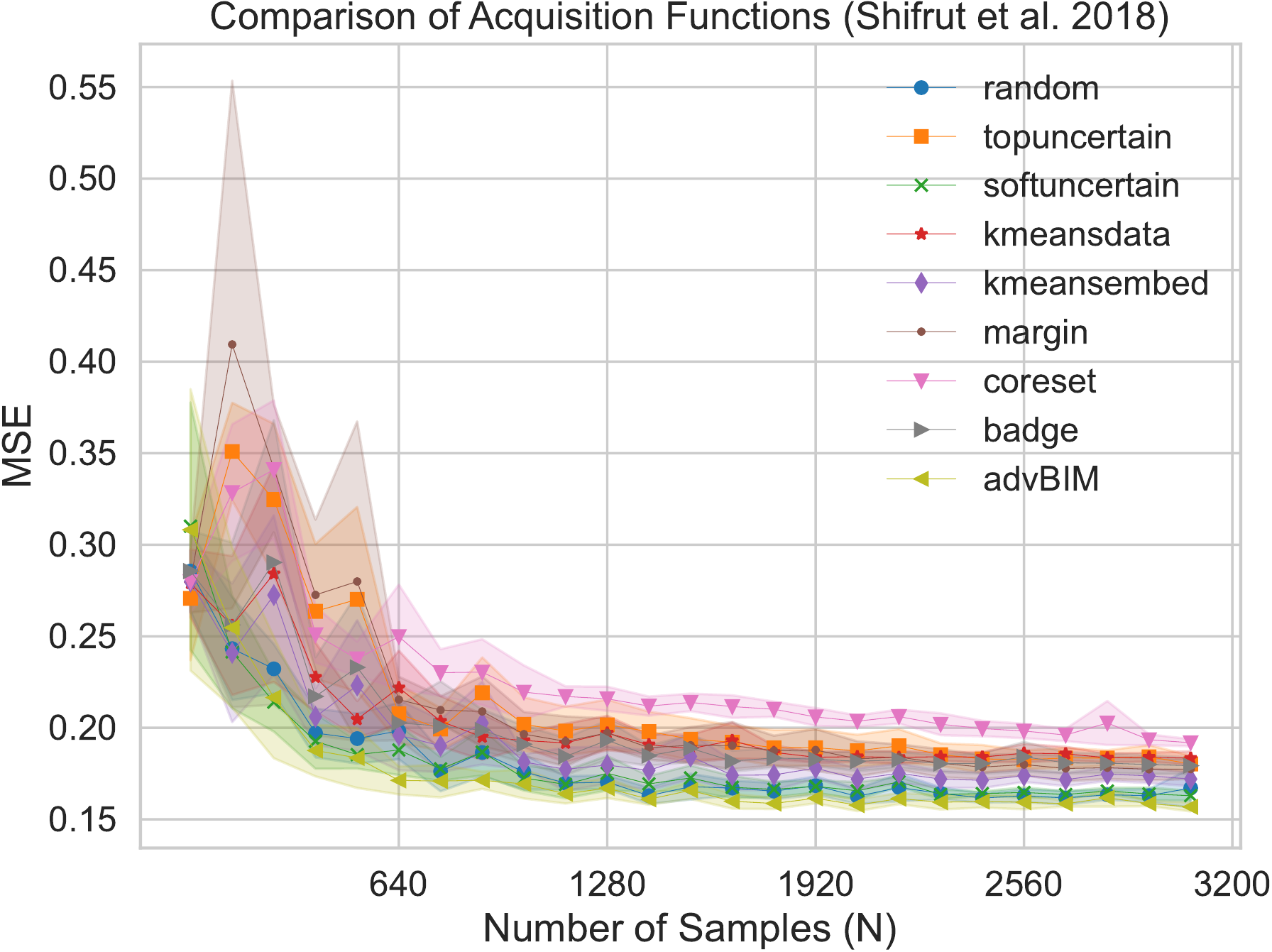}};
                        \end{tikzpicture}
                    }
                \end{subfigure}
                \&
                \begin{subfigure}{0.27\columnwidth}
                    \hspace{-23mm}
                    \centering
                    \resizebox{\linewidth}{!}{
                        \begin{tikzpicture}
                            \node (img)  {\includegraphics[width=\textwidth]{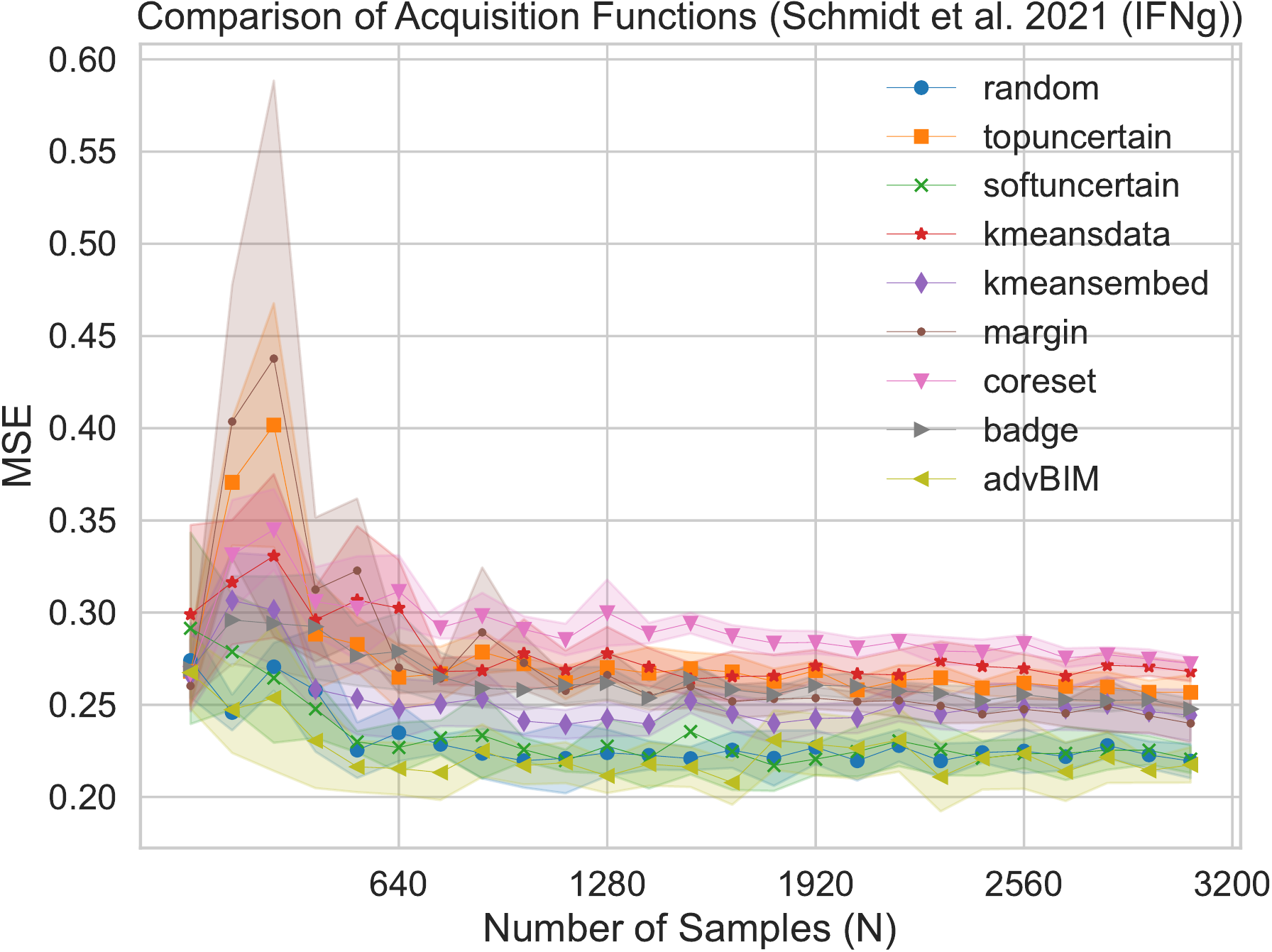}};
                        \end{tikzpicture}
                    }
                \end{subfigure}
                \&
                \begin{subfigure}{0.28\columnwidth}
                    \hspace{-28mm}
                    \centering
                    \resizebox{\linewidth}{!}{
                        \begin{tikzpicture}
                            \node (img)  {\includegraphics[width=\textwidth]{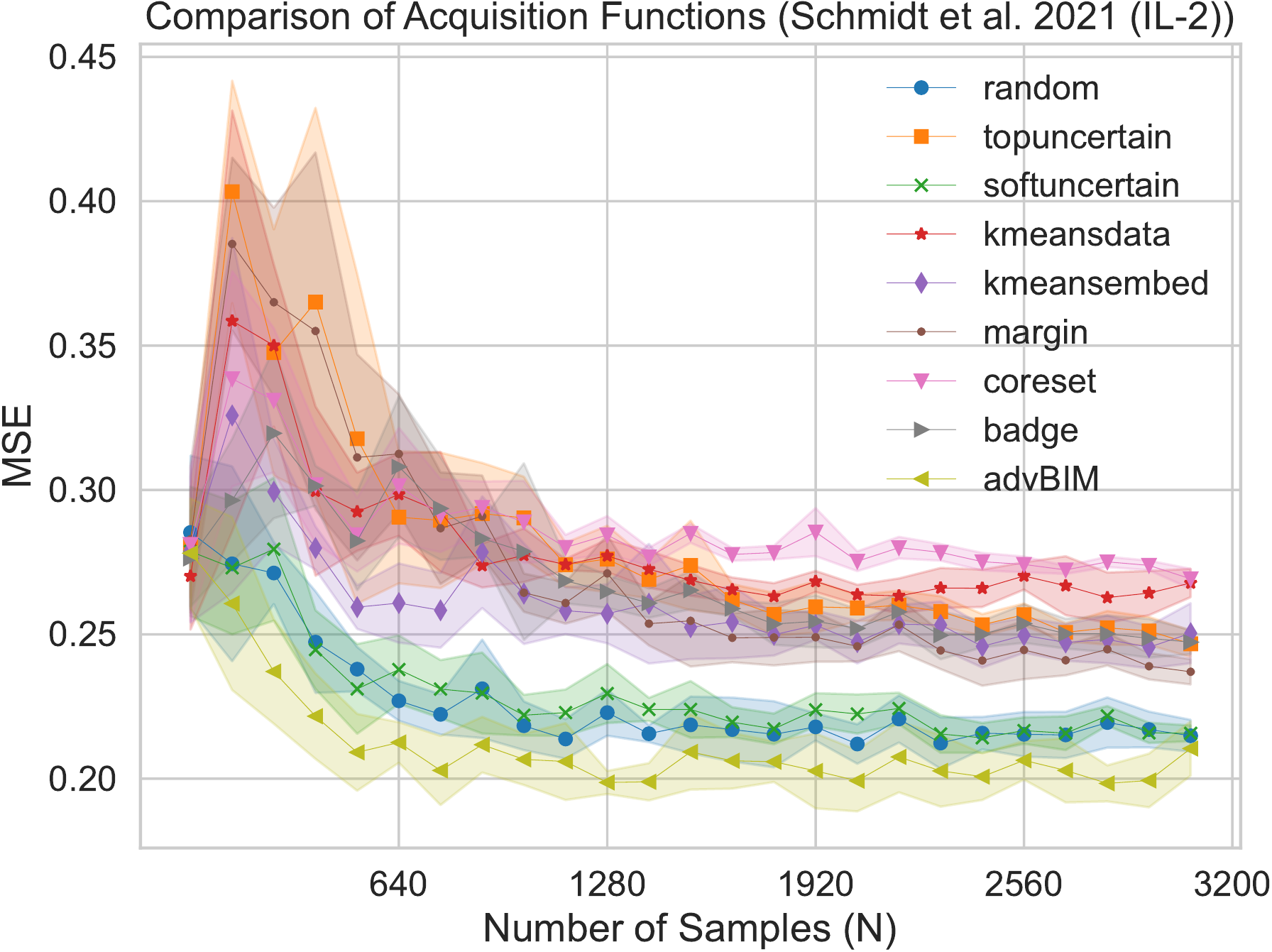}};
                        \end{tikzpicture}
                    }
                \end{subfigure}
                \&
                \begin{subfigure}{0.29\columnwidth}
                    \hspace{-32mm}
                    \centering
                    \resizebox{\linewidth}{!}{
                        \begin{tikzpicture}
                            \node (img)  {\includegraphics[width=\textwidth]{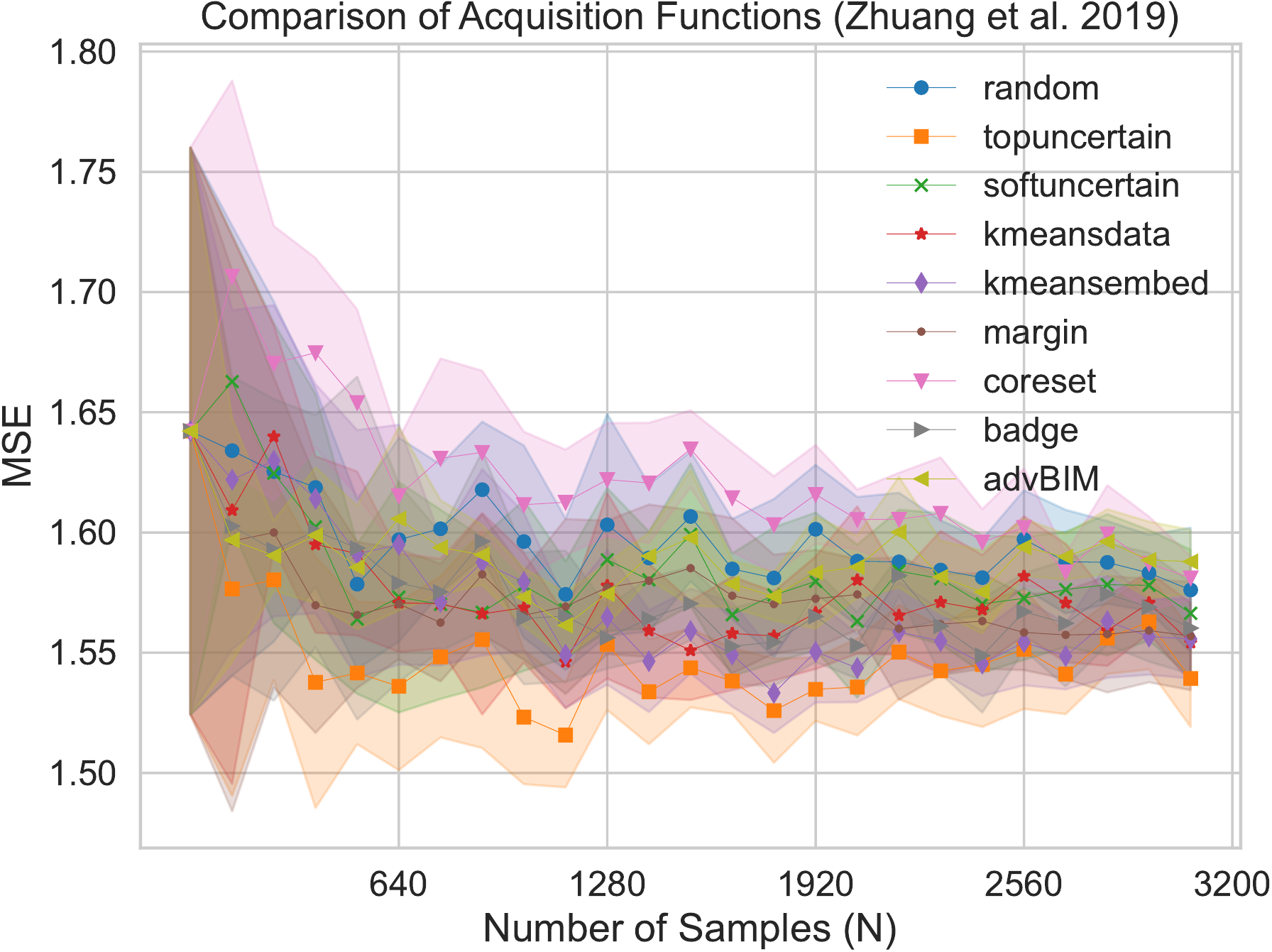}};
                        \end{tikzpicture}
                    }
                \end{subfigure}
                \&
                \\
\begin{subfigure}{0.275\columnwidth}
                    \hspace{-17mm}
                    \centering
                    \resizebox{\linewidth}{!}{
                        \begin{tikzpicture}
                            \node (img)  {\includegraphics[width=\textwidth]{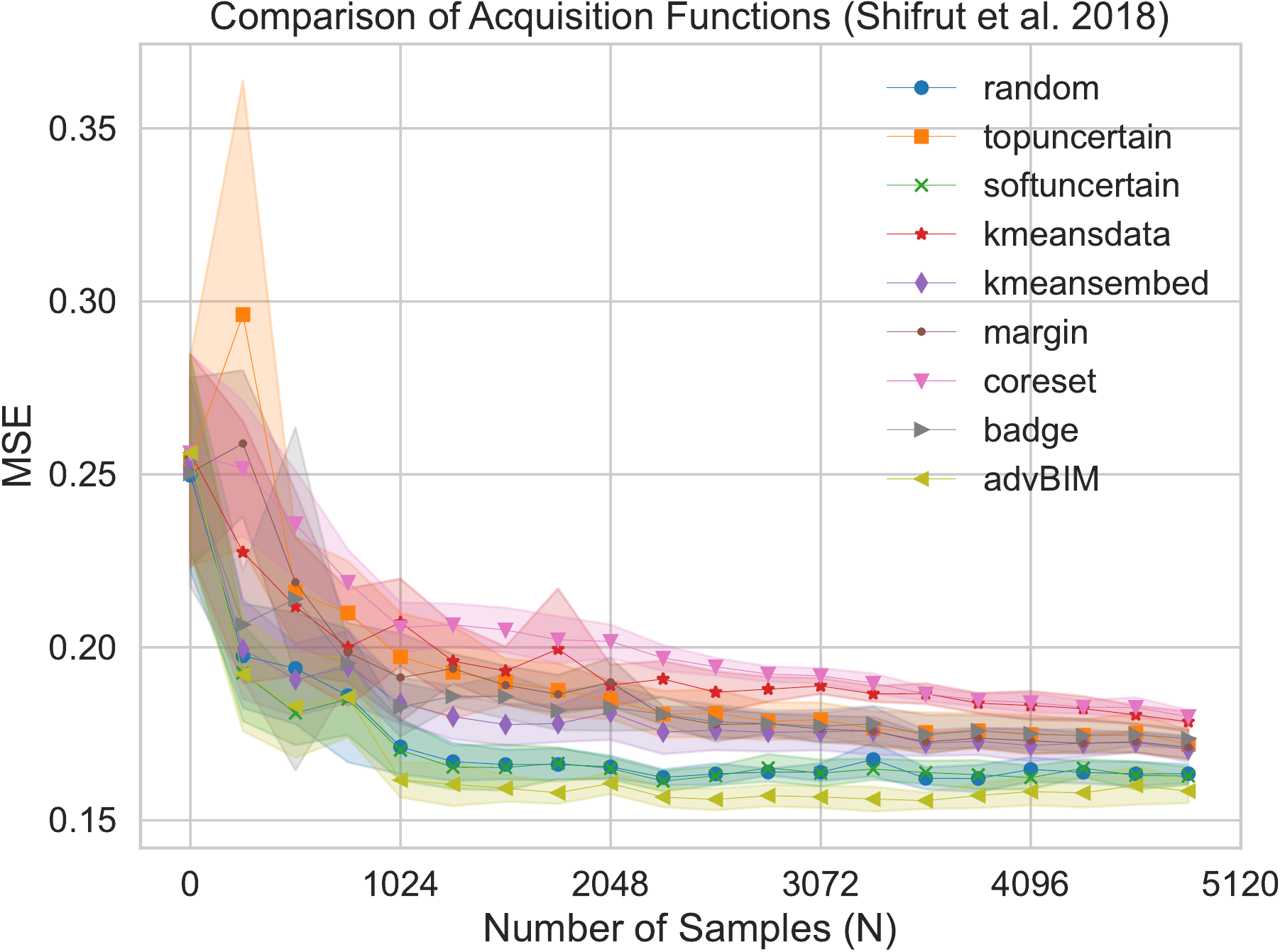}};
                        \end{tikzpicture}
                    }
                \end{subfigure}
                \&
                \begin{subfigure}{0.27\columnwidth}
                    \hspace{-23mm}
                    \centering
                    \resizebox{\linewidth}{!}{
                        \begin{tikzpicture}
                            \node (img)  {\includegraphics[width=\textwidth]{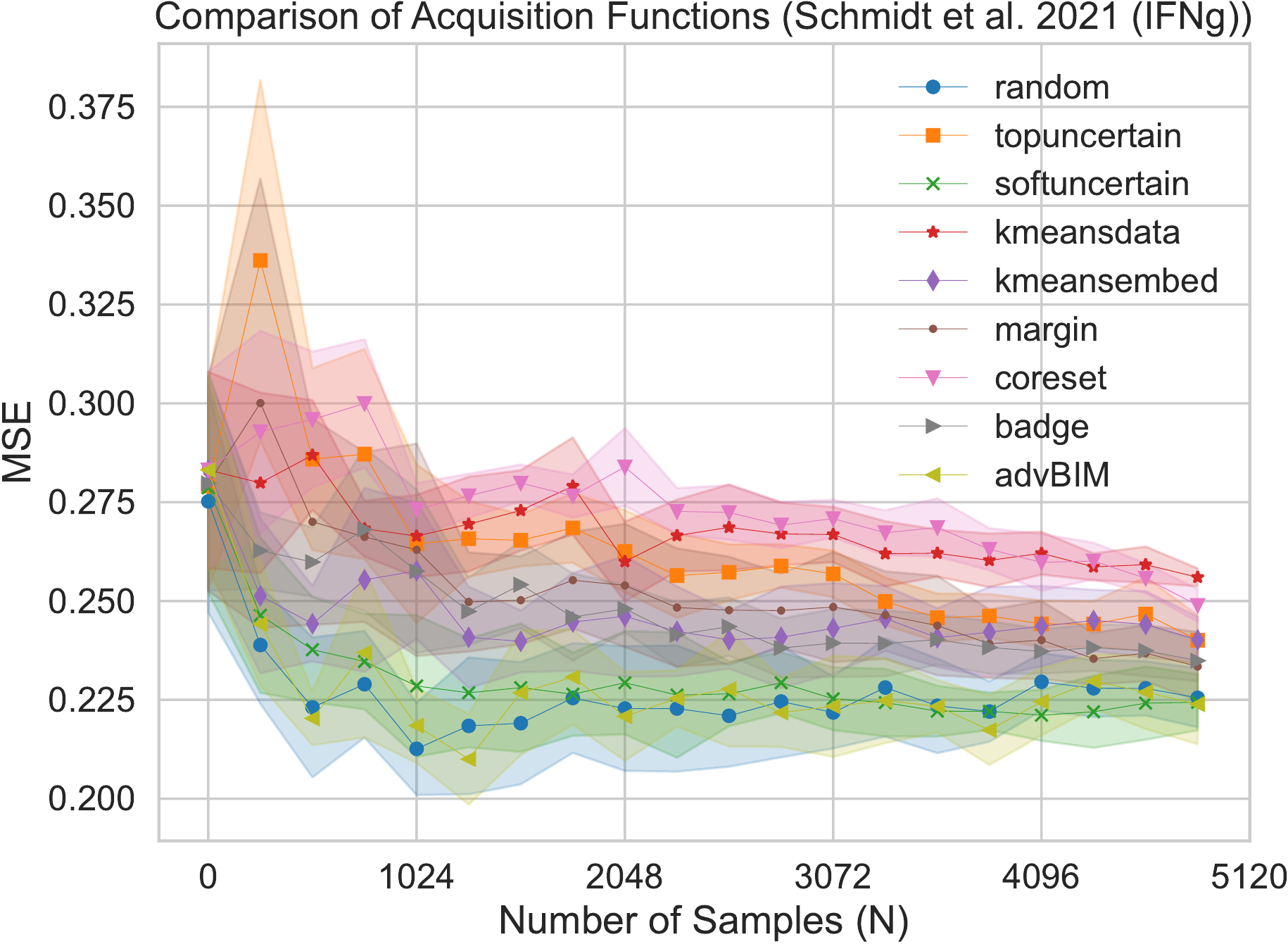}};
                        \end{tikzpicture}
                    }
                \end{subfigure}
                \&
                \begin{subfigure}{0.27\columnwidth}
                    \hspace{-28mm}
                    \centering
                    \resizebox{\linewidth}{!}{
                        \begin{tikzpicture}
                            \node (img)  {\includegraphics[width=\textwidth]{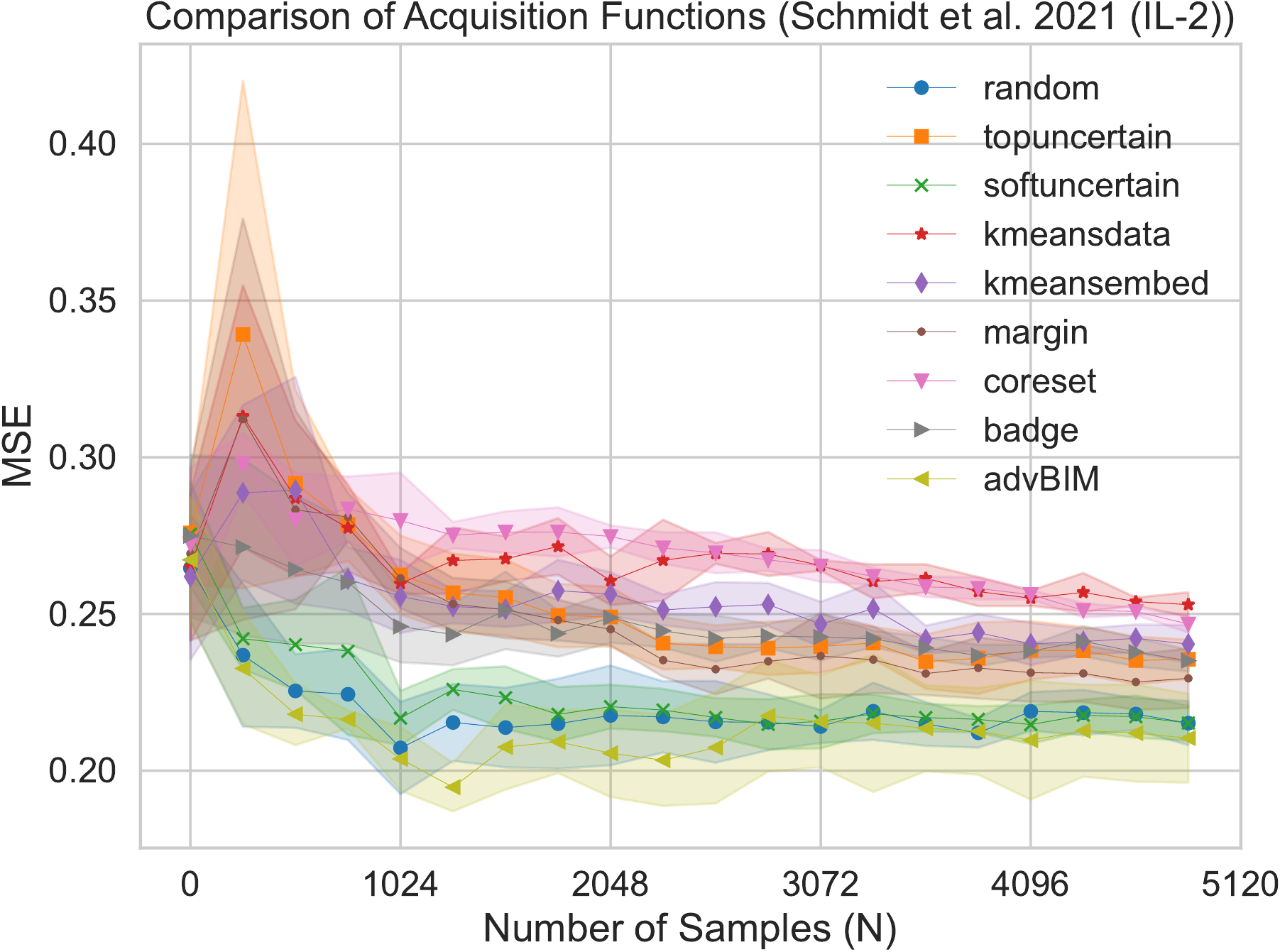}};
                        \end{tikzpicture}
                    }
                \end{subfigure}
                \&
                \begin{subfigure}{0.29\columnwidth}
                    \hspace{-32mm}
                    \centering
                    \resizebox{\linewidth}{!}{
                        \begin{tikzpicture}
                            \node (img)  {\includegraphics[width=\textwidth]{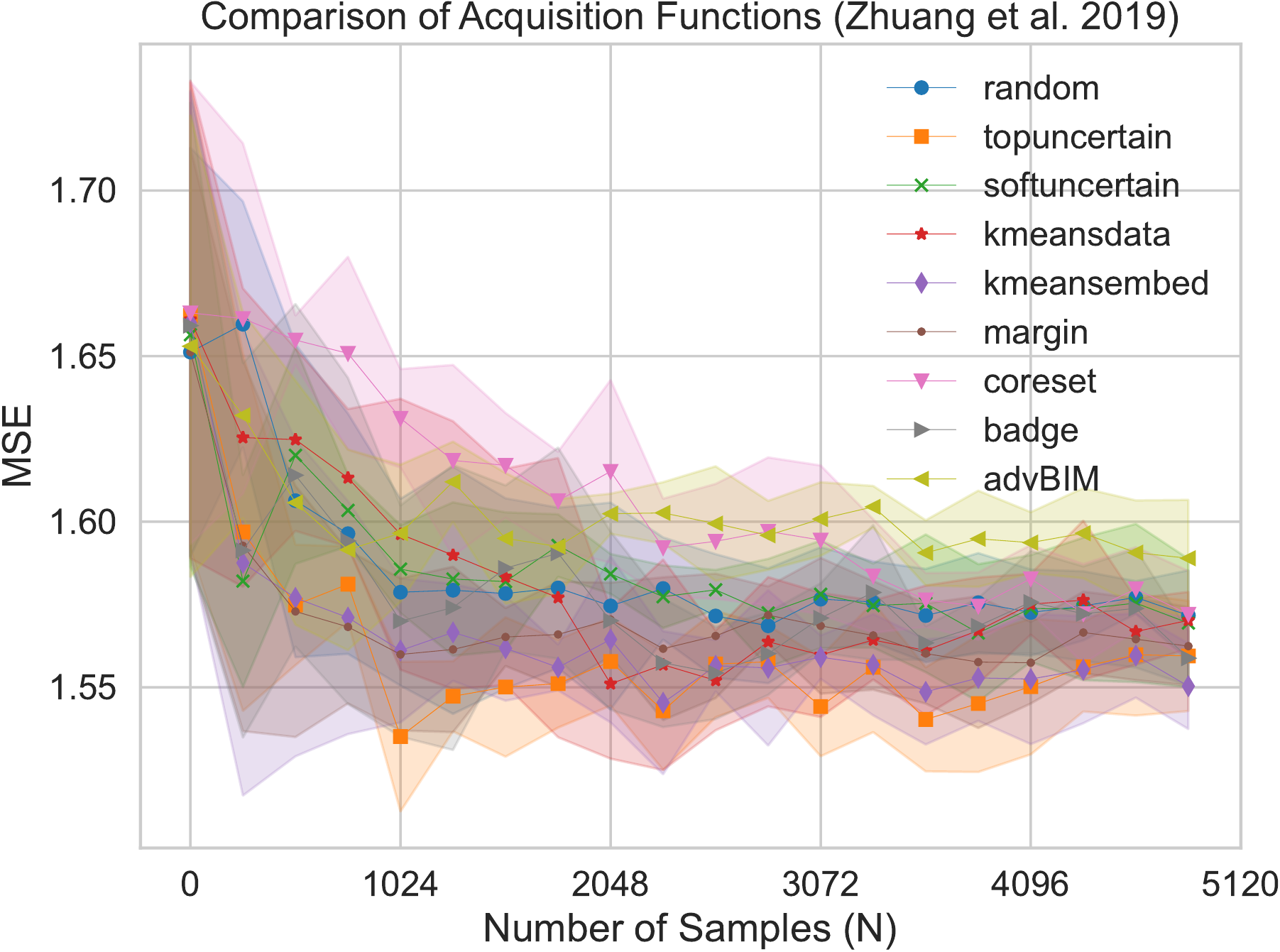}};
                        \end{tikzpicture}
                    }
                \end{subfigure}
                \&
                \\
\begin{subfigure}{0.28\columnwidth}
                    \hspace{-17mm}
                    \centering
                    \resizebox{\linewidth}{!}{
                        \begin{tikzpicture}
                            \node (img)  {\includegraphics[width=\textwidth]{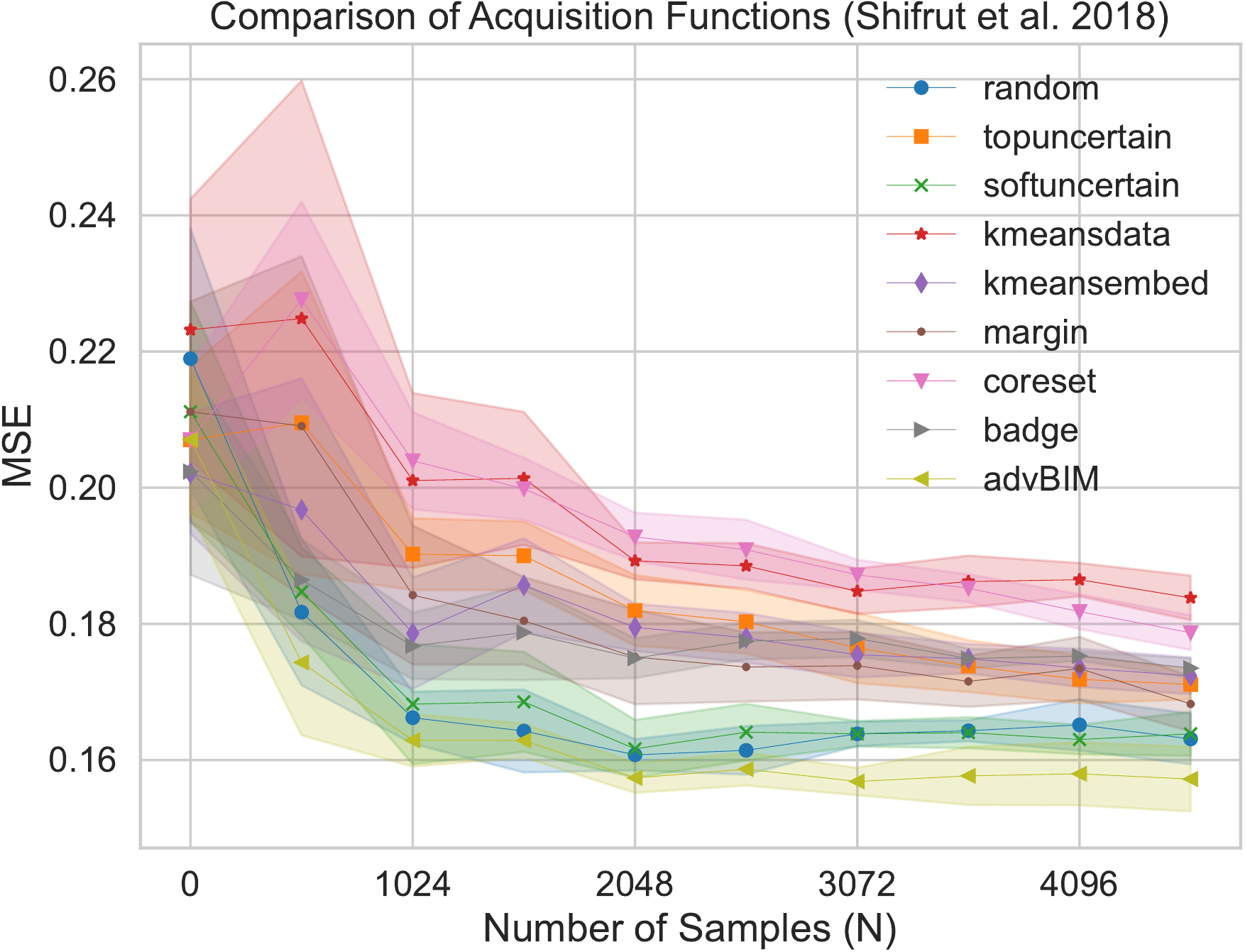}};
                        \end{tikzpicture}
                    }
                \end{subfigure}
                \&
                \begin{subfigure}{0.27\columnwidth}
                    \hspace{-23mm}
                    \centering
                    \resizebox{\linewidth}{!}{
                        \begin{tikzpicture}
                            \node (img)  {\includegraphics[width=\textwidth]{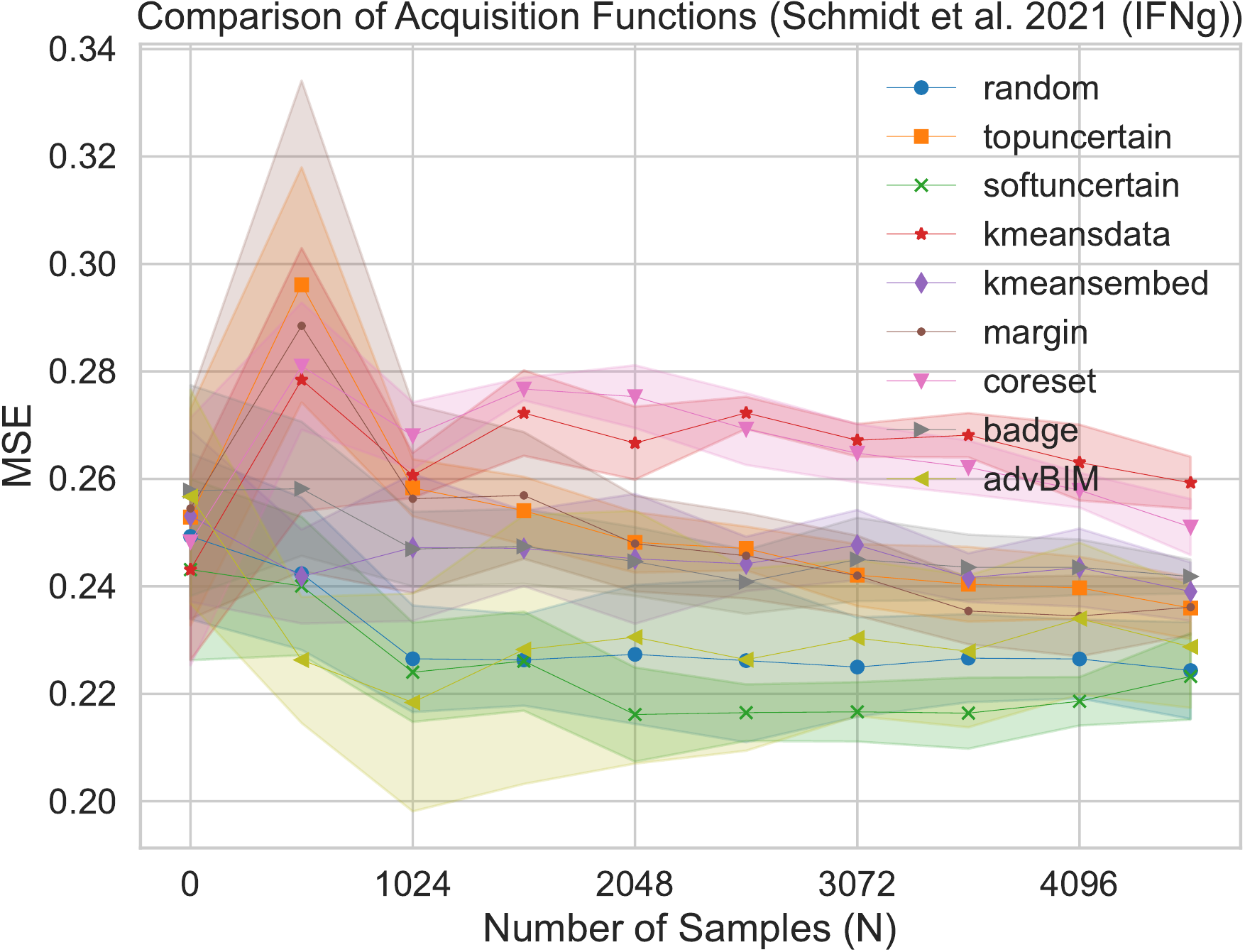}};
                        \end{tikzpicture}
                    }
                \end{subfigure}
                \&
                \begin{subfigure}{0.27\columnwidth}
                    \hspace{-28mm}
                    \centering
                    \resizebox{\linewidth}{!}{
                        \begin{tikzpicture}
                            \node (img)  {\includegraphics[width=\textwidth]{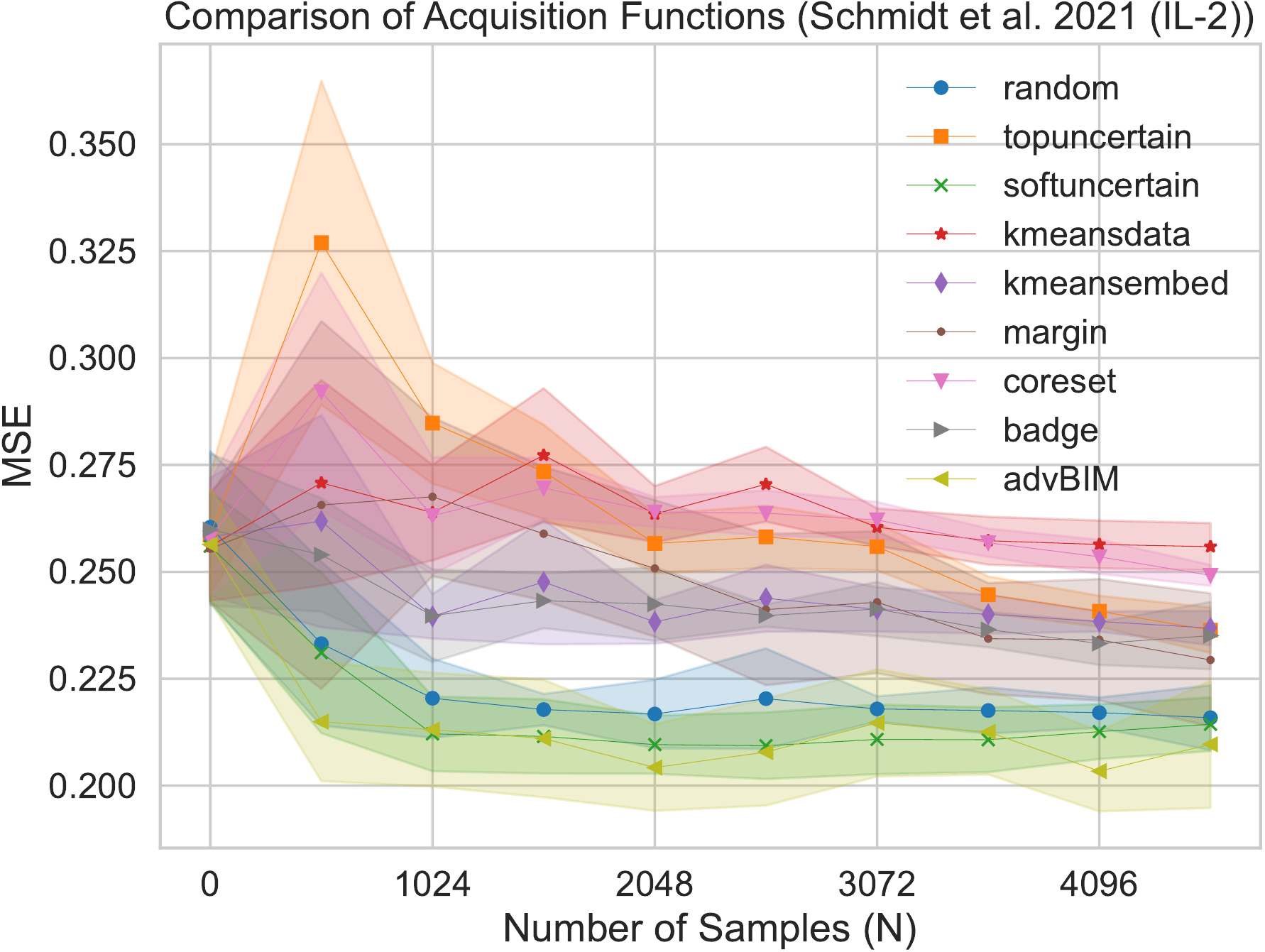}};
                        \end{tikzpicture}
                    }
                \end{subfigure}
                \&
                \begin{subfigure}{0.28\columnwidth}
                    \hspace{-32mm}
                    \centering
                    \resizebox{\linewidth}{!}{
                        \begin{tikzpicture}
                            \node (img)  {\includegraphics[width=\textwidth]{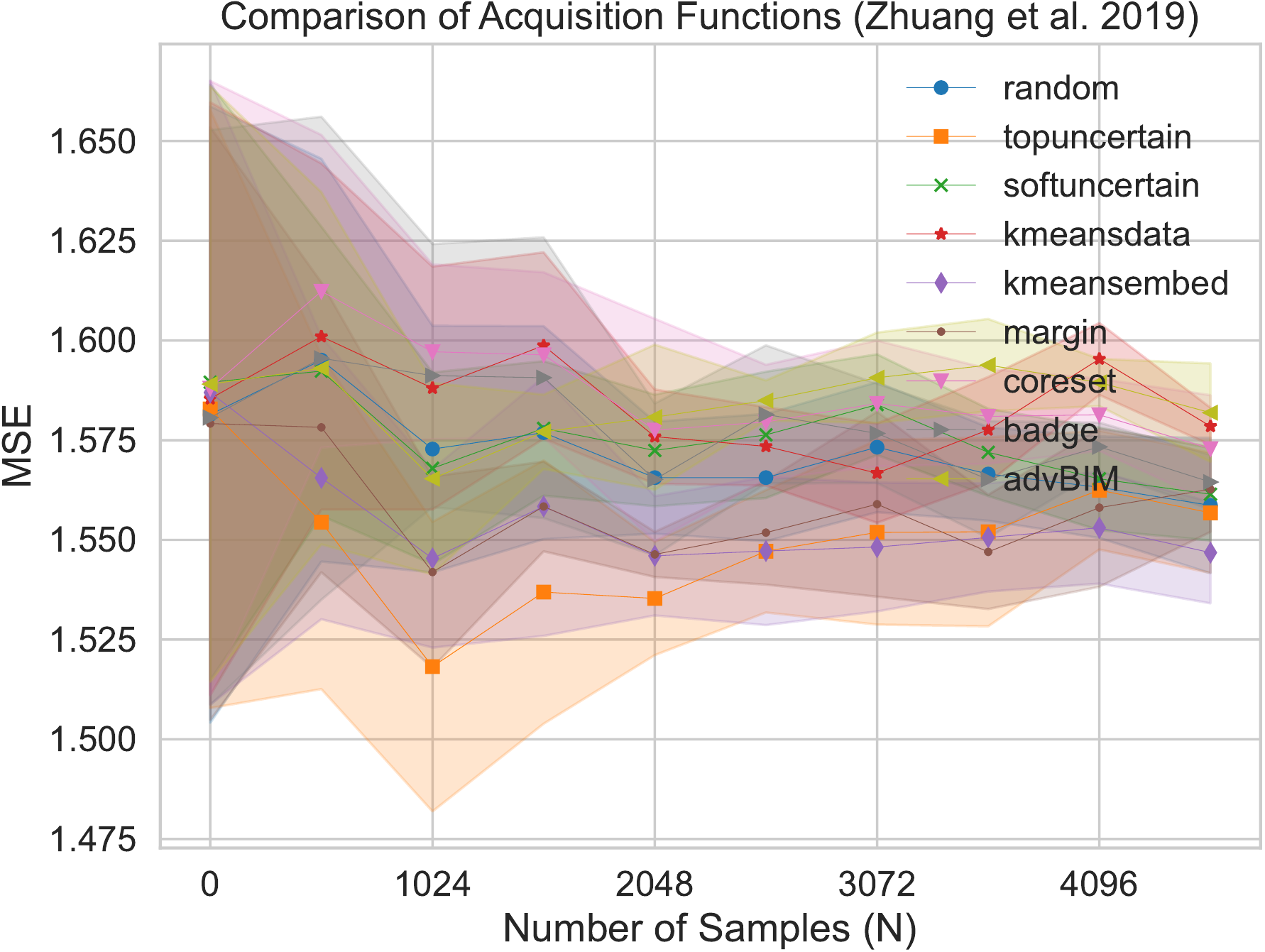}};
                        \end{tikzpicture}
                    }
                \end{subfigure}
                \&
                \\
            \\
           
            \\
            };
            \node [draw=none, rotate=90] at ([xshift=-8mm, yshift=2mm]fig-1-1.west) {\small batch size = 16};
            \node [draw=none, rotate=90] at ([xshift=-8mm, yshift=2mm]fig-2-1.west) {\small batch size = 32};
            \node [draw=none, rotate=90] at ([xshift=-8mm, yshift=2mm]fig-3-1.west) {\small batch size = 64};
            \node [draw=none, rotate=90] at ([xshift=-8mm, yshift=2mm]fig-4-1.west) {\small batch size = 128};
            \node [draw=none, rotate=90] at ([xshift=-8mm, yshift=2mm]fig-5-1.west) {\small batch size = 256};
            \node [draw=none, rotate=90] at ([xshift=-8mm, yshift=2mm]fig-6-1.west) {\small batch size = 512};
            \node [draw=none] at ([xshift=-6mm, yshift=3mm]fig-1-1.north) {\small Shifrut et al. 2018};
            \node [draw=none] at ([xshift=-9mm, yshift=3mm]fig-1-2.north) {\small Schmidt et al. 2021 (IFNg)};
            \node [draw=none] at ([xshift=-11mm, yshift=3mm]fig-1-3.north) {\small Schmidt et al. 2021 (IL-2)};
            \node [draw=none] at ([xshift=-13mm, yshift=2.5mm]fig-1-4.north) {\small Zhuang et al. 2019};
\end{tikzpicture}}
        \vspace{-7mm}
        \caption{The evaluation of the model trained with {Achilles} treatment descriptors at each active learning cycle for 4 datasets and 6 acquisition batch sizes. In each plot, the x-axis is the active learning cycles multiplied by the acquisition bath size that gives the total number of data points collected so far. The y-axis is the test MSE error evaluated on the test data.}
        \vspace{-5mm}
        \label{fig:bnn_feat_achilles_alldatasets_allbathcsizes}
    \end{figure*}

\begin{figure*}
    \vspace{-2mm}
        \centering
        \makebox[0.72\paperwidth]{\begin{tikzpicture}[ampersand replacement=\&]
            \matrix (fig) [matrix of nodes]{ 
\begin{subfigure}{0.27\columnwidth}
                    \hspace{-17mm}
                    \centering
                    \resizebox{\linewidth}{!}{
                        \begin{tikzpicture}
                            \node (img)  {\includegraphics[width=\textwidth]{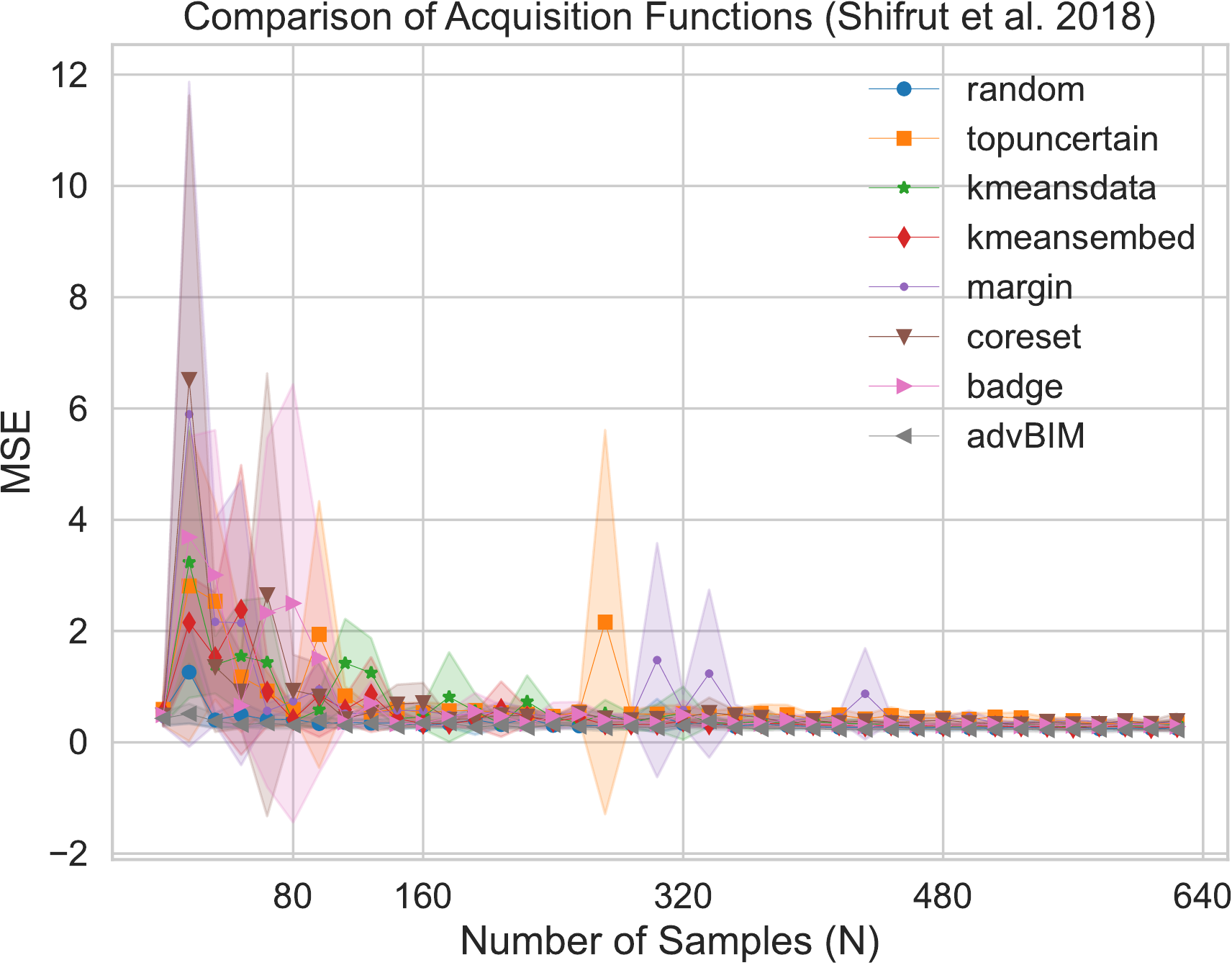}};
                        \end{tikzpicture}
                    }
                \end{subfigure}
                \&
                 \begin{subfigure}{0.27\columnwidth}
                    \hspace{-23mm}
                    \centering
                    \resizebox{\linewidth}{!}{
                        \begin{tikzpicture}
                            \node (img)  {\includegraphics[width=\textwidth]{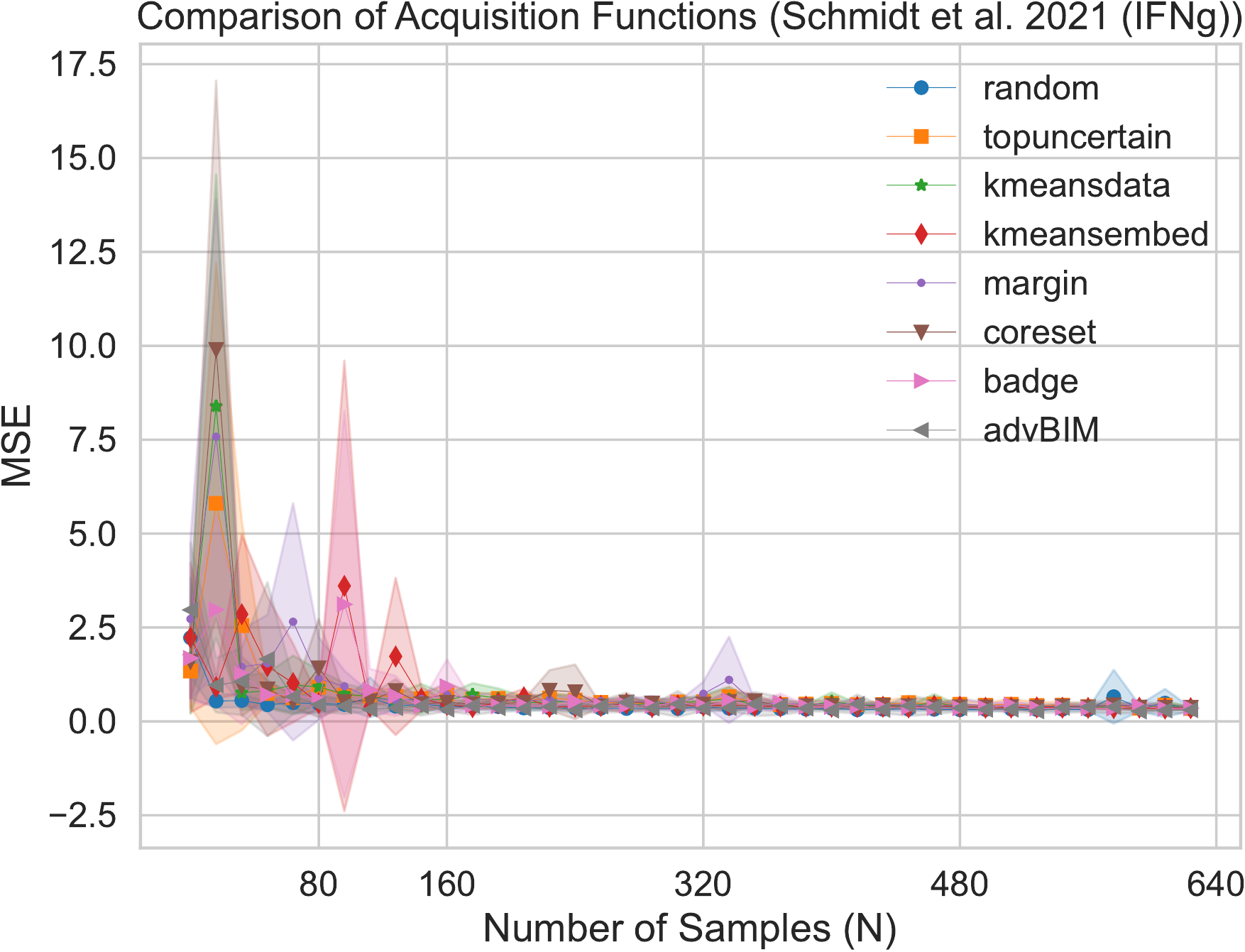}};
                        \end{tikzpicture}
                    }
                \end{subfigure}
                \&
                 \begin{subfigure}{0.27\columnwidth}
                    \hspace{-28mm}
                    \centering
                    \resizebox{\linewidth}{!}{
                        \begin{tikzpicture}
                            \node (img)  {\includegraphics[width=\textwidth]{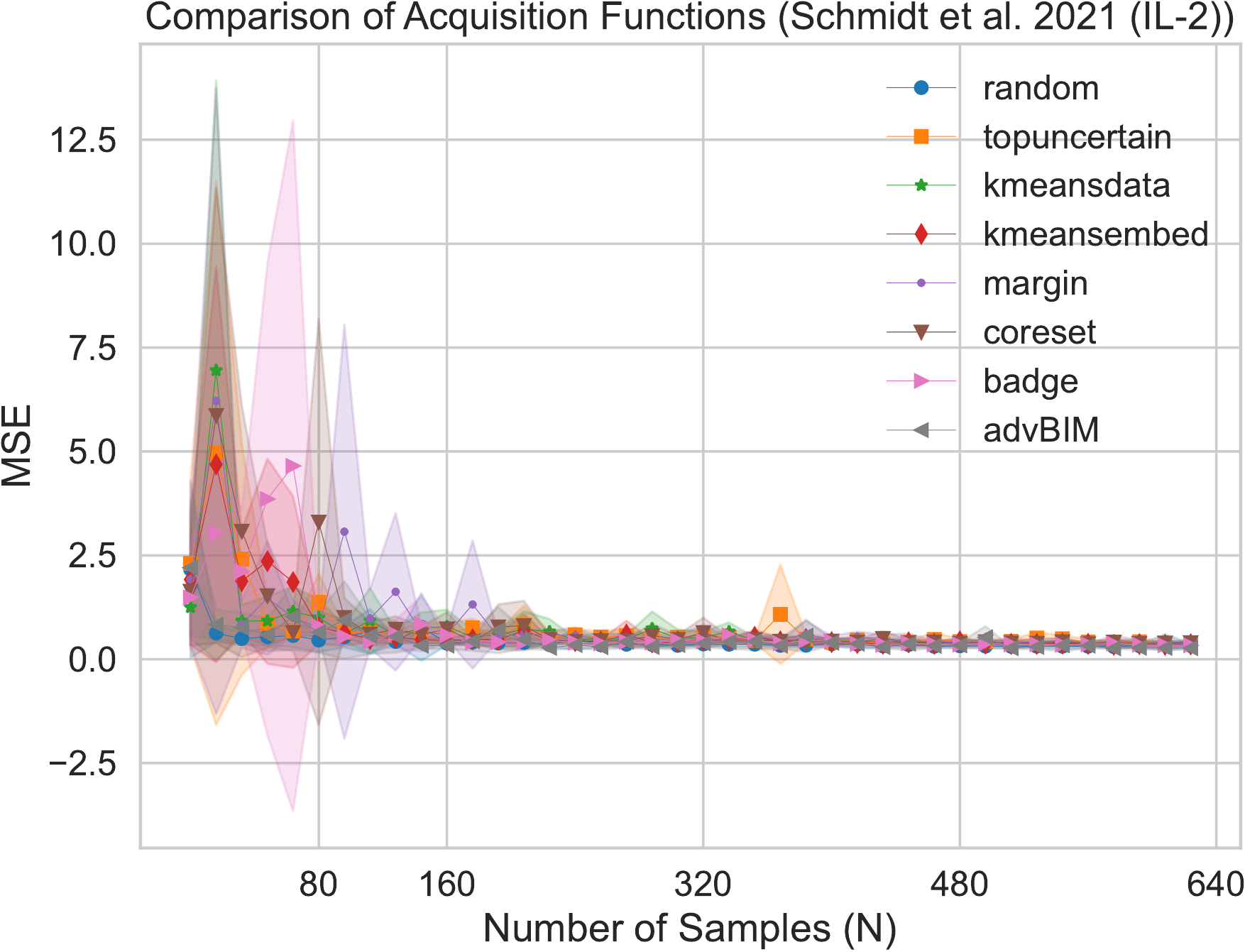}};
                        \end{tikzpicture}
                    }
                \end{subfigure}
                \&
                \begin{subfigure}{0.28\columnwidth}
                    \hspace{-32mm}
                    \centering
                    \resizebox{\linewidth}{!}{
                        \begin{tikzpicture}
                            \node (img)  {\includegraphics[width=\textwidth]{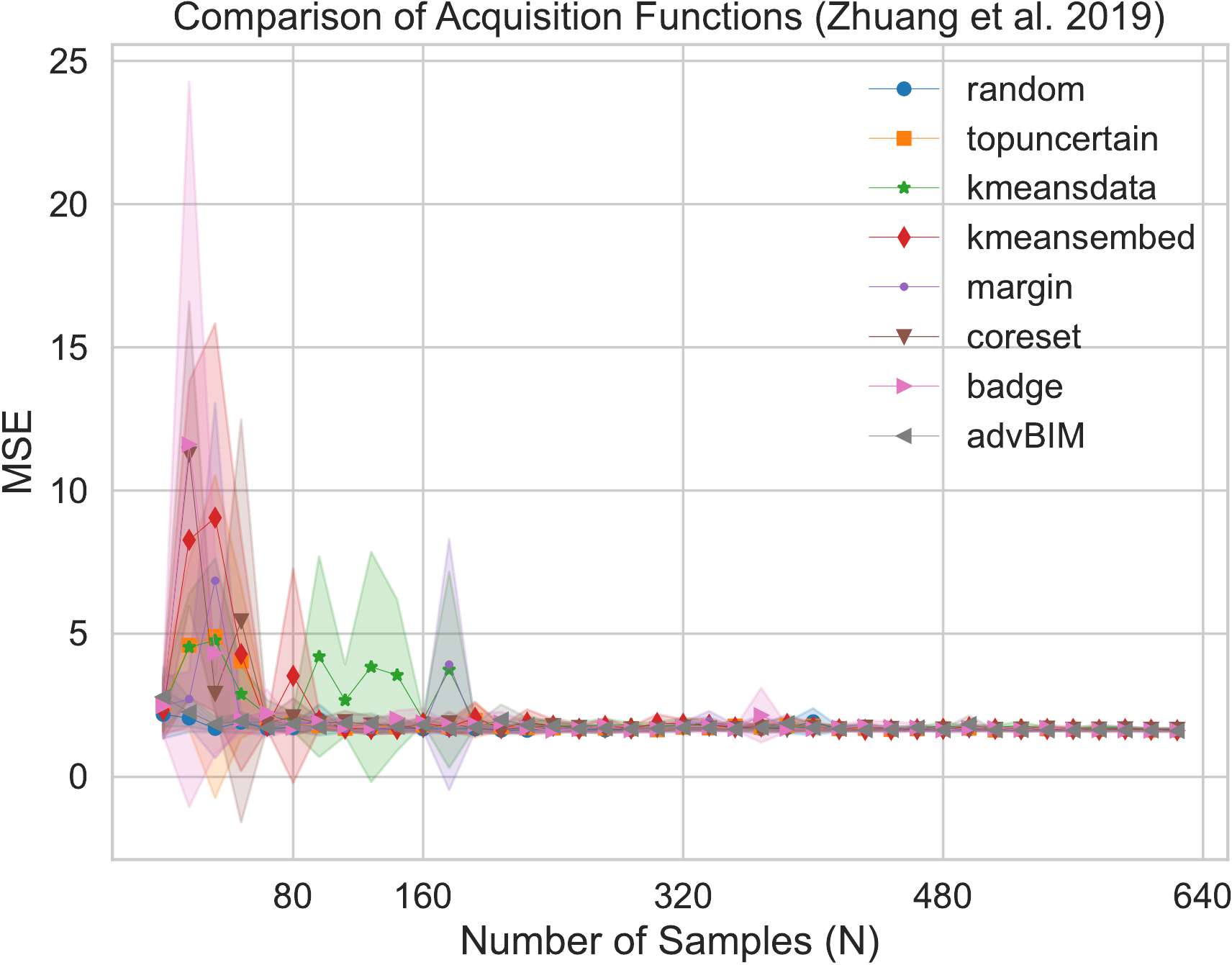}};
                        \end{tikzpicture}
                    }
                \end{subfigure}
                \&
            \\
\begin{subfigure}{0.27\columnwidth}
                    \hspace{-17mm}
                    \centering
                    \resizebox{\linewidth}{!}{
                        \begin{tikzpicture}
                            \node (img)  {\includegraphics[width=\textwidth]{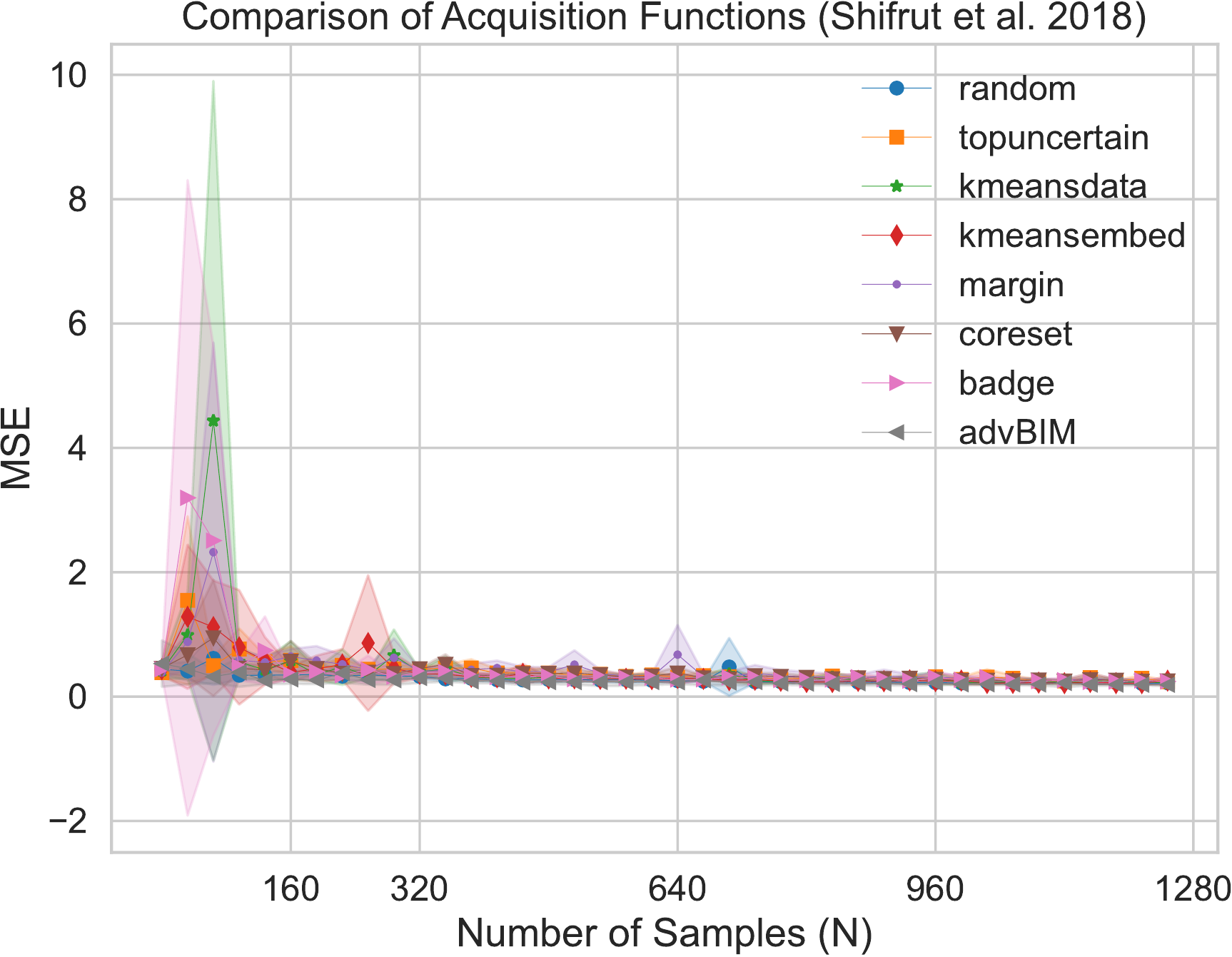}};
                        \end{tikzpicture}
                    }
                \end{subfigure}
                \&
                \begin{subfigure}{0.27\columnwidth}
                    \hspace{-23mm}
                    \centering
                    \resizebox{\linewidth}{!}{
                        \begin{tikzpicture}
                            \node (img)  {\includegraphics[width=\textwidth]{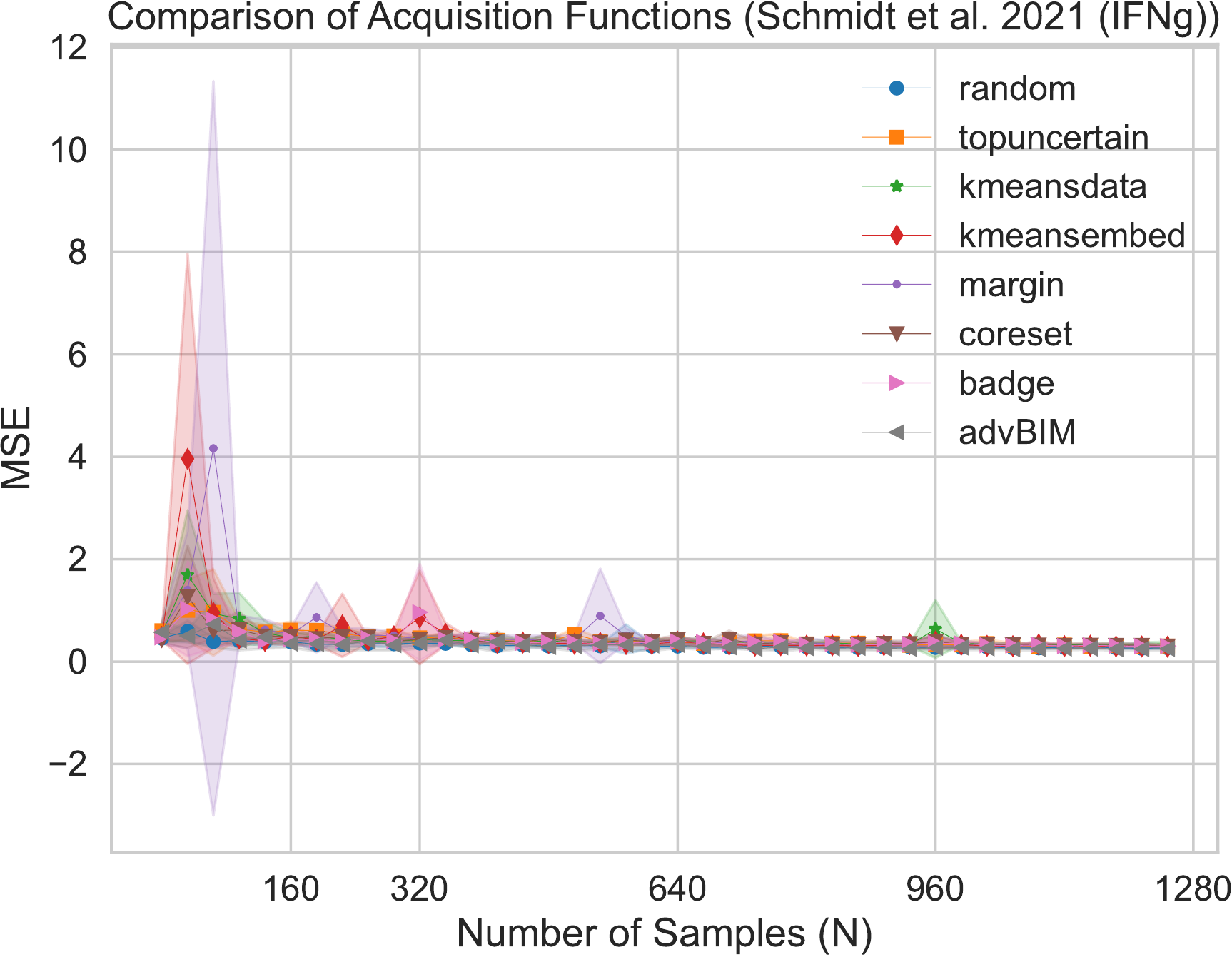}};
                        \end{tikzpicture}
                    }
                \end{subfigure}
                \&
                \begin{subfigure}{0.27\columnwidth}
                    \hspace{-28mm}
                    \centering
                    \resizebox{\linewidth}{!}{
                        \begin{tikzpicture}
                            \node (img)  {\includegraphics[width=\textwidth]{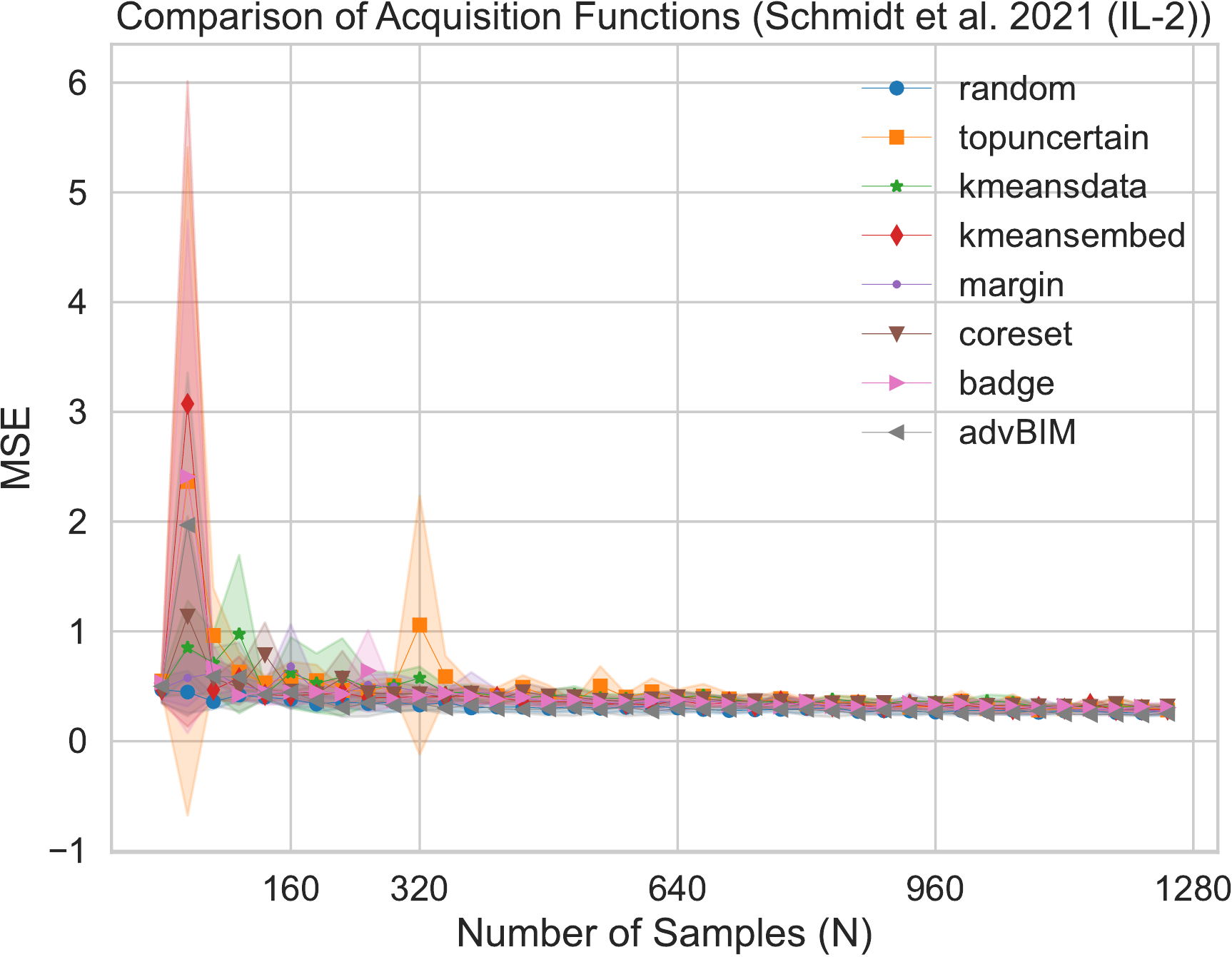}};
                        \end{tikzpicture}
                    }
                \end{subfigure}
                \&
                \begin{subfigure}{0.28\columnwidth}
                    \hspace{-32mm}
                    \centering
                    \resizebox{\linewidth}{!}{
                        \begin{tikzpicture}
                            \node (img)  {\includegraphics[width=\textwidth]{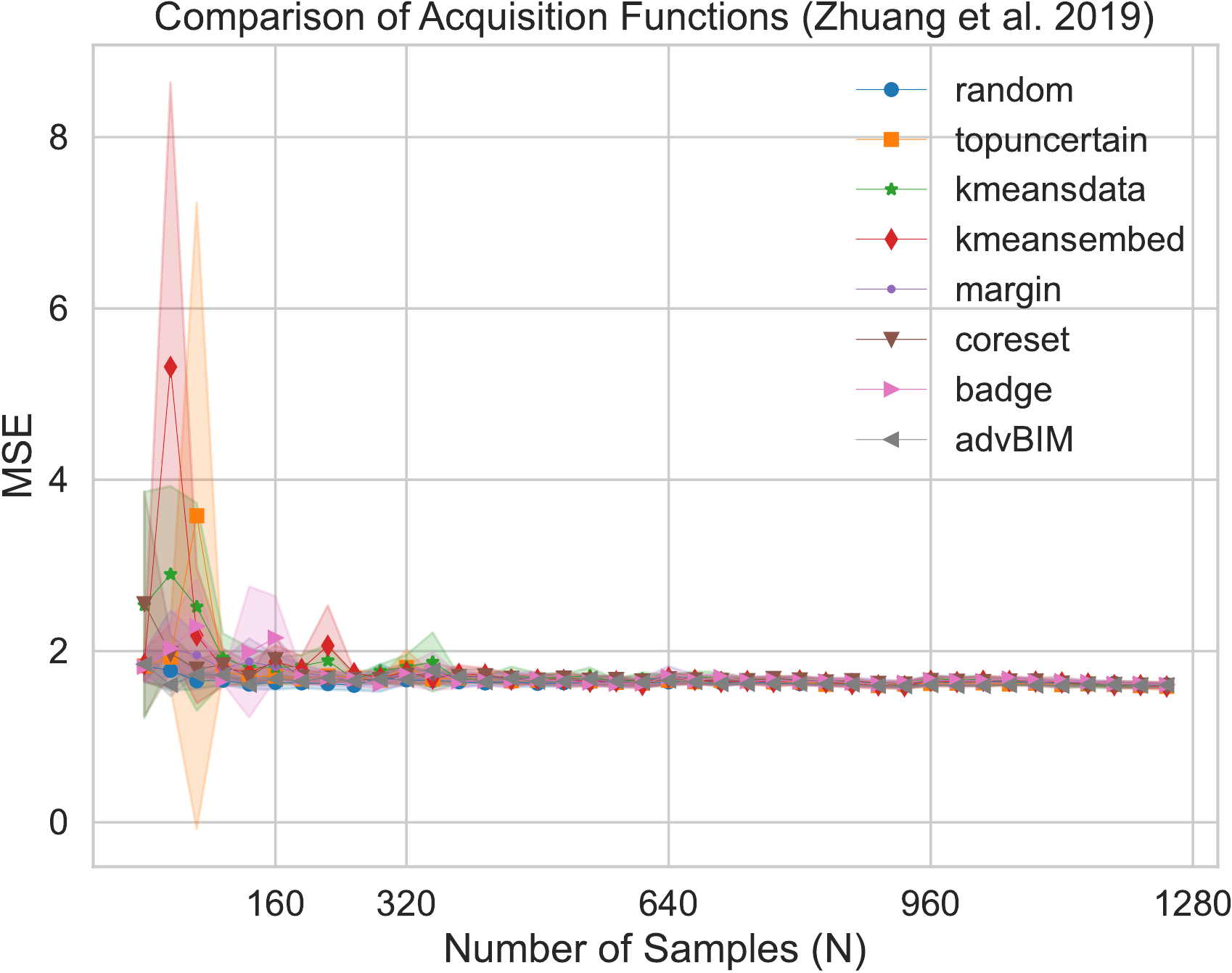}};
                        \end{tikzpicture}
                    }
                \end{subfigure}
                \&
                \\
\begin{subfigure}{0.27\columnwidth}
                    \hspace{-17mm}
                    \centering
                    \resizebox{\linewidth}{!}{
                        \begin{tikzpicture}
                            \node (img)  {\includegraphics[width=\textwidth]{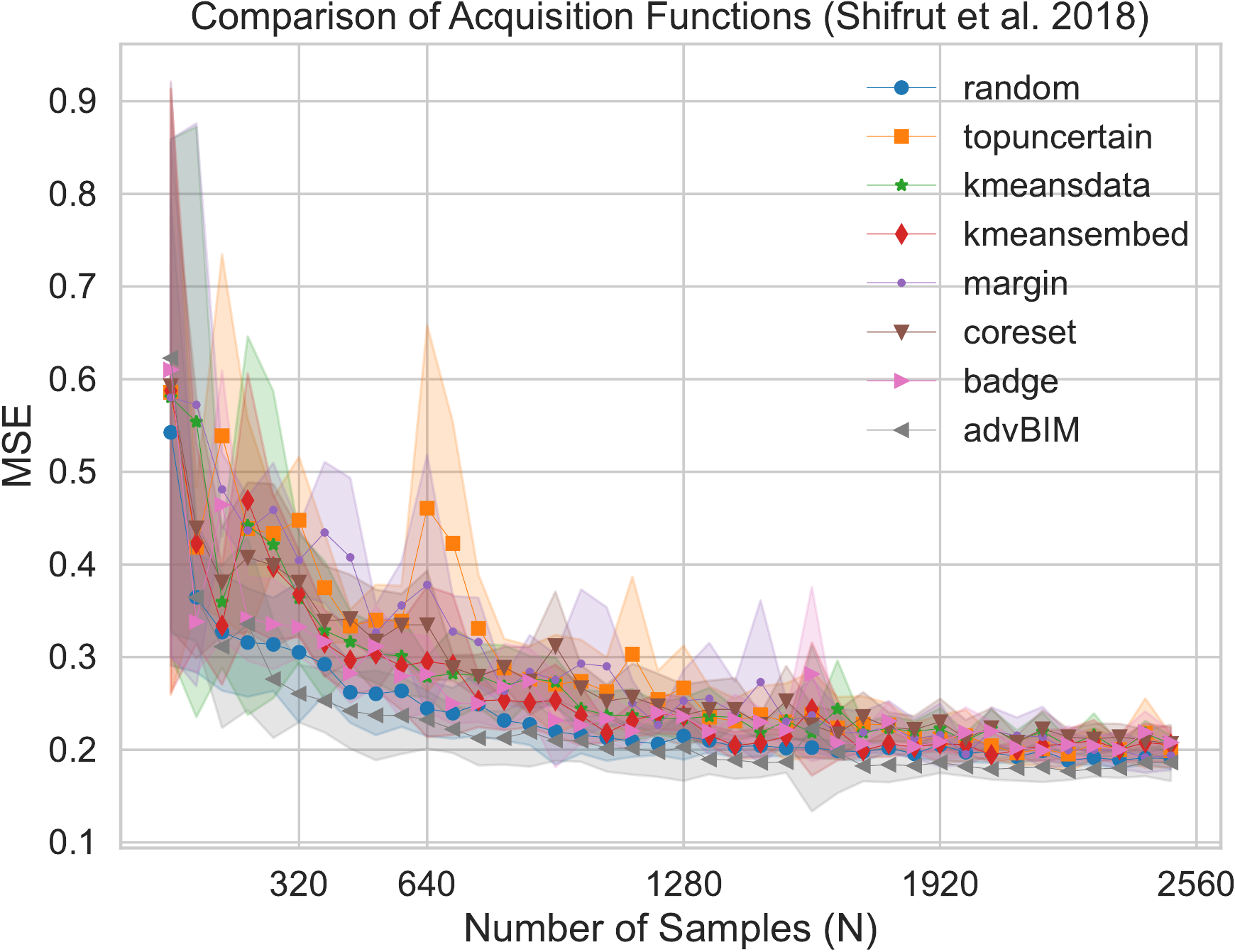}};
                        \end{tikzpicture}
                    }
                \end{subfigure}
                \&
                \begin{subfigure}{0.27\columnwidth}
                    \hspace{-23mm}
                    \centering
                    \resizebox{\linewidth}{!}{
                        \begin{tikzpicture}
                            \node (img)  {\includegraphics[width=\textwidth]{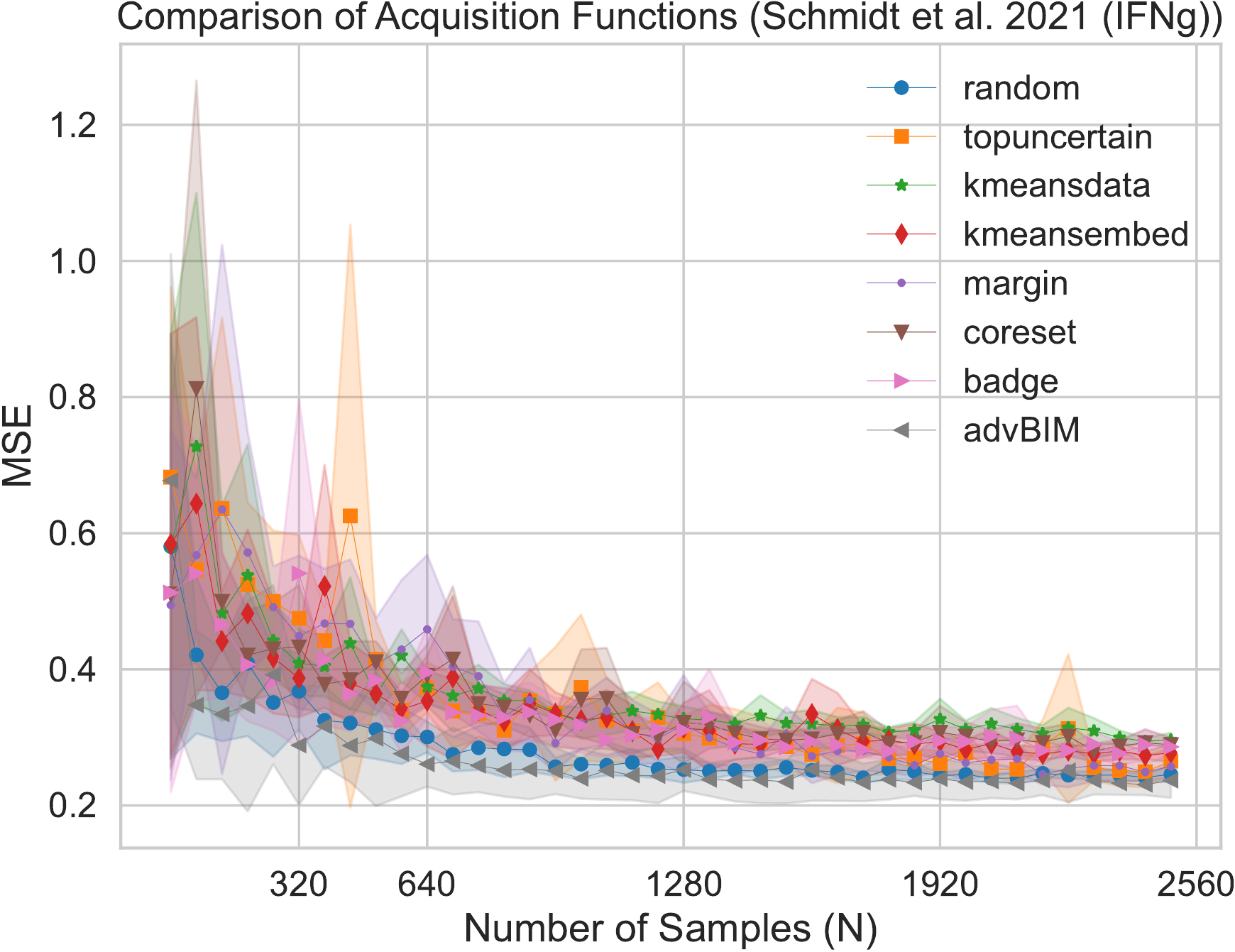}};
                        \end{tikzpicture}
                    }
                \end{subfigure}
                \&
                \begin{subfigure}{0.27\columnwidth}
                    \hspace{-28mm}
                    \centering
                    \resizebox{\linewidth}{!}{
                        \begin{tikzpicture}
                            \node (img)  {\includegraphics[width=\textwidth]{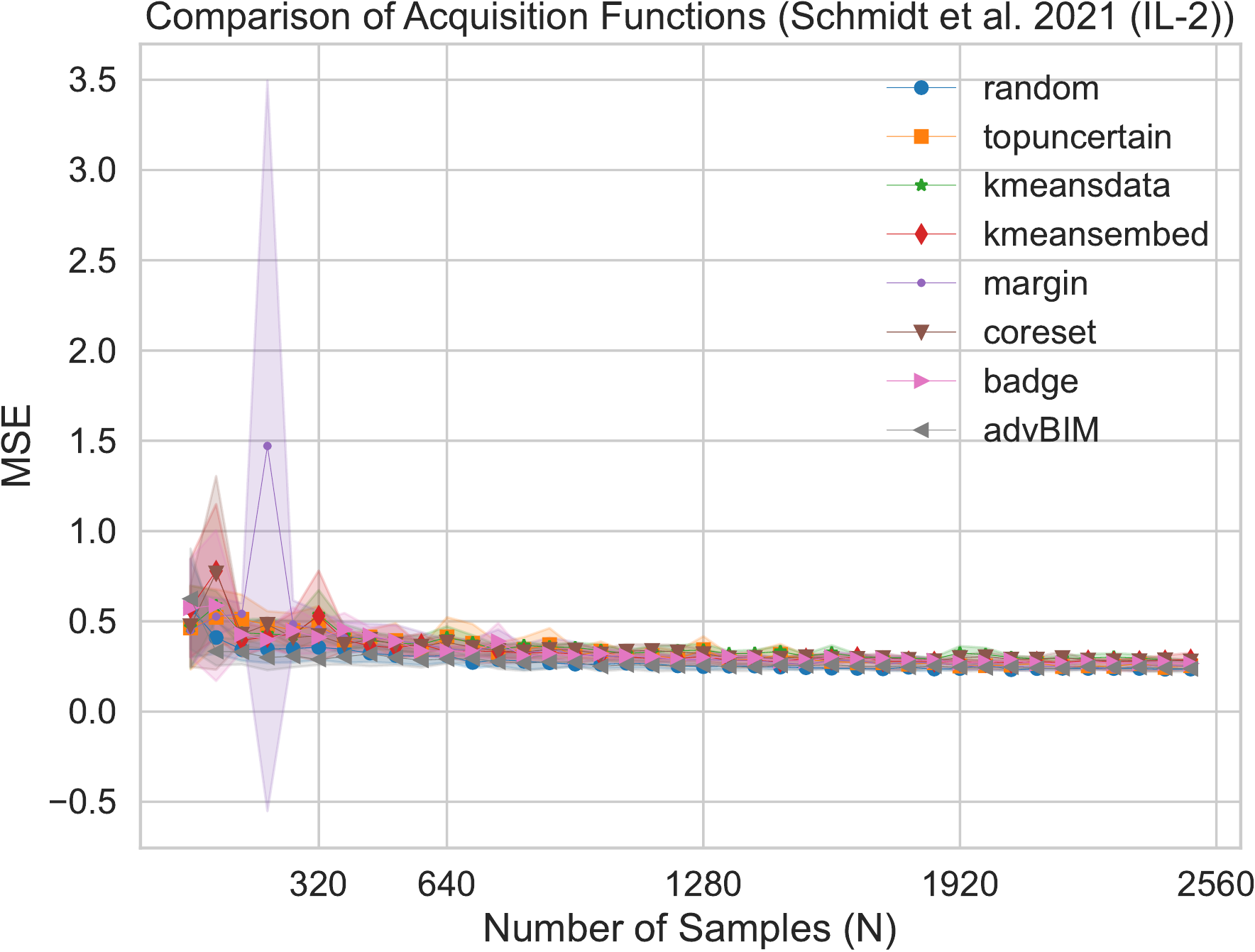}};
                        \end{tikzpicture}
                    }
                \end{subfigure}
                \&
                \begin{subfigure}{0.28\columnwidth}
                    \hspace{-32mm}
                    \centering
                    \resizebox{\linewidth}{!}{
                        \begin{tikzpicture}
                            \node (img)  {\includegraphics[width=\textwidth]{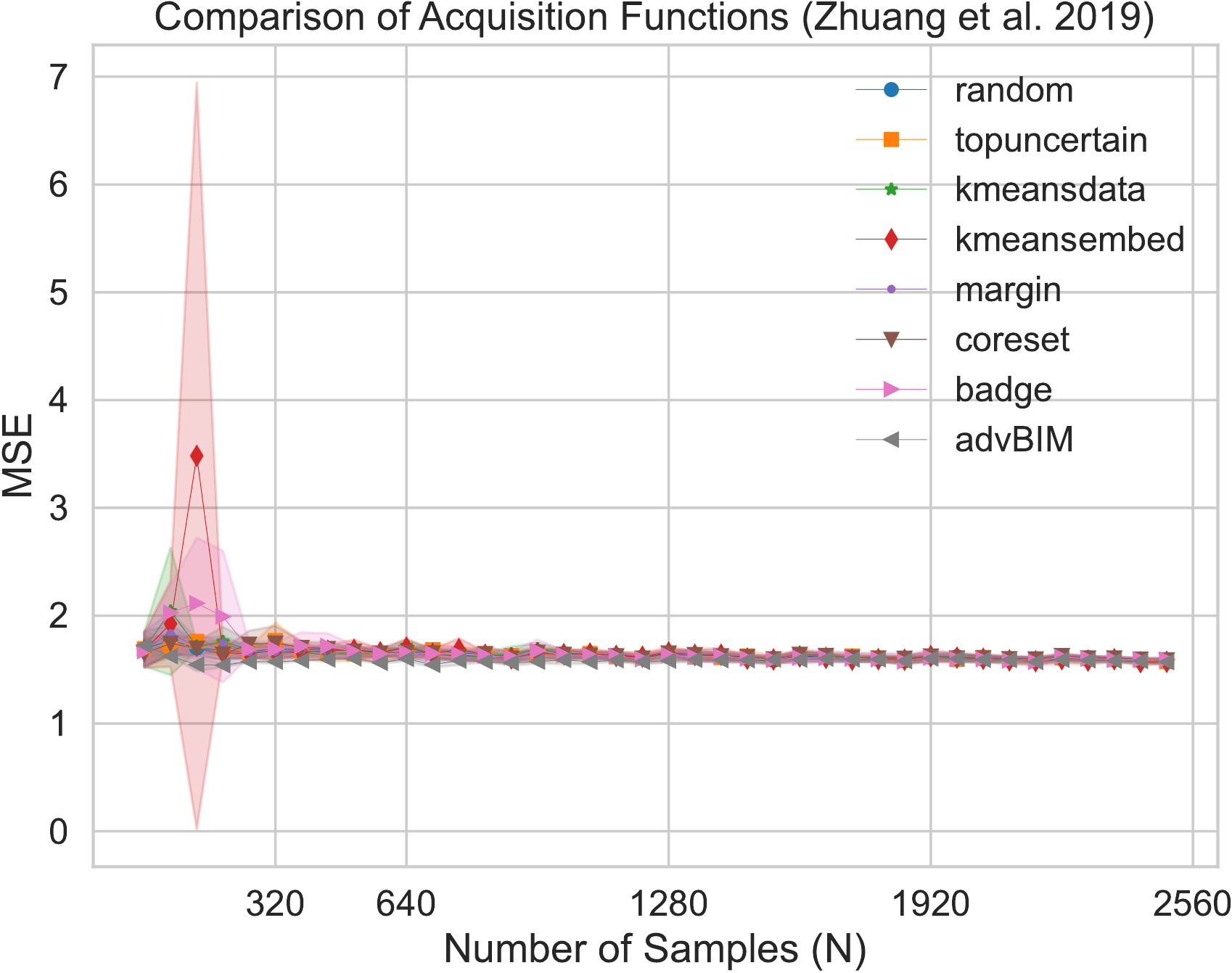}};
                        \end{tikzpicture}
                    }
                \end{subfigure}
                \&
                \\
\begin{subfigure}{0.27\columnwidth}
                    \hspace{-17mm}
                    \centering
                    \resizebox{\linewidth}{!}{
                        \begin{tikzpicture}
                            \node (img)  {\includegraphics[width=\textwidth]{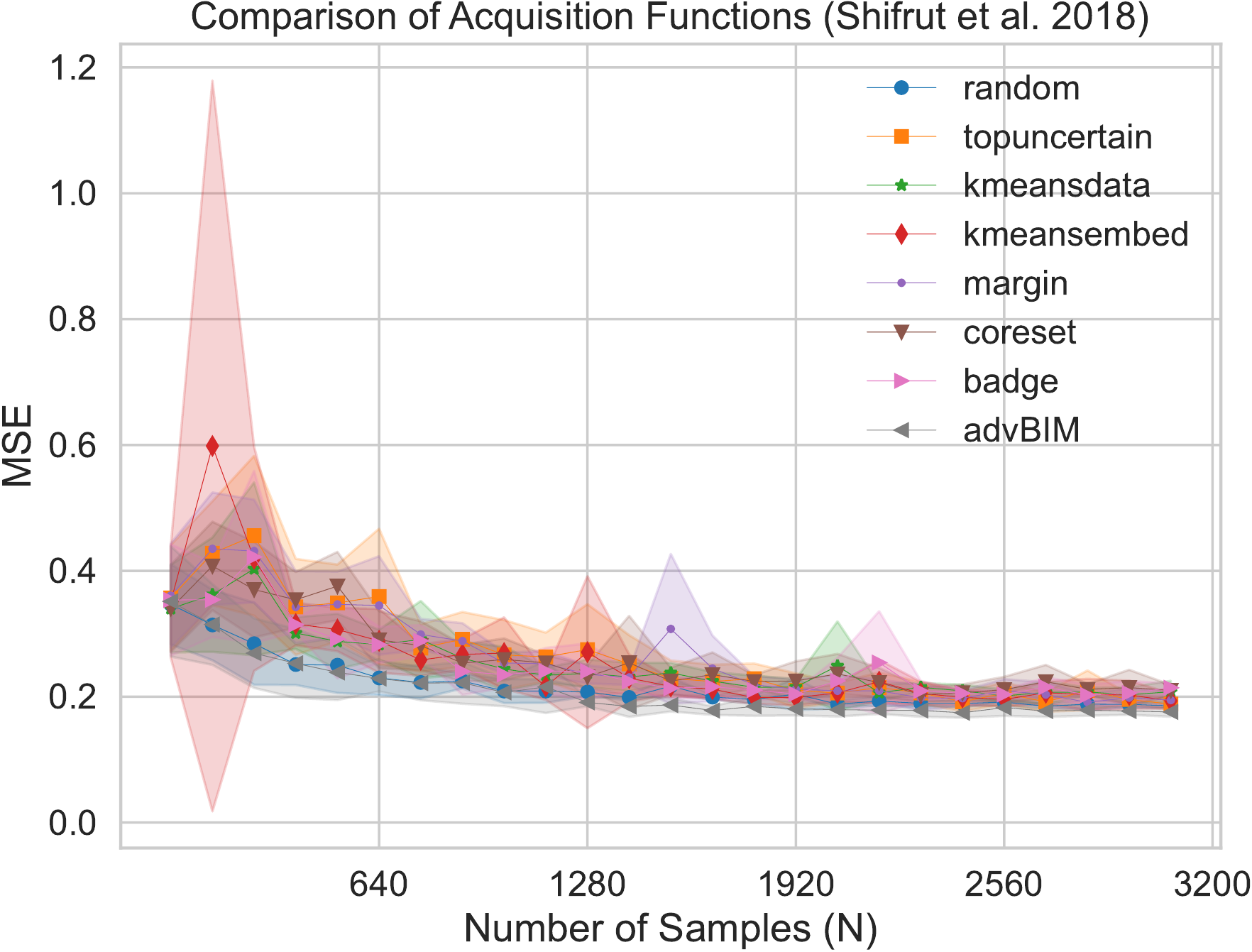}};
                        \end{tikzpicture}
                    }
                \end{subfigure}
                \&
                \begin{subfigure}{0.27\columnwidth}
                    \hspace{-23mm}
                    \centering
                    \resizebox{\linewidth}{!}{
                        \begin{tikzpicture}
                            \node (img)  {\includegraphics[width=\textwidth]{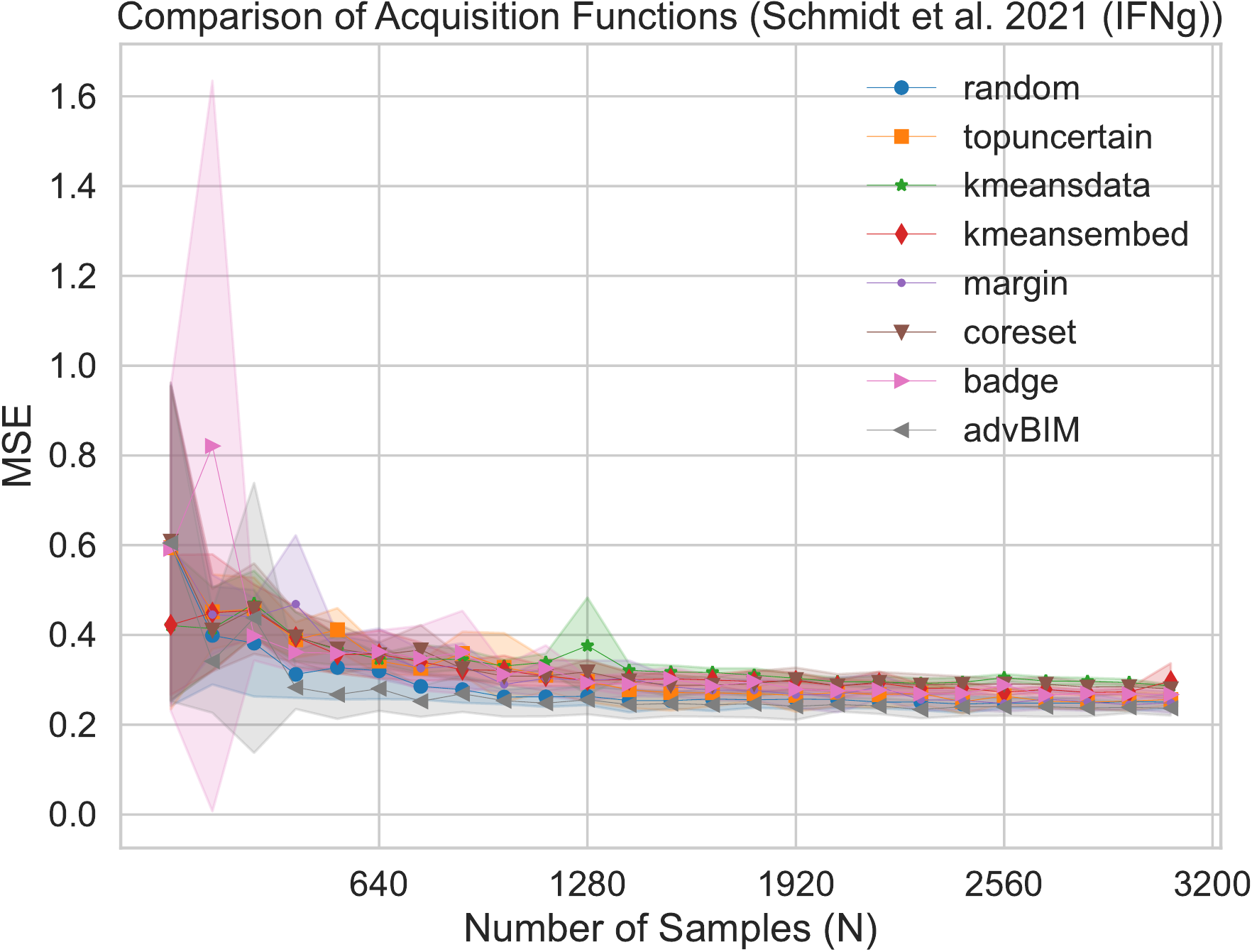}};
                        \end{tikzpicture}
                    }
                \end{subfigure}
                \&
                \begin{subfigure}{0.28\columnwidth}
                    \hspace{-28mm}
                    \centering
                    \resizebox{\linewidth}{!}{
                        \begin{tikzpicture}
                            \node (img)  {\includegraphics[width=\textwidth]{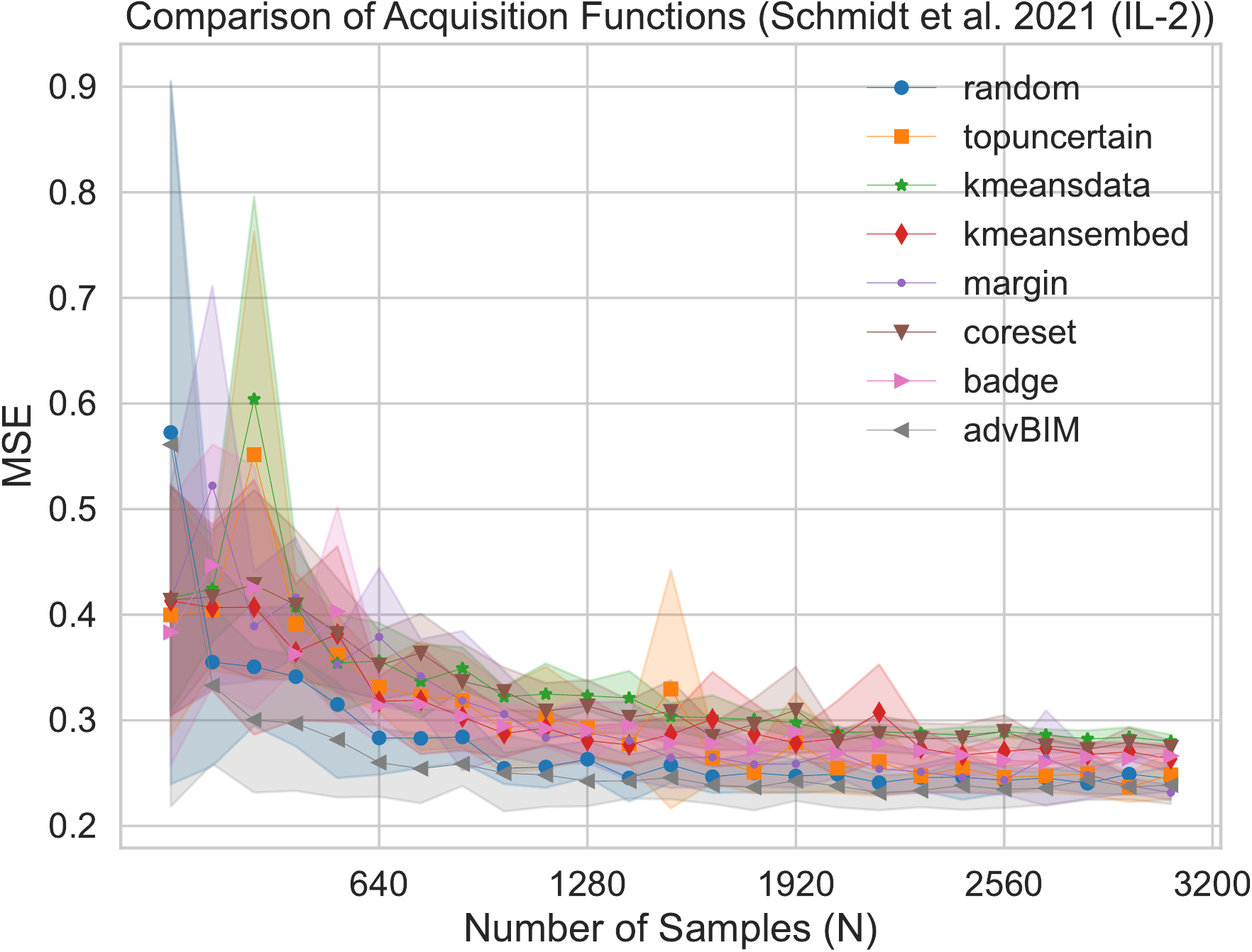}};
                        \end{tikzpicture}
                    }
                \end{subfigure}
                \&
                \begin{subfigure}{0.29\columnwidth}
                    \hspace{-32mm}
                    \centering
                    \resizebox{\linewidth}{!}{
                        \begin{tikzpicture}
                            \node (img)  {\includegraphics[width=\textwidth]{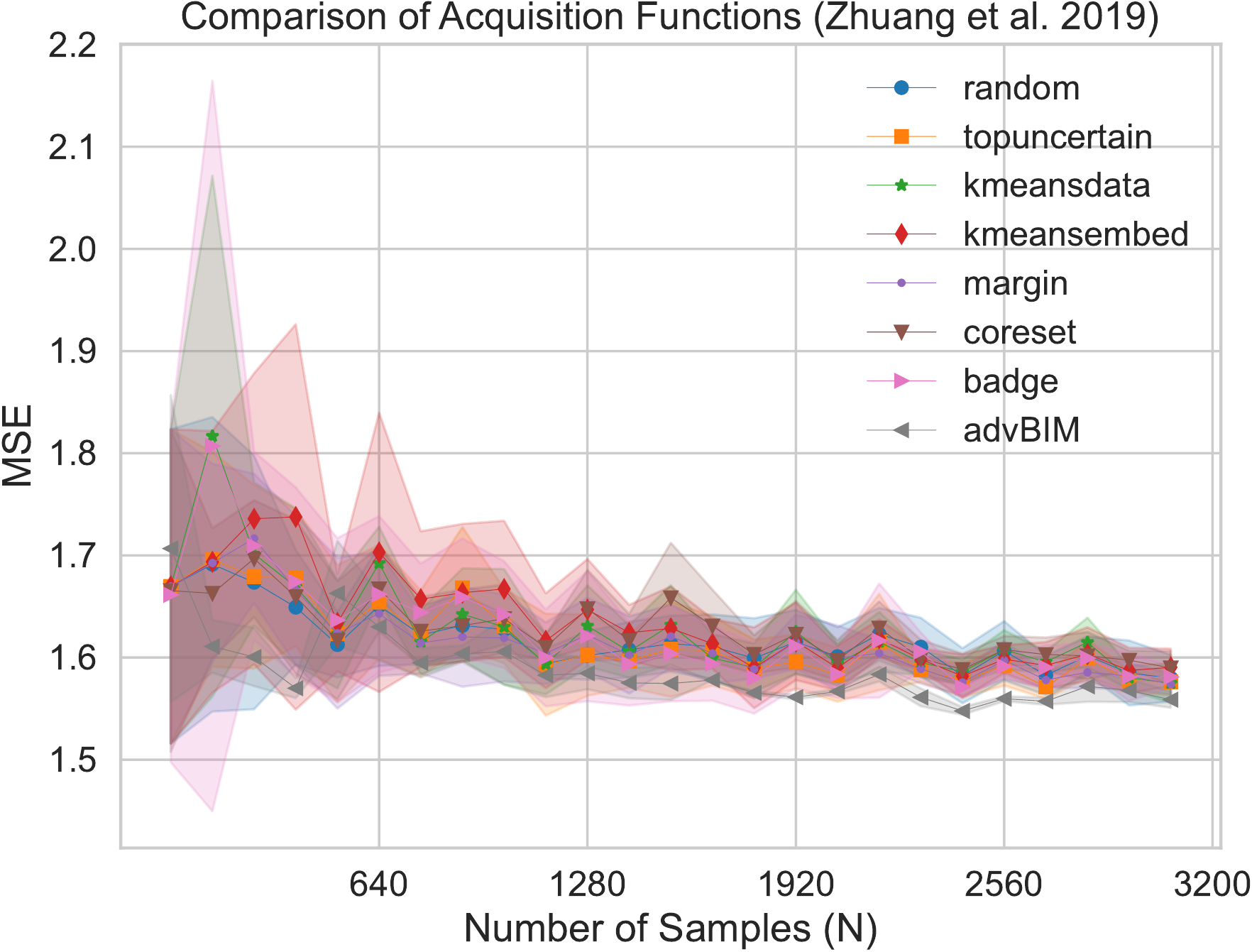}};
                        \end{tikzpicture}
                    }
                \end{subfigure}
                \&
                \\
\begin{subfigure}{0.275\columnwidth}
                    \hspace{-17mm}
                    \centering
                    \resizebox{\linewidth}{!}{
                        \begin{tikzpicture}
                            \node (img)  {\includegraphics[width=\textwidth]{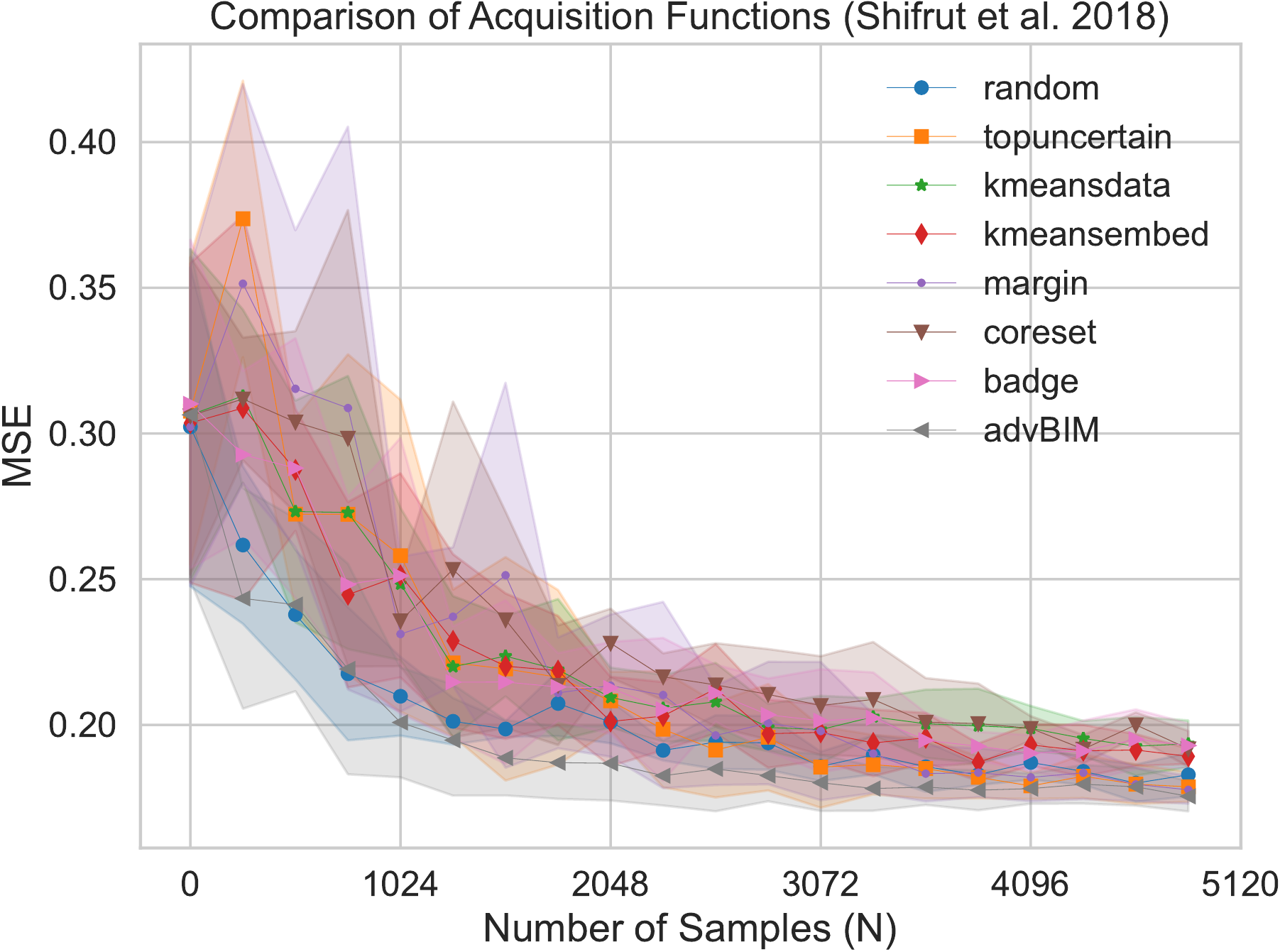}};
                        \end{tikzpicture}
                    }
                \end{subfigure}
                \&
                \begin{subfigure}{0.27\columnwidth}
                    \hspace{-23mm}
                    \centering
                    \resizebox{\linewidth}{!}{
                        \begin{tikzpicture}
                            \node (img)  {\includegraphics[width=\textwidth]{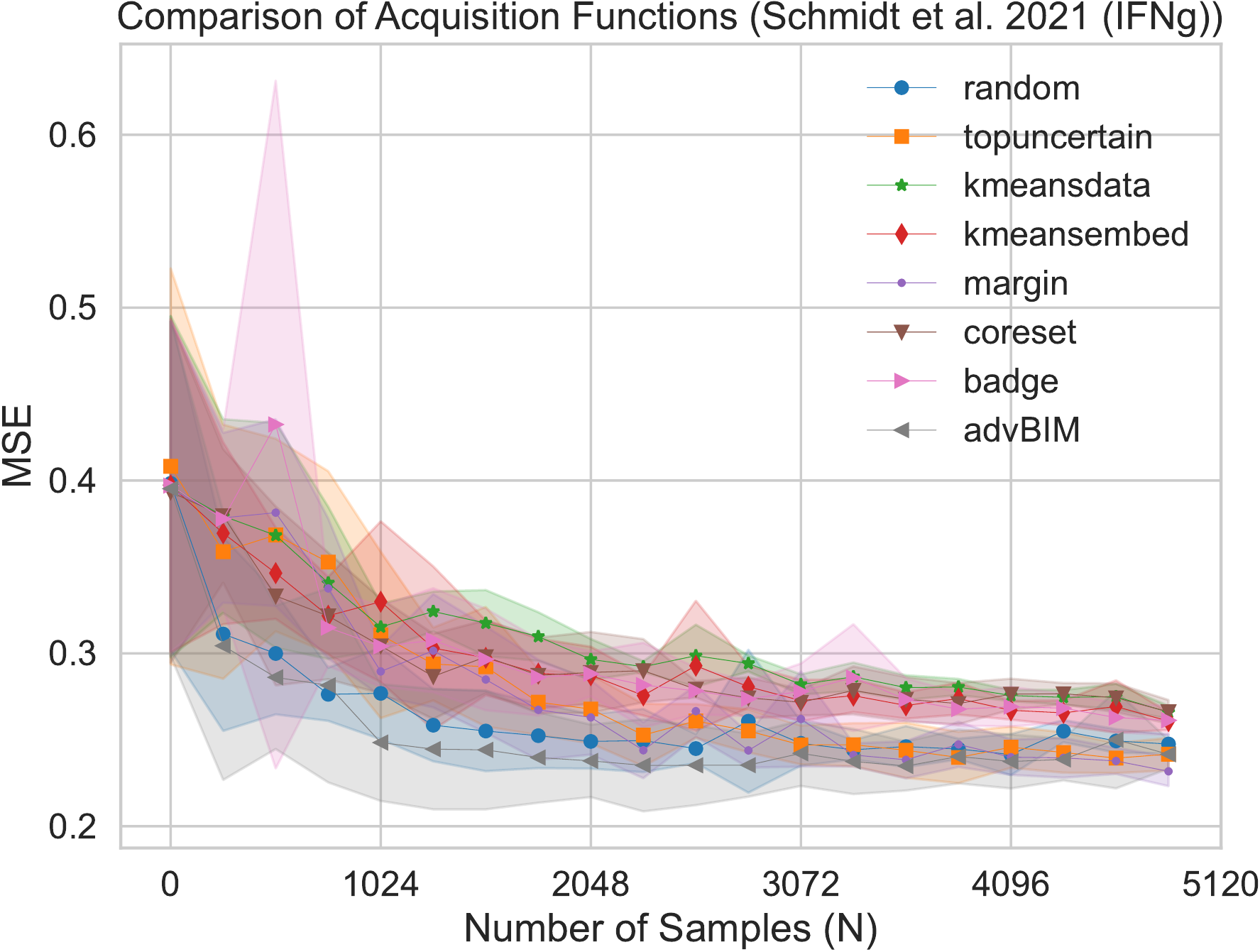}};
                        \end{tikzpicture}
                    }
                \end{subfigure}
                \&
                \begin{subfigure}{0.27\columnwidth}
                    \hspace{-28mm}
                    \centering
                    \resizebox{\linewidth}{!}{
                        \begin{tikzpicture}
                            \node (img)  {\includegraphics[width=\textwidth]{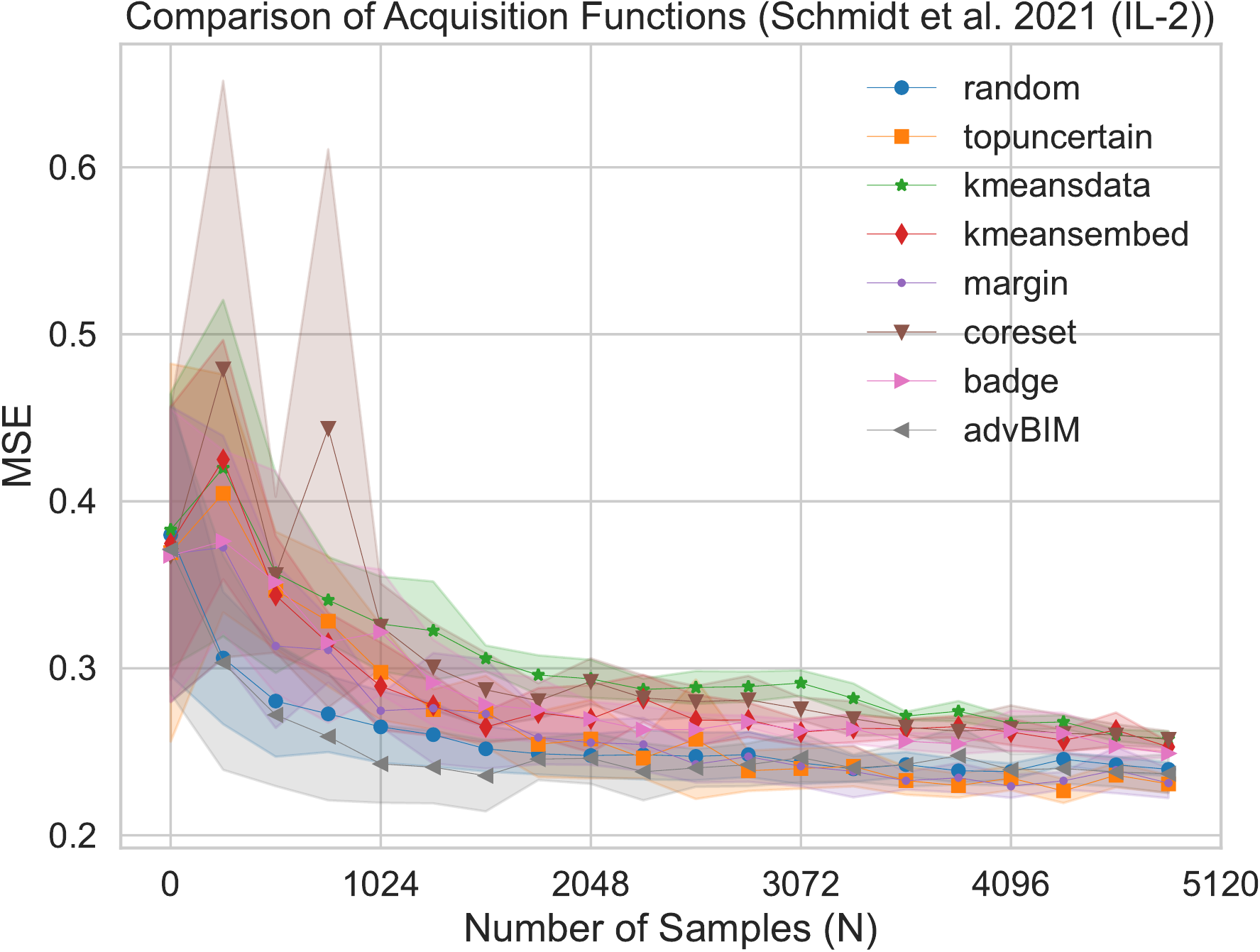}};
                        \end{tikzpicture}
                    }
                \end{subfigure}
                \&
                \begin{subfigure}{0.29\columnwidth}
                    \hspace{-32mm}
                    \centering
                    \resizebox{\linewidth}{!}{
                        \begin{tikzpicture}
                            \node (img)  {\includegraphics[width=\textwidth]{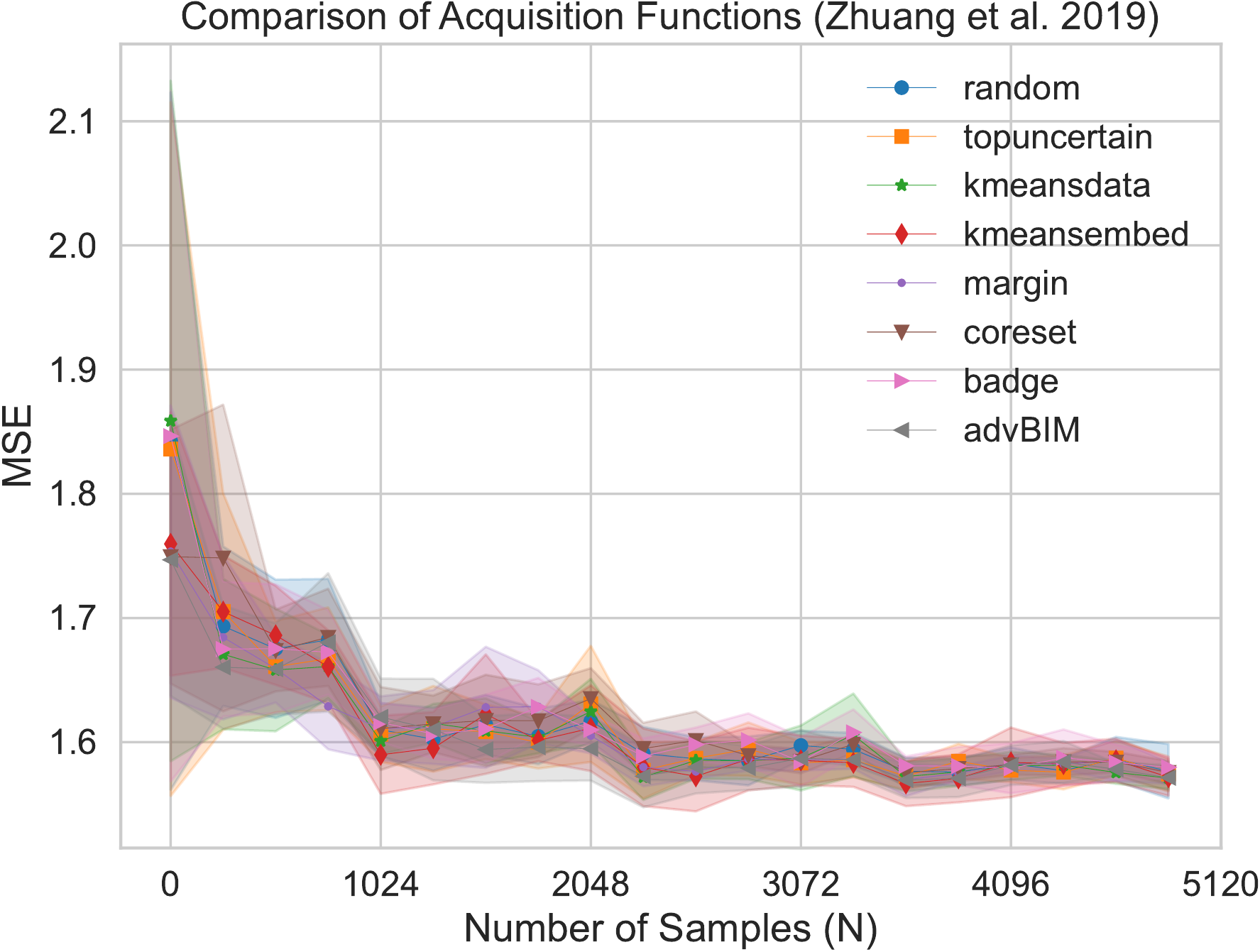}};
                        \end{tikzpicture}
                    }
                \end{subfigure}
                \&
                \\
\begin{subfigure}{0.28\columnwidth}
                    \hspace{-17mm}
                    \centering
                    \resizebox{\linewidth}{!}{
                        \begin{tikzpicture}
                            \node (img)  {\includegraphics[width=\textwidth]{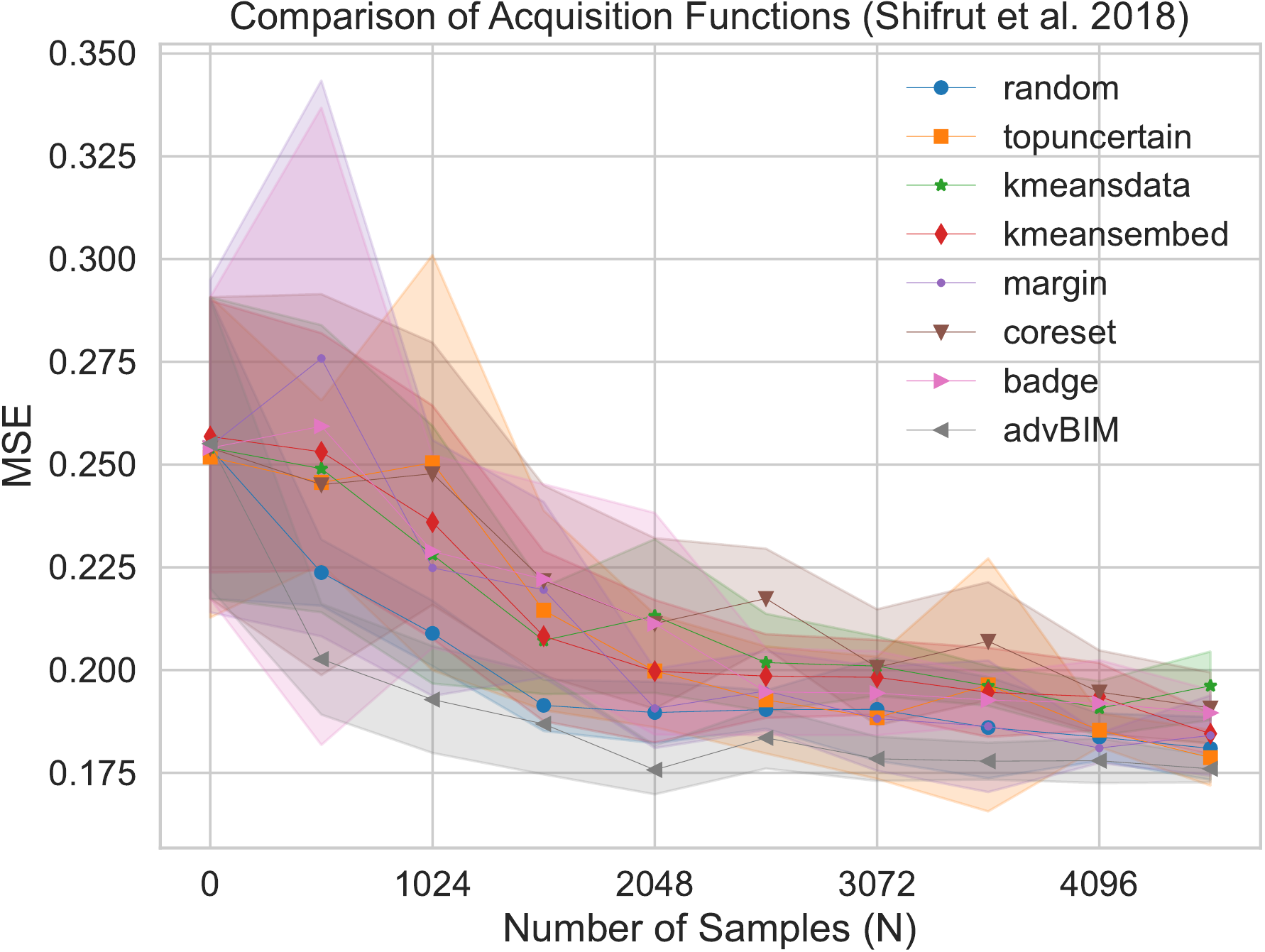}};
                        \end{tikzpicture}
                    }
                \end{subfigure}
                \&
                \begin{subfigure}{0.27\columnwidth}
                    \hspace{-23mm}
                    \centering
                    \resizebox{\linewidth}{!}{
                        \begin{tikzpicture}
                            \node (img)  {\includegraphics[width=\textwidth]{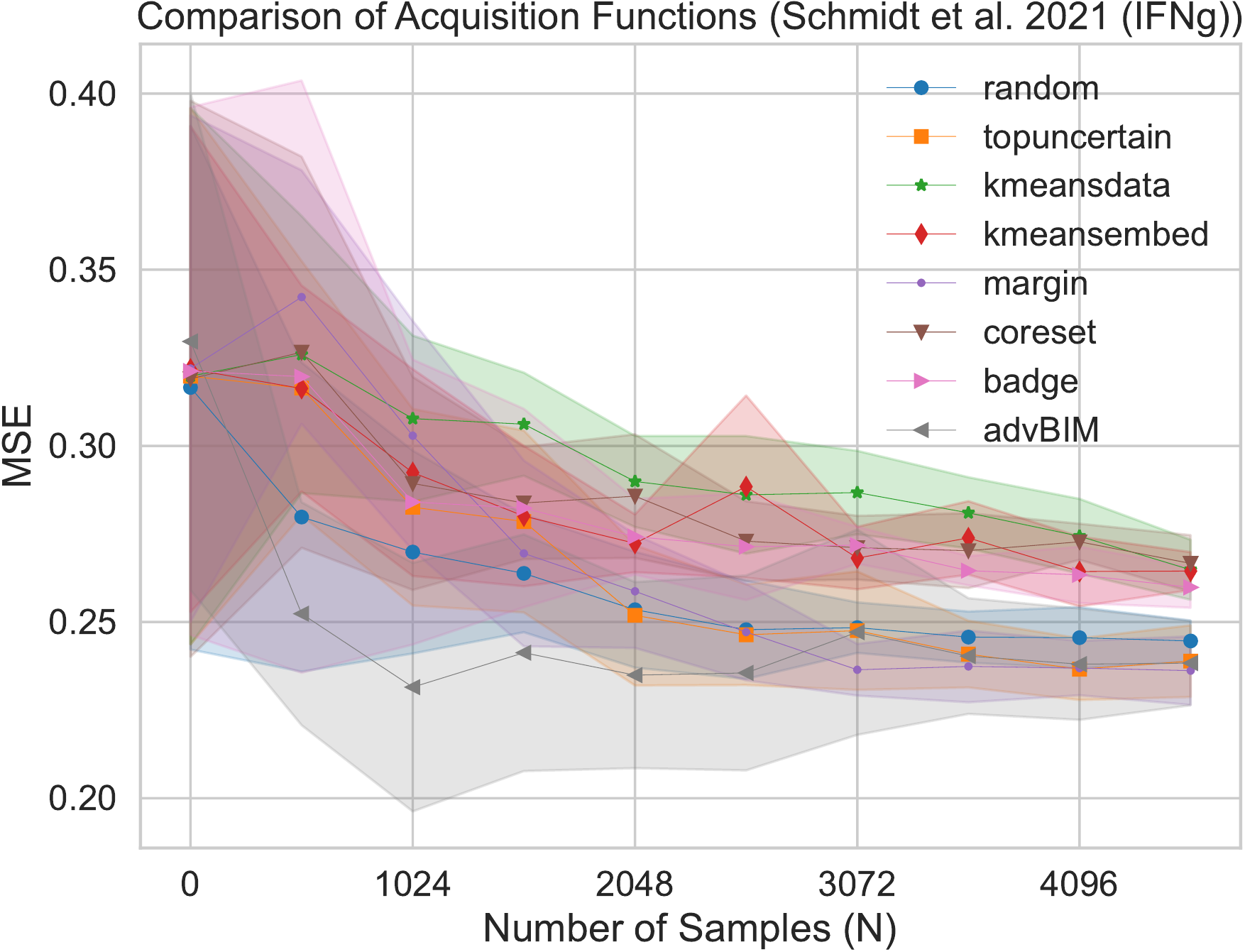}};
                        \end{tikzpicture}
                    }
                \end{subfigure}
                \&
                \begin{subfigure}{0.27\columnwidth}
                    \hspace{-28mm}
                    \centering
                    \resizebox{\linewidth}{!}{
                        \begin{tikzpicture}
                            \node (img)  {\includegraphics[width=\textwidth]{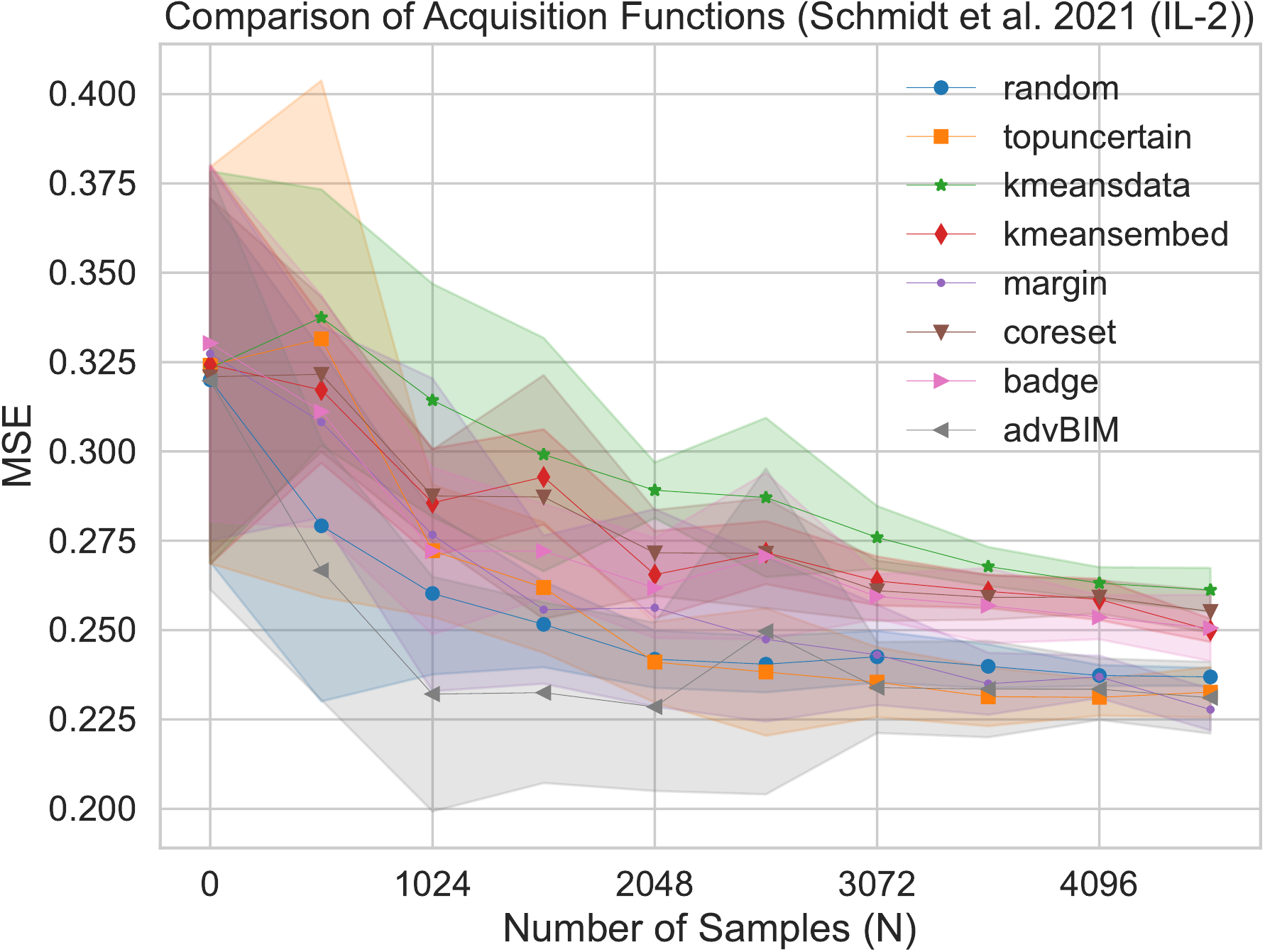}};
                        \end{tikzpicture}
                    }
                \end{subfigure}
                \&
                \begin{subfigure}{0.28\columnwidth}
                    \hspace{-32mm}
                    \centering
                    \resizebox{\linewidth}{!}{
                        \begin{tikzpicture}
                            \node (img)  {\includegraphics[width=\textwidth]{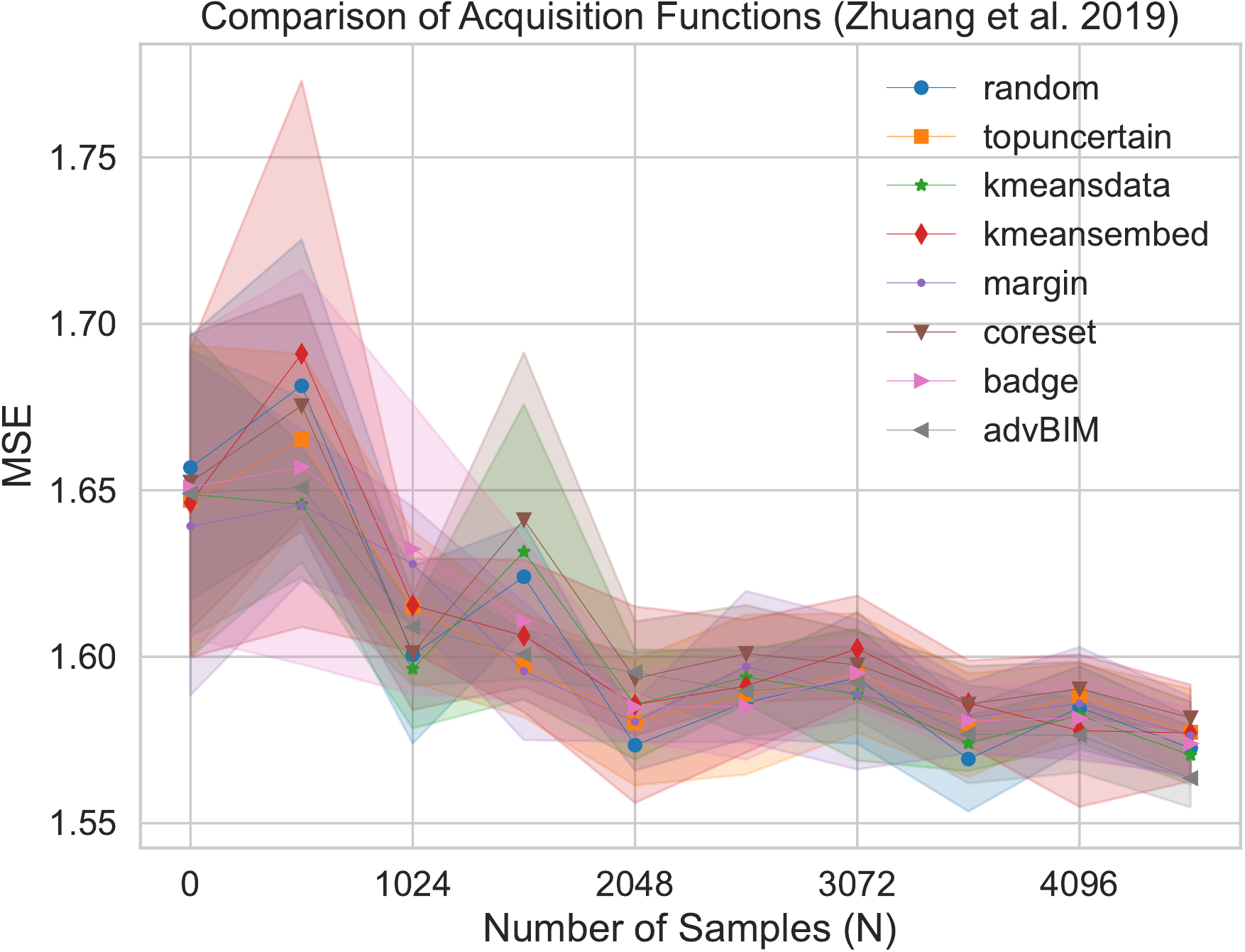}};
                        \end{tikzpicture}
                    }
                \end{subfigure}
                \&
                \\
            \\
           
            \\
            };
            \node [draw=none, rotate=90] at ([xshift=-8mm, yshift=2mm]fig-1-1.west) {\small batch size = 16};
            \node [draw=none, rotate=90] at ([xshift=-8mm, yshift=2mm]fig-2-1.west) {\small batch size = 32};
            \node [draw=none, rotate=90] at ([xshift=-8mm, yshift=2mm]fig-3-1.west) {\small batch size = 64};
            \node [draw=none, rotate=90] at ([xshift=-8mm, yshift=2mm]fig-4-1.west) {\small batch size = 128};
            \node [draw=none, rotate=90] at ([xshift=-8mm, yshift=2mm]fig-5-1.west) {\small batch size = 256};
            \node [draw=none, rotate=90] at ([xshift=-8mm, yshift=2mm]fig-6-1.west) {\small batch size = 512};
            \node [draw=none] at ([xshift=-6mm, yshift=3mm]fig-1-1.north) {\small Shifrut et al. 2018};
            \node [draw=none] at ([xshift=-9mm, yshift=3mm]fig-1-2.north) {\small Schmidt et al. 2021 (IFNg)};
            \node [draw=none] at ([xshift=-11mm, yshift=3mm]fig-1-3.north) {\small Schmidt et al. 2021 (IL-2)};
            \node [draw=none] at ([xshift=-13mm, yshift=2.5mm]fig-1-4.north) {\small Zhuang et al. 2019};
\end{tikzpicture}}
        \vspace{-7mm}
        \caption{The evaluation of the model trained with {CCLE} treatment descriptors at each active learning cycle for 4 datasets and 6 acquisition batch sizes. In each plot, the x-axis is the active learning cycles multiplied by the acquisition bath size that gives the total number of data points collected so far. The y-axis is the test MSE error evaluated on the test data.}
        \vspace{-5mm}
        \label{fig:bnn_feat_ccle_alldatasets_allbathcsizes}
    \end{figure*}

We provide here detailed experimental results across all hyperparameter settings. The result of \cref{fig:bnn_feat_string_alldatasets_allbathcsizes_small} that was presented for 3 batch sizes are provided for 6 batch sizes in \cref{fig:bnn_feat_string_alldatasets_allbathcsizes}. Similarly, the results of \cref{fig:hitratio_bnn_feat_string_alldatasets_allbathcsizes_small} are provided for additional batch sizes in \cref{fig:hitratio_bnn_feat_string_alldatasets_allbathcsizes}. In addition, both \cref{fig:bnn_feat_string_alldatasets_allbathcsizes_small} and \cref{fig:hitratio_bnn_feat_string_alldatasets_allbathcsizes_small} report the results for the STRING treatment descriptors. All experiments are repeated for two other sets of input treatment descriptors (Achilles and CCLE) whose results are provided in \cref{fig:bnn_feat_achilles_alldatasets_allbathcsizes,fig:bnn_feat_ccle_alldatasets_allbathcsizes,fig:hitratio_bnn_feat_achilles_alldatasets_allbathcsizes,fig:hitratio_bnn_feat_ccle_alldatasets_allbathcsizes}.

\subsection{Random Forest Model}
\label{sec:appendix_experiments_rf}
\begin{figure*}
    \vspace{-2mm}
        \centering
        \makebox[0.72\paperwidth]{\begin{tikzpicture}[ampersand replacement=\&]
            \matrix (fig) [matrix of nodes]{ 
\begin{subfigure}{0.27\columnwidth}
                    \hspace{-17mm}
                    \centering
                    \resizebox{\linewidth}{!}{
                        \begin{tikzpicture}
                            \node (img)  {\includegraphics[width=\textwidth]{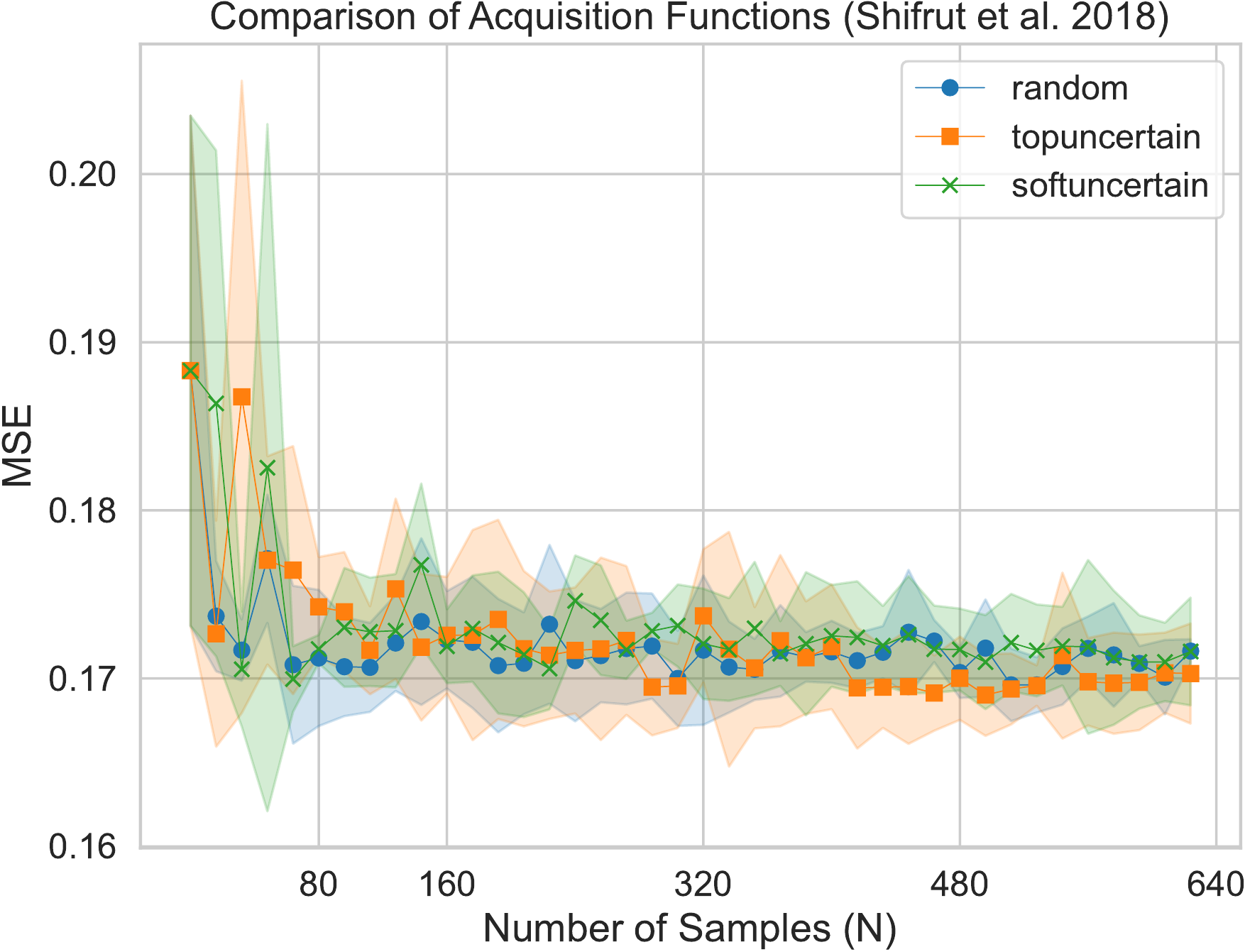}};
                        \end{tikzpicture}
                    }
                \end{subfigure}
                \&
                 \begin{subfigure}{0.27\columnwidth}
                    \hspace{-23mm}
                    \centering
                    \resizebox{\linewidth}{!}{
                        \begin{tikzpicture}
                            \node (img)  {\includegraphics[width=\textwidth]{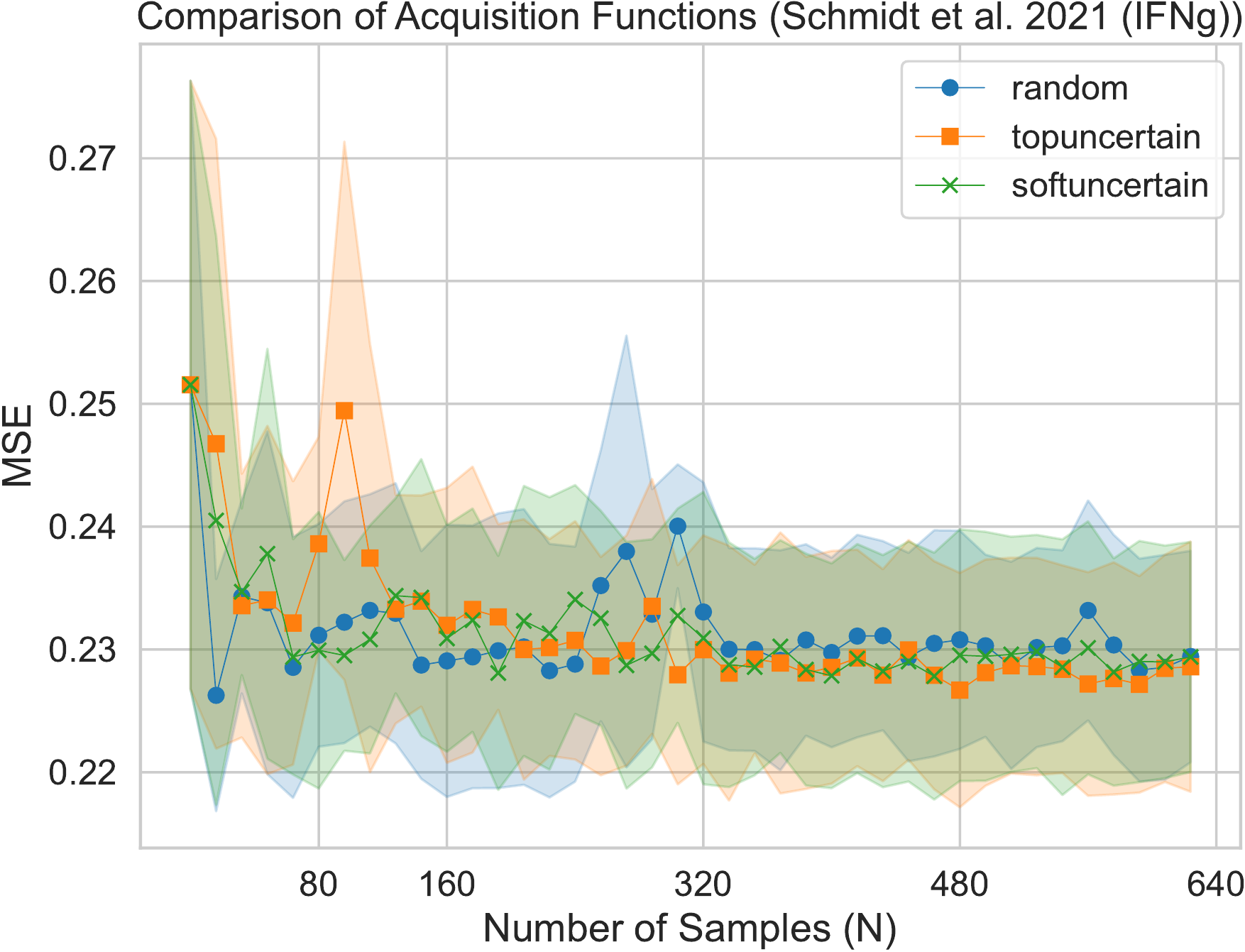}};
                        \end{tikzpicture}
                    }
                \end{subfigure}
                \&
                 \begin{subfigure}{0.27\columnwidth}
                    \hspace{-28mm}
                    \centering
                    \resizebox{\linewidth}{!}{
                        \begin{tikzpicture}
                            \node (img)  {\includegraphics[width=\textwidth]{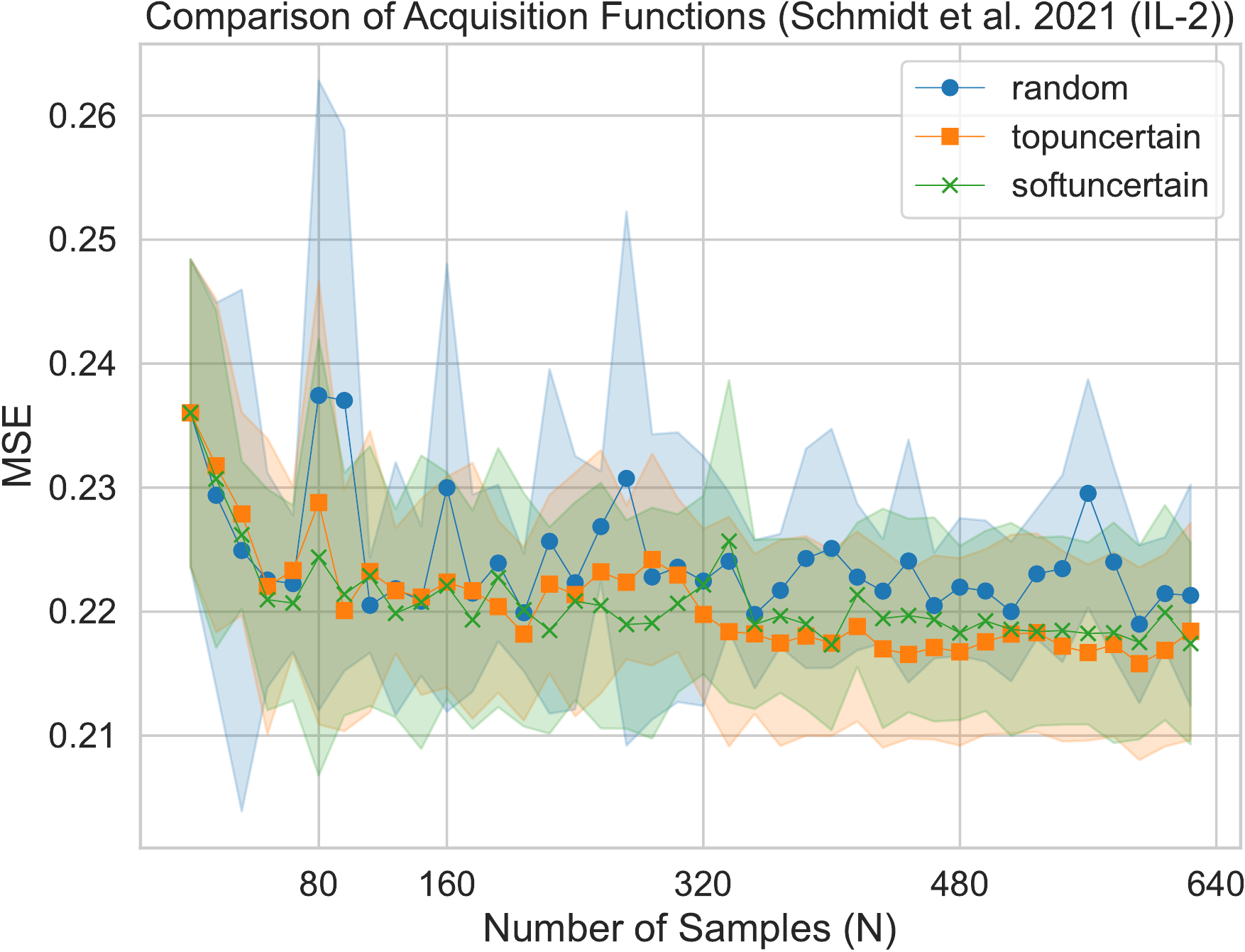}};
                        \end{tikzpicture}
                    }
                \end{subfigure}
                \&
                \begin{subfigure}{0.28\columnwidth}
                    \hspace{-32mm}
                    \centering
                    \resizebox{\linewidth}{!}{
                        \begin{tikzpicture}
                            \node (img)  {\includegraphics[width=\textwidth]{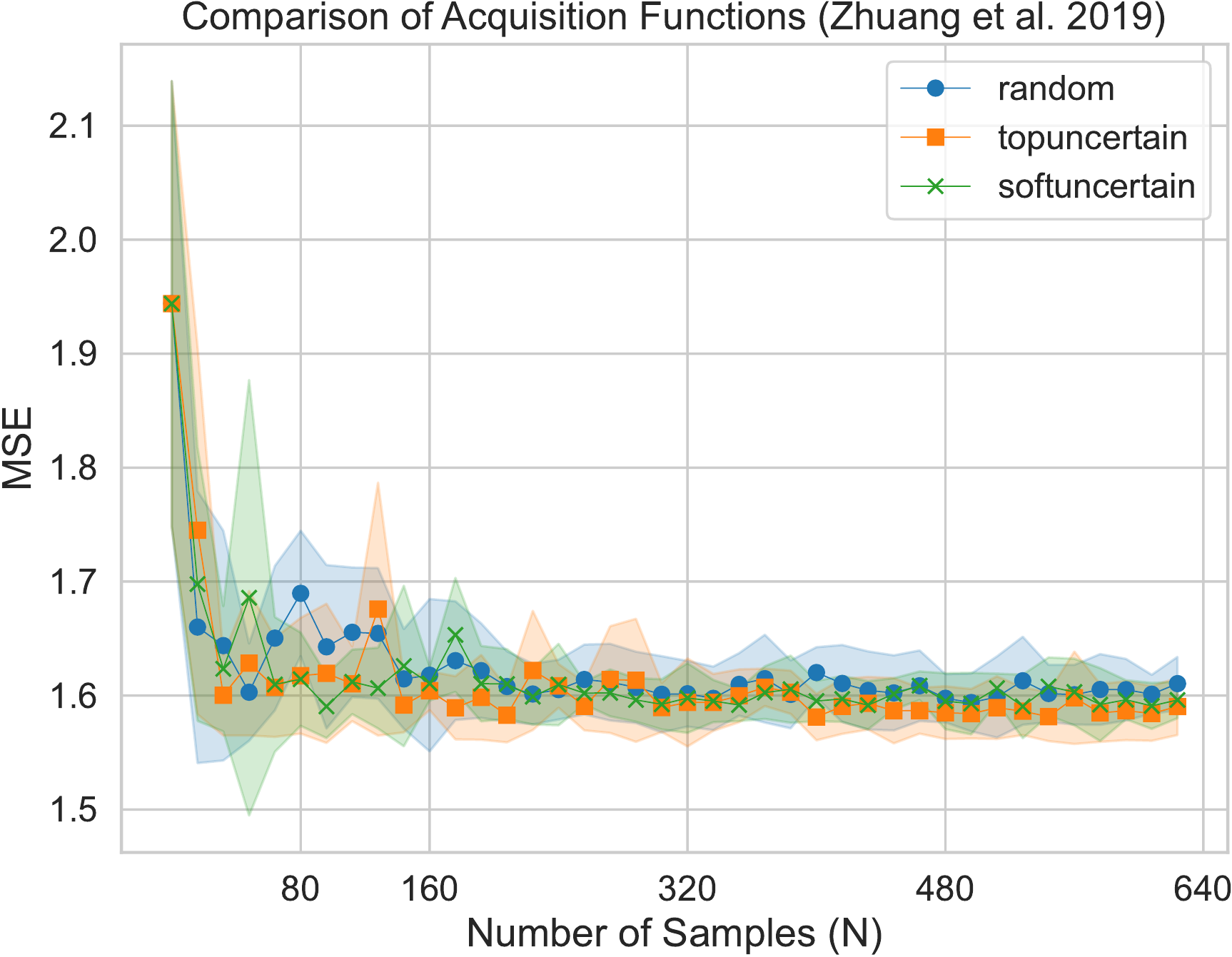}};
                        \end{tikzpicture}
                    }
                \end{subfigure}
                \&
            \\
\begin{subfigure}{0.27\columnwidth}
                    \hspace{-17mm}
                    \centering
                    \resizebox{\linewidth}{!}{
                        \begin{tikzpicture}
                            \node (img)  {\includegraphics[width=\textwidth]{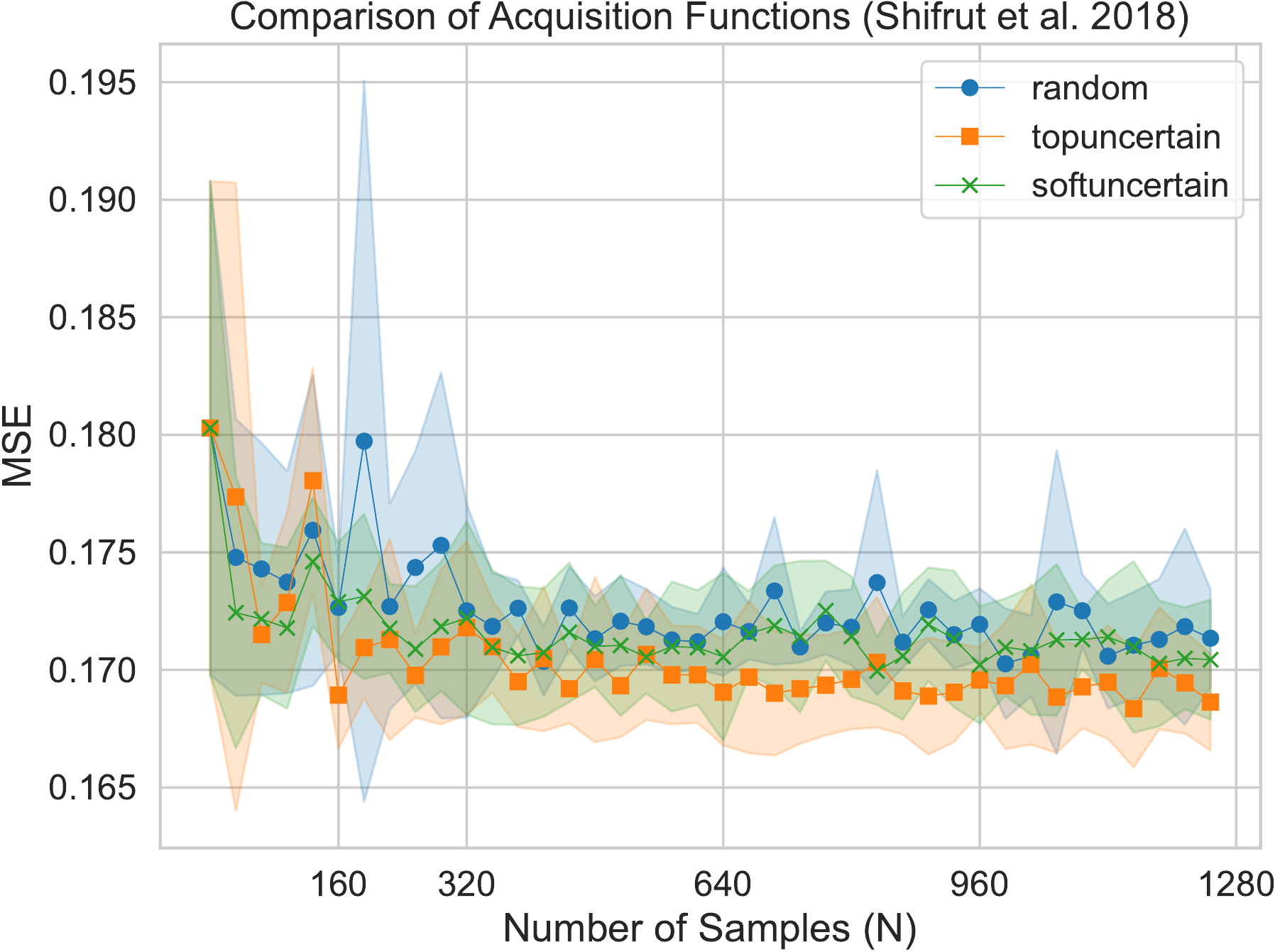}};
                        \end{tikzpicture}
                    }
                \end{subfigure}
                \&
                \begin{subfigure}{0.27\columnwidth}
                    \hspace{-23mm}
                    \centering
                    \resizebox{\linewidth}{!}{
                        \begin{tikzpicture}
                            \node (img)  {\includegraphics[width=\textwidth]{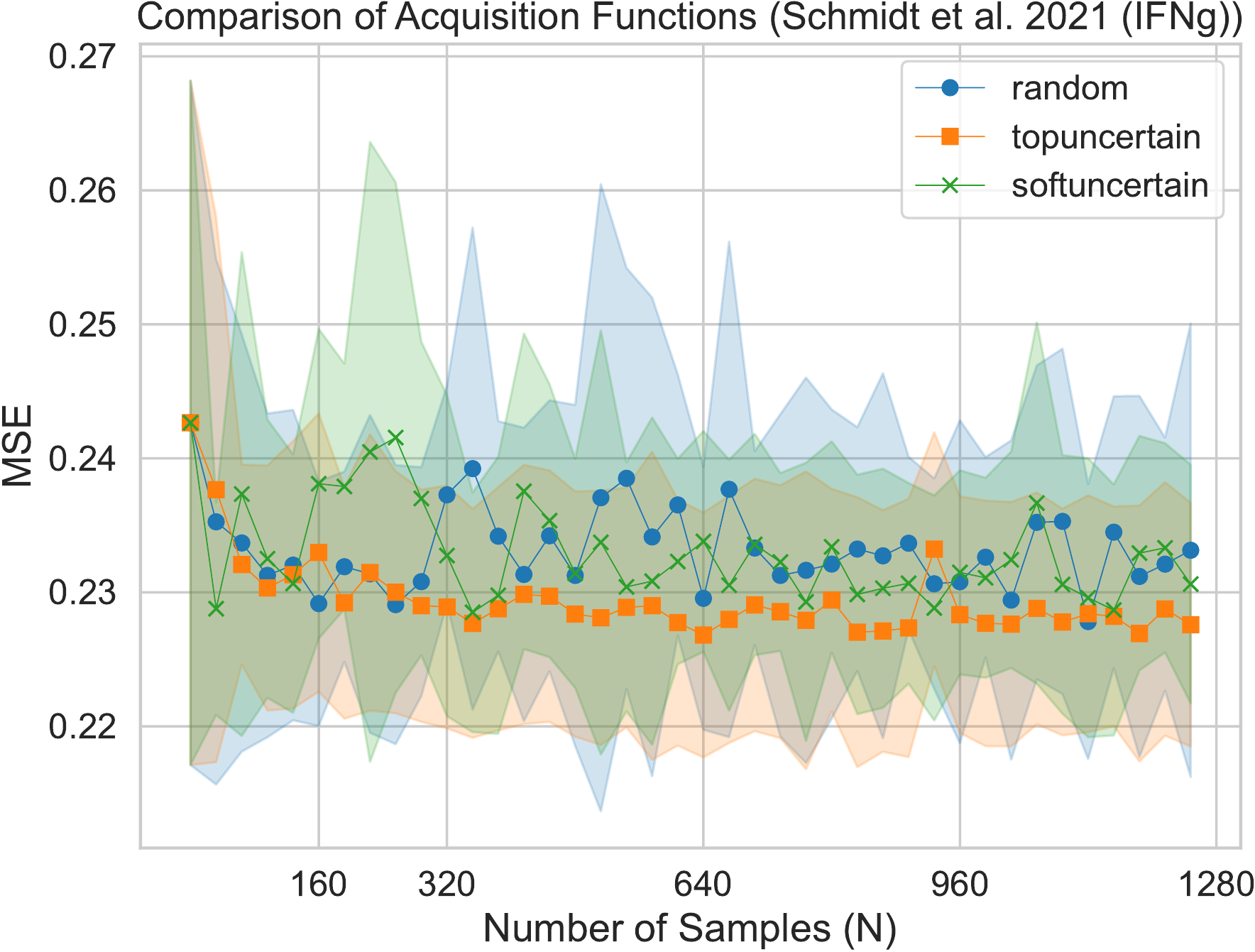}};
                        \end{tikzpicture}
                    }
                \end{subfigure}
                \&
                \begin{subfigure}{0.27\columnwidth}
                    \hspace{-28mm}
                    \centering
                    \resizebox{\linewidth}{!}{
                        \begin{tikzpicture}
                            \node (img)  {\includegraphics[width=\textwidth]{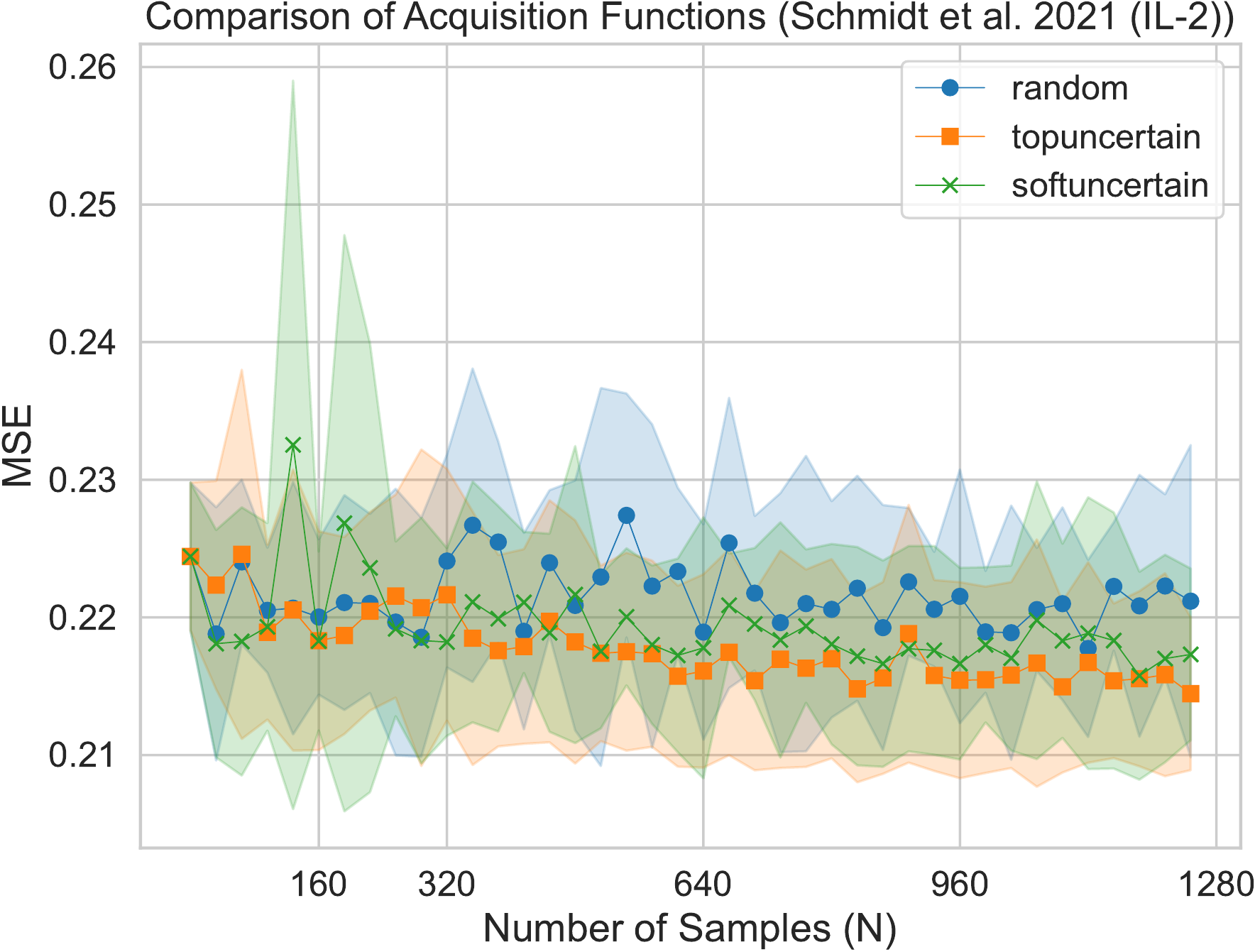}};
                        \end{tikzpicture}
                    }
                \end{subfigure}
                \&
                \begin{subfigure}{0.28\columnwidth}
                    \hspace{-32mm}
                    \centering
                    \resizebox{\linewidth}{!}{
                        \begin{tikzpicture}
                            \node (img)  {\includegraphics[width=\textwidth]{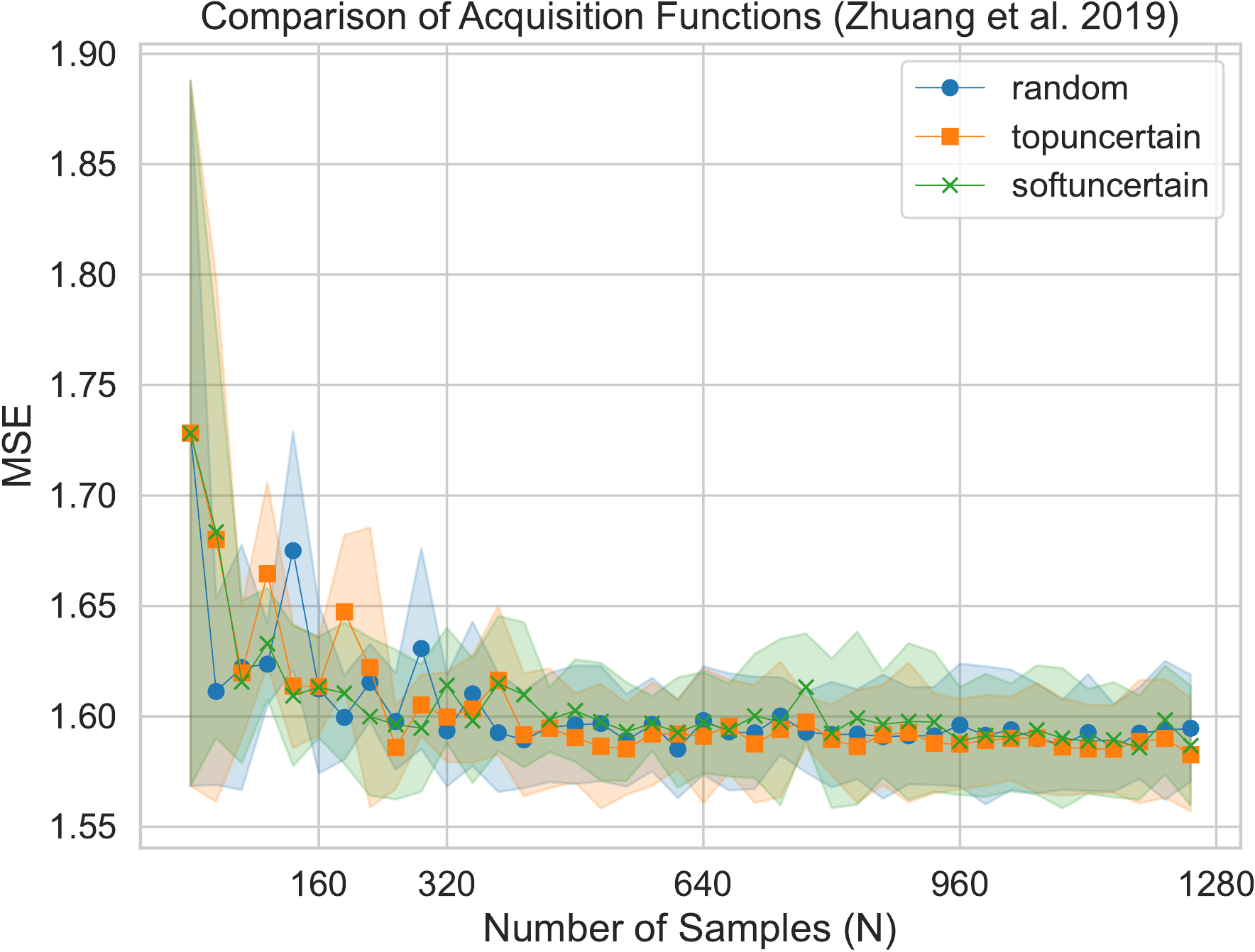}};
                        \end{tikzpicture}
                    }
                \end{subfigure}
                \&
                \\
\begin{subfigure}{0.27\columnwidth}
                    \hspace{-17mm}
                    \centering
                    \resizebox{\linewidth}{!}{
                        \begin{tikzpicture}
                            \node (img)  {\includegraphics[width=\textwidth]{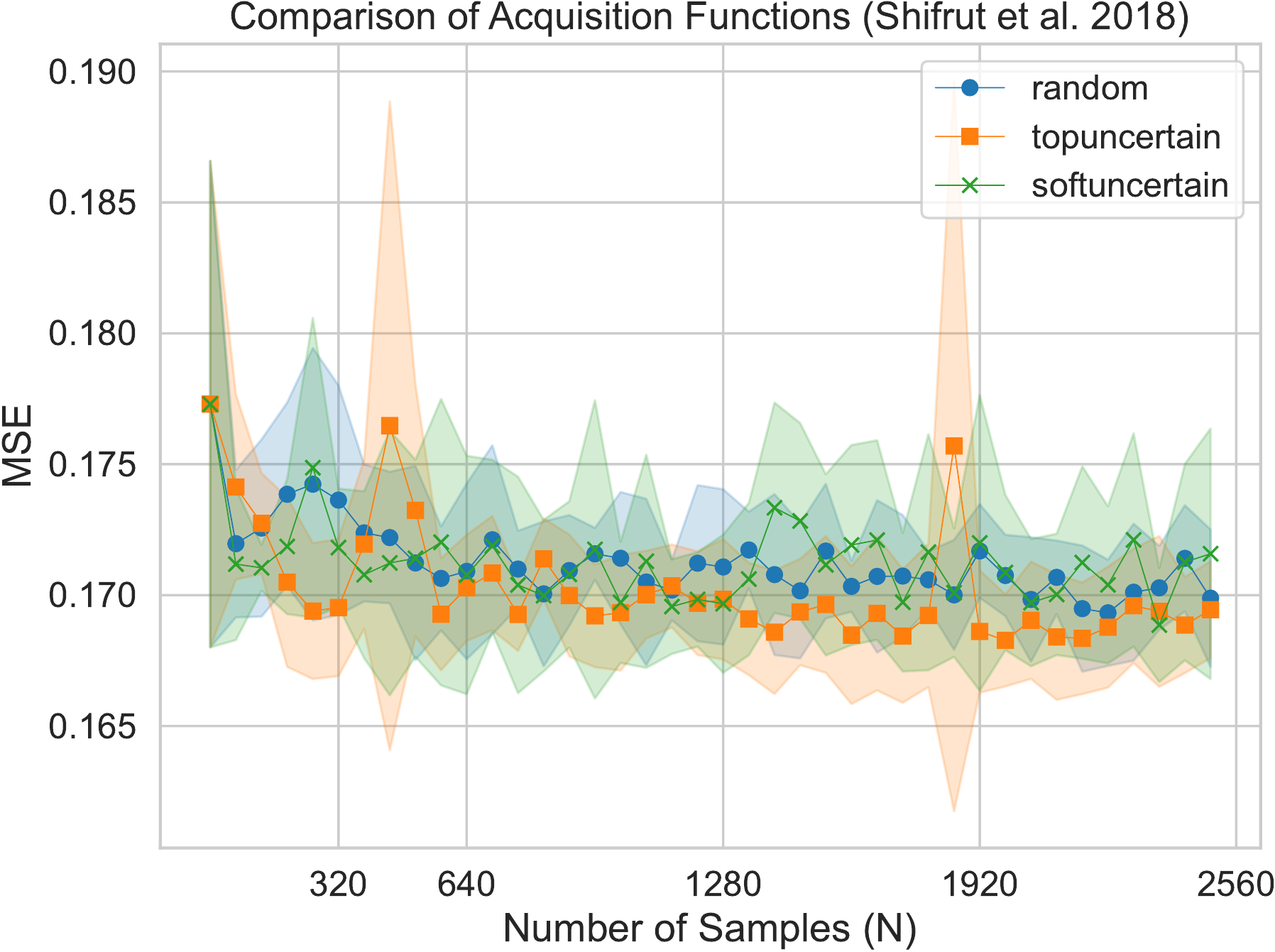}};
                        \end{tikzpicture}
                    }
                \end{subfigure}
                \&
                \begin{subfigure}{0.27\columnwidth}
                    \hspace{-23mm}
                    \centering
                    \resizebox{\linewidth}{!}{
                        \begin{tikzpicture}
                            \node (img)  {\includegraphics[width=\textwidth]{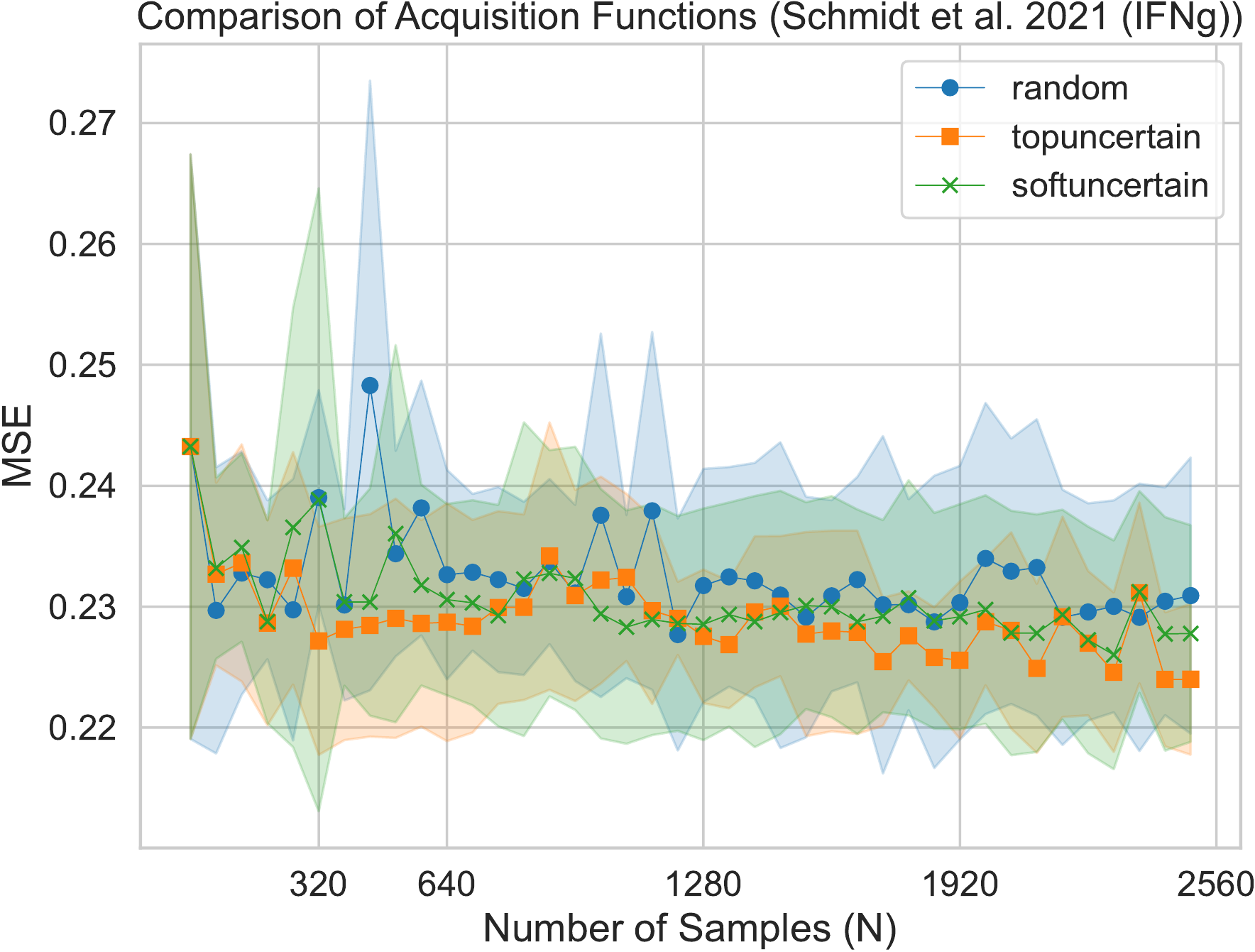}};
                        \end{tikzpicture}
                    }
                \end{subfigure}
                \&
                \begin{subfigure}{0.27\columnwidth}
                    \hspace{-28mm}
                    \centering
                    \resizebox{\linewidth}{!}{
                        \begin{tikzpicture}
                            \node (img)  {\includegraphics[width=\textwidth]{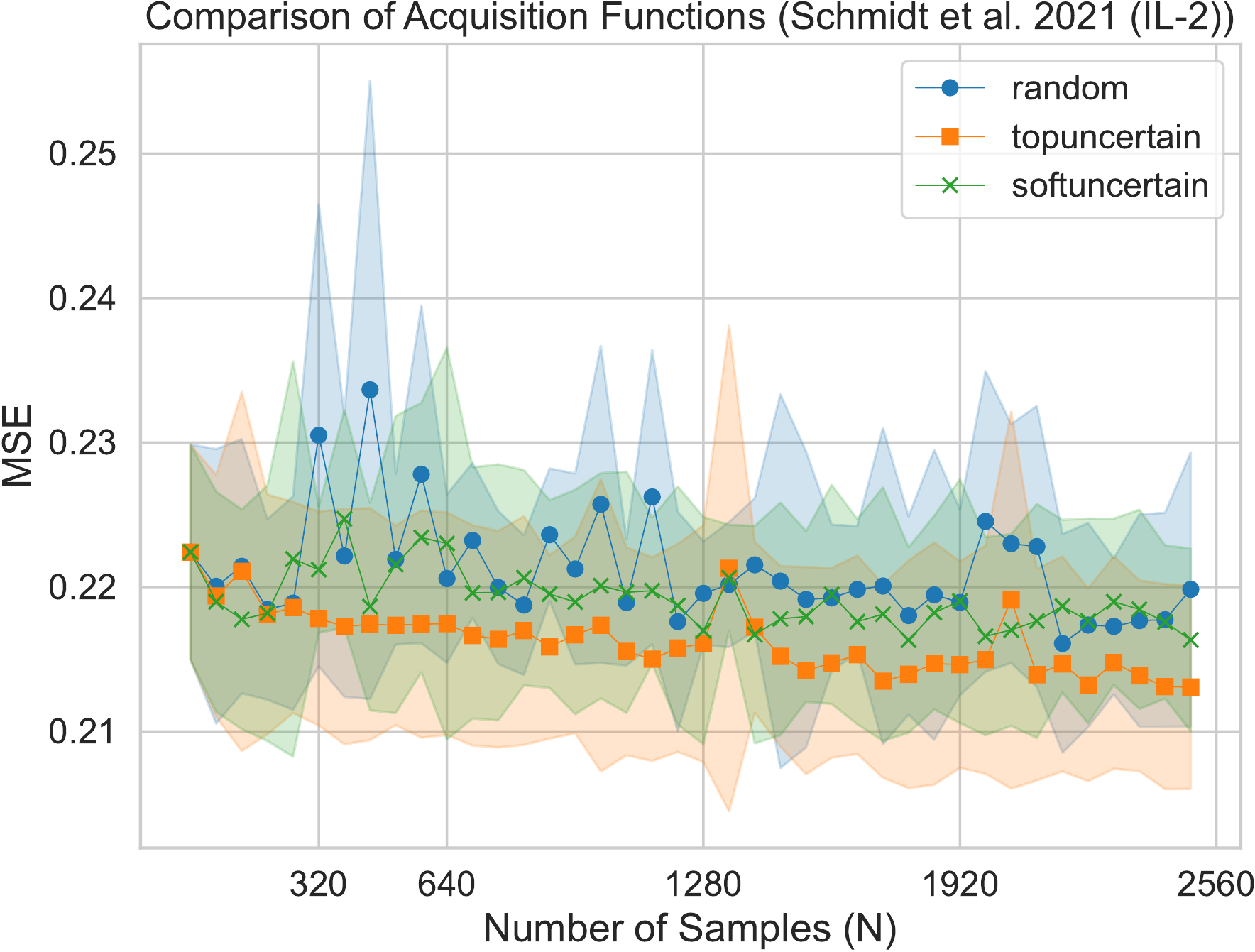}};
                        \end{tikzpicture}
                    }
                \end{subfigure}
                \&
                \begin{subfigure}{0.28\columnwidth}
                    \hspace{-32mm}
                    \centering
                    \resizebox{\linewidth}{!}{
                        \begin{tikzpicture}
                            \node (img)  {\includegraphics[width=\textwidth]{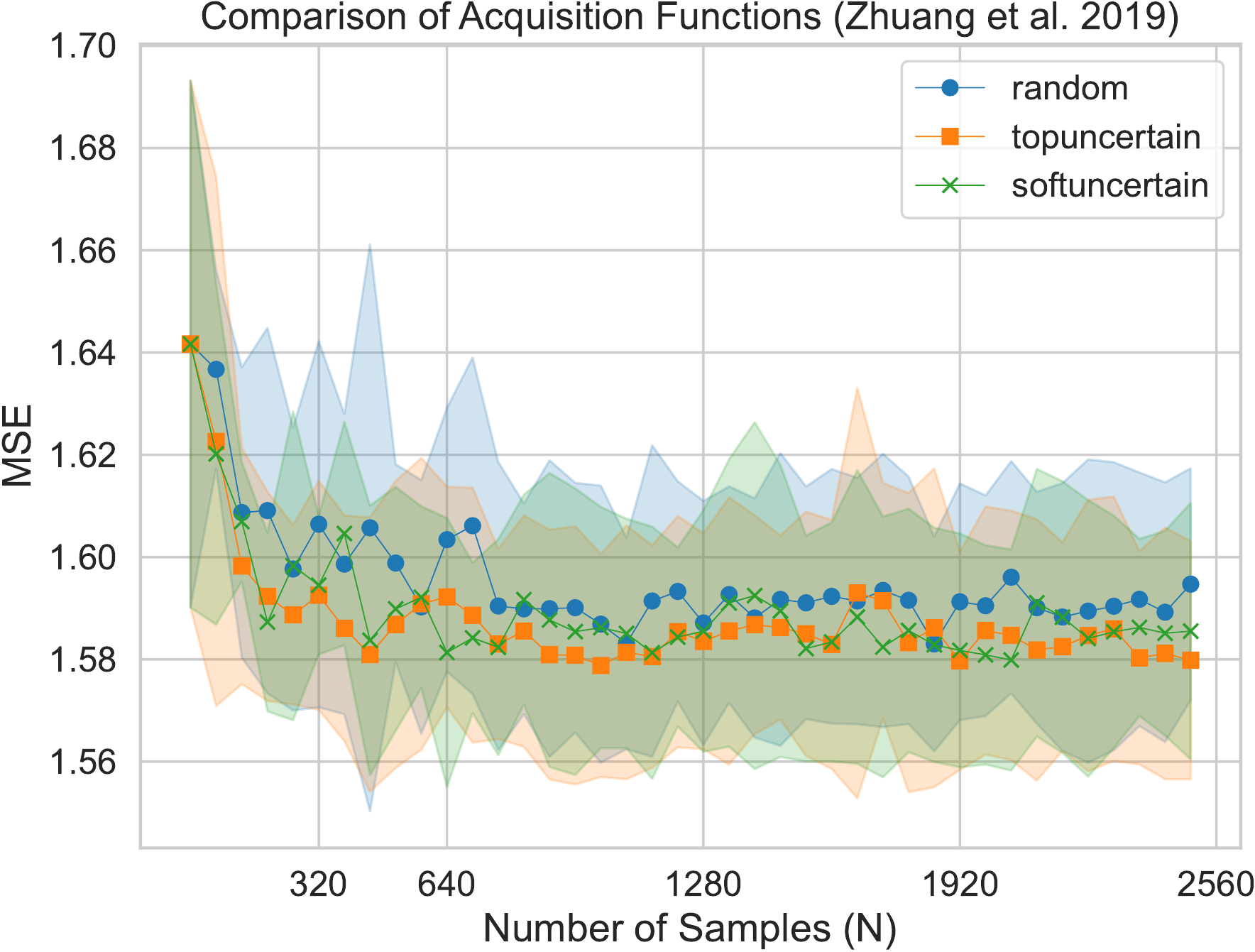}};
                        \end{tikzpicture}
                    }
                \end{subfigure}
                \&
                \\
\begin{subfigure}{0.27\columnwidth}
                    \hspace{-17mm}
                    \centering
                    \resizebox{\linewidth}{!}{
                        \begin{tikzpicture}
                            \node (img)  {\includegraphics[width=\textwidth]{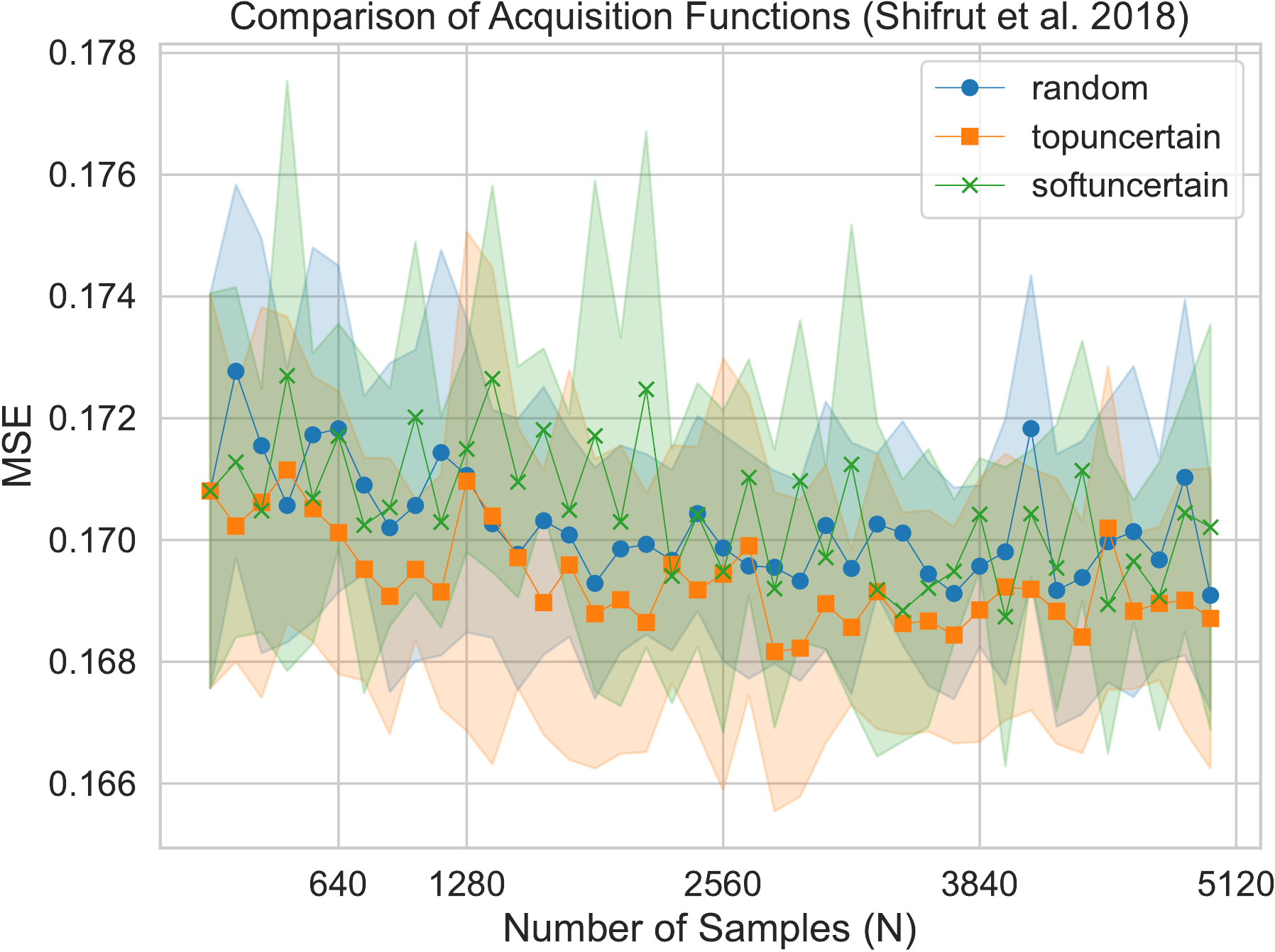}};
                        \end{tikzpicture}
                    }
                \end{subfigure}
                \&
                \begin{subfigure}{0.27\columnwidth}
                    \hspace{-23mm}
                    \centering
                    \resizebox{\linewidth}{!}{
                        \begin{tikzpicture}
                            \node (img)  {\includegraphics[width=\textwidth]{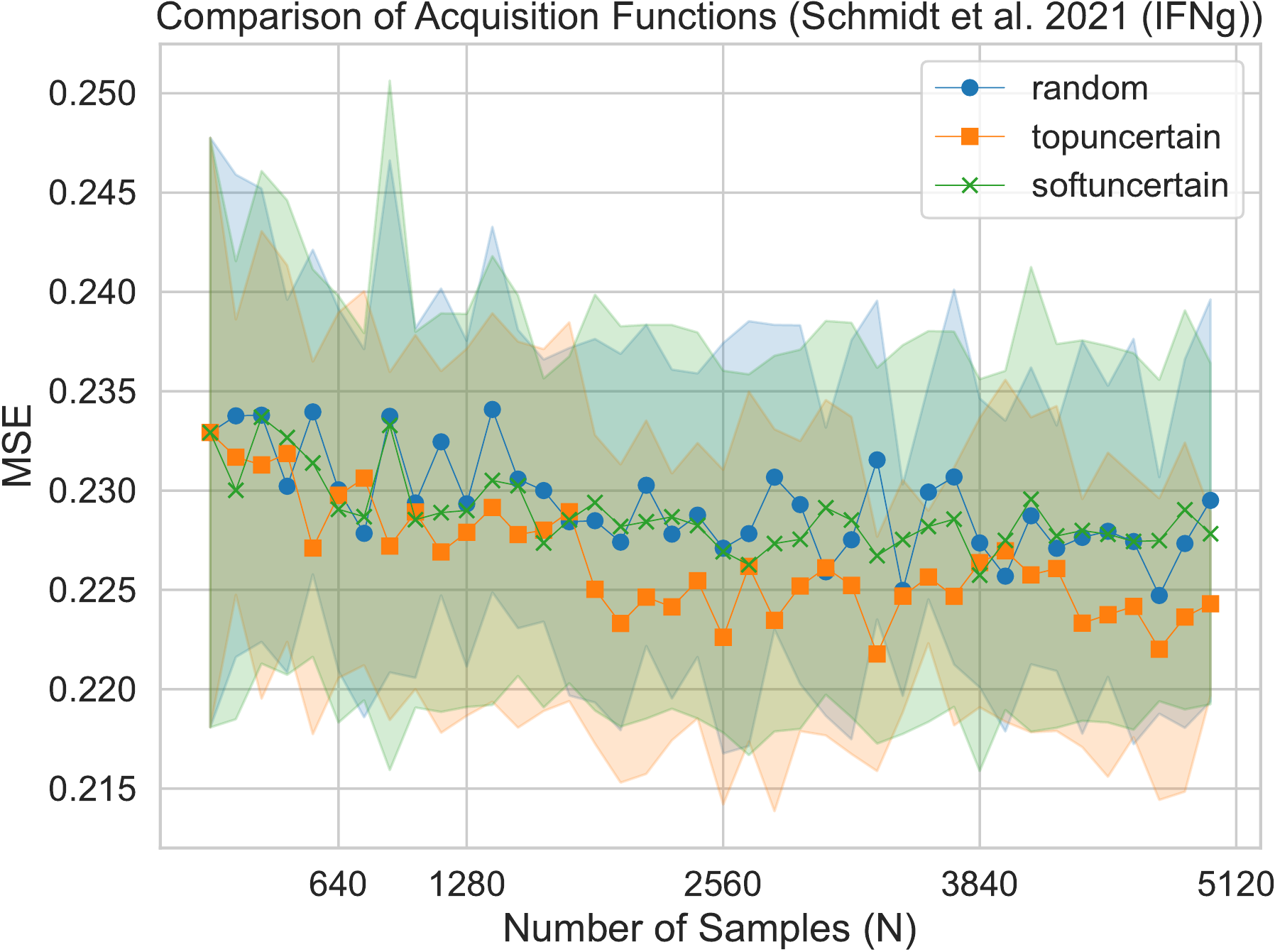}};
                        \end{tikzpicture}
                    }
                \end{subfigure}
                \&
                \begin{subfigure}{0.28\columnwidth}
                    \hspace{-28mm}
                    \centering
                    \resizebox{\linewidth}{!}{
                        \begin{tikzpicture}
                            \node (img)  {\includegraphics[width=\textwidth]{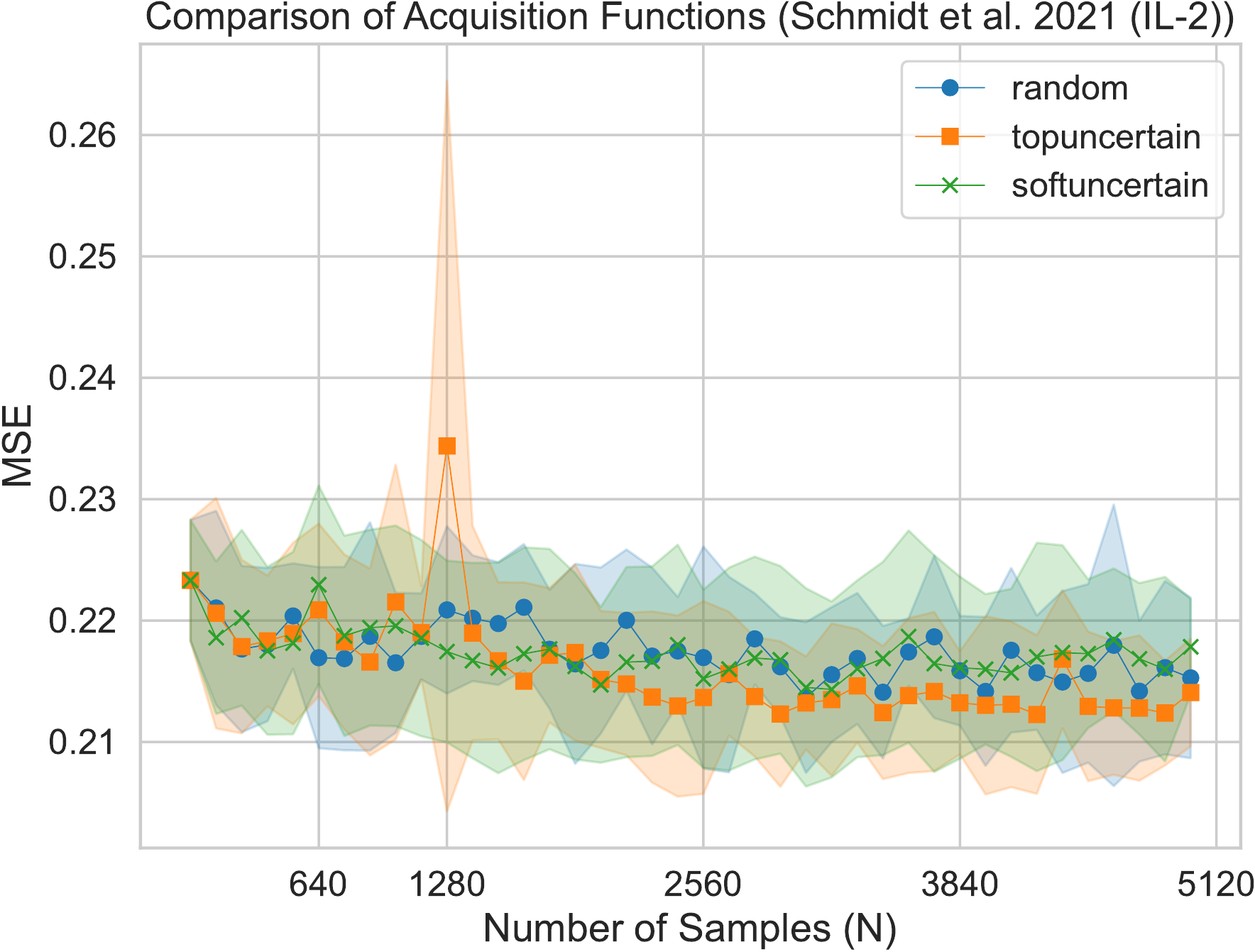}};
                        \end{tikzpicture}
                    }
                \end{subfigure}
                \&
                \begin{subfigure}{0.29\columnwidth}
                    \hspace{-32mm}
                    \centering
                    \resizebox{\linewidth}{!}{
                        \begin{tikzpicture}
                            \node (img)  {\includegraphics[width=\textwidth]{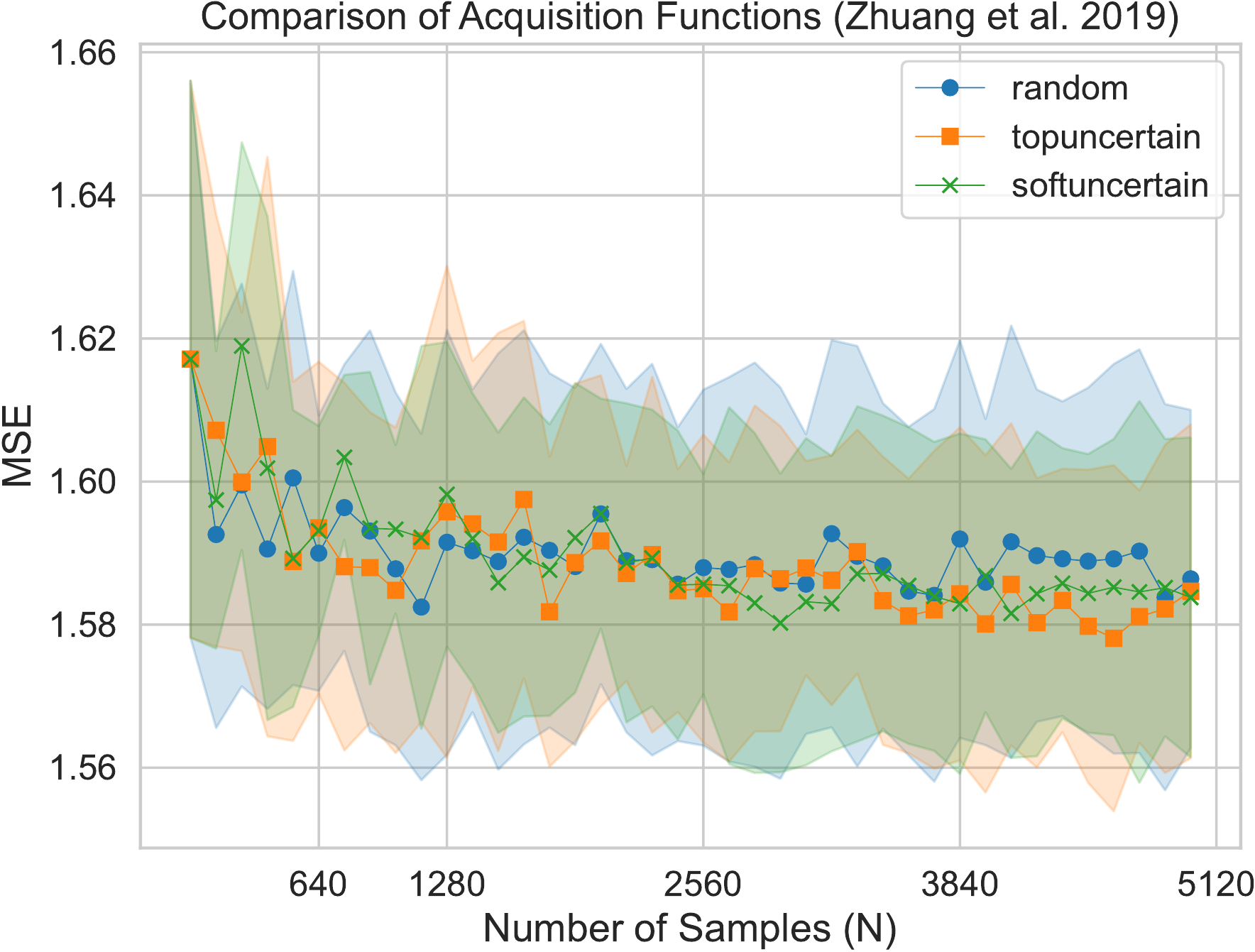}};
                        \end{tikzpicture}
                    }
                \end{subfigure}
                \&
                \\
\begin{subfigure}{0.275\columnwidth}
                    \hspace{-17mm}
                    \centering
                    \resizebox{\linewidth}{!}{
                        \begin{tikzpicture}
                            \node (img)  {\includegraphics[width=\textwidth]{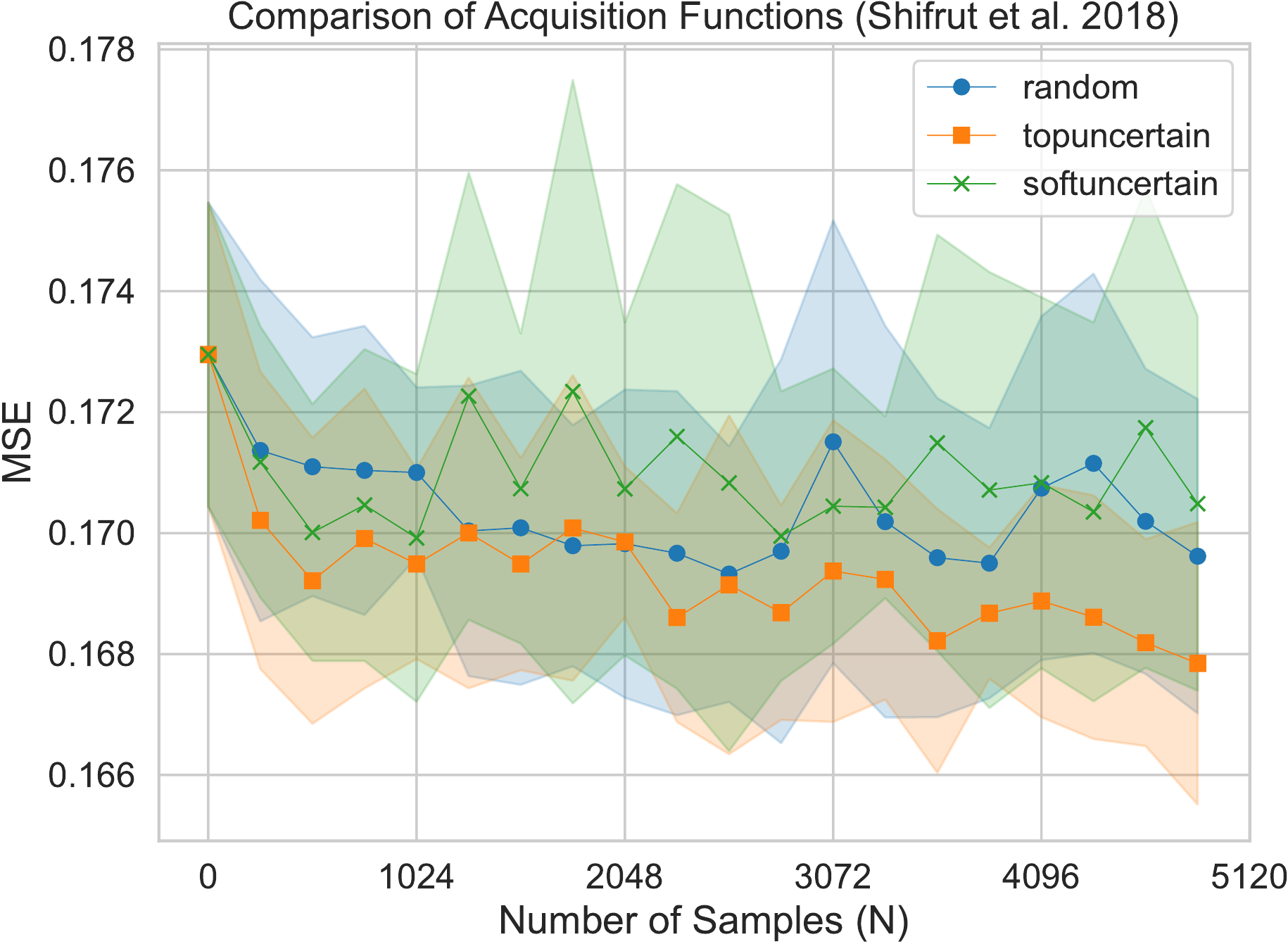}};
                        \end{tikzpicture}
                    }
                \end{subfigure}
                \&
                \begin{subfigure}{0.27\columnwidth}
                    \hspace{-23mm}
                    \centering
                    \resizebox{\linewidth}{!}{
                        \begin{tikzpicture}
                            \node (img)  {\includegraphics[width=\textwidth]{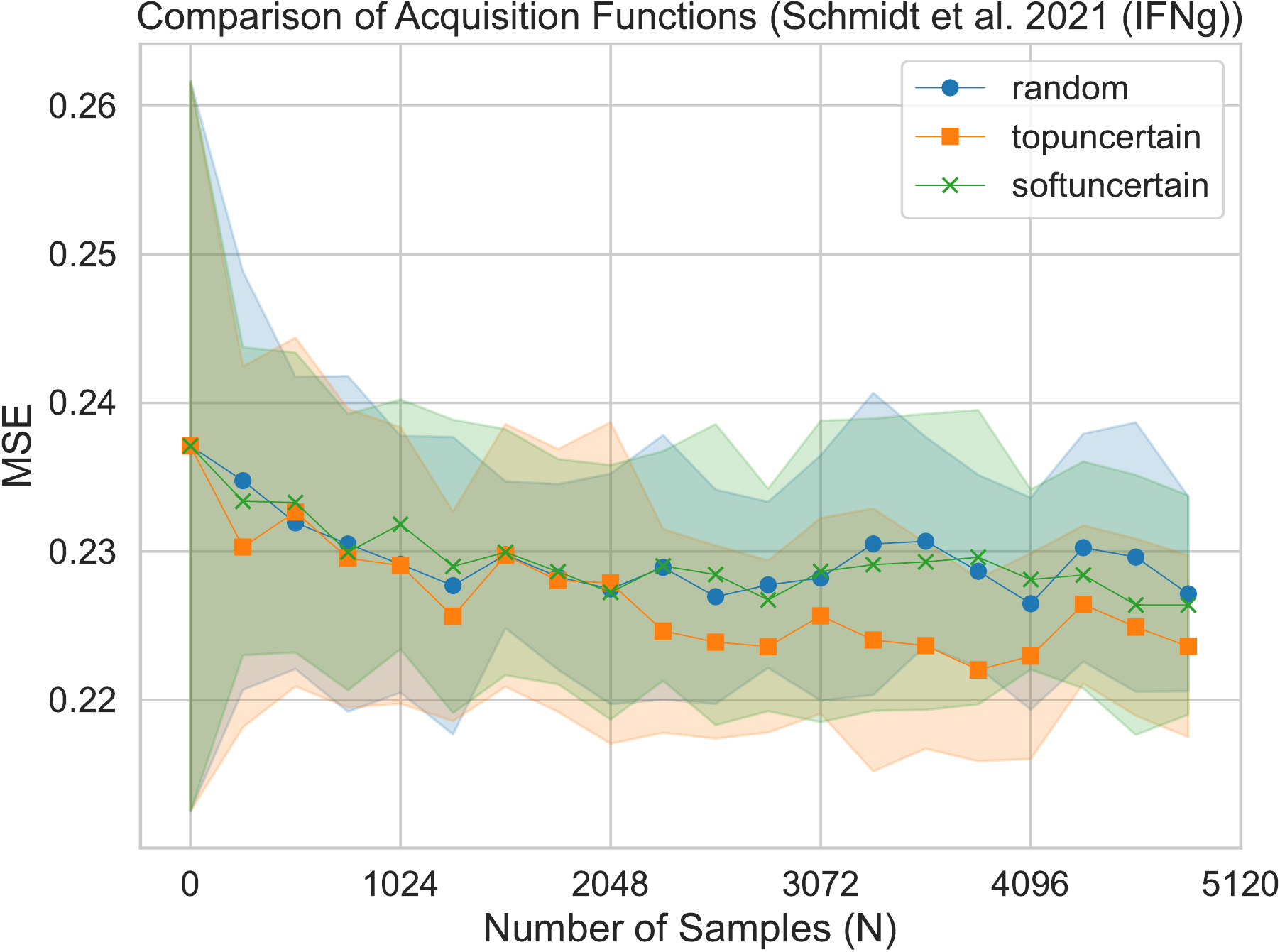}};
                        \end{tikzpicture}
                    }
                \end{subfigure}
                \&
                \begin{subfigure}{0.27\columnwidth}
                    \hspace{-28mm}
                    \centering
                    \resizebox{\linewidth}{!}{
                        \begin{tikzpicture}
                            \node (img)  {\includegraphics[width=\textwidth]{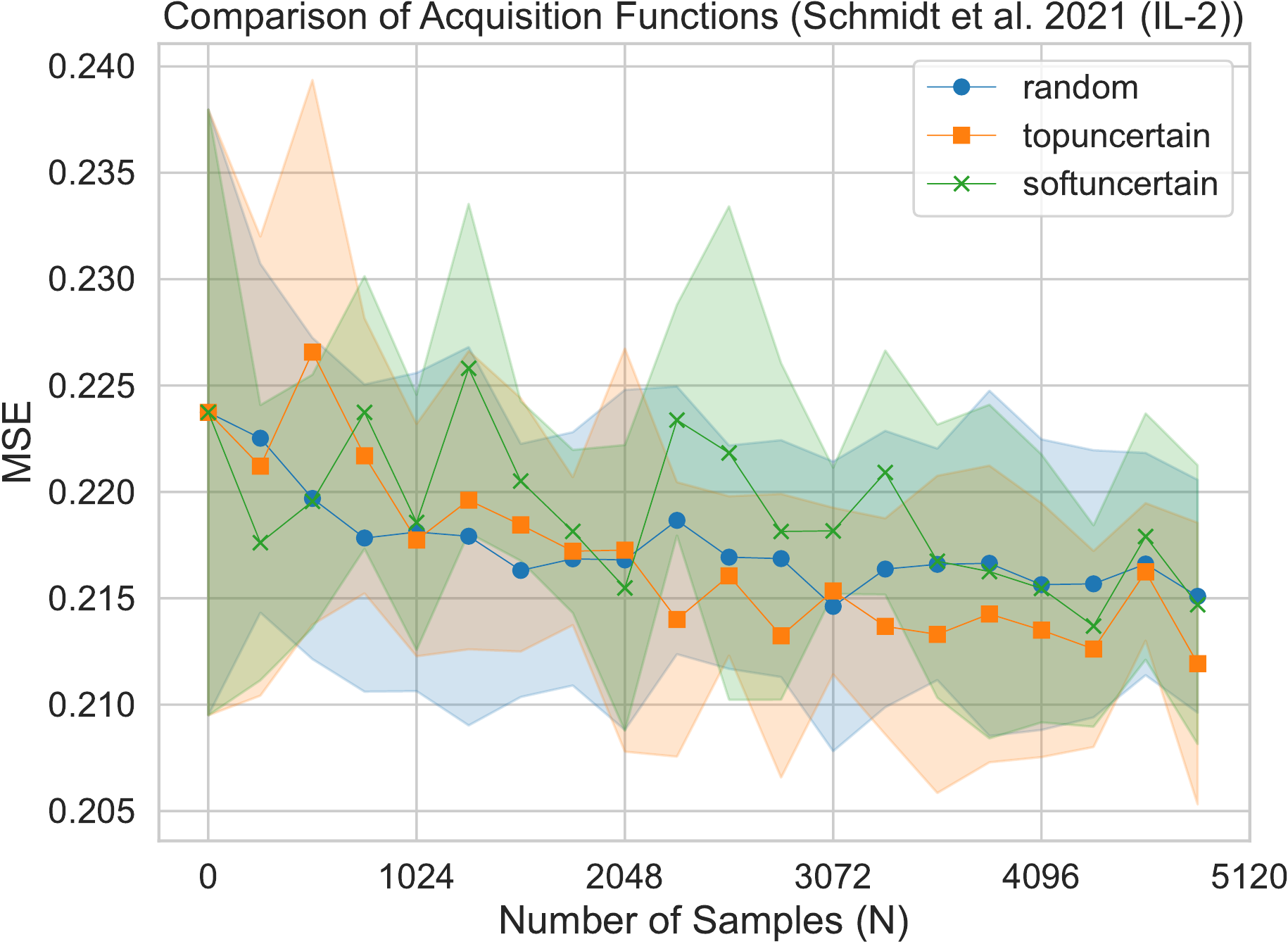}};
                        \end{tikzpicture}
                    }
                \end{subfigure}
                \&
                \begin{subfigure}{0.29\columnwidth}
                    \hspace{-32mm}
                    \centering
                    \resizebox{\linewidth}{!}{
                        \begin{tikzpicture}
                            \node (img)  {\includegraphics[width=\textwidth]{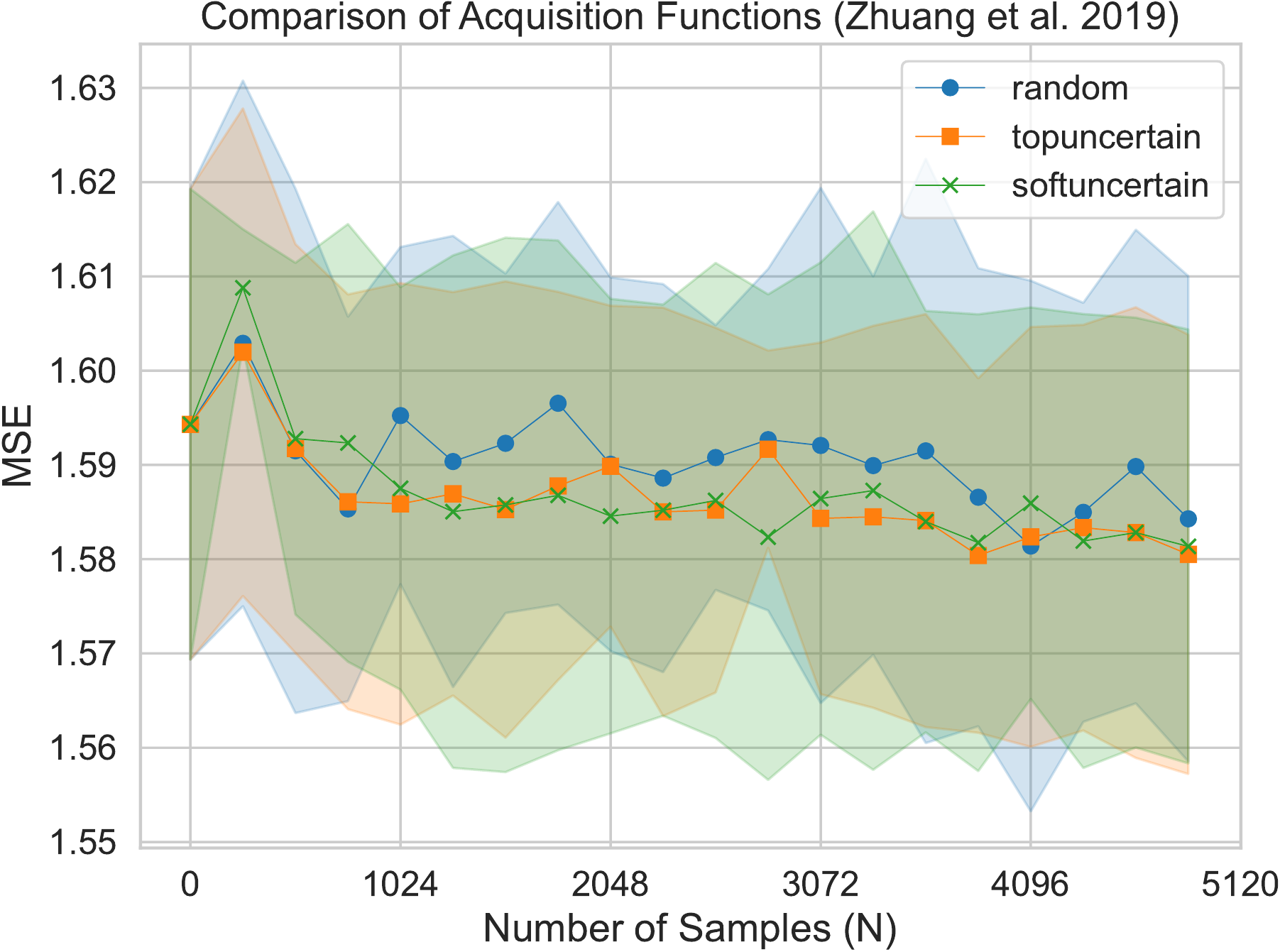}};
                        \end{tikzpicture}
                    }
                \end{subfigure}
                \&
                \\
\begin{subfigure}{0.28\columnwidth}
                    \hspace{-17mm}
                    \centering
                    \resizebox{\linewidth}{!}{
                        \begin{tikzpicture}
                            \node (img)  {\includegraphics[width=\textwidth]{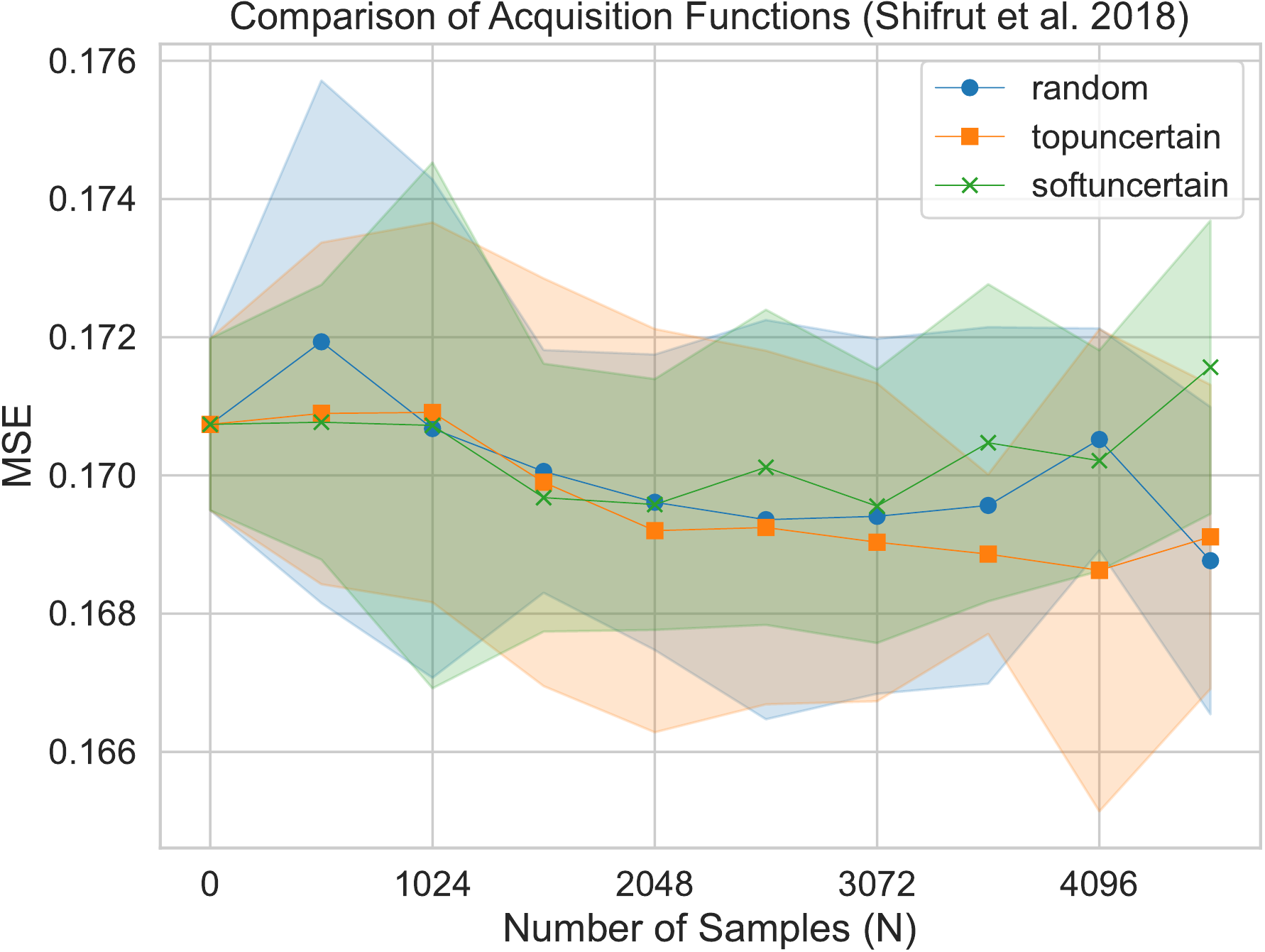}};
                        \end{tikzpicture}
                    }
                \end{subfigure}
                \&
                \begin{subfigure}{0.27\columnwidth}
                    \hspace{-23mm}
                    \centering
                    \resizebox{\linewidth}{!}{
                        \begin{tikzpicture}
                            \node (img)  {\includegraphics[width=\textwidth]{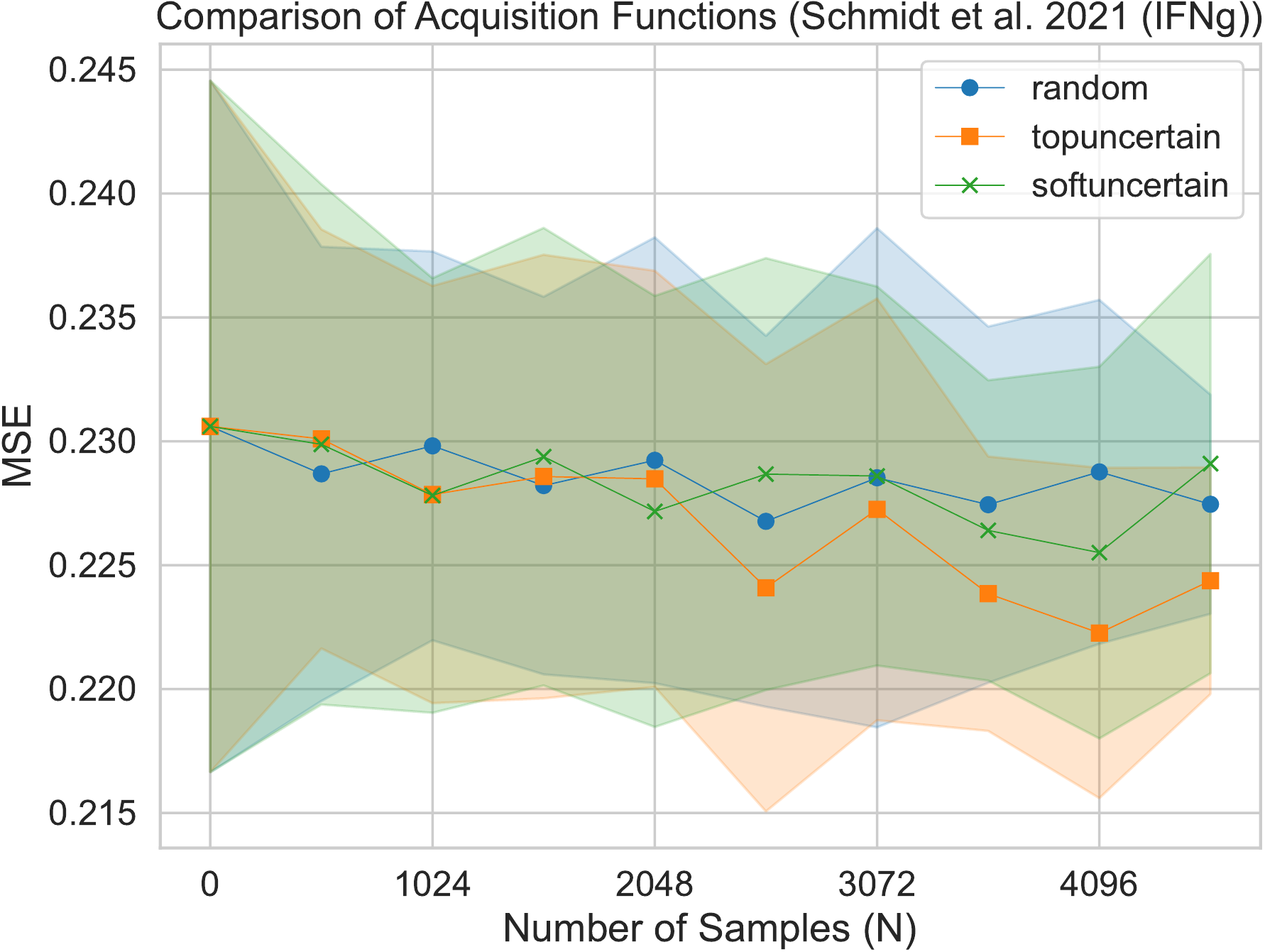}};
                        \end{tikzpicture}
                    }
                \end{subfigure}
                \&
                \begin{subfigure}{0.27\columnwidth}
                    \hspace{-28mm}
                    \centering
                    \resizebox{\linewidth}{!}{
                        \begin{tikzpicture}
                            \node (img)  {\includegraphics[width=\textwidth]{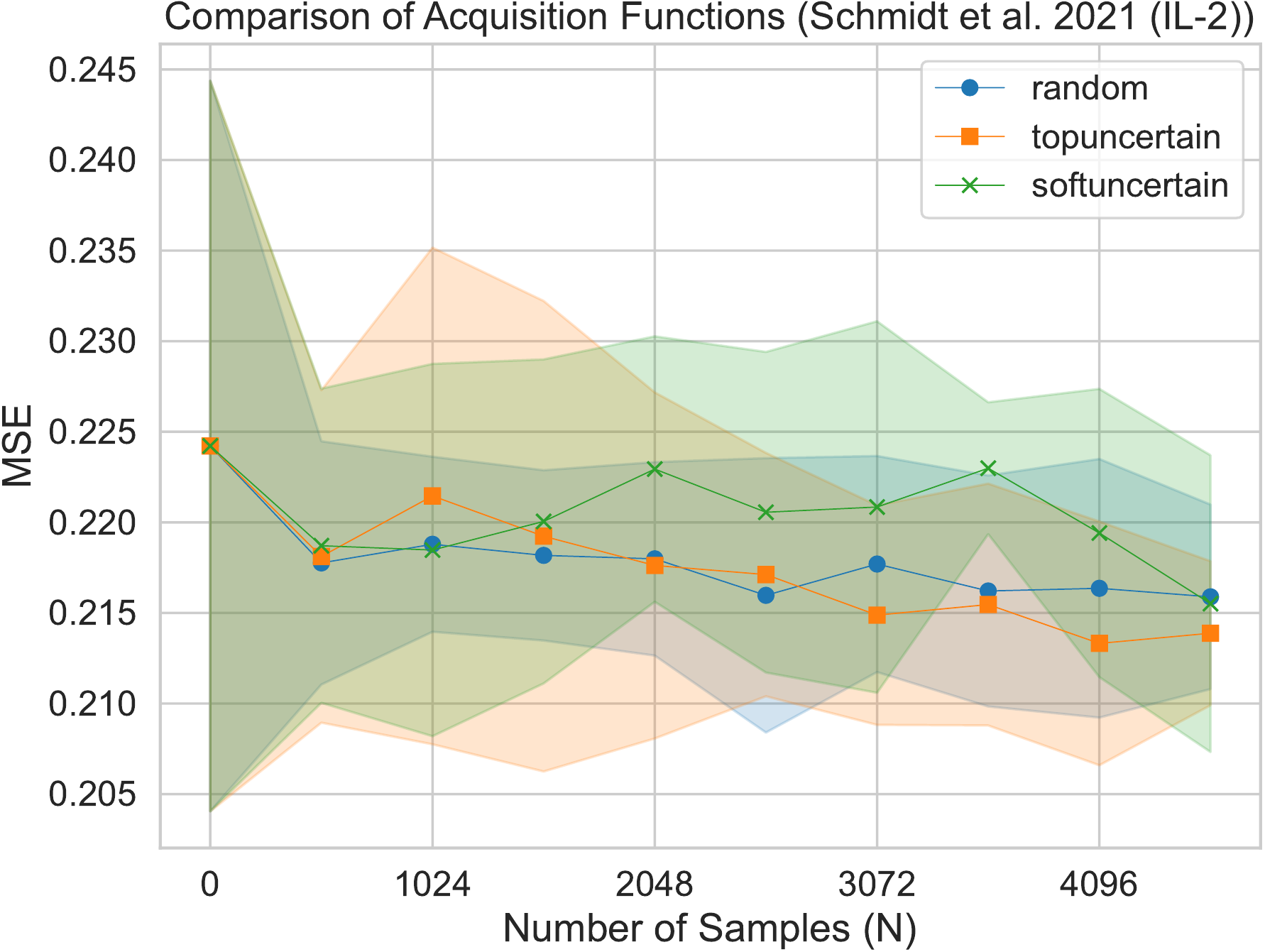}};
                        \end{tikzpicture}
                    }
                \end{subfigure}
                \&
                \begin{subfigure}{0.28\columnwidth}
                    \hspace{-32mm}
                    \centering
                    \resizebox{\linewidth}{!}{
                        \begin{tikzpicture}
                            \node (img)  {\includegraphics[width=\textwidth]{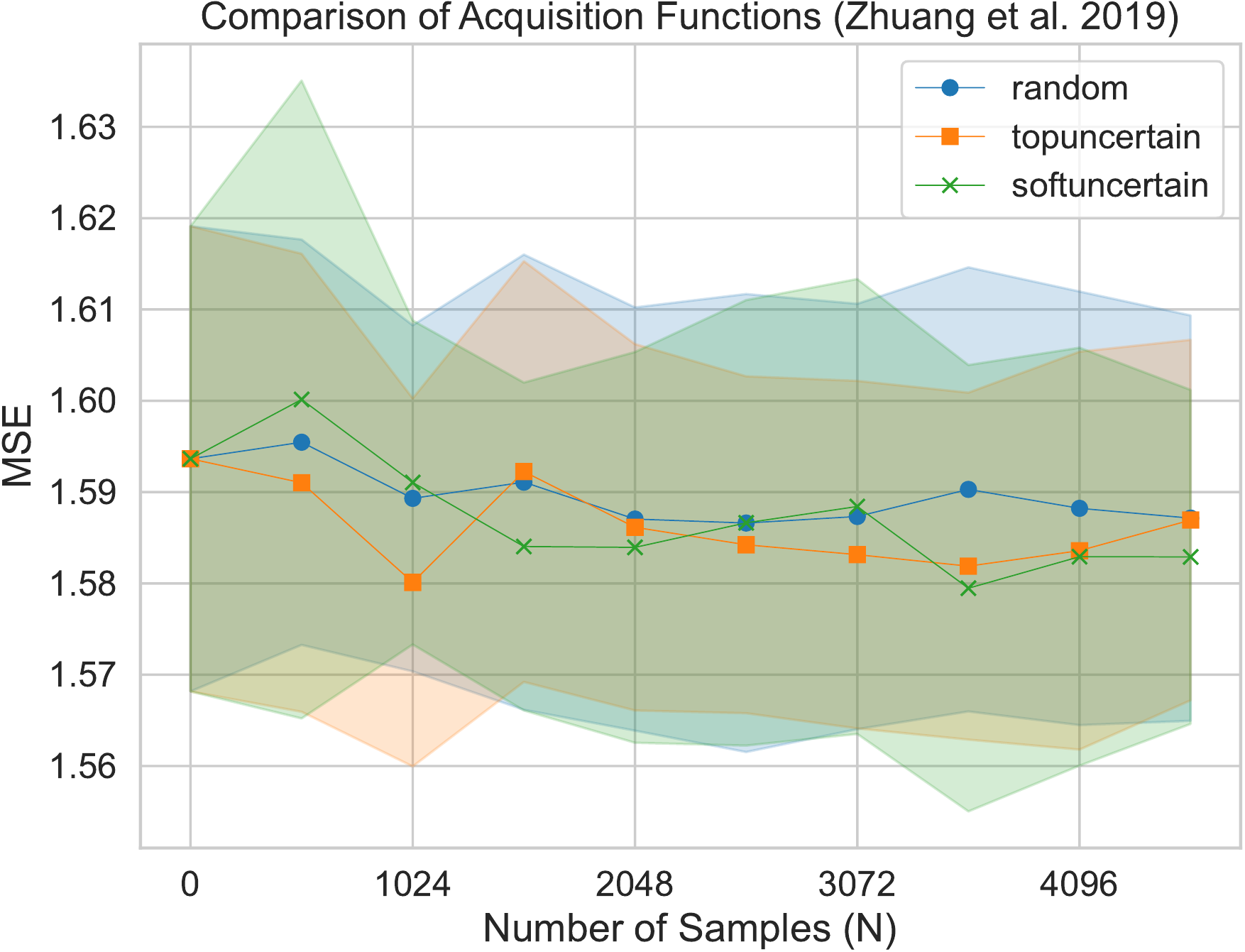}};
                        \end{tikzpicture}
                    }
                \end{subfigure}
                \&
                \\
            \\
           
            \\
            };
            \node [draw=none, rotate=90] at ([xshift=-8mm, yshift=2mm]fig-1-1.west) {\small batch size = 16};
            \node [draw=none, rotate=90] at ([xshift=-8mm, yshift=2mm]fig-2-1.west) {\small batch size = 32};
            \node [draw=none, rotate=90] at ([xshift=-8mm, yshift=2mm]fig-3-1.west) {\small batch size = 64};
            \node [draw=none, rotate=90] at ([xshift=-8mm, yshift=2mm]fig-4-1.west) {\small batch size = 128};
            \node [draw=none, rotate=90] at ([xshift=-8mm, yshift=2mm]fig-5-1.west) {\small batch size = 256};
            \node [draw=none, rotate=90] at ([xshift=-8mm, yshift=2mm]fig-6-1.west) {\small batch size = 512};
            \node [draw=none] at ([xshift=-6mm, yshift=3mm]fig-1-1.north) {\small Shifrut et al. 2018};
            \node [draw=none] at ([xshift=-9mm, yshift=3mm]fig-1-2.north) {\small Schmidt et al. 2021 (IFNg)};
            \node [draw=none] at ([xshift=-11mm, yshift=3mm]fig-1-3.north) {\small Schmidt et al. 2021 (IL-2)};
            \node [draw=none] at ([xshift=-13mm, yshift=2.5mm]fig-1-4.north) {\small Zhuang et al. 2019};
\end{tikzpicture}}
        \vspace{-7mm}
        \caption{The evaluation of the random forest model trained with {STRING} treatment descriptors at each active learning cycle for 4 datasets and 6 acquisition batch sizes. In each plot, the x-axis is the active learning cycles multiplied by the acquisition bath size that gives the total number of data points collected so far. The y-axis is the test MSE error evaluated on the test data.}
        \vspace{-5mm}
        \label{fig:rf_feat_string_alldatasets_allbathcsizes}
    \end{figure*}

\begin{figure*}
    \vspace{-2mm}
        \centering
        \makebox[0.72\paperwidth]{\begin{tikzpicture}[ampersand replacement=\&]
            \matrix (fig) [matrix of nodes]{ 
\begin{subfigure}{0.27\columnwidth}
                    \hspace{-17mm}
                    \centering
                    \resizebox{\linewidth}{!}{
                        \begin{tikzpicture}
                            \node (img)  {\includegraphics[width=\textwidth]{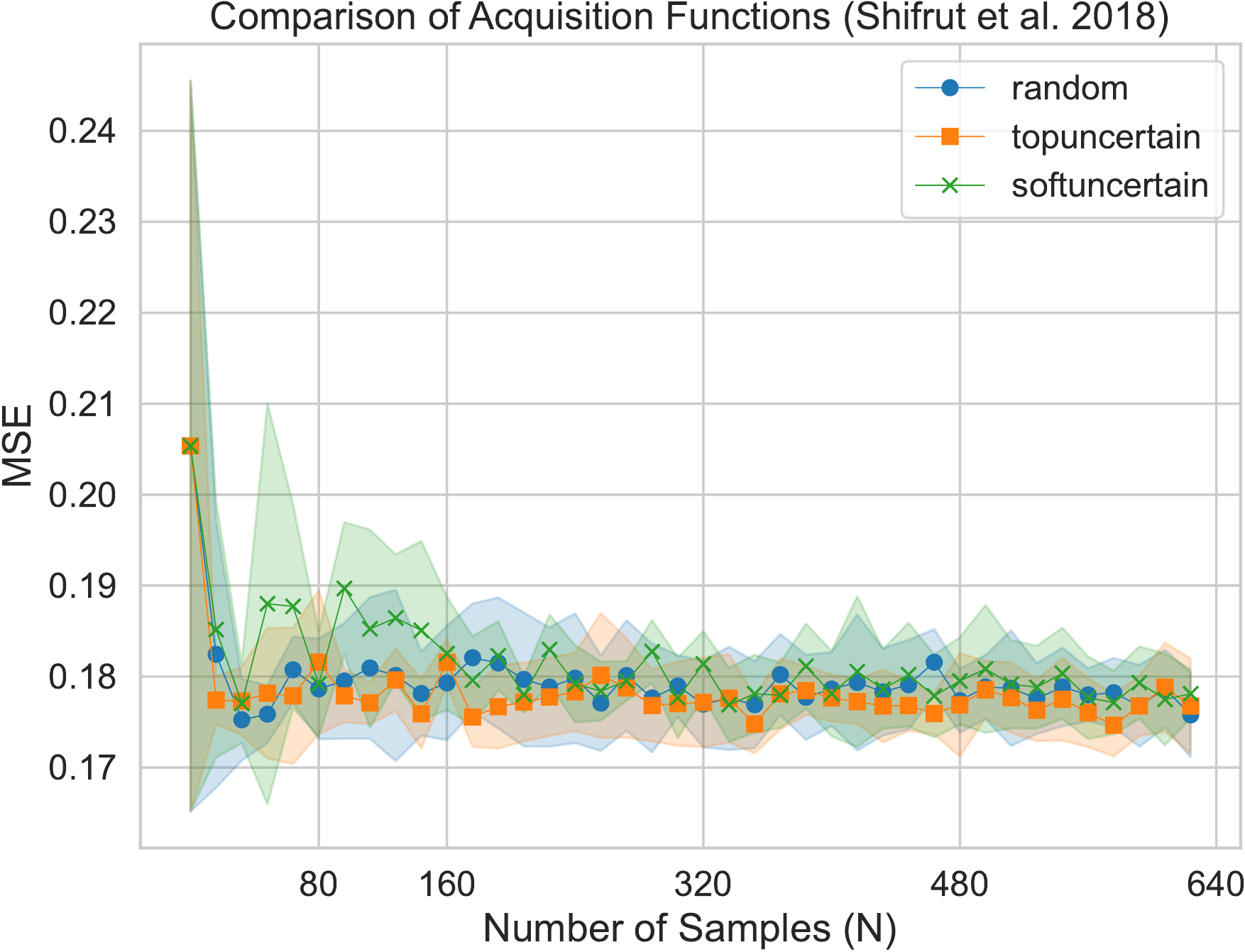}};
                            \node[below=of img, node distance=0cm, yshift=1.2cm, font=\color{black}] {\scalebox{0.35}{Sparsity}};
                        \end{tikzpicture}
                    }
                \end{subfigure}
                \&
                 \begin{subfigure}{0.27\columnwidth}
                    \hspace{-23mm}
                    \centering
                    \resizebox{\linewidth}{!}{
                        \begin{tikzpicture}
                            \node (img)  {\includegraphics[width=\textwidth]{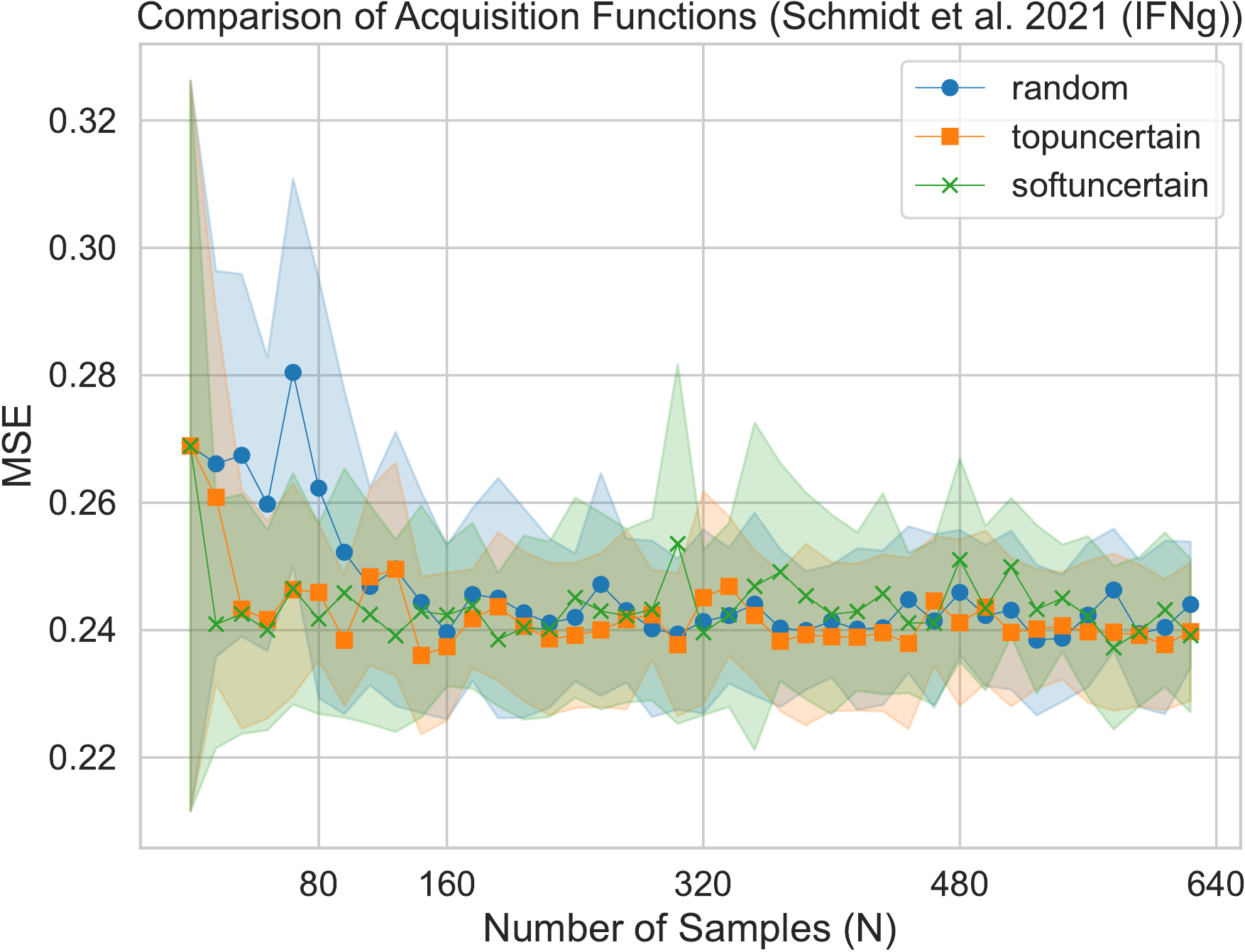}};
                        \end{tikzpicture}
                    }
                \end{subfigure}
                \&
                 \begin{subfigure}{0.27\columnwidth}
                    \hspace{-28mm}
                    \centering
                    \resizebox{\linewidth}{!}{
                        \begin{tikzpicture}
                            \node (img)  {\includegraphics[width=\textwidth]{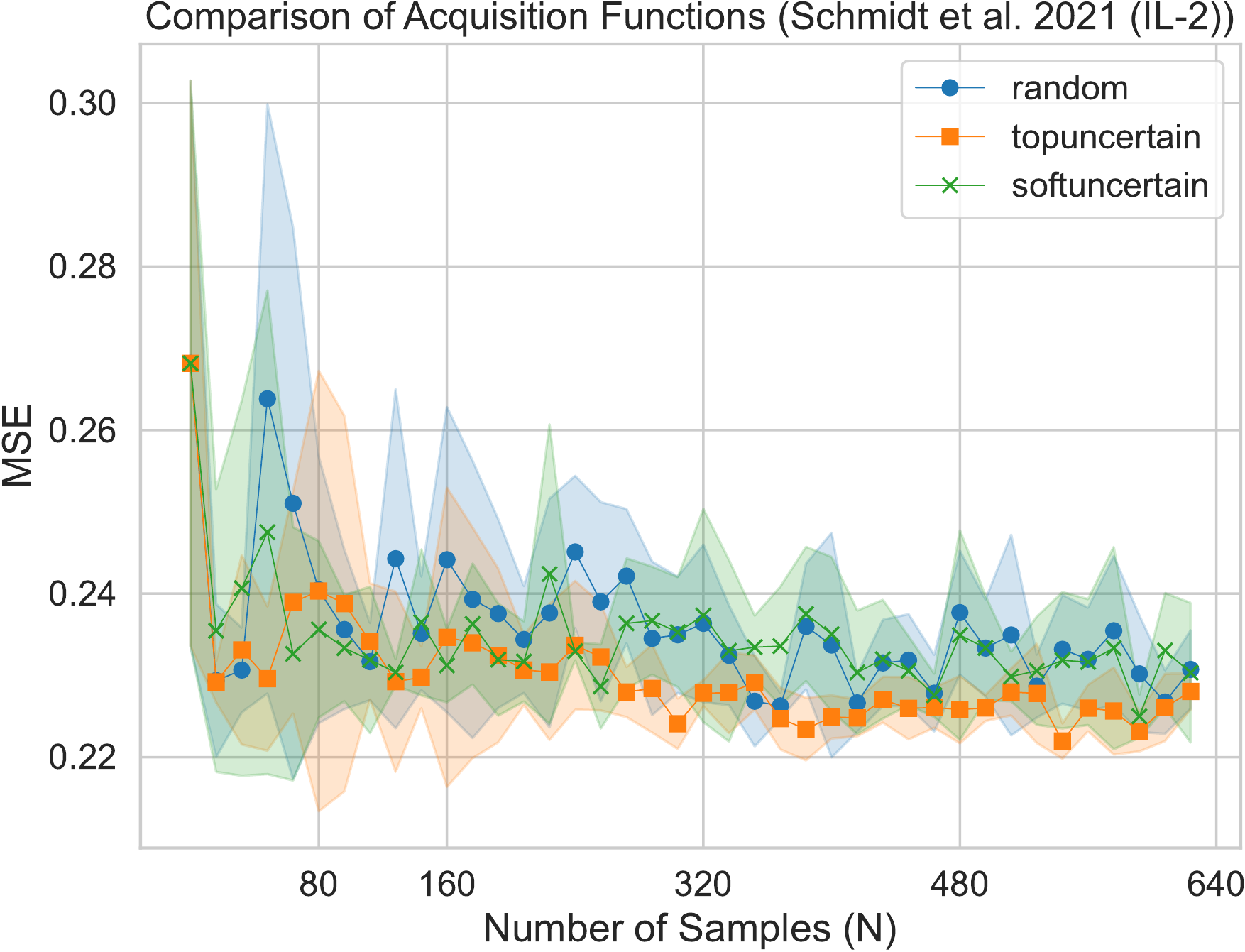}};
                        \end{tikzpicture}
                    }
                \end{subfigure}
                \&
                \begin{subfigure}{0.28\columnwidth}
                    \hspace{-32mm}
                    \centering
                    \resizebox{\linewidth}{!}{
                        \begin{tikzpicture}
                            \node (img)  {\includegraphics[width=\textwidth]{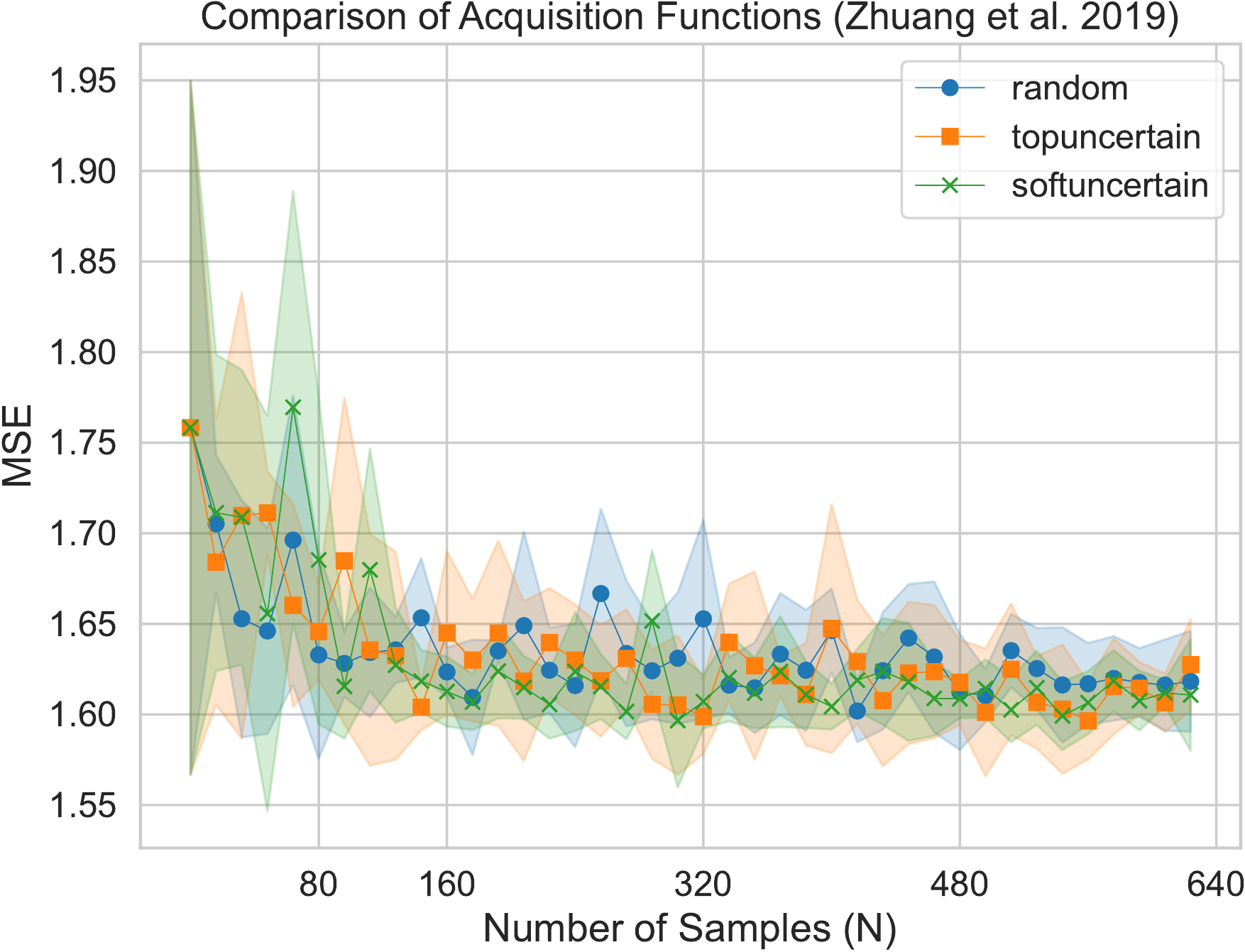}};
                        \end{tikzpicture}
                    }
                \end{subfigure}
                \&
            \\
\begin{subfigure}{0.27\columnwidth}
                    \hspace{-17mm}
                    \centering
                    \resizebox{\linewidth}{!}{
                        \begin{tikzpicture}
                            \node (img)  {\includegraphics[width=\textwidth]{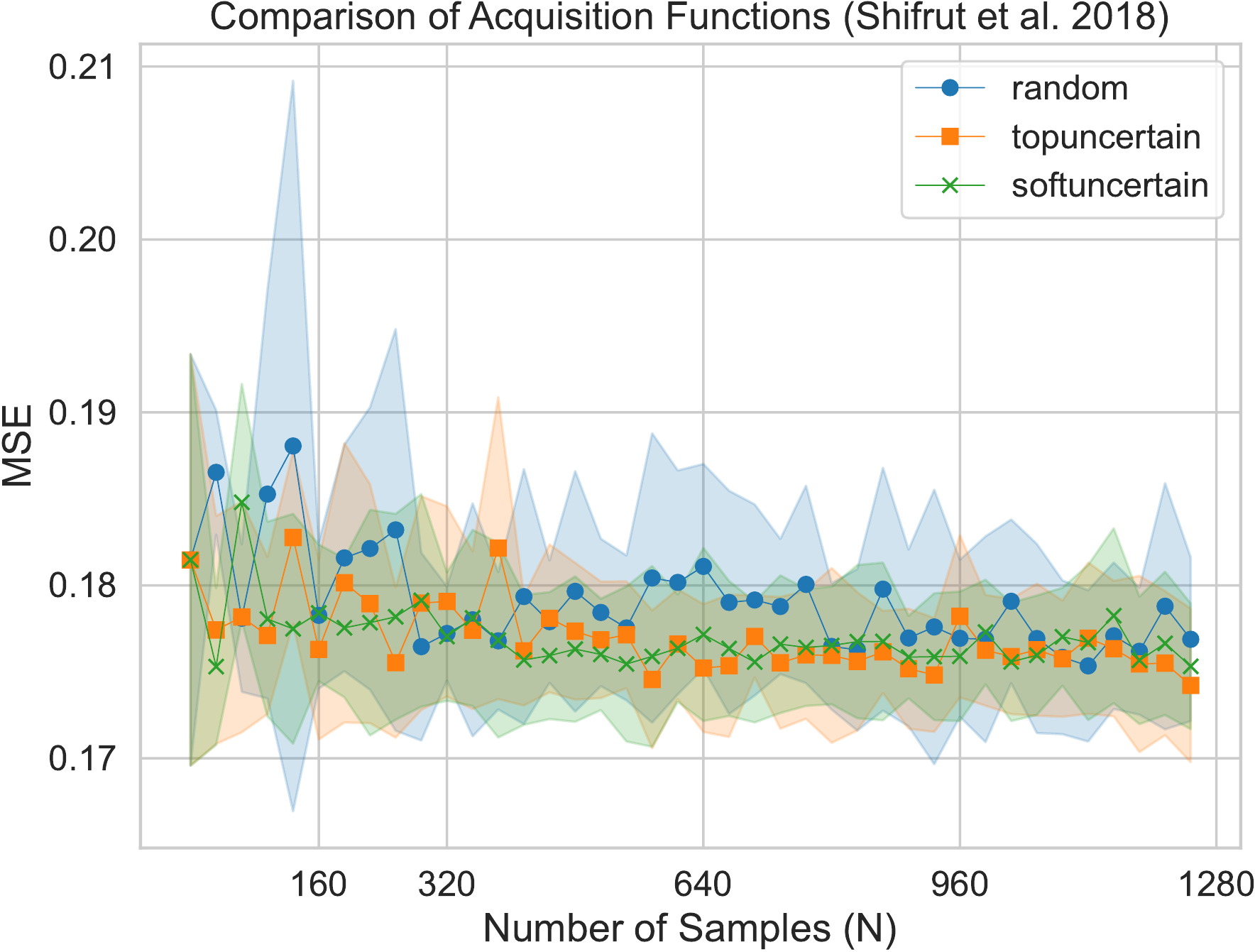}};
                        \end{tikzpicture}
                    }
                \end{subfigure}
                \&
                \begin{subfigure}{0.27\columnwidth}
                    \hspace{-23mm}
                    \centering
                    \resizebox{\linewidth}{!}{
                        \begin{tikzpicture}
                            \node (img)  {\includegraphics[width=\textwidth]{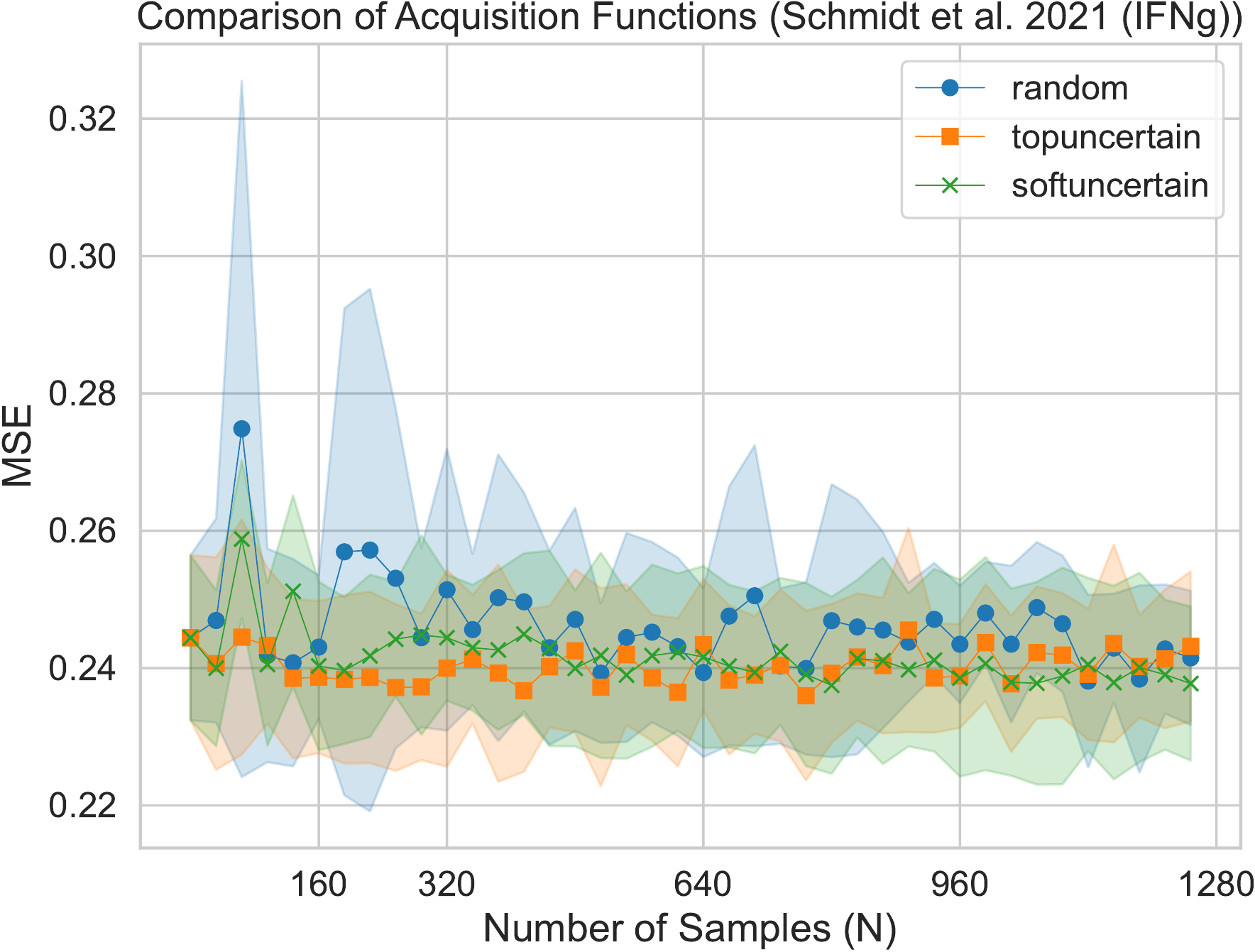}};
                        \end{tikzpicture}
                    }
                \end{subfigure}
                \&
                \begin{subfigure}{0.27\columnwidth}
                    \hspace{-28mm}
                    \centering
                    \resizebox{\linewidth}{!}{
                        \begin{tikzpicture}
                            \node (img)  {\includegraphics[width=\textwidth]{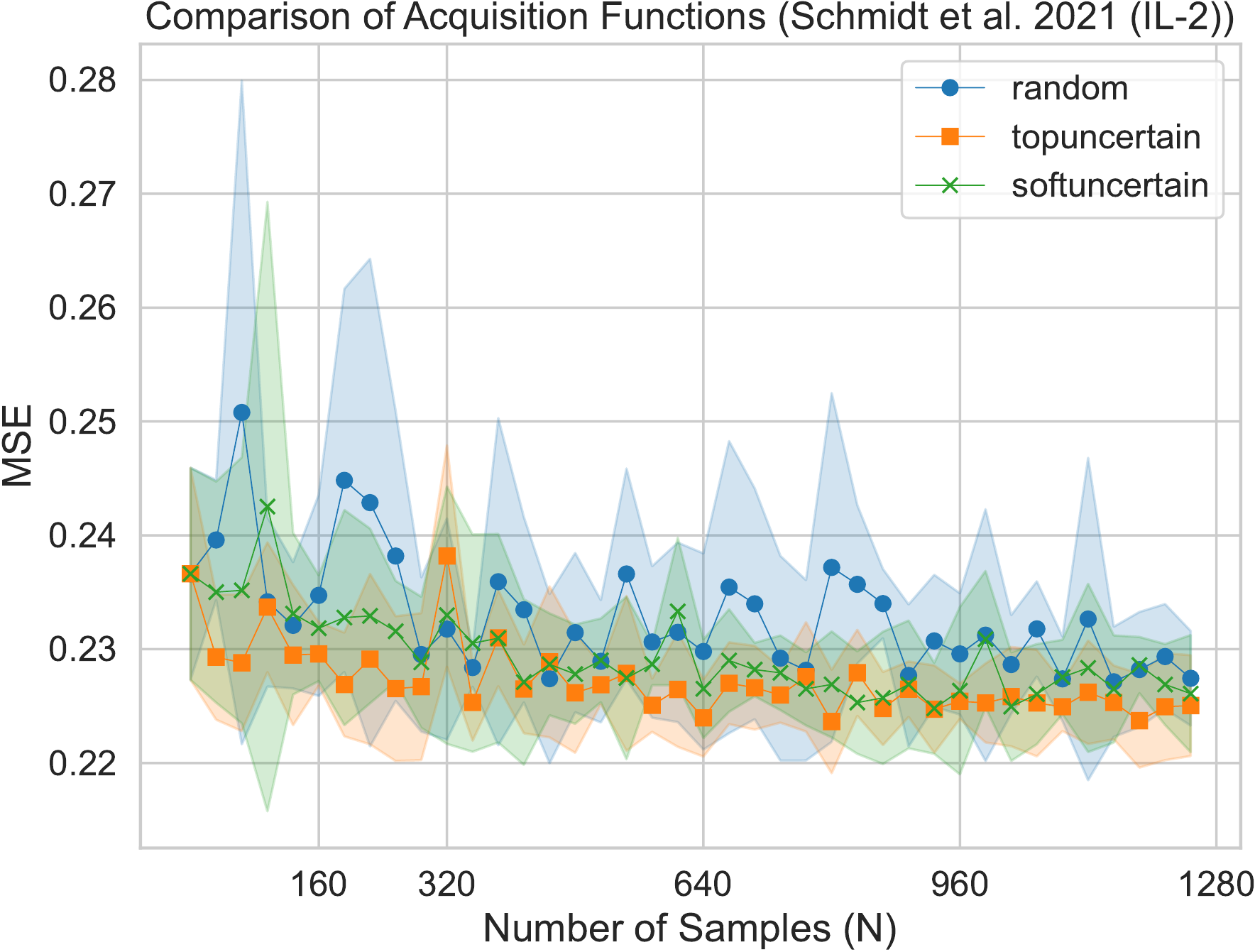}};
                        \end{tikzpicture}
                    }
                \end{subfigure}
                \&
                \begin{subfigure}{0.28\columnwidth}
                    \hspace{-32mm}
                    \centering
                    \resizebox{\linewidth}{!}{
                        \begin{tikzpicture}
                            \node (img)  {\includegraphics[width=\textwidth]{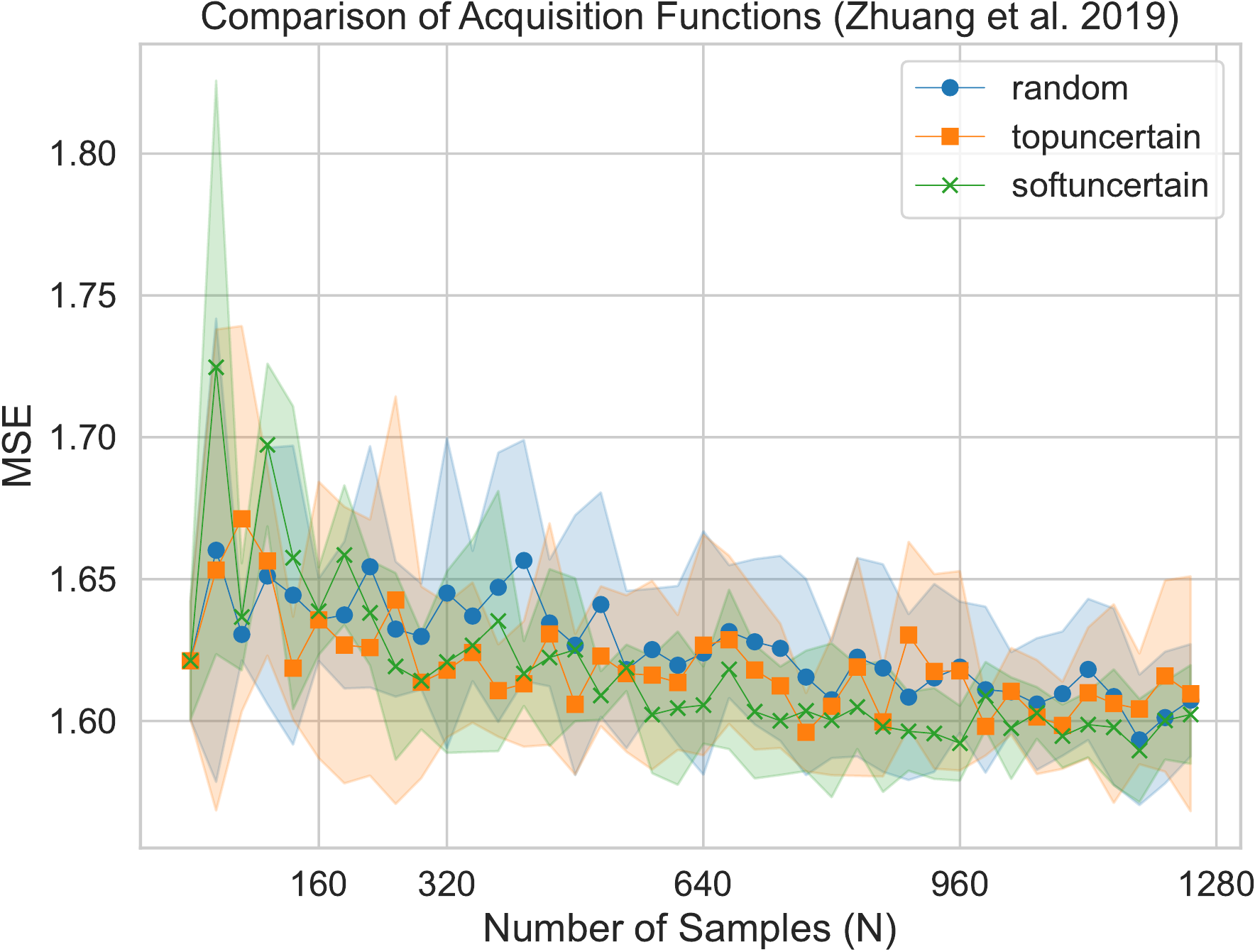}};
                        \end{tikzpicture}
                    }
                \end{subfigure}
                \&
                \\
\begin{subfigure}{0.27\columnwidth}
                    \hspace{-17mm}
                    \centering
                    \resizebox{\linewidth}{!}{
                        \begin{tikzpicture}
                            \node (img)  {\includegraphics[width=\textwidth]{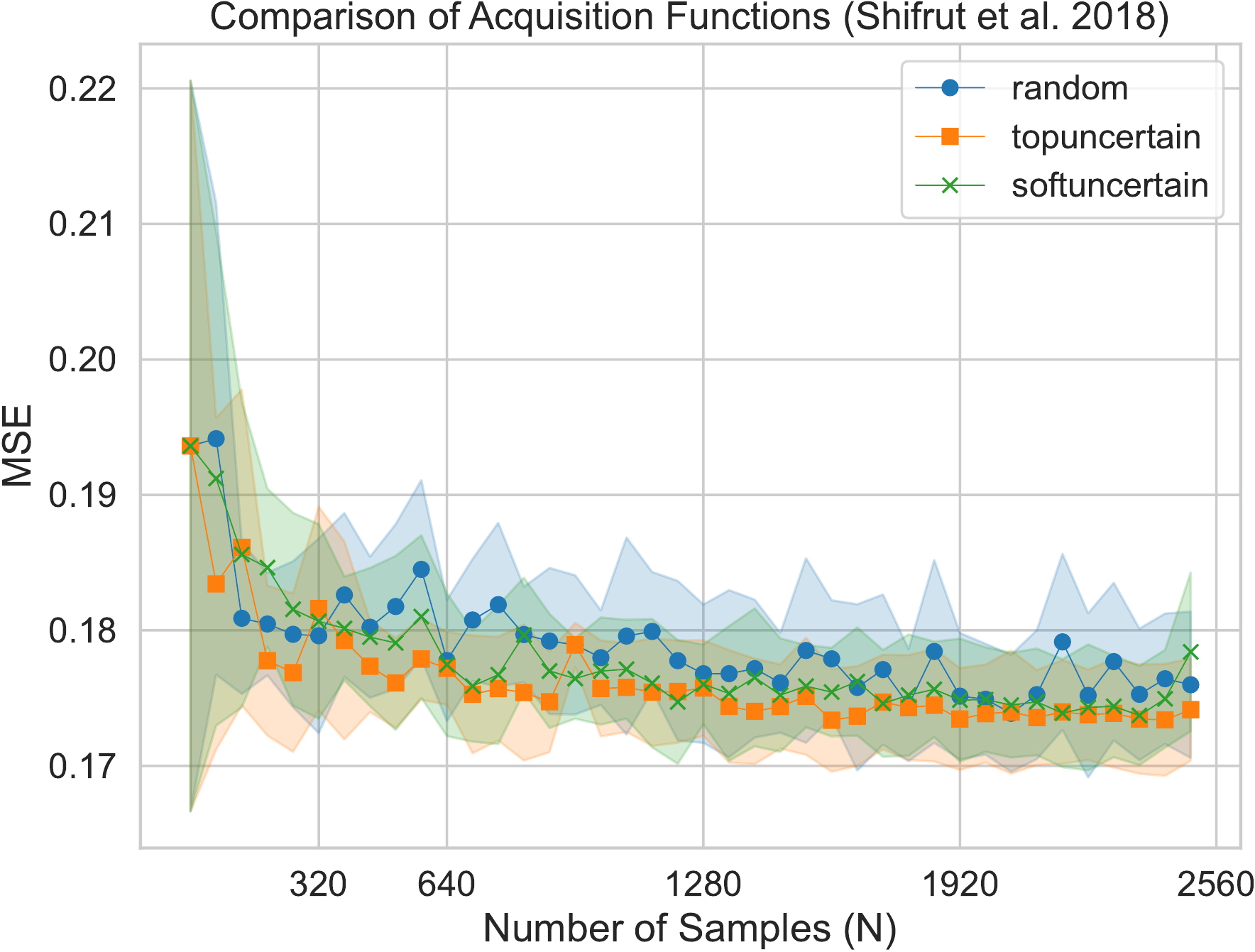}};
                        \end{tikzpicture}
                    }
                \end{subfigure}
                \&
                \begin{subfigure}{0.27\columnwidth}
                    \hspace{-23mm}
                    \centering
                    \resizebox{\linewidth}{!}{
                        \begin{tikzpicture}
                            \node (img)  {\includegraphics[width=\textwidth]{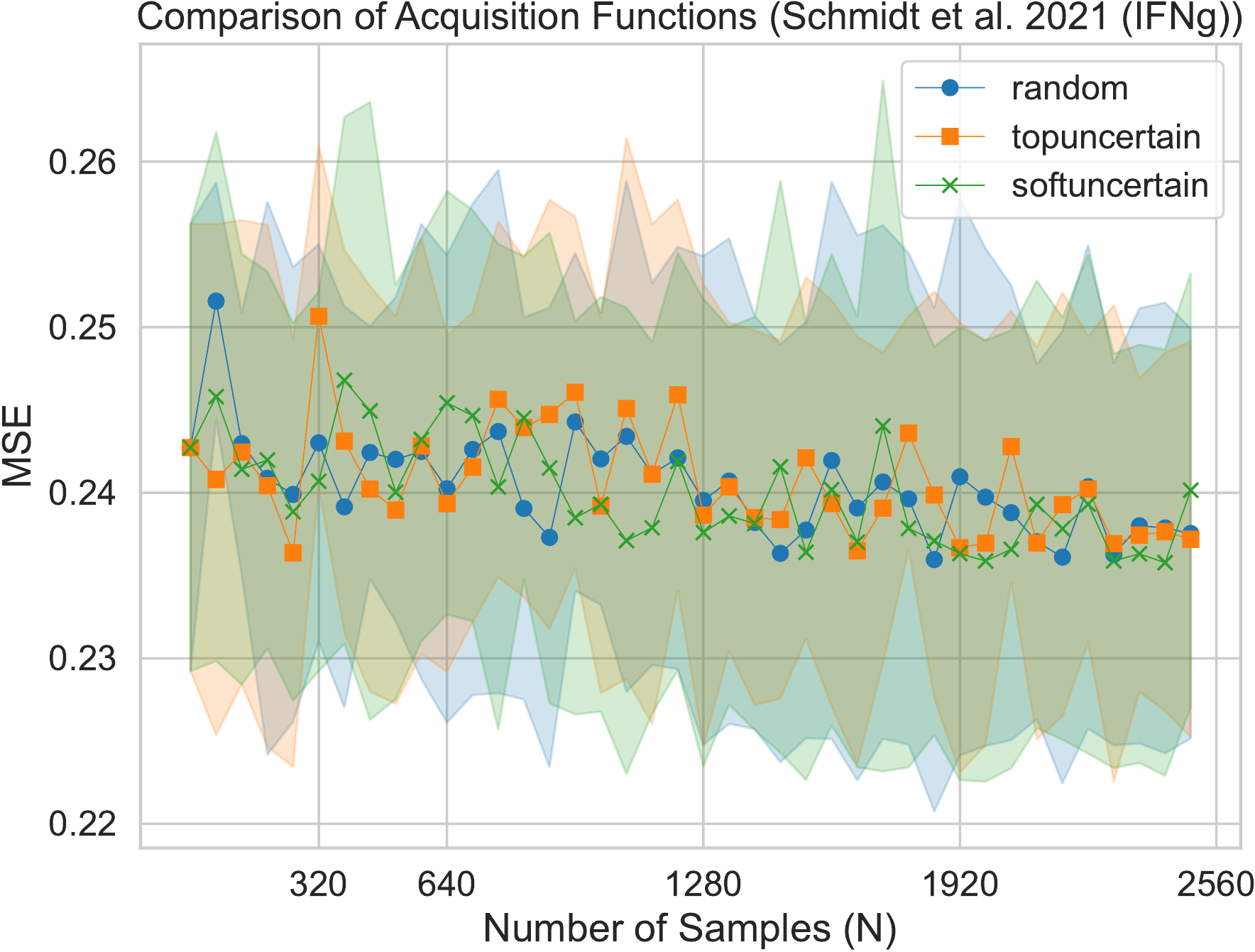}};
                        \end{tikzpicture}
                    }
                \end{subfigure}
                \&
                \begin{subfigure}{0.27\columnwidth}
                    \hspace{-28mm}
                    \centering
                    \resizebox{\linewidth}{!}{
                        \begin{tikzpicture}
                            \node (img)  {\includegraphics[width=\textwidth]{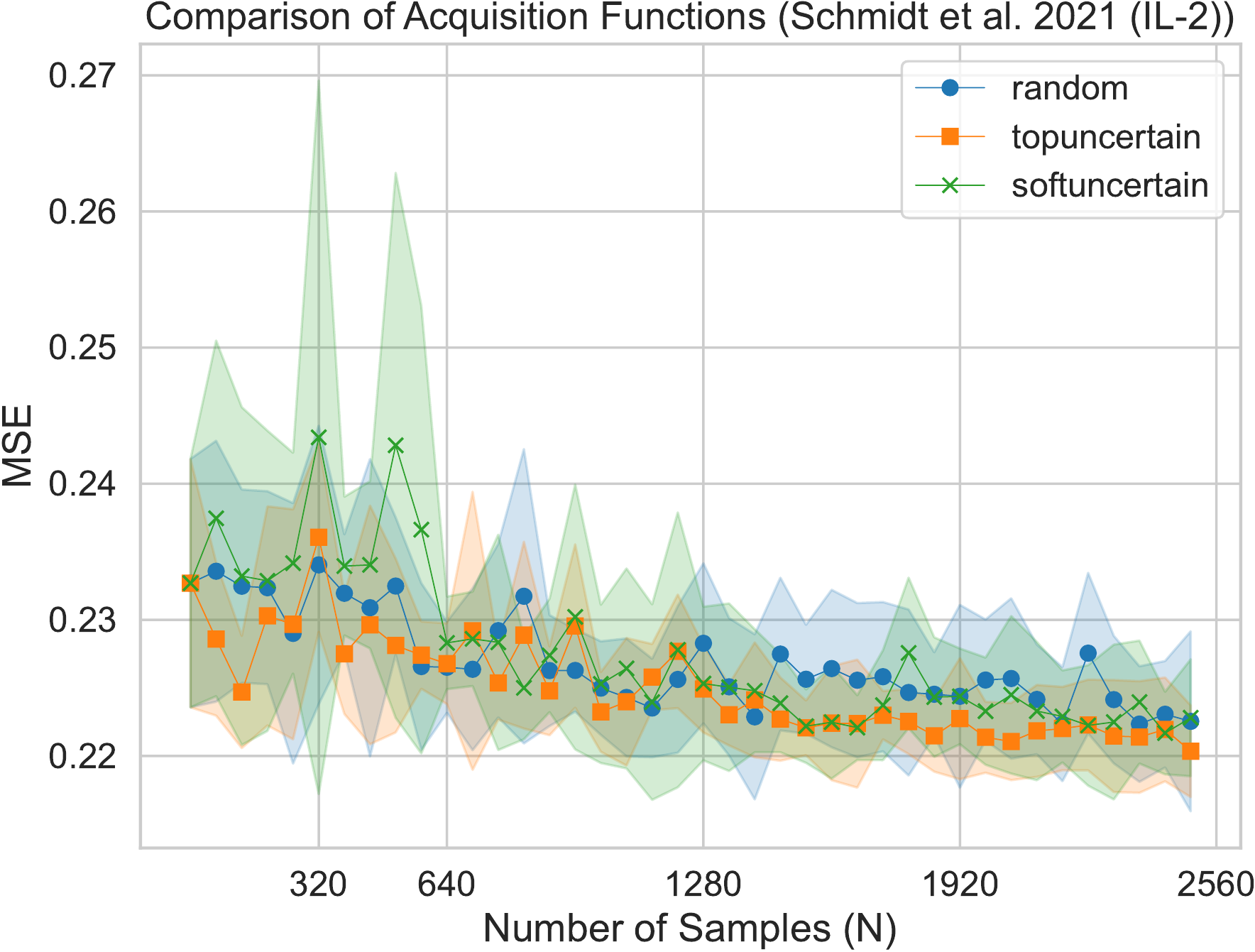}};
                        \end{tikzpicture}
                    }
                \end{subfigure}
                \&
                \begin{subfigure}{0.28\columnwidth}
                    \hspace{-32mm}
                    \centering
                    \resizebox{\linewidth}{!}{
                        \begin{tikzpicture}
                            \node (img)  {\includegraphics[width=\textwidth]{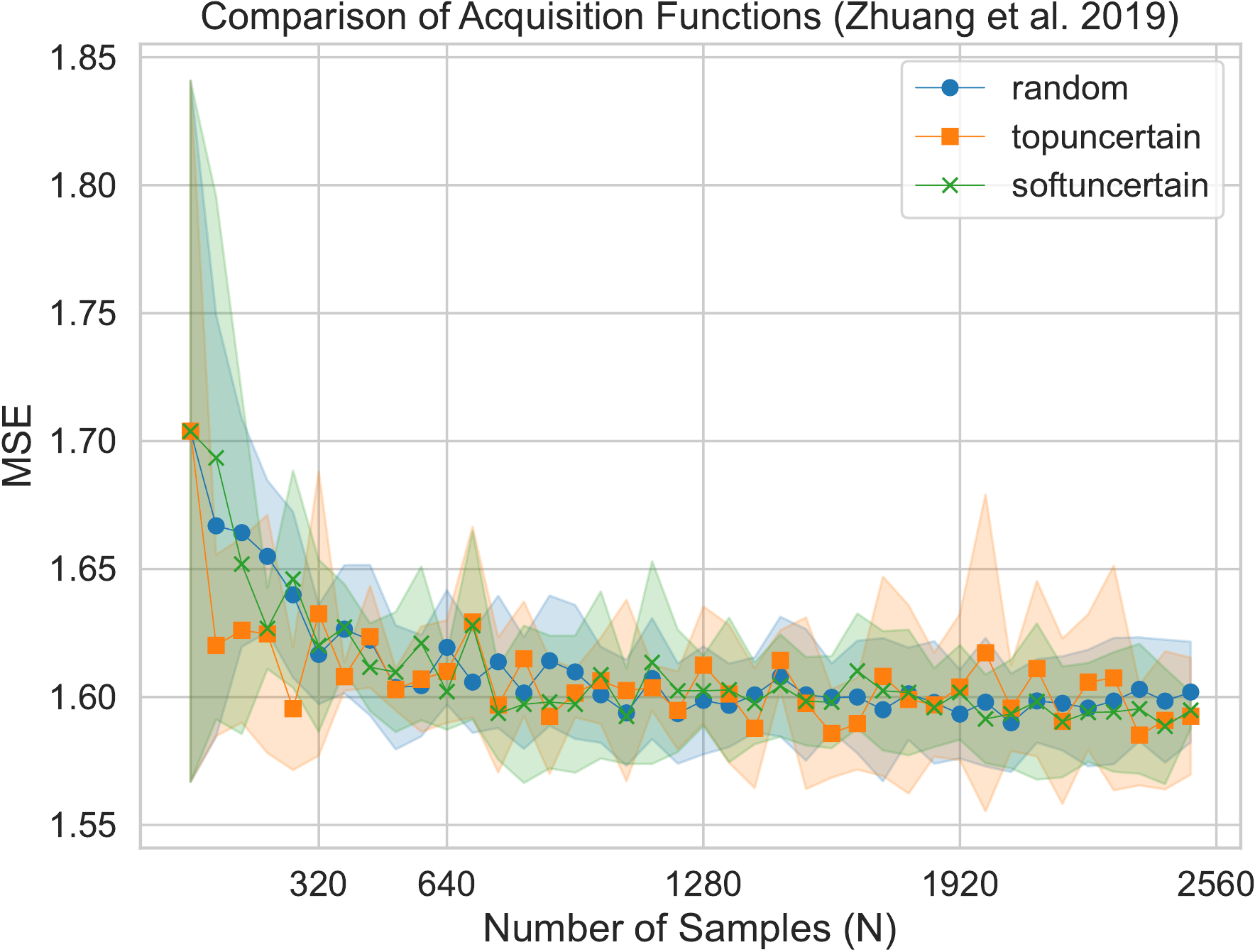}};
                        \end{tikzpicture}
                    }
                \end{subfigure}
                \&
                \\
\begin{subfigure}{0.27\columnwidth}
                    \hspace{-17mm}
                    \centering
                    \resizebox{\linewidth}{!}{
                        \begin{tikzpicture}
                            \node (img)  {\includegraphics[width=\textwidth]{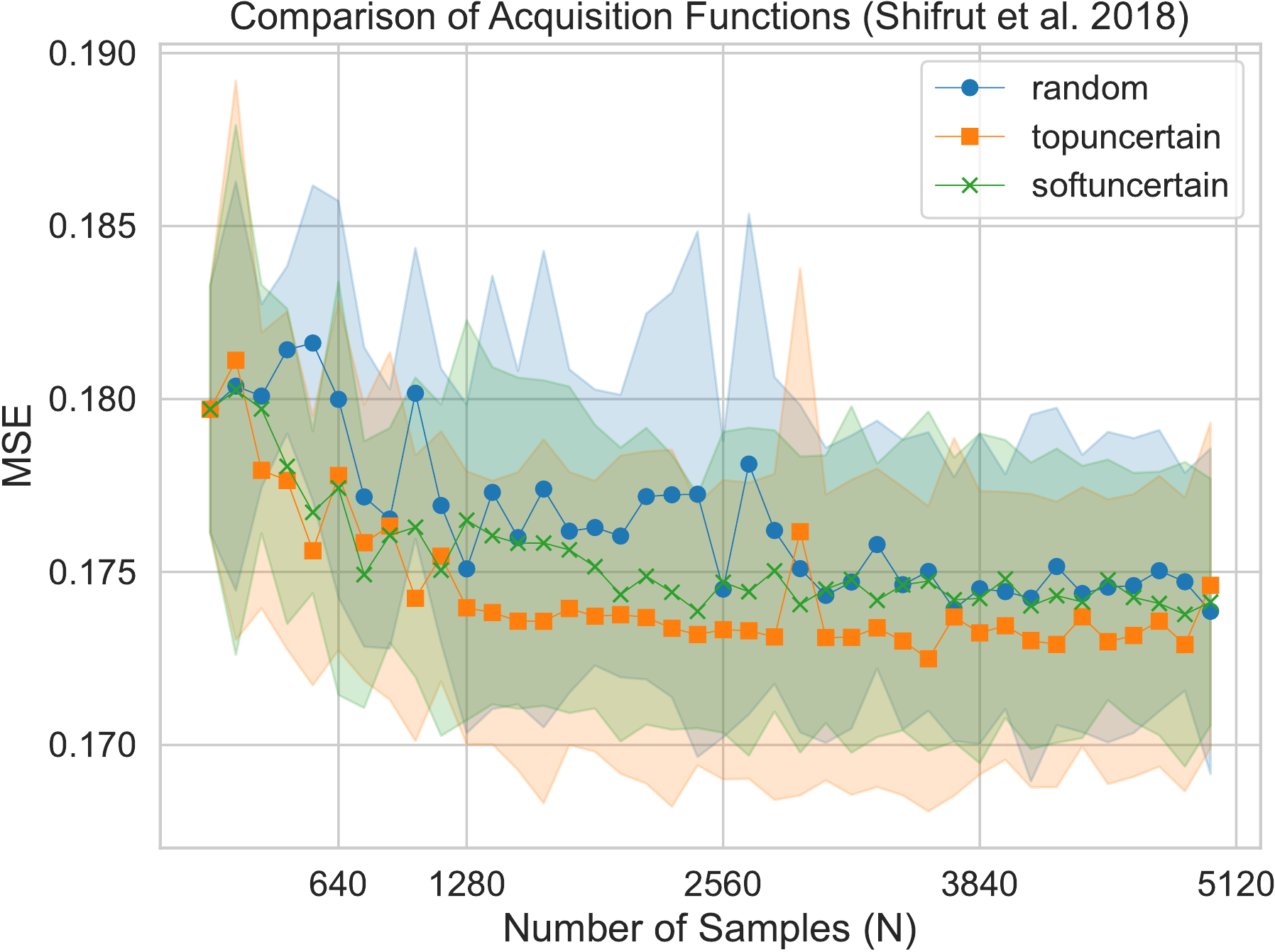}};
                        \end{tikzpicture}
                    }
                \end{subfigure}
                \&
                \begin{subfigure}{0.27\columnwidth}
                    \hspace{-23mm}
                    \centering
                    \resizebox{\linewidth}{!}{
                        \begin{tikzpicture}
                            \node (img)  {\includegraphics[width=\textwidth]{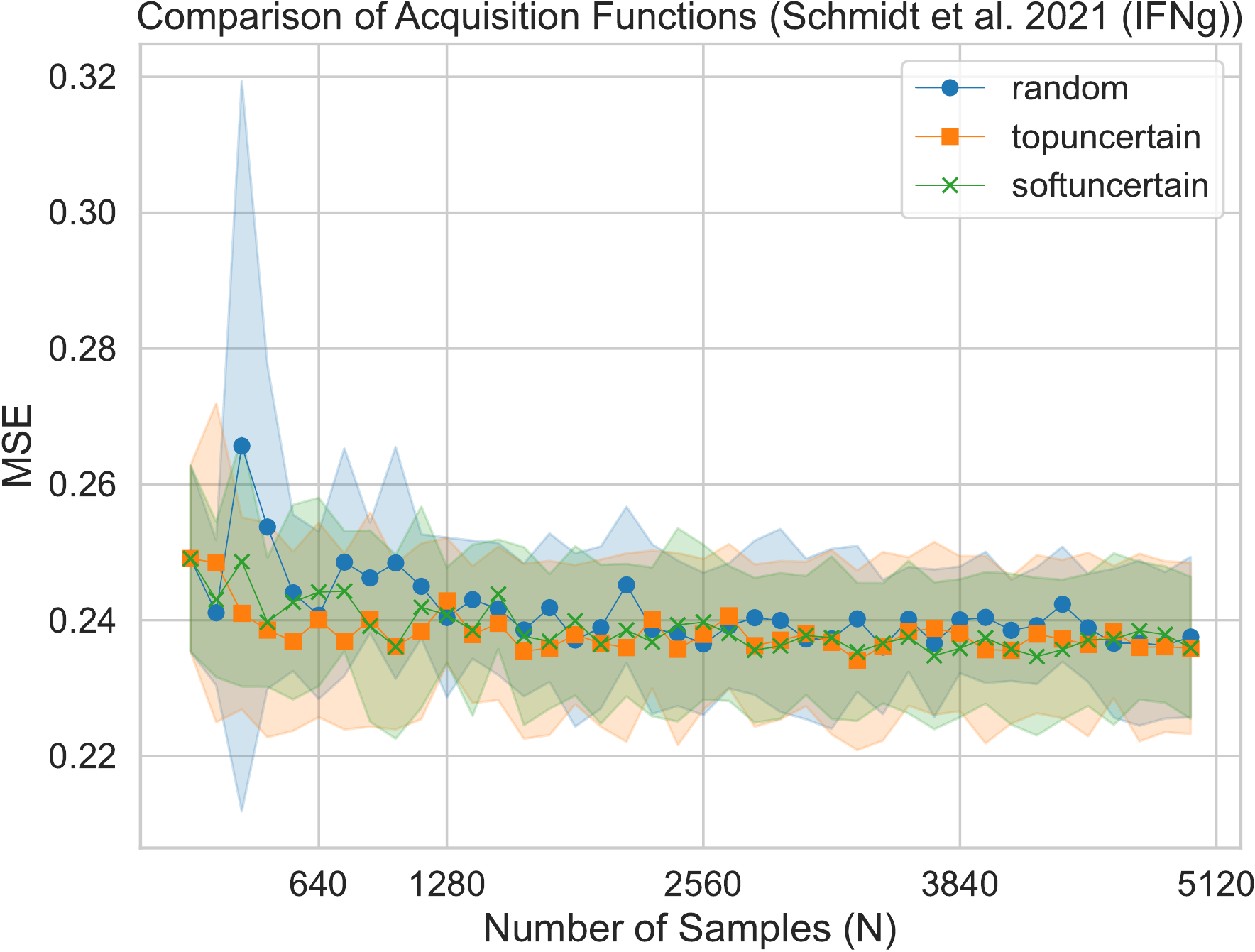}};
                        \end{tikzpicture}
                    }
                \end{subfigure}
                \&
                \begin{subfigure}{0.28\columnwidth}
                    \hspace{-28mm}
                    \centering
                    \resizebox{\linewidth}{!}{
                        \begin{tikzpicture}
                            \node (img)  {\includegraphics[width=\textwidth]{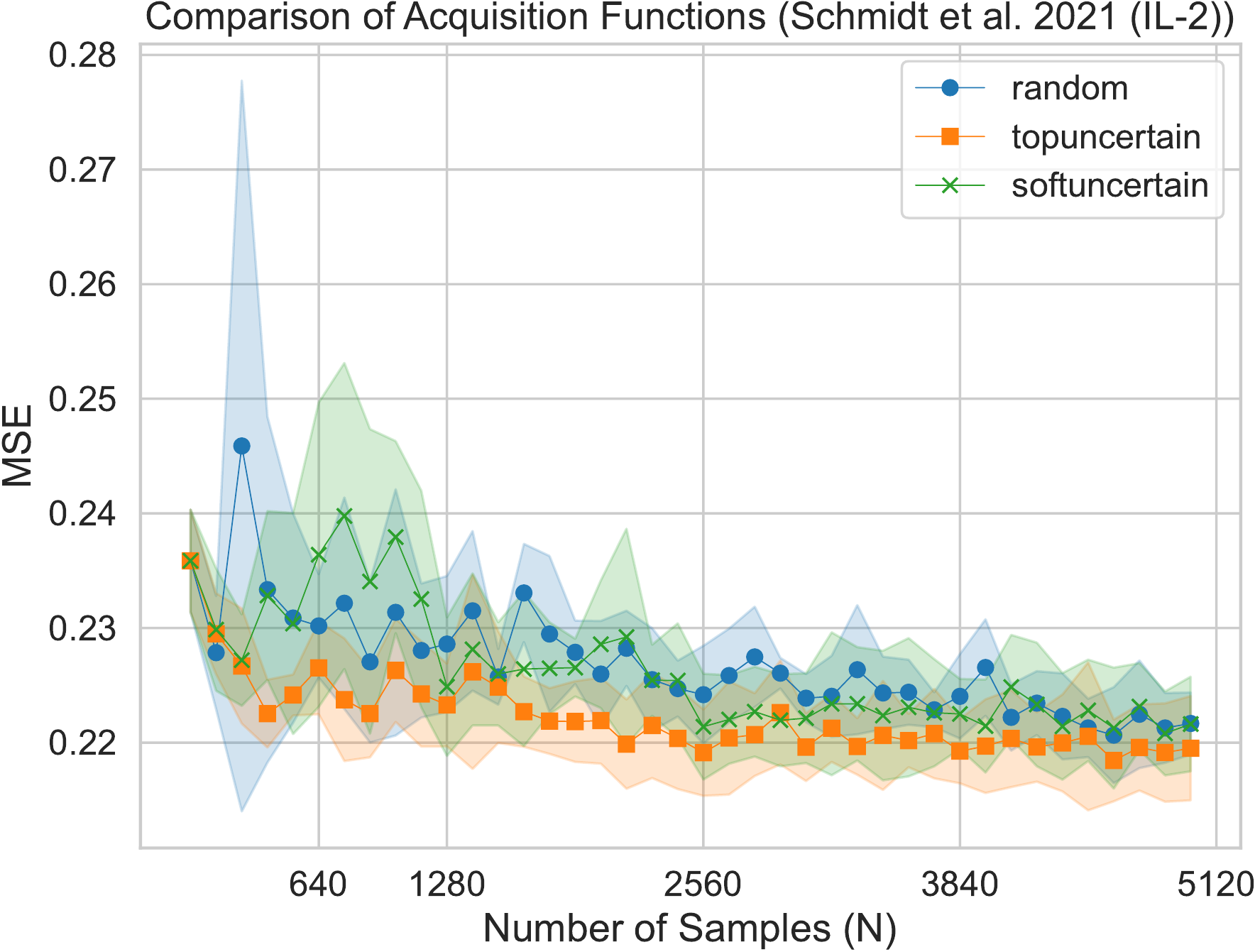}};
                        \end{tikzpicture}
                    }
                \end{subfigure}
                \&
                \begin{subfigure}{0.29\columnwidth}
                    \hspace{-32mm}
                    \centering
                    \resizebox{\linewidth}{!}{
                        \begin{tikzpicture}
                            \node (img)  {\includegraphics[width=\textwidth]{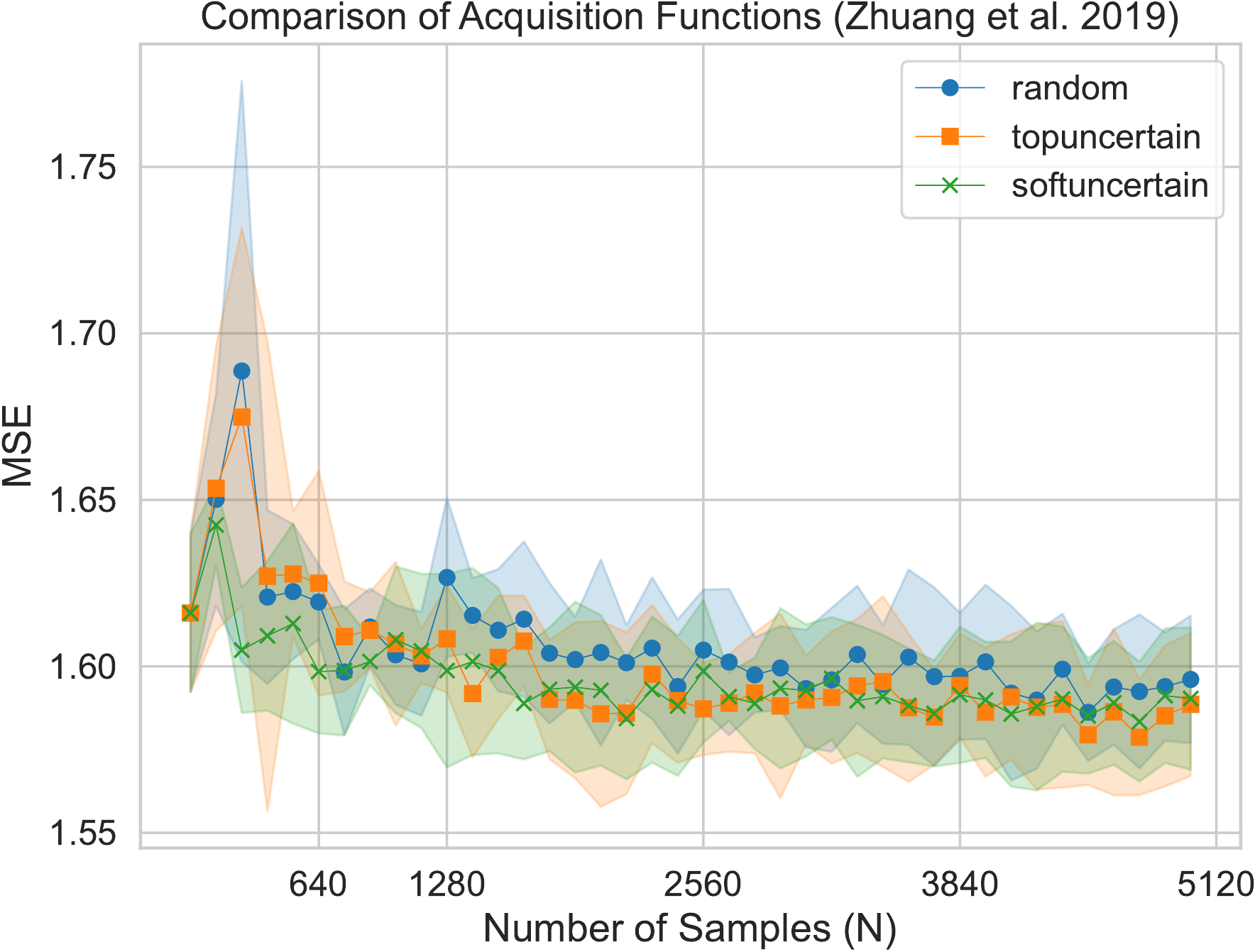}};
                        \end{tikzpicture}
                    }
                \end{subfigure}
                \&
                \\
\begin{subfigure}{0.275\columnwidth}
                    \hspace{-17mm}
                    \centering
                    \resizebox{\linewidth}{!}{
                        \begin{tikzpicture}
                            \node (img)  {\includegraphics[width=\textwidth]{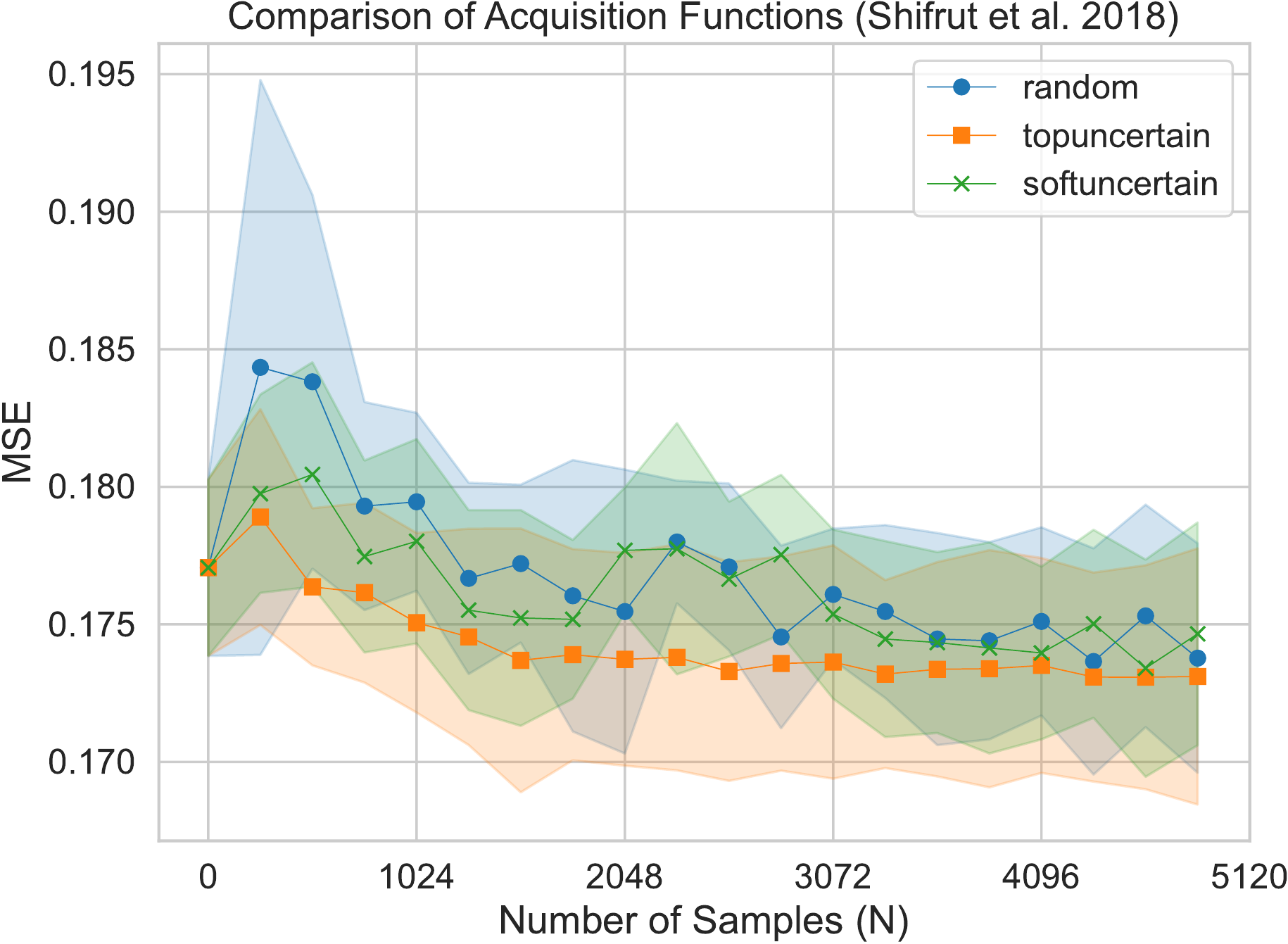}};
                        \end{tikzpicture}
                    }
                \end{subfigure}
                \&
                \begin{subfigure}{0.27\columnwidth}
                    \hspace{-23mm}
                    \centering
                    \resizebox{\linewidth}{!}{
                        \begin{tikzpicture}
                            \node (img)  {\includegraphics[width=\textwidth]{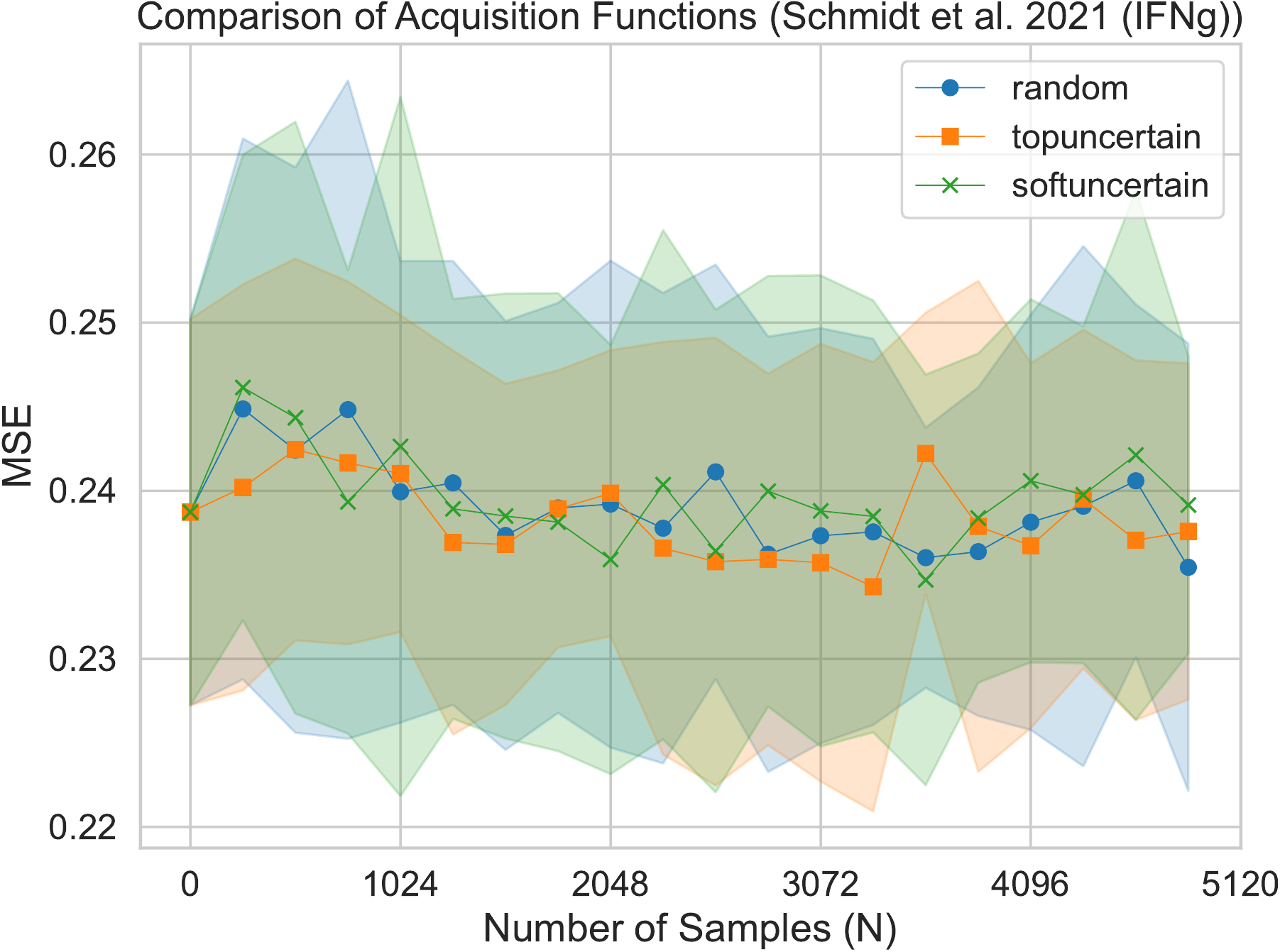}};
                        \end{tikzpicture}
                    }
                \end{subfigure}
                \&
                \begin{subfigure}{0.27\columnwidth}
                    \hspace{-28mm}
                    \centering
                    \resizebox{\linewidth}{!}{
                        \begin{tikzpicture}
                            \node (img)  {\includegraphics[width=\textwidth]{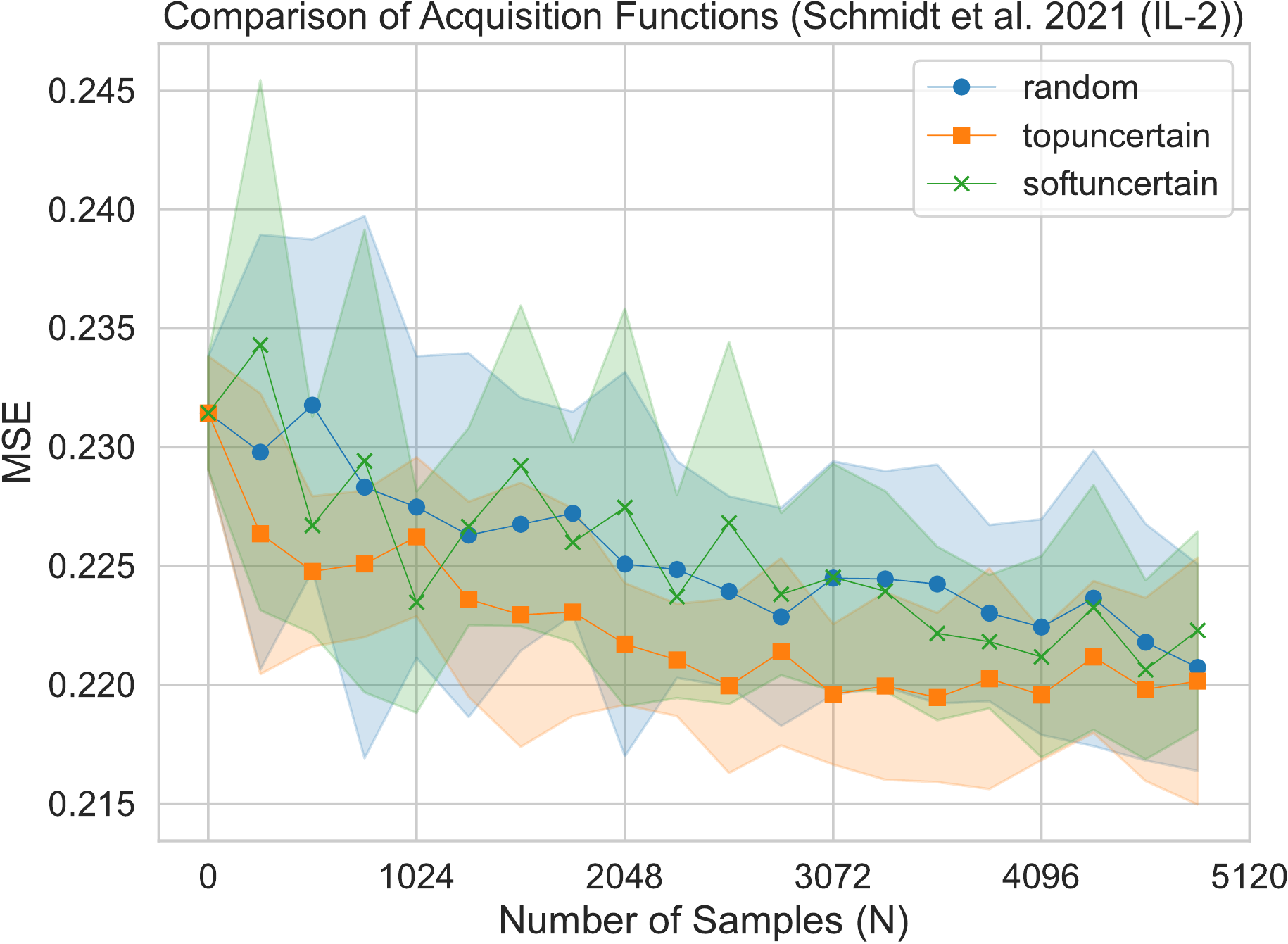}};
                        \end{tikzpicture}
                    }
                \end{subfigure}
                \&
                \begin{subfigure}{0.29\columnwidth}
                    \hspace{-32mm}
                    \centering
                    \resizebox{\linewidth}{!}{
                        \begin{tikzpicture}
                            \node (img)  {\includegraphics[width=\textwidth]{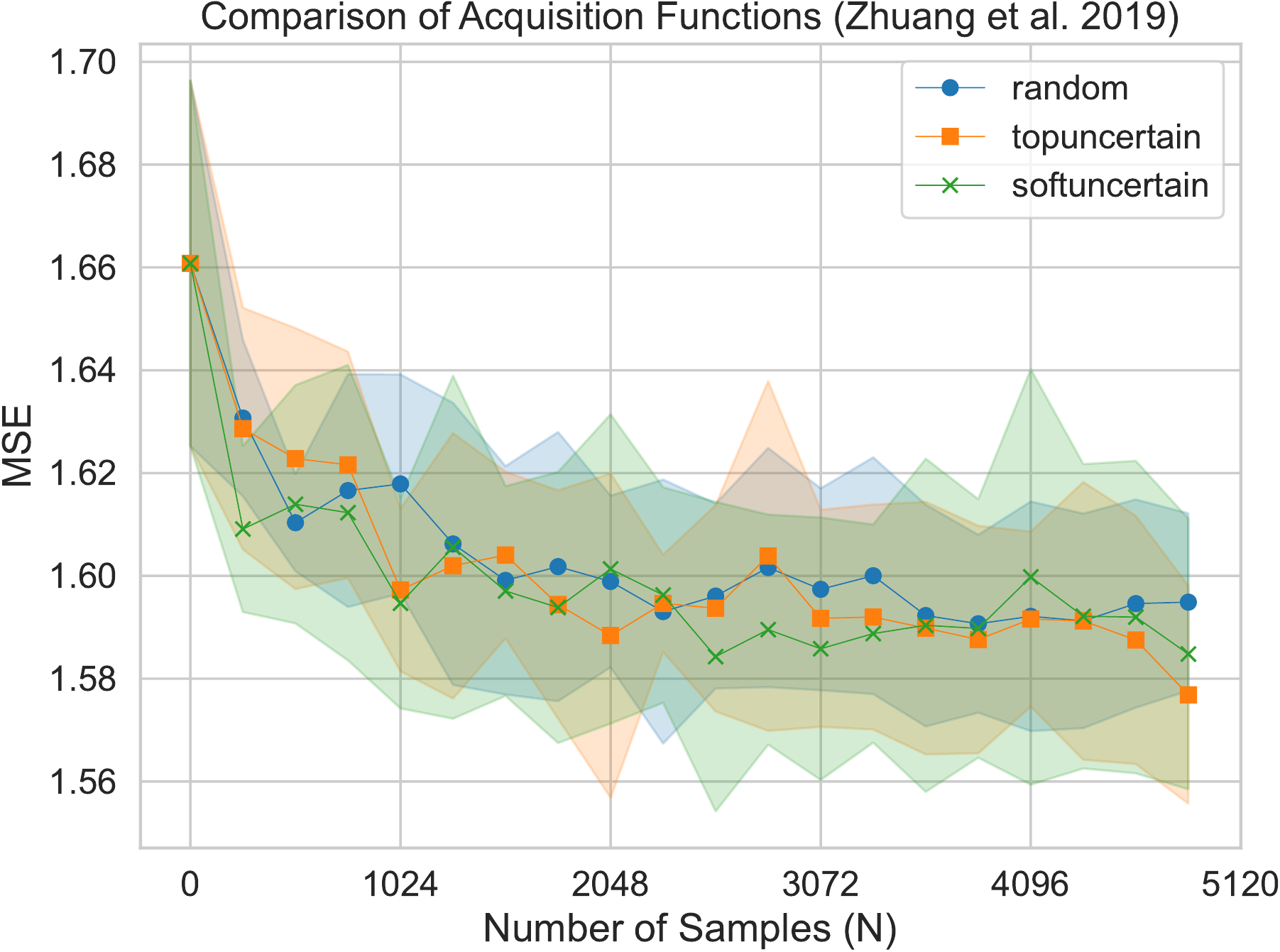}};
                        \end{tikzpicture}
                    }
                \end{subfigure}
                \&
                \\
\begin{subfigure}{0.28\columnwidth}
                    \hspace{-17mm}
                    \centering
                    \resizebox{\linewidth}{!}{
                        \begin{tikzpicture}
                            \node (img)  {\includegraphics[width=\textwidth]{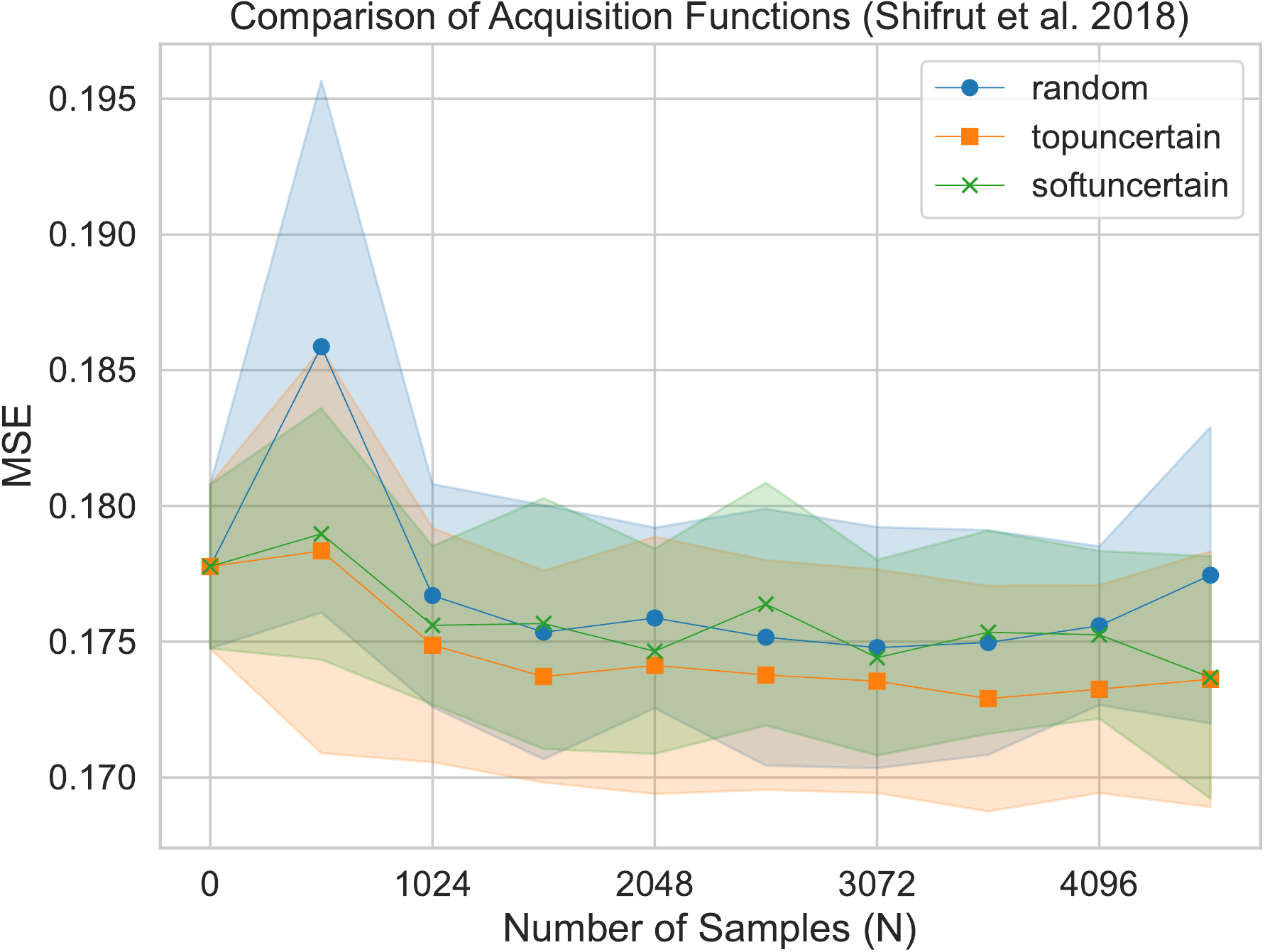}};
                        \end{tikzpicture}
                    }
                \end{subfigure}
                \&
                \begin{subfigure}{0.27\columnwidth}
                    \hspace{-23mm}
                    \centering
                    \resizebox{\linewidth}{!}{
                        \begin{tikzpicture}
                            \node (img)  {\includegraphics[width=\textwidth]{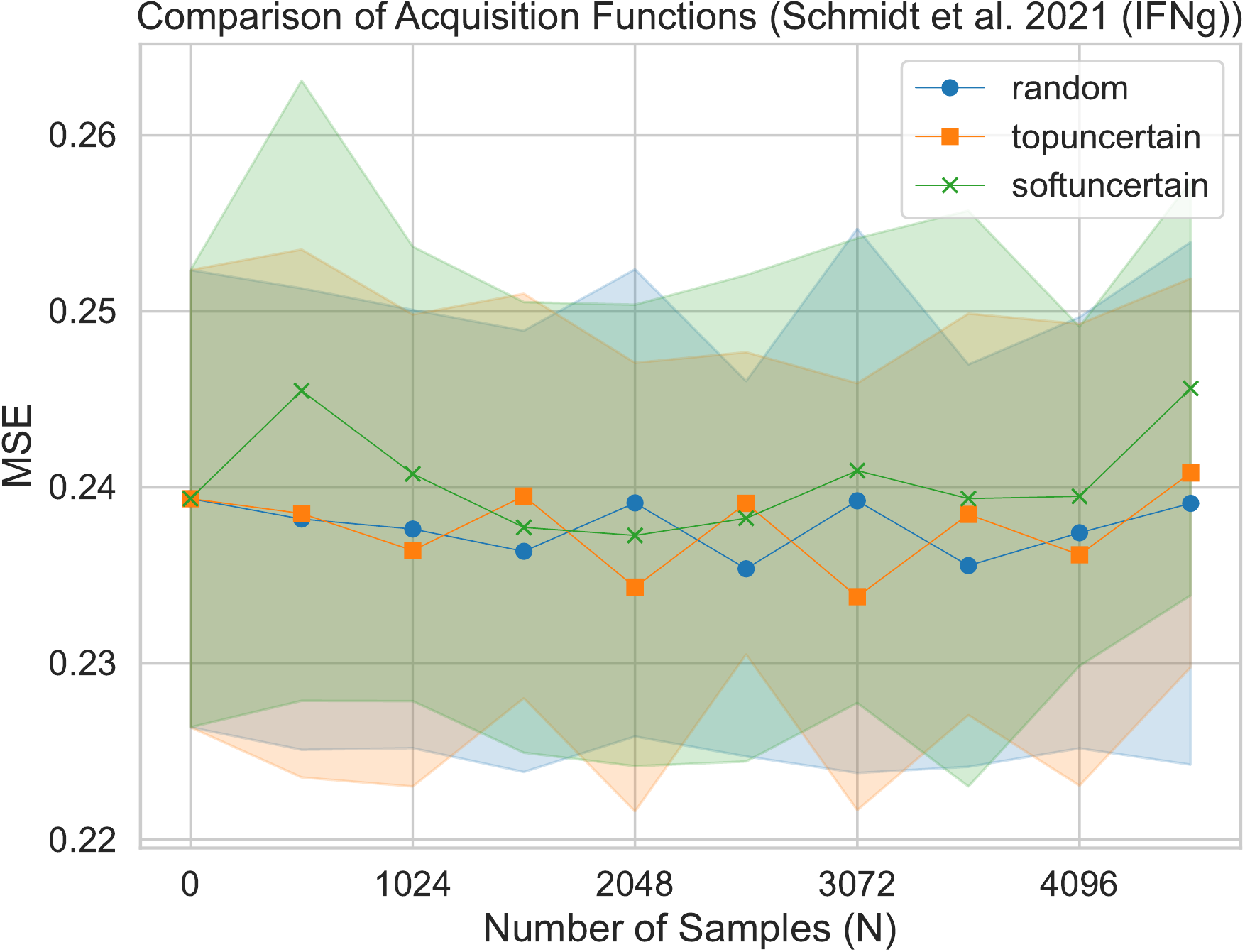}};
                        \end{tikzpicture}
                    }
                \end{subfigure}
                \&
                \begin{subfigure}{0.27\columnwidth}
                    \hspace{-28mm}
                    \centering
                    \resizebox{\linewidth}{!}{
                        \begin{tikzpicture}
                            \node (img)  {\includegraphics[width=\textwidth]{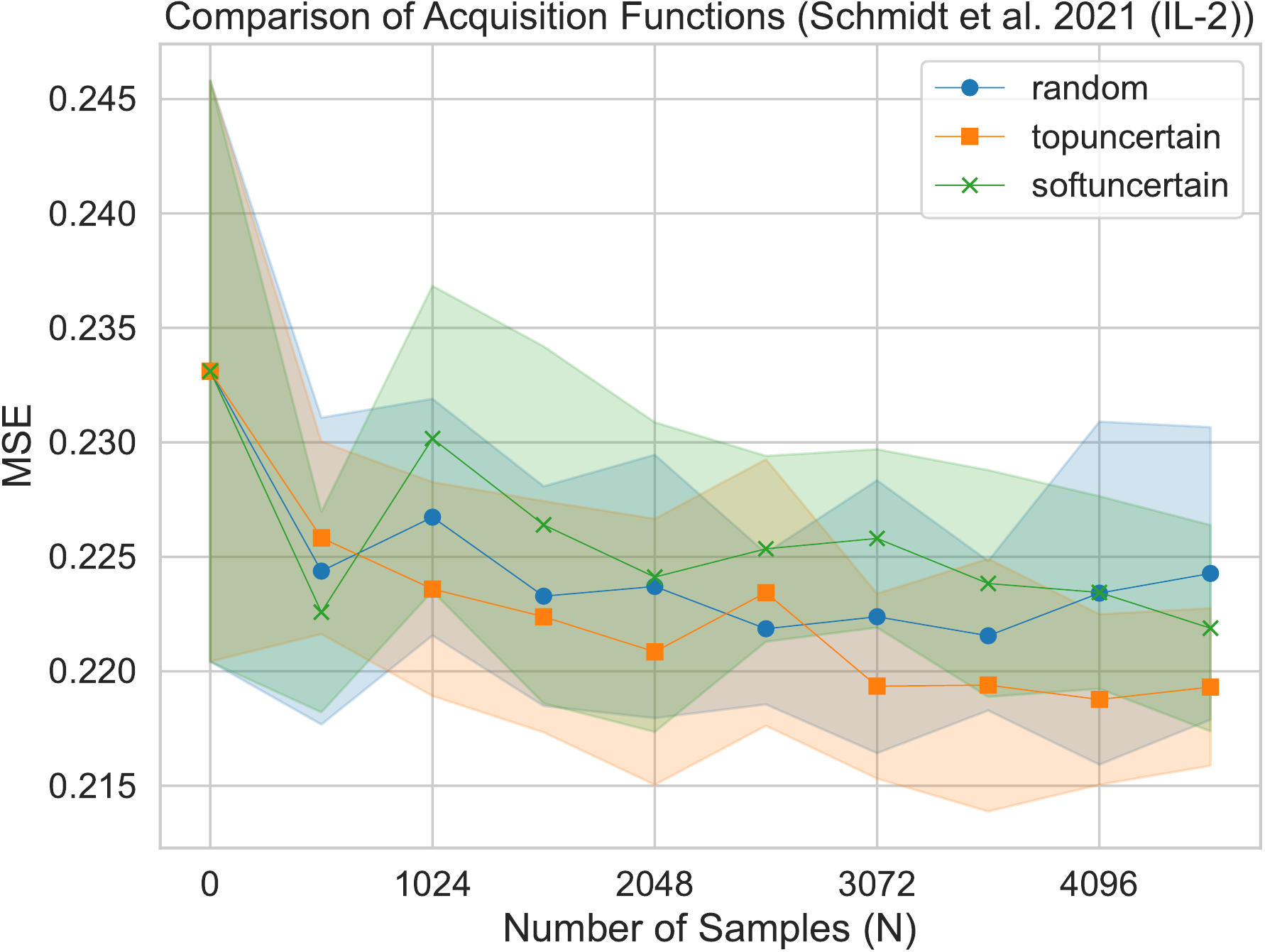}};
                        \end{tikzpicture}
                    }
                \end{subfigure}
                \&
                \begin{subfigure}{0.28\columnwidth}
                    \hspace{-32mm}
                    \centering
                    \resizebox{\linewidth}{!}{
                        \begin{tikzpicture}
                            \node (img)  {\includegraphics[width=\textwidth]{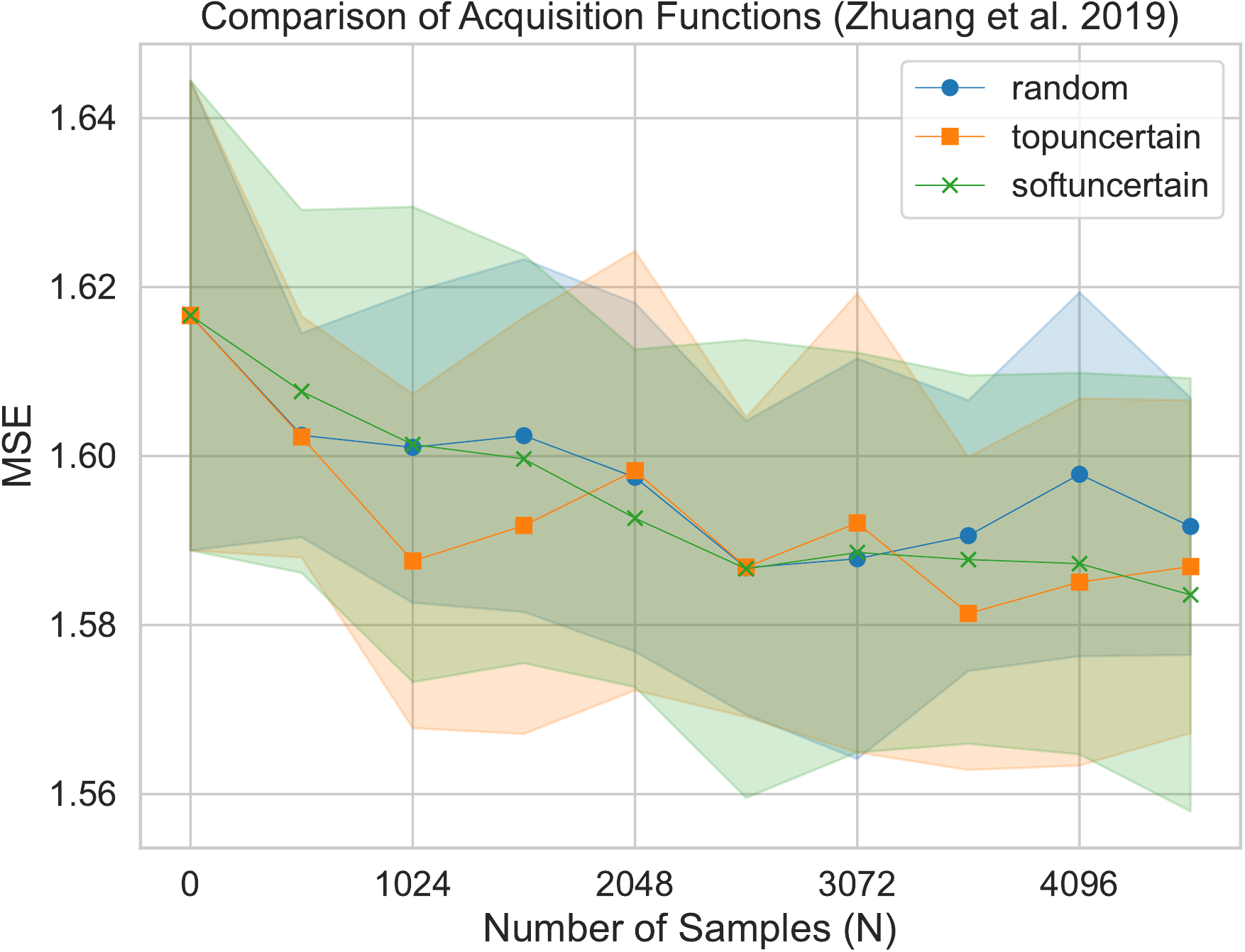}};
                        \end{tikzpicture}
                    }
                \end{subfigure}
                \&
                \\
            \\
           
            \\
            };
            \node [draw=none, rotate=90] at ([xshift=-8mm, yshift=2mm]fig-1-1.west) {\small batch size = 16};
            \node [draw=none, rotate=90] at ([xshift=-8mm, yshift=2mm]fig-2-1.west) {\small batch size = 32};
            \node [draw=none, rotate=90] at ([xshift=-8mm, yshift=2mm]fig-3-1.west) {\small batch size = 64};
            \node [draw=none, rotate=90] at ([xshift=-8mm, yshift=2mm]fig-4-1.west) {\small batch size = 128};
            \node [draw=none, rotate=90] at ([xshift=-8mm, yshift=2mm]fig-5-1.west) {\small batch size = 256};
            \node [draw=none, rotate=90] at ([xshift=-8mm, yshift=2mm]fig-6-1.west) {\small batch size = 512};
            \node [draw=none] at ([xshift=-6mm, yshift=3mm]fig-1-1.north) {\small Shifrut et al. 2018};
            \node [draw=none] at ([xshift=-9mm, yshift=3mm]fig-1-2.north) {\small Schmidt et al. 2021 (IFNg)};
            \node [draw=none] at ([xshift=-11mm, yshift=3mm]fig-1-3.north) {\small Schmidt et al. 2021 (IL-2)};
            \node [draw=none] at ([xshift=-13mm, yshift=2.5mm]fig-1-4.north) {\small Zhuang et al. 2019};
\end{tikzpicture}}
        \vspace{-7mm}
        \caption{The evaluation of the random forest model trained with {CCLE} treatment descriptors at each active learning cycle for 4 datasets and 6 acquisition batch sizes. In each plot, the x-axis is the active learning cycles multiplied by the acquisition bath size that gives the total number of data points collected so far. The y-axis is the test MSE error evaluated on the test data.}
        \vspace{-5mm}
        \label{fig:rf_feat_ccle_alldatasets_allbathcsizes}
    \end{figure*}

In addition to the BNN model, we carried out thorough analyses for a different model class. The experiments are repeated for the random forest as an uncertainty aware ensemble model. The uncertainty in random forests, similar to other ensemble methods, is originated from the prediction made by each model instance in the ensemble. We use the random forest implementation in the Scikit-learn package \citep{scikit-learn} with 100 trees and set the option max\_depth=None so that the depth of the trees are determined automatically. The performance of the model trained over the active learning cycles can be seen in~\cref{fig:rf_feat_string_alldatasets_allbathcsizes} for different acquisition functions, different batch sizes, different target datasets, and the STRING treatment descriptors. Similarly, the hit ratio of the interesting genes for a random forest model is reported in~\cref{fig:hitratio_rf_feat_string_alldatasets_allbathcsizes}. The same experiment was repeated for CCLE treatment descriptors whose results are provided in~\cref{fig:rf_feat_ccle_alldatasets_allbathcsizes} and~\cref{fig:hitratio_rf_feat_ccle_alldatasets_allbathcsizes}. Notice that random forest experiments are done with a reduced set of acquisition functions that could be adjusted to the random forest model.

\subsection{In-depth Description of the \emph{Hit Ratio} Experiment}
\label{sec:appendix:hit_ratio_description}
\begin{figure*}
    \vspace{-2mm}
        \centering
        \makebox[0.72\paperwidth]{\begin{tikzpicture}[ampersand replacement=\&]
            \matrix (fig) [matrix of nodes]{ 
\begin{subfigure}{0.27\columnwidth}
                    \hspace{-17mm}
                    \centering
                    \resizebox{\linewidth}{!}{
                        \begin{tikzpicture}
                            \node (img)  {\includegraphics[width=\textwidth]{figs/hitratio_bnnplots/hitratio_data_shifrut_2018_feat_string_bs16.pdf}};
                        \end{tikzpicture}
                    }
                \end{subfigure}
                \&
                 \begin{subfigure}{0.27\columnwidth}
                    \hspace{-23mm}
                    \centering
                    \resizebox{\linewidth}{!}{
                        \begin{tikzpicture}
                            \node (img)  {\includegraphics[width=\textwidth]{figs/hitratio_bnnplots/hitratio_data_schmidt_2021_ifng_feat_string_bs16.pdf}};
                        \end{tikzpicture}
                    }
                \end{subfigure}
                \&
                 \begin{subfigure}{0.27\columnwidth}
                    \hspace{-28mm}
                    \centering
                    \resizebox{\linewidth}{!}{
                        \begin{tikzpicture}
                            \node (img)  {\includegraphics[width=\textwidth]{figs/hitratio_bnnplots/hitratio_data_schmidt_2021_il2_feat_string_bs16.pdf}};
                        \end{tikzpicture}
                    }
                \end{subfigure}
                \&
                \begin{subfigure}{0.28\columnwidth}
                    \hspace{-32mm}
                    \centering
                    \resizebox{\linewidth}{!}{
                        \begin{tikzpicture}
                            \node (img)  {\includegraphics[width=\textwidth]{figs/hitratio_bnnplots/hitratio_data_zhuang_2019_nk_feat_string_bs16.pdf}};
                        \end{tikzpicture}
                    }
                \end{subfigure}
                \&
            \\
\begin{subfigure}{0.27\columnwidth}
                    \hspace{-17mm}
                    \centering
                    \resizebox{\linewidth}{!}{
                        \begin{tikzpicture}
                            \node (img)  {\includegraphics[width=\textwidth]{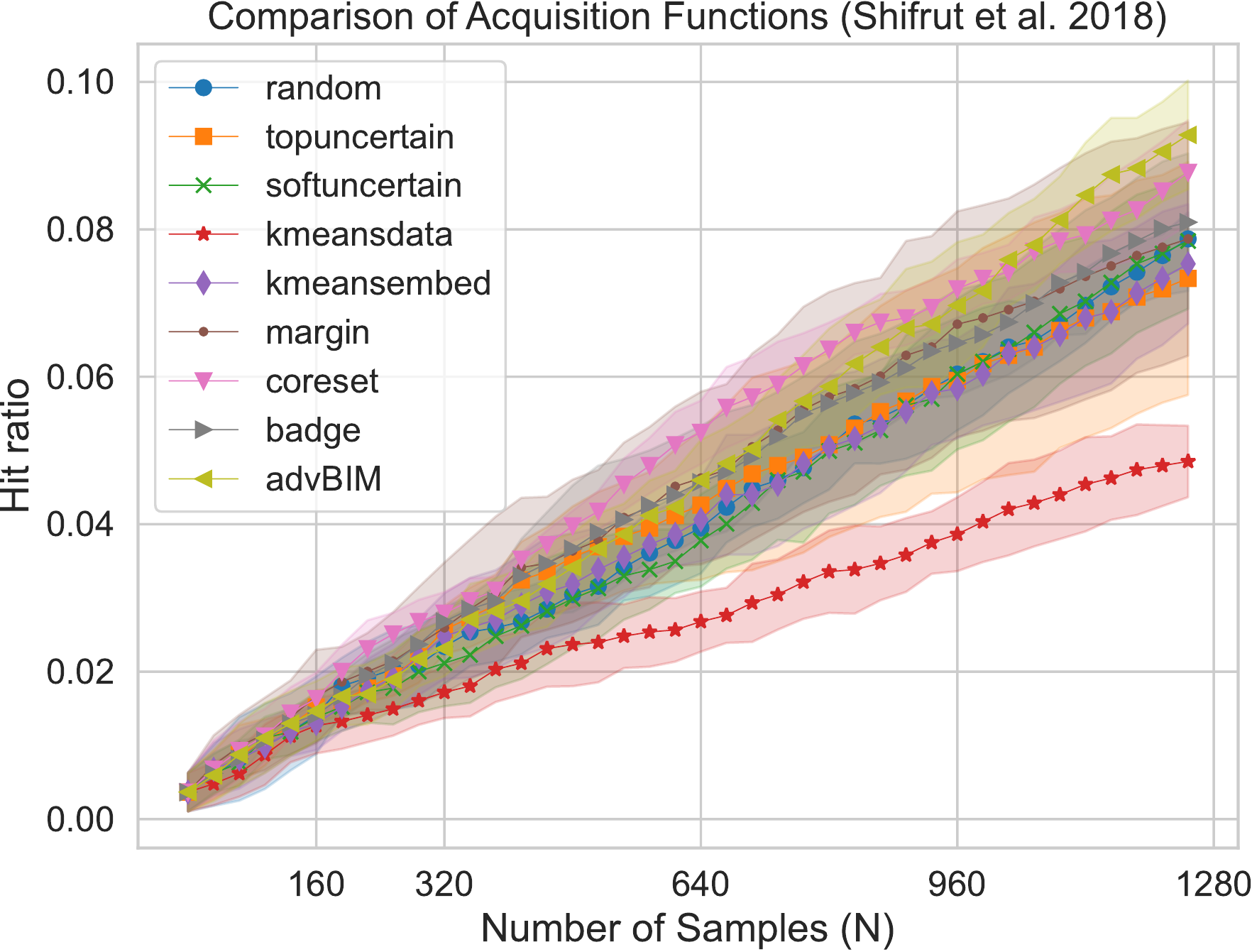}};
                        \end{tikzpicture}
                    }
                \end{subfigure}
                \&
                \begin{subfigure}{0.27\columnwidth}
                    \hspace{-23mm}
                    \centering
                    \resizebox{\linewidth}{!}{
                        \begin{tikzpicture}
                            \node (img)  {\includegraphics[width=\textwidth]{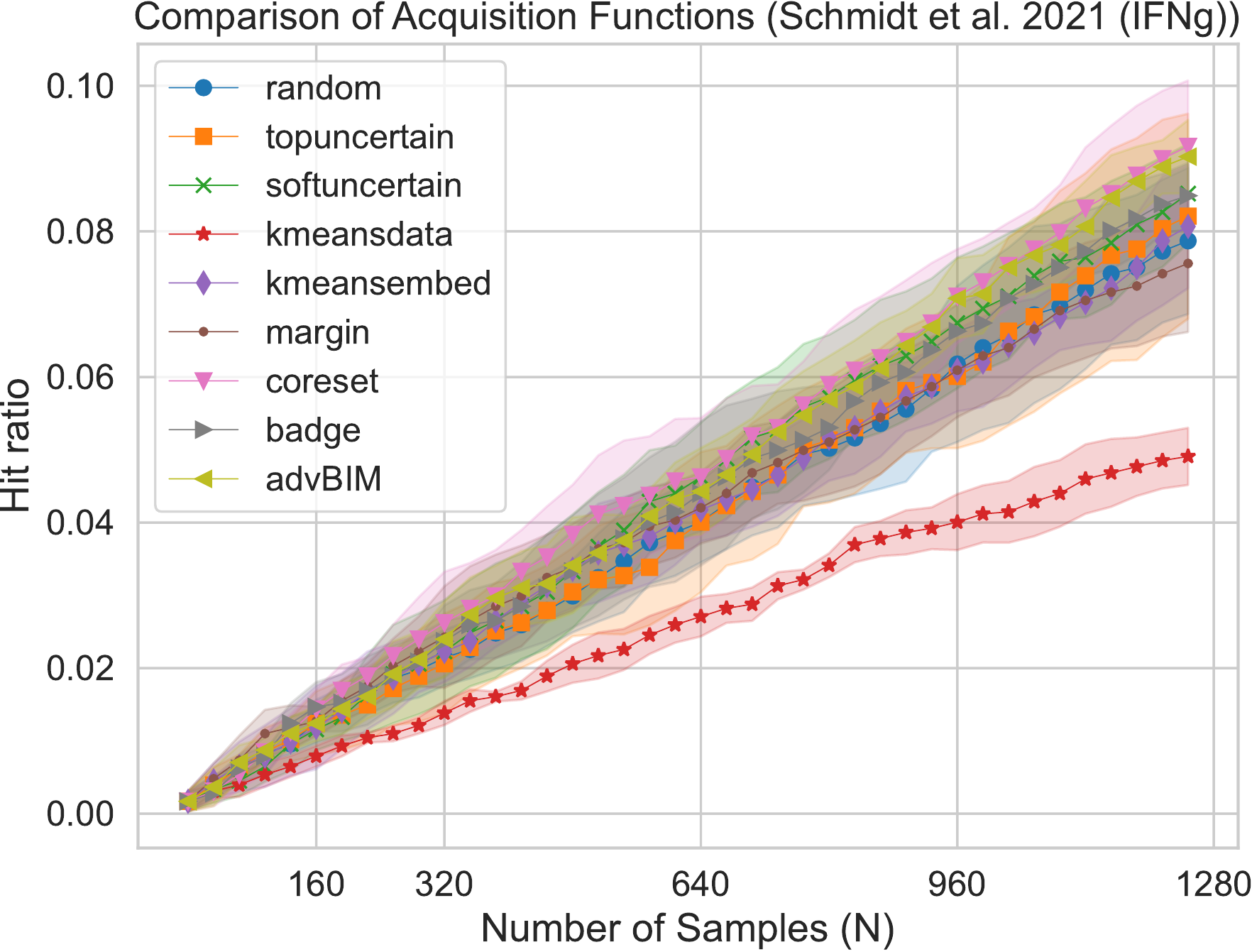}};
                        \end{tikzpicture}
                    }
                \end{subfigure}
                \&
                \begin{subfigure}{0.27\columnwidth}
                    \hspace{-28mm}
                    \centering
                    \resizebox{\linewidth}{!}{
                        \begin{tikzpicture}
                            \node (img)  {\includegraphics[width=\textwidth]{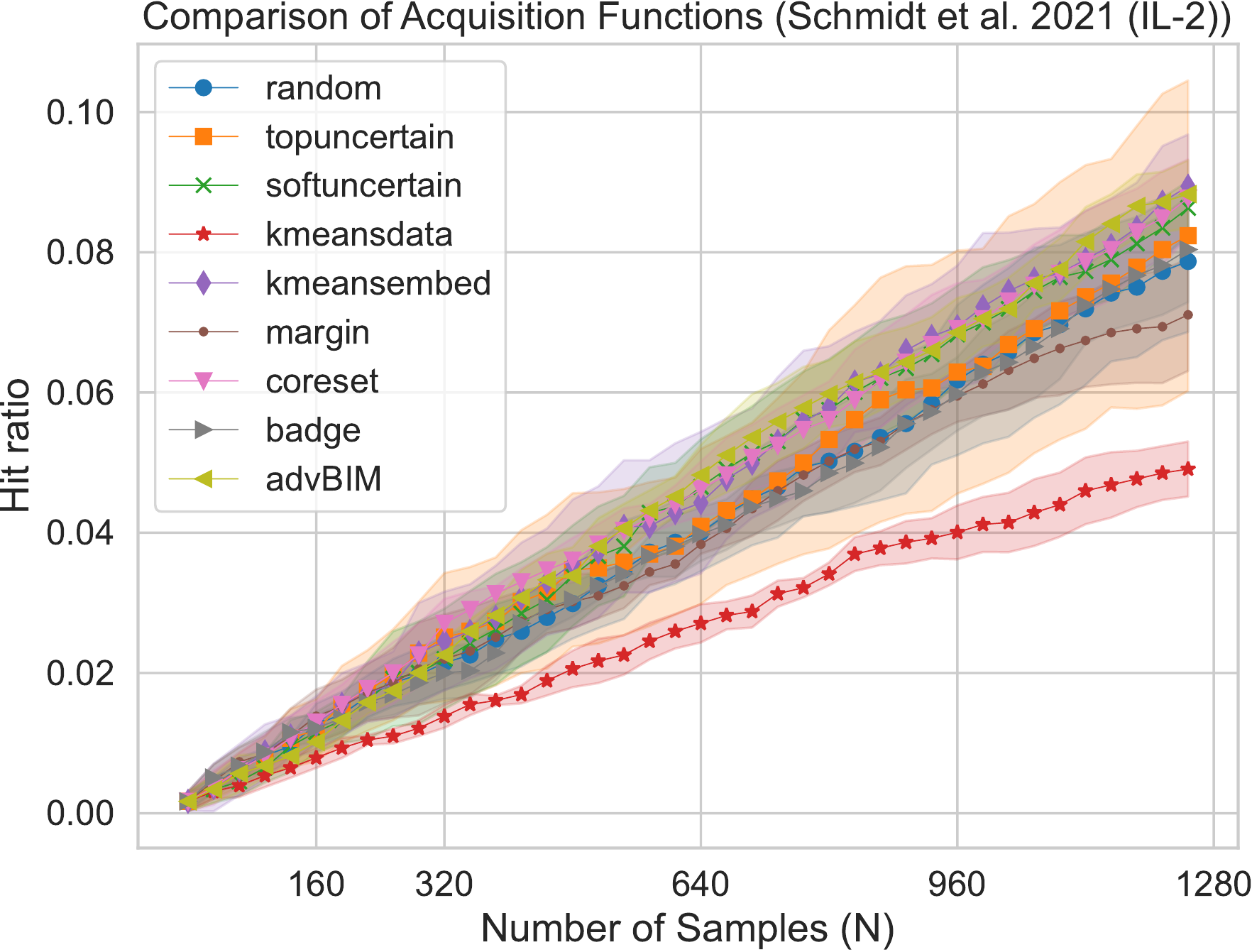}};
                        \end{tikzpicture}
                    }
                \end{subfigure}
                \&
                \begin{subfigure}{0.28\columnwidth}
                    \hspace{-32mm}
                    \centering
                    \resizebox{\linewidth}{!}{
                        \begin{tikzpicture}
                            \node (img)  {\includegraphics[width=\textwidth]{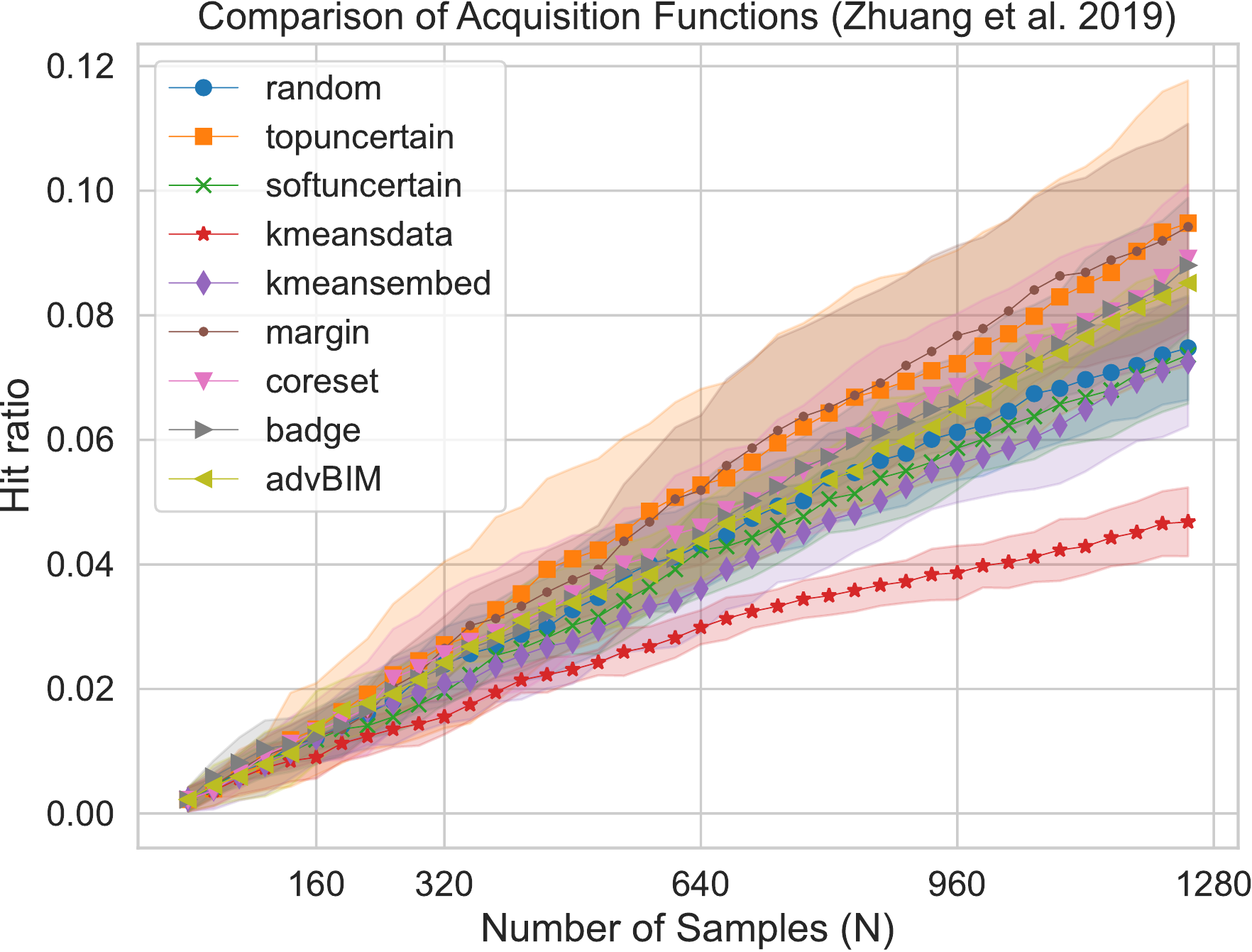}};
                        \end{tikzpicture}
                    }
                \end{subfigure}
                \&
                \\
\begin{subfigure}{0.27\columnwidth}
                    \hspace{-17mm}
                    \centering
                    \resizebox{\linewidth}{!}{
                        \begin{tikzpicture}
                            \node (img)  {\includegraphics[width=\textwidth]{figs/hitratio_bnnplots/hitratio_data_shifrut_2018_feat_string_bs64.pdf}};
                        \end{tikzpicture}
                    }
                \end{subfigure}
                \&
                \begin{subfigure}{0.27\columnwidth}
                    \hspace{-23mm}
                    \centering
                    \resizebox{\linewidth}{!}{
                        \begin{tikzpicture}
                            \node (img)  {\includegraphics[width=\textwidth]{figs/hitratio_bnnplots/hitratio_data_schmidt_2021_ifng_feat_string_bs64.pdf}};
                        \end{tikzpicture}
                    }
                \end{subfigure}
                \&
                \begin{subfigure}{0.27\columnwidth}
                    \hspace{-28mm}
                    \centering
                    \resizebox{\linewidth}{!}{
                        \begin{tikzpicture}
                            \node (img)  {\includegraphics[width=\textwidth]{figs/hitratio_bnnplots/hitratio_data_schmidt_2021_il2_feat_string_bs64.pdf}};
                        \end{tikzpicture}
                    }
                \end{subfigure}
                \&
                \begin{subfigure}{0.28\columnwidth}
                    \hspace{-32mm}
                    \centering
                    \resizebox{\linewidth}{!}{
                        \begin{tikzpicture}
                            \node (img)  {\includegraphics[width=\textwidth]{figs/hitratio_bnnplots/hitratio_data_zhuang_2019_nk_feat_string_bs64.pdf}};
                        \end{tikzpicture}
                    }
                \end{subfigure}
                \&
                \\
\begin{subfigure}{0.27\columnwidth}
                    \hspace{-17mm}
                    \centering
                    \resizebox{\linewidth}{!}{
                        \begin{tikzpicture}
                            \node (img)  {\includegraphics[width=\textwidth]{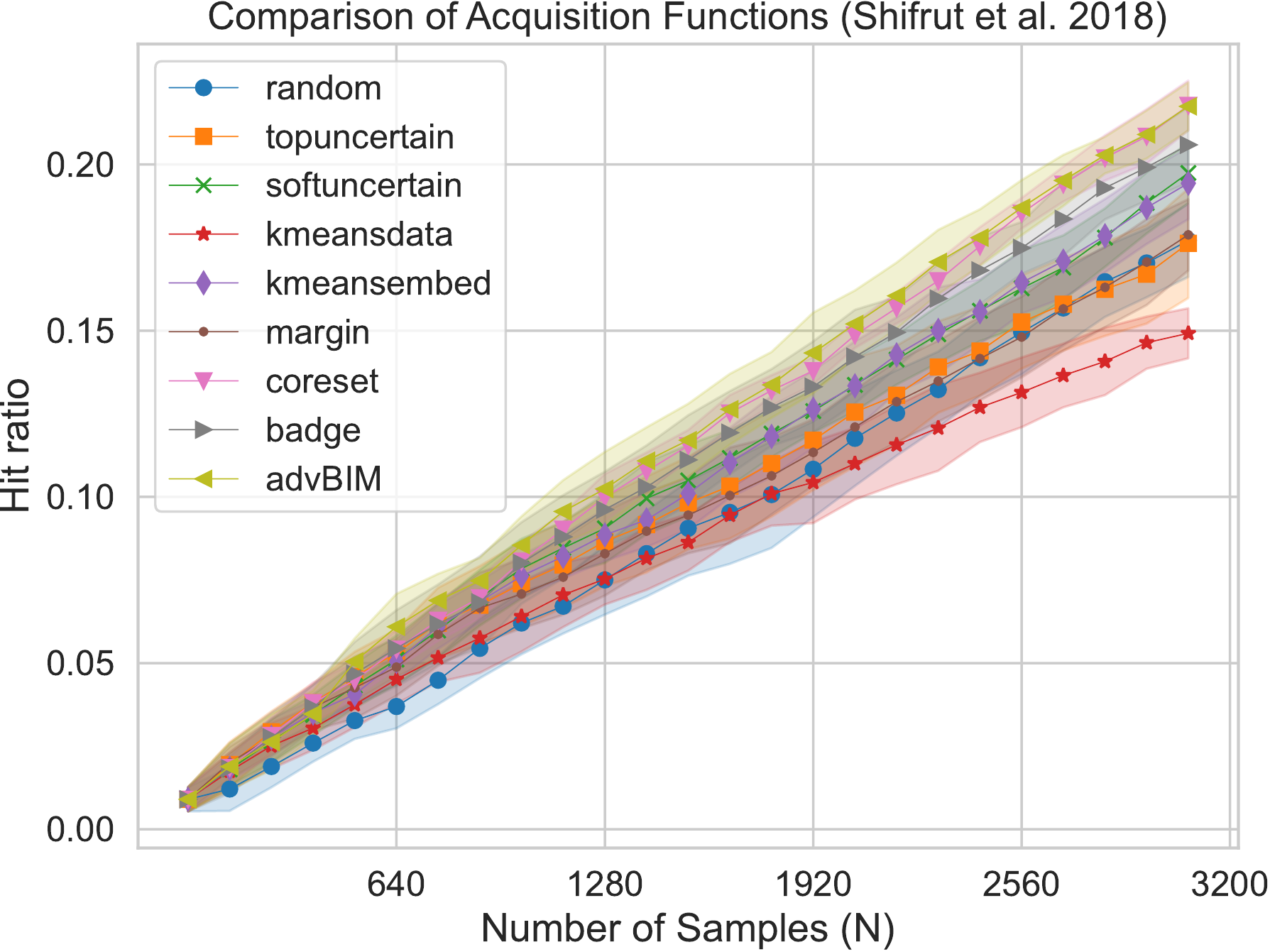}};
                        \end{tikzpicture}
                    }
                \end{subfigure}
                \&
                \begin{subfigure}{0.27\columnwidth}
                    \hspace{-23mm}
                    \centering
                    \resizebox{\linewidth}{!}{
                        \begin{tikzpicture}
                            \node (img)  {\includegraphics[width=\textwidth]{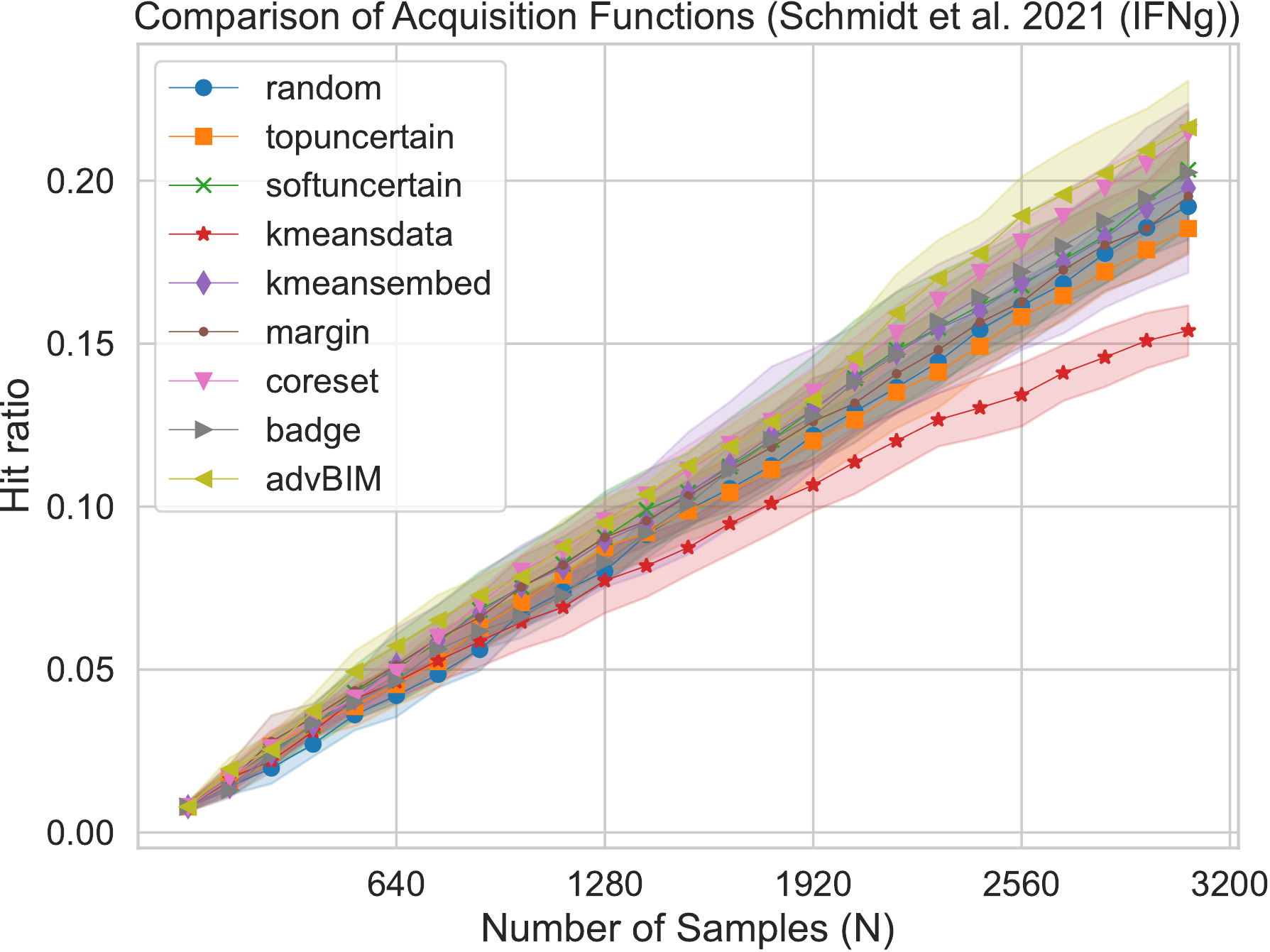}};
                        \end{tikzpicture}
                    }
                \end{subfigure}
                \&
                \begin{subfigure}{0.28\columnwidth}
                    \hspace{-28mm}
                    \centering
                    \resizebox{\linewidth}{!}{
                        \begin{tikzpicture}
                            \node (img)  {\includegraphics[width=\textwidth]{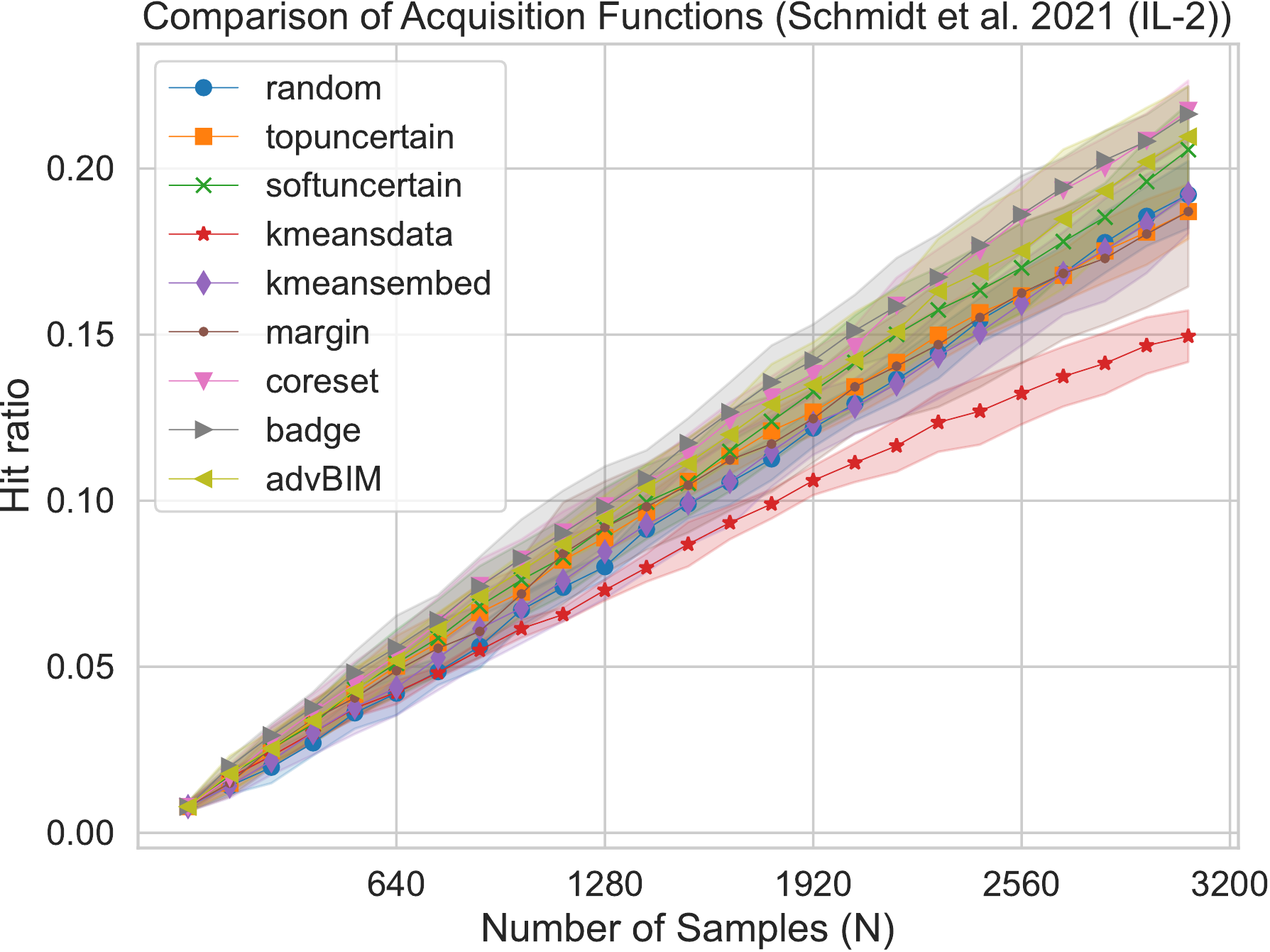}};
                        \end{tikzpicture}
                    }
                \end{subfigure}
                \&
                \begin{subfigure}{0.29\columnwidth}
                    \hspace{-32mm}
                    \centering
                    \resizebox{\linewidth}{!}{
                        \begin{tikzpicture}
                            \node (img)  {\includegraphics[width=\textwidth]{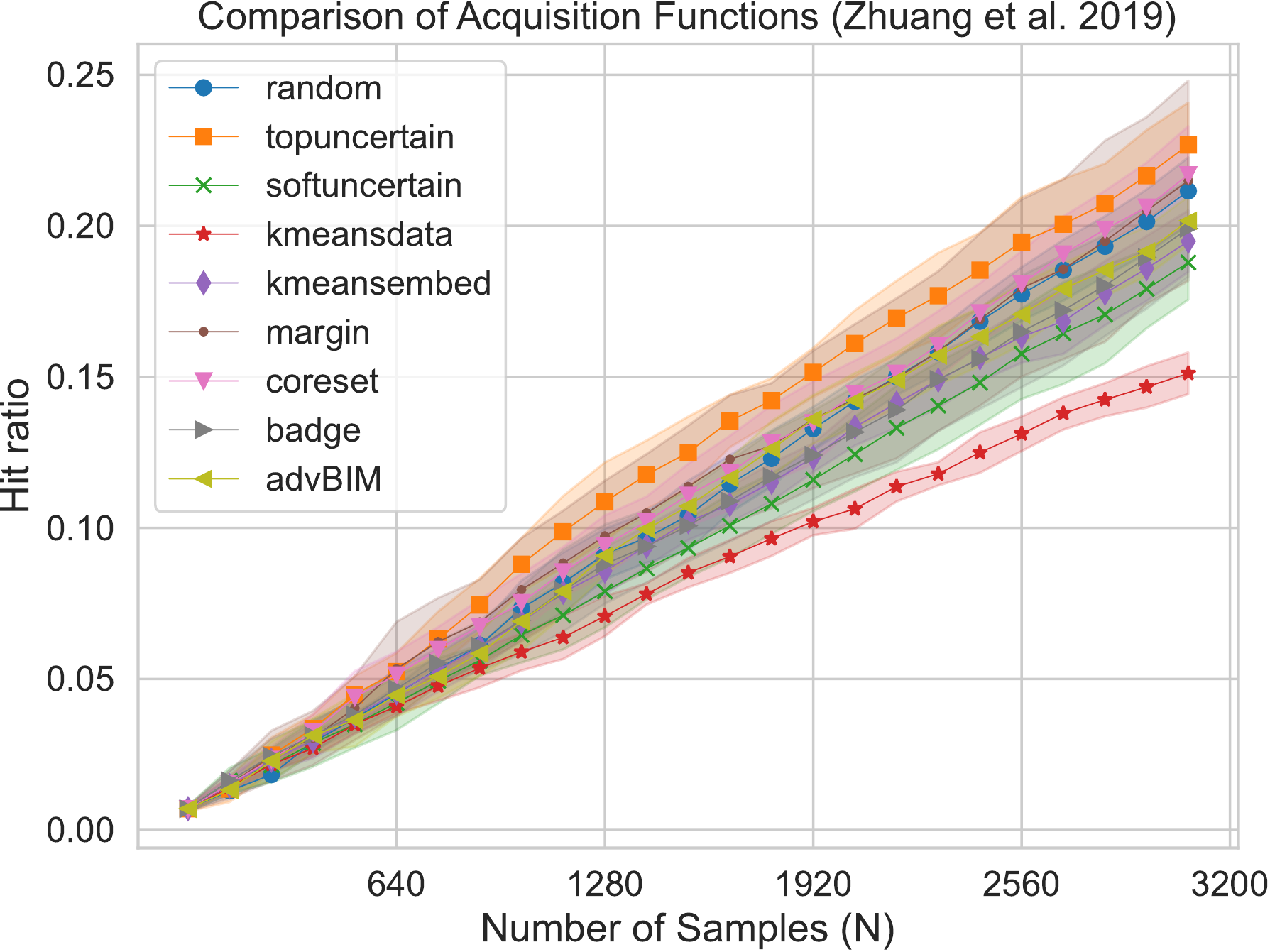}};
                        \end{tikzpicture}
                    }
                \end{subfigure}
                \&
                \\
\begin{subfigure}{0.275\columnwidth}
                    \hspace{-17mm}
                    \centering
                    \resizebox{\linewidth}{!}{
                        \begin{tikzpicture}
                            \node (img)  {\includegraphics[width=\textwidth]{figs/hitratio_bnnplots/hitratio_data_shifrut_2018_feat_string_bs256.pdf}};
                        \end{tikzpicture}
                    }
                \end{subfigure}
                \&
                \begin{subfigure}{0.27\columnwidth}
                    \hspace{-23mm}
                    \centering
                    \resizebox{\linewidth}{!}{
                        \begin{tikzpicture}
                            \node (img)  {\includegraphics[width=\textwidth]{figs/hitratio_bnnplots/hitratio_data_schmidt_2021_ifng_feat_string_bs256.pdf}};
                        \end{tikzpicture}
                    }
                \end{subfigure}
                \&
                \begin{subfigure}{0.27\columnwidth}
                    \hspace{-28mm}
                    \centering
                    \resizebox{\linewidth}{!}{
                        \begin{tikzpicture}
                            \node (img)  {\includegraphics[width=\textwidth]{figs/hitratio_bnnplots/hitratio_data_schmidt_2021_il2_feat_string_bs256.pdf}};
                        \end{tikzpicture}
                    }
                \end{subfigure}
                \&
                \begin{subfigure}{0.29\columnwidth}
                    \hspace{-32mm}
                    \centering
                    \resizebox{\linewidth}{!}{
                        \begin{tikzpicture}
                            \node (img)  {\includegraphics[width=\textwidth]{figs/hitratio_bnnplots/hitratio_data_zhuang_2019_nk_feat_string_bs256.pdf}};
                        \end{tikzpicture}
                    }
                \end{subfigure}
                \&
                \\
\begin{subfigure}{0.28\columnwidth}
                    \hspace{-17mm}
                    \centering
                    \resizebox{\linewidth}{!}{
                        \begin{tikzpicture}
                            \node (img)  {\includegraphics[width=\textwidth]{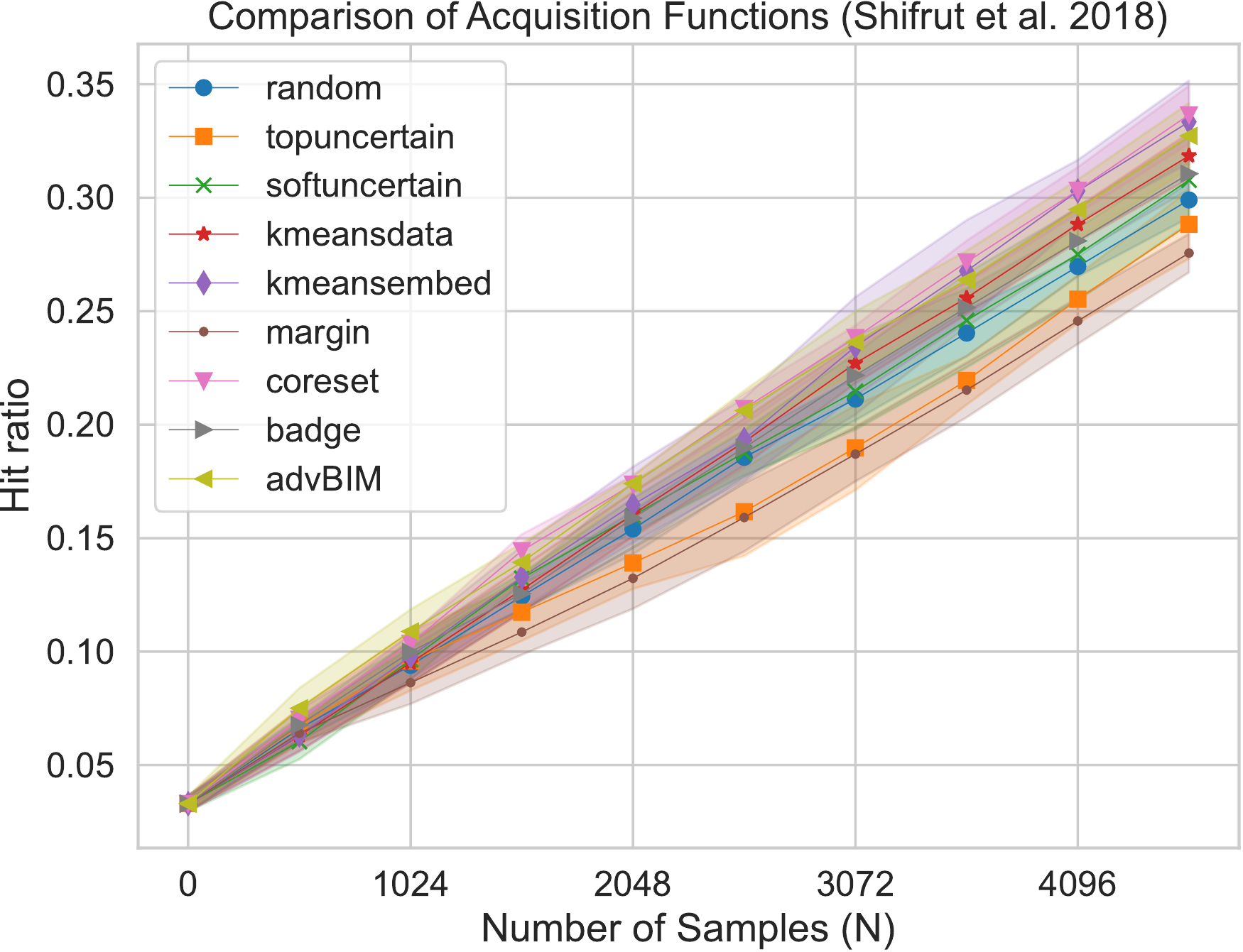}};
                        \end{tikzpicture}
                    }
                \end{subfigure}
                \&
                \begin{subfigure}{0.27\columnwidth}
                    \hspace{-23mm}
                    \centering
                    \resizebox{\linewidth}{!}{
                        \begin{tikzpicture}
                            \node (img)  {\includegraphics[width=\textwidth]{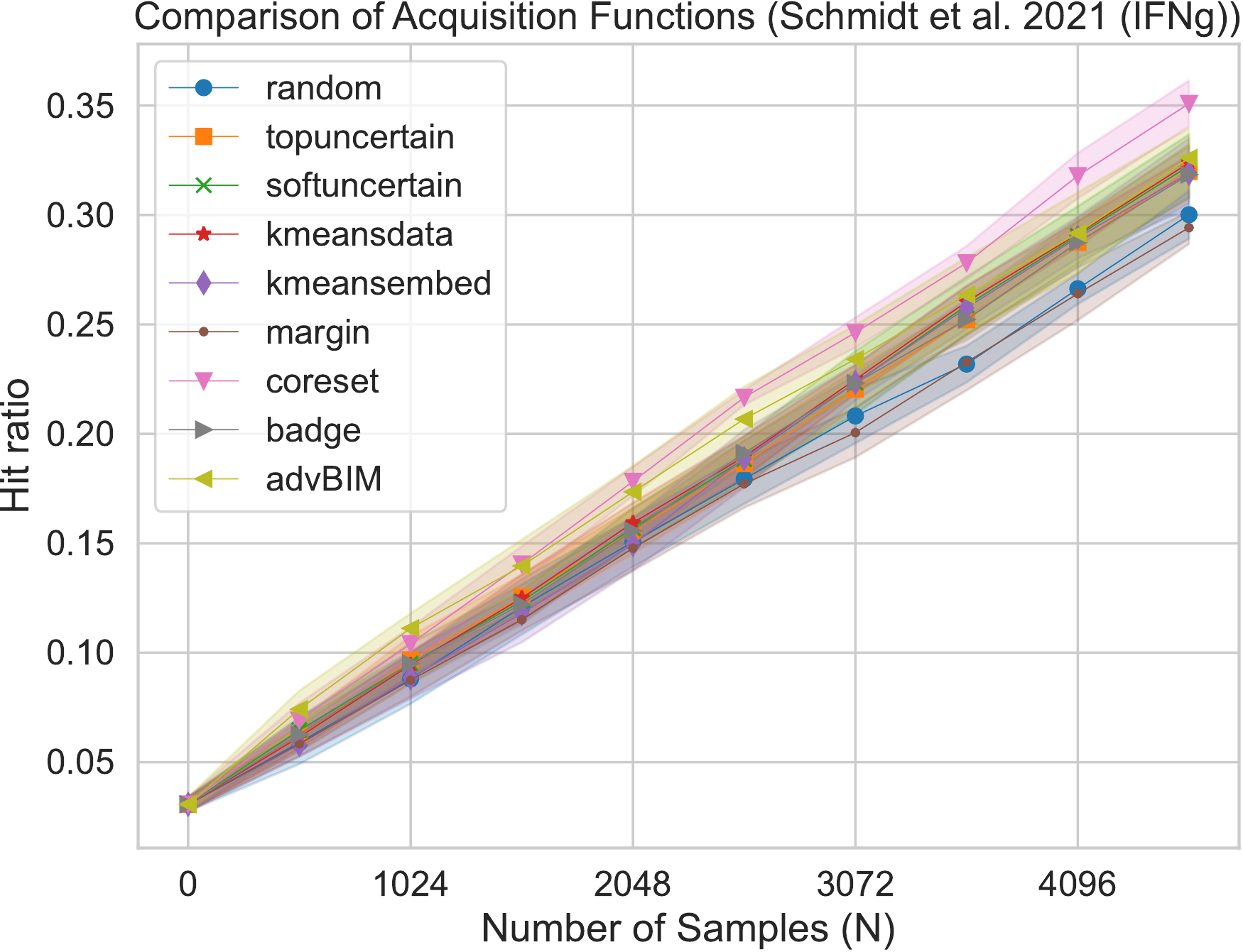}};
                        \end{tikzpicture}
                    }
                \end{subfigure}
                \&
                \begin{subfigure}{0.27\columnwidth}
                    \hspace{-28mm}
                    \centering
                    \resizebox{\linewidth}{!}{
                        \begin{tikzpicture}
                            \node (img)  {\includegraphics[width=\textwidth]{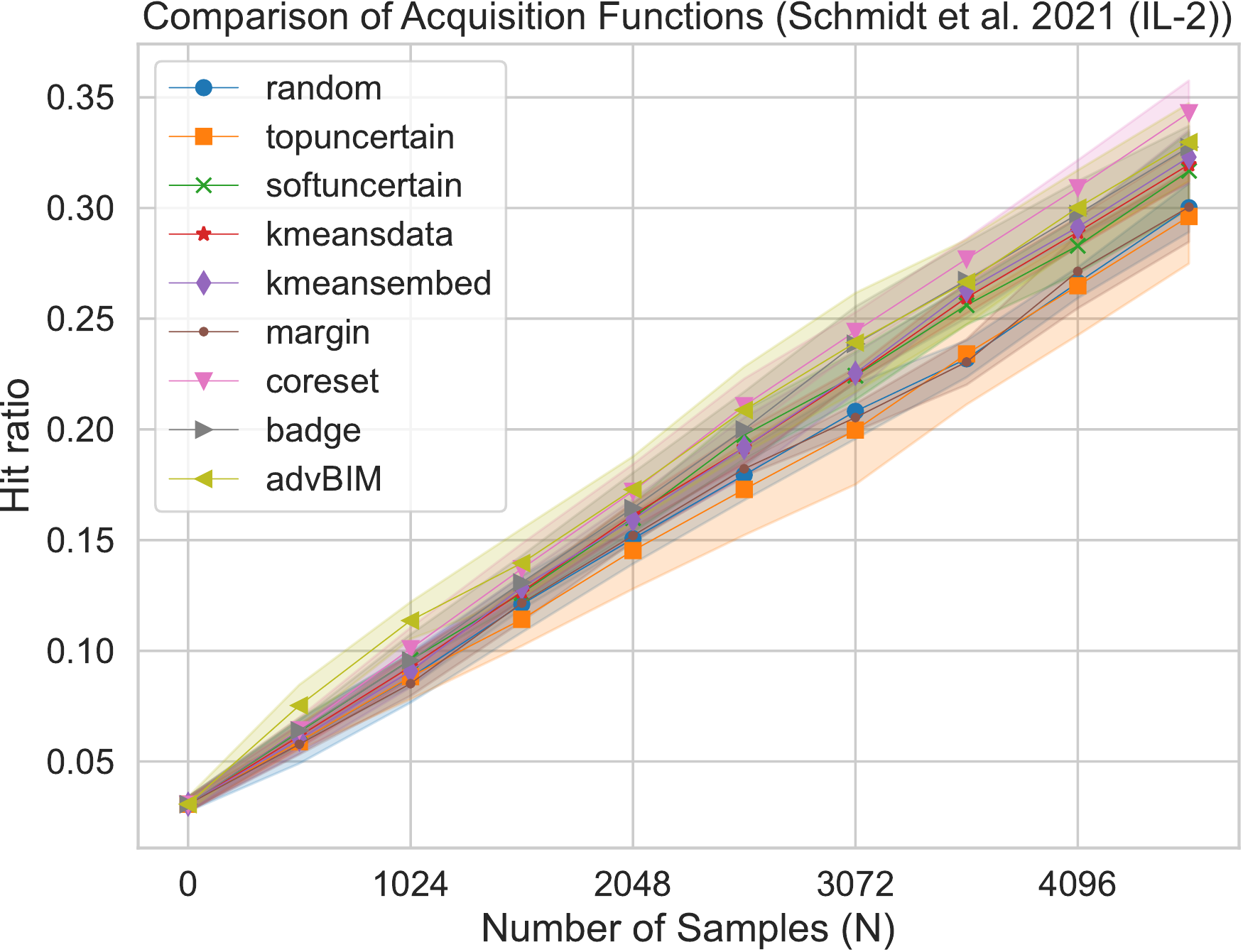}};
                        \end{tikzpicture}
                    }
                \end{subfigure}
                \&
                \begin{subfigure}{0.28\columnwidth}
                    \hspace{-32mm}
                    \centering
                    \resizebox{\linewidth}{!}{
                        \begin{tikzpicture}
                            \node (img)  {\includegraphics[width=\textwidth]{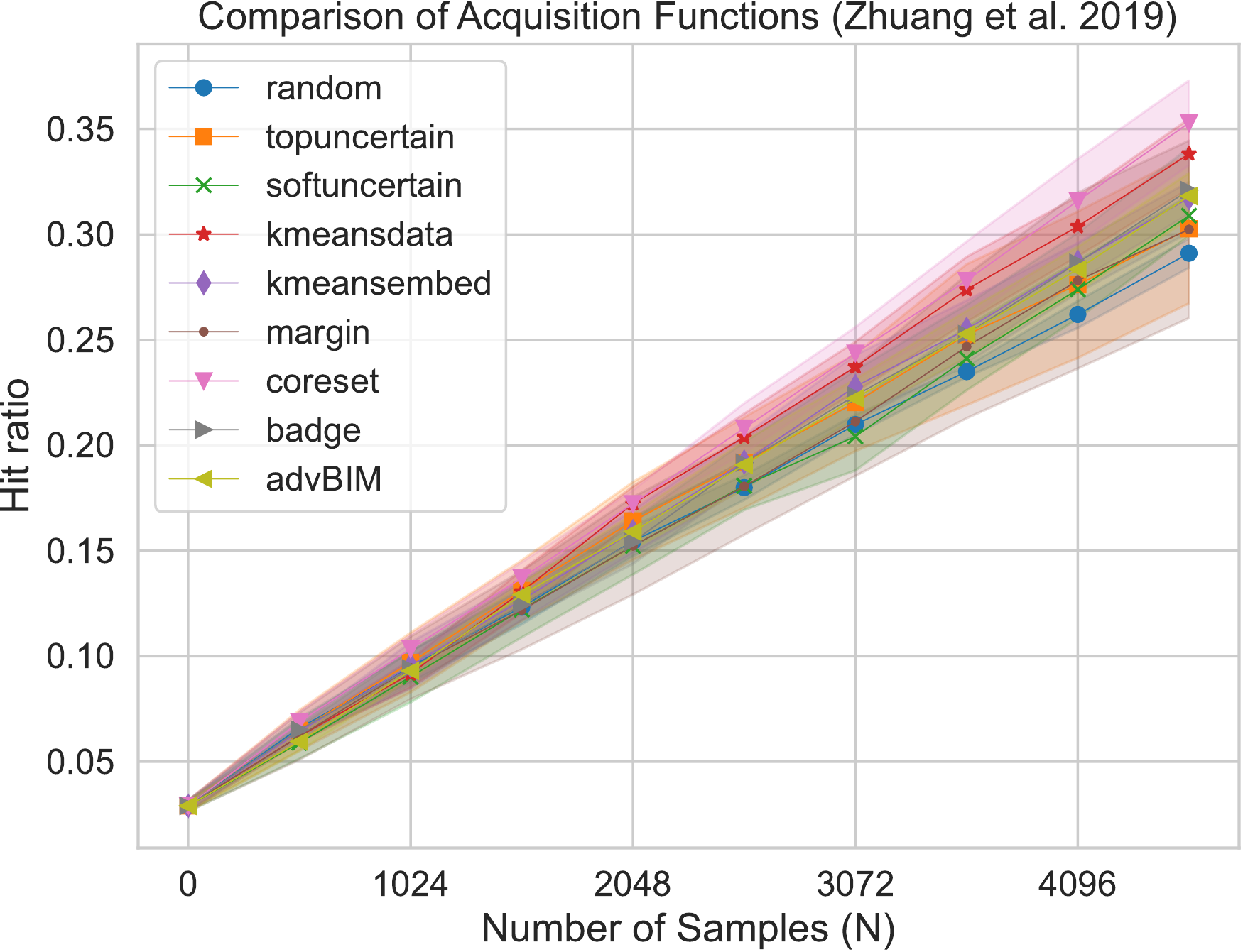}};
                        \end{tikzpicture}
                    }
                \end{subfigure}
                \&
                \\
            \\
           
            \\
            };
            \node [draw=none, rotate=90] at ([xshift=-8mm, yshift=2mm]fig-1-1.west) {\small batch size = 16};
            \node [draw=none, rotate=90] at ([xshift=-8mm, yshift=2mm]fig-2-1.west) {\small batch size = 32};
            \node [draw=none, rotate=90] at ([xshift=-8mm, yshift=2mm]fig-3-1.west) {\small batch size = 64};
            \node [draw=none, rotate=90] at ([xshift=-8mm, yshift=2mm]fig-4-1.west) {\small batch size = 128};
            \node [draw=none, rotate=90] at ([xshift=-8mm, yshift=2mm]fig-5-1.west) {\small batch size = 256};
            \node [draw=none, rotate=90] at ([xshift=-8mm, yshift=2mm]fig-6-1.west) {\small batch size = 512};
            \node [draw=none] at ([xshift=-6mm, yshift=3mm]fig-1-1.north) {\small Shifrut et al. 2018};
            \node [draw=none] at ([xshift=-9mm, yshift=3mm]fig-1-2.north) {\small Schmidt et al. 2021 (IFNg)};
            \node [draw=none] at ([xshift=-11mm, yshift=3mm]fig-1-3.north) {\small Schmidt et al. 2021 (IL-2)};
            \node [draw=none] at ([xshift=-13mm, yshift=2.5mm]fig-1-4.north) {\small Zhuang et al. 2019};
\end{tikzpicture}}
        \vspace{-2em}
        \caption{The hit ratio of different acquisition for BNN model, different target datasets, and different acquisition batch sizes. We use {STRING} treatment descriptors here. The x-axis shows the number of data points collected so far during the active learning cycles. The y-axis shows the ratio of the set of interesting genes that have been found by the acquisition function up until each cycle.}
        \vspace{-5mm}
        \label{fig:hitratio_bnn_feat_string_alldatasets_allbathcsizes}
    \end{figure*}

\begin{figure*}
    \vspace{-2mm}
        \centering
        \makebox[0.72\paperwidth]{\begin{tikzpicture}[ampersand replacement=\&]
            \matrix (fig) [matrix of nodes]{ 
\begin{subfigure}{0.27\columnwidth}
                    \hspace{-17mm}
                    \centering
                    \resizebox{\linewidth}{!}{
                        \begin{tikzpicture}
                            \node (img)  {\includegraphics[width=\textwidth]{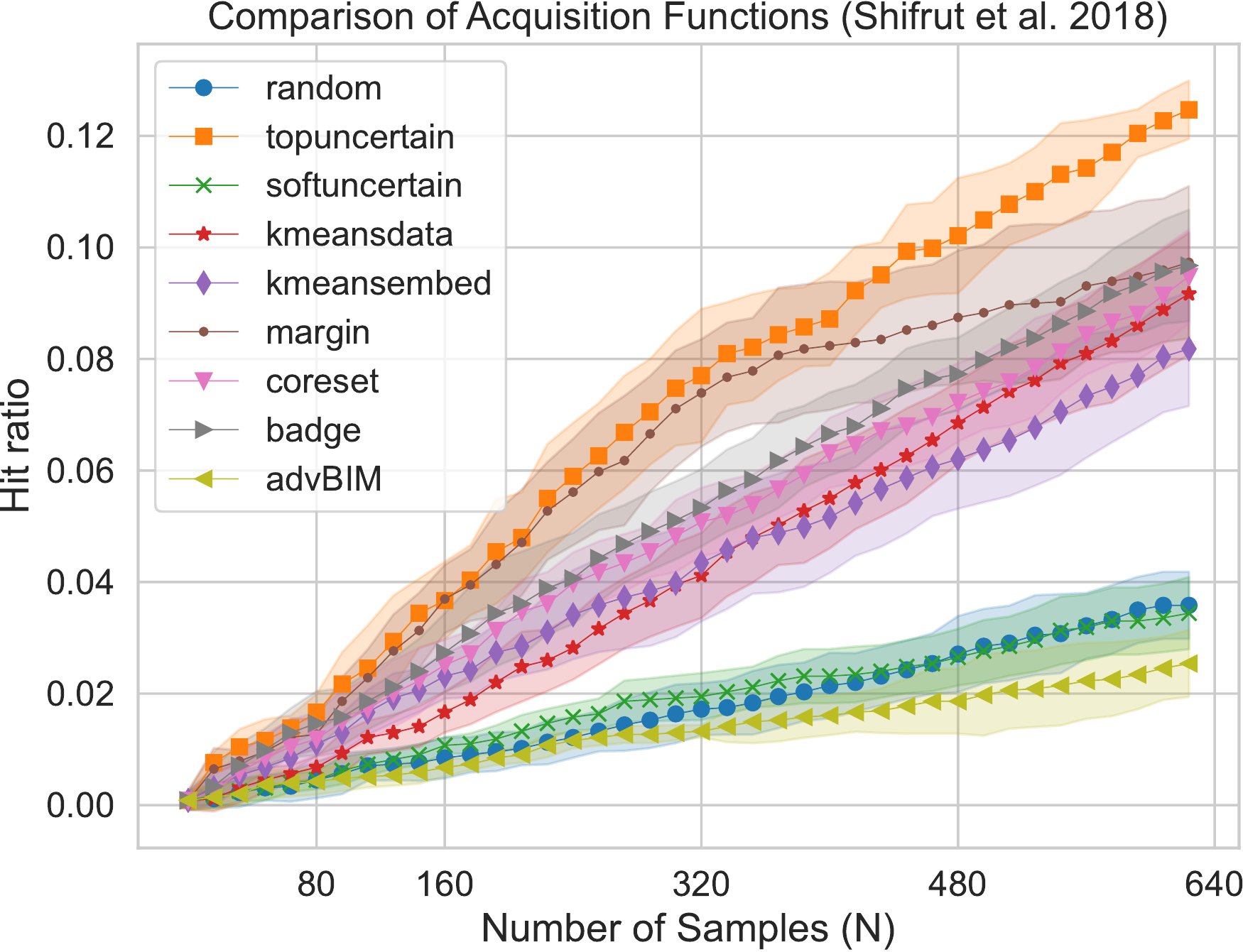}};
                        \end{tikzpicture}
                    }
                \end{subfigure}
                \&
                 \begin{subfigure}{0.27\columnwidth}
                    \hspace{-23mm}
                    \centering
                    \resizebox{\linewidth}{!}{
                        \begin{tikzpicture}
                            \node (img)  {\includegraphics[width=\textwidth]{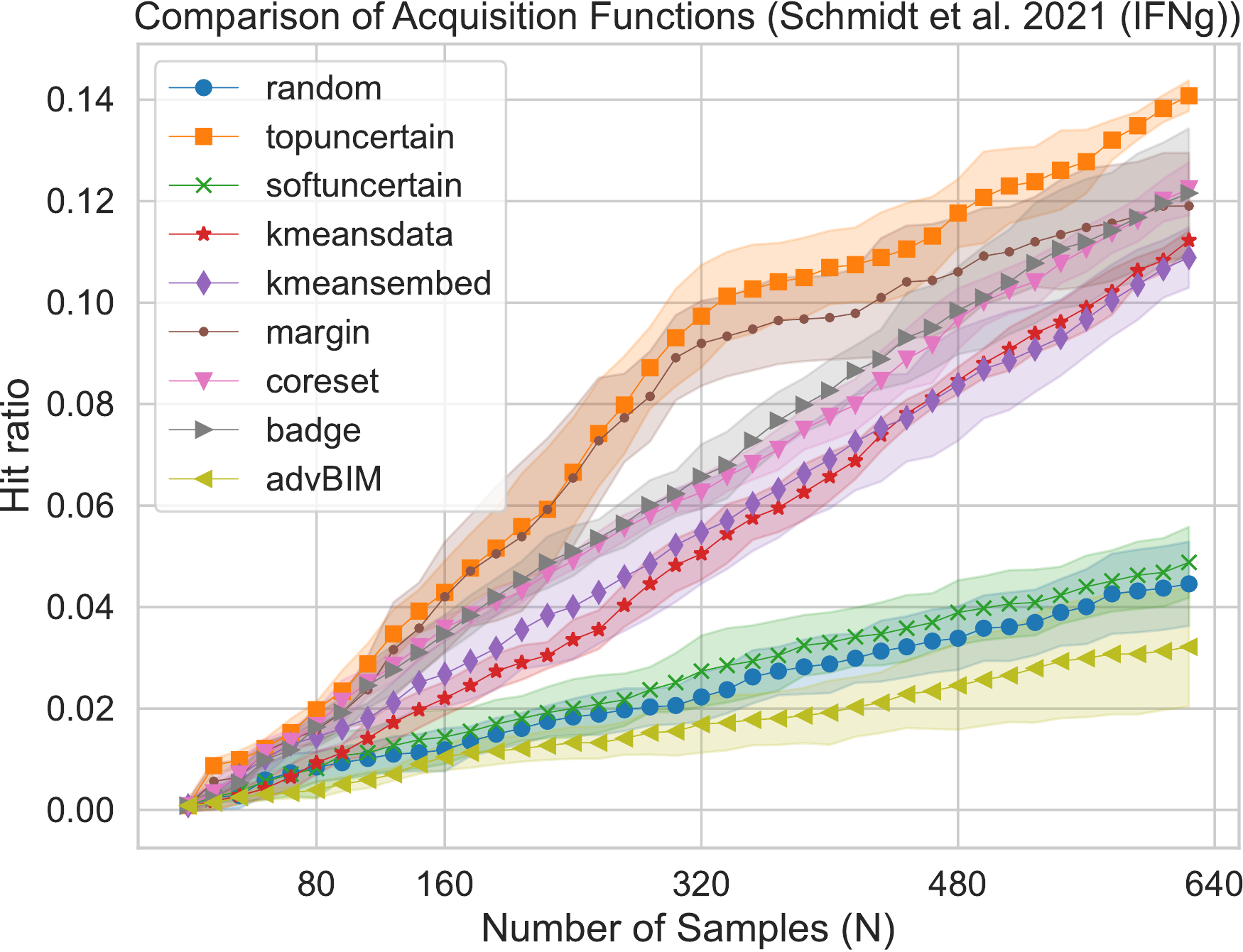}};
                        \end{tikzpicture}
                    }
                \end{subfigure}
                \&
                 \begin{subfigure}{0.27\columnwidth}
                    \hspace{-28mm}
                    \centering
                    \resizebox{\linewidth}{!}{
                        \begin{tikzpicture}
                            \node (img)  {\includegraphics[width=\textwidth]{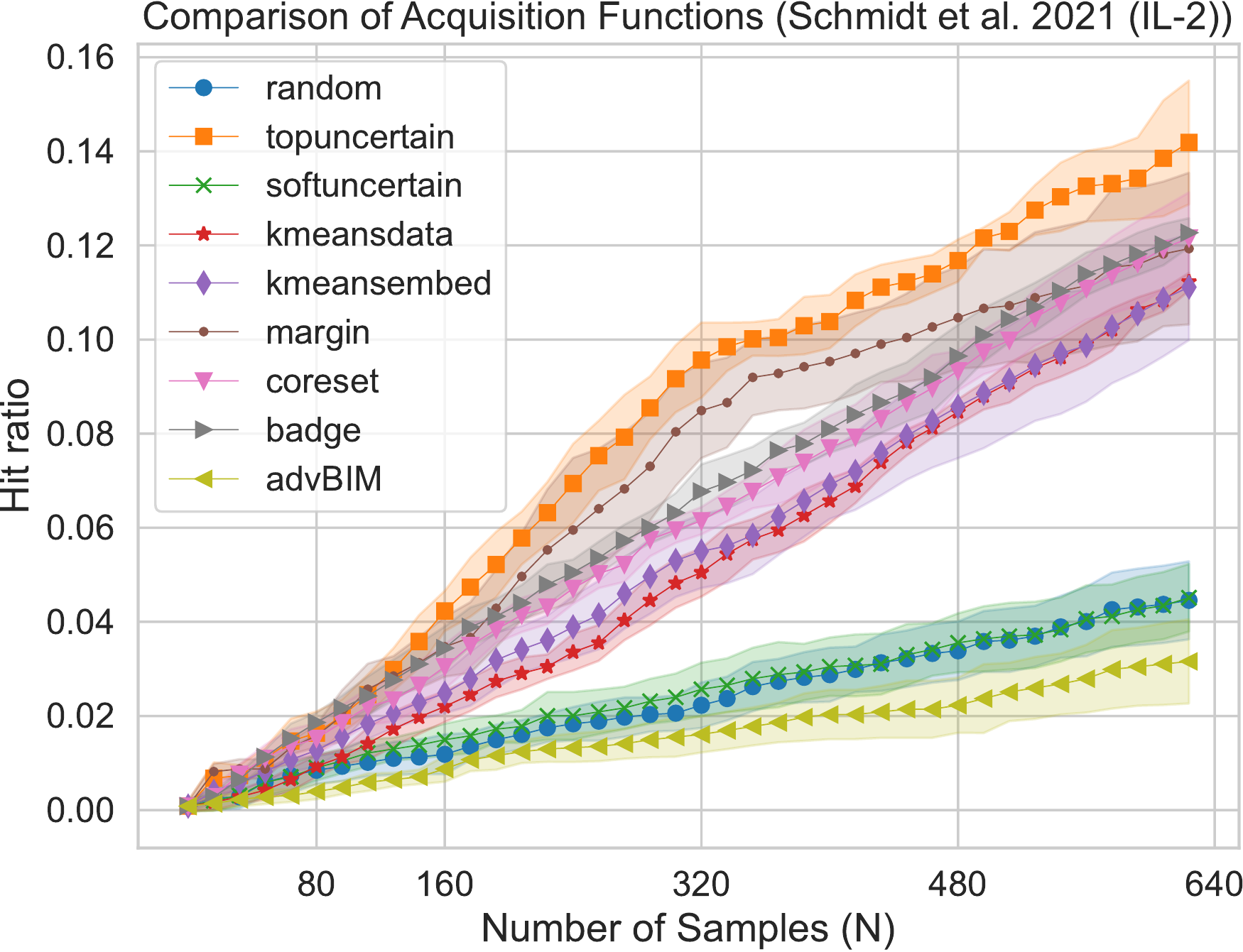}};
                        \end{tikzpicture}
                    }
                \end{subfigure}
                \&
                \begin{subfigure}{0.28\columnwidth}
                    \hspace{-32mm}
                    \centering
                    \resizebox{\linewidth}{!}{
                        \begin{tikzpicture}
                            \node (img)  {\includegraphics[width=\textwidth]{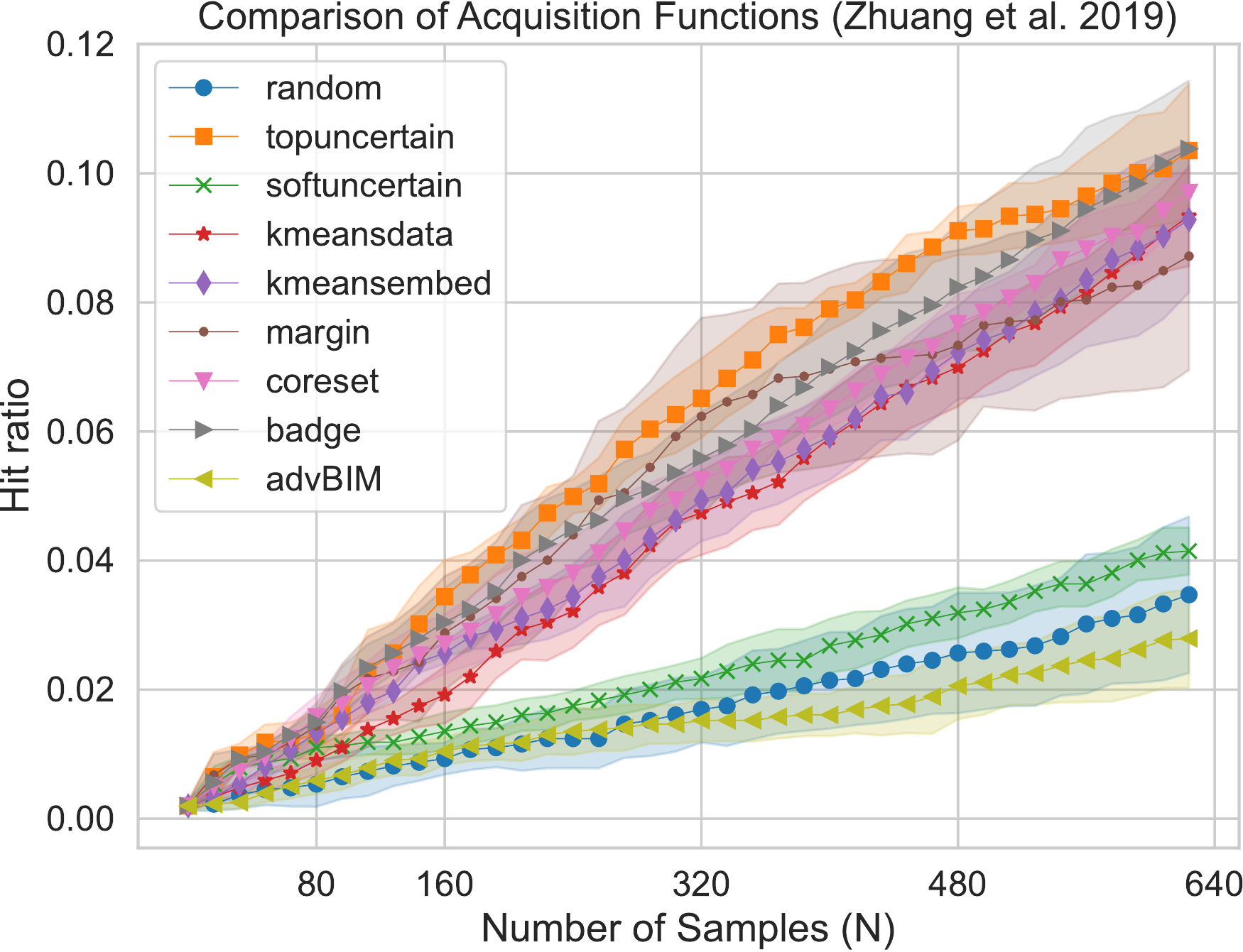}};
                        \end{tikzpicture}
                    }
                \end{subfigure}
                \&
            \\
\begin{subfigure}{0.27\columnwidth}
                    \hspace{-17mm}
                    \centering
                    \resizebox{\linewidth}{!}{
                        \begin{tikzpicture}
                            \node (img)  {\includegraphics[width=\textwidth]{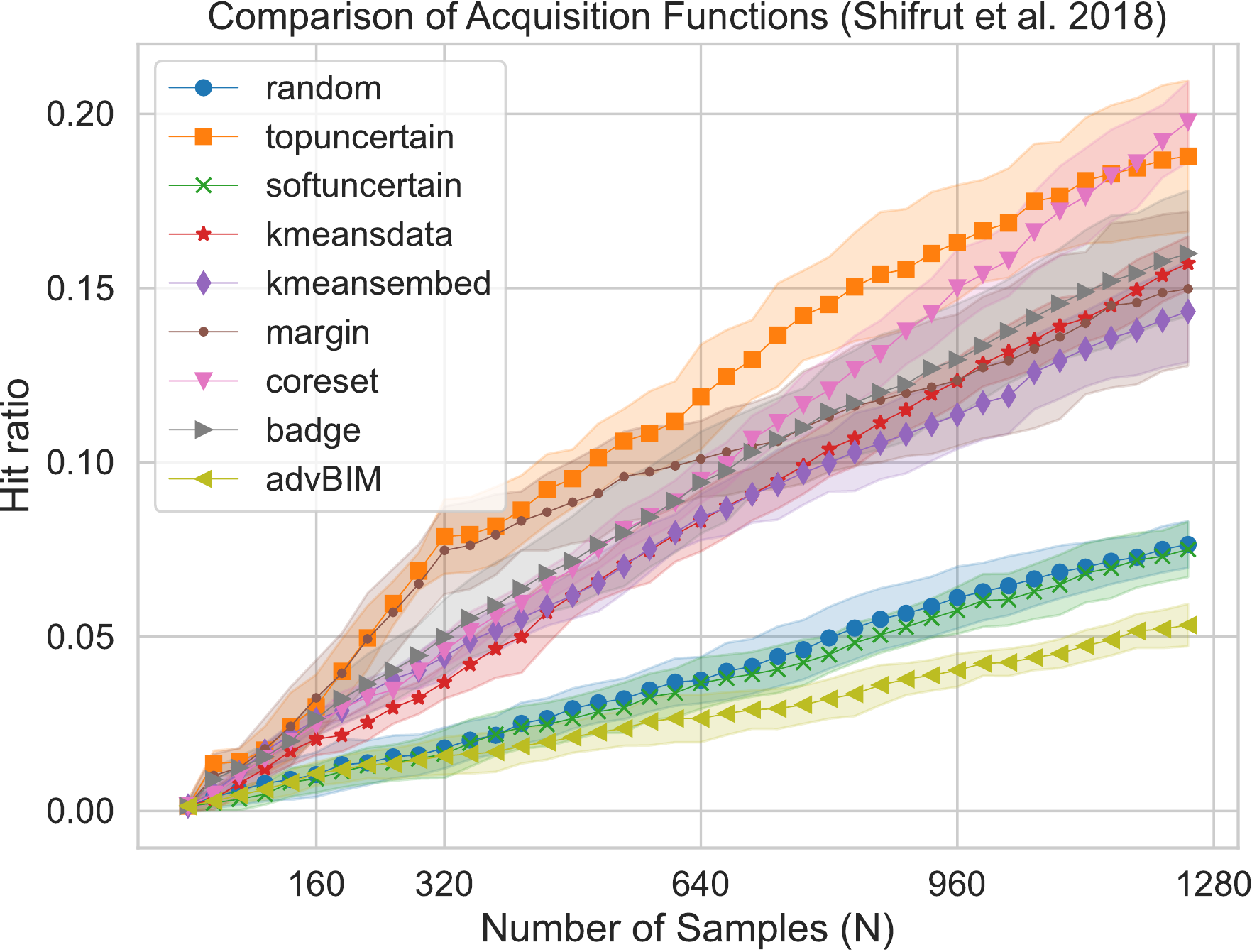}};
                        \end{tikzpicture}
                    }
                \end{subfigure}
                \&
                \begin{subfigure}{0.27\columnwidth}
                    \hspace{-23mm}
                    \centering
                    \resizebox{\linewidth}{!}{
                        \begin{tikzpicture}
                            \node (img)  {\includegraphics[width=\textwidth]{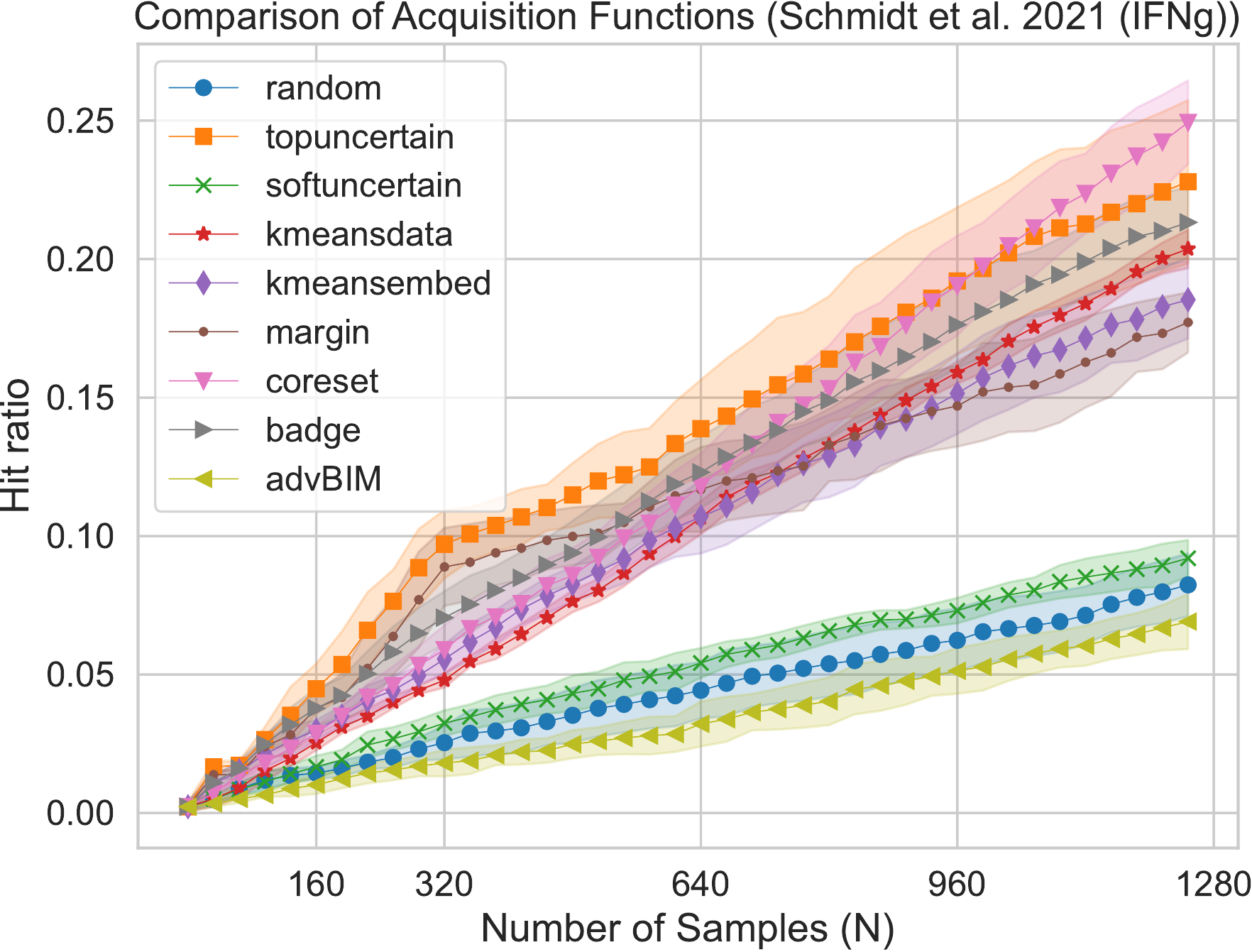}};
                        \end{tikzpicture}
                    }
                \end{subfigure}
                \&
                \begin{subfigure}{0.27\columnwidth}
                    \hspace{-28mm}
                    \centering
                    \resizebox{\linewidth}{!}{
                        \begin{tikzpicture}
                            \node (img)  {\includegraphics[width=\textwidth]{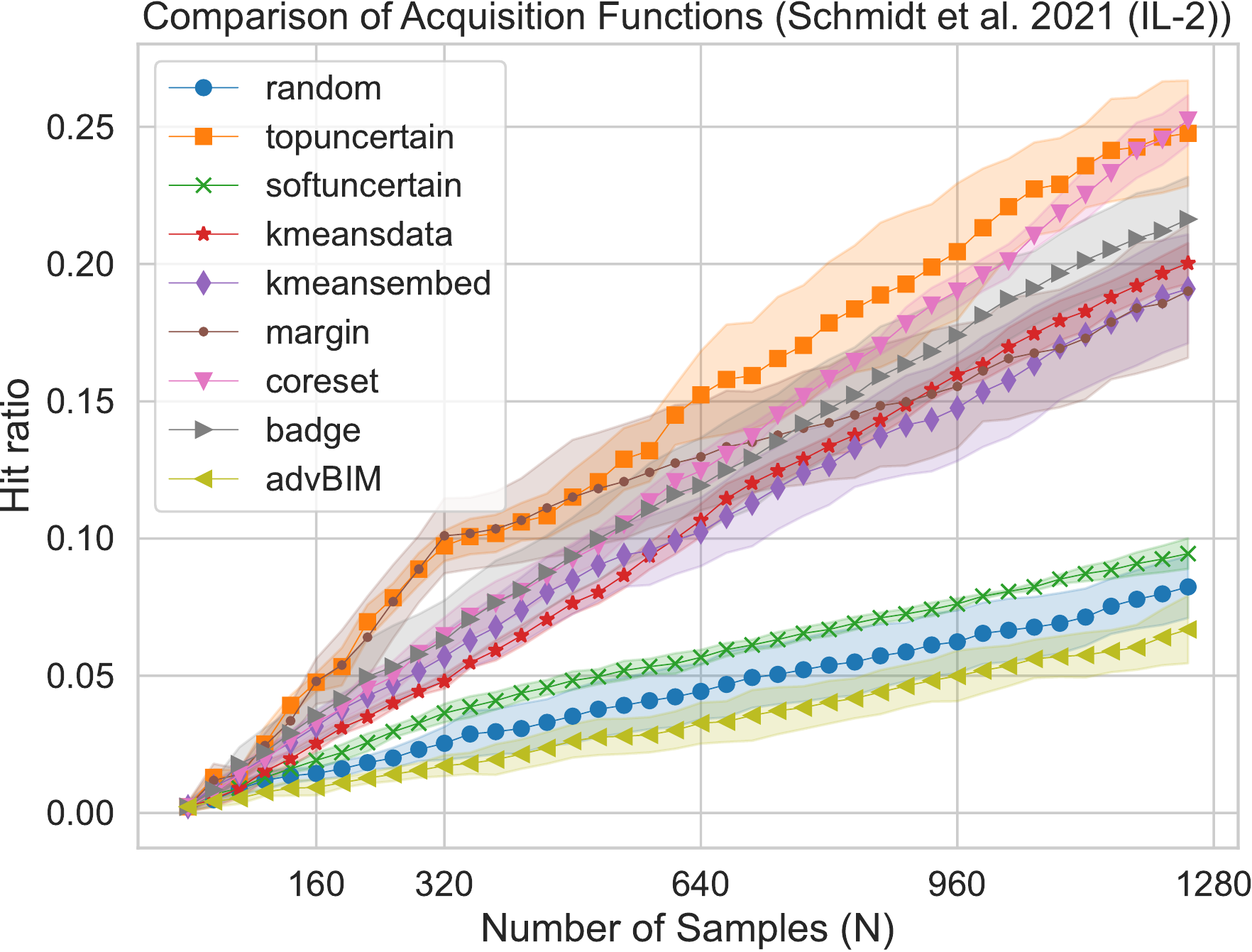}};
                        \end{tikzpicture}
                    }
                \end{subfigure}
                \&
                \begin{subfigure}{0.28\columnwidth}
                    \hspace{-32mm}
                    \centering
                    \resizebox{\linewidth}{!}{
                        \begin{tikzpicture}
                            \node (img)  {\includegraphics[width=\textwidth]{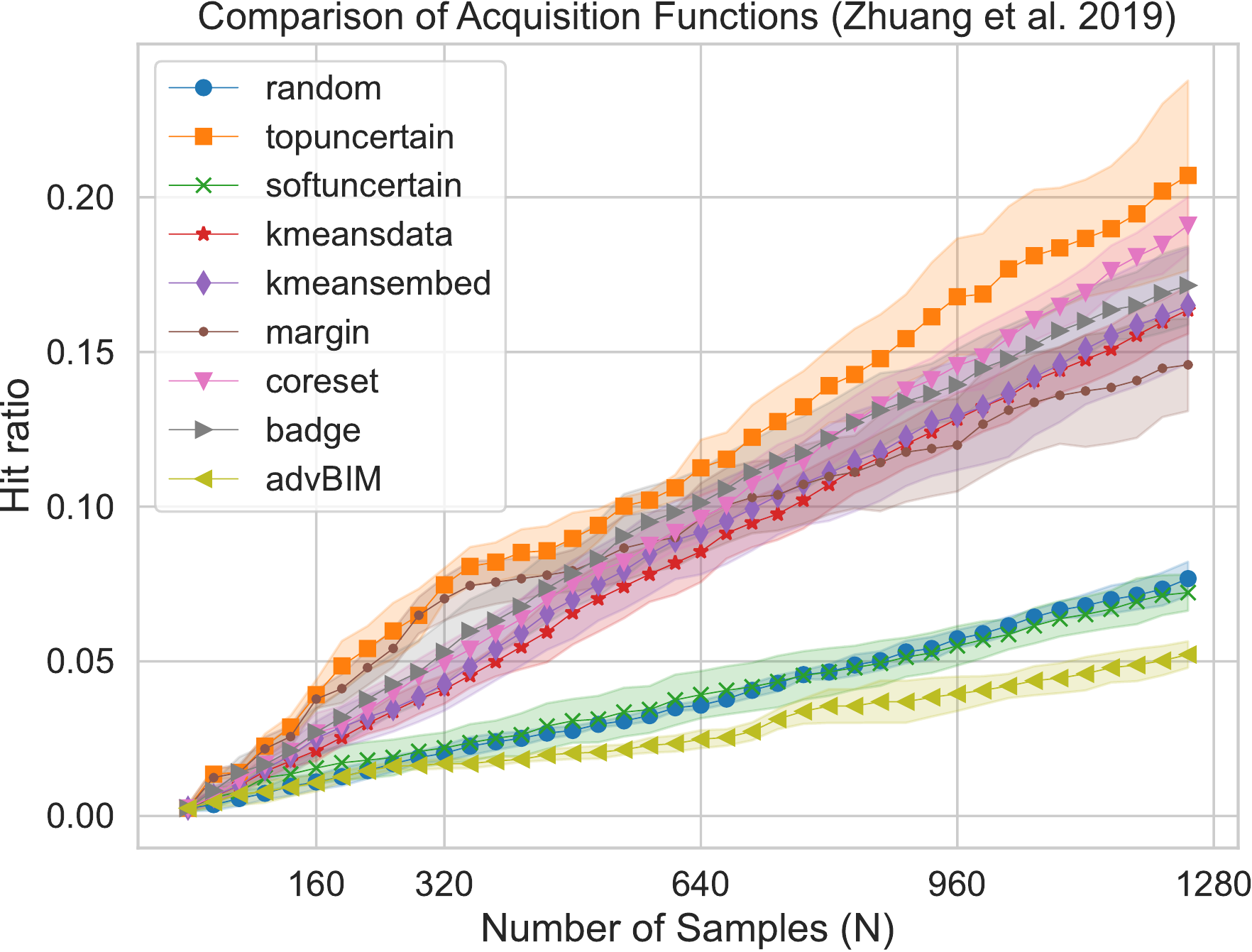}};
                        \end{tikzpicture}
                    }
                \end{subfigure}
                \&
                \\
\begin{subfigure}{0.27\columnwidth}
                    \hspace{-17mm}
                    \centering
                    \resizebox{\linewidth}{!}{
                        \begin{tikzpicture}
                            \node (img)  {\includegraphics[width=\textwidth]{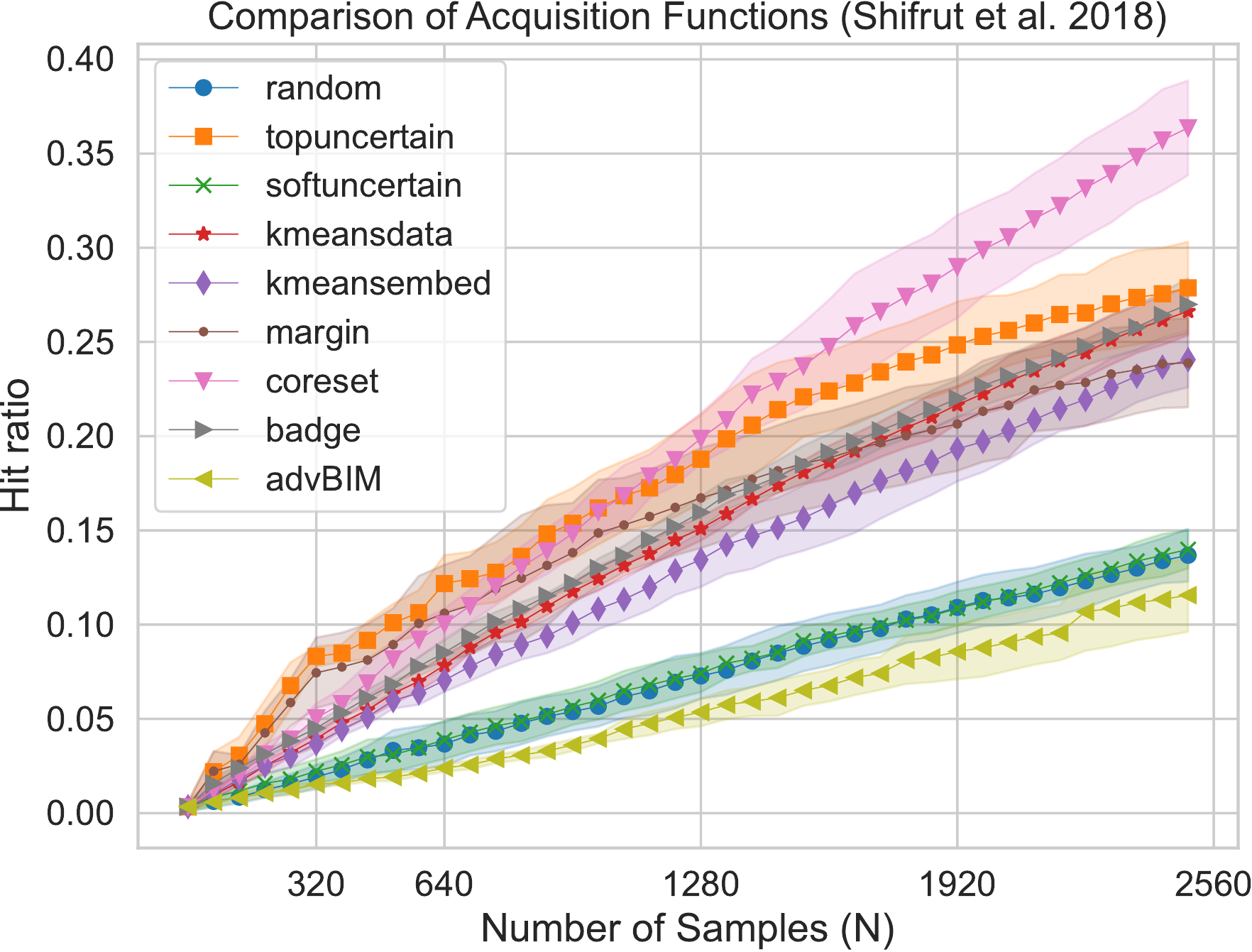}};
                        \end{tikzpicture}
                    }
                \end{subfigure}
                \&
                \begin{subfigure}{0.27\columnwidth}
                    \hspace{-23mm}
                    \centering
                    \resizebox{\linewidth}{!}{
                        \begin{tikzpicture}
                            \node (img)  {\includegraphics[width=\textwidth]{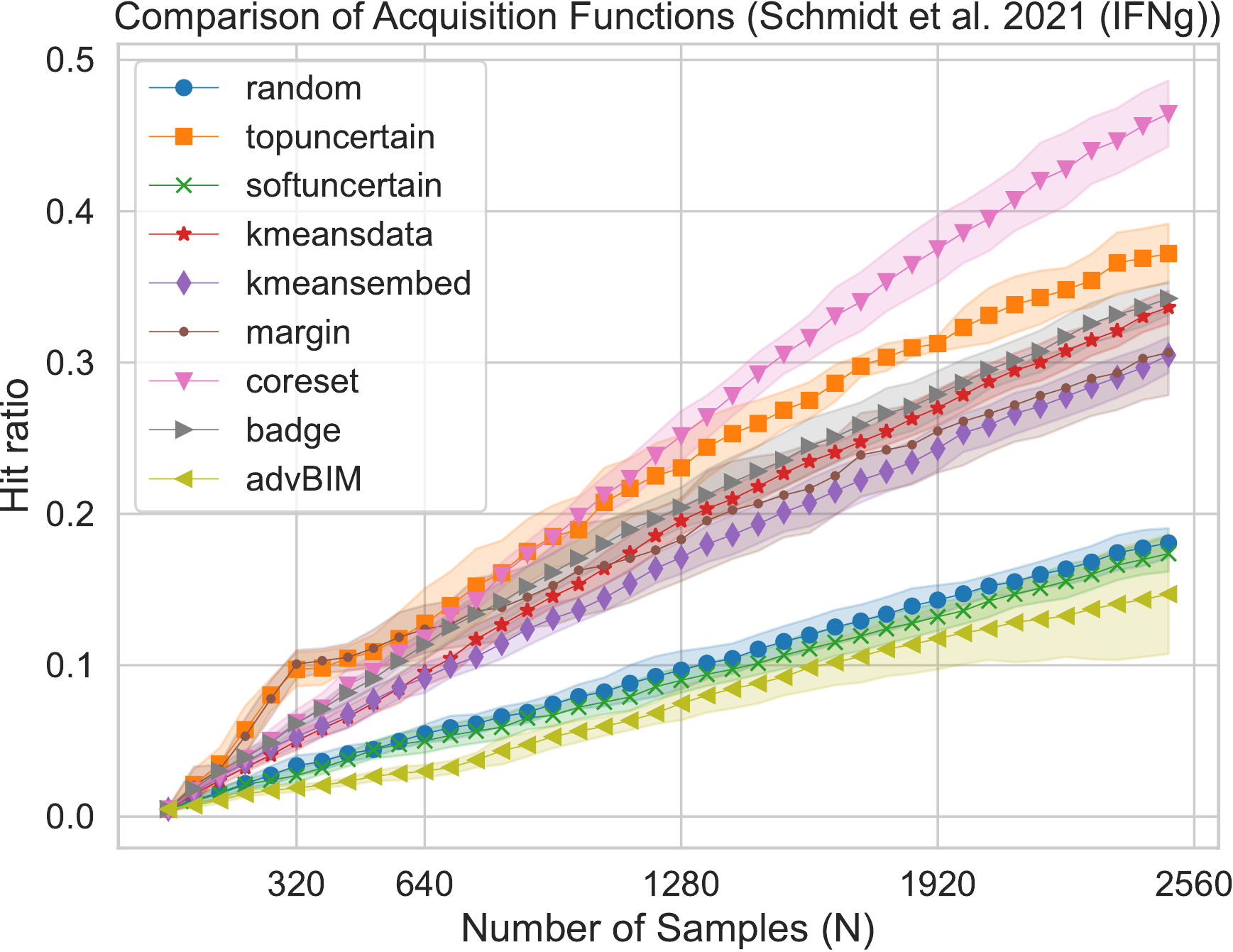}};
                        \end{tikzpicture}
                    }
                \end{subfigure}
                \&
                \begin{subfigure}{0.27\columnwidth}
                    \hspace{-28mm}
                    \centering
                    \resizebox{\linewidth}{!}{
                        \begin{tikzpicture}
                            \node (img)  {\includegraphics[width=\textwidth]{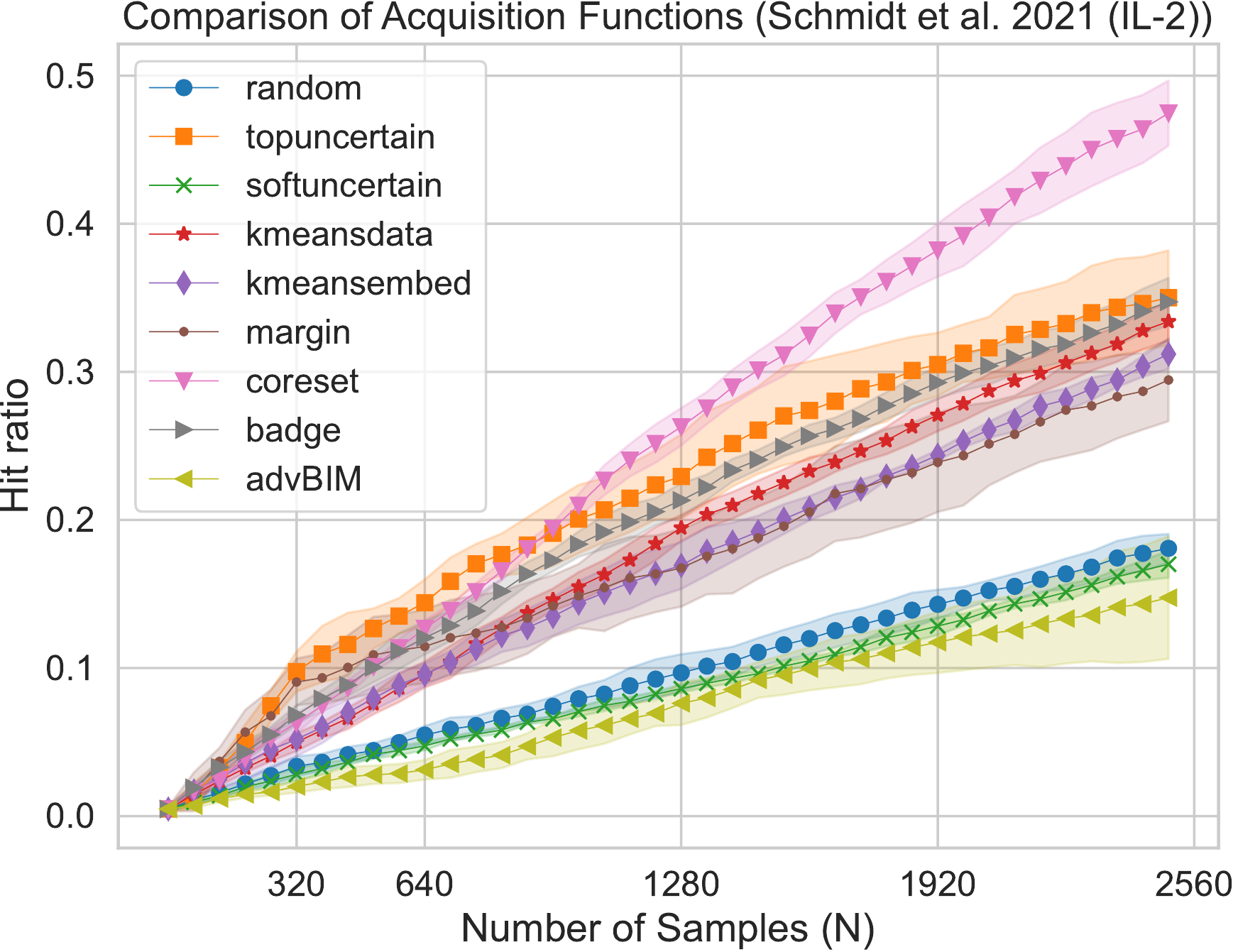}};
                        \end{tikzpicture}
                    }
                \end{subfigure}
                \&
                \begin{subfigure}{0.28\columnwidth}
                    \hspace{-32mm}
                    \centering
                    \resizebox{\linewidth}{!}{
                        \begin{tikzpicture}
                            \node (img)  {\includegraphics[width=\textwidth]{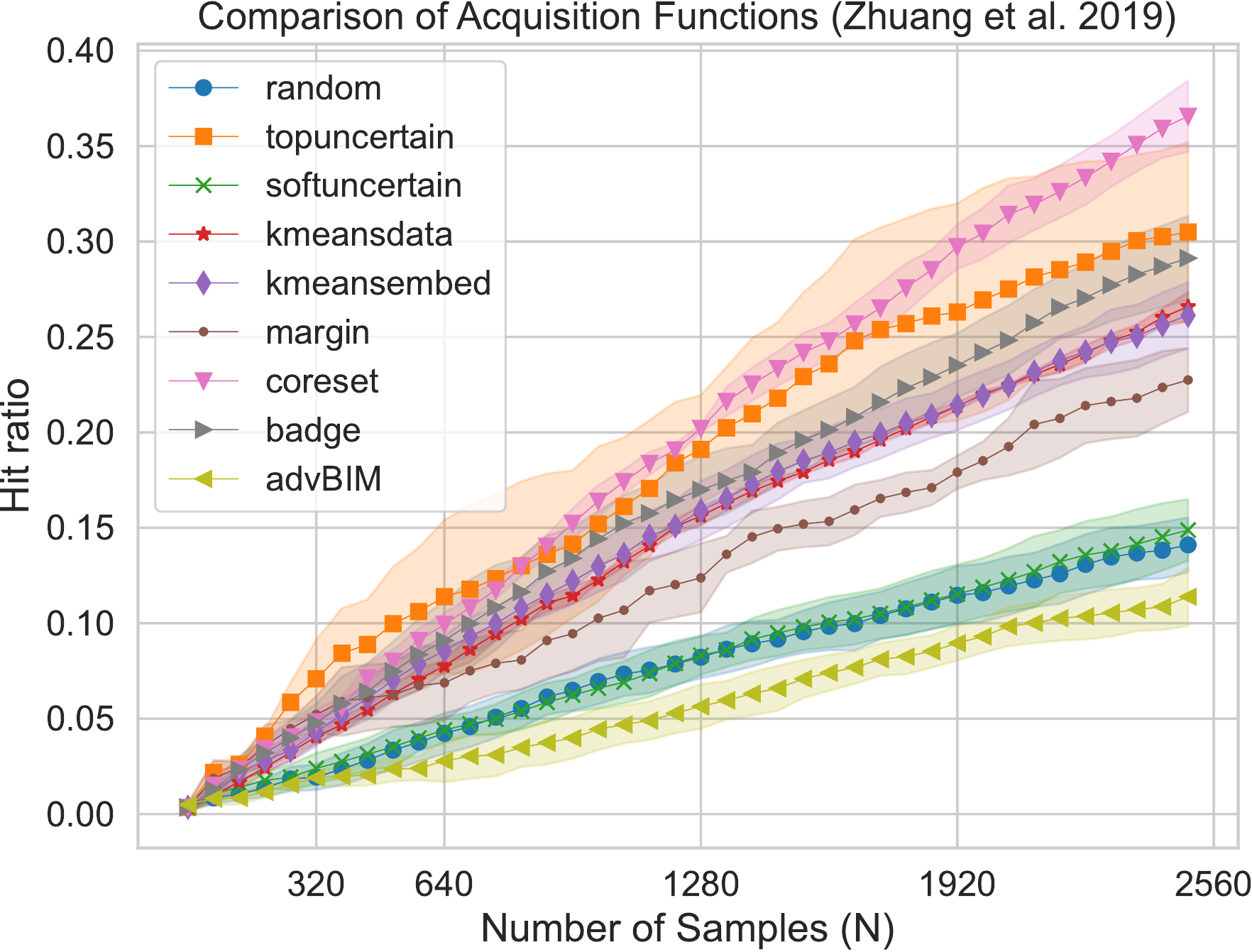}};
                        \end{tikzpicture}
                    }
                \end{subfigure}
                \&
                \\
\begin{subfigure}{0.27\columnwidth}
                    \hspace{-17mm}
                    \centering
                    \resizebox{\linewidth}{!}{
                        \begin{tikzpicture}
                            \node (img)  {\includegraphics[width=\textwidth]{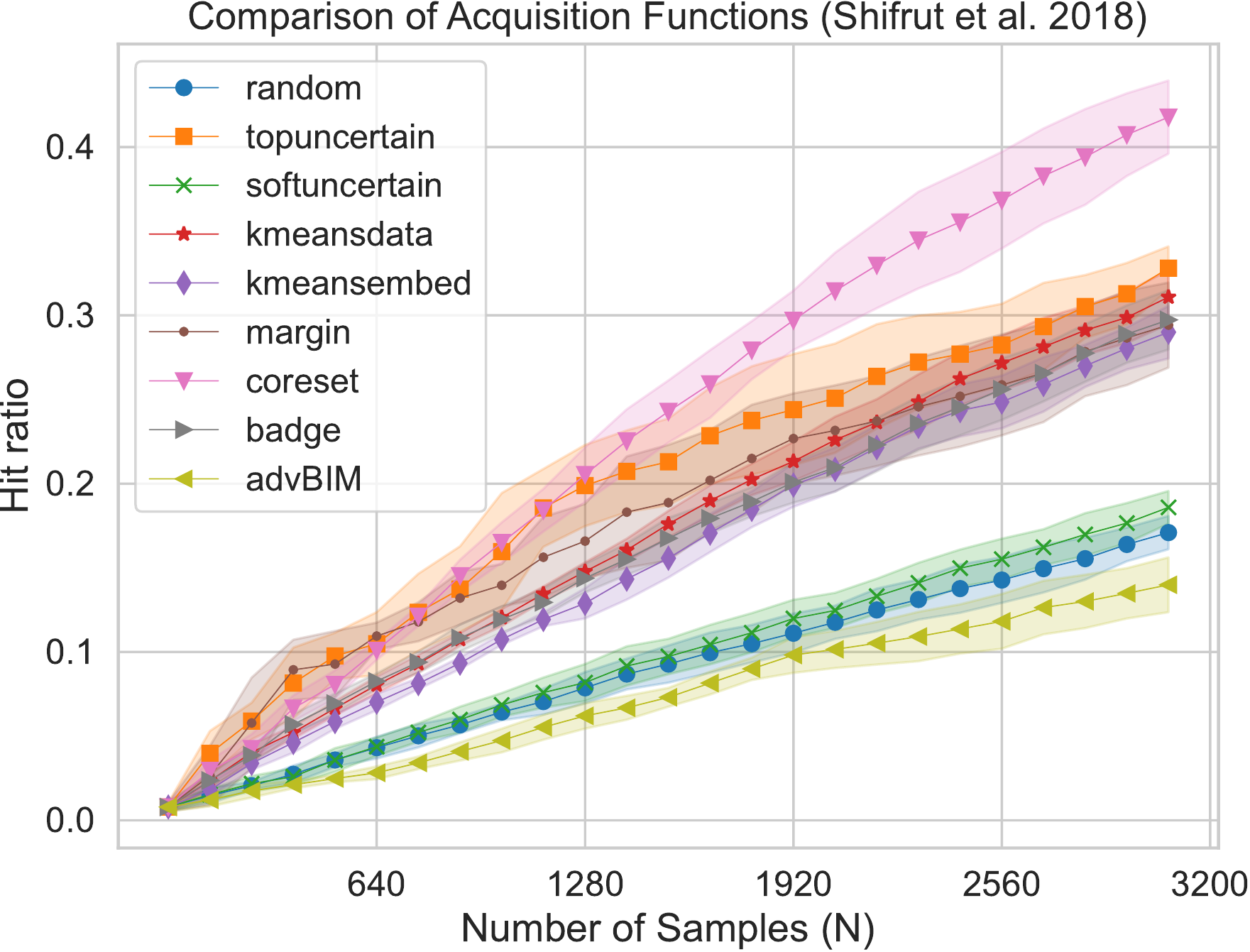}};
                        \end{tikzpicture}
                    }
                \end{subfigure}
                \&
                \begin{subfigure}{0.27\columnwidth}
                    \hspace{-23mm}
                    \centering
                    \resizebox{\linewidth}{!}{
                        \begin{tikzpicture}
                            \node (img)  {\includegraphics[width=\textwidth]{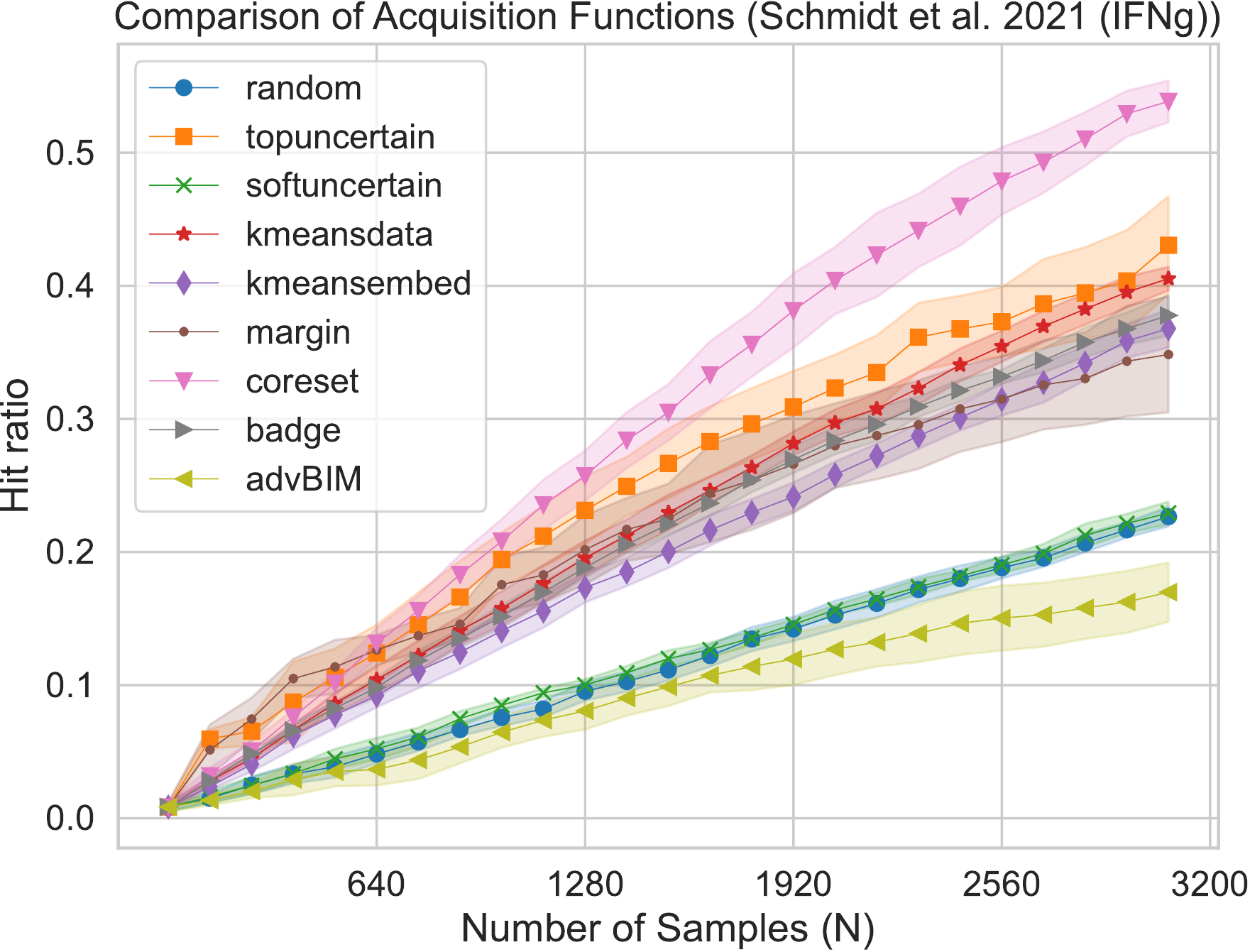}};
                        \end{tikzpicture}
                    }
                \end{subfigure}
                \&
                \begin{subfigure}{0.28\columnwidth}
                    \hspace{-28mm}
                    \centering
                    \resizebox{\linewidth}{!}{
                        \begin{tikzpicture}
                            \node (img)  {\includegraphics[width=\textwidth]{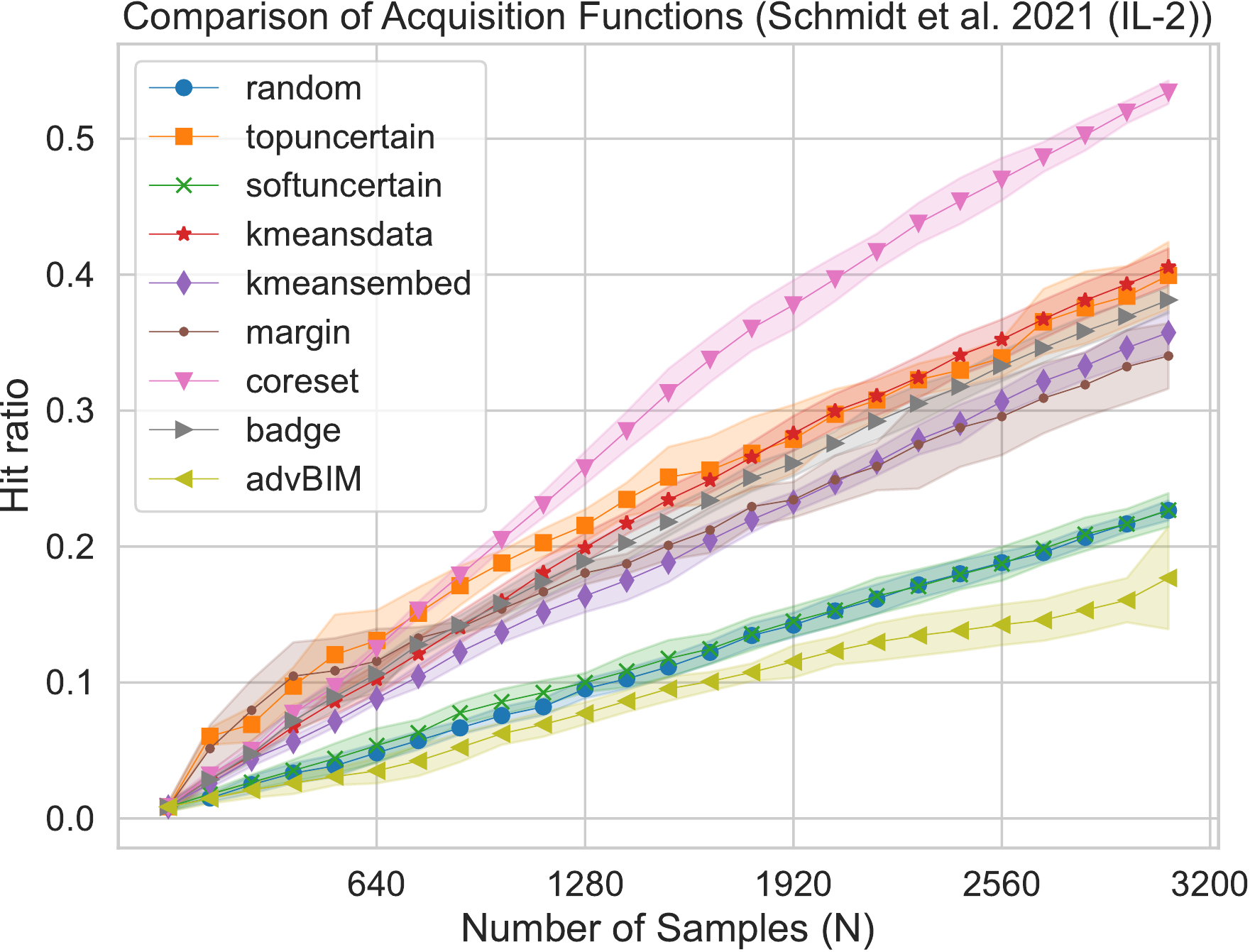}};
                        \end{tikzpicture}
                    }
                \end{subfigure}
                \&
                \begin{subfigure}{0.29\columnwidth}
                    \hspace{-32mm}
                    \centering
                    \resizebox{\linewidth}{!}{
                        \begin{tikzpicture}
                            \node (img)  {\includegraphics[width=\textwidth]{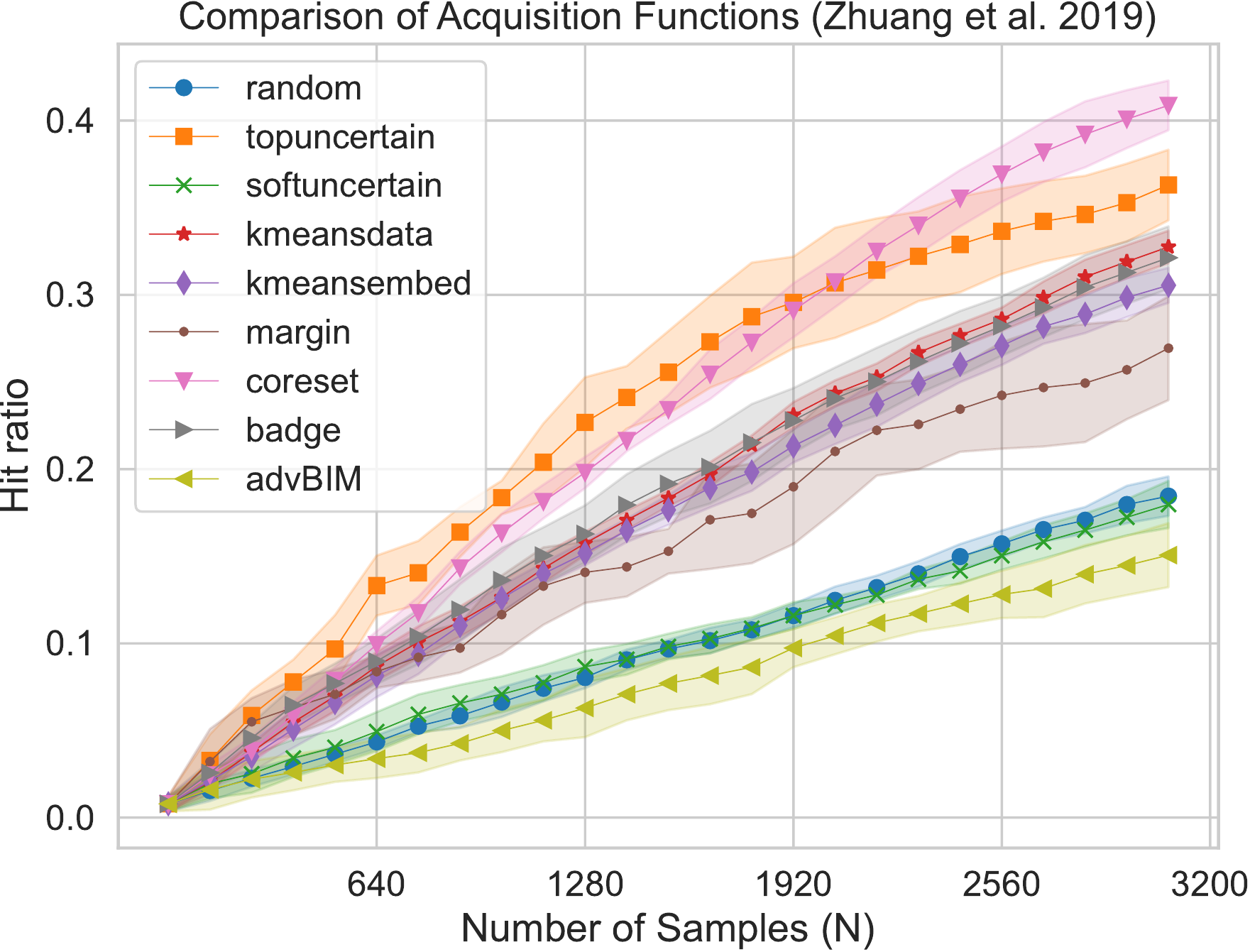}};
                        \end{tikzpicture}
                    }
                \end{subfigure}
                \&
                \\
\begin{subfigure}{0.275\columnwidth}
                    \hspace{-17mm}
                    \centering
                    \resizebox{\linewidth}{!}{
                        \begin{tikzpicture}
                            \node (img)  {\includegraphics[width=\textwidth]{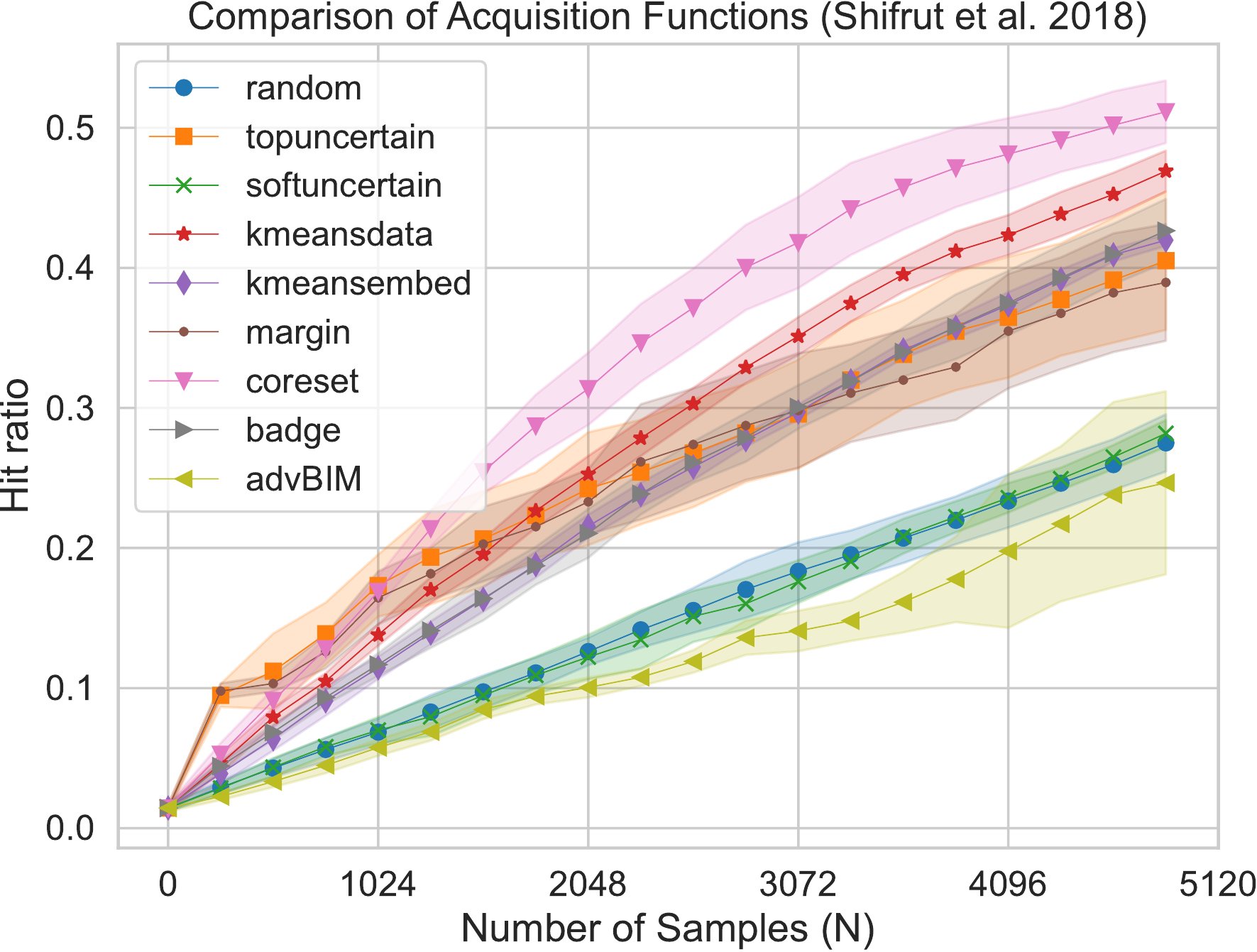}};
                        \end{tikzpicture}
                    }
                \end{subfigure}
                \&
                \begin{subfigure}{0.27\columnwidth}
                    \hspace{-23mm}
                    \centering
                    \resizebox{\linewidth}{!}{
                        \begin{tikzpicture}
                            \node (img)  {\includegraphics[width=\textwidth]{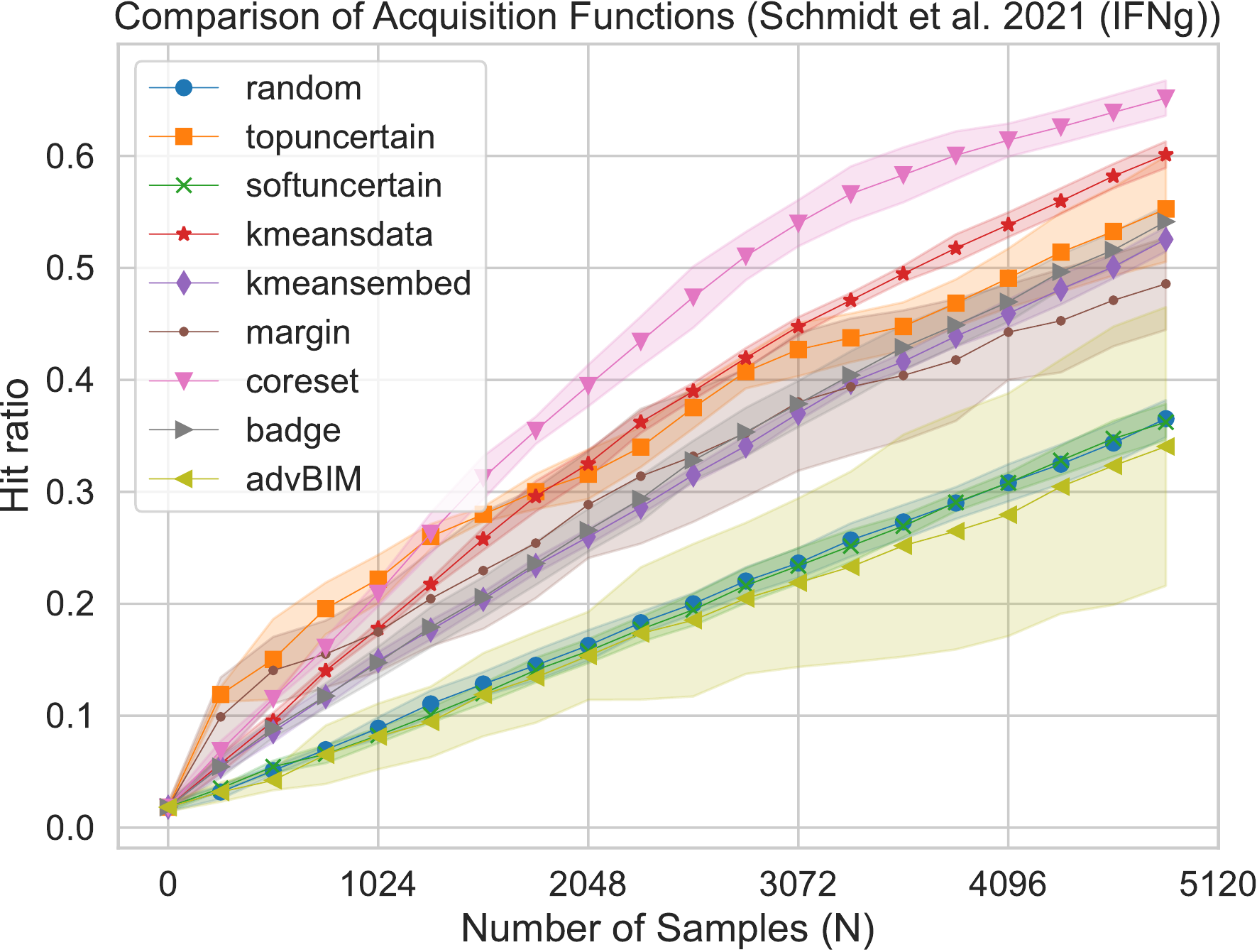}};
                        \end{tikzpicture}
                    }
                \end{subfigure}
                \&
                \begin{subfigure}{0.27\columnwidth}
                    \hspace{-28mm}
                    \centering
                    \resizebox{\linewidth}{!}{
                        \begin{tikzpicture}
                            \node (img)  {\includegraphics[width=\textwidth]{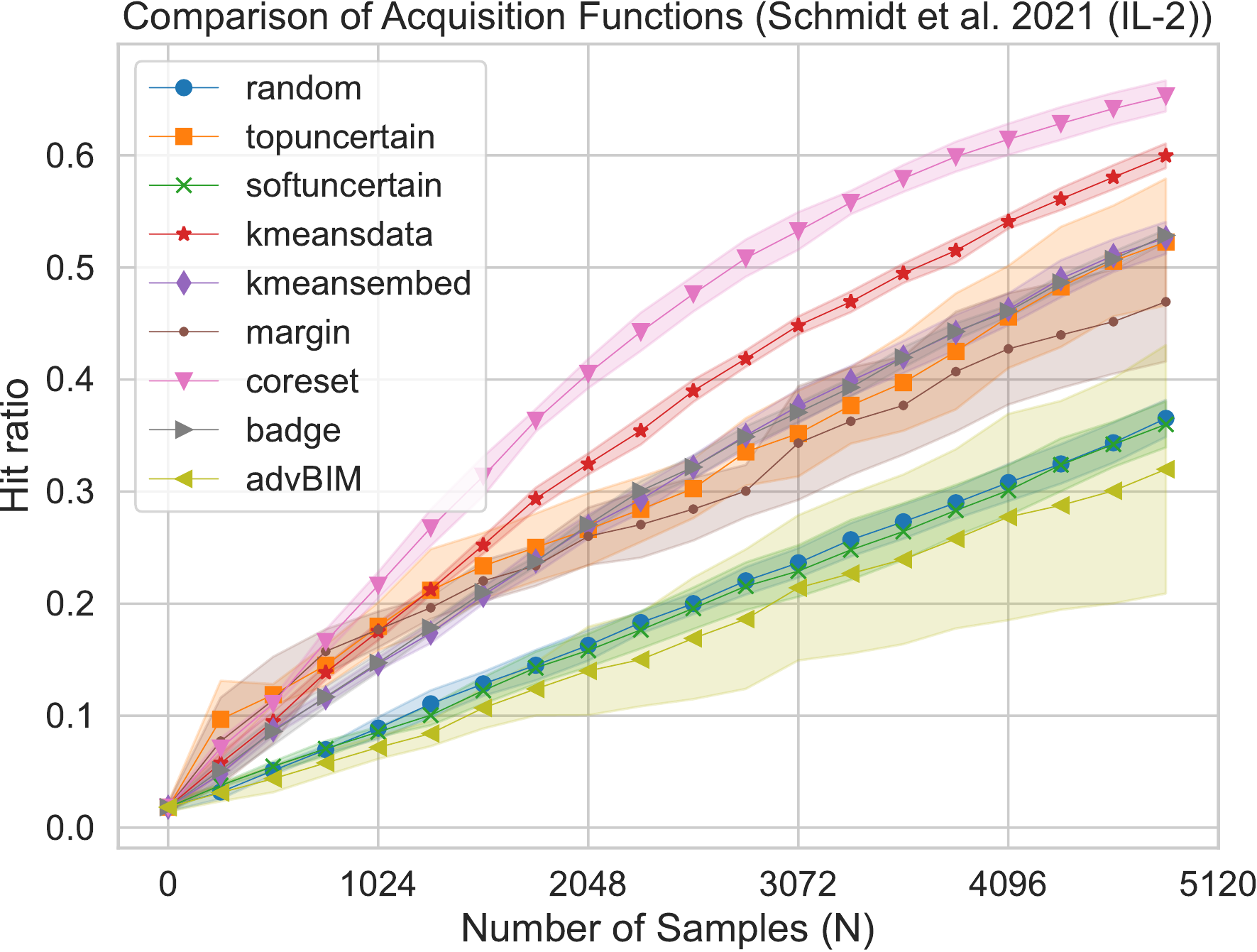}};
                        \end{tikzpicture}
                    }
                \end{subfigure}
                \&
                \begin{subfigure}{0.29\columnwidth}
                    \hspace{-32mm}
                    \centering
                    \resizebox{\linewidth}{!}{
                        \begin{tikzpicture}
                            \node (img)  {\includegraphics[width=\textwidth]{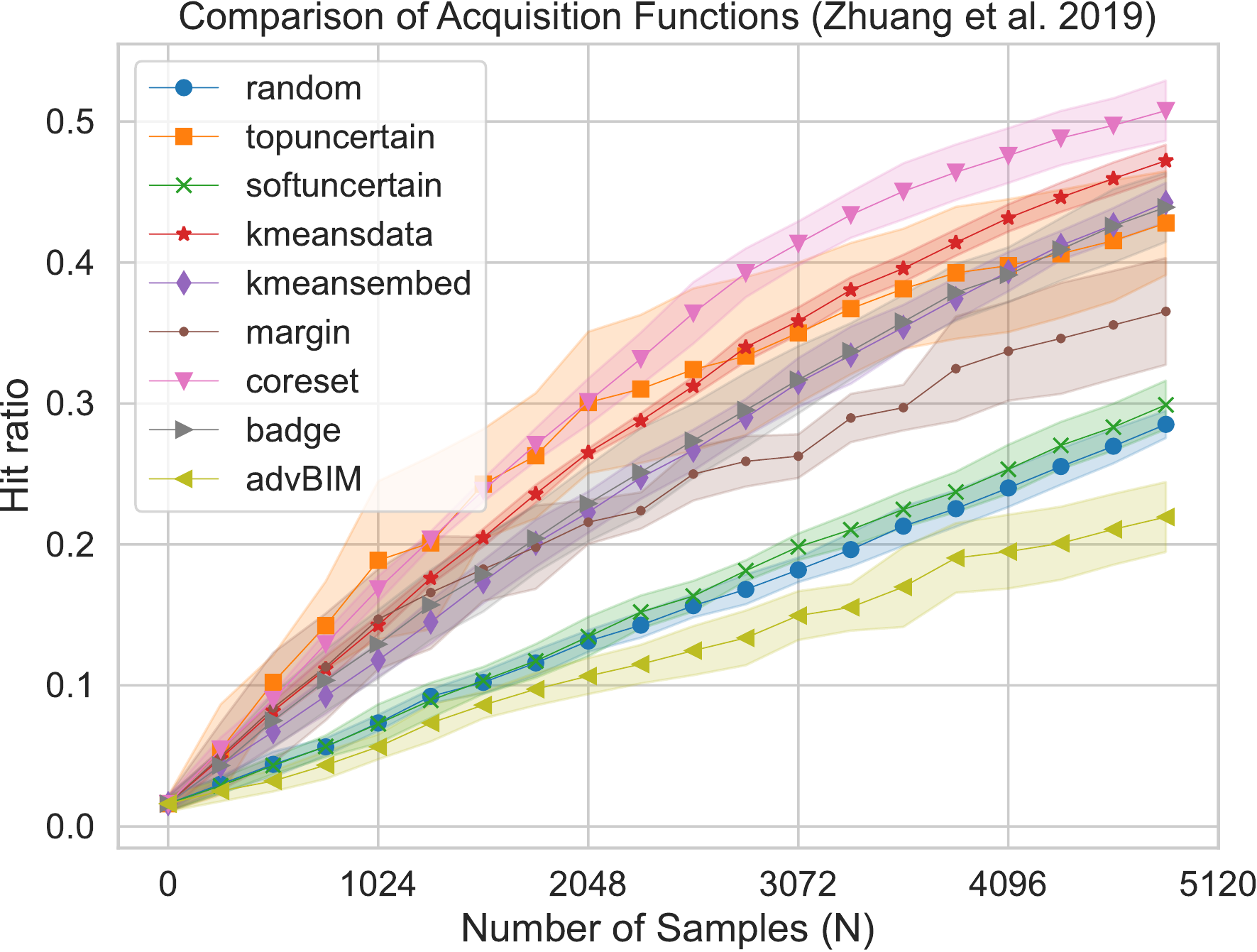}};
                        \end{tikzpicture}
                    }
                \end{subfigure}
                \&
                \\
\begin{subfigure}{0.28\columnwidth}
                    \hspace{-17mm}
                    \centering
                    \resizebox{\linewidth}{!}{
                        \begin{tikzpicture}
                            \node (img)  {\includegraphics[width=\textwidth]{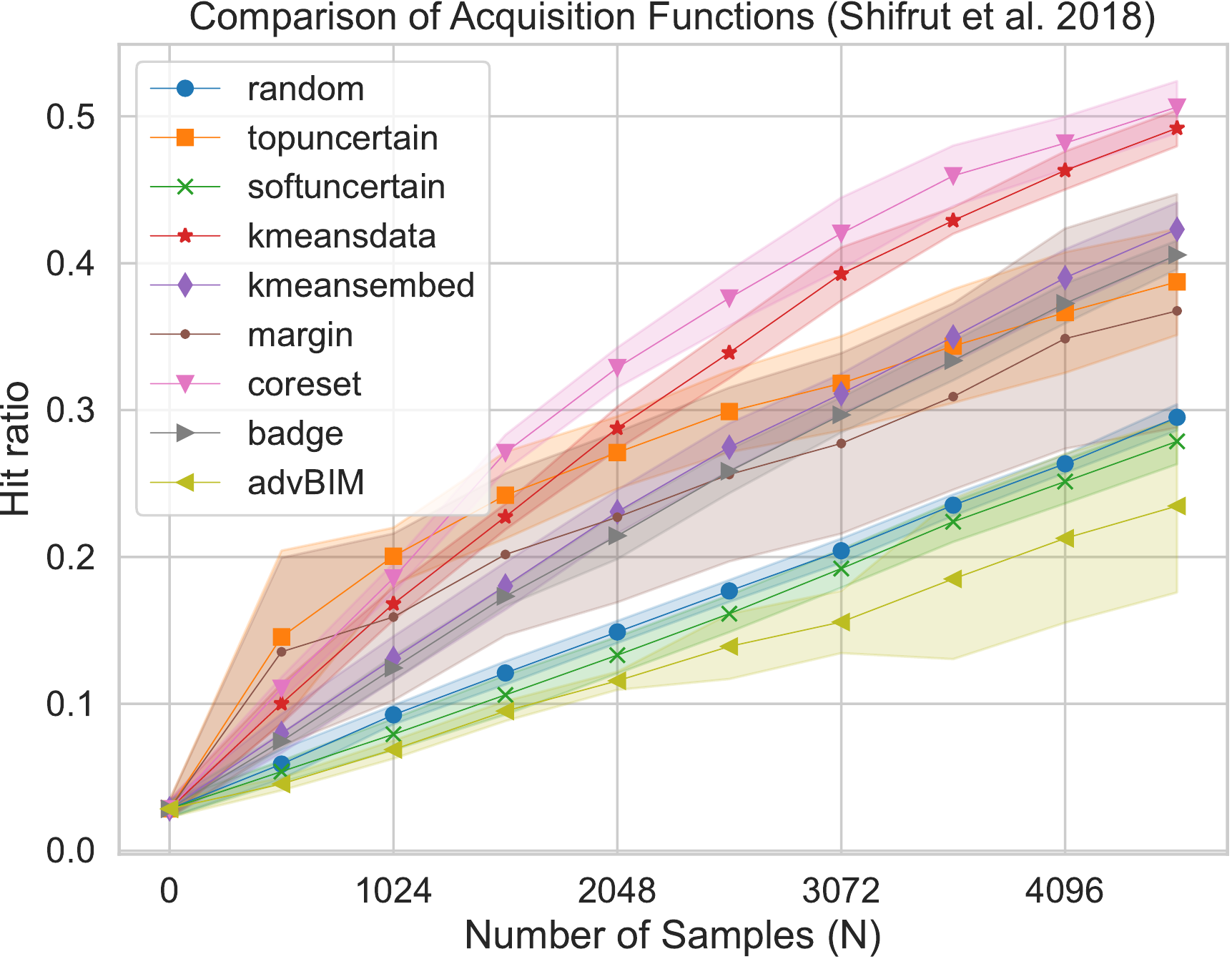}};
                        \end{tikzpicture}
                    }
                \end{subfigure}
                \&
                \begin{subfigure}{0.27\columnwidth}
                    \hspace{-23mm}
                    \centering
                    \resizebox{\linewidth}{!}{
                        \begin{tikzpicture}
                            \node (img)  {\includegraphics[width=\textwidth]{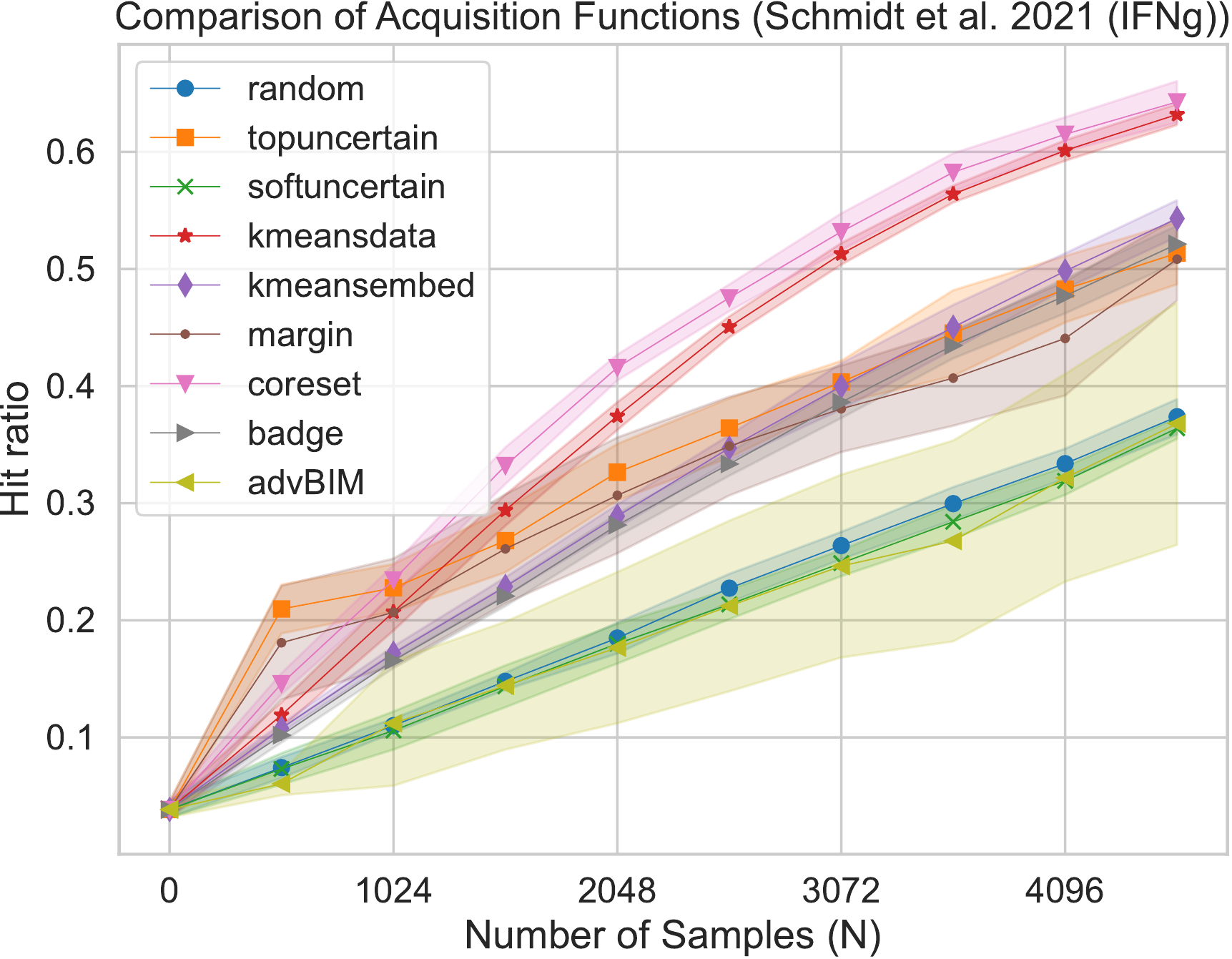}};
                        \end{tikzpicture}
                    }
                \end{subfigure}
                \&
                \begin{subfigure}{0.27\columnwidth}
                    \hspace{-28mm}
                    \centering
                    \resizebox{\linewidth}{!}{
                        \begin{tikzpicture}
                            \node (img)  {\includegraphics[width=\textwidth]{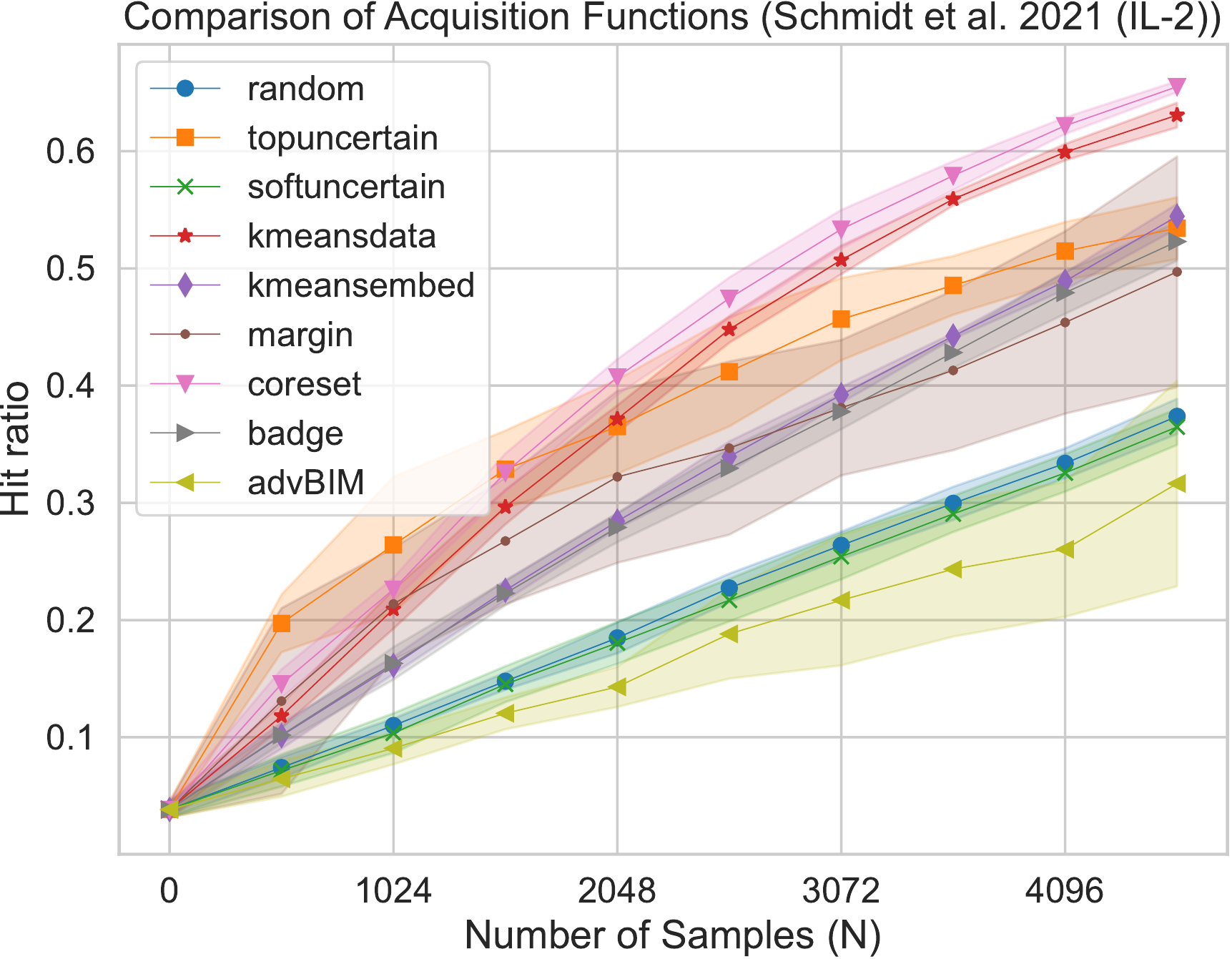}};
                        \end{tikzpicture}
                    }
                \end{subfigure}
                \&
                \begin{subfigure}{0.28\columnwidth}
                    \hspace{-32mm}
                    \centering
                    \resizebox{\linewidth}{!}{
                        \begin{tikzpicture}
                            \node (img)  {\includegraphics[width=\textwidth]{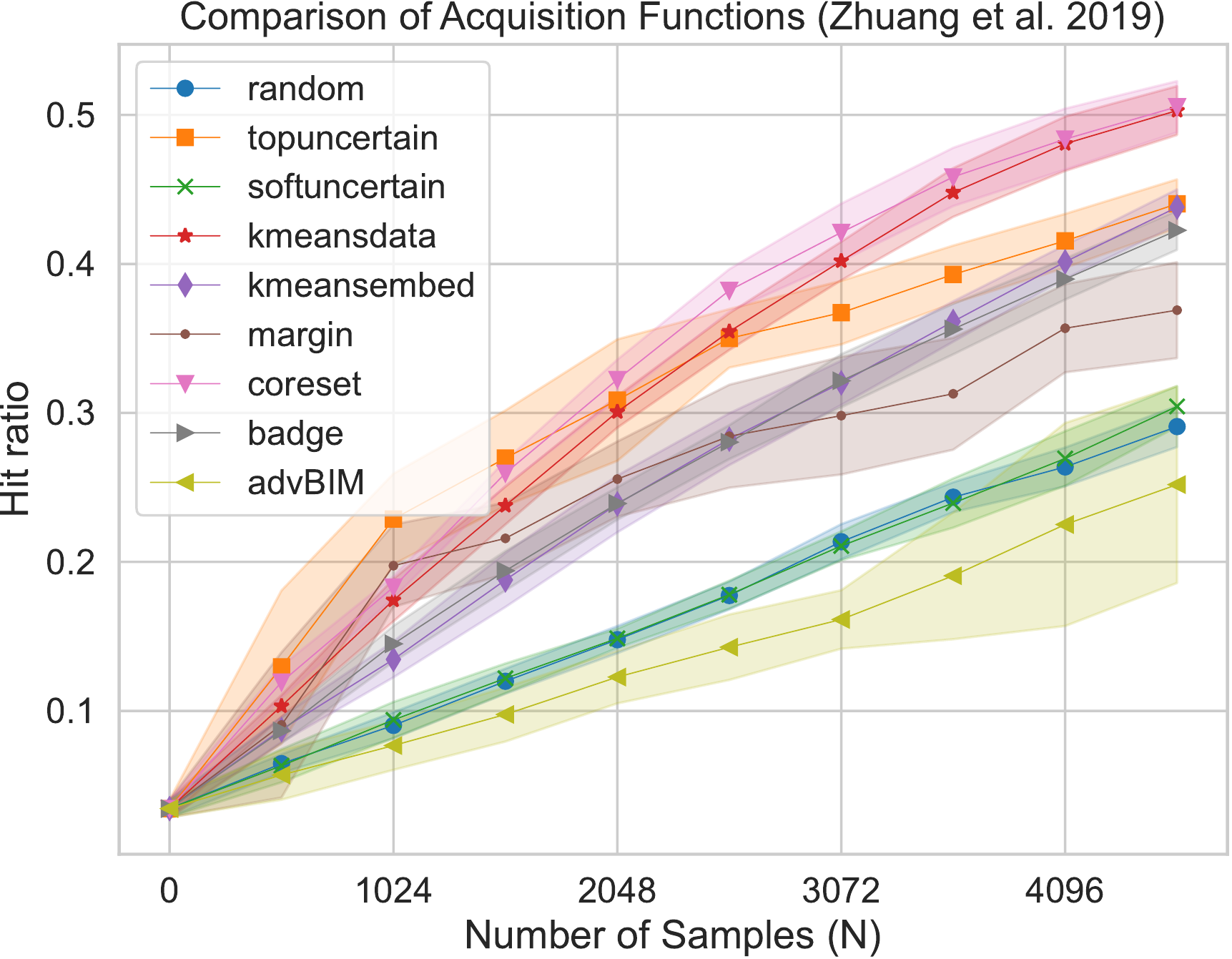}};
                        \end{tikzpicture}
                    }
                \end{subfigure}
                \&
                \\
            \\
           
            \\
            };
            \node [draw=none, rotate=90] at ([xshift=-8mm, yshift=2mm]fig-1-1.west) {\small batch size = 16};
            \node [draw=none, rotate=90] at ([xshift=-8mm, yshift=2mm]fig-2-1.west) {\small batch size = 32};
            \node [draw=none, rotate=90] at ([xshift=-8mm, yshift=2mm]fig-3-1.west) {\small batch size = 64};
            \node [draw=none, rotate=90] at ([xshift=-8mm, yshift=2mm]fig-4-1.west) {\small batch size = 128};
            \node [draw=none, rotate=90] at ([xshift=-8mm, yshift=2mm]fig-5-1.west) {\small batch size = 256};
            \node [draw=none, rotate=90] at ([xshift=-8mm, yshift=2mm]fig-6-1.west) {\small batch size = 512};
            \node [draw=none] at ([xshift=-6mm, yshift=3mm]fig-1-1.north) {\small Shifrut et al. 2018};
            \node [draw=none] at ([xshift=-9mm, yshift=3mm]fig-1-2.north) {\small Schmidt et al. 2021 (IFNg)};
            \node [draw=none] at ([xshift=-11mm, yshift=3mm]fig-1-3.north) {\small Schmidt et al. 2021 (IL-2)};
            \node [draw=none] at ([xshift=-13mm, yshift=2.5mm]fig-1-4.north) {\small Zhuang et al. 2019};
\end{tikzpicture}}
        \vspace{-7mm}
        \caption{The hit ratio of different acquisition for BNN model, different target datasets, and different acquisition batch sizes. We use {Achilles} treatment descriptors here. The x-axis shows the number of data points collected so far during the active learning cycles. The y-axis shows the ratio of the set of interesting genes that have been found by the acquisition function up until each cycle.}        
        \vspace{-5mm}
        \label{fig:hitratio_bnn_feat_achilles_alldatasets_allbathcsizes}
    \end{figure*}

\begin{figure*}
    \vspace{-2mm}
        \centering
        \makebox[0.72\paperwidth]{\begin{tikzpicture}[ampersand replacement=\&]
            \matrix (fig) [matrix of nodes]{ 
\begin{subfigure}{0.27\columnwidth}
                    \hspace{-17mm}
                    \centering
                    \resizebox{\linewidth}{!}{
                        \begin{tikzpicture}
                            \node (img)  {\includegraphics[width=\textwidth]{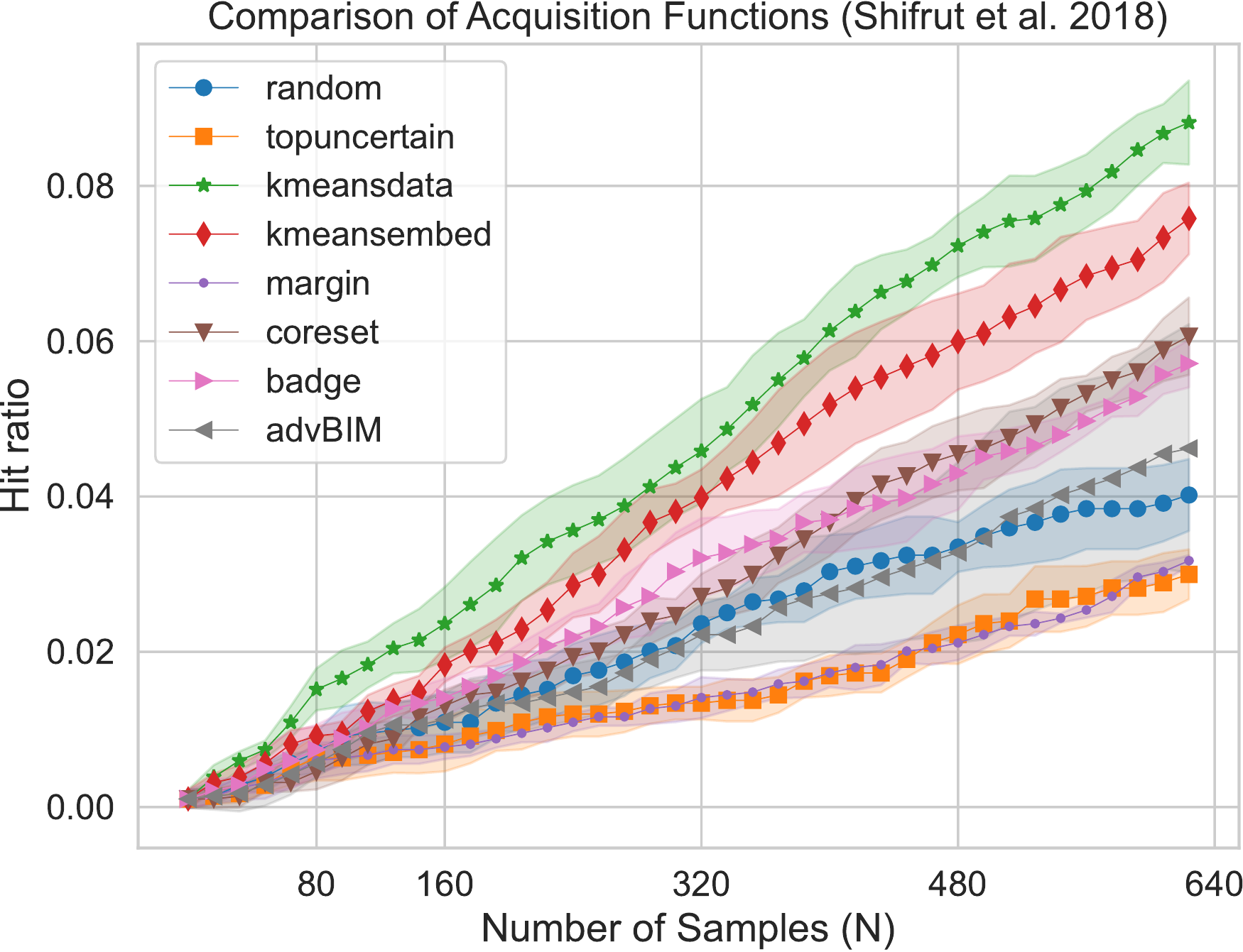}};
                        \end{tikzpicture}
                    }
                \end{subfigure}
                \&
                 \begin{subfigure}{0.27\columnwidth}
                    \hspace{-23mm}
                    \centering
                    \resizebox{\linewidth}{!}{
                        \begin{tikzpicture}
                            \node (img)  {\includegraphics[width=\textwidth]{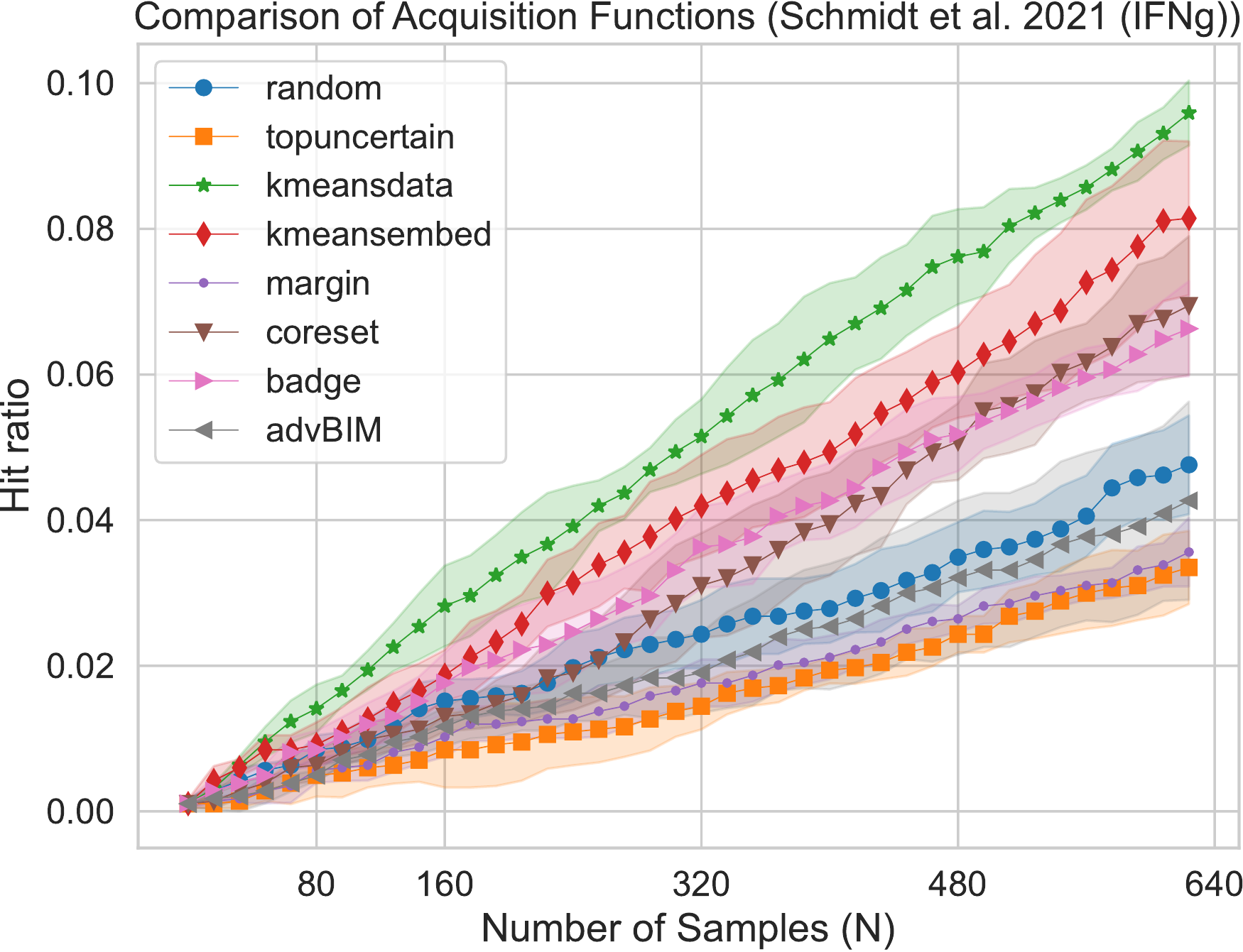}};
                        \end{tikzpicture}
                    }
                \end{subfigure}
                \&
                 \begin{subfigure}{0.27\columnwidth}
                    \hspace{-28mm}
                    \centering
                    \resizebox{\linewidth}{!}{
                        \begin{tikzpicture}
                            \node (img)  {\includegraphics[width=\textwidth]{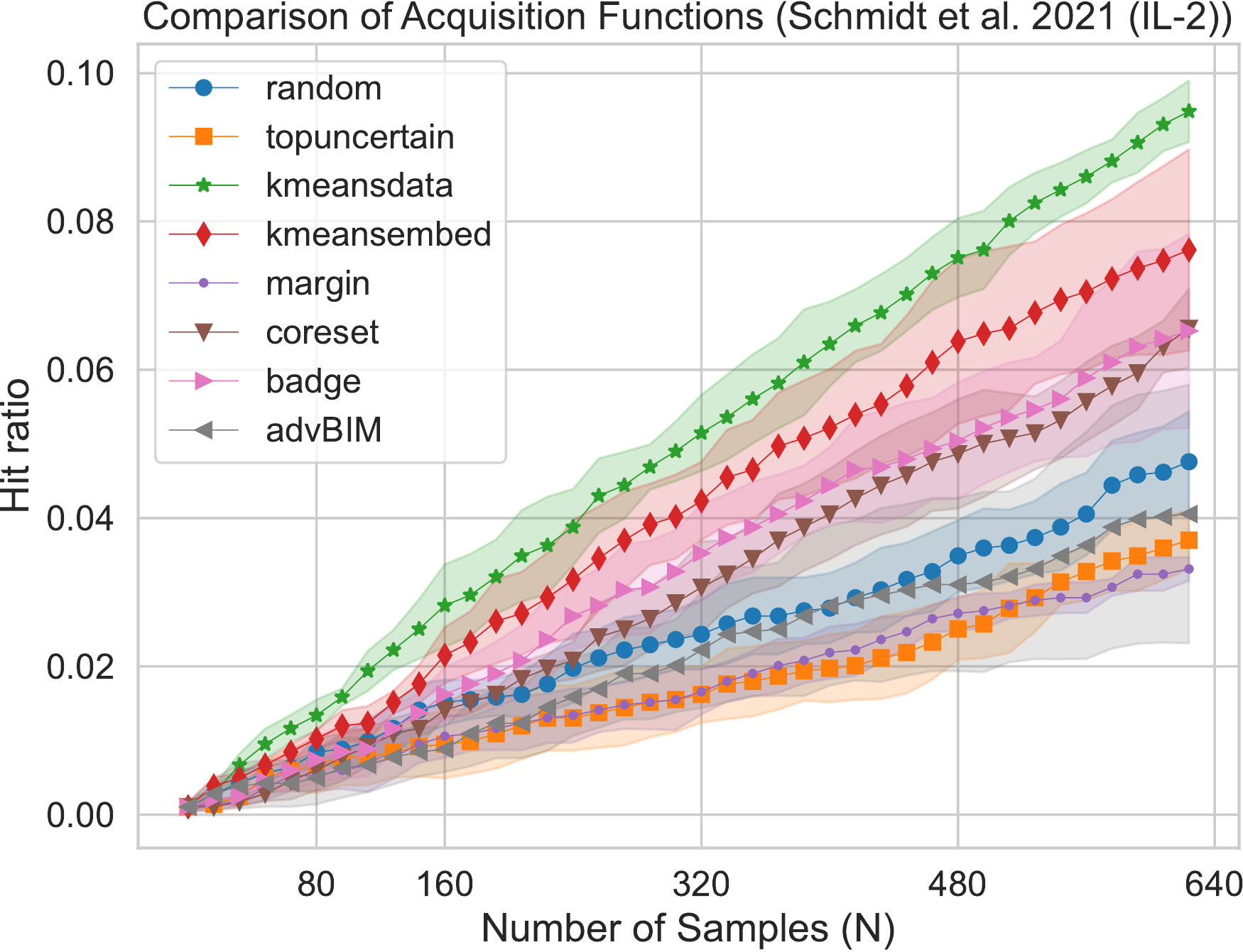}};
                        \end{tikzpicture}
                    }
                \end{subfigure}
                \&
                \begin{subfigure}{0.28\columnwidth}
                    \hspace{-32mm}
                    \centering
                    \resizebox{\linewidth}{!}{
                        \begin{tikzpicture}
                            \node (img)  {\includegraphics[width=\textwidth]{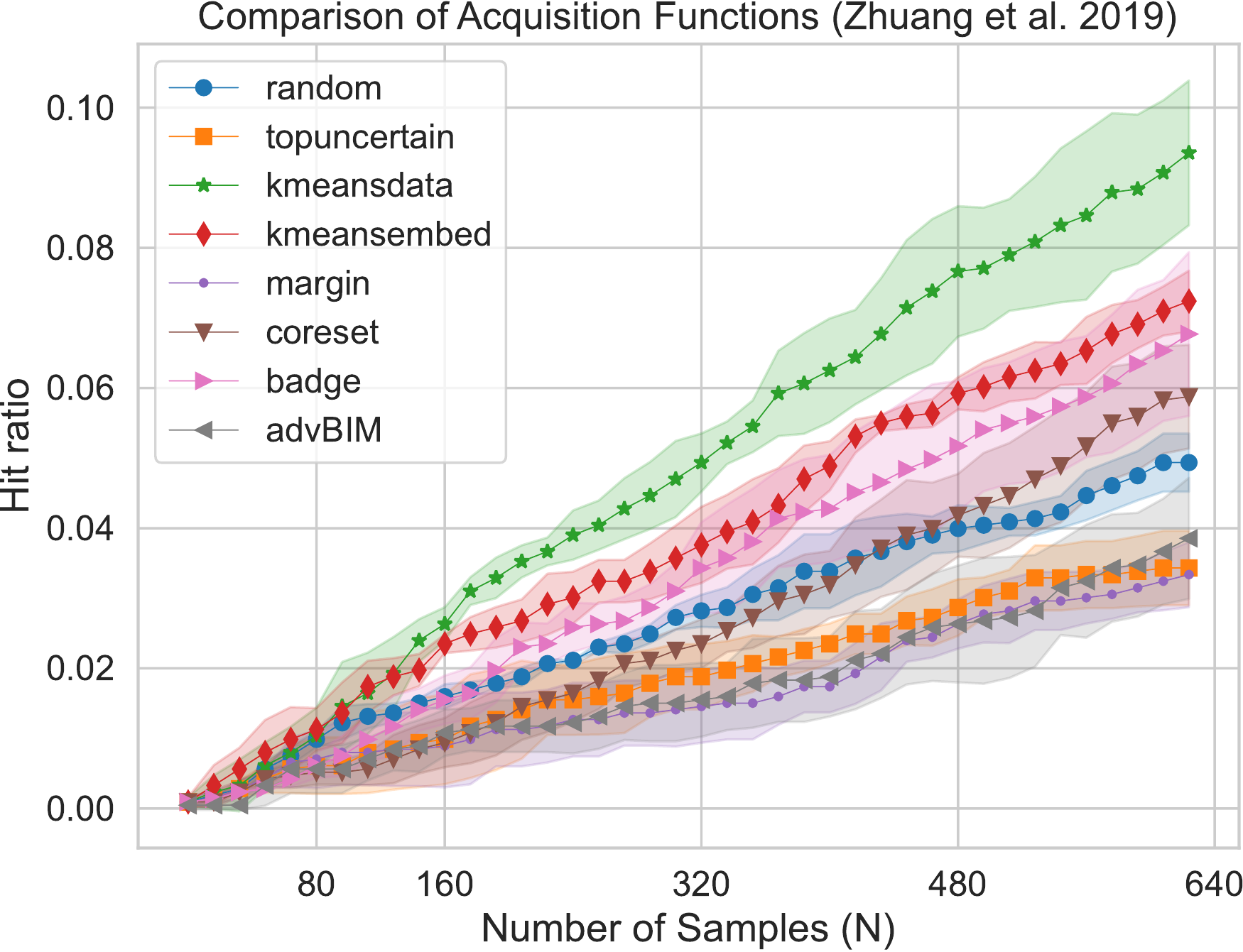}};
                        \end{tikzpicture}
                    }
                \end{subfigure}
                \&
            \\
\begin{subfigure}{0.27\columnwidth}
                    \hspace{-17mm}
                    \centering
                    \resizebox{\linewidth}{!}{
                        \begin{tikzpicture}
                            \node (img)  {\includegraphics[width=\textwidth]{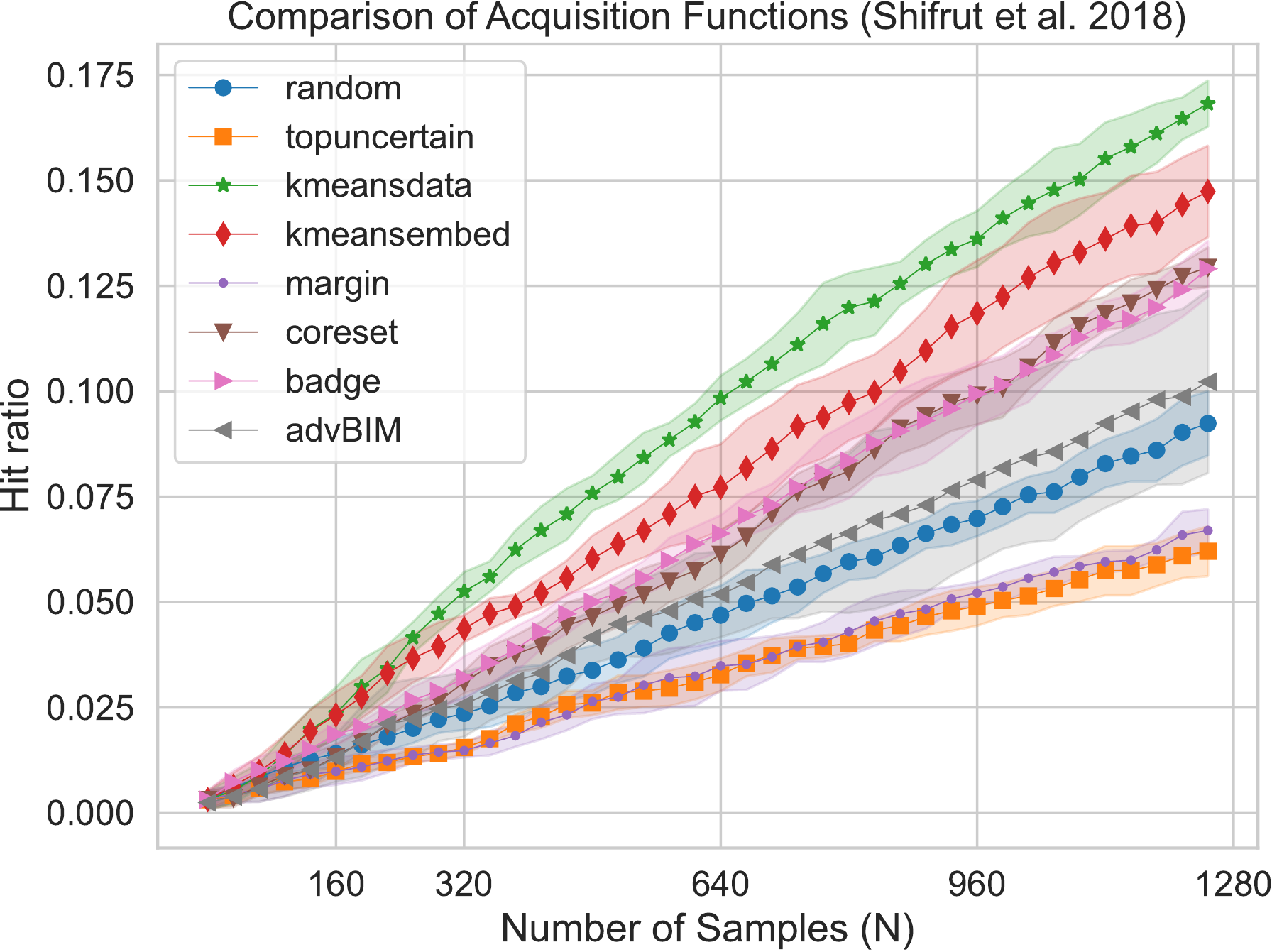}};
                        \end{tikzpicture}
                    }
                \end{subfigure}
                \&
                \begin{subfigure}{0.27\columnwidth}
                    \hspace{-23mm}
                    \centering
                    \resizebox{\linewidth}{!}{
                        \begin{tikzpicture}
                            \node (img)  {\includegraphics[width=\textwidth]{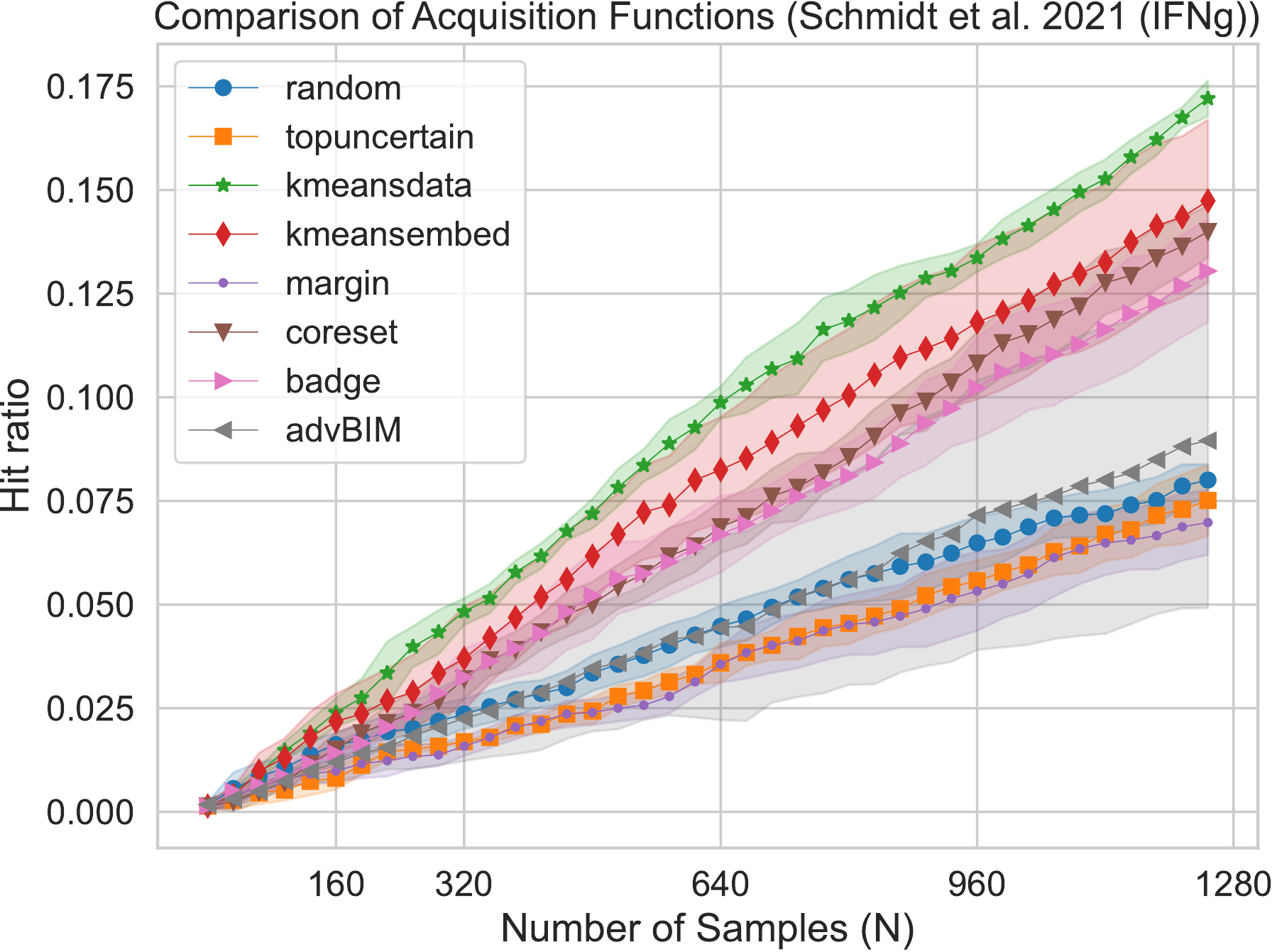}};
                        \end{tikzpicture}
                    }
                \end{subfigure}
                \&
                \begin{subfigure}{0.27\columnwidth}
                    \hspace{-28mm}
                    \centering
                    \resizebox{\linewidth}{!}{
                        \begin{tikzpicture}
                            \node (img)  {\includegraphics[width=\textwidth]{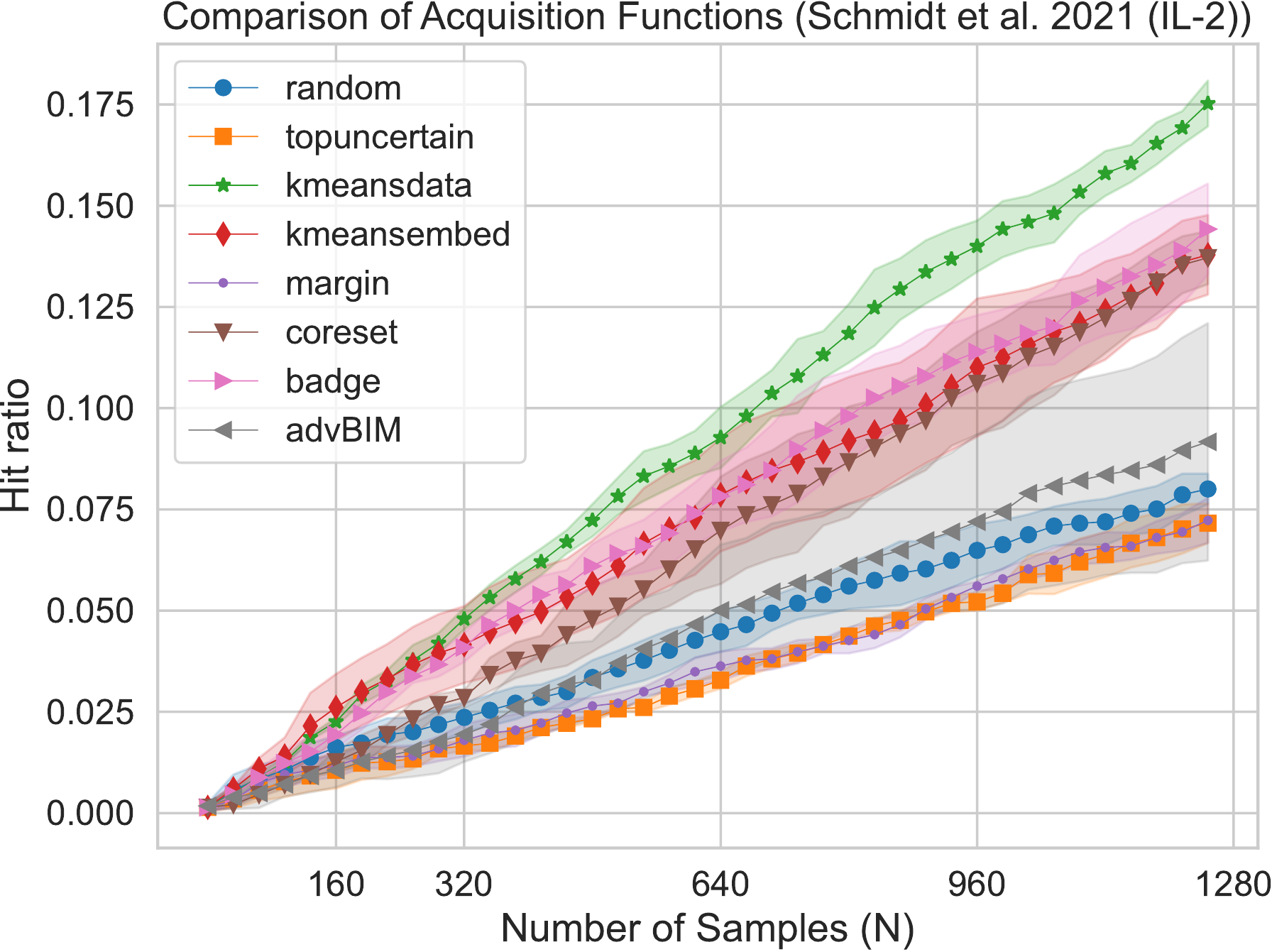}};
                        \end{tikzpicture}
                    }
                \end{subfigure}
                \&
                \begin{subfigure}{0.28\columnwidth}
                    \hspace{-32mm}
                    \centering
                    \resizebox{\linewidth}{!}{
                        \begin{tikzpicture}
                            \node (img)  {\includegraphics[width=\textwidth]{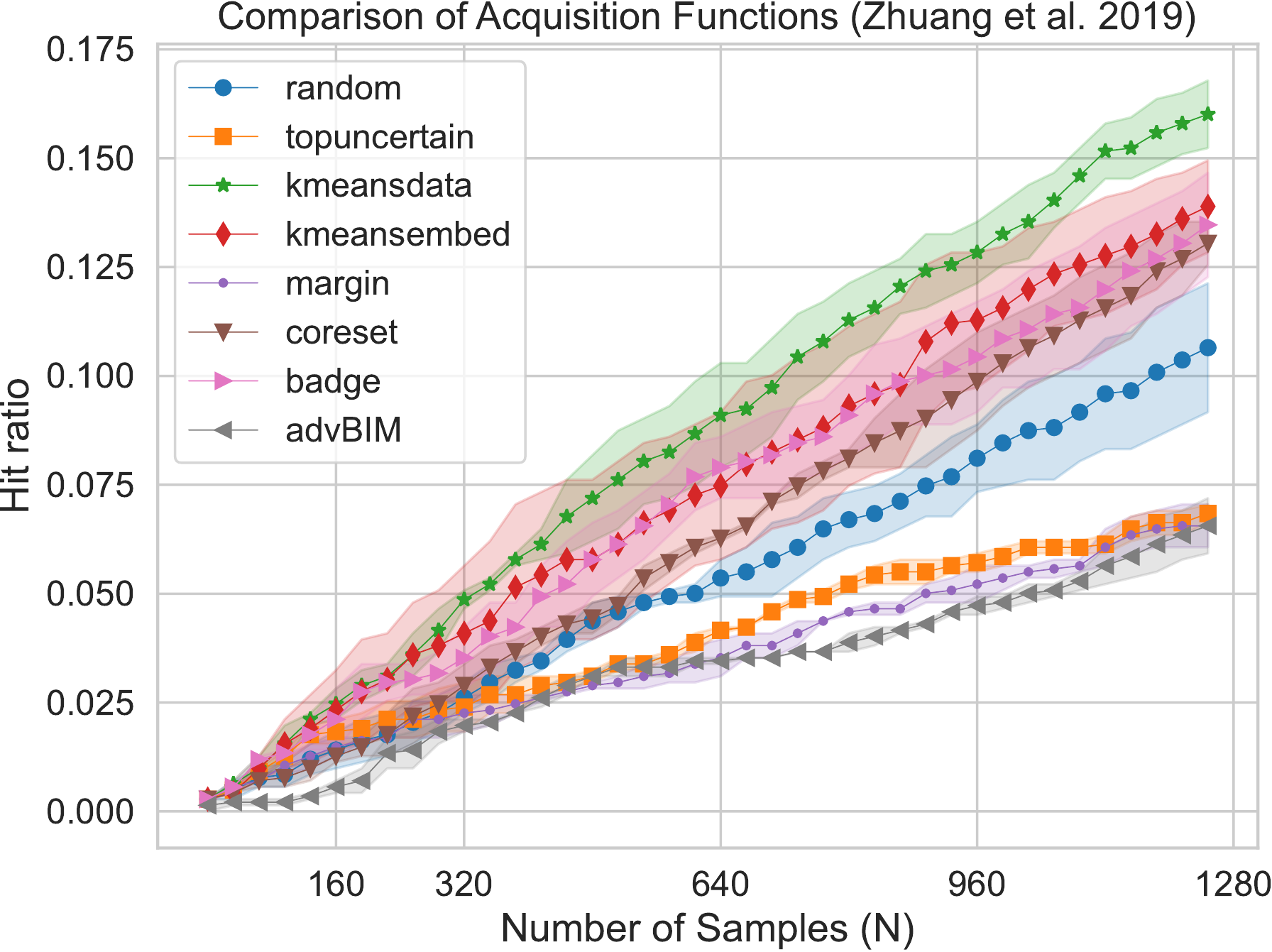}};
                        \end{tikzpicture}
                    }
                \end{subfigure}
                \&
                \\
\begin{subfigure}{0.27\columnwidth}
                    \hspace{-17mm}
                    \centering
                    \resizebox{\linewidth}{!}{
                        \begin{tikzpicture}
                            \node (img)  {\includegraphics[width=\textwidth]{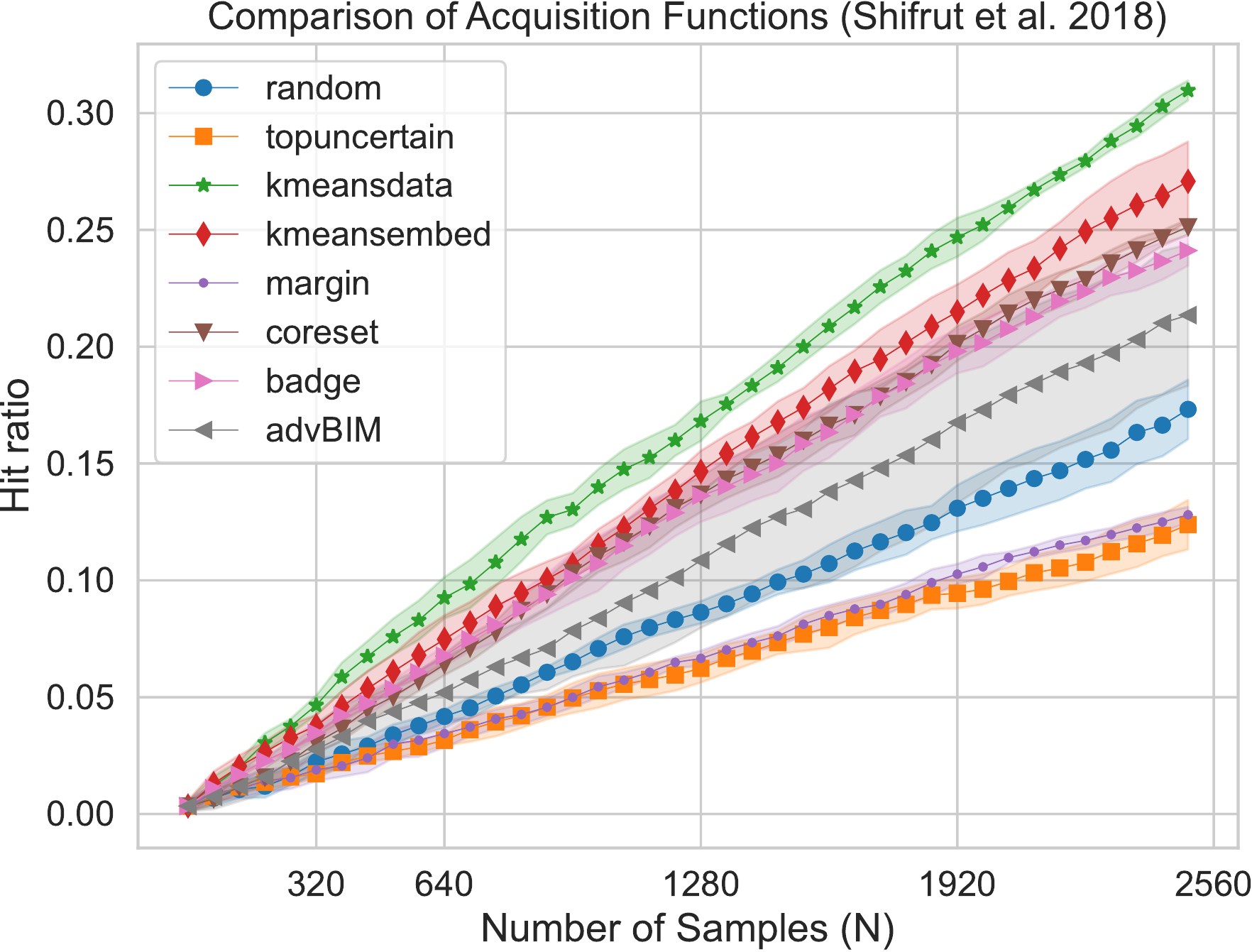}};
                        \end{tikzpicture}
                    }
                \end{subfigure}
                \&
                \begin{subfigure}{0.27\columnwidth}
                    \hspace{-23mm}
                    \centering
                    \resizebox{\linewidth}{!}{
                        \begin{tikzpicture}
                            \node (img)  {\includegraphics[width=\textwidth]{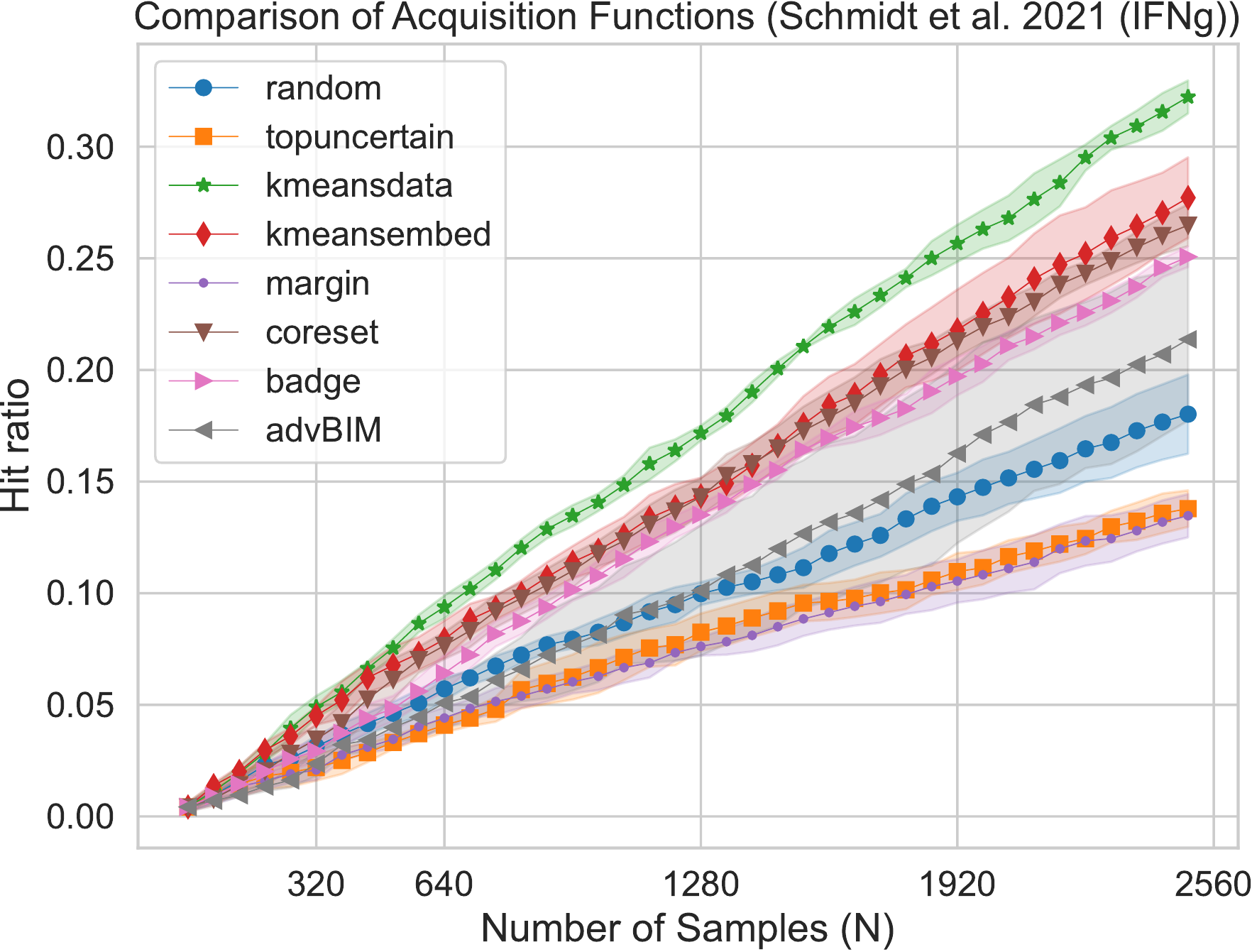}};
                        \end{tikzpicture}
                    }
                \end{subfigure}
                \&
                \begin{subfigure}{0.27\columnwidth}
                    \hspace{-28mm}
                    \centering
                    \resizebox{\linewidth}{!}{
                        \begin{tikzpicture}
                            \node (img)  {\includegraphics[width=\textwidth]{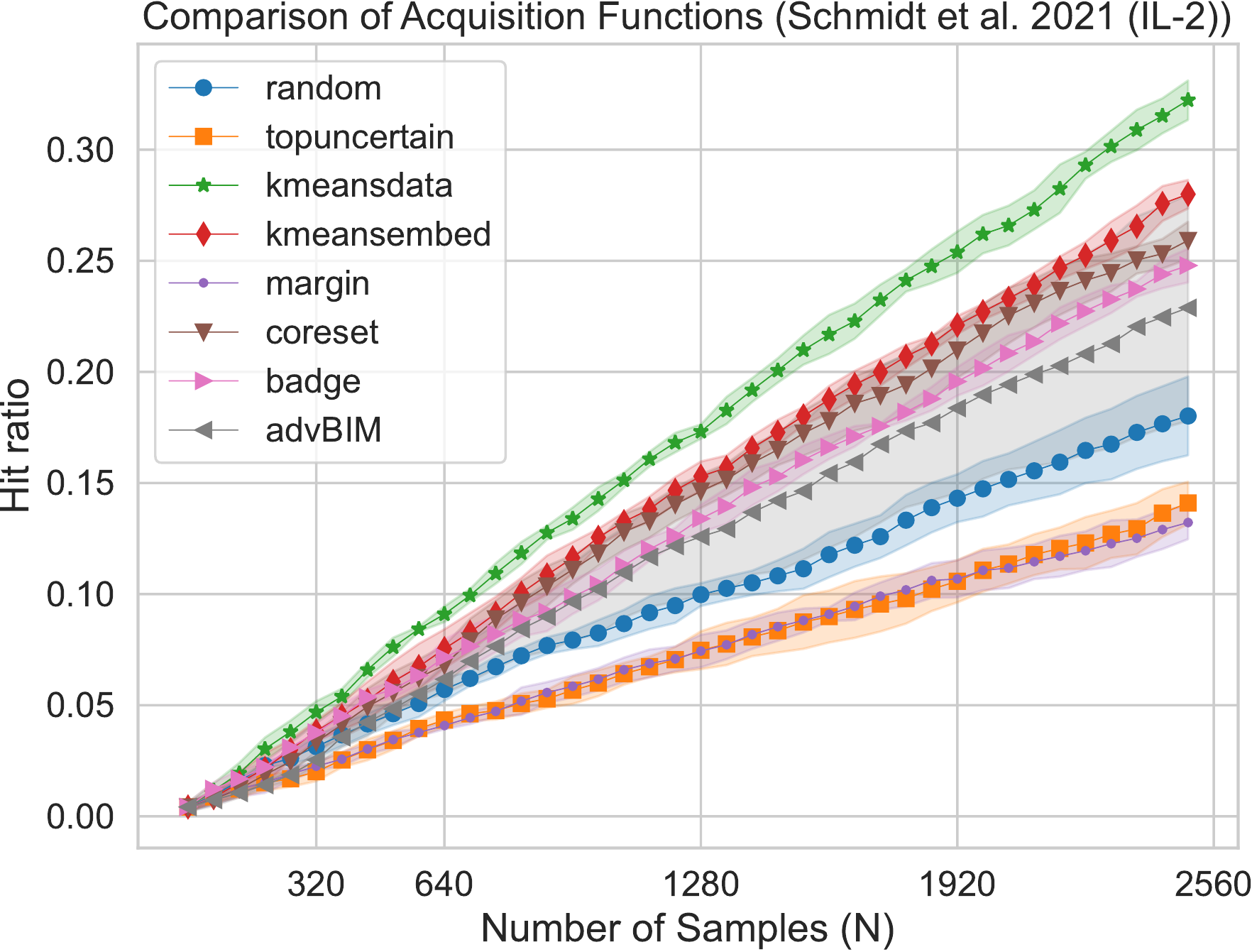}};
                        \end{tikzpicture}
                    }
                \end{subfigure}
                \&
                \begin{subfigure}{0.28\columnwidth}
                    \hspace{-32mm}
                    \centering
                    \resizebox{\linewidth}{!}{
                        \begin{tikzpicture}
                            \node (img)  {\includegraphics[width=\textwidth]{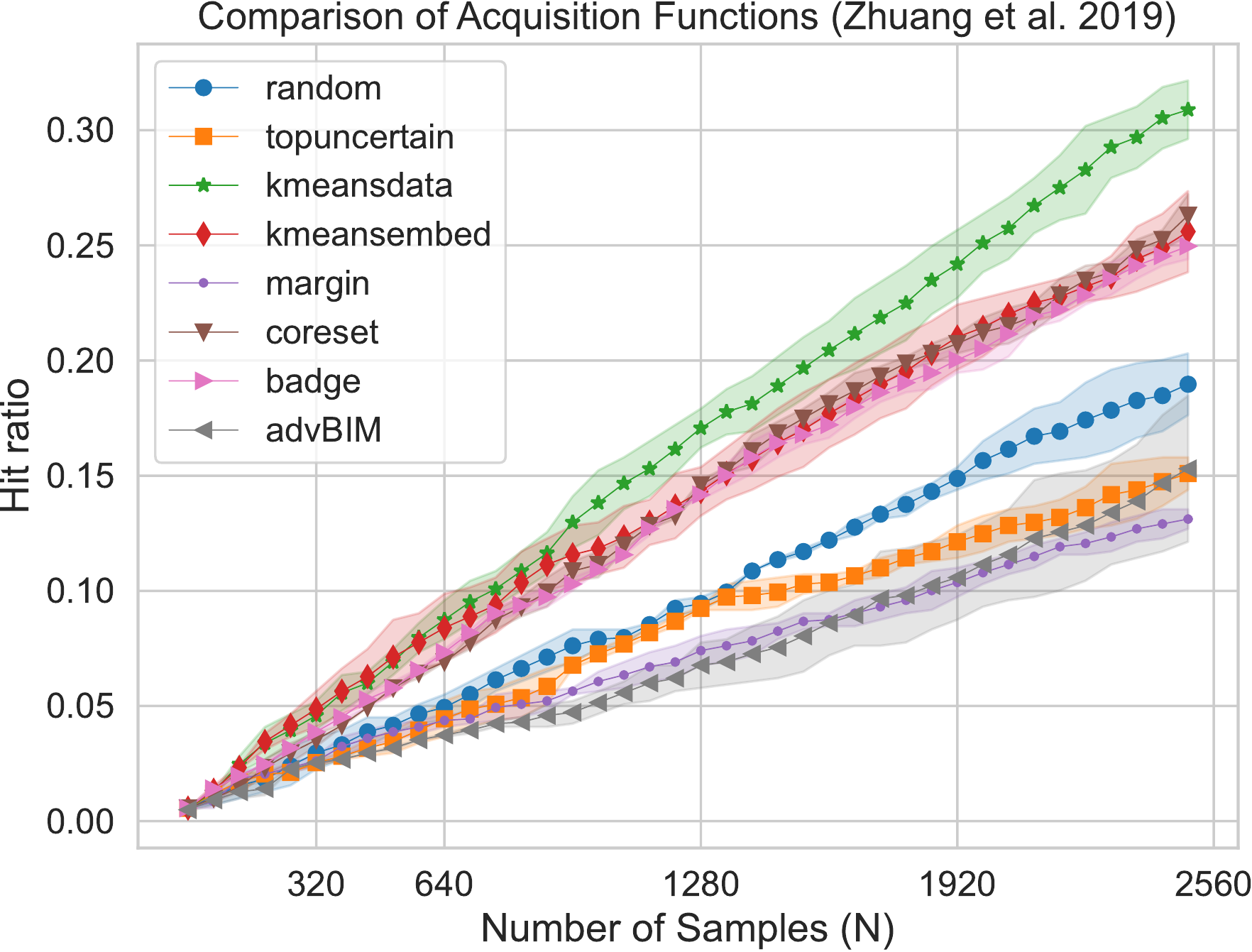}};
                        \end{tikzpicture}
                    }
                \end{subfigure}
                \&
                \\
\begin{subfigure}{0.27\columnwidth}
                    \hspace{-17mm}
                    \centering
                    \resizebox{\linewidth}{!}{
                        \begin{tikzpicture}
                            \node (img)  {\includegraphics[width=\textwidth]{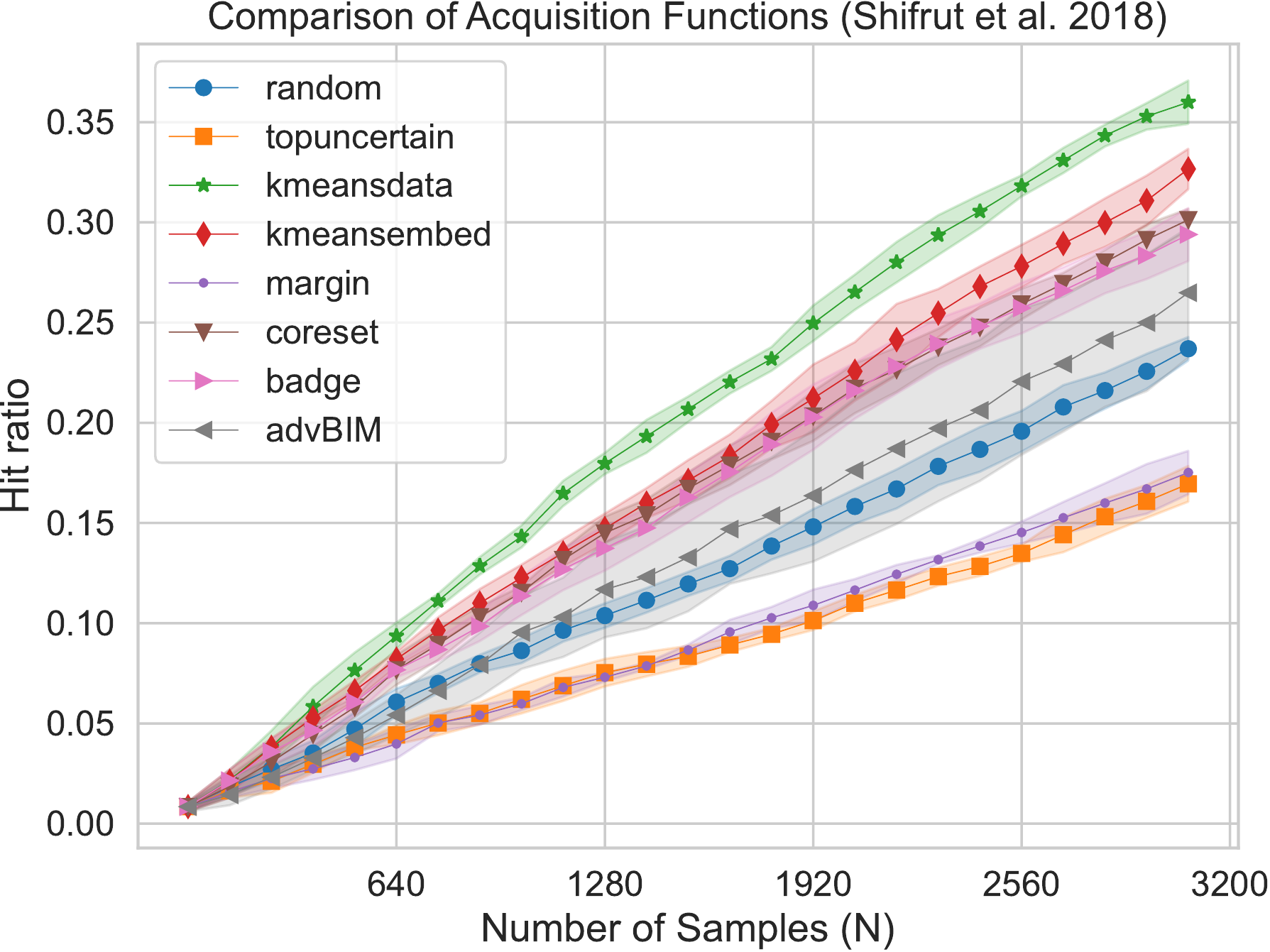}};
                        \end{tikzpicture}
                    }
                \end{subfigure}
                \&
                \begin{subfigure}{0.27\columnwidth}
                    \hspace{-23mm}
                    \centering
                    \resizebox{\linewidth}{!}{
                        \begin{tikzpicture}
                            \node (img)  {\includegraphics[width=\textwidth]{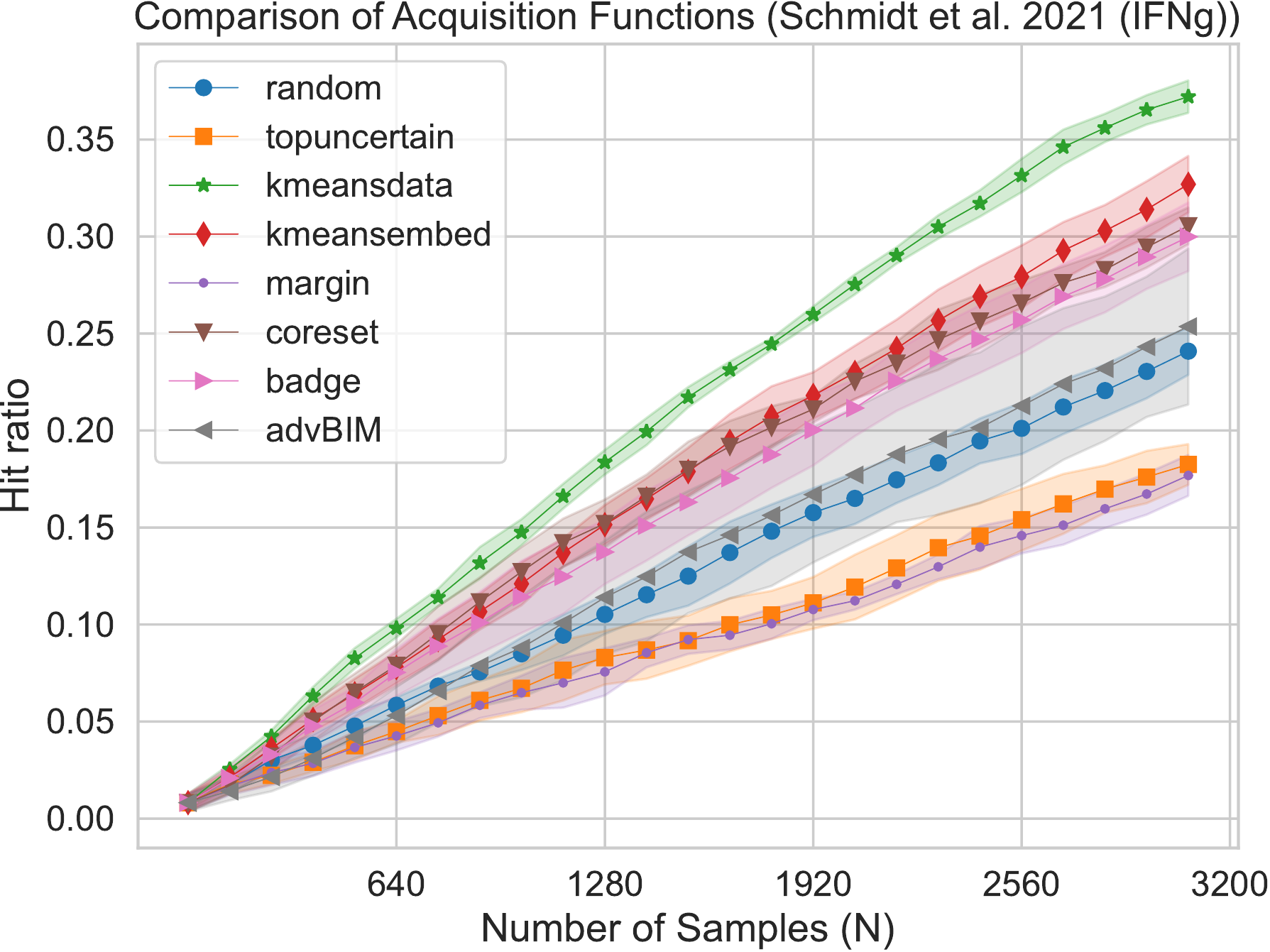}};
                        \end{tikzpicture}
                    }
                \end{subfigure}
                \&
                \begin{subfigure}{0.28\columnwidth}
                    \hspace{-28mm}
                    \centering
                    \resizebox{\linewidth}{!}{
                        \begin{tikzpicture}
                            \node (img)  {\includegraphics[width=\textwidth]{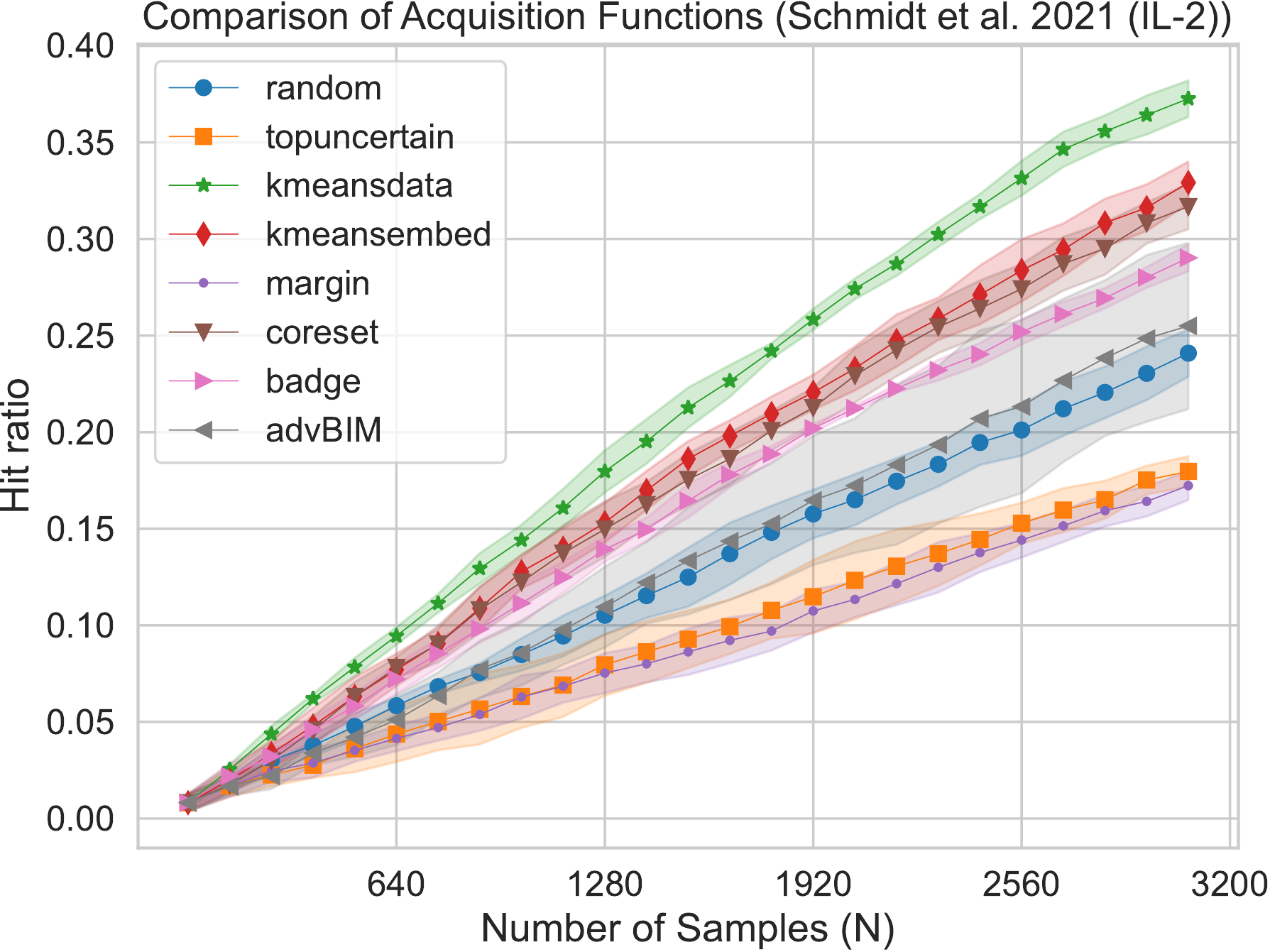}};
                        \end{tikzpicture}
                    }
                \end{subfigure}
                \&
                \begin{subfigure}{0.29\columnwidth}
                    \hspace{-32mm}
                    \centering
                    \resizebox{\linewidth}{!}{
                        \begin{tikzpicture}
                            \node (img)  {\includegraphics[width=\textwidth]{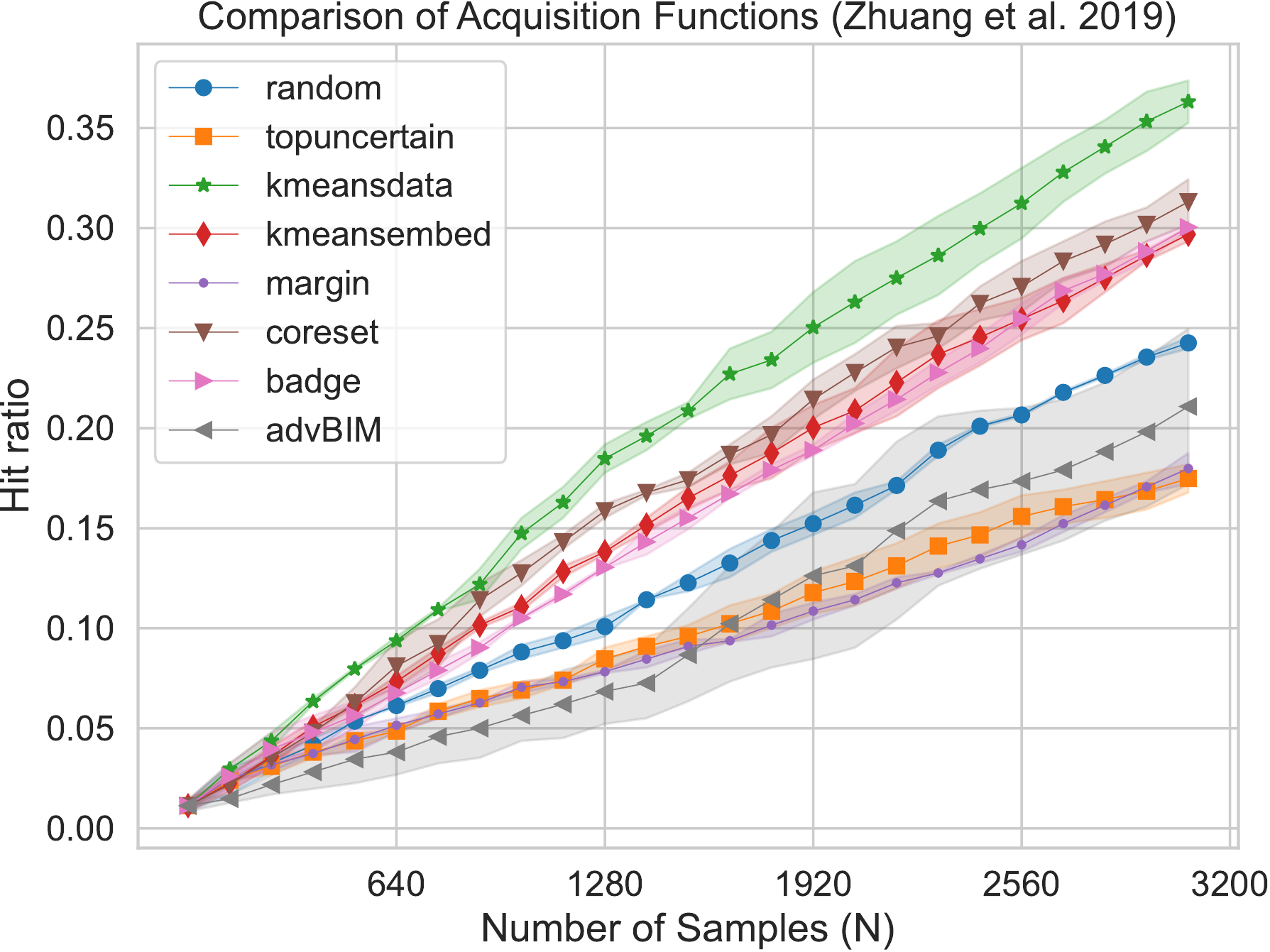}};
                        \end{tikzpicture}
                    }
                \end{subfigure}
                \&
                \\
\begin{subfigure}{0.275\columnwidth}
                    \hspace{-17mm}
                    \centering
                    \resizebox{\linewidth}{!}{
                        \begin{tikzpicture}
                            \node (img)  {\includegraphics[width=\textwidth]{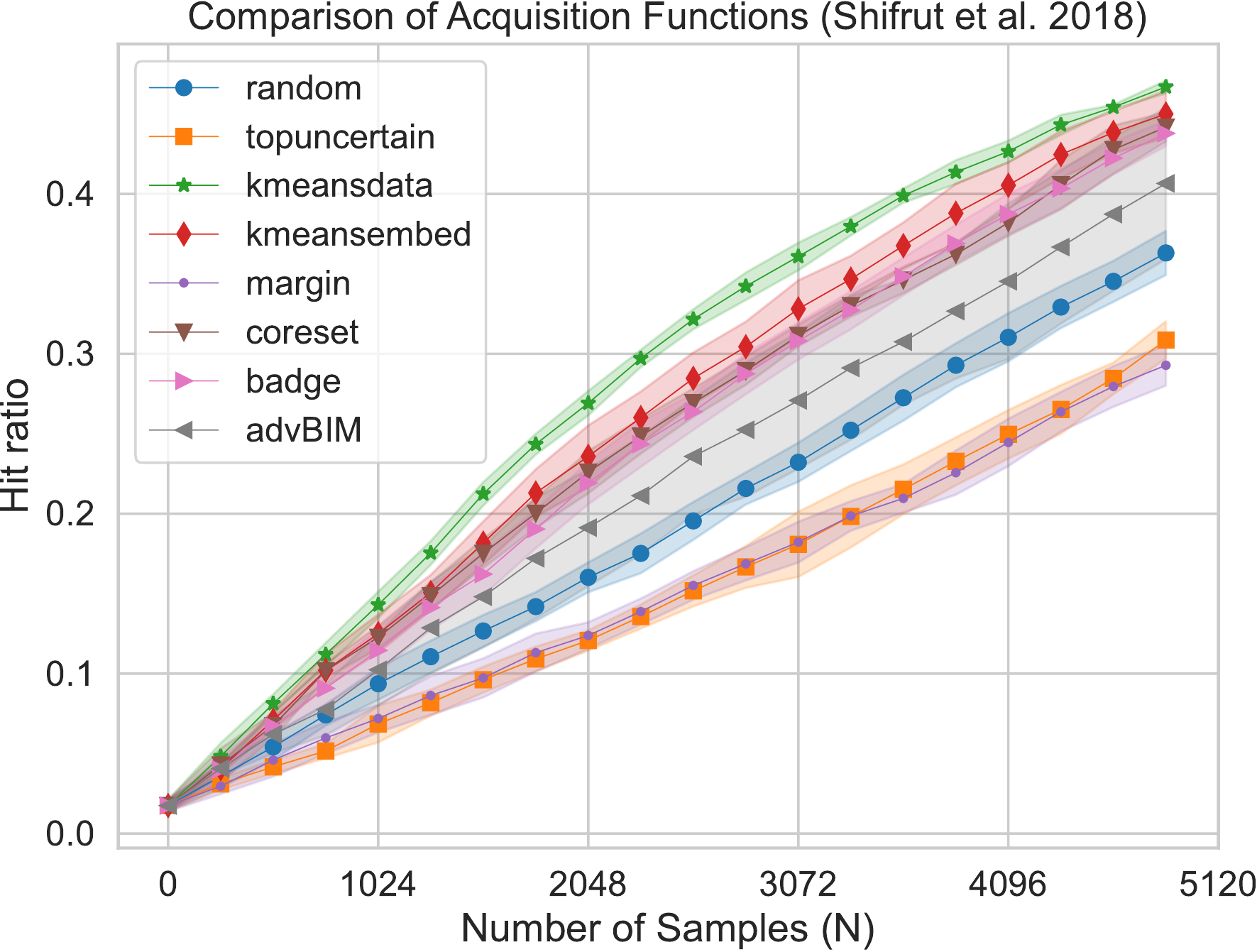}};
                        \end{tikzpicture}
                    }
                \end{subfigure}
                \&
                \begin{subfigure}{0.27\columnwidth}
                    \hspace{-23mm}
                    \centering
                    \resizebox{\linewidth}{!}{
                        \begin{tikzpicture}
                            \node (img)  {\includegraphics[width=\textwidth]{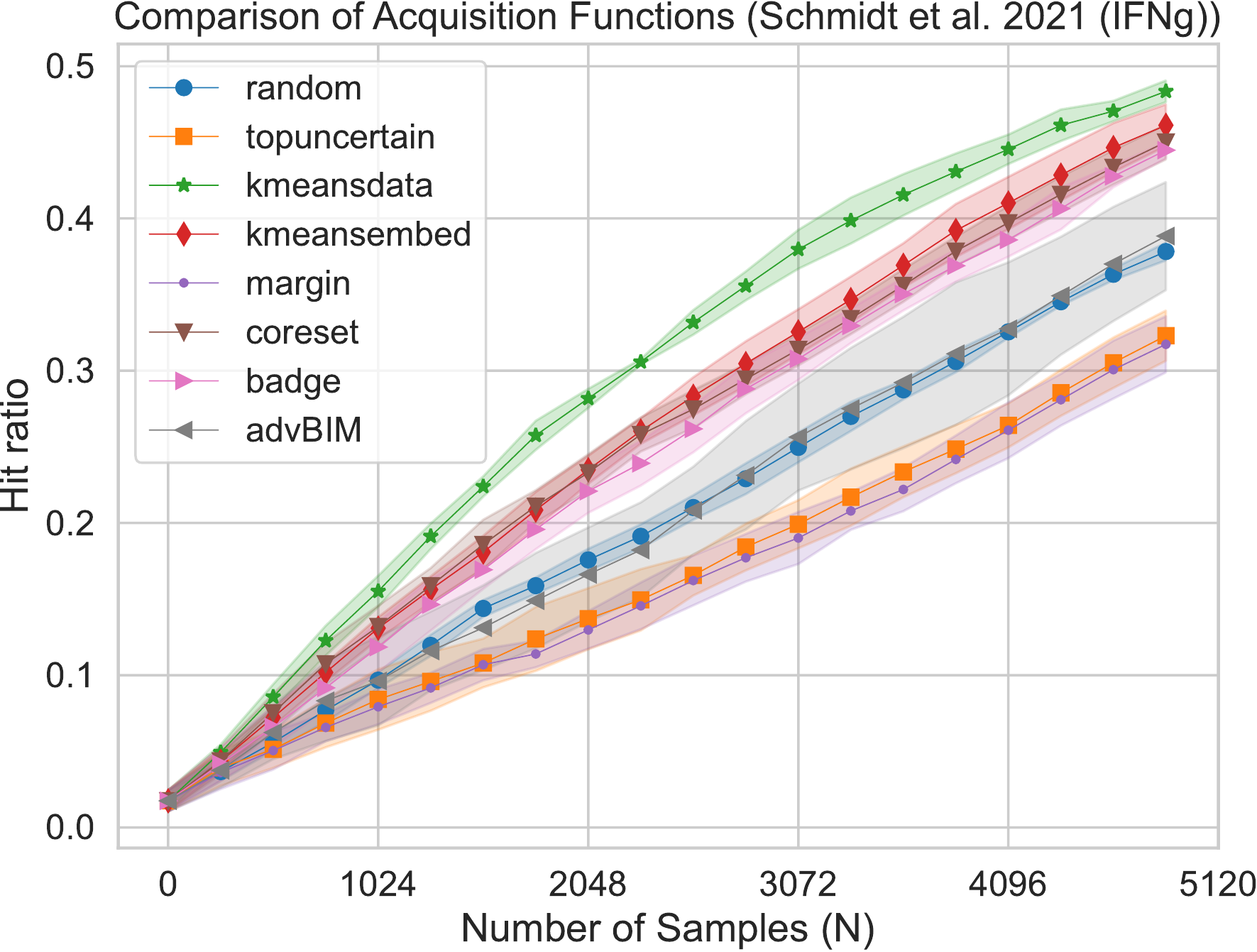}};
                        \end{tikzpicture}
                    }
                \end{subfigure}
                \&
                \begin{subfigure}{0.27\columnwidth}
                    \hspace{-28mm}
                    \centering
                    \resizebox{\linewidth}{!}{
                        \begin{tikzpicture}
                            \node (img)  {\includegraphics[width=\textwidth]{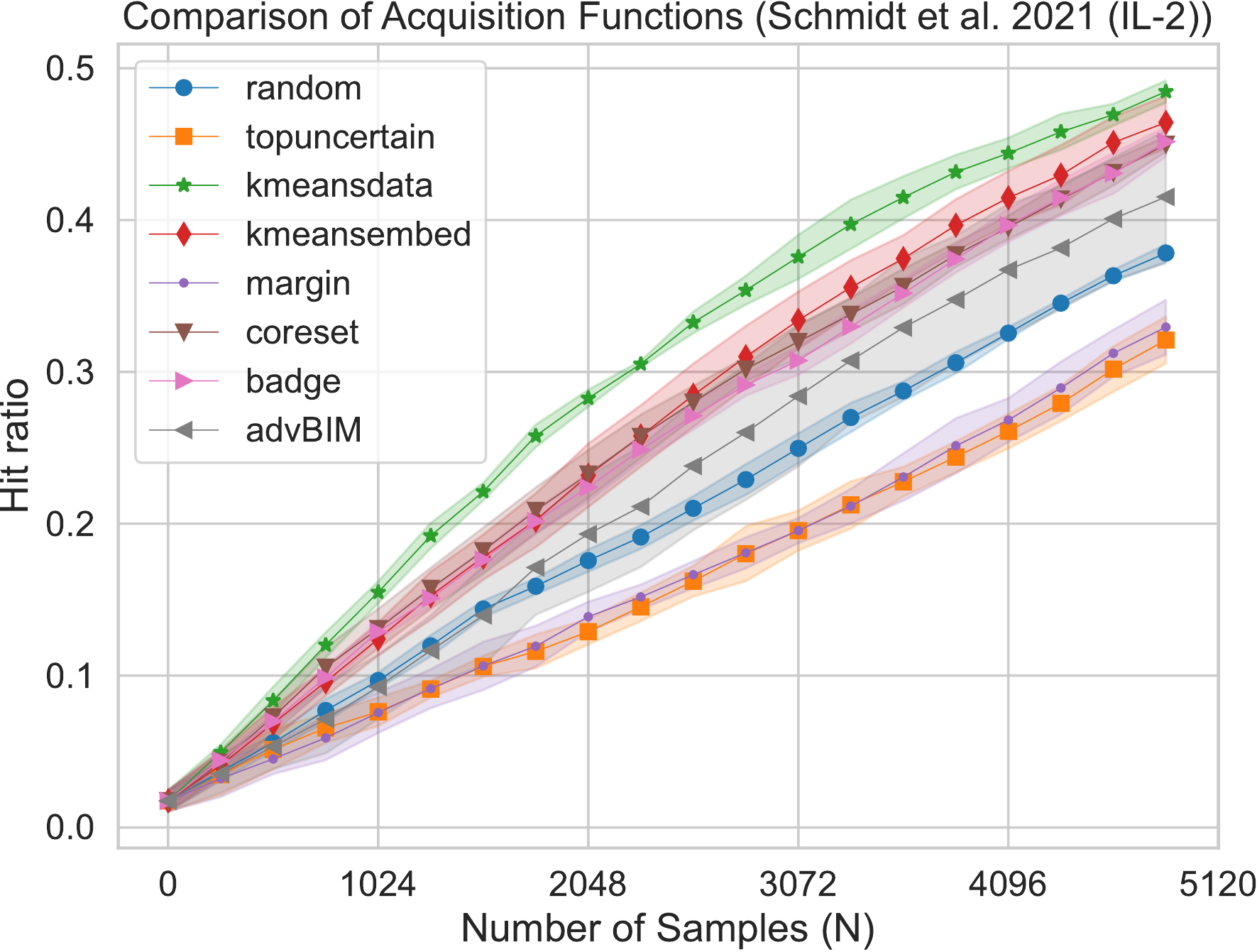}};
                        \end{tikzpicture}
                    }
                \end{subfigure}
                \&
                \begin{subfigure}{0.29\columnwidth}
                    \hspace{-32mm}
                    \centering
                    \resizebox{\linewidth}{!}{
                        \begin{tikzpicture}
                            \node (img)  {\includegraphics[width=\textwidth]{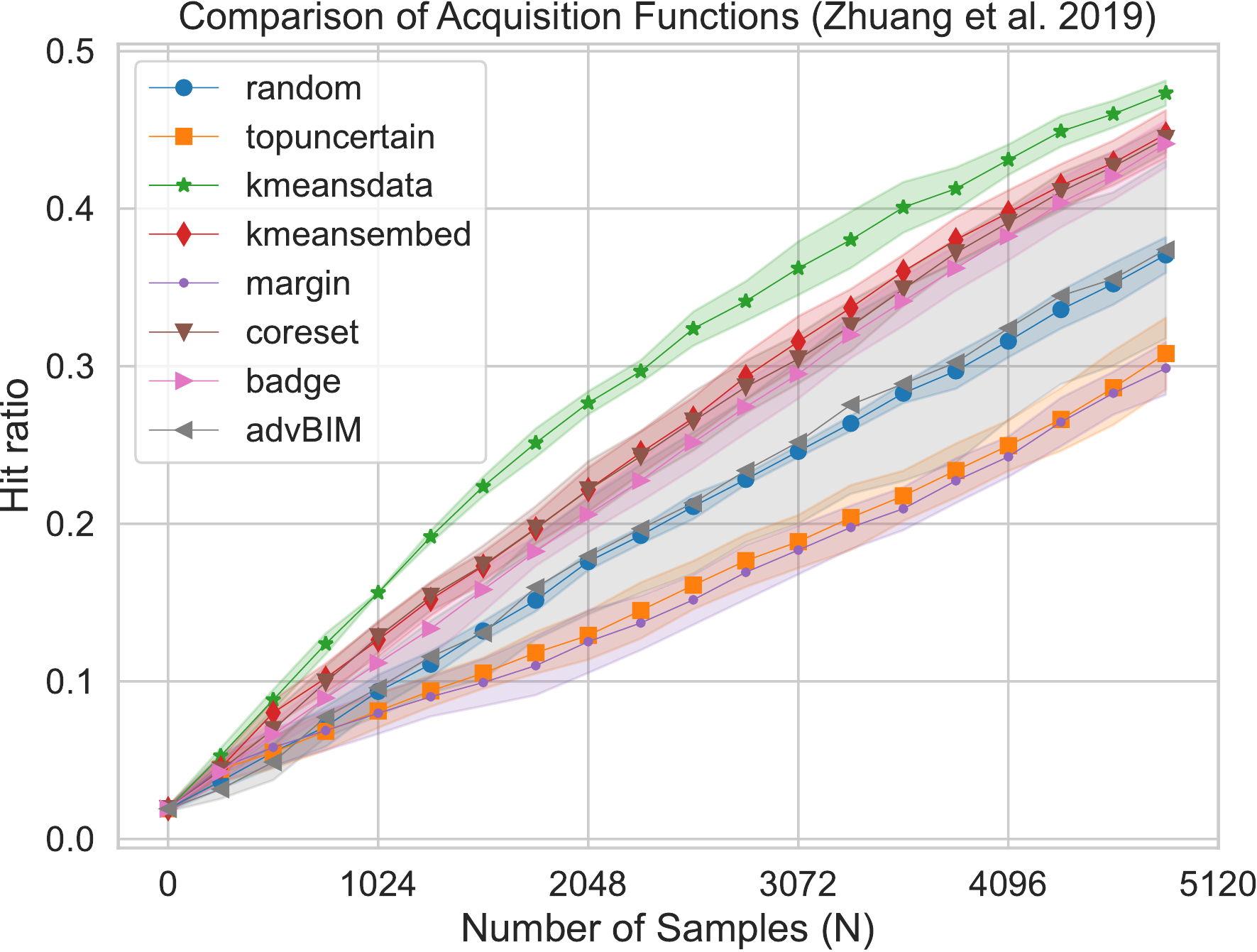}};
                        \end{tikzpicture}
                    }
                \end{subfigure}
                \&
                \\
\begin{subfigure}{0.28\columnwidth}
                    \hspace{-17mm}
                    \centering
                    \resizebox{\linewidth}{!}{
                        \begin{tikzpicture}
                            \node (img)  {\includegraphics[width=\textwidth]{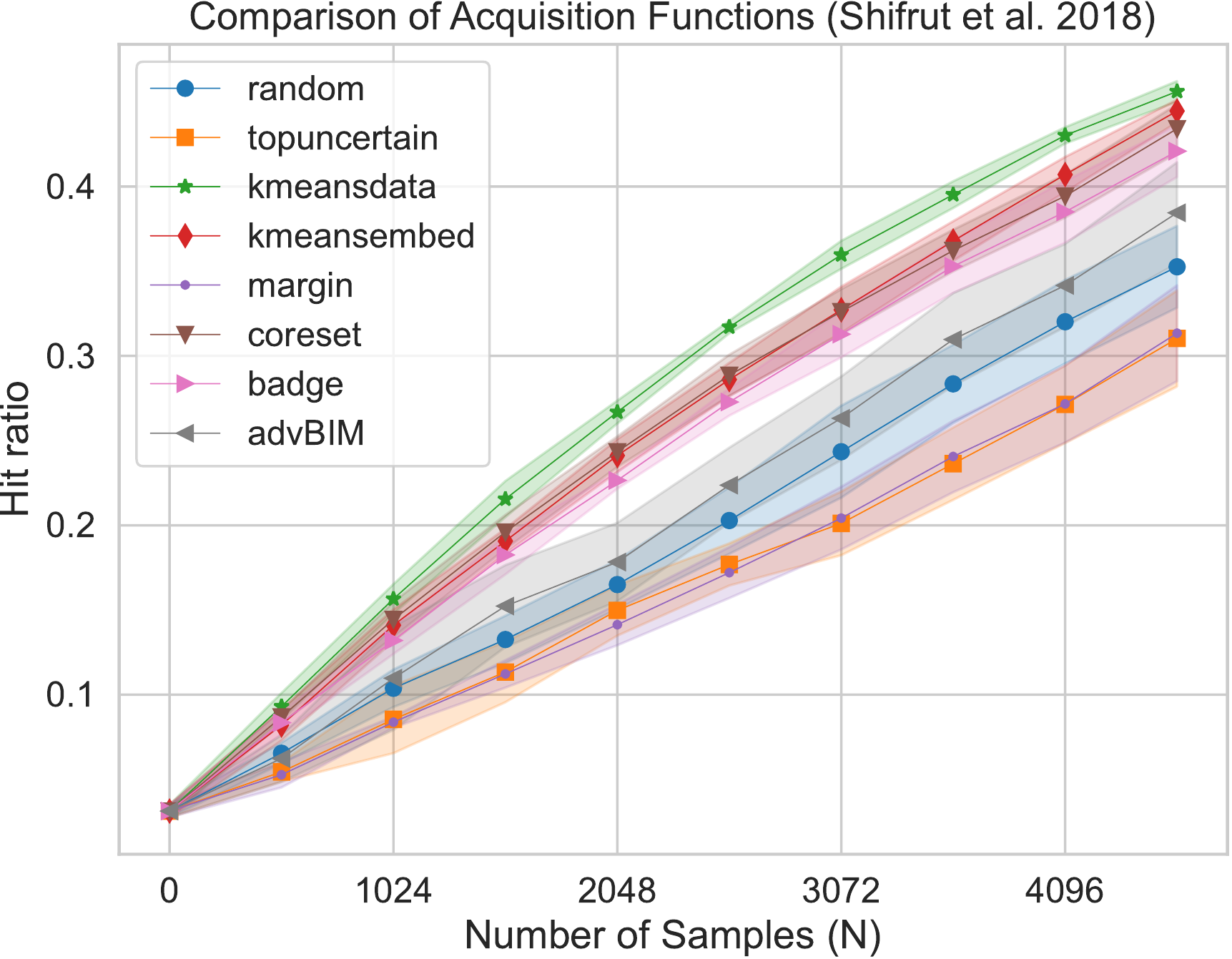}};
                        \end{tikzpicture}
                    }
                \end{subfigure}
                \&
                \begin{subfigure}{0.27\columnwidth}
                    \hspace{-23mm}
                    \centering
                    \resizebox{\linewidth}{!}{
                        \begin{tikzpicture}
                            \node (img)  {\includegraphics[width=\textwidth]{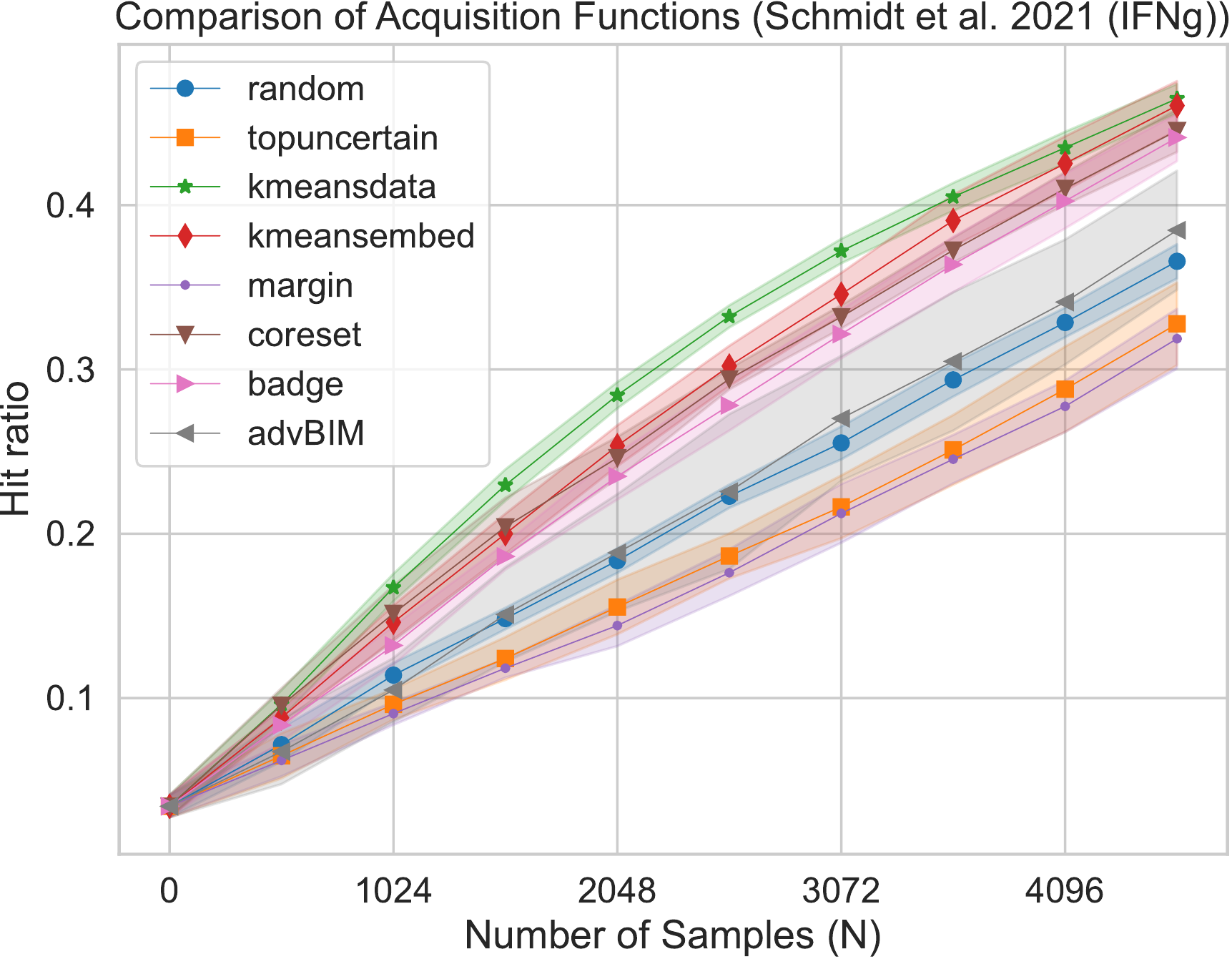}};
                        \end{tikzpicture}
                    }
                \end{subfigure}
                \&
                \begin{subfigure}{0.27\columnwidth}
                    \hspace{-28mm}
                    \centering
                    \resizebox{\linewidth}{!}{
                        \begin{tikzpicture}
                            \node (img)  {\includegraphics[width=\textwidth]{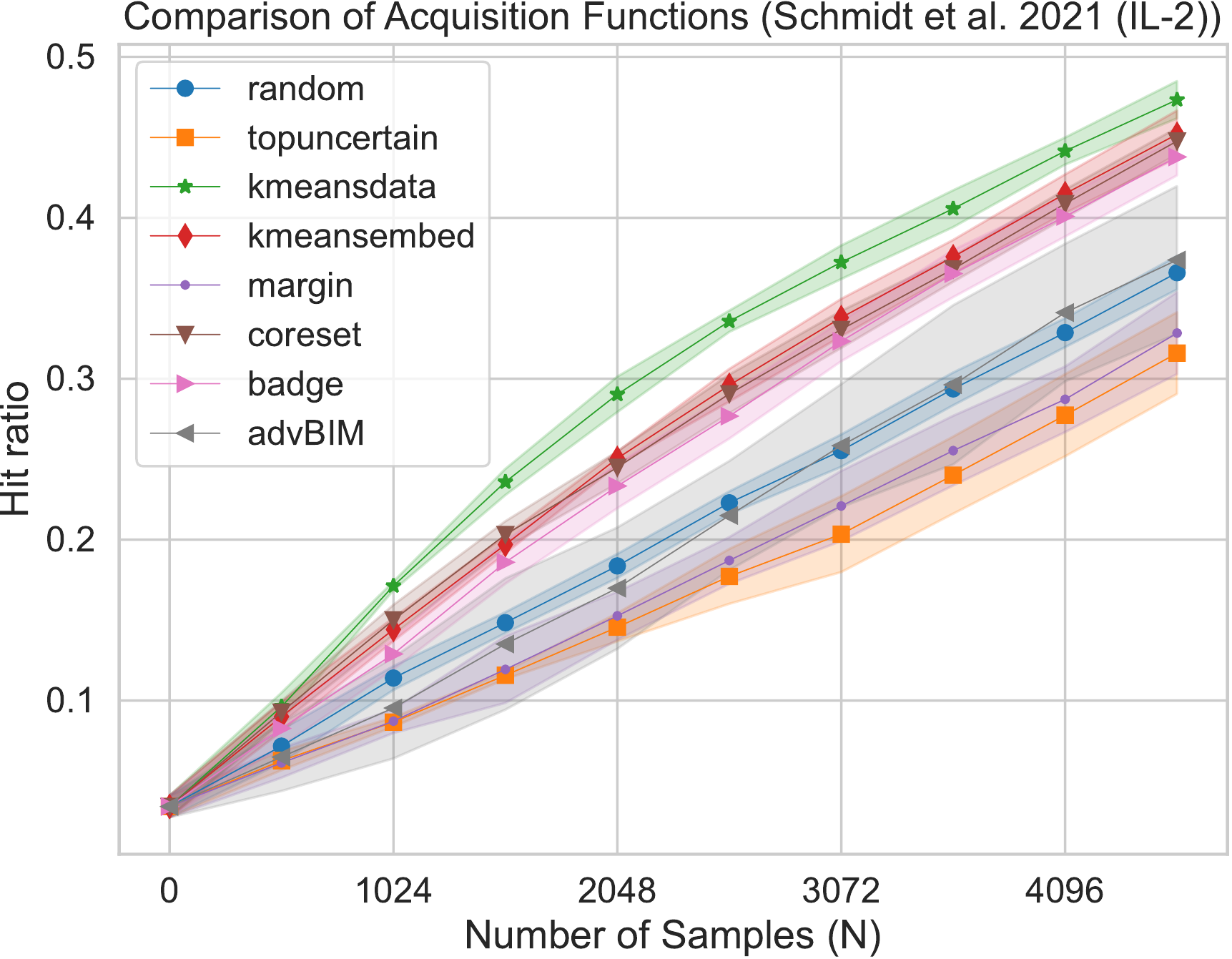}};
                        \end{tikzpicture}
                    }
                \end{subfigure}
                \&
                \begin{subfigure}{0.28\columnwidth}
                    \hspace{-32mm}
                    \centering
                    \resizebox{\linewidth}{!}{
                        \begin{tikzpicture}
                            \node (img)  {\includegraphics[width=\textwidth]{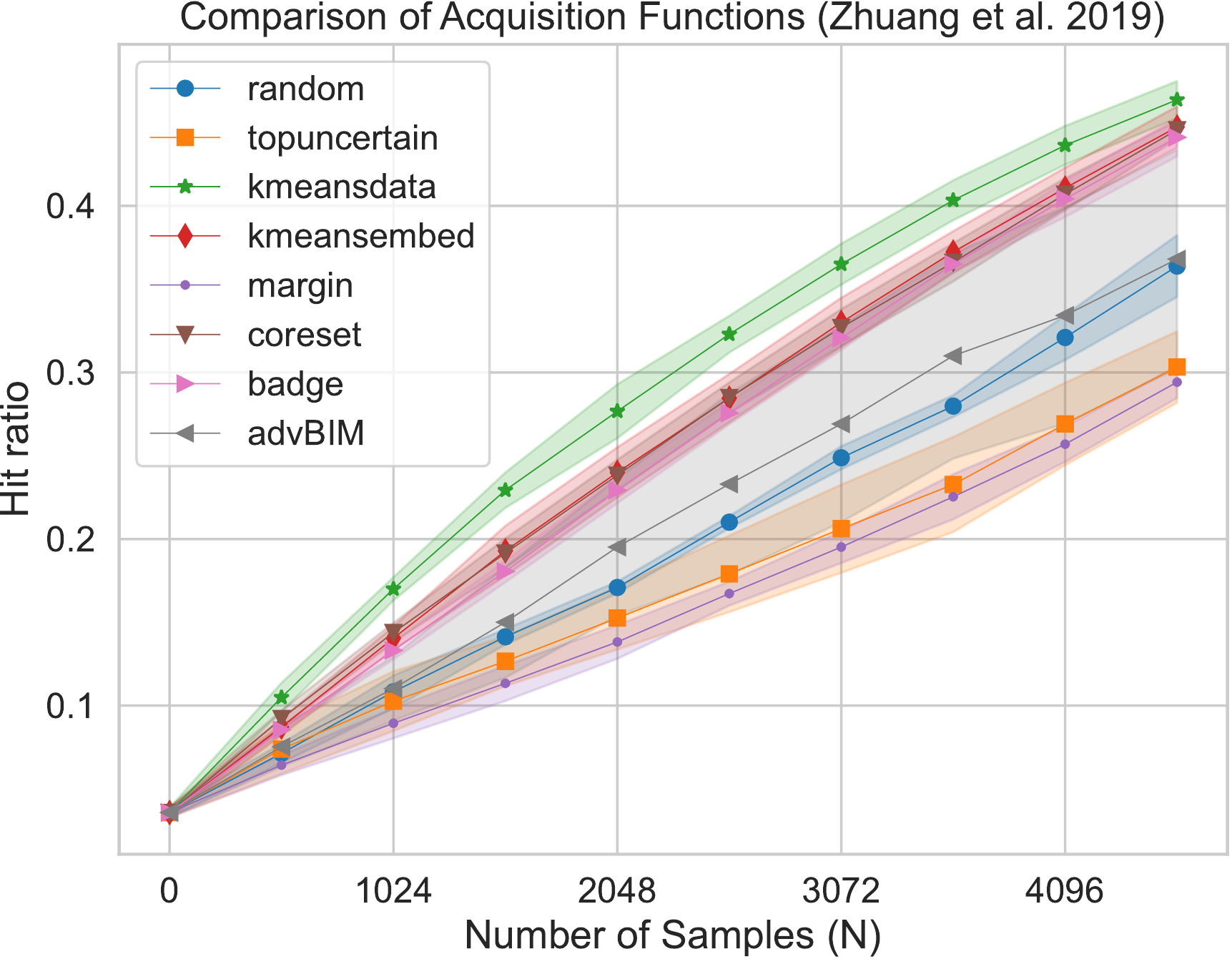}};
                        \end{tikzpicture}
                    }
                \end{subfigure}
                \&
                \\
            \\
           
            \\
            };
            \node [draw=none, rotate=90] at ([xshift=-8mm, yshift=2mm]fig-1-1.west) {\small batch size = 16};
            \node [draw=none, rotate=90] at ([xshift=-8mm, yshift=2mm]fig-2-1.west) {\small batch size = 32};
            \node [draw=none, rotate=90] at ([xshift=-8mm, yshift=2mm]fig-3-1.west) {\small batch size = 64};
            \node [draw=none, rotate=90] at ([xshift=-8mm, yshift=2mm]fig-4-1.west) {\small batch size = 128};
            \node [draw=none, rotate=90] at ([xshift=-8mm, yshift=2mm]fig-5-1.west) {\small batch size = 256};
            \node [draw=none, rotate=90] at ([xshift=-8mm, yshift=2mm]fig-6-1.west) {\small batch size = 512};
            \node [draw=none] at ([xshift=-6mm, yshift=3mm]fig-1-1.north) {\small Shifrut et al. 2018};
            \node [draw=none] at ([xshift=-9mm, yshift=3mm]fig-1-2.north) {\small Schmidt et al. 2021 (IFNg)};
            \node [draw=none] at ([xshift=-11mm, yshift=3mm]fig-1-3.north) {\small Schmidt et al. 2021 (IL-2)};
            \node [draw=none] at ([xshift=-13mm, yshift=2.5mm]fig-1-4.north) {\small Zhuang et al. 2019};
\end{tikzpicture}}
        \vspace{-7mm}
        \caption{The hit ratio of different acquisition for BNN model, different target datasets, and different acquisition batch sizes. We use {CCLE} treatment descriptors here. The x-axis shows the number of data points collected so far during the active learning cycles. The y-axis shows the ratio of the set of interesting genes that have been found by the acquisition function up until each cycle.}        
        \vspace{-5mm}
        \label{fig:hitratio_bnn_feat_ccle_alldatasets_allbathcsizes}
    \end{figure*}

\begin{figure*}
    \vspace{-2mm}
        \centering
        \makebox[0.72\paperwidth]{\begin{tikzpicture}[ampersand replacement=\&]
            \matrix (fig) [matrix of nodes]{ 
\begin{subfigure}{0.27\columnwidth}
                    \hspace{-17mm}
                    \centering
                    \resizebox{\linewidth}{!}{
                        \begin{tikzpicture}
                            \node (img)  {\includegraphics[width=\textwidth]{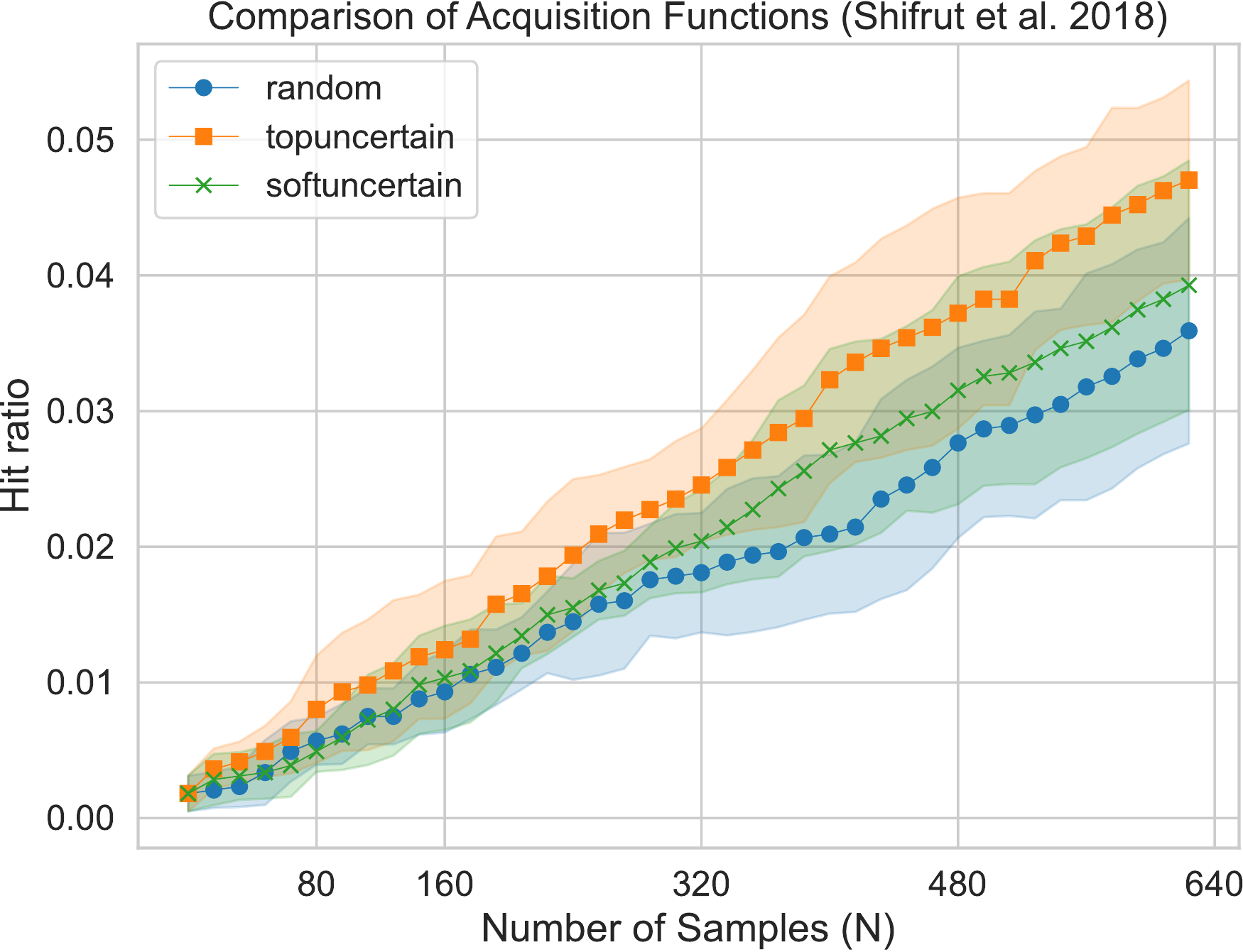}};
                        \end{tikzpicture}
                    }
                \end{subfigure}
                \&
                 \begin{subfigure}{0.27\columnwidth}
                    \hspace{-23mm}
                    \centering
                    \resizebox{\linewidth}{!}{
                        \begin{tikzpicture}
                            \node (img)  {\includegraphics[width=\textwidth]{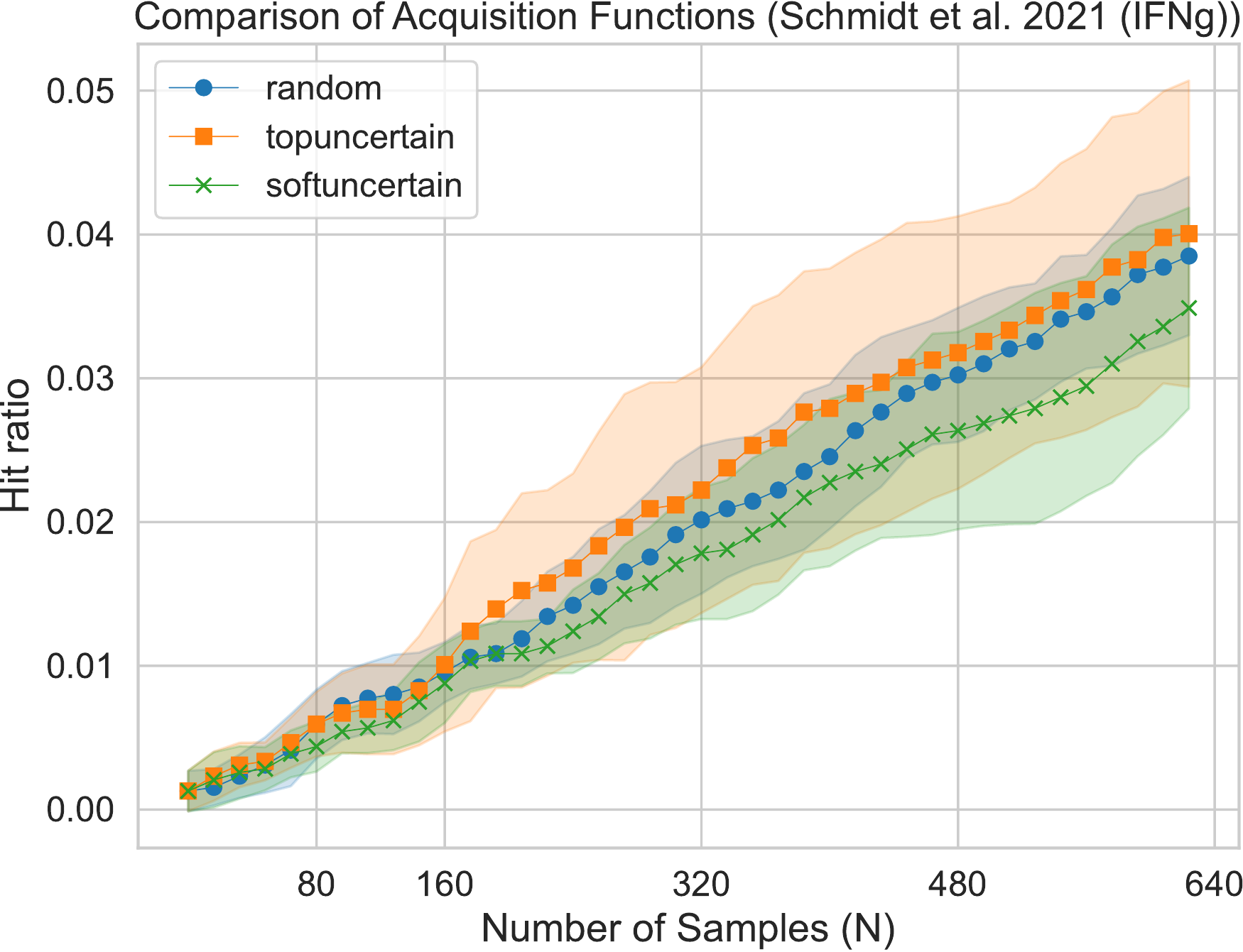}};
                        \end{tikzpicture}
                    }
                \end{subfigure}
                \&
                 \begin{subfigure}{0.27\columnwidth}
                    \hspace{-28mm}
                    \centering
                    \resizebox{\linewidth}{!}{
                        \begin{tikzpicture}
                            \node (img)  {\includegraphics[width=\textwidth]{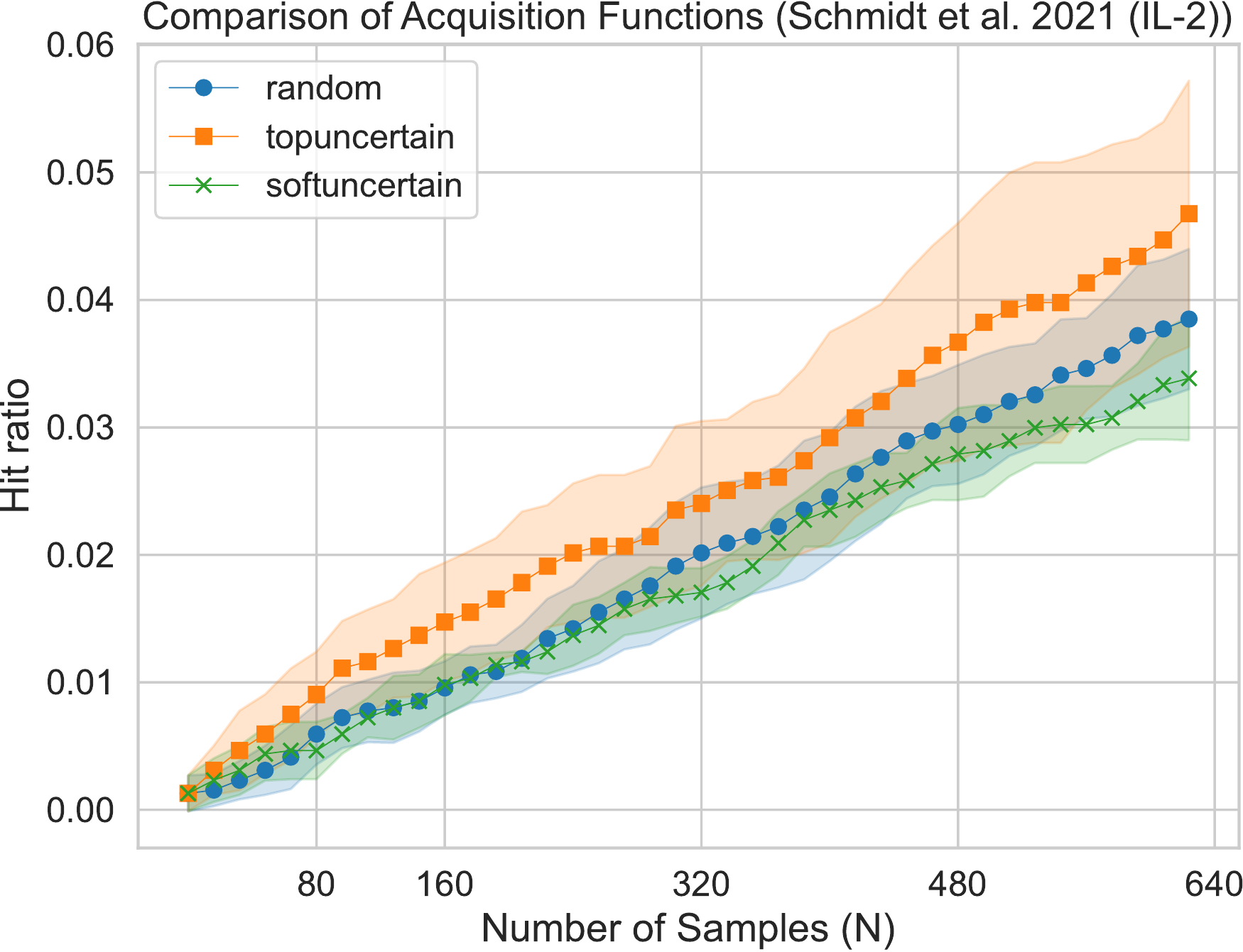}};
                        \end{tikzpicture}
                    }
                \end{subfigure}
                \&
                \begin{subfigure}{0.28\columnwidth}
                    \hspace{-32mm}
                    \centering
                    \resizebox{\linewidth}{!}{
                        \begin{tikzpicture}
                            \node (img)  {\includegraphics[width=\textwidth]{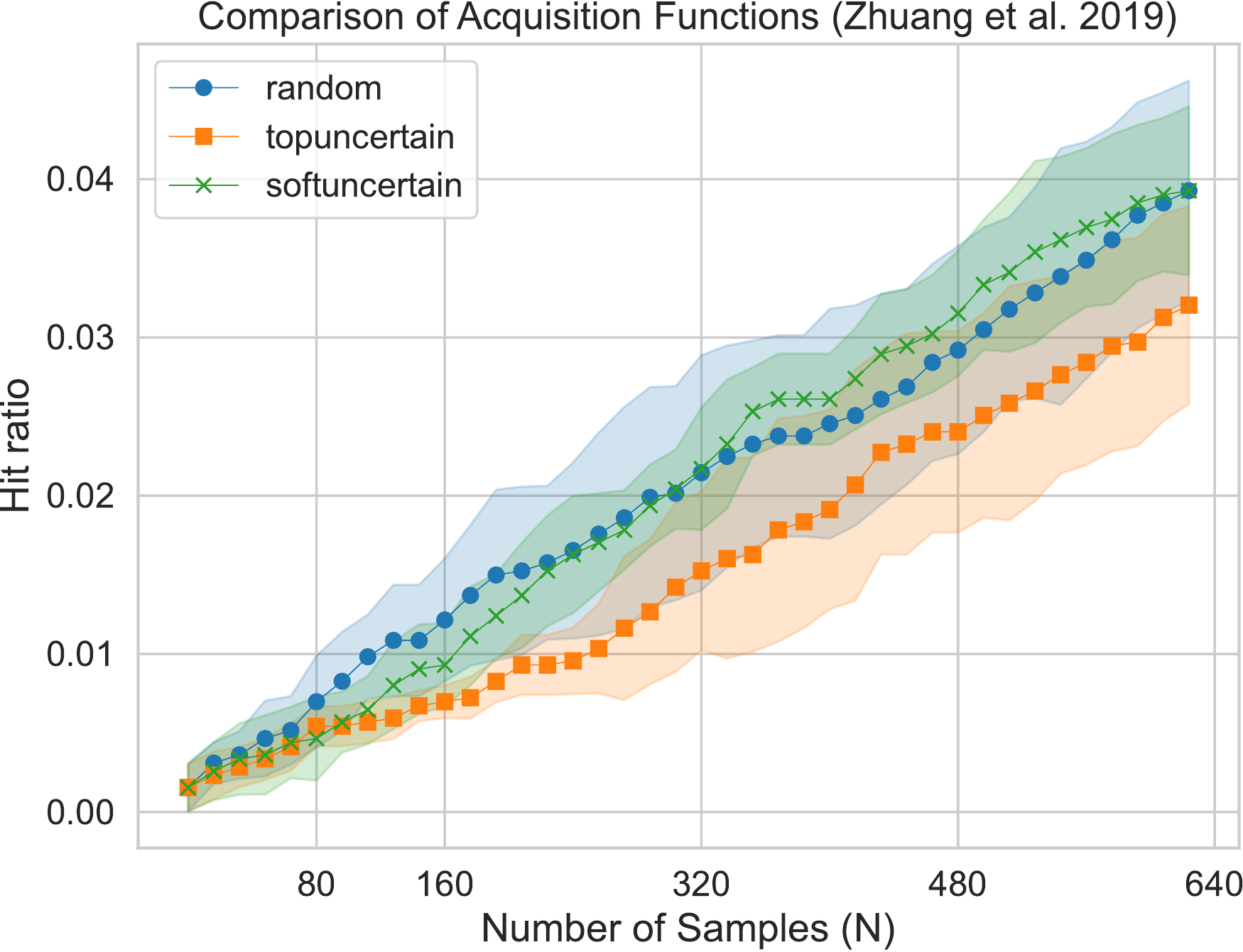}};
                        \end{tikzpicture}
                    }
                \end{subfigure}
                \&
            \\
\begin{subfigure}{0.27\columnwidth}
                    \hspace{-17mm}
                    \centering
                    \resizebox{\linewidth}{!}{
                        \begin{tikzpicture}
                            \node (img)  {\includegraphics[width=\textwidth]{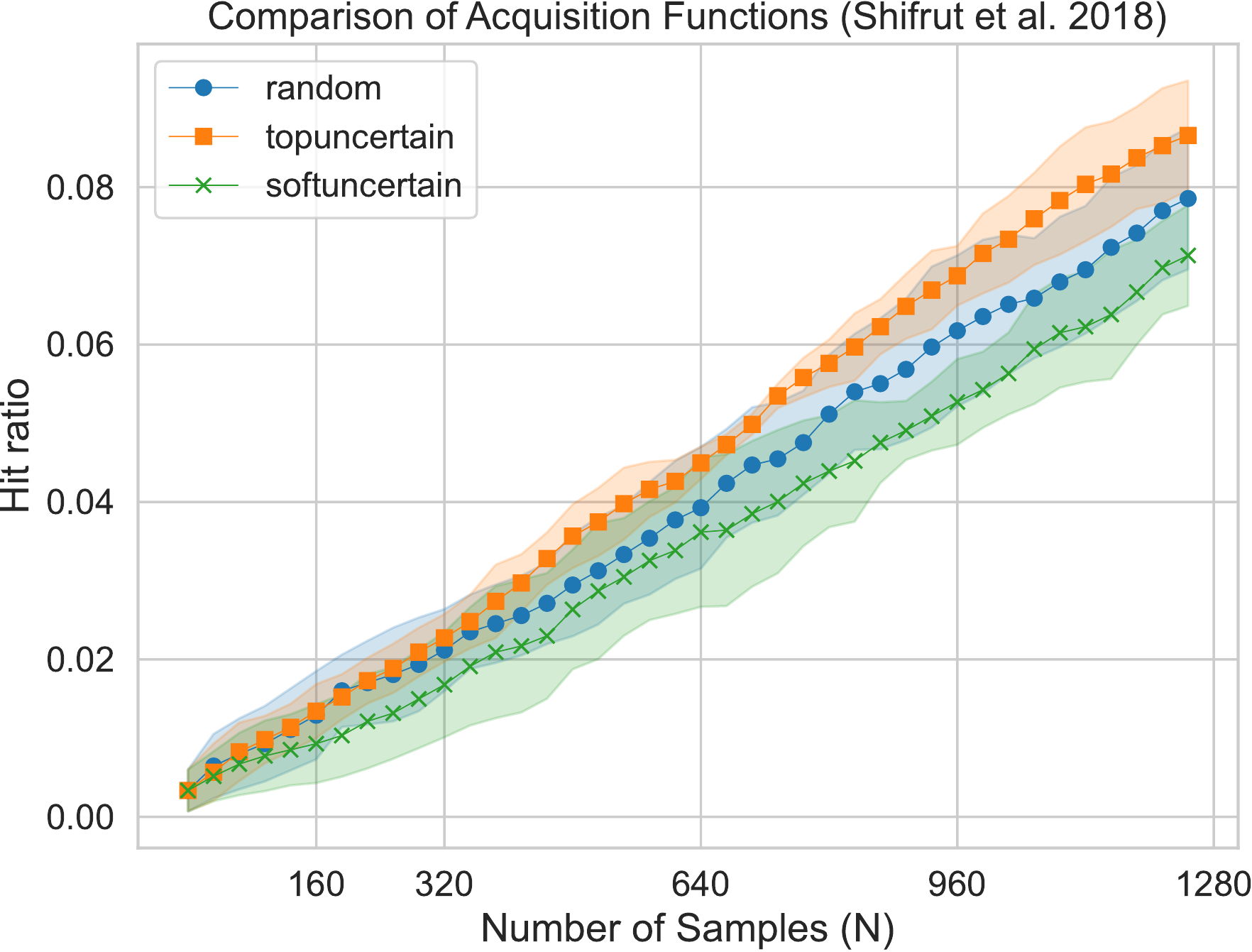}};
                        \end{tikzpicture}
                    }
                \end{subfigure}
                \&
                \begin{subfigure}{0.27\columnwidth}
                    \hspace{-23mm}
                    \centering
                    \resizebox{\linewidth}{!}{
                        \begin{tikzpicture}
                            \node (img)  {\includegraphics[width=\textwidth]{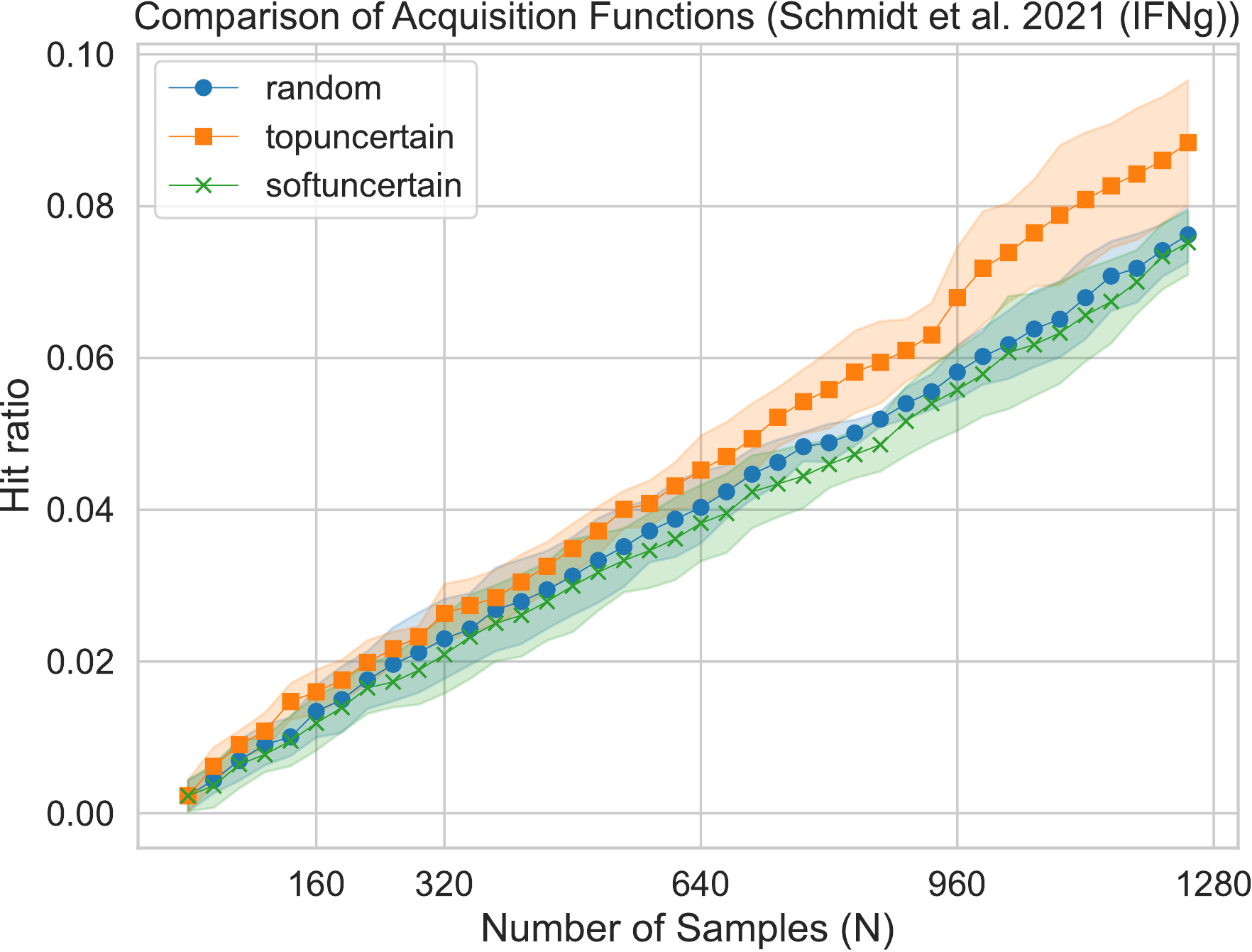}};
                        \end{tikzpicture}
                    }
                \end{subfigure}
                \&
                \begin{subfigure}{0.27\columnwidth}
                    \hspace{-28mm}
                    \centering
                    \resizebox{\linewidth}{!}{
                        \begin{tikzpicture}
                            \node (img)  {\includegraphics[width=\textwidth]{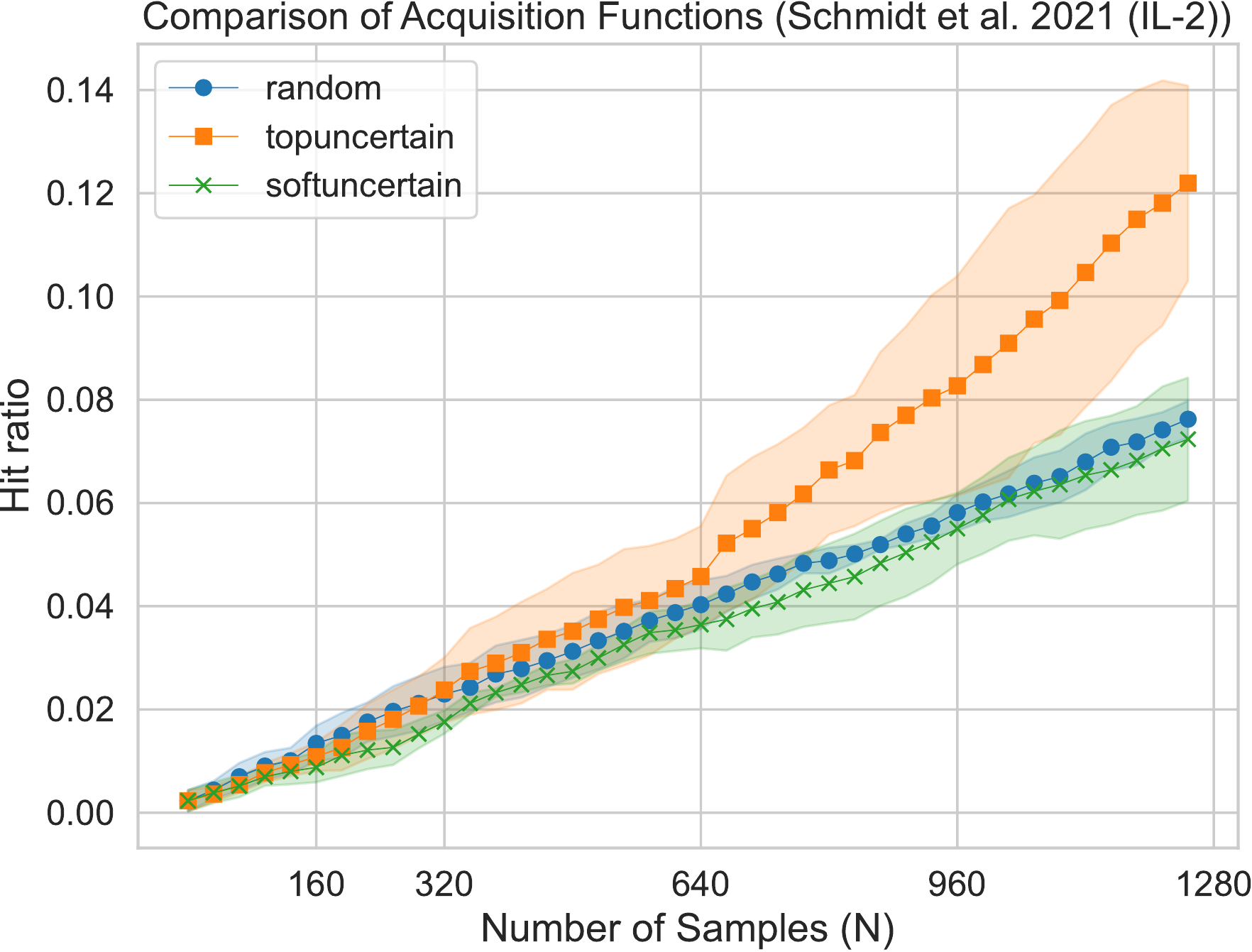}};
                        \end{tikzpicture}
                    }
                \end{subfigure}
                \&
                \begin{subfigure}{0.28\columnwidth}
                    \hspace{-32mm}
                    \centering
                    \resizebox{\linewidth}{!}{
                        \begin{tikzpicture}
                            \node (img)  {\includegraphics[width=\textwidth]{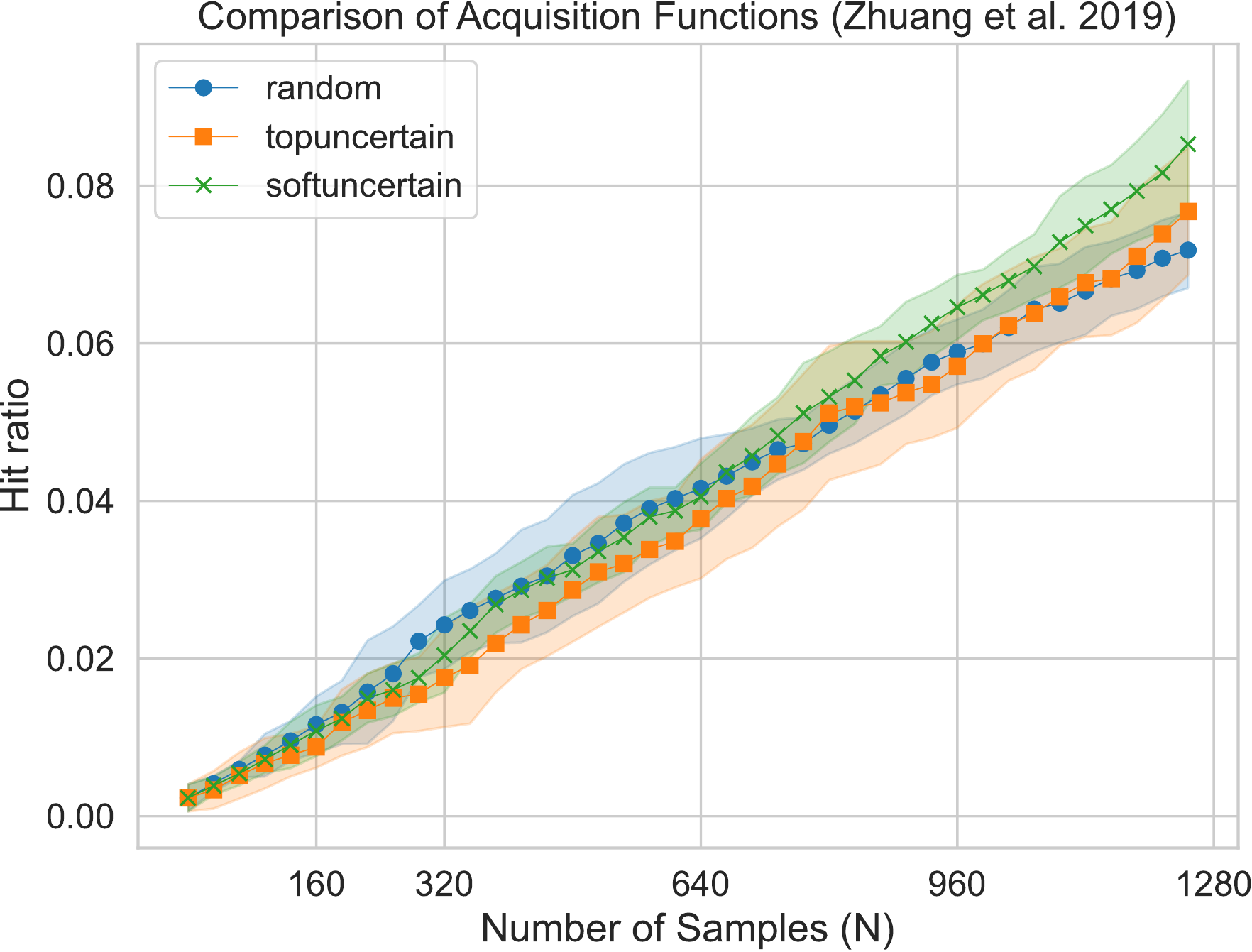}};
                        \end{tikzpicture}
                    }
                \end{subfigure}
                \&
                \\
\begin{subfigure}{0.27\columnwidth}
                    \hspace{-17mm}
                    \centering
                    \resizebox{\linewidth}{!}{
                        \begin{tikzpicture}
                            \node (img)  {\includegraphics[width=\textwidth]{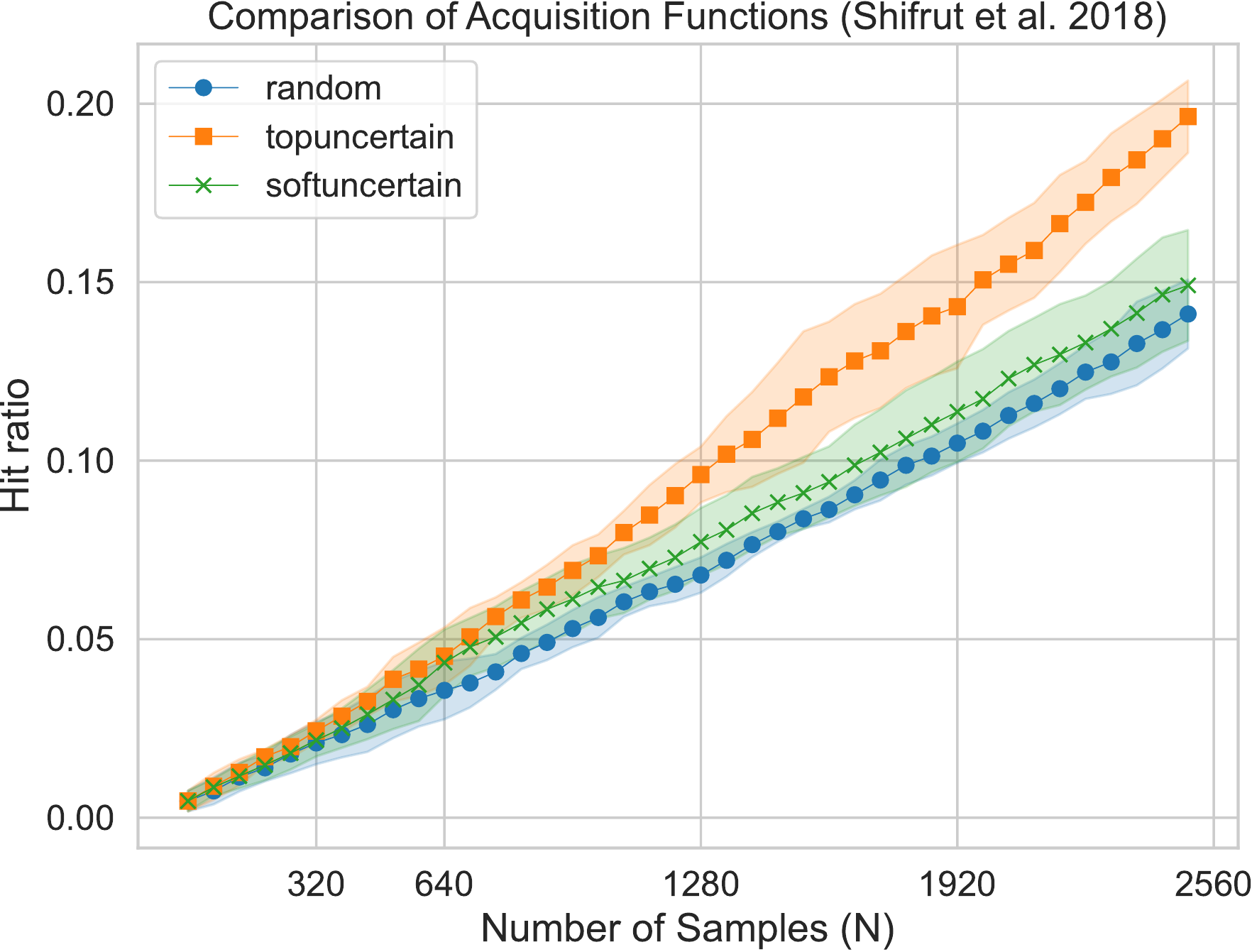}};
                        \end{tikzpicture}
                    }
                \end{subfigure}
                \&
                \begin{subfigure}{0.27\columnwidth}
                    \hspace{-23mm}
                    \centering
                    \resizebox{\linewidth}{!}{
                        \begin{tikzpicture}
                            \node (img)  {\includegraphics[width=\textwidth]{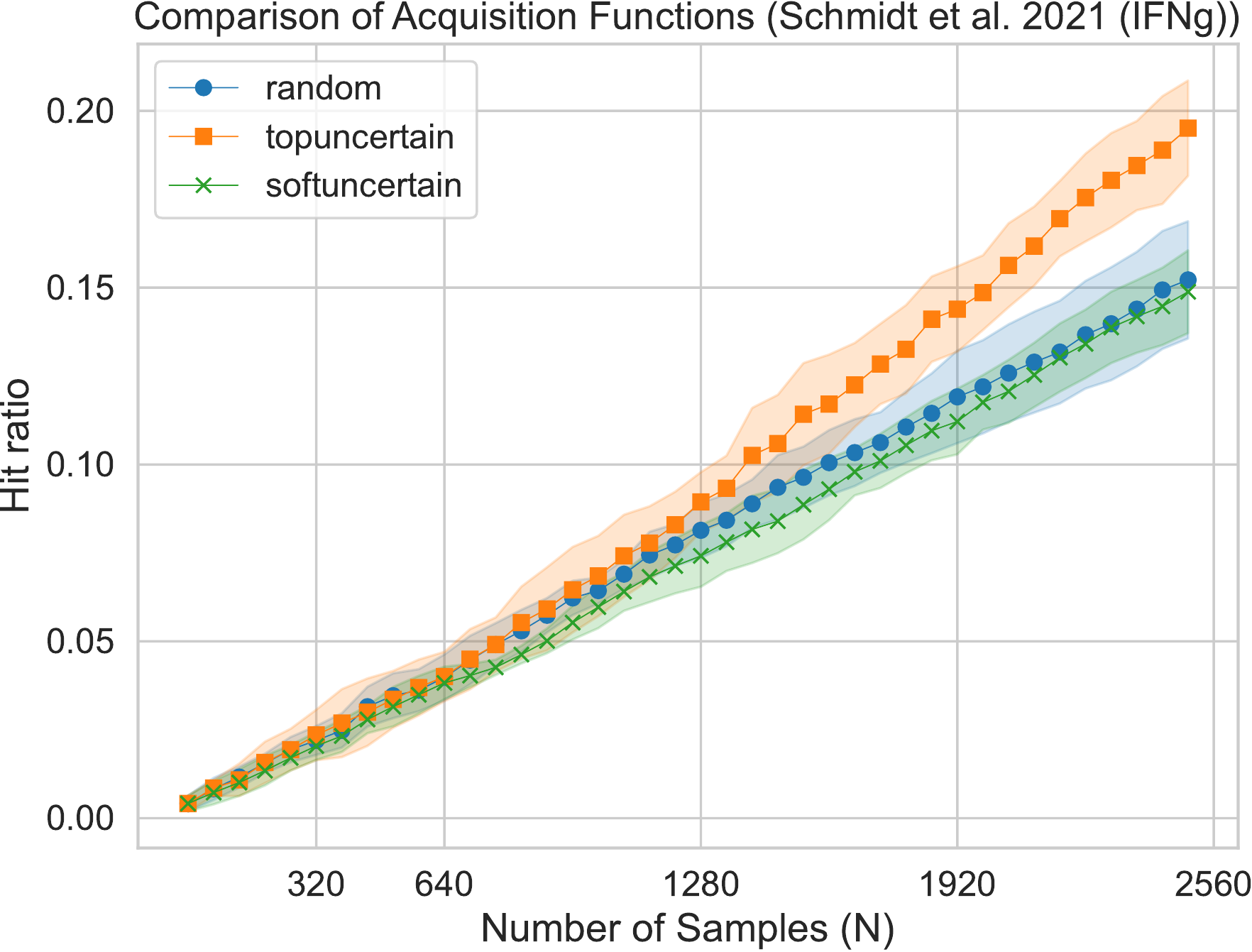}};
                        \end{tikzpicture}
                    }
                \end{subfigure}
                \&
                \begin{subfigure}{0.27\columnwidth}
                    \hspace{-28mm}
                    \centering
                    \resizebox{\linewidth}{!}{
                        \begin{tikzpicture}
                            \node (img)  {\includegraphics[width=\textwidth]{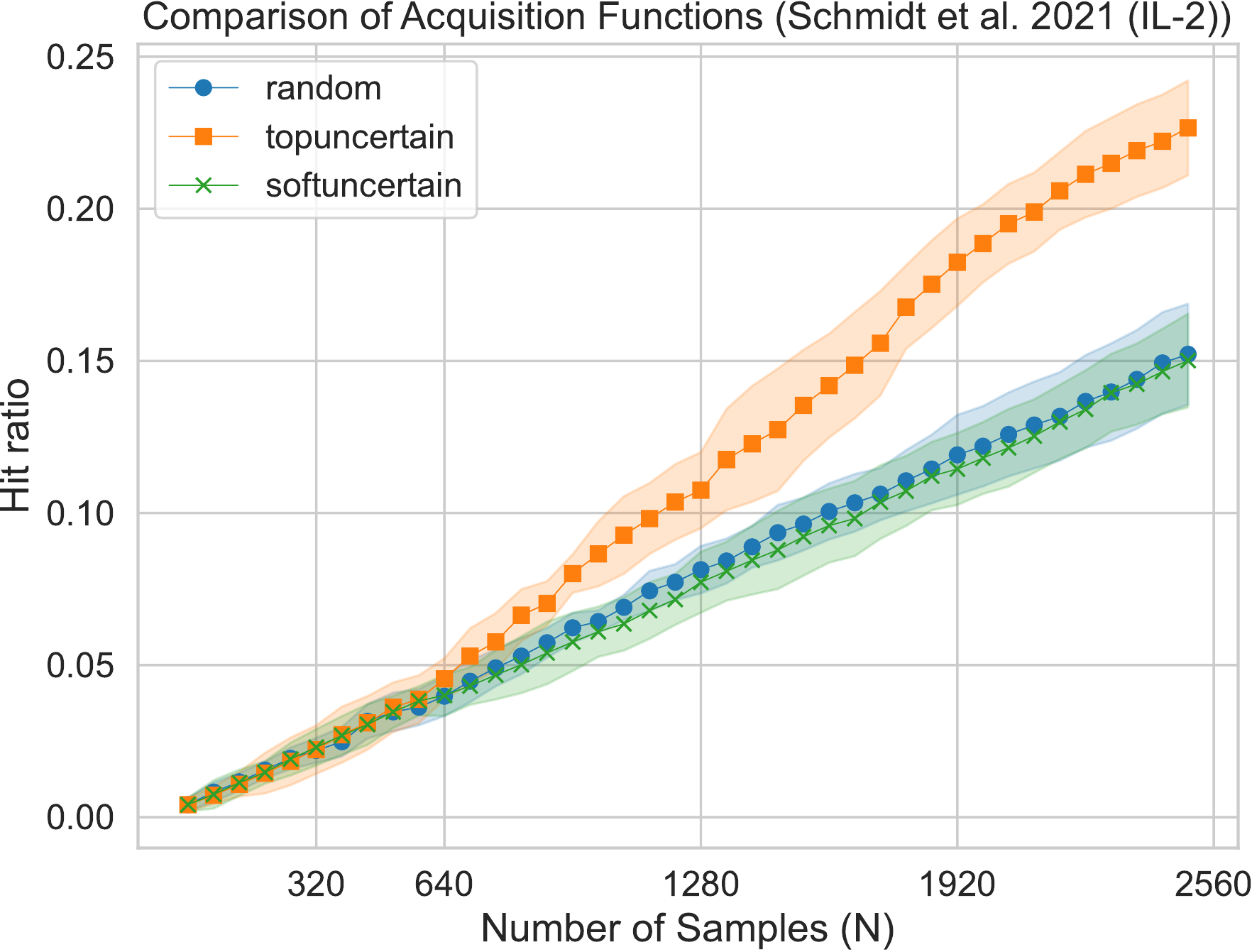}};
                        \end{tikzpicture}
                    }
                \end{subfigure}
                \&
                \begin{subfigure}{0.28\columnwidth}
                    \hspace{-32mm}
                    \centering
                    \resizebox{\linewidth}{!}{
                        \begin{tikzpicture}
                            \node (img)  {\includegraphics[width=\textwidth]{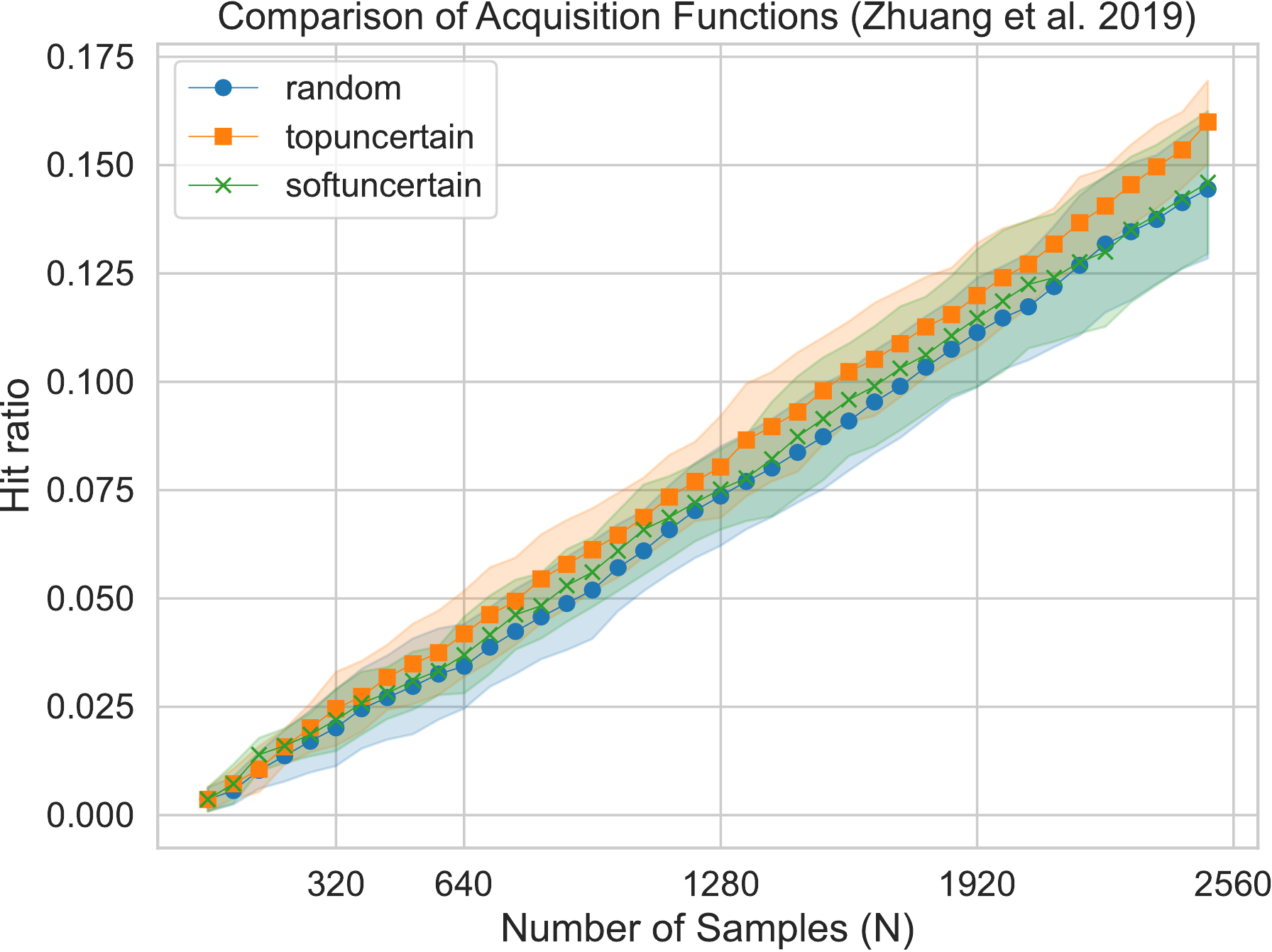}};
                        \end{tikzpicture}
                    }
                \end{subfigure}
                \&
                \\
\begin{subfigure}{0.27\columnwidth}
                    \hspace{-17mm}
                    \centering
                    \resizebox{\linewidth}{!}{
                        \begin{tikzpicture}
                            \node (img)  {\includegraphics[width=\textwidth]{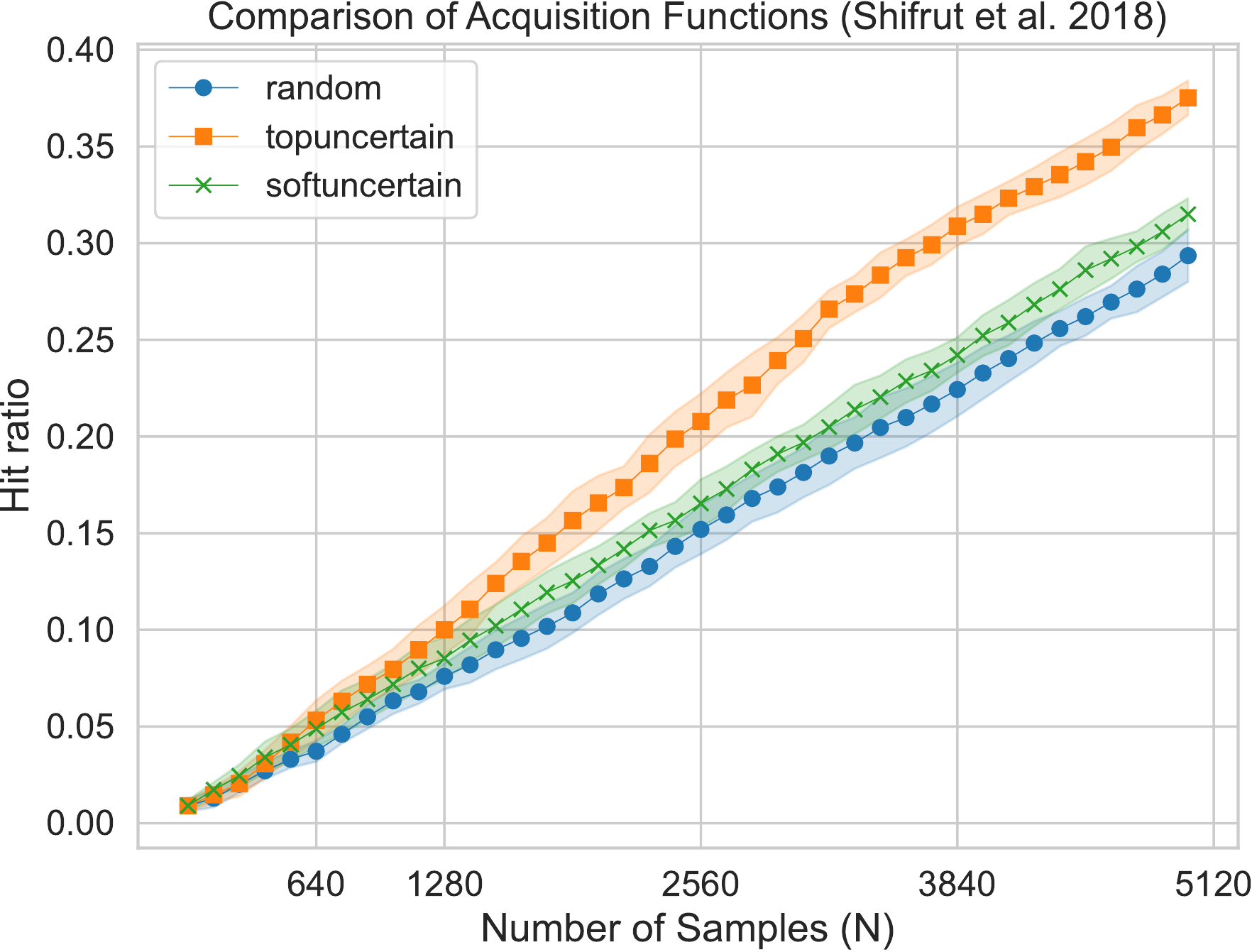}};
                        \end{tikzpicture}
                    }
                \end{subfigure}
                \&
                \begin{subfigure}{0.27\columnwidth}
                    \hspace{-23mm}
                    \centering
                    \resizebox{\linewidth}{!}{
                        \begin{tikzpicture}
                            \node (img)  {\includegraphics[width=\textwidth]{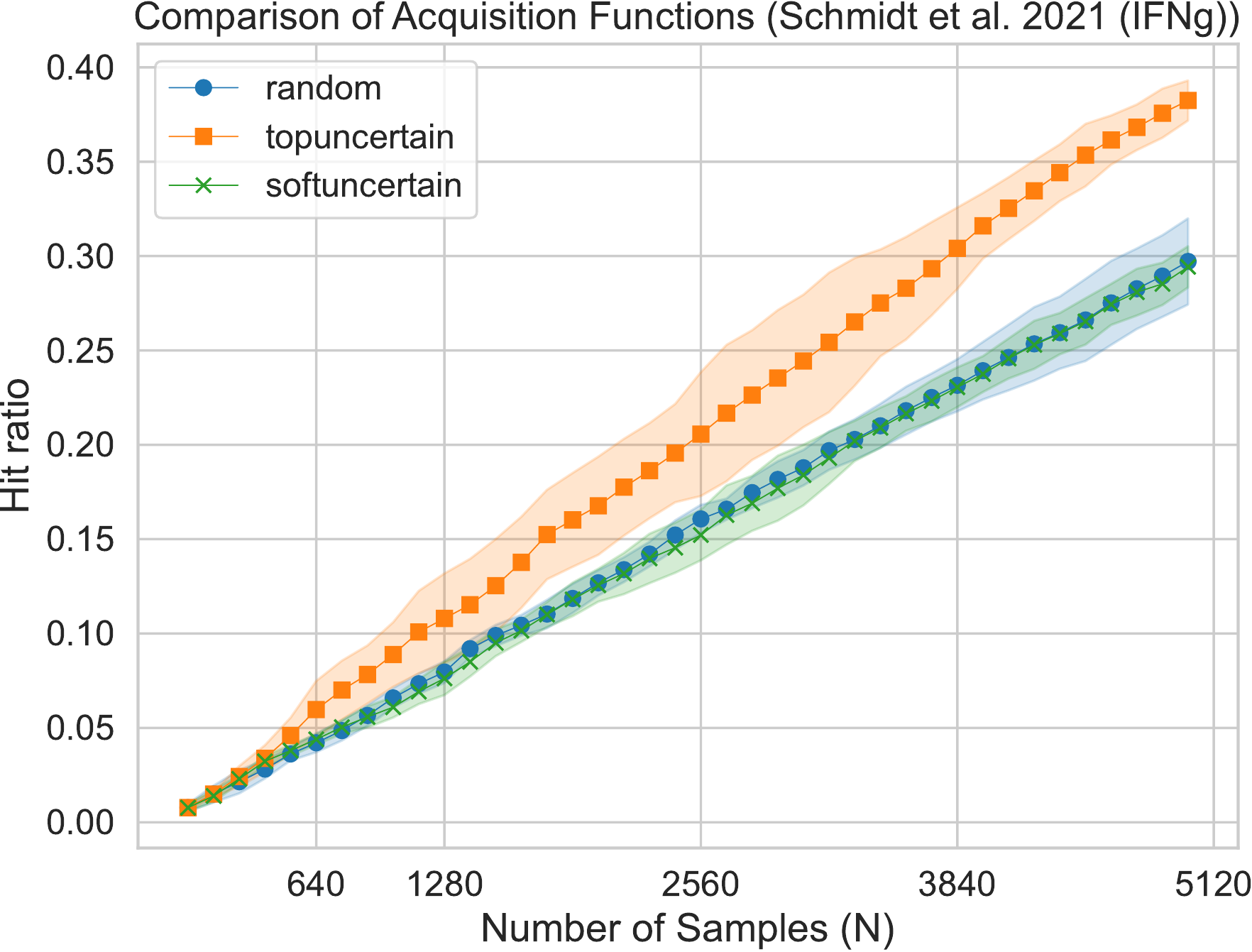}};
                        \end{tikzpicture}
                    }
                \end{subfigure}
                \&
                \begin{subfigure}{0.28\columnwidth}
                    \hspace{-28mm}
                    \centering
                    \resizebox{\linewidth}{!}{
                        \begin{tikzpicture}
                            \node (img)  {\includegraphics[width=\textwidth]{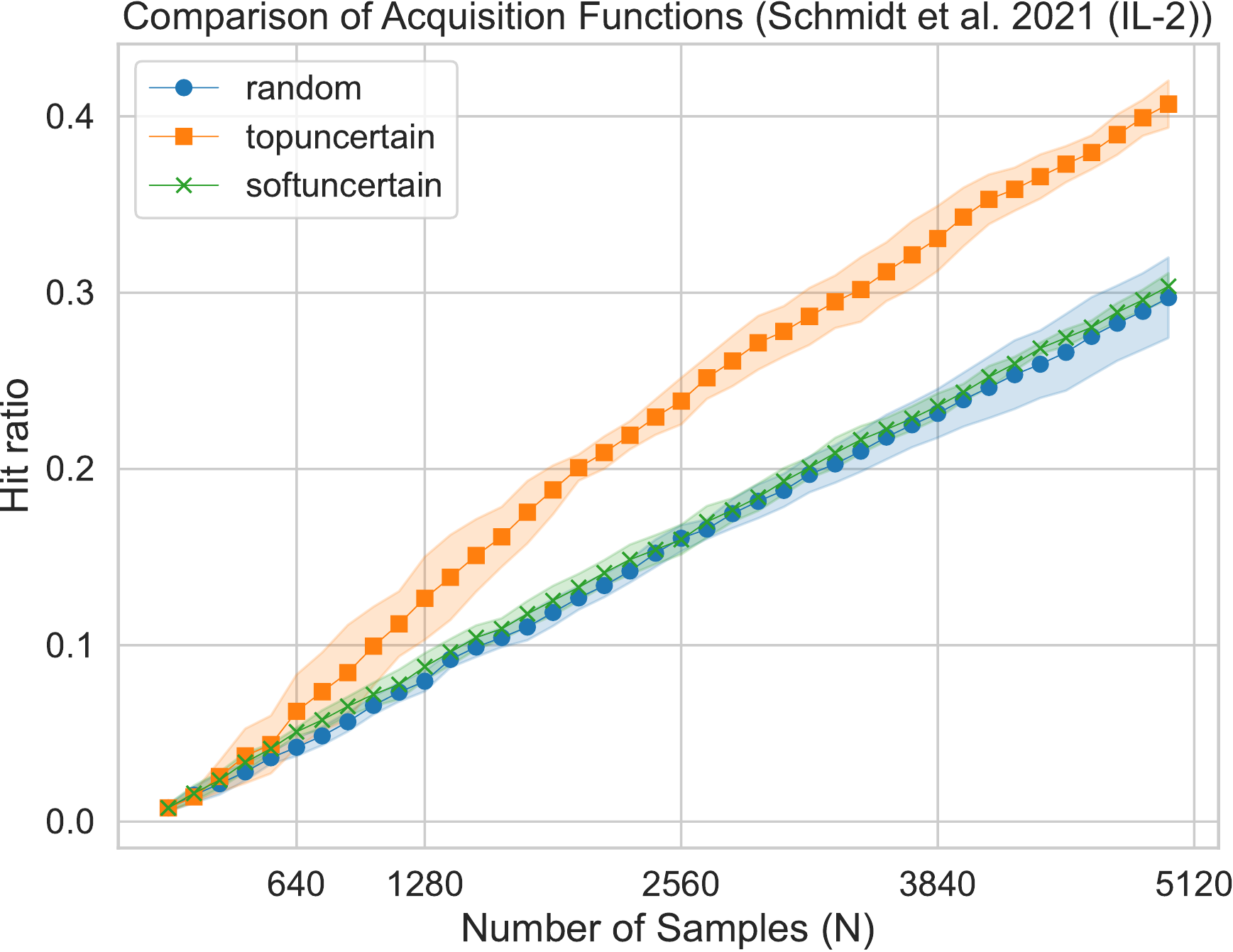}};
                        \end{tikzpicture}
                    }
                \end{subfigure}
                \&
                \begin{subfigure}{0.29\columnwidth}
                    \hspace{-32mm}
                    \centering
                    \resizebox{\linewidth}{!}{
                        \begin{tikzpicture}
                            \node (img)  {\includegraphics[width=\textwidth]{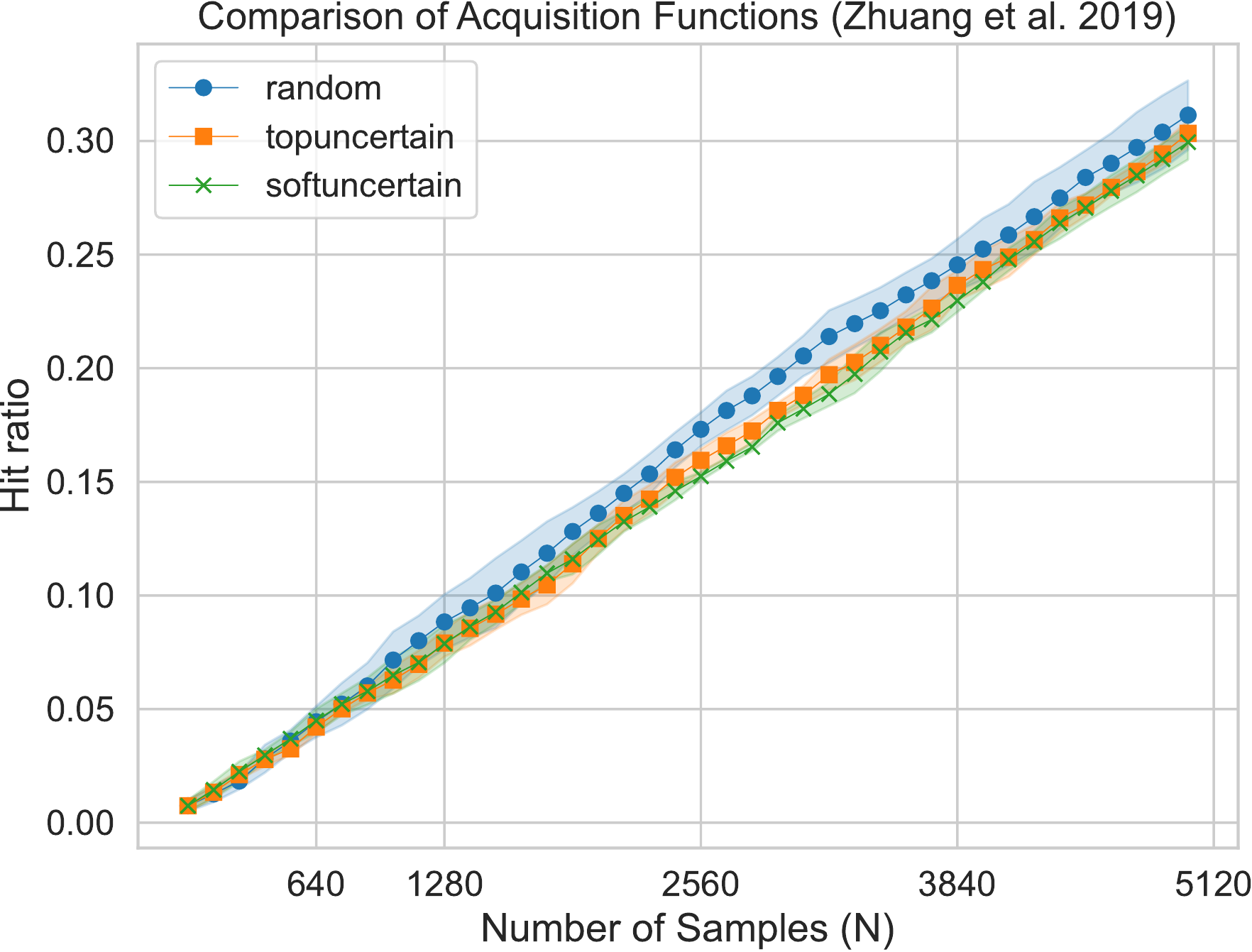}};
                        \end{tikzpicture}
                    }
                \end{subfigure}
                \&
                \\
\begin{subfigure}{0.275\columnwidth}
                    \hspace{-17mm}
                    \centering
                    \resizebox{\linewidth}{!}{
                        \begin{tikzpicture}
                            \node (img)  {\includegraphics[width=\textwidth]{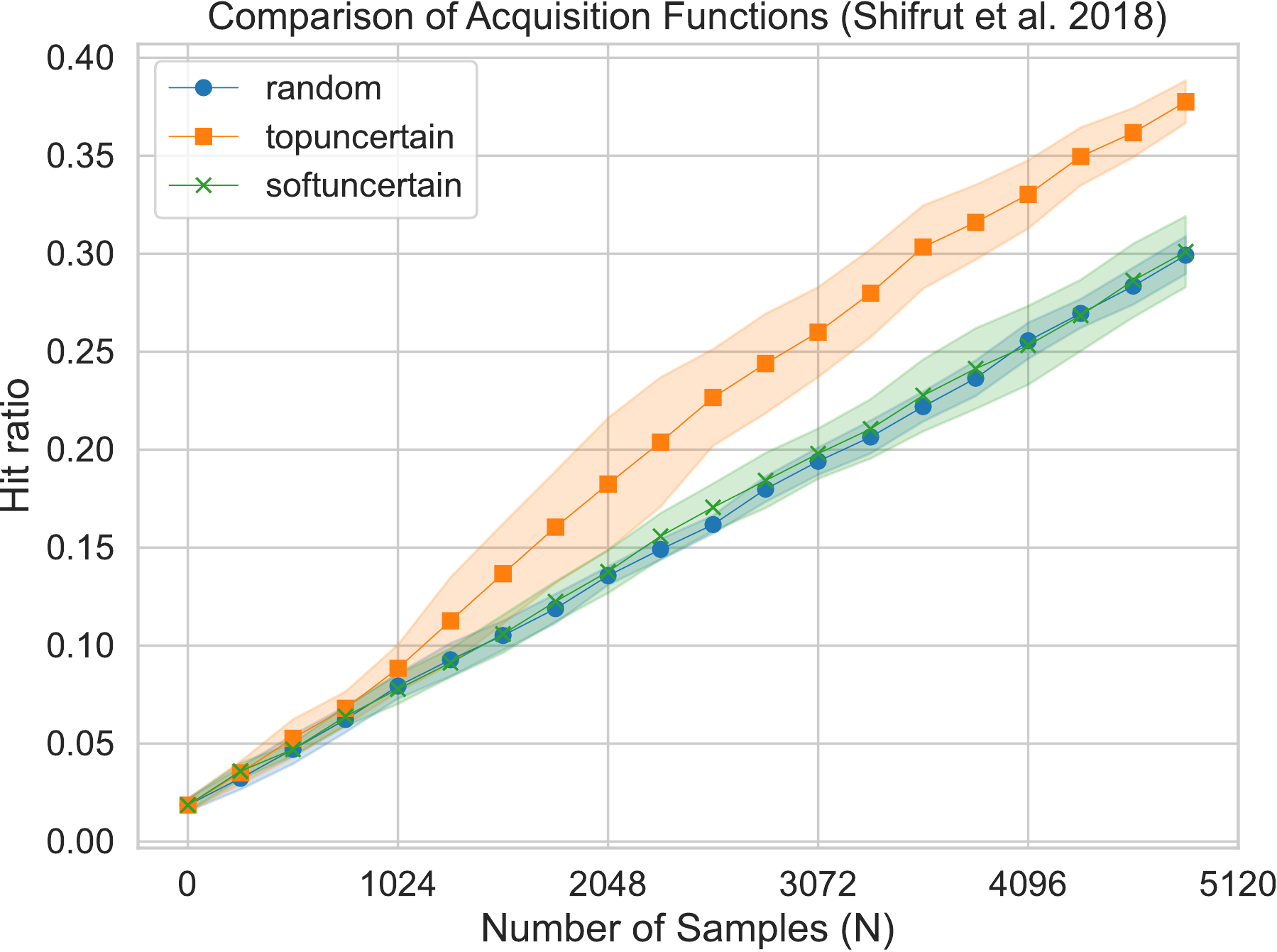}};
                        \end{tikzpicture}
                    }
                \end{subfigure}
                \&
                \begin{subfigure}{0.27\columnwidth}
                    \hspace{-23mm}
                    \centering
                    \resizebox{\linewidth}{!}{
                        \begin{tikzpicture}
                            \node (img)  {\includegraphics[width=\textwidth]{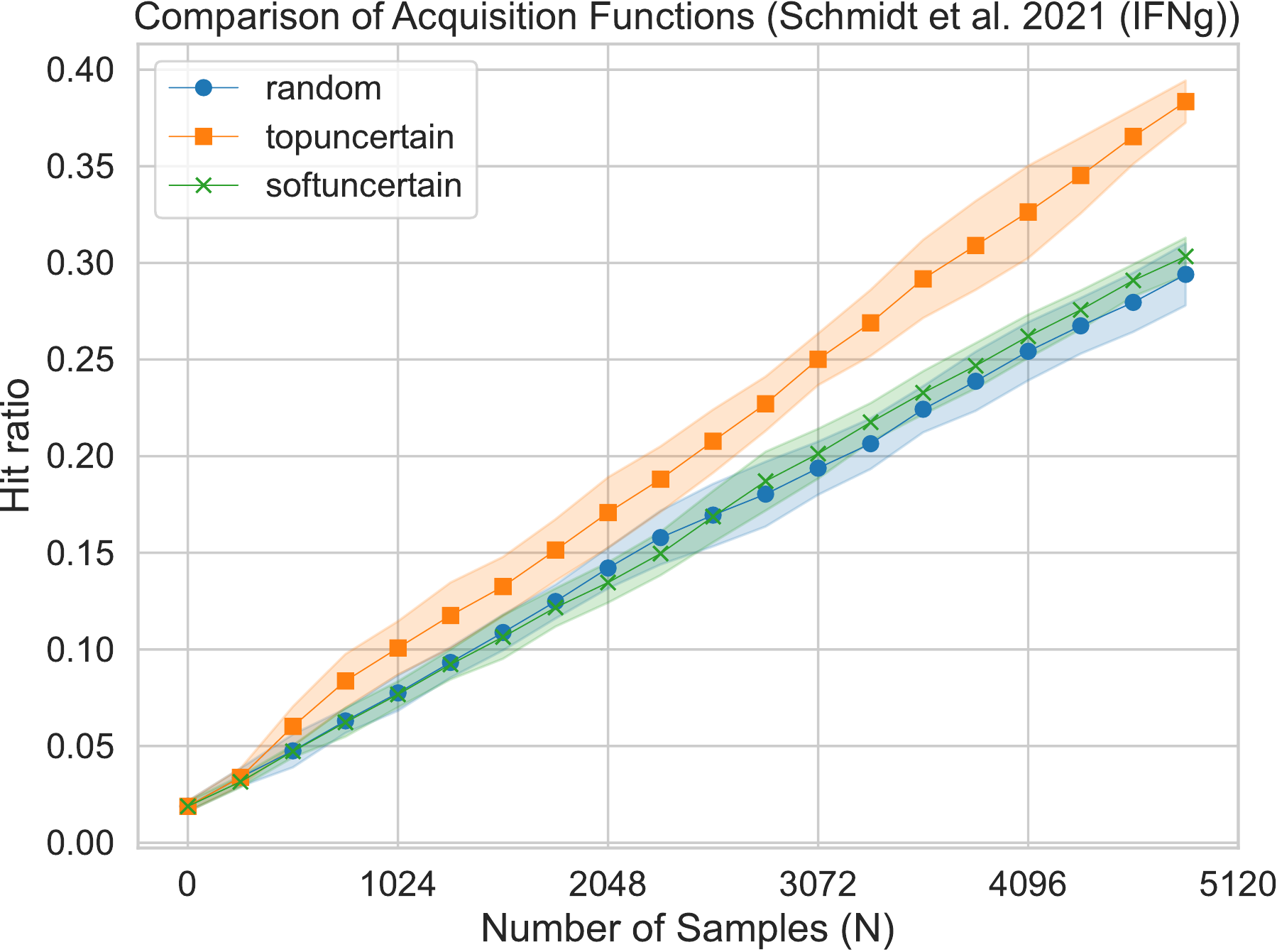}};
                        \end{tikzpicture}
                    }
                \end{subfigure}
                \&
                \begin{subfigure}{0.27\columnwidth}
                    \hspace{-28mm}
                    \centering
                    \resizebox{\linewidth}{!}{
                        \begin{tikzpicture}
                            \node (img)  {\includegraphics[width=\textwidth]{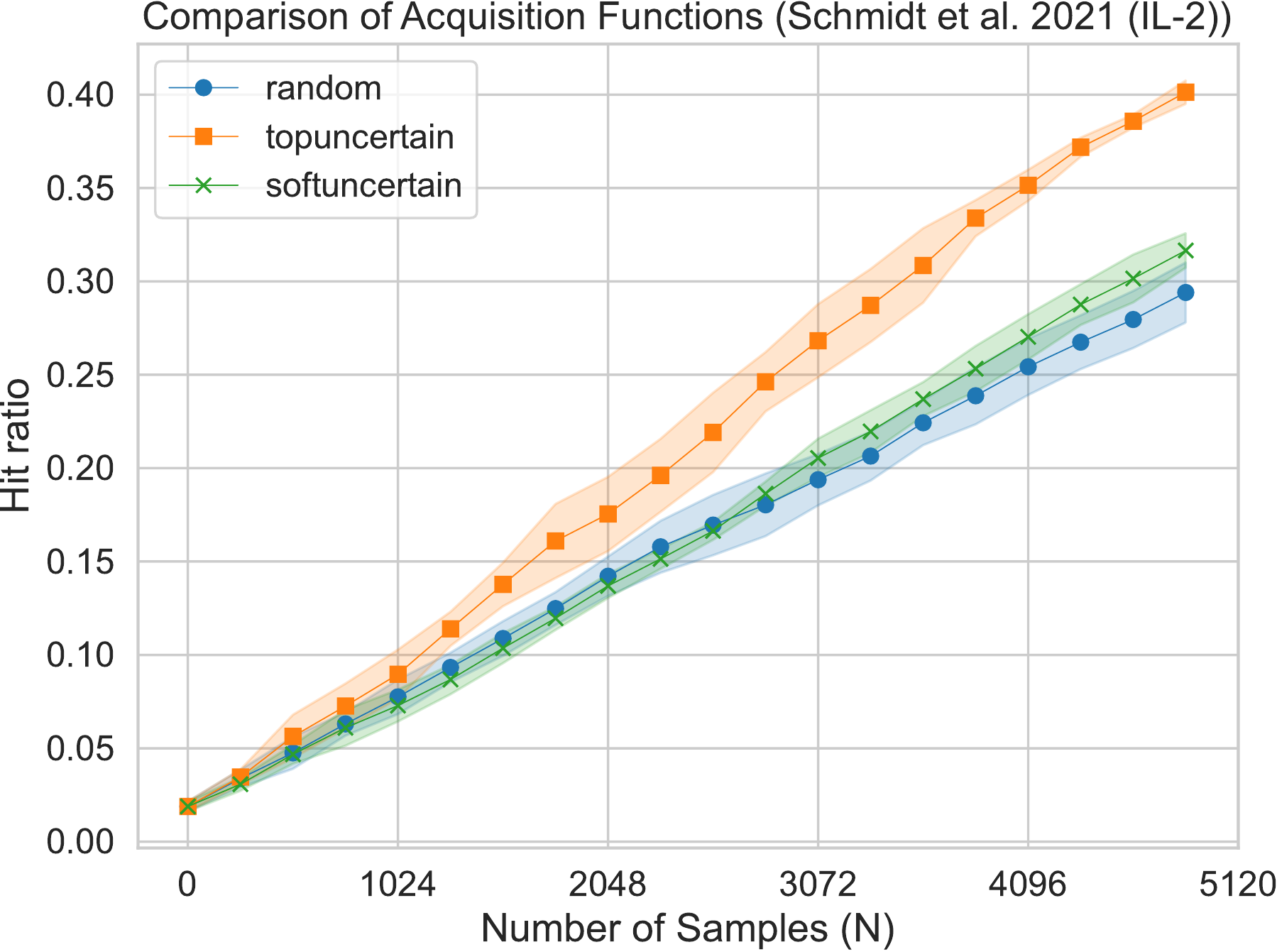}};
                        \end{tikzpicture}
                    }
                \end{subfigure}
                \&
                \begin{subfigure}{0.29\columnwidth}
                    \hspace{-32mm}
                    \centering
                    \resizebox{\linewidth}{!}{
                        \begin{tikzpicture}
                            \node (img)  {\includegraphics[width=\textwidth]{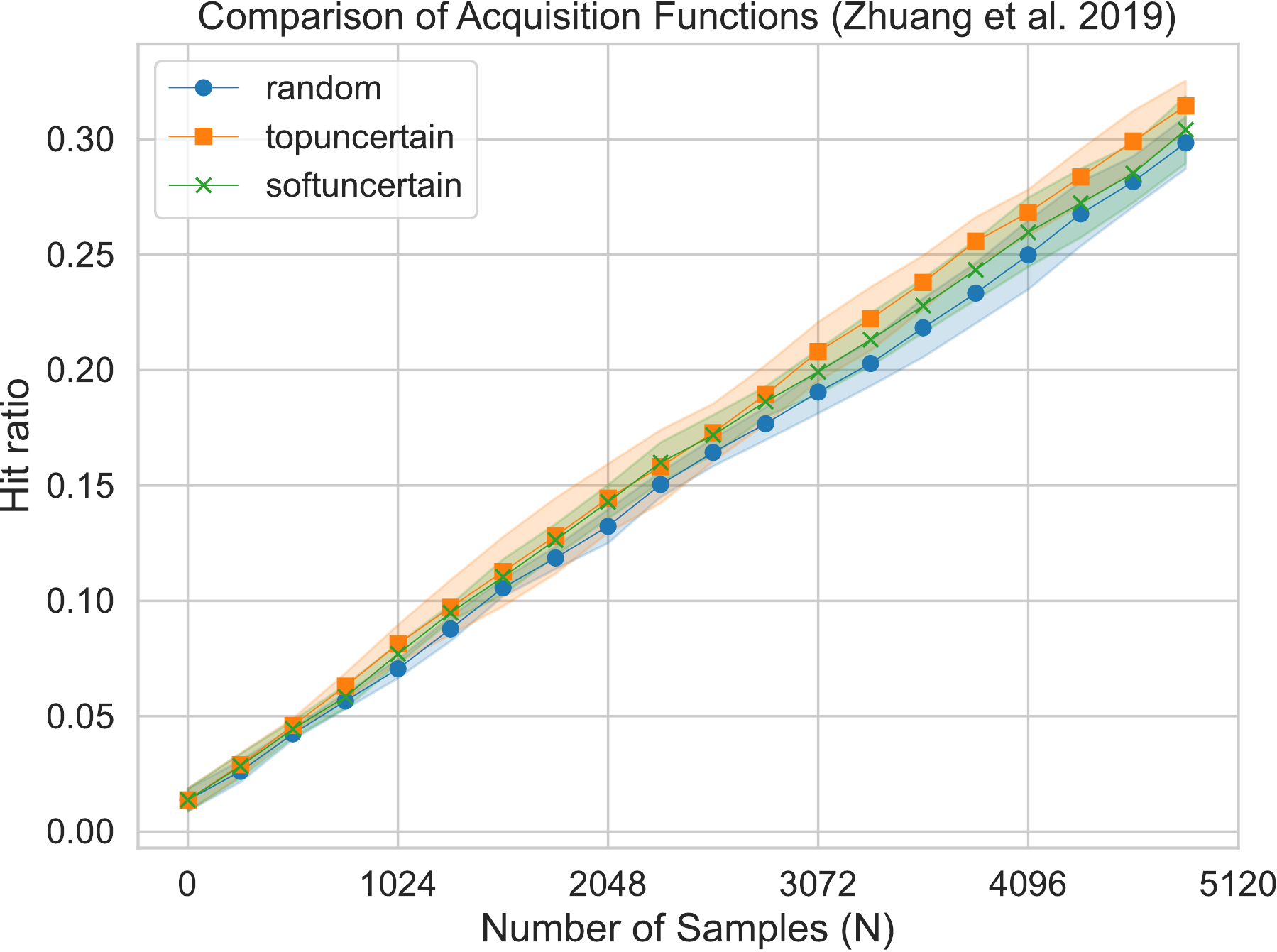}};
                        \end{tikzpicture}
                    }
                \end{subfigure}
                \&
                \\
\begin{subfigure}{0.28\columnwidth}
                    \hspace{-17mm}
                    \centering
                    \resizebox{\linewidth}{!}{
                        \begin{tikzpicture}
                            \node (img)  {\includegraphics[width=\textwidth]{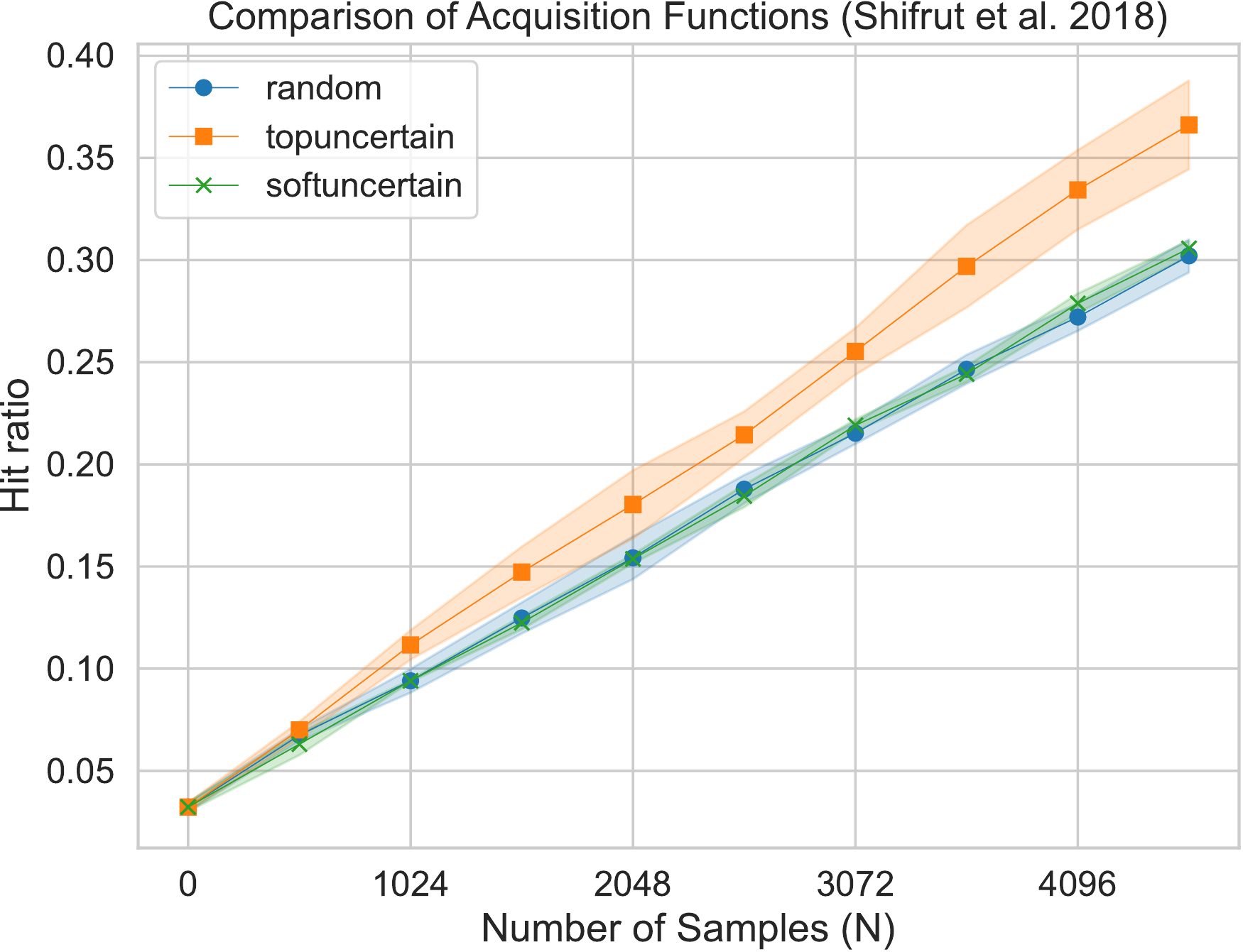}};
                        \end{tikzpicture}
                    }
                \end{subfigure}
                \&
                \begin{subfigure}{0.27\columnwidth}
                    \hspace{-23mm}
                    \centering
                    \resizebox{\linewidth}{!}{
                        \begin{tikzpicture}
                            \node (img)  {\includegraphics[width=\textwidth]{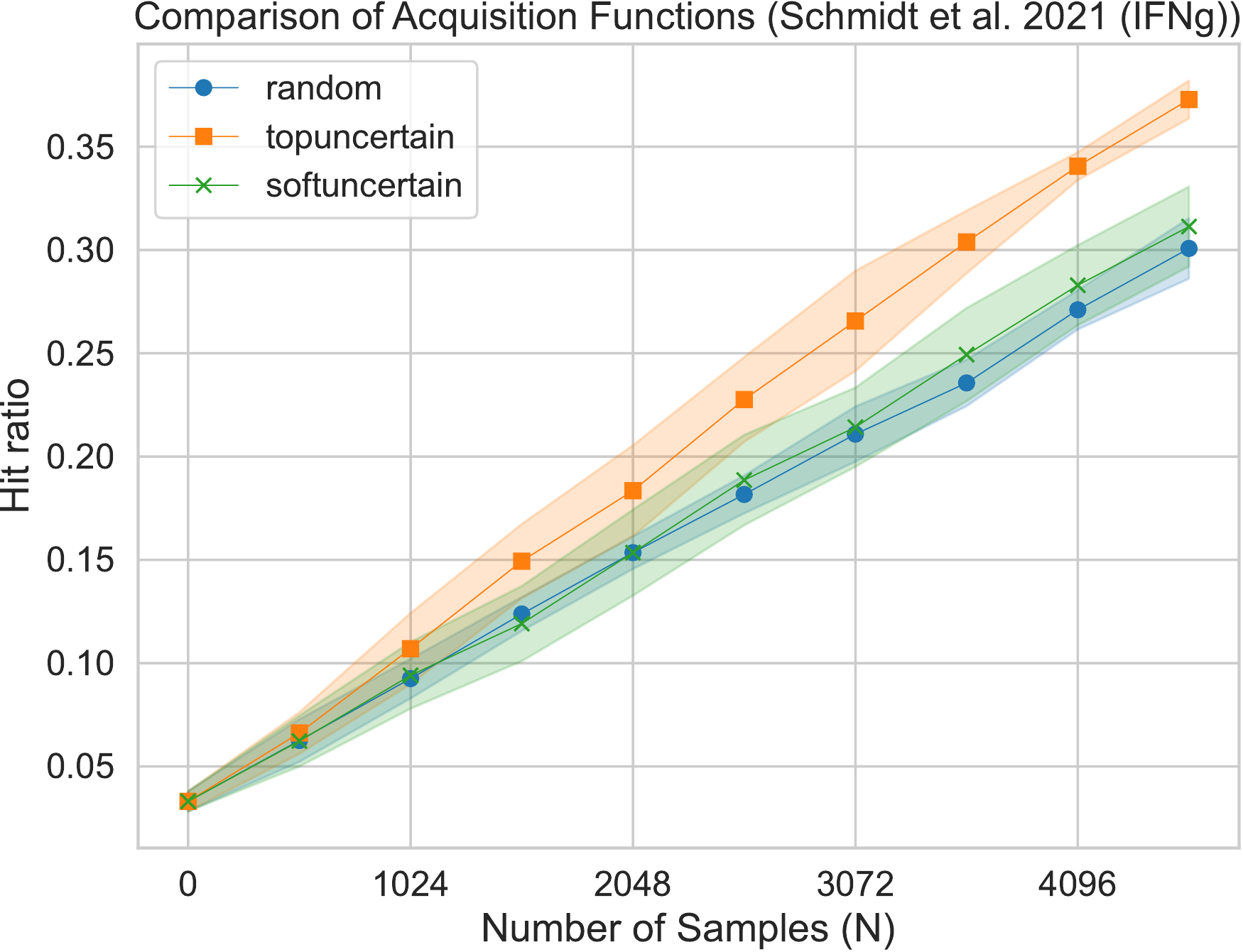}};
                        \end{tikzpicture}
                    }
                \end{subfigure}
                \&
                \begin{subfigure}{0.27\columnwidth}
                    \hspace{-28mm}
                    \centering
                    \resizebox{\linewidth}{!}{
                        \begin{tikzpicture}
                            \node (img)  {\includegraphics[width=\textwidth]{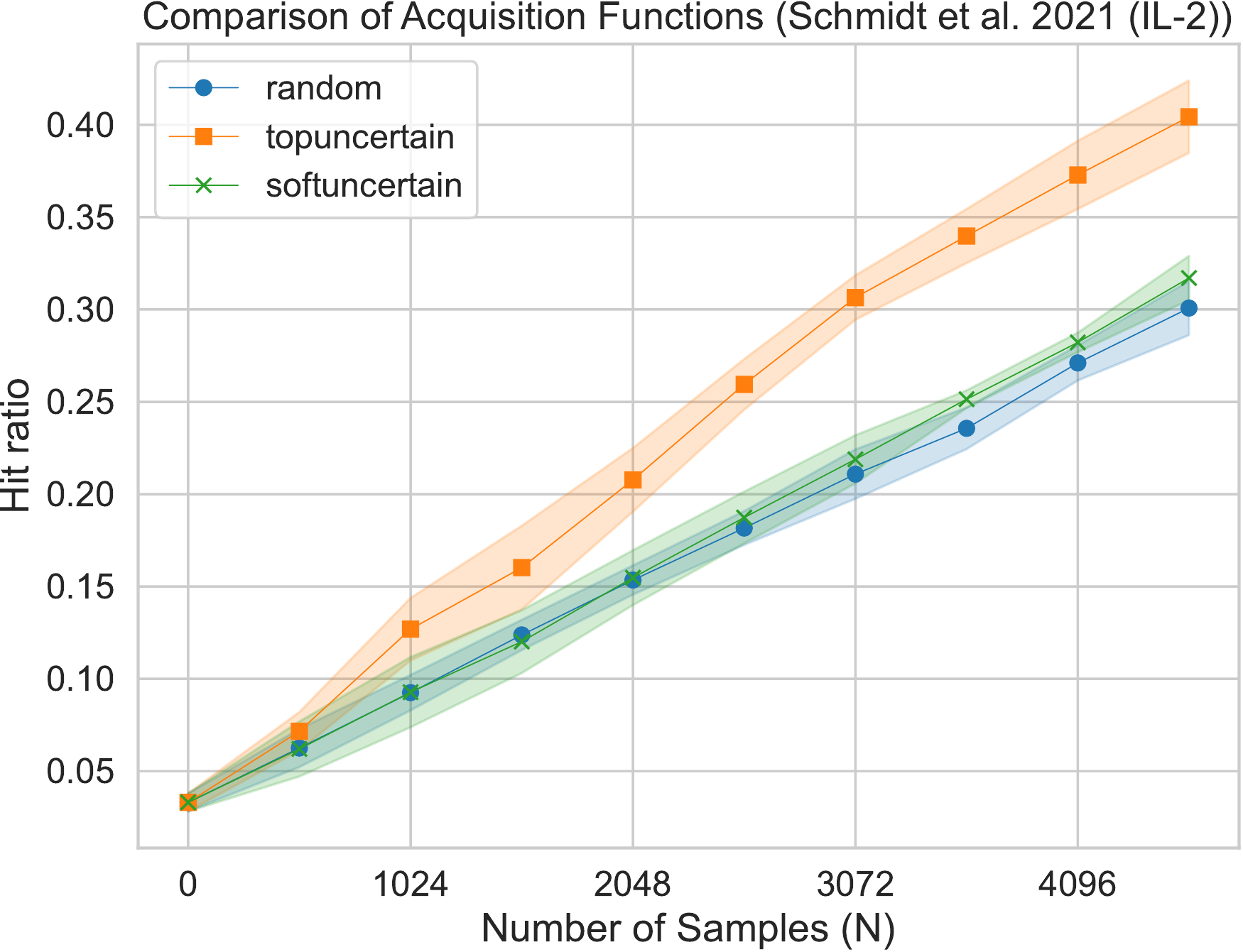}};
                        \end{tikzpicture}
                    }
                \end{subfigure}
                \&
                \begin{subfigure}{0.28\columnwidth}
                    \hspace{-32mm}
                    \centering
                    \resizebox{\linewidth}{!}{
                        \begin{tikzpicture}
                            \node (img)  {\includegraphics[width=\textwidth]{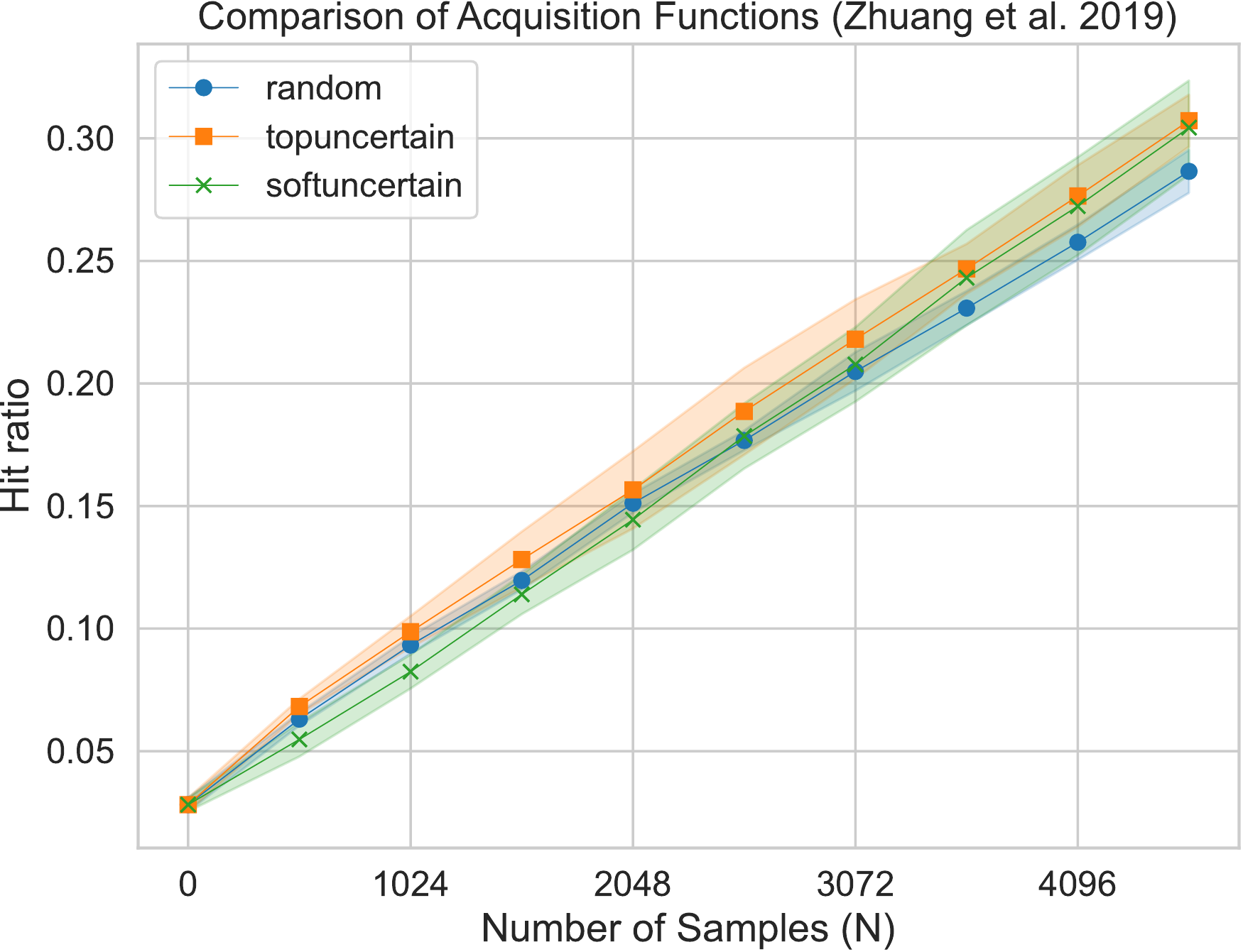}};
                        \end{tikzpicture}
                    }
                \end{subfigure}
                \&
                \\
            \\
           
            \\
            };
            \node [draw=none, rotate=90] at ([xshift=-8mm, yshift=2mm]fig-1-1.west) {\small batch size = 16};
            \node [draw=none, rotate=90] at ([xshift=-8mm, yshift=2mm]fig-2-1.west) {\small batch size = 32};
            \node [draw=none, rotate=90] at ([xshift=-8mm, yshift=2mm]fig-3-1.west) {\small batch size = 64};
            \node [draw=none, rotate=90] at ([xshift=-8mm, yshift=2mm]fig-4-1.west) {\small batch size = 128};
            \node [draw=none, rotate=90] at ([xshift=-8mm, yshift=2mm]fig-5-1.west) {\small batch size = 256};
            \node [draw=none, rotate=90] at ([xshift=-8mm, yshift=2mm]fig-6-1.west) {\small batch size = 512};
            \node [draw=none] at ([xshift=-6mm, yshift=3mm]fig-1-1.north) {\small Shifrut et al. 2018};
            \node [draw=none] at ([xshift=-9mm, yshift=3mm]fig-1-2.north) {\small Schmidt et al. 2021 (IFNg)};
            \node [draw=none] at ([xshift=-11mm, yshift=3mm]fig-1-3.north) {\small Schmidt et al. 2021 (IL-2)};
            \node [draw=none] at ([xshift=-13mm, yshift=2.5mm]fig-1-4.north) {\small Zhuang et al. 2019};
\end{tikzpicture}}
        \vspace{-7mm}
        \caption{The hit ratio of different acquisition for random forest model, different target datasets, and different acquisition batch sizes. We use {STRING} treatment descriptors here. The x-axis shows the number of data points collected so far during the active learning cycles. The y-axis shows the ratio of the set of interesting genes that have been found by the acquisition function up until each cycle.}
        \vspace{-5mm}
        \label{fig:hitratio_rf_feat_string_alldatasets_allbathcsizes}
    \end{figure*}

\begin{figure*}
    \vspace{-2mm}
        \centering
        \makebox[0.72\paperwidth]{\begin{tikzpicture}[ampersand replacement=\&]
            \matrix (fig) [matrix of nodes]{ 
\begin{subfigure}{0.27\columnwidth}
                    \hspace{-17mm}
                    \centering
                    \resizebox{\linewidth}{!}{
                        \begin{tikzpicture}
                            \node (img)  {\includegraphics[width=\textwidth]{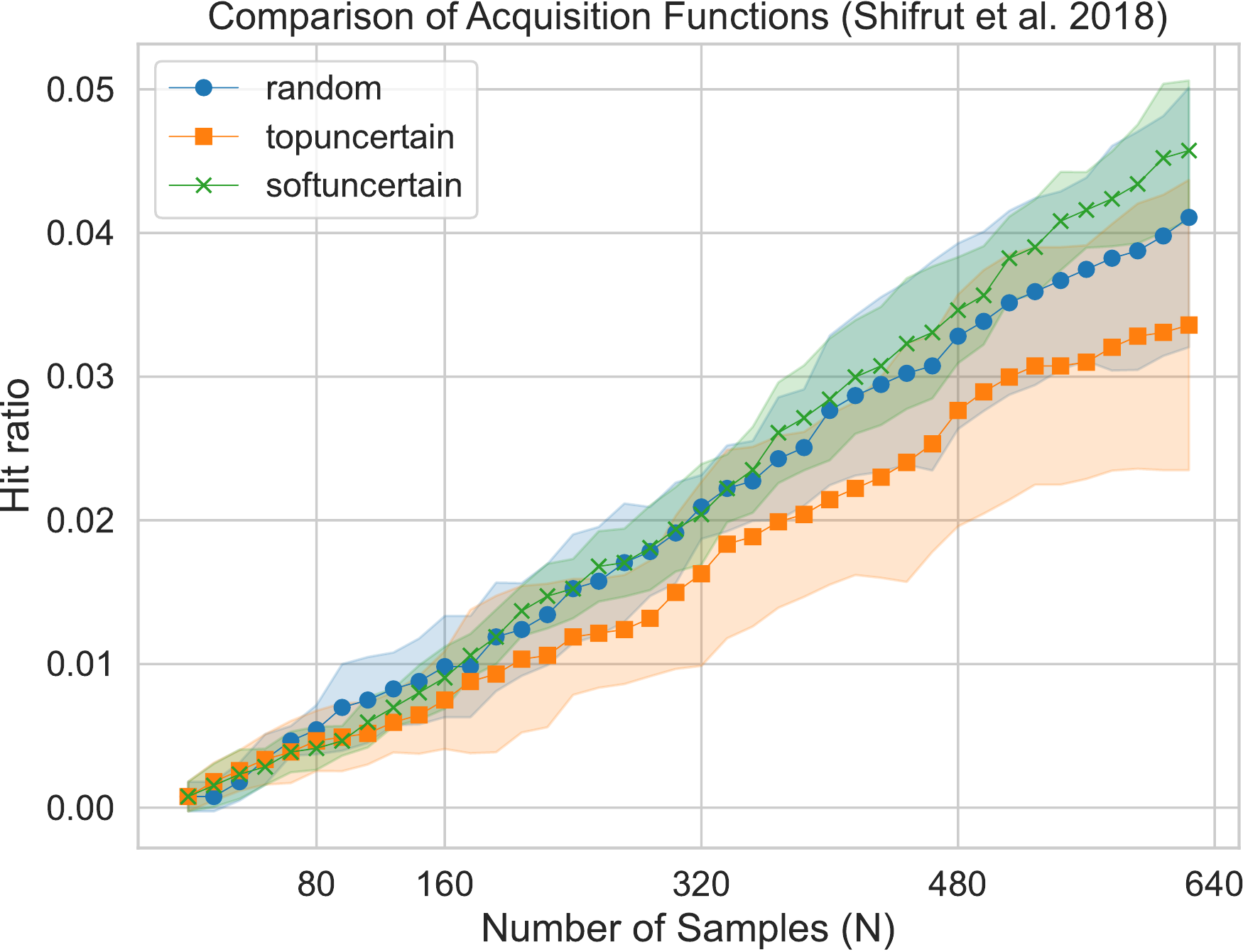}};
                        \end{tikzpicture}
                    }
                \end{subfigure}
                \&
                 \begin{subfigure}{0.27\columnwidth}
                    \hspace{-23mm}
                    \centering
                    \resizebox{\linewidth}{!}{
                        \begin{tikzpicture}
                            \node (img)  {\includegraphics[width=\textwidth]{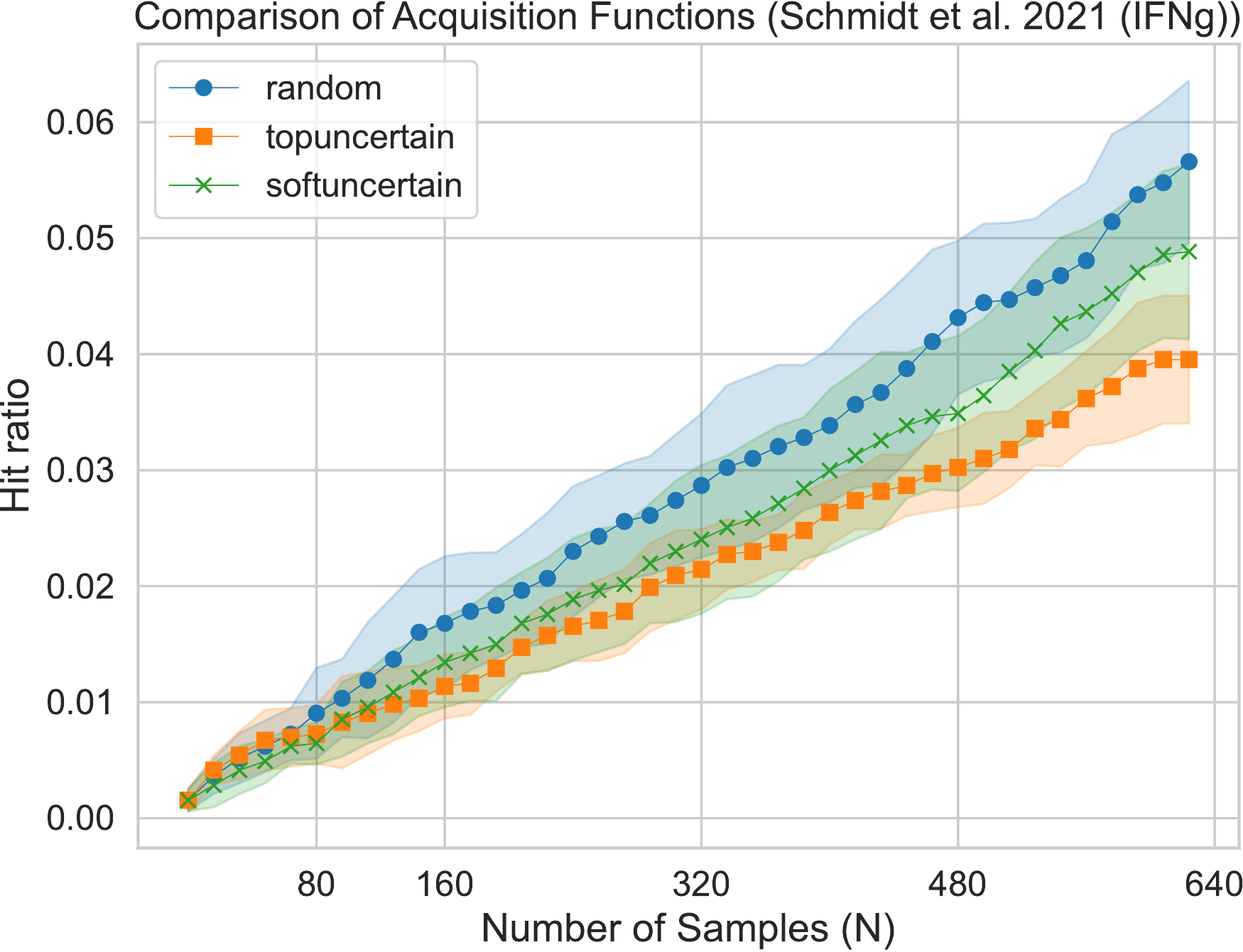}};
                        \end{tikzpicture}
                    }
                \end{subfigure}
                \&
                 \begin{subfigure}{0.27\columnwidth}
                    \hspace{-28mm}
                    \centering
                    \resizebox{\linewidth}{!}{
                        \begin{tikzpicture}
                            \node (img)  {\includegraphics[width=\textwidth]{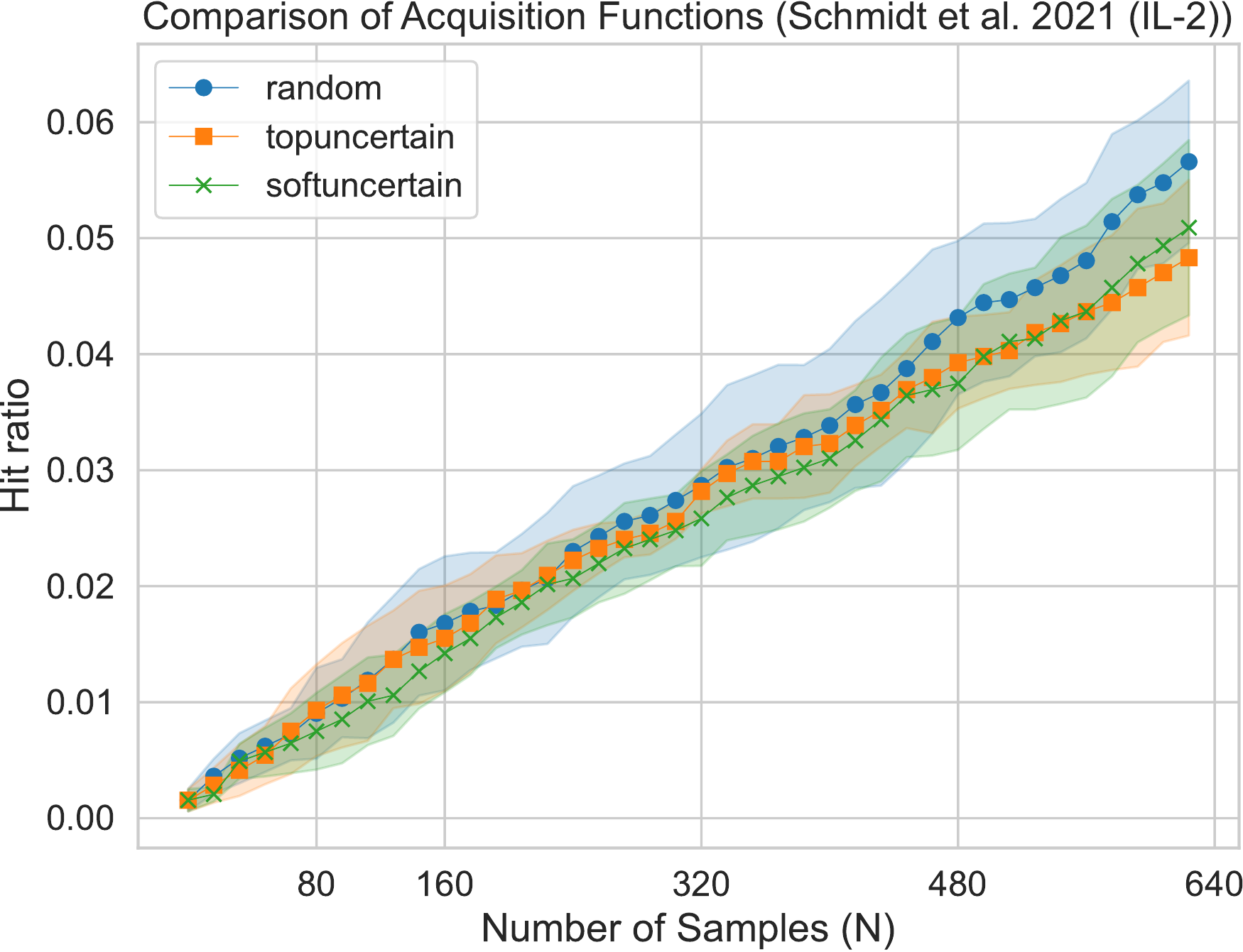}};
                        \end{tikzpicture}
                    }
                \end{subfigure}
                \&
                \begin{subfigure}{0.28\columnwidth}
                    \hspace{-32mm}
                    \centering
                    \resizebox{\linewidth}{!}{
                        \begin{tikzpicture}
                            \node (img)  {\includegraphics[width=\textwidth]{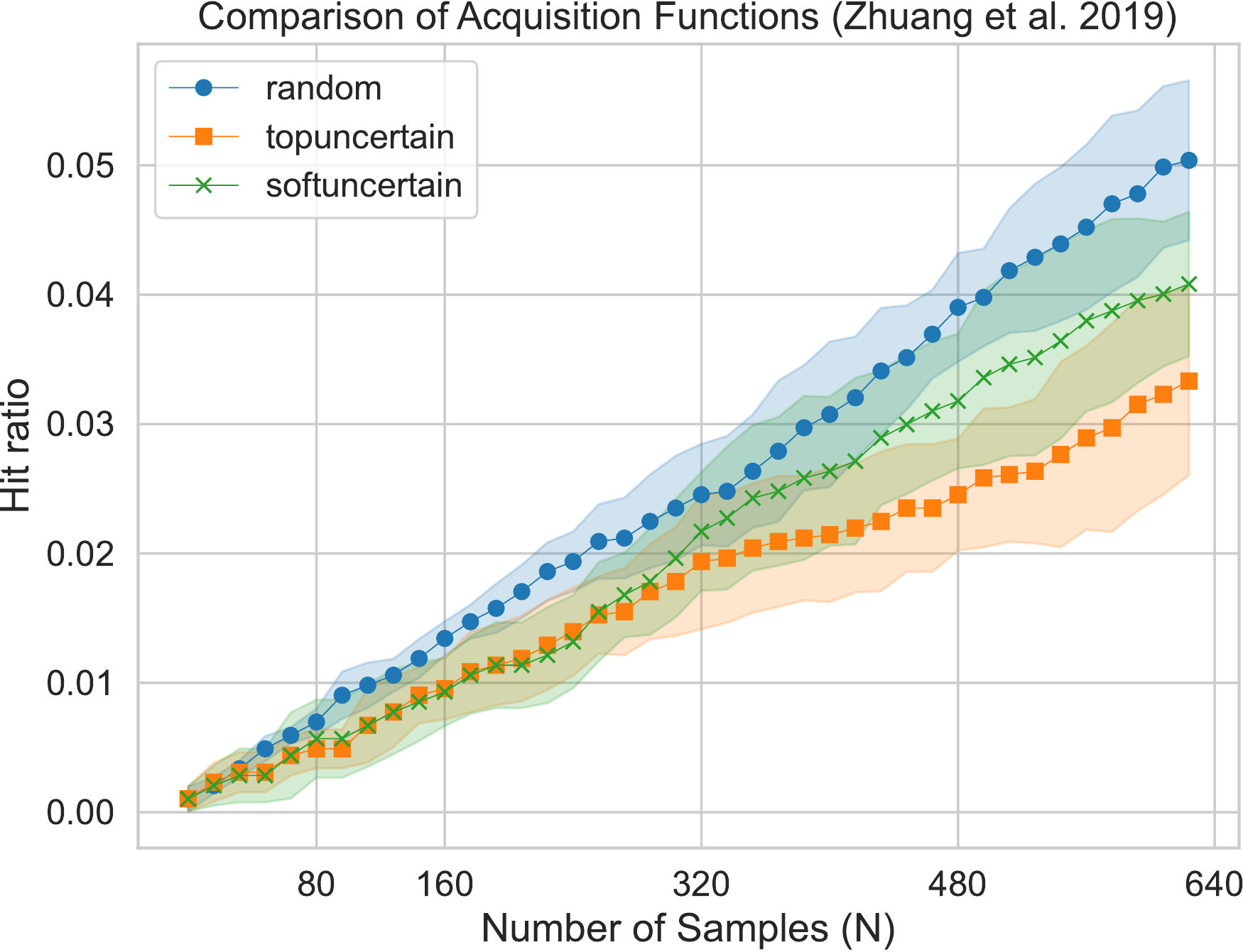}};
                        \end{tikzpicture}
                    }
                \end{subfigure}
                \&
            \\
\begin{subfigure}{0.27\columnwidth}
                    \hspace{-17mm}
                    \centering
                    \resizebox{\linewidth}{!}{
                        \begin{tikzpicture}
                            \node (img)  {\includegraphics[width=\textwidth]{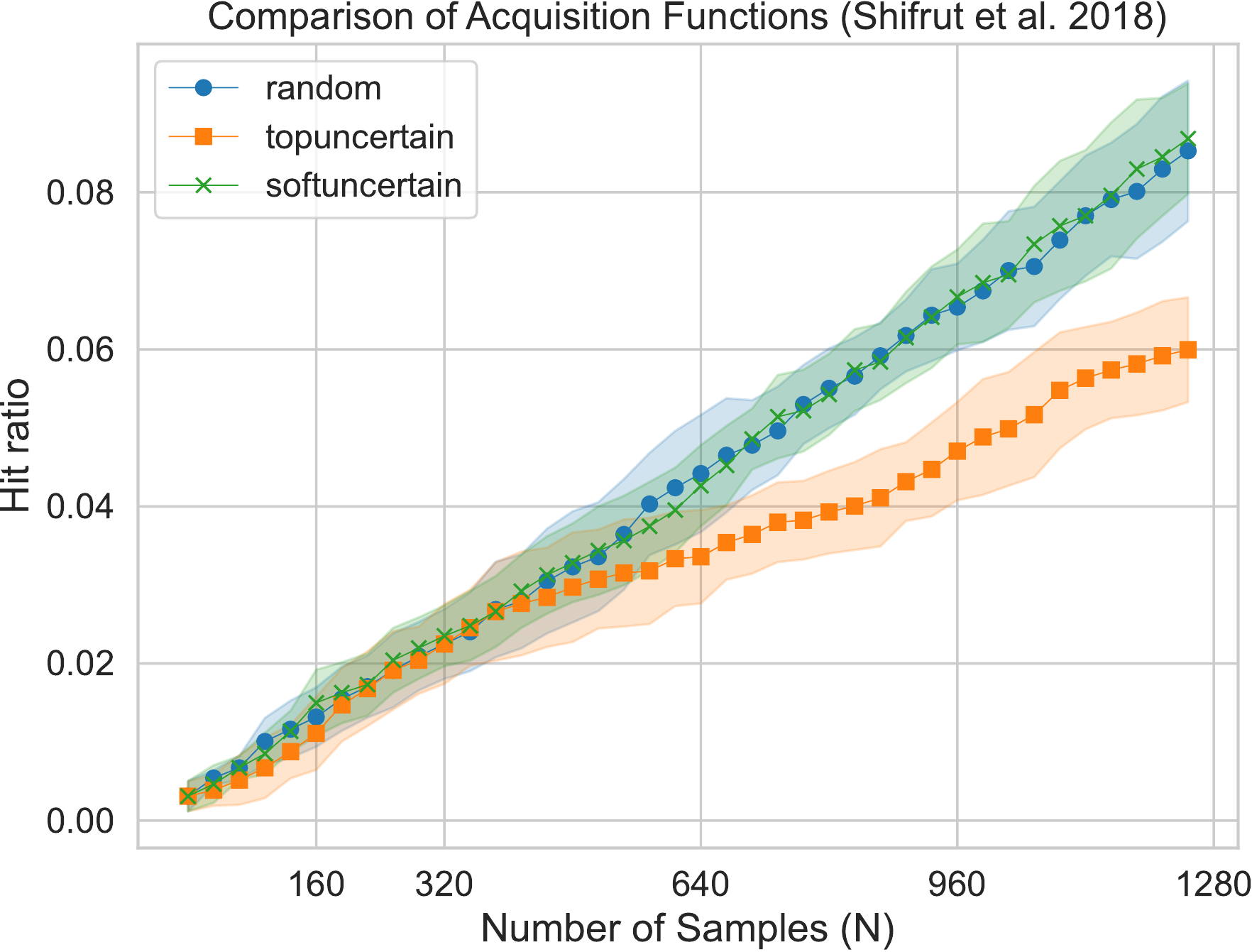}};
                        \end{tikzpicture}
                    }
                \end{subfigure}
                \&
                \begin{subfigure}{0.27\columnwidth}
                    \hspace{-23mm}
                    \centering
                    \resizebox{\linewidth}{!}{
                        \begin{tikzpicture}
                            \node (img)  {\includegraphics[width=\textwidth]{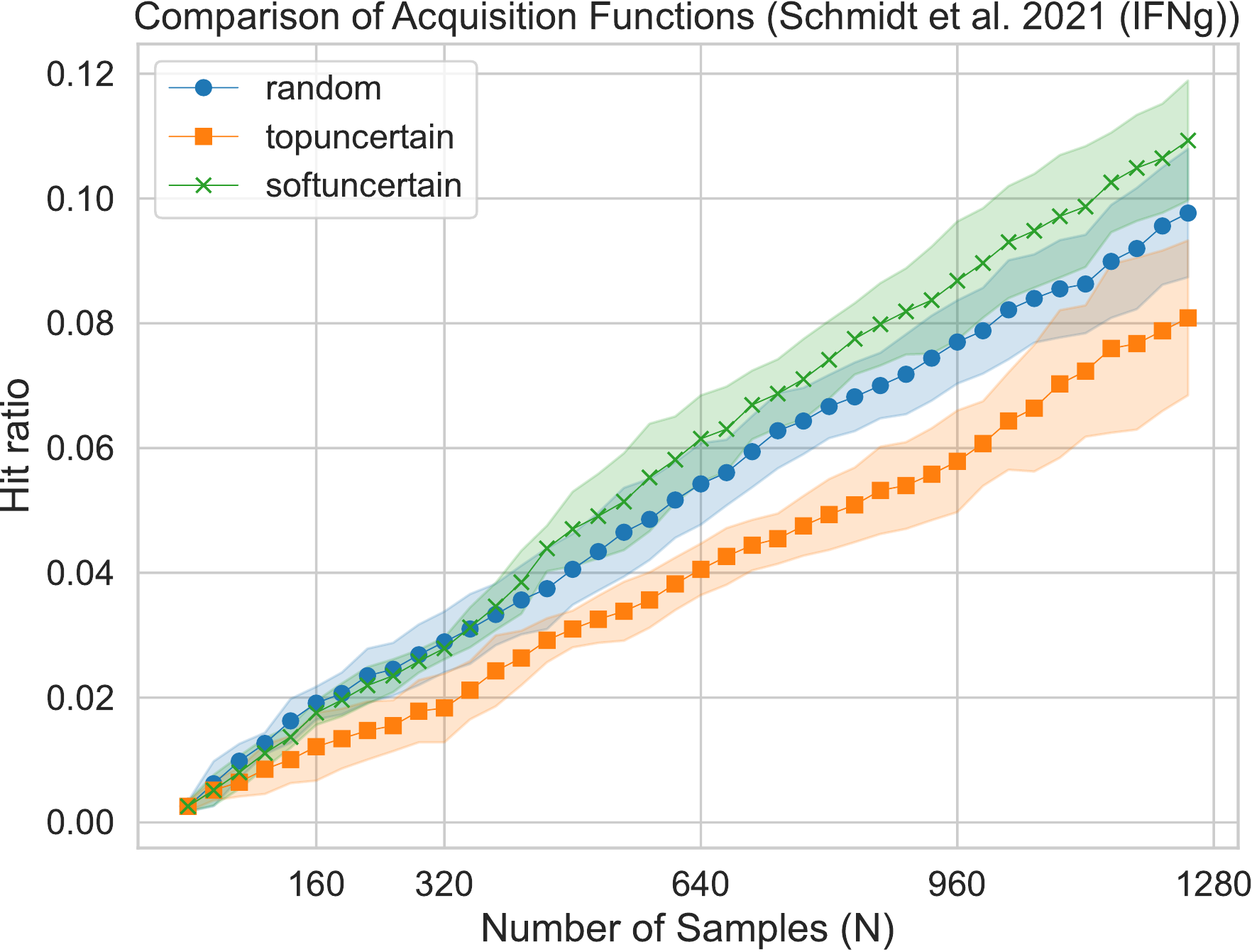}};
                        \end{tikzpicture}
                    }
                \end{subfigure}
                \&
                \begin{subfigure}{0.27\columnwidth}
                    \hspace{-28mm}
                    \centering
                    \resizebox{\linewidth}{!}{
                        \begin{tikzpicture}
                            \node (img)  {\includegraphics[width=\textwidth]{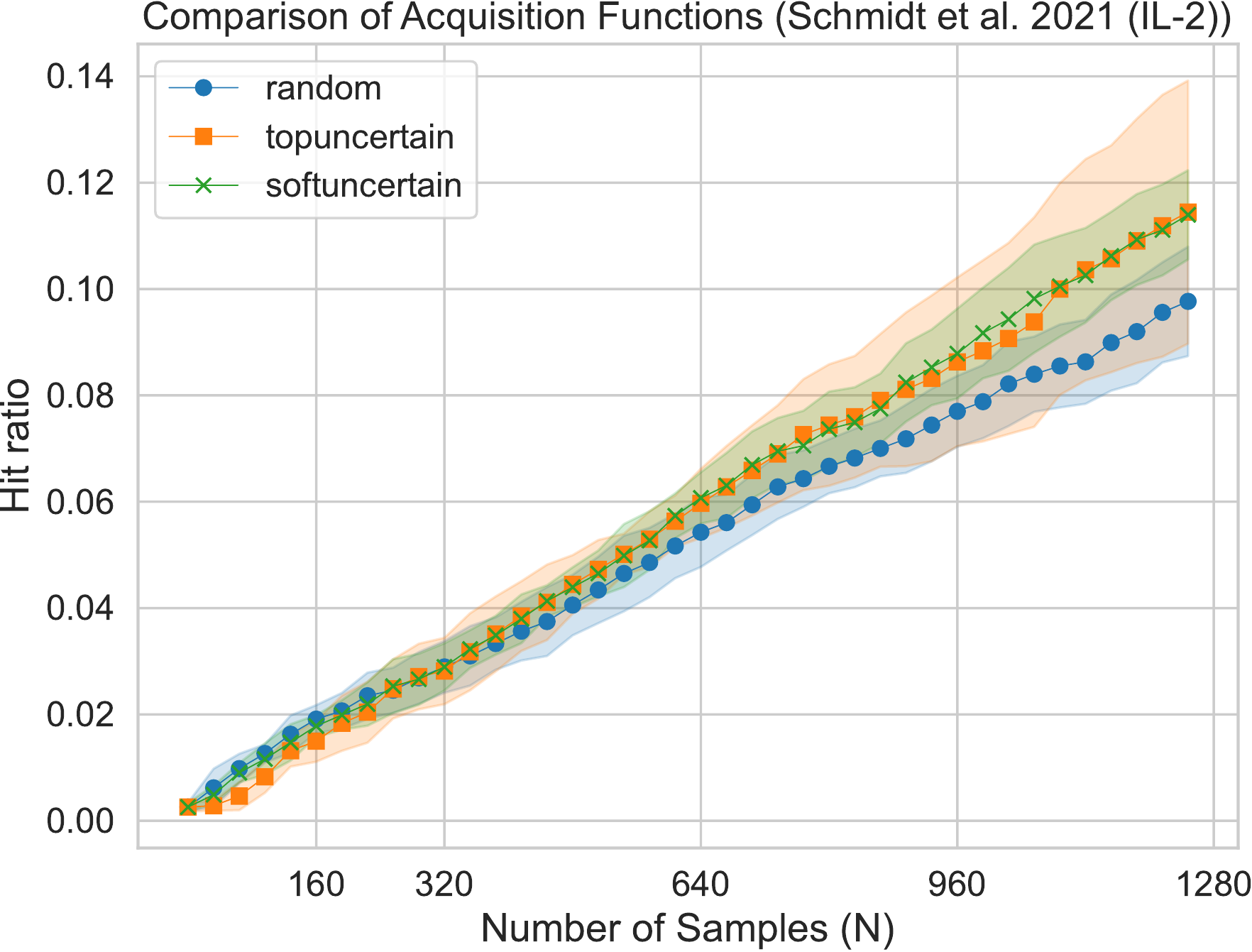}};
                        \end{tikzpicture}
                    }
                \end{subfigure}
                \&
                \begin{subfigure}{0.28\columnwidth}
                    \hspace{-32mm}
                    \centering
                    \resizebox{\linewidth}{!}{
                        \begin{tikzpicture}
                            \node (img)  {\includegraphics[width=\textwidth]{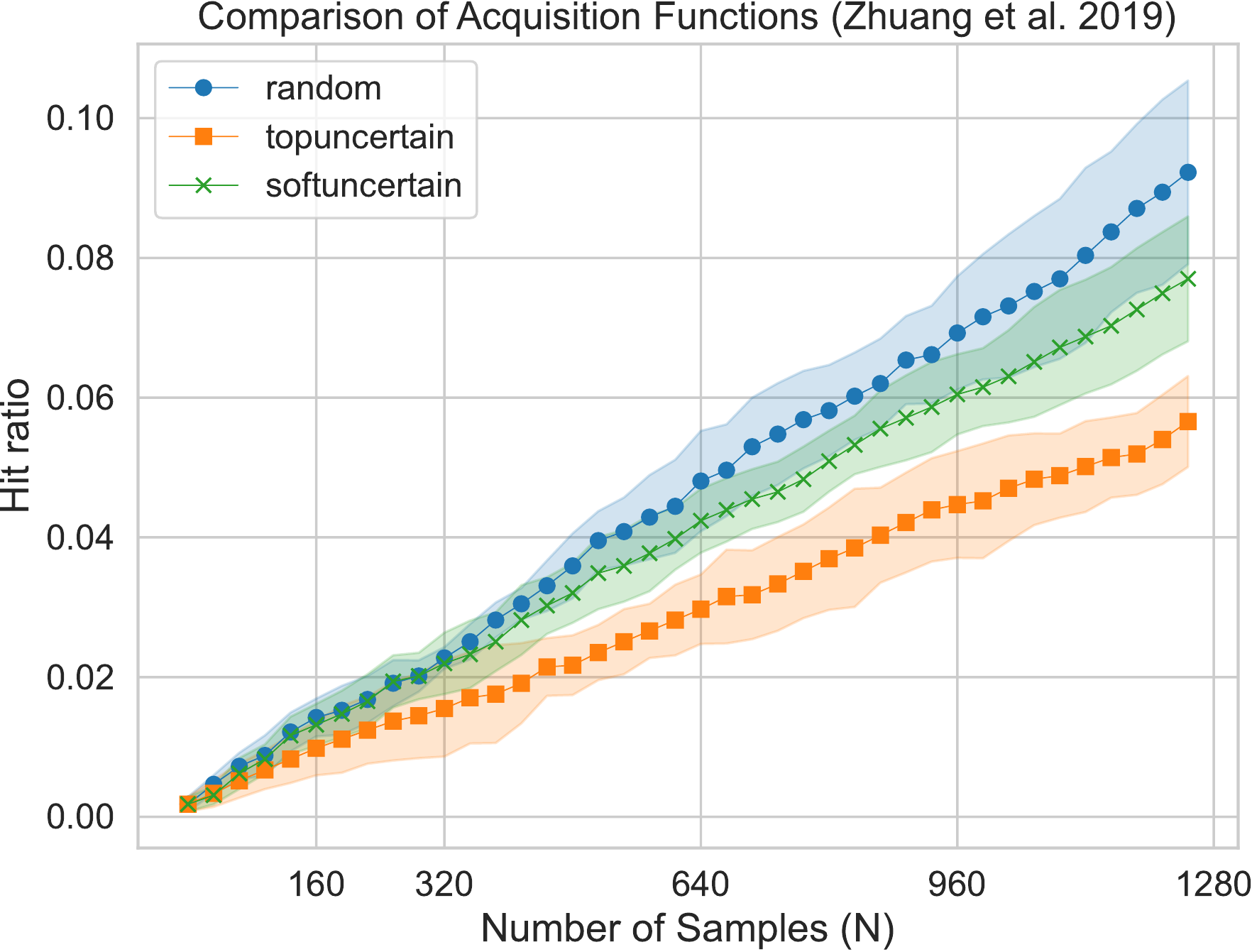}};
                        \end{tikzpicture}
                    }
                \end{subfigure}
                \&
                \\
\begin{subfigure}{0.27\columnwidth}
                    \hspace{-17mm}
                    \centering
                    \resizebox{\linewidth}{!}{
                        \begin{tikzpicture}
                            \node (img)  {\includegraphics[width=\textwidth]{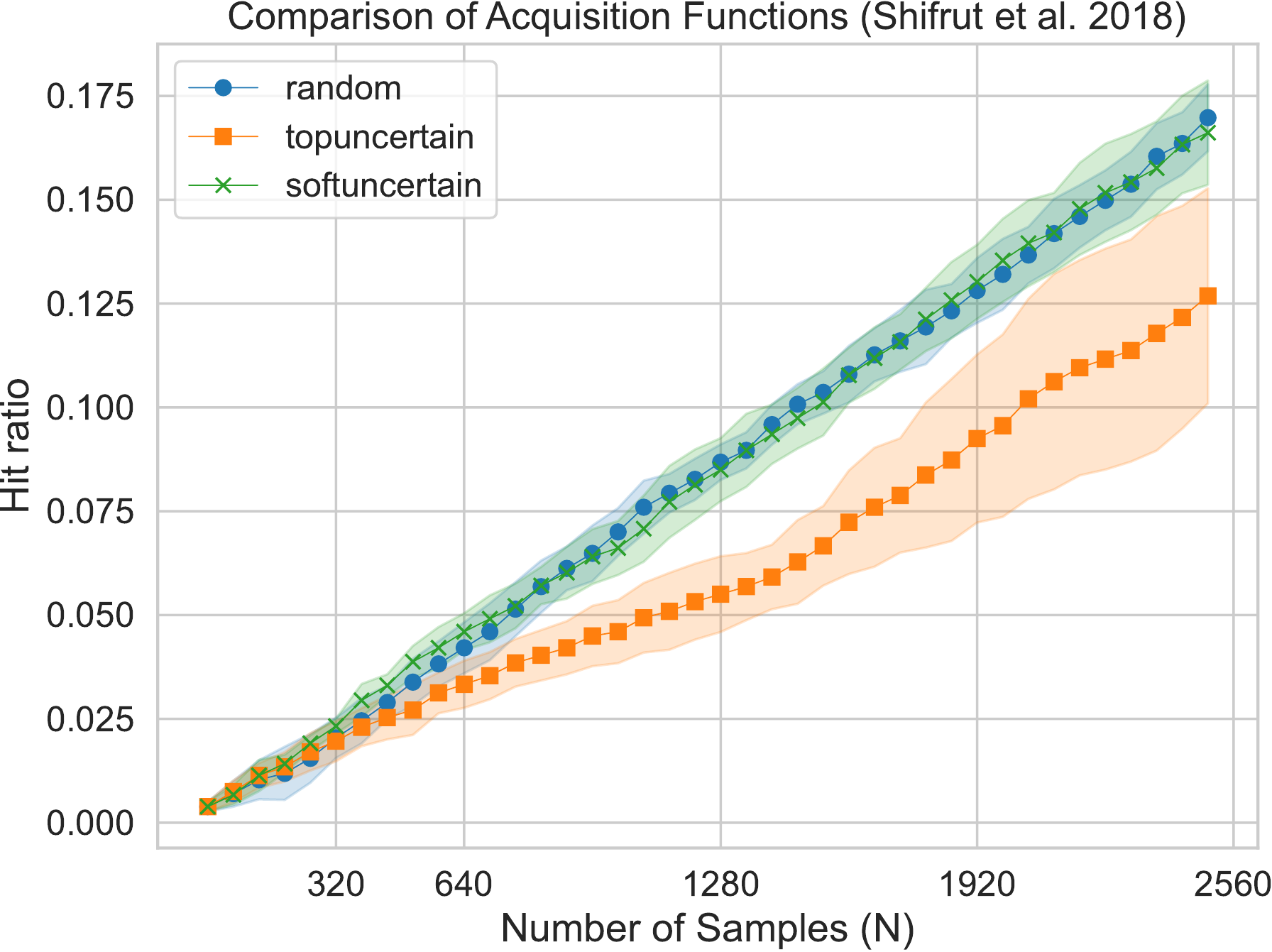}};
                        \end{tikzpicture}
                    }
                \end{subfigure}
                \&
                \begin{subfigure}{0.27\columnwidth}
                    \hspace{-23mm}
                    \centering
                    \resizebox{\linewidth}{!}{
                        \begin{tikzpicture}
                            \node (img)  {\includegraphics[width=\textwidth]{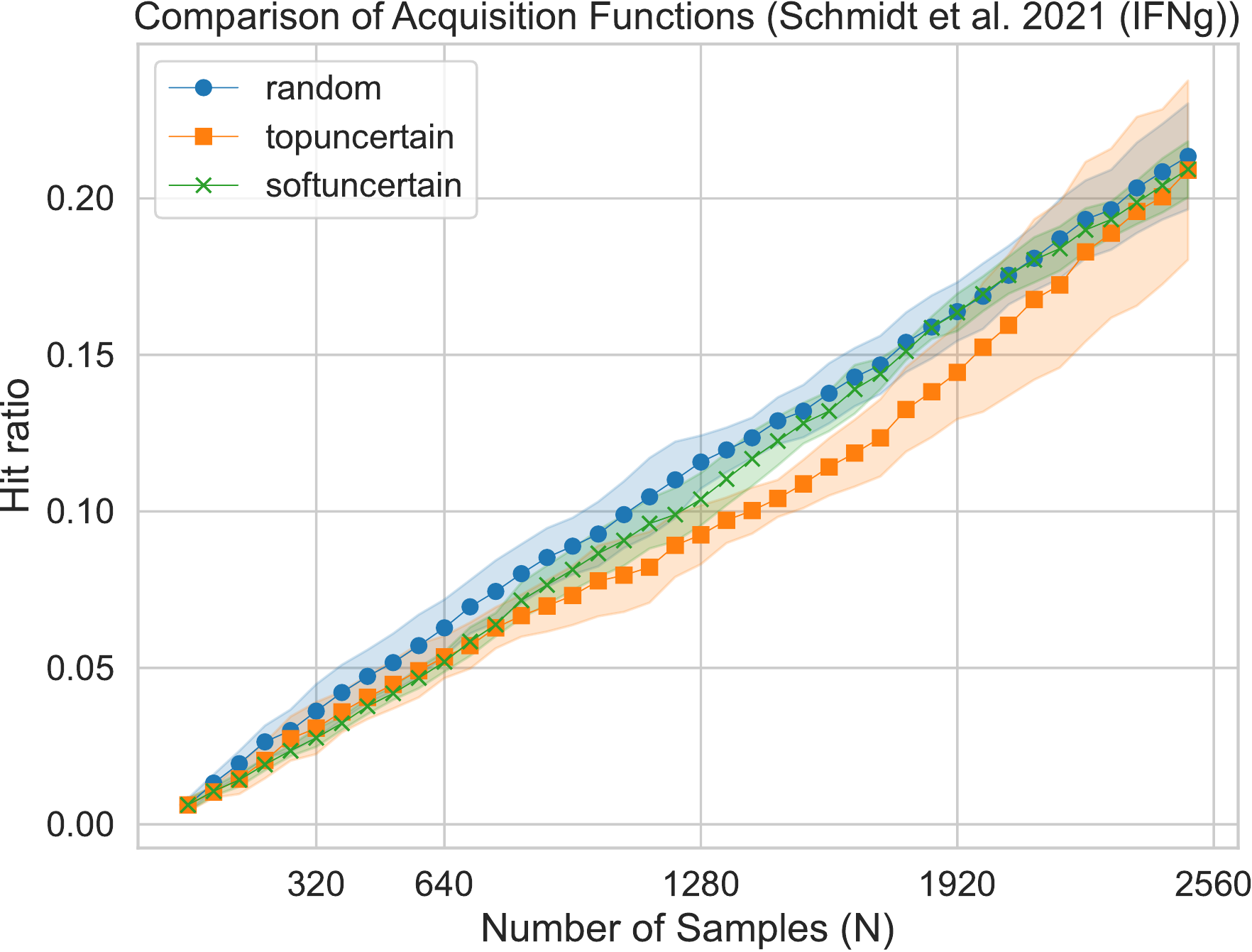}};
                        \end{tikzpicture}
                    }
                \end{subfigure}
                \&
                \begin{subfigure}{0.27\columnwidth}
                    \hspace{-28mm}
                    \centering
                    \resizebox{\linewidth}{!}{
                        \begin{tikzpicture}
                            \node (img)  {\includegraphics[width=\textwidth]{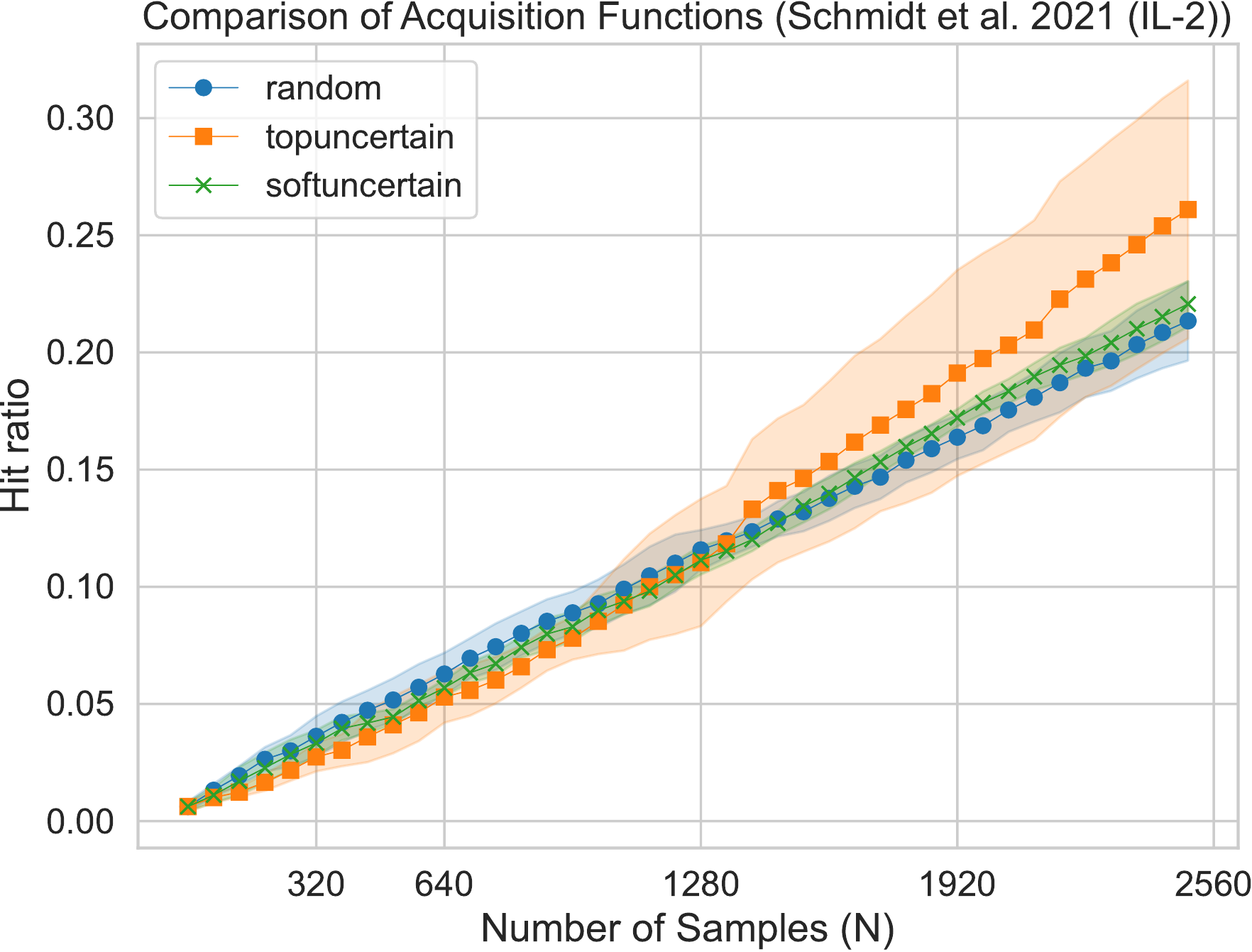}};
                        \end{tikzpicture}
                    }
                \end{subfigure}
                \&
                \begin{subfigure}{0.28\columnwidth}
                    \hspace{-32mm}
                    \centering
                    \resizebox{\linewidth}{!}{
                        \begin{tikzpicture}
                            \node (img)  {\includegraphics[width=\textwidth]{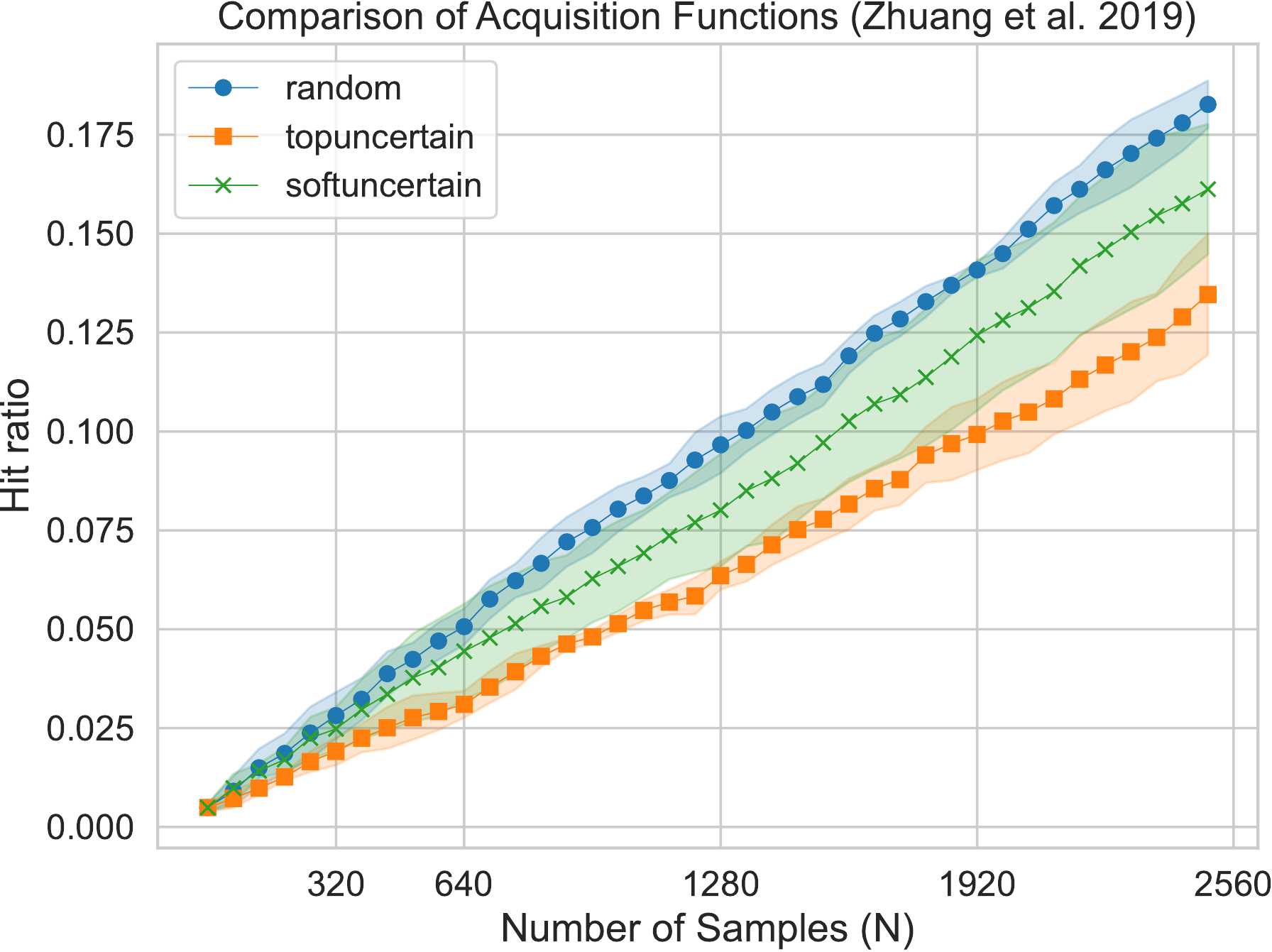}};
                        \end{tikzpicture}
                    }
                \end{subfigure}
                \&
                \\
\begin{subfigure}{0.27\columnwidth}
                    \hspace{-17mm}
                    \centering
                    \resizebox{\linewidth}{!}{
                        \begin{tikzpicture}
                            \node (img)  {\includegraphics[width=\textwidth]{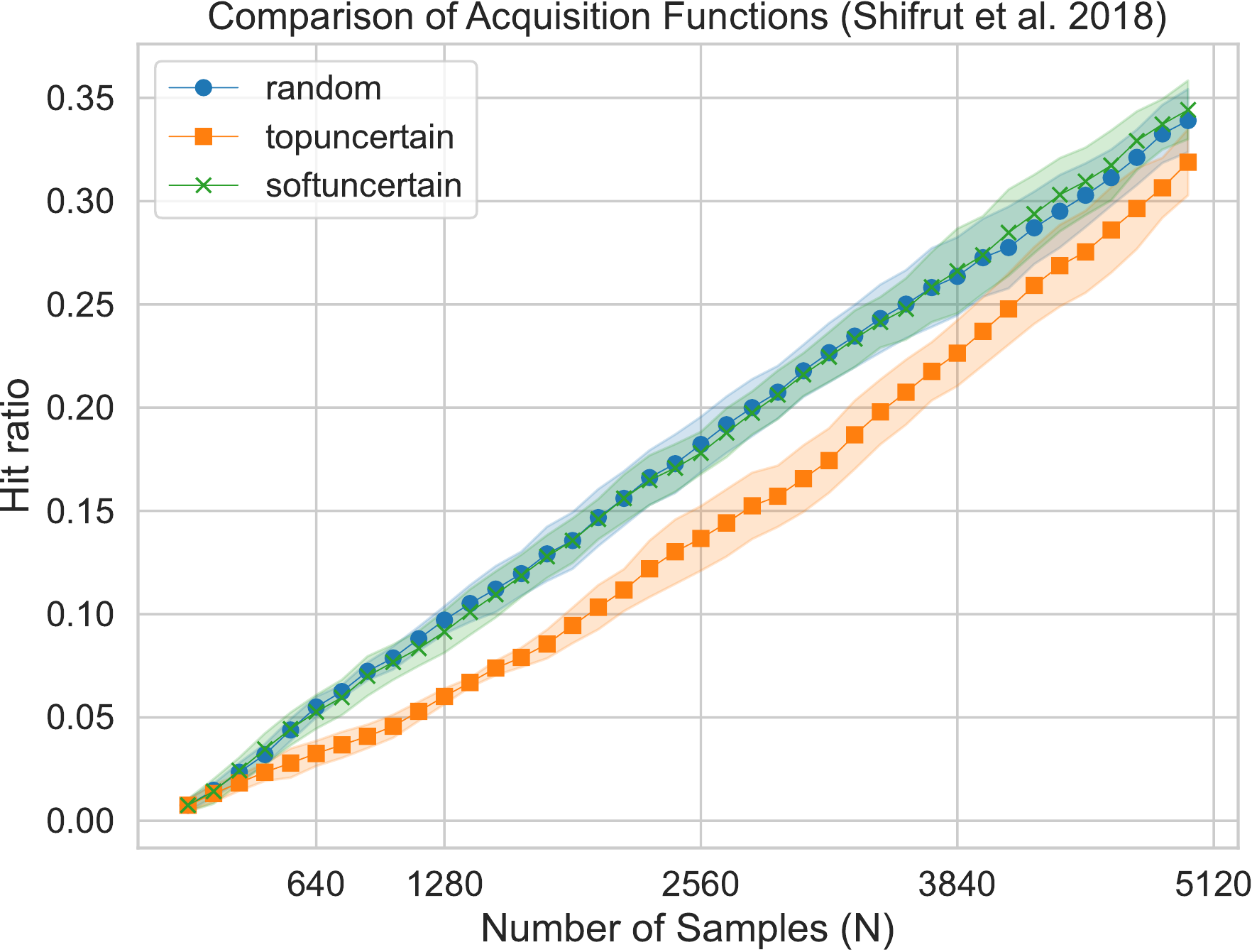}};
                        \end{tikzpicture}
                    }
                \end{subfigure}
                \&
                \begin{subfigure}{0.27\columnwidth}
                    \hspace{-23mm}
                    \centering
                    \resizebox{\linewidth}{!}{
                        \begin{tikzpicture}
                            \node (img)  {\includegraphics[width=\textwidth]{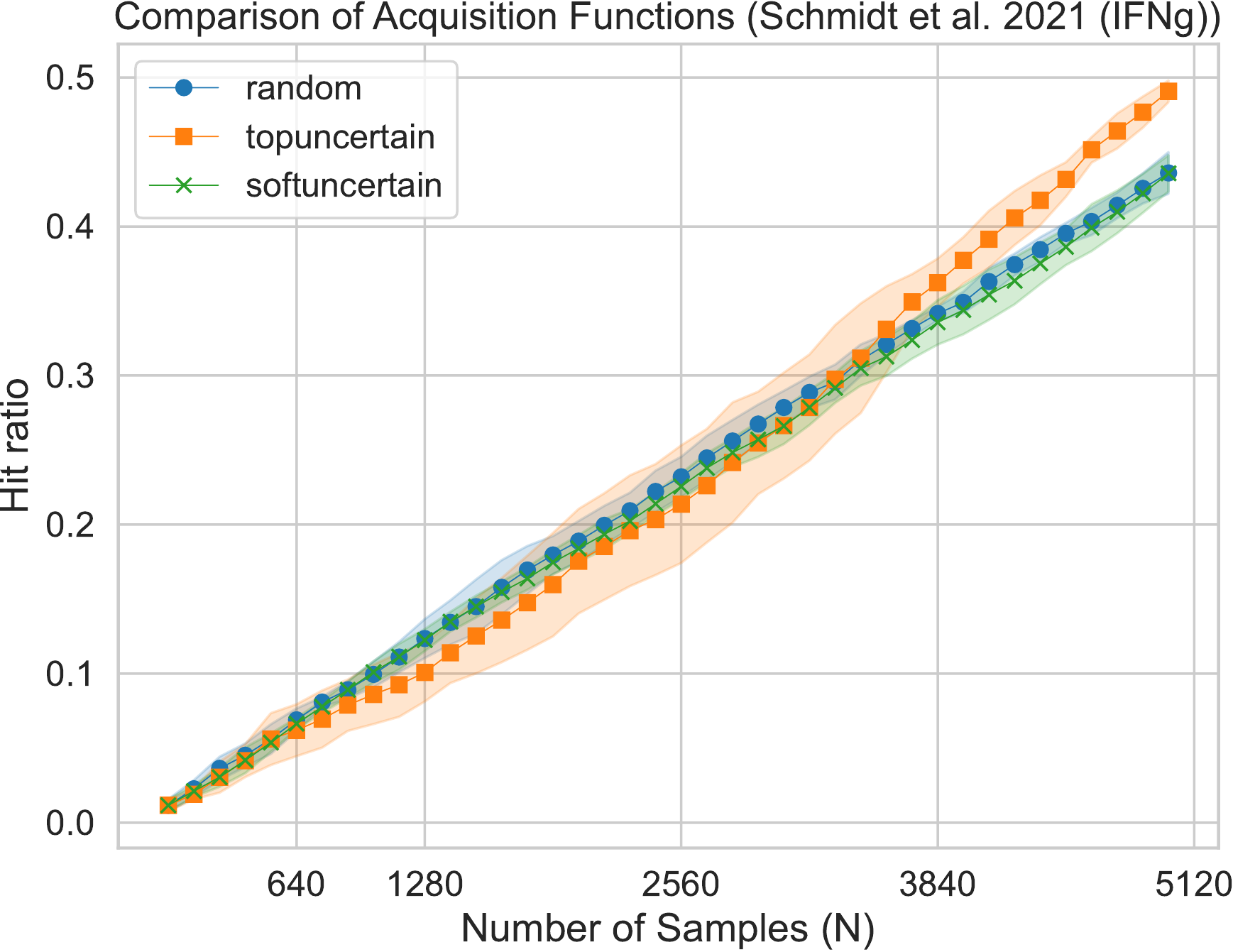}};
                        \end{tikzpicture}
                    }
                \end{subfigure}
                \&
                \begin{subfigure}{0.28\columnwidth}
                    \hspace{-28mm}
                    \centering
                    \resizebox{\linewidth}{!}{
                        \begin{tikzpicture}
                            \node (img)  {\includegraphics[width=\textwidth]{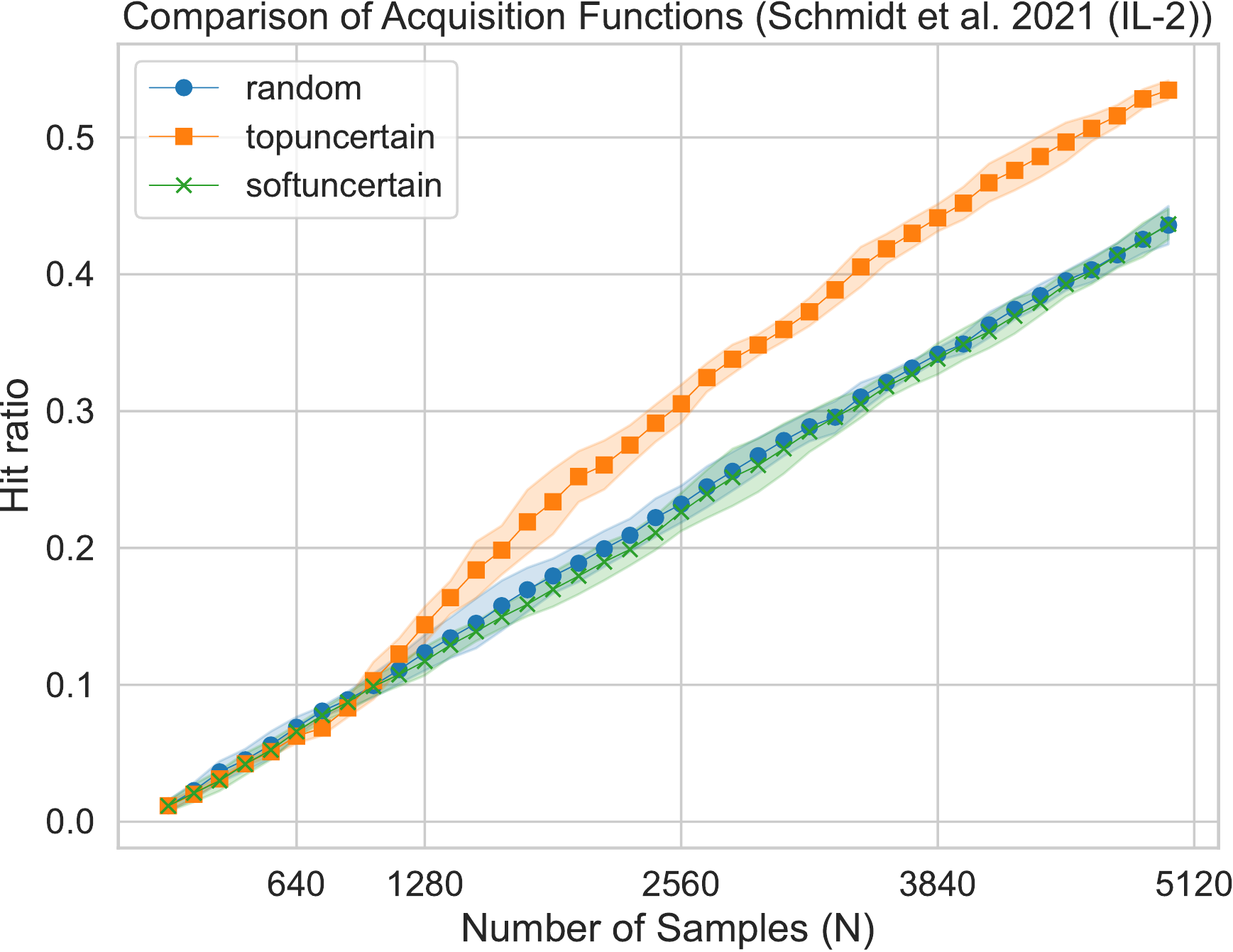}};
                        \end{tikzpicture}
                    }
                \end{subfigure}
                \&
                \begin{subfigure}{0.29\columnwidth}
                    \hspace{-32mm}
                    \centering
                    \resizebox{\linewidth}{!}{
                        \begin{tikzpicture}
                            \node (img)  {\includegraphics[width=\textwidth]{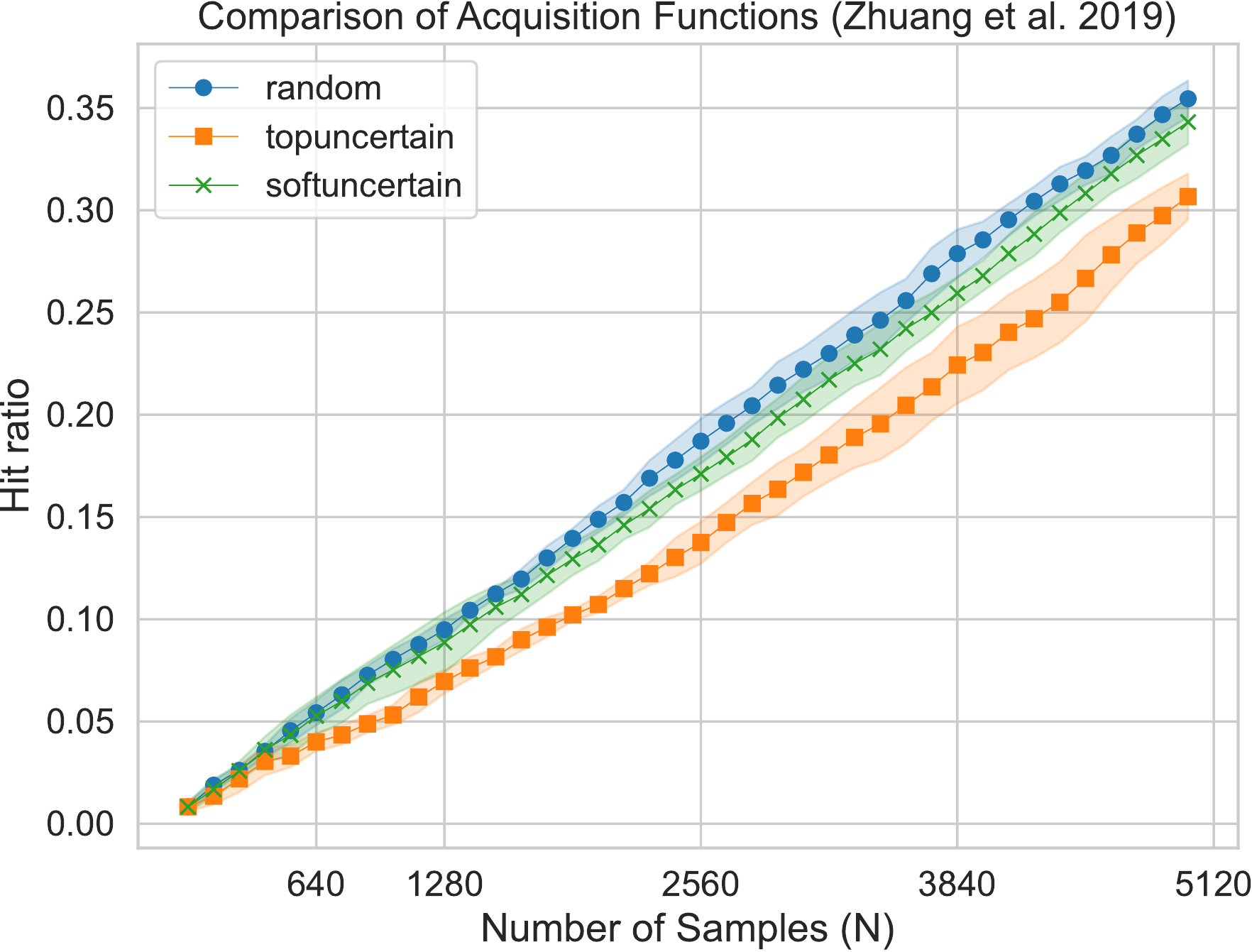}};
                        \end{tikzpicture}
                    }
                \end{subfigure}
                \&
                \\
\begin{subfigure}{0.275\columnwidth}
                    \hspace{-17mm}
                    \centering
                    \resizebox{\linewidth}{!}{
                        \begin{tikzpicture}
                            \node (img)  {\includegraphics[width=\textwidth]{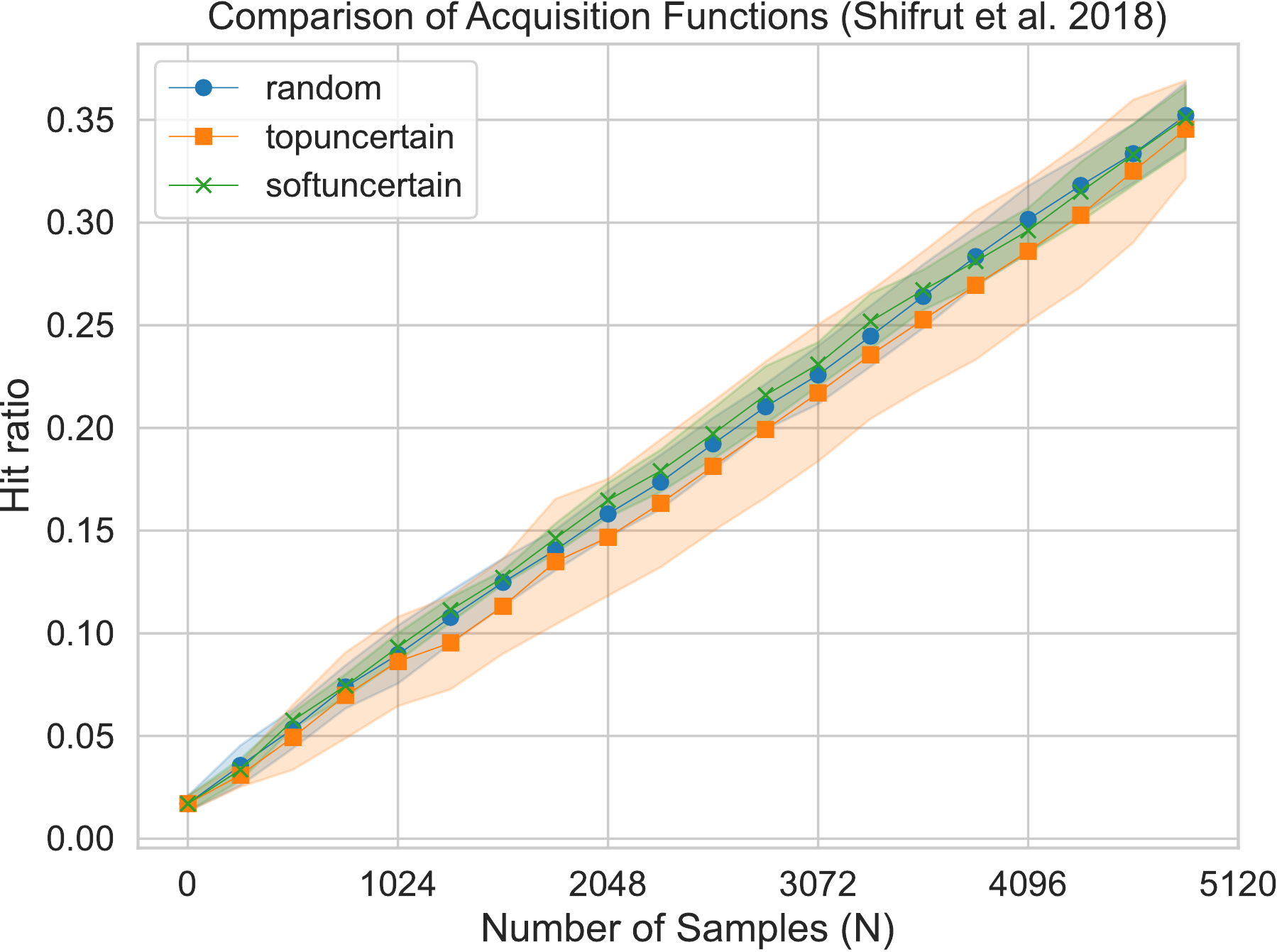}};
                        \end{tikzpicture}
                    }
                \end{subfigure}
                \&
                \begin{subfigure}{0.27\columnwidth}
                    \hspace{-23mm}
                    \centering
                    \resizebox{\linewidth}{!}{
                        \begin{tikzpicture}
                            \node (img)  {\includegraphics[width=\textwidth]{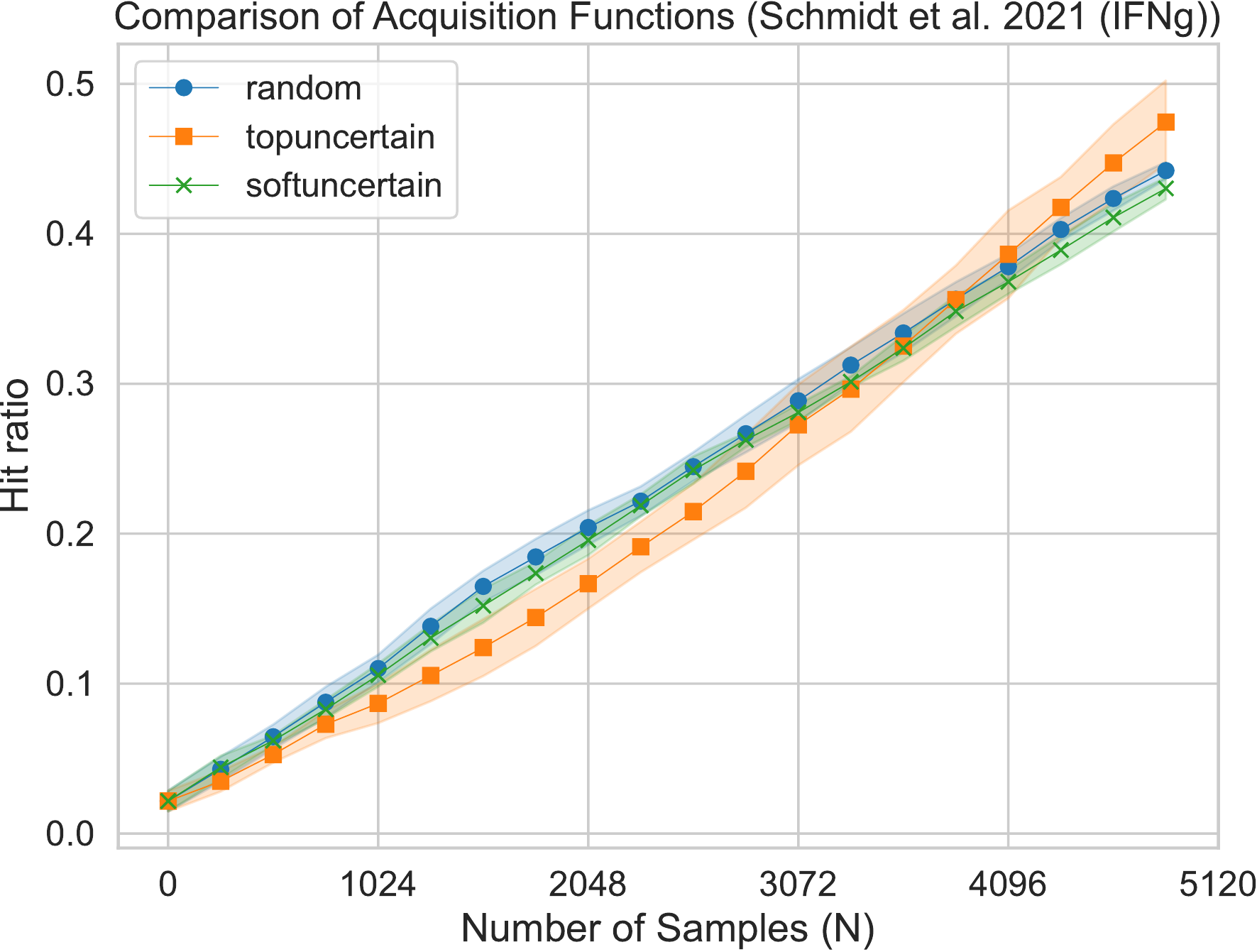}};
                        \end{tikzpicture}
                    }
                \end{subfigure}
                \&
                \begin{subfigure}{0.27\columnwidth}
                    \hspace{-28mm}
                    \centering
                    \resizebox{\linewidth}{!}{
                        \begin{tikzpicture}
                            \node (img)  {\includegraphics[width=\textwidth]{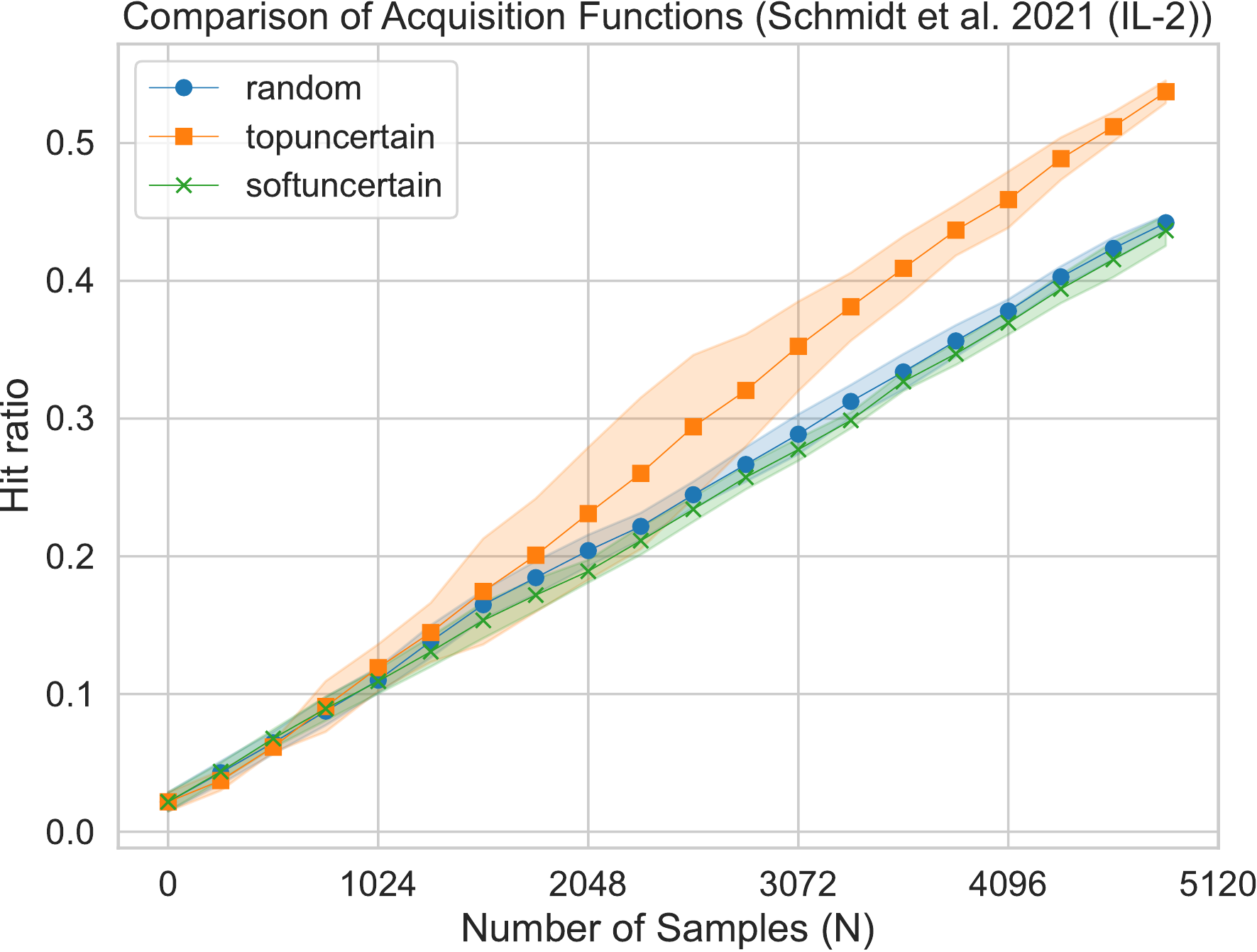}};
                        \end{tikzpicture}
                    }
                \end{subfigure}
                \&
                \begin{subfigure}{0.29\columnwidth}
                    \hspace{-32mm}
                    \centering
                    \resizebox{\linewidth}{!}{
                        \begin{tikzpicture}
                            \node (img)  {\includegraphics[width=\textwidth]{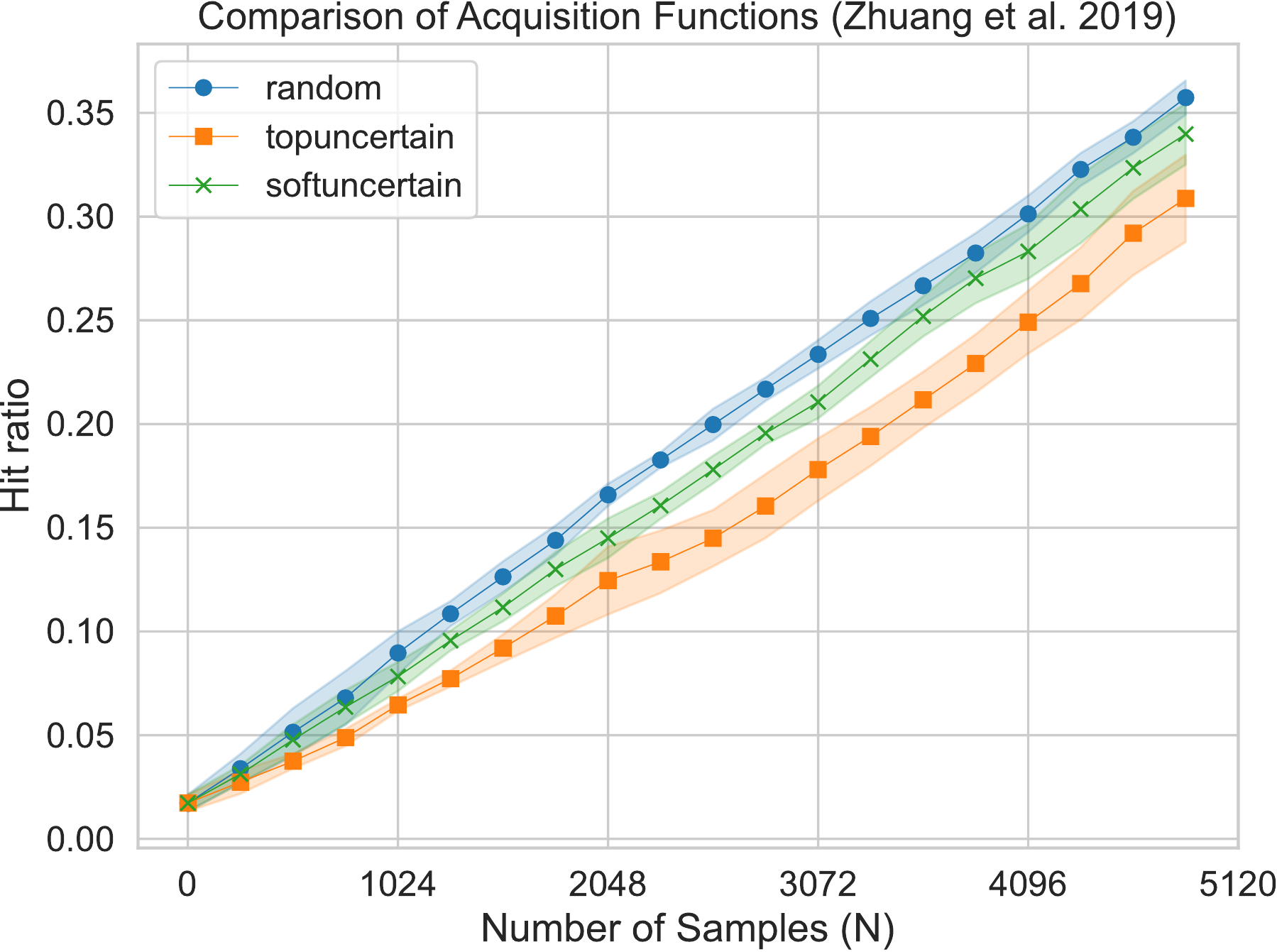}};
                        \end{tikzpicture}
                    }
                \end{subfigure}
                \&
                \\
\begin{subfigure}{0.28\columnwidth}
                    \hspace{-17mm}
                    \centering
                    \resizebox{\linewidth}{!}{
                        \begin{tikzpicture}
                            \node (img)  {\includegraphics[width=\textwidth]{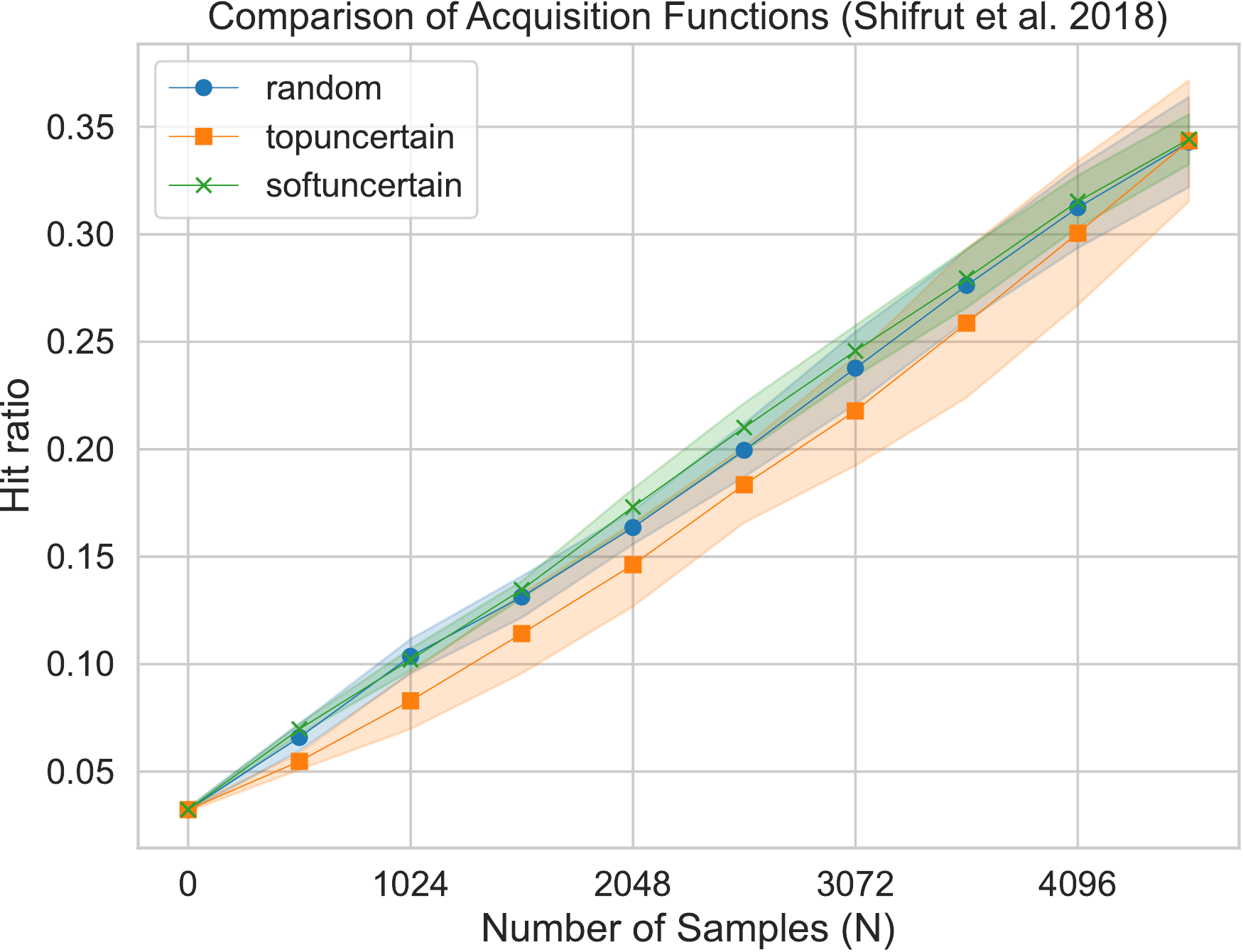}};
                        \end{tikzpicture}
                    }
                \end{subfigure}
                \&
                \begin{subfigure}{0.27\columnwidth}
                    \hspace{-23mm}
                    \centering
                    \resizebox{\linewidth}{!}{
                        \begin{tikzpicture}
                            \node (img)  {\includegraphics[width=\textwidth]{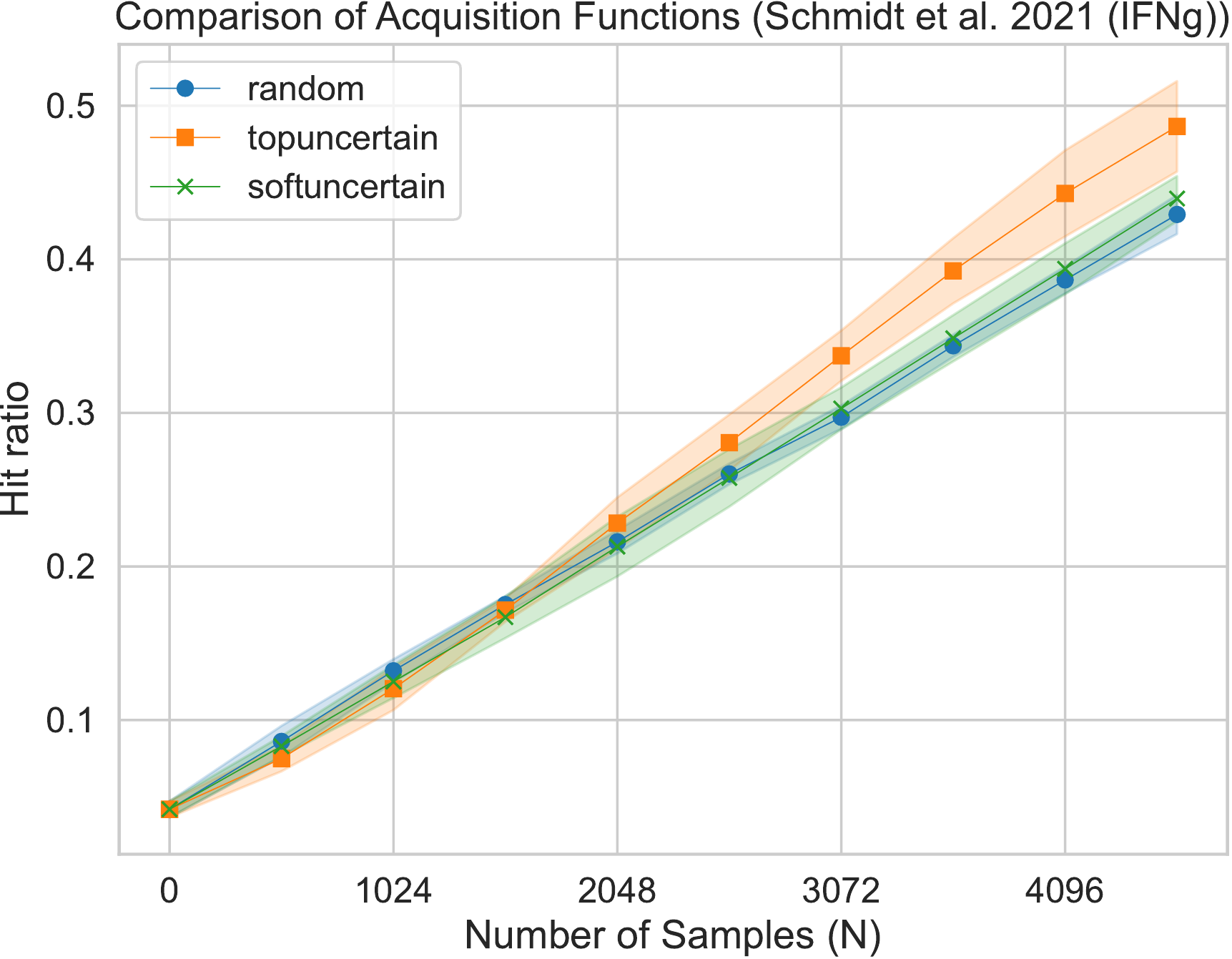}};
                        \end{tikzpicture}
                    }
                \end{subfigure}
                \&
                \begin{subfigure}{0.27\columnwidth}
                    \hspace{-28mm}
                    \centering
                    \resizebox{\linewidth}{!}{
                        \begin{tikzpicture}
                            \node (img)  {\includegraphics[width=\textwidth]{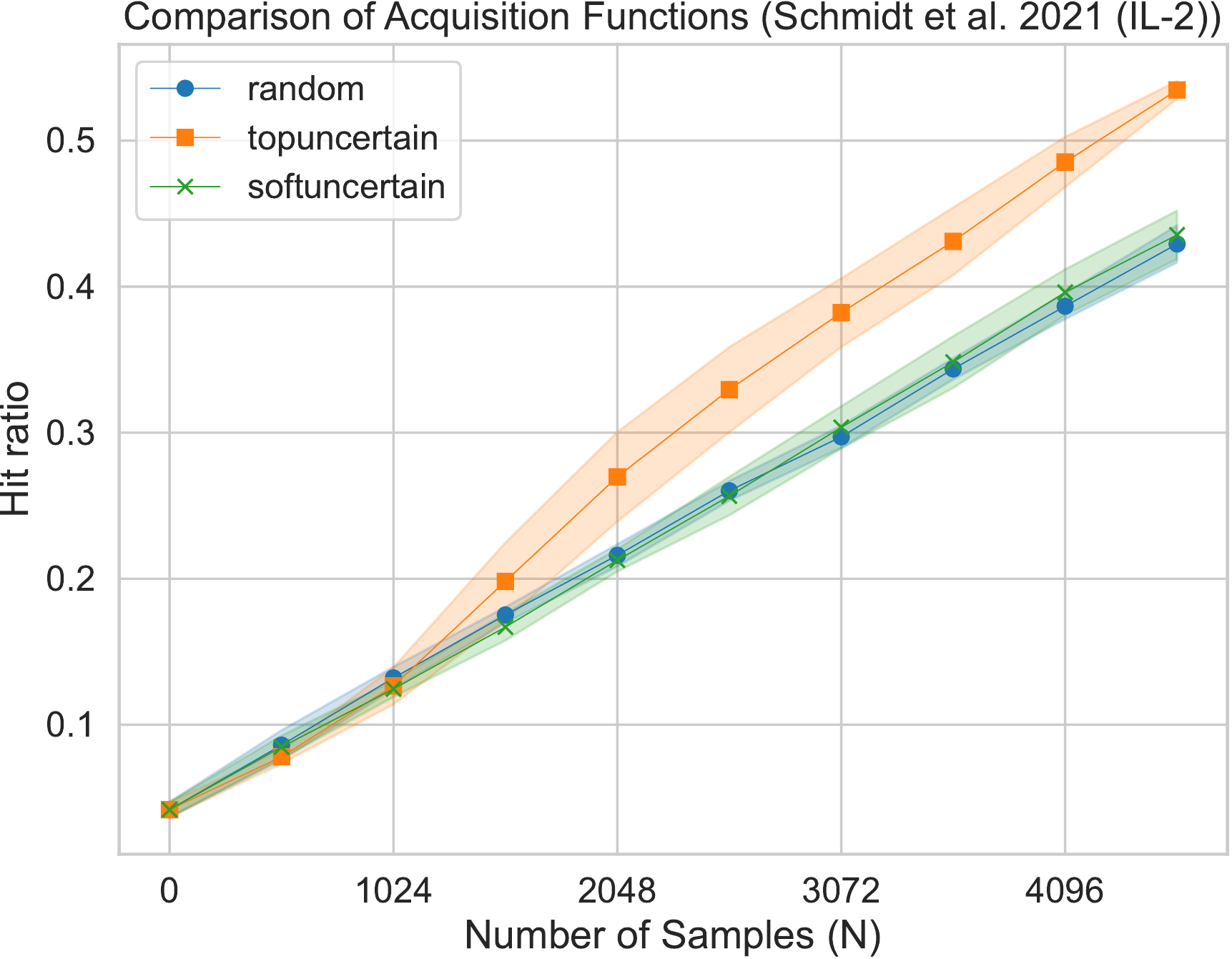}};
                        \end{tikzpicture}
                    }
                \end{subfigure}
                \&
                \begin{subfigure}{0.28\columnwidth}
                    \hspace{-32mm}
                    \centering
                    \resizebox{\linewidth}{!}{
                        \begin{tikzpicture}
                            \node (img)  {\includegraphics[width=\textwidth]{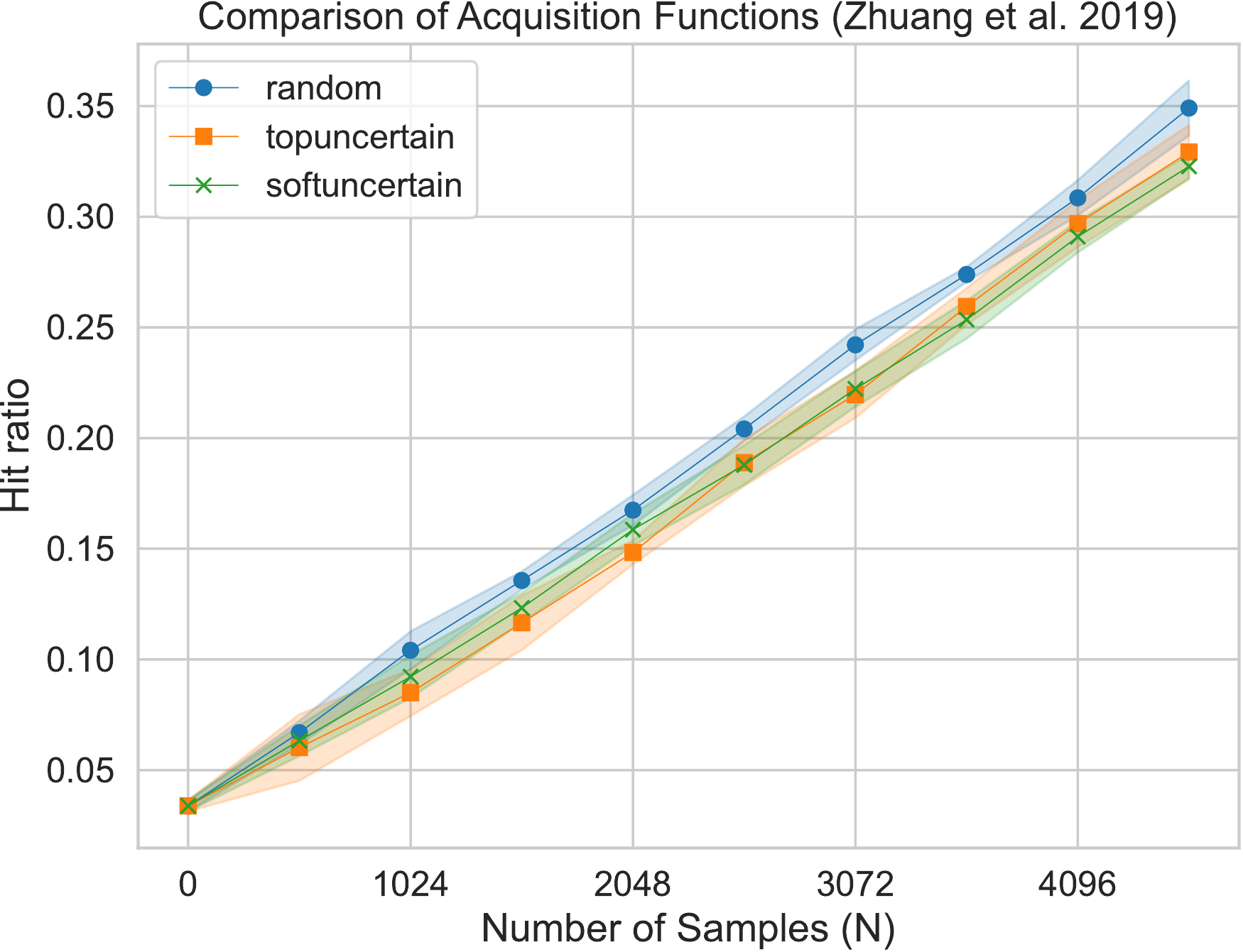}};
                        \end{tikzpicture}
                    }
                \end{subfigure}
                \&
                \\
            \\
           
            \\
            };
            \node [draw=none, rotate=90] at ([xshift=-8mm, yshift=2mm]fig-1-1.west) {\small batch size = 16};
            \node [draw=none, rotate=90] at ([xshift=-8mm, yshift=2mm]fig-2-1.west) {\small batch size = 32};
            \node [draw=none, rotate=90] at ([xshift=-8mm, yshift=2mm]fig-3-1.west) {\small batch size = 64};
            \node [draw=none, rotate=90] at ([xshift=-8mm, yshift=2mm]fig-4-1.west) {\small batch size = 128};
            \node [draw=none, rotate=90] at ([xshift=-8mm, yshift=2mm]fig-5-1.west) {\small batch size = 256};
            \node [draw=none, rotate=90] at ([xshift=-8mm, yshift=2mm]fig-6-1.west) {\small batch size = 512};
            \node [draw=none] at ([xshift=-6mm, yshift=3mm]fig-1-1.north) {\small Shifrut et al. 2018};
            \node [draw=none] at ([xshift=-9mm, yshift=3mm]fig-1-2.north) {\small Schmidt et al. 2021 (IFNg)};
            \node [draw=none] at ([xshift=-11mm, yshift=3mm]fig-1-3.north) {\small Schmidt et al. 2021 (IL-2)};
            \node [draw=none] at ([xshift=-13mm, yshift=2.5mm]fig-1-4.north) {\small Zhuang et al. 2019};
\end{tikzpicture}}
        \vspace{-7mm}
        \caption{The hit ratio of different acquisition for random forest model, different target datasets, and different acquisition batch sizes. We use {CCLE} treatment descriptors here. The x-axis shows the number of data points collected so far during the active learning cycles. The y-axis shows the ratio of the set of interesting genes that have been found by the acquisition function up until each cycle.}
        \vspace{-5mm}
        \label{fig:hitratio_rf_feat_ccle_alldatasets_allbathcsizes}
    \end{figure*}

Here we elaborate more on the purpose and the message of the hit ratio experiment whose results are reported in \cref{fig:hitratio_bnn_feat_string_alldatasets_allbathcsizes,fig:hitratio_bnn_feat_achilles_alldatasets_allbathcsizes,fig:hitratio_bnn_feat_ccle_alldatasets_allbathcsizes,fig:hitratio_rf_feat_string_alldatasets_allbathcsizes,fig:hitratio_rf_feat_ccle_alldatasets_allbathcsizes} for various settings. The purpose of these experiments is to compare the performance of different acquisition functions in different settings of batch sizes and input/output datasets to hit the gene targets that are known to be interesting by genomics experts. To choose the set of interesting genes, we sort them based on their absolute target values. Then we choose the top $5\%$ of this list that corresponds to both extremes of positive and negative values (both extremes are considered to be good targets by experts.) The experiments are repeated for 5 different random seeds to obtain the error bars.

\newpage
 \newpage
 \newpage
 \newpage
 \newpage
 \newpage
 \newpage
 \newpage
 \newpage
 \newpage

\end{document}